\documentclass[table]{ai2style/ai2}

\usepackage{microtype}
\usepackage{url}
\usepackage{booktabs, tabularx} %
\usepackage{graphicx}
\usepackage{lineno}
\usepackage{enumitem}
\usepackage{listings} %

\definecolor{ darkblue}{rgb}{0, 0, 0.5}
\hypersetup{colorlinks=true, citecolor=darkblue, linkcolor=darkblue, urlcolor=darkblue}

\usepackage{float}
\usepackage{amssymb}
\usepackage{multirow}
\usepackage{bigdelim}
\usepackage{todonotes}
\usepackage{longtable}
\usepackage{tabularray}
\usepackage{wrapfig}
\usepackage[most]{tcolorbox}
\usepackage{url}
\usepackage{xspace}
\usepackage[absolute]{textpos} %

\usepackage{fdsymbol}   %

\usepackage[utf8]{inputenc} %
\usepackage[T1]{fontenc}    %
\usepackage{url}            %
\usepackage{booktabs}       %
\usepackage{amsfonts}       %
\usepackage{nicefrac}       %
\usepackage{microtype}      %
\usepackage[table,xcdraw]{xcolor}         %
\usepackage{amsmath}
\usepackage[most]{tcolorbox}
\usepackage{csquotes}

\usepackage{siunitx}
\usepackage{graphicx}
\usepackage{arydshln}
\usepackage{wrapfig}
\usepackage{enumitem}
\usepackage{soul} %

\usepackage{multirow}
\usepackage{xspace}
\usepackage{adjustbox}
\usepackage{pifont}
\usepackage{caption}
\usepackage{makecell}
\usepackage{subcaption}
\usepackage{bold-extra}
\usepackage{url}

\usepackage{pgf-pie}

\usepackage{pifont}%
\usepackage{listings}

\definecolor{prompt}{RGB}{255,221,87}   %
\definecolor{gen}{RGB}{57,122,204}      %
\definecolor{eos}{RGB}{226,64,64}       %
\definecolor{idleshade}{gray}{0.90}     %

\setlist[itemize]{leftmargin=*,itemsep=0.3em,parsep=0em,topsep=0em}
\setlist[enumerate]{label={\bf{\arabic*.}},leftmargin=*,itemsep=0.3em,parsep=0em,topsep=0em}

\DeclareUnicodeCharacter{2212}{\ensuremath{-}}

\definecolor{maroon}{HTML}{F26035}
\definecolor{yellow}{HTML}{FDBC42}
\definecolor{lavender}{HTML}{734f96}
\definecolor{darkergrey}{HTML}{444444}
\definecolor{midgrey}{HTML}{e6eded}
\definecolor{ai2pink}{HTML}{f0529c}%
\definecolor{ai2midpink}{HTML}{fad3e5}
\definecolor{ai2lightpink}{HTML}{fbecf3}
\definecolor{ai2midwhite}{HTML}{f2e5d9}
\definecolor{ai2offwhite}{HTML}{fbf4ee}
\definecolor{ai2green}{HTML}{0fcb8c}
\definecolor{ai2lightgreen}{HTML}{e7f9f3}
\definecolor{ai2darkgreen}{HTML}{105257}
\definecolor{ai2purple}{HTML}{B932EB}
\definecolor{ai2lightpurple}{HTML}{f7e8fc}
\definecolor{neutralEight}{HTML}{343434}
\definecolor{neutralFive}{HTML}{838383}
\definecolor{neutralThree}{HTML}{bebebe}
\definecolor{neutralOne}{HTML}{dedede}
\definecolor{lightgrey}{HTML}{fafcfc}
\definecolor{plum}{rgb}{0.56,0.27,0.52}
\definecolor{mulberry}{HTML}{A93C93}
\definecolor{periwinkle}{HTML}{665fd1}

\usepackage{tikz}

\usetikzlibrary{arrows.meta,positioning,calc,backgrounds,shadows.blur}
\definecolor{LearnerMain}{RGB}{30,102,200}
\definecolor{LearnerLite}{RGB}{120,175,255}
\definecolor{ActorMain}{RGB}{200,57,43}
\definecolor{ActorLite}{RGB}{244,170,160}
\definecolor{QueueMain}{RGB}{34,121,60}

\tikzset{
  every node/.style={font=\sffamily},
  panel/.style={
    rounded corners=2mm,
    minimum width=58mm, minimum height=36mm,
    inner sep=6mm, align=center, text=white,
    blur shadow={shadow blur steps=4, shadow opacity=.45}
  },
  title/.style={font=\bfseries\Large, text=white},
  flow/.style={-{Latex[length=3mm,width=2.3mm]}, line width=.9pt},
  lbl/.style={font=\normalsize, fill=white, fill opacity=.85, text opacity=1, inner sep=1pt},
  flowlbl/.style={ %
    midway, sloped,            %
    inner sep=1.5pt,
    font=\sffamily\Large,
    fill=white, fill opacity=.98, text opacity=1,
  }
}

\definecolor{maroon}{HTML}{F26035}
\definecolor{yellow}{HTML}{FDBC42}
\definecolor{darkred}{RGB}{156, 39, 33}
\definecolor{darkblue}{RGB}{31, 90, 153}
\definecolor{forestgreen}{rgb}{0.13, 0.55, 0.13}
\definecolor{brickred}{rgb}{0.8, 0.25, 0.33}
\definecolor{olmoDarkBlue}{HTML}{012e59}
\definecolor{olmoBlue}{HTML}{265ed4}
\definecolor{olmoLightBlue}{HTML}{012e59}
\definecolor{olmoTeal}{HTML}{00d5ff}
\definecolor{olmoYellow}{HTML}{ffbb00}
\definecolor{olmoOrange}{HTML}{ff9100}

\usepackage{setspace}

\usepackage{nicematrix}
\newcolumntype{L}[1]{>{\raggedright\let\newline\\\arraybackslash\hspace{0pt}}m{#1}}
\newcolumntype{C}[1]{>{\centering\let\newline\\\arraybackslash\hspace{0pt}}m{#1}}
\newcolumntype{R}[1]{>{\raggedleft\let\newline\\\arraybackslash\hspace{0pt}}m{#1}}
\newcolumntype{P}[1]{>{\centering\let\newline\\\arraybackslash\columncolor{ai2lightpink}}m{#1}}
\newcolumntype{Q}[1]{>{\centering\let\newline\\\arraybackslash}m{#1}}

\newcolumntype{H}{>{\setbox0=\hbox\bgroup}c<{\egroup}@{}} %
\newcommand{\aitoo}{\raisebox{-1.5pt}{\includegraphics[height=1.05em]{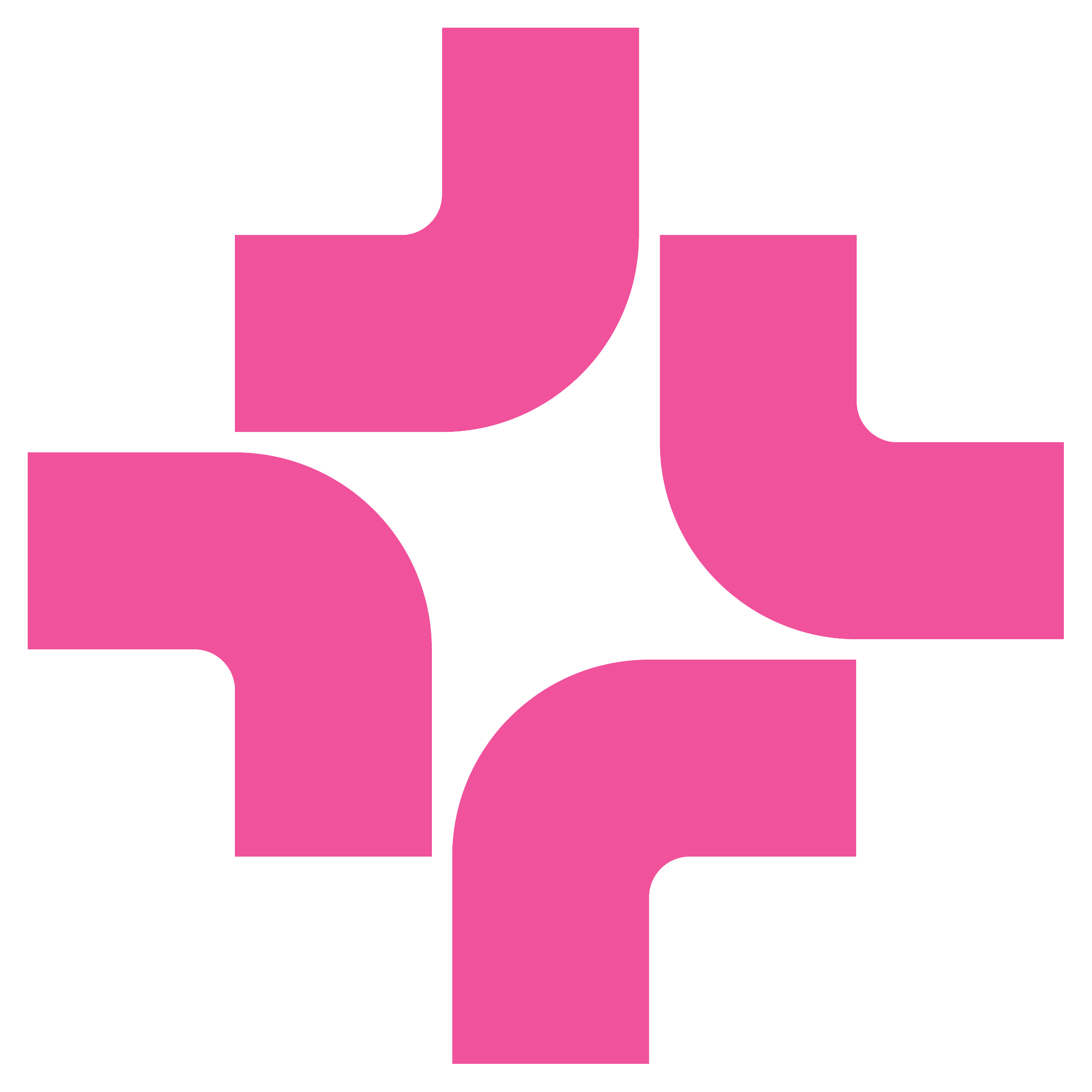}}\xspace}
\newcommand{\allenAiAff}{\raisebox{.28em}{\hspace{.02em}\scalebox{0.7}{\textbf{1}}}}

\newcommand{\uwAff}{\raisebox{.28em}{\hspace{.02em}\scalebox{0.7}{\textbf{2}}}}
\newcommand{\cmuAff}{\raisebox{.28em}{\hspace{.02em}\scalebox{0.7}{\textbf{3}}}}
\newcommand{\stanfordAff}{\raisebox{.28em}{\hspace{.02em}\scalebox{0.7}{\textbf{4}}}}
\newcommand{\milaAff}{\raisebox{.28em}{\hspace{.02em}\scalebox{0.7}{\textbf{5}}}}
\newcommand{\montrealAff}{\raisebox{.28em}{\hspace{.02em}\scalebox{0.7}{\textbf{6}}}}
\newcommand{\princetonAff}{\raisebox{.28em}{\hspace{.02em}\scalebox{0.7}{\textbf{7}}}}
\newcommand{\mitAff}{\raisebox{.28em}{\hspace{.02em}\scalebox{0.7}{\textbf{8}}}}
\newcommand{\umdAff}{\raisebox{.28em}{\hspace{.02em}\scalebox{0.7}{\textbf{9}}}}

\newcommand{\commaAff}{\raisebox{.28em}{\hspace{.02em}\scalebox{0.7}{\textbf{,}\hspace{0.1em}}}}
\newcommand{\coreContrib}{\raisebox{.28em}{\hspace{.05em}\includegraphics[height=.45em]{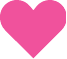}}\hspace{0.1em}}
\newcommand{\starOlmo}{\raisebox{.28em}{\hspace{.05em}\includegraphics[height=.5em]{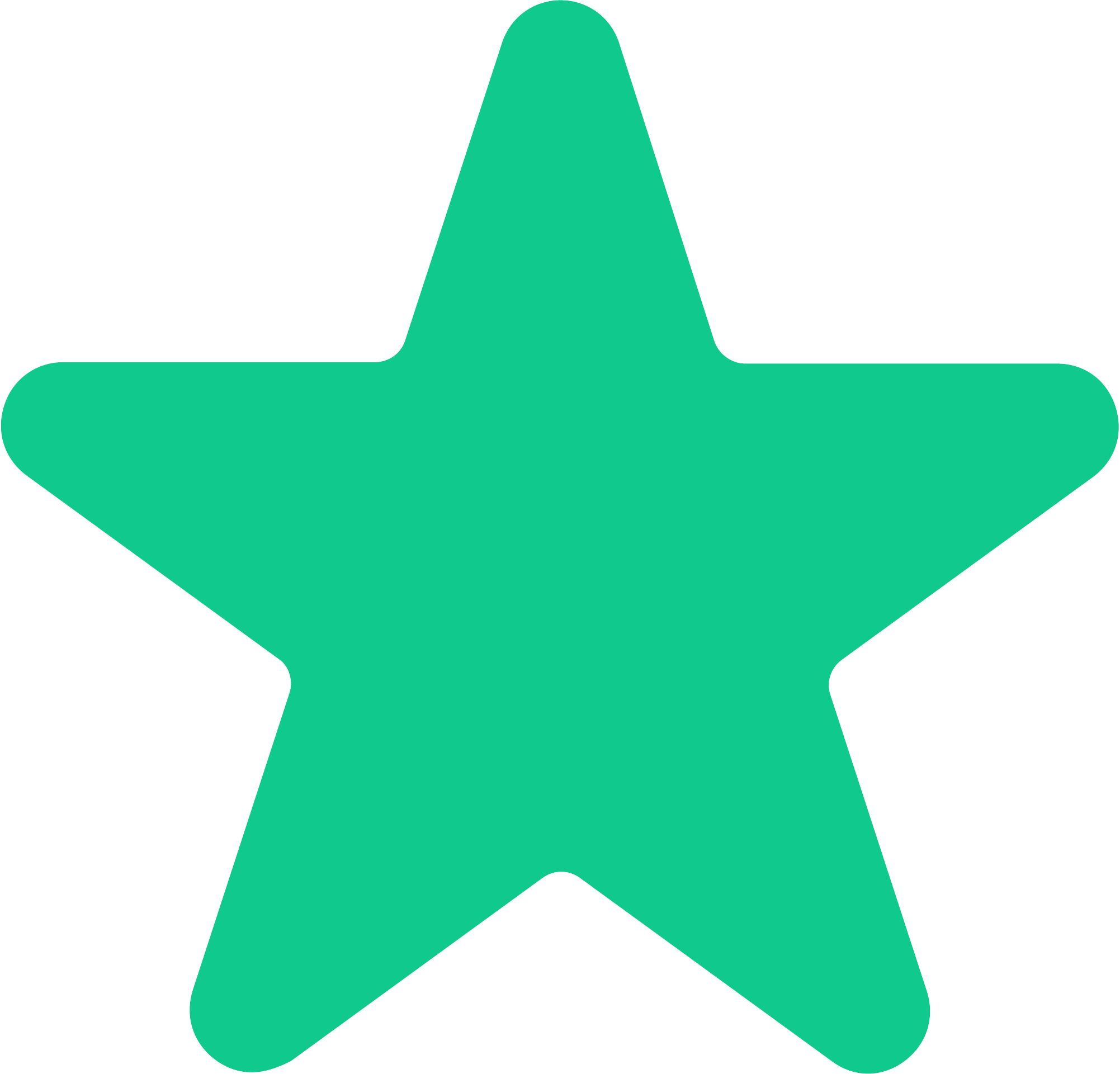}}}
\newcommand{\huggingface}{\raisebox{-1.5pt}{\includegraphics[height=1.05em]{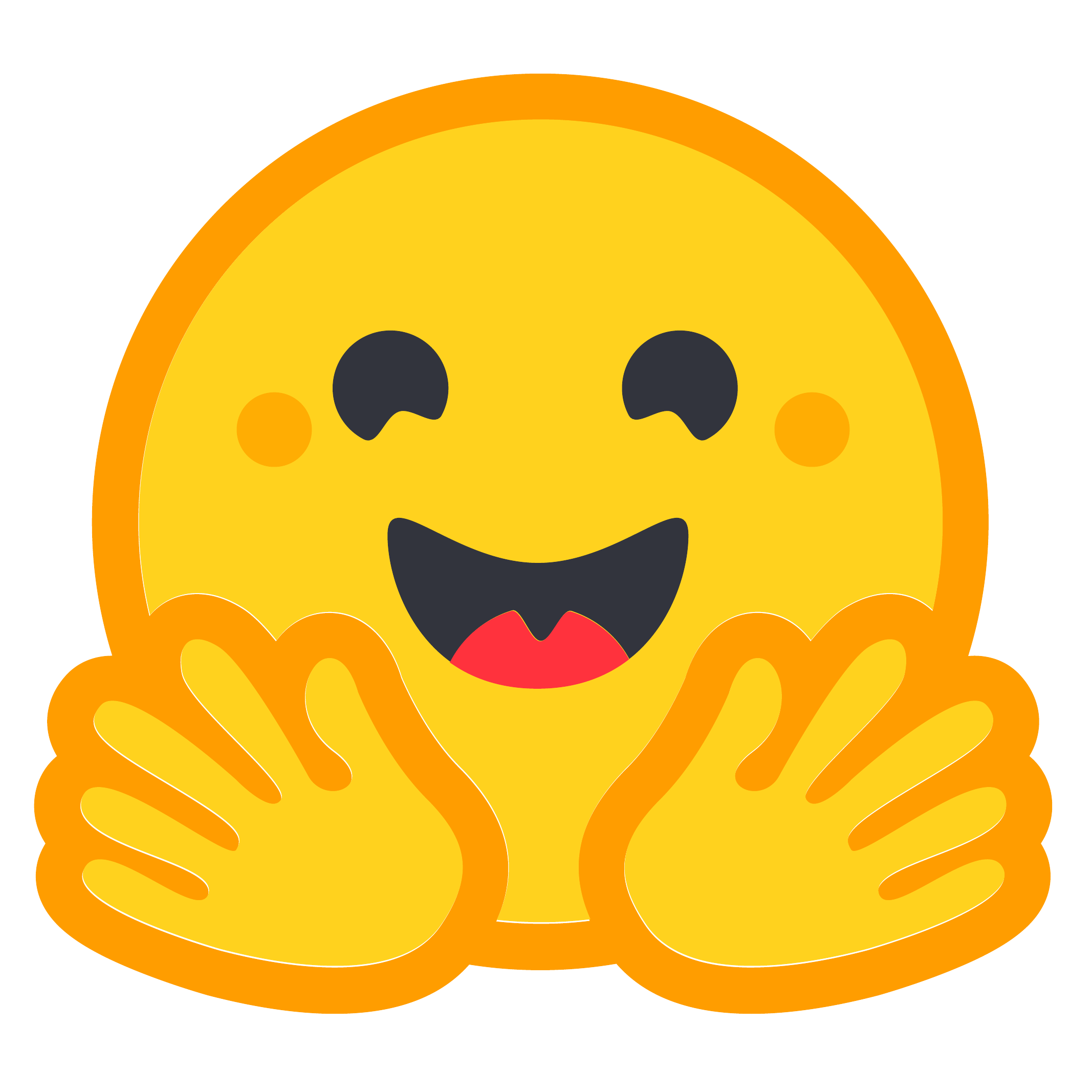}}\xspace}
\newcommand{\hfdataset}{\raisebox{-1.5pt}{\includegraphics[height=1.05em]{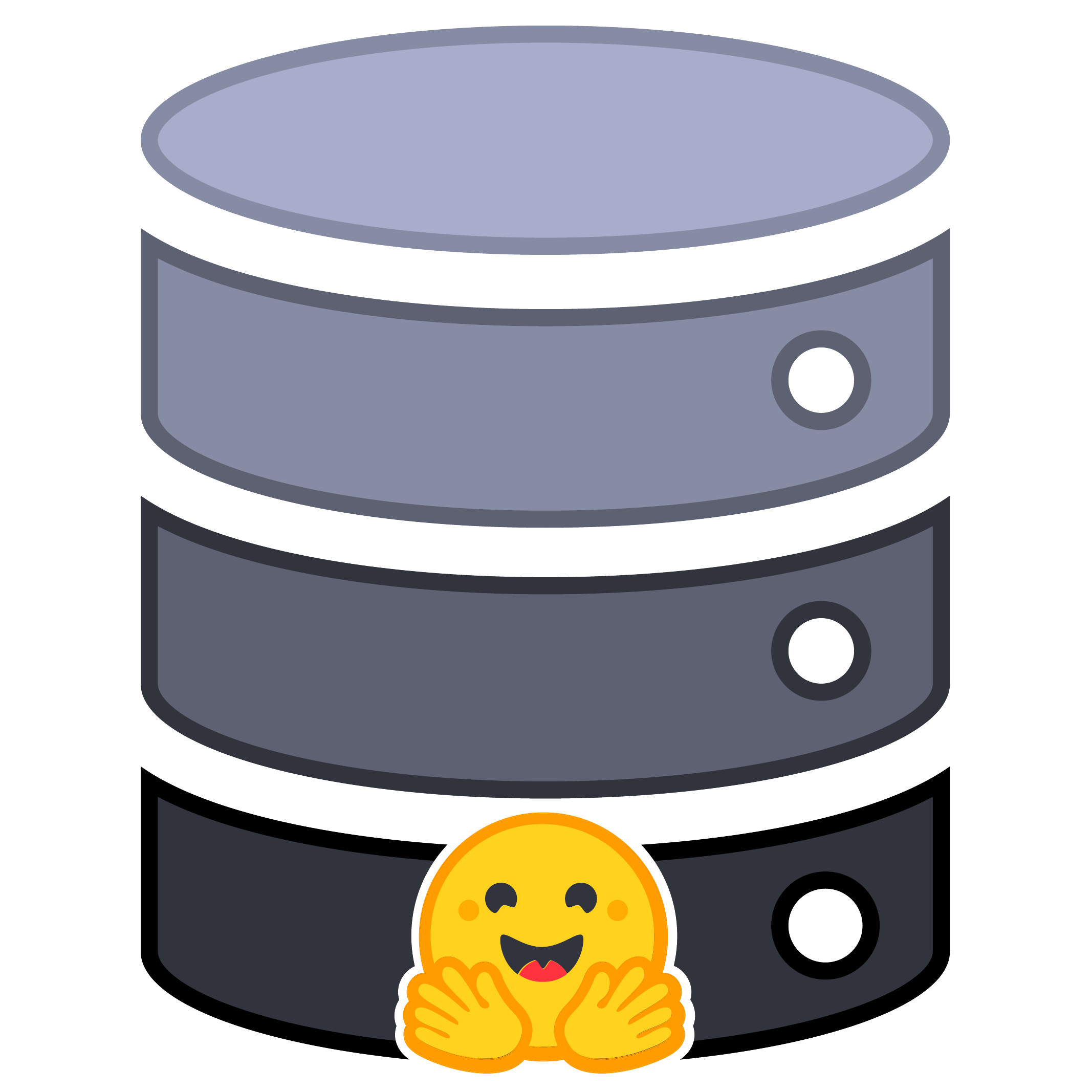}}\xspace}
\newcommand{\emailLogo}{\raisebox{-1.5pt}{\includegraphics[height=1.05em]{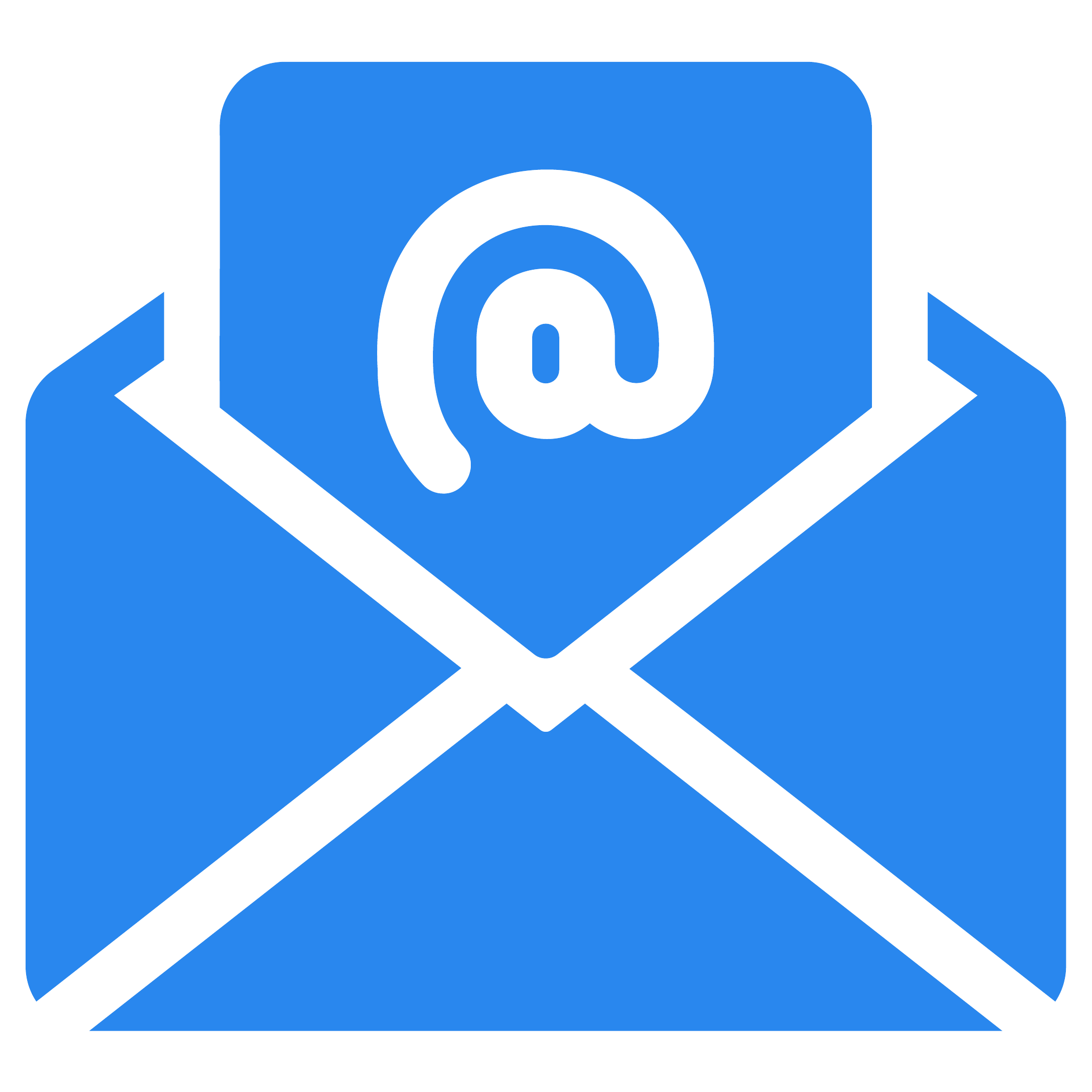}}\xspace}
\newcommand{\github}{\raisebox{-1.5pt}{\includegraphics[height=1.05em]{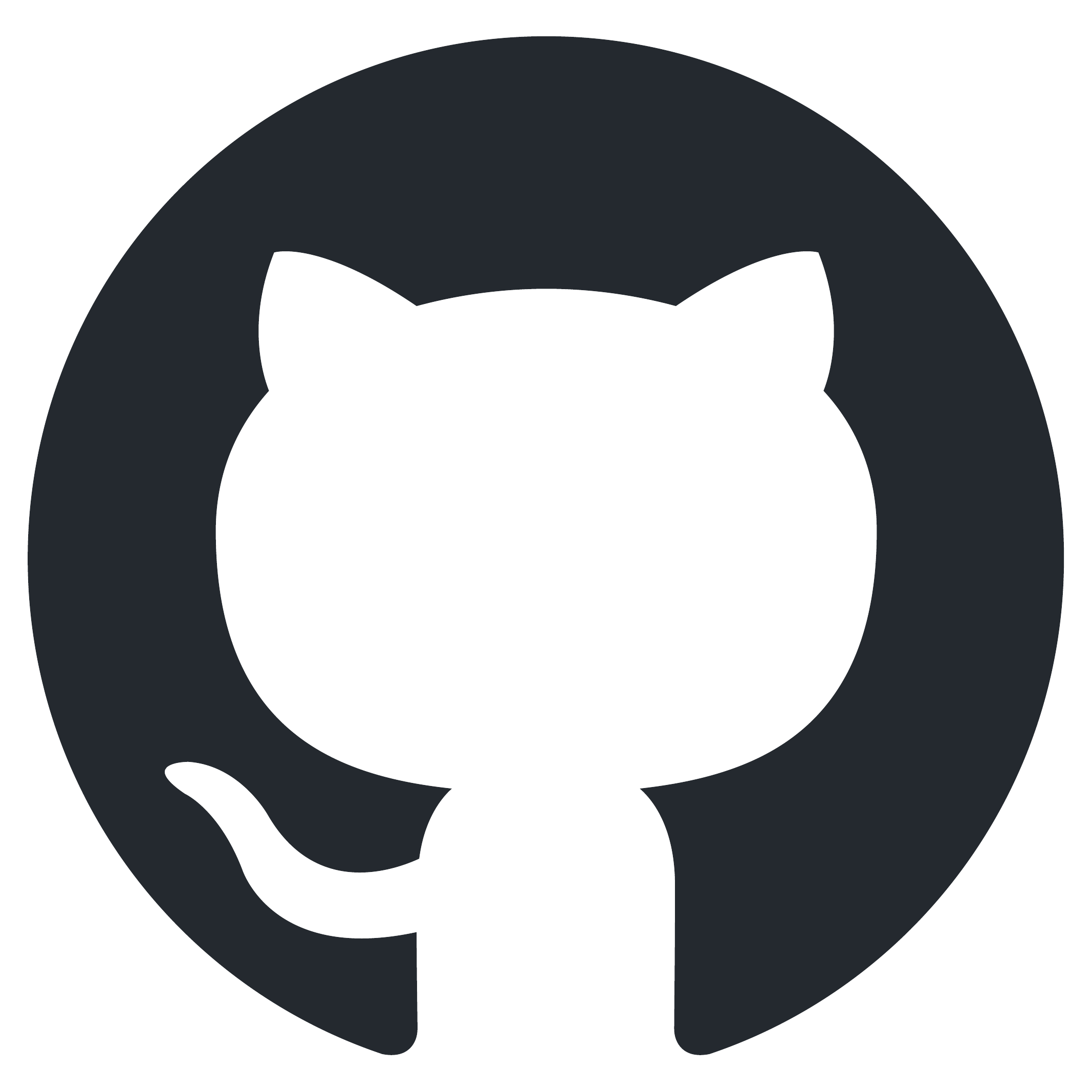}}\xspace}
\newcommand{\wandb}{\raisebox{-1.5pt}{\includegraphics[height=1.05em]{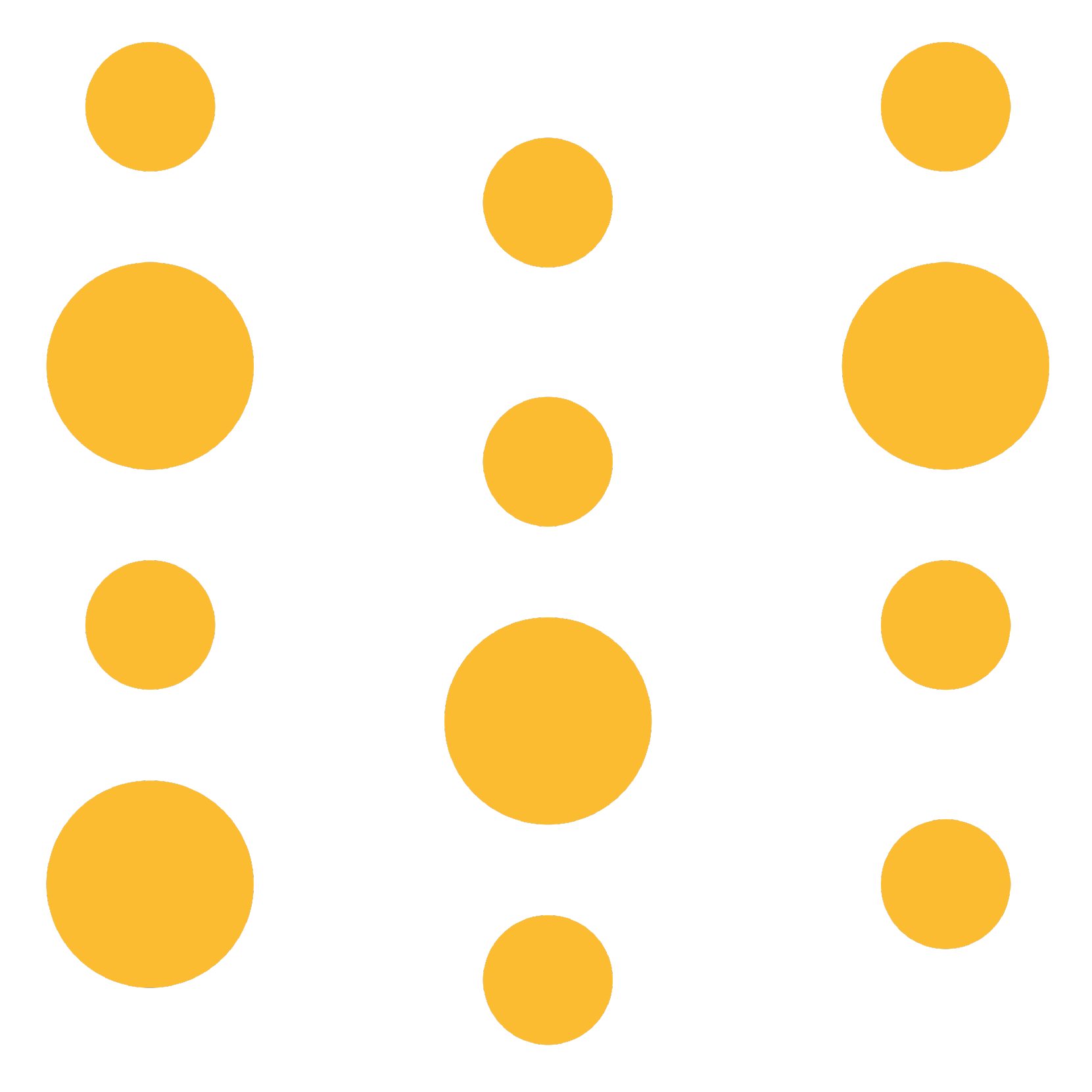}}\xspace}

\newcommand{\scale}{\raisebox{-1.5pt}{\includegraphics[height=1.05em]{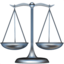}}\xspace}
\newcommand{\shrug}{\raisebox{-1.5pt}{\includegraphics[height=1.05em]{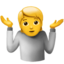}}\xspace}
\newcommand{\trophy}{\raisebox{-1.5pt}{\includegraphics[height=1.05em]{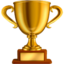}}\xspace}
\newcommand{\cow}{\raisebox{-1.5pt}{\includegraphics[height=1.05em]{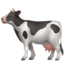}}\xspace}
\newcommand{\cowface}{\raisebox{-1.5pt}{\includegraphics[height=1.05em]{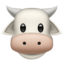}}\xspace}
\newcommand{\trex}{\raisebox{-1.5pt}{\includegraphics[height=1.05em]{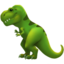}}\xspace}
\newcommand{\sparkles}{\raisebox{-1.5pt}{\includegraphics[height=1.05em]{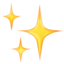}}\xspace}
\newcommand{\devilsmile}{\raisebox{-1.5pt}{\includegraphics[height=1.05em]{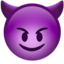}}\xspace}

\newcommand{\oldOlmo}{OLMo\xspace}
\newcommand{\newOlmo}{Olmo\xspace}

\newcommand{\tulu}{T\"ulu~3\xspace}

\newcommand{\olmotoo}{\textsc{\oldOlmo~2}\xspace}
\newcommand{\olmotooinstruct}{\textsc{\oldOlmo~2 Instruct}\xspace}
\newcommand{\olmothree}{\textsc{\newOlmo~3}\xspace}
\newcommand{\olmothreebase}{\textsc{\newOlmo~3 Base}\xspace}
\newcommand{\olmothreeinstruct}{\textsc{\newOlmo~3 Instruct}\xspace}

\newcommand{\olmothreerl}{\textsc{\newOlmo{}RL}\xspace}
\newcommand{\olmothreerlzero}{\textsc{\newOlmo~3 RL-Zero}\xspace}
\newcommand{\olmothreerlzeromath}{\olmothreerlzero \textsc{Math}\xspace}

\newcommand{\olmothreeonethink}{\textsc{\newOlmo~3.1~Think}\xspace}
\newcommand{\olmothreeoneinstruct}{\textsc{\newOlmo~3.1 Instruct}\xspace}
\newcommand{\olmothreeonerlzero}{\textsc{\newOlmo~3.1~RL-Zero}\xspace}

\newcommand{\olmocore}{{\oldOlmo-core}\xspace}
\newcommand{\openinstruct}{{Open~Instruct}\xspace}

\newcommand{\roundOne}{{Round~1}\xspace}
\newcommand{\roundThree}{{Round~3}\xspace}
\newcommand{\roundFive}{{Round~5}\xspace}

\newcommand{\metric}[1]{\hspace{0.7em}#1}

\newcommand{\olmothreethinking}{\textsc{Olmo~3 Think}\xspace}
\newcommand{\olmothreethinkingsft}{\textsc{Olmo~3 Think SFT}\xspace}
\newcommand{\dolma}{\textsc{Dolma}\xspace}

\newcommand{\dolmatoo}{\textsc{Dolma~3}\xspace}
\newcommand{\dolmathree}{\dolmatoo}
\newcommand{\dolci}{\textsc{Dolci}\xspace}
\newcommand{\dolcithink}{\textsc{Dolci Think}\xspace}
\newcommand{\dolcithinksft}{\textsc{Dolci Think SFT}\xspace}
\newcommand{\dolcithinkdpo}{\textsc{Dolci Think DPO}\xspace}
\newcommand{\dolcithinkrl}{\textsc{Dolci Think RL}\xspace}
\newcommand{\dolciinstruct}{\textsc{Dolci Instruct}\xspace}
\newcommand{\dolciinstructsft}{\textsc{Dolci Instruct SFT}\xspace}
\newcommand{\dolciinstructdpo}{\textsc{Dolci Instruct DPO}\xspace}

\newcommand{\dolcirlzero}{\textsc{Dolci RL-Zero}\xspace}

\newcommand{\dolmatoomix}{\textsc{Dolma 3 Mix}\xspace}
\newcommand{\dolminos}{\textsc{\oldOlmo~2 Dolmino Mix}\xspace}
\newcommand{\dolminostoo}{\textsc{Dolma~3 Dolmino Mix}\xspace}
\newcommand{\longminomix}{\textsc{Dolma~3 Longmino Mix}\xspace}
\newcommand{\longminoPool}{\textsc{Dolma~3 Longmino Pool}\xspace}

\newcommand{\olmocr}{\textsc{olmOCR}\xspace}
\newcommand{\olmOCR}{\olmocr}
\newcommand{\olmothreeeval}{\textsc{OlmoBaseEval}\xspace}
\newcommand{\olmix}{\textsc{Olmix}\xspace}
\newcommand{\olmorl}{\olmothreerl}
\newcommand{\olmocrPDF}{{\olmocr science PDFs}\xspace}

\title{Olmo 3}%

\authorOne[\starOlmo]{Olmo~Team}

\authorTwo[\allenAiAff]{Allyson Ettinger\coreContrib}
\authorTwo[\allenAiAff\commaAff\cmuAff]{Amanda Bertsch\coreContrib}
\authorTwo[\allenAiAff]{Bailey Kuehl\coreContrib}
\authorTwo[\allenAiAff]{David Graham\coreContrib}
\authorTwo[\allenAiAff]{David Heineman\coreContrib}
\authorTwo[\allenAiAff]{Dirk Groeneveld\coreContrib}
\authorTwo[\allenAiAff]{Faeze Brahman\coreContrib}
\authorTwo[\allenAiAff]{Finbarr Timbers\coreContrib}
\authorTwo[\allenAiAff\commaAff\uwAff]{Hamish Ivison\coreContrib}
\authorTwo[\allenAiAff\commaAff\uwAff]{Jacob Morrison\coreContrib}
\authorTwo[\allenAiAff]{Jake Poznanski\coreContrib}
\authorTwo[\allenAiAff\commaAff\uwAff]{Kyle Lo\coreContrib}
\authorTwo[\allenAiAff]{Luca Soldaini\coreContrib}
\authorTwo[\allenAiAff]{Matt Jordan\coreContrib}
\authorTwo[\allenAiAff\commaAff\stanfordAff]{Mayee Chen\coreContrib}
\authorTwo[\allenAiAff\commaAff\milaAff\commaAff\montrealAff]{Michael Noukhovitch\coreContrib}
\authorTwo[\allenAiAff]{Nathan Lambert\coreContrib}
\authorTwo[\allenAiAff]{Pete Walsh\coreContrib}
\authorTwo[\allenAiAff]{Pradeep Dasigi\coreContrib}
\authorTwo[\allenAiAff]{Robert Berry\coreContrib}
\authorTwo[\allenAiAff]{Saumya Malik\coreContrib}
\authorTwo[\allenAiAff]{Saurabh Shah\coreContrib}
\authorTwo[\allenAiAff\commaAff\uwAff]{Scott Geng\coreContrib}
\authorTwo[\allenAiAff]{Shane Arora\coreContrib}
\authorTwo[\allenAiAff]{Shashank Gupta\coreContrib}
\authorTwo[\allenAiAff]{Taira Anderson\coreContrib}
\authorTwo[\allenAiAff]{Teng Xiao\coreContrib}
\authorTwo[\allenAiAff]{Tyler Murray\coreContrib}
\authorTwo[\allenAiAff]{Tyler Romero\coreContrib}
\authorTwo[\allenAiAff\commaAff\uwAff]{Victoria Graf\coreContrib}

\authorThree[\allenAiAff\commaAff\cmuAff]{Akari Asai}
\authorThree[\allenAiAff]{Akshita Bhagia}
\authorThree[\princetonAff]{Alexander Wettig}
\authorThree[\uwAff]{Alisa Liu}
\authorThree[\allenAiAff]{Aman Rangapur}
\authorThree[\allenAiAff]{Chloe Anastasiades}
\authorThree[\allenAiAff]{Costa Huang}
\authorThree[\allenAiAff]{Dustin Schwenk}
\authorThree[\allenAiAff]{Harsh Trivedi}
\authorThree[\allenAiAff\commaAff\uwAff]{Ian Magnusson}
\authorThree[\allenAiAff]{Jaron Lochner}
\authorThree[\allenAiAff]{Jiacheng Liu}
\authorThree[\allenAiAff]{Lester James V. Miranda}
\authorThree[\allenAiAff\commaAff\cmuAff]{Maarten Sap}
\authorThree[\allenAiAff]{Malia Morgan}
\authorThree[\allenAiAff]{Michael Schmitz}
\authorThree[\allenAiAff]{Michal Guerquin}
\authorThree[\allenAiAff]{Michael Wilson}
\authorThree[\allenAiAff]{Regan Huff}
\authorThree[\allenAiAff]{Ronan Le Bras}
\authorThree[\uwAff]{Rui Xin}
\authorThree[\uwAff]{Rulin Shao}
\authorThree[\allenAiAff]{Sam Skjonsberg}
\authorThree[\mitAff]{Shannon Zejiang Shen}
\authorThree[\uwAff]{Shuyue Stella Li}
\authorThree[\allenAiAff]{Tucker Wilde}
\authorThree[\allenAiAff]{Valentina Pyatkin}
\authorThree[\allenAiAff]{Will Merrill}
\authorThree[\umdAff]{Yapei Chang}
\authorThree[\allenAiAff]{Yuling Gu}
\authorThree[\allenAiAff\commaAff\uwAff]{Zhiyuan Zeng}

\authorFour[\allenAiAff]{Ashish Sabharwal}
\authorFour[\uwAff]{Luke Zettlemoyer}
\authorFour[\allenAiAff\commaAff\uwAff]{Pang Wei Koh}

\authorFive[\allenAiAff\commaAff\uwAff]{Ali Farhadi}
\authorFive[\allenAiAff\commaAff\uwAff]{Noah A.~Smith\coreContrib}
\authorFive[\allenAiAff\commaAff\uwAff]{Hannaneh Hajishirzi\coreContrib}

\affiliation[\allenAiAff]{Allen~Institute~for~AI}
\affiliation[\uwAff]{University~of~Washington}
\affiliation[\cmuAff]{Carnegie~Mellon~University}
\affiliation[\stanfordAff]{Stanford~University}
\affiliation[\milaAff]{Mila}
\affiliation[\montrealAff]{Universit\'{e}~de~Montr\'{e}al}
\affiliation[\princetonAff]{Princeton~University}
\affiliation[\mitAff]{Massachusetts~Institute~of~Technology}
\affiliation[\umdAff]{University~of~Maryland}

\contribution[]{\vspace{-6pt}\starOlmo\olmothree was a team effort; authors sorted alphabetically. \coreContrib{marks core contributors}. See author contributions \hyperref[sec:contrib]{here}.}

\newtcolorbox{prompt}[1]{colback=gray!5,colframe=ai2pink!80,fonttitle=\bfseries, title={#1}}

\abstract{
We introduce {\bf{\olmothree}}, a family of state-of-the-art, fully-open language models at the 7B and 32B parameter scales.  \olmothree model construction targets long-context reasoning, function calling, coding, instruction following, general chat, and knowledge recall.  This release includes the entire {\bf{model flow}}, i.e., the full lifecycle of the family of  models, including every stage, checkpoint, data point, and dependency used to build it.  Our flagship model, {\bf{\olmothreeonethink 32B}}, is the strongest fully-open thinking model released to-date.
}

\metadata[\quad\huggingface Olmo 3 Base:]{
\href{https://huggingface.co/allenai/Olmo-3-1025-7B}{\texttt{Olmo-3-1025-7B}}\quad \href{https://huggingface.co/allenai/Olmo-3-1125-32B}{\texttt{Olmo-3-1125-32B}}}

\metadata[\quad\huggingface Olmo 3 Think:]{
\href{https://huggingface.co/allenai/Olmo-3-7B-Think}{\texttt{Olmo-3-7B-Think}}\quad
\texttt{Olmo-\{\href{https://huggingface.co/allenai/Olmo-3-32B-Think}{3}|\href{https://huggingface.co/allenai/Olmo-3.1-32B-Think}{3.1}\}-32B-Think}
}

\metadata[\quad\huggingface Olmo 3 Instruct:]{
\href{https://huggingface.co/allenai/olmo-3-7b-instruct}{\texttt{Olmo-3-7B-Instruct}}
\quad
\href{https://huggingface.co/allenai/olmo-3.1-32B-Instruct}{\texttt{Olmo-3.1-32B-Instruct}}
}

\metadata[\quad\huggingface Olmo 3 RL Zero:]{
\texttt{Olmo-3-7B-RL-Zero-\{\href{https://huggingface.co/allenai/Olmo-3-7B-RL-Zero-Math}{Math}|\href{https://huggingface.co/allenai/Olmo-3-7B-RL-Zero-Code}{Code}|\href{https://huggingface.co/allenai/Olmo-3-7B-RL-Zero-IF}{IF}|\href{https://huggingface.co/allenai/Olmo-3-7B-RL-Zero-General}{General}|\href{https://huggingface.co/allenai/Olmo-3-7B-RL-Zero-Mix}{\texttt{Mix}}\}}
\quad
\texttt{Olmo-3.1-7B-RL-Zero-\{\href{https://huggingface.co/allenai/Olmo-3.1-7B-RL-Zero-Math}{Math}|\href{https://huggingface.co/allenai/Olmo-3.1-7B-RL-Zero-Code}{Code}\}
}
}

\metadata[\vspace{.5em}\quad\hfdataset Base Data:]{
    {\texttt{Pretrain:}}
    \href{https://huggingface.co/datasets/allenai/dolma3_mix-6T-1025}{\texttt{Dolma 3 Mix}}
    \quad
    {\texttt{Midtrain:}}
    \href{https://huggingface.co/datasets/allenai/dolma3_dolmino_mix-100B-1125}{\texttt{Dolma 3 Dolmino Mix}}
    \quad
    {\texttt{Long-ctx:}}
    \href{https://huggingface.co/datasets/allenai/dolma3_longmino_mix-100B-1125}{\texttt{Dolma 3 Longmino Mix}}
}

\metadata[\vspace{.25em}\quad\hfdataset Think Data:]{
    \texttt{Dolci-Think-\{\href{https://huggingface.co/datasets/allenai/Dolci-Think-SFT-7B}{SFT}|\href{https://huggingface.co/datasets/allenai/Dolci-Think-DPO-7B}{DPO}|\href{https://huggingface.co/datasets/allenai/Dolci-Think-RL-7B}{RL}\}-7B}
    \quad
    \texttt{Dolci-Think-\{\href{https://huggingface.co/datasets/allenai/Dolci-Think-SFT-32B}{SFT}|\href{https://huggingface.co/datasets/allenai/Dolci-Think-DPO-32B}{DPO}|\href{https://huggingface.co/datasets/allenai/Dolci-Think-RL-32B}{RL}\}-32B}
}

\metadata[\vspace{.25em}\quad\hfdataset Instruct Data:]{
    \texttt{Dolci-Instruct-\{\href{https://huggingface.co/datasets/allenai/Dolci-Instruct-SFT-7B}{SFT}|\href{https://huggingface.co/datasets/allenai/Dolci-Instruct-DPO-7B}{DPO}|\href{https://huggingface.co/datasets/allenai/Dolci-Instruct-RL-7B}{RL}\}}
}

\metadata[\vspace{.25em}\quad\hfdataset RL-Zero Data:]{
    \texttt{Dolci-RL-Zero-\{\href{https://huggingface.co/datasets/allenai/Dolci-RL-Zero-Math-7B}{Math}|\href{https://huggingface.co/datasets/allenai/Dolci-RL-Zero-Code-7B}{Code}|\href{https://huggingface.co/datasets/allenai/Dolci-RL-Zero-IF-7B}{IF}|\href{https://huggingface.co/datasets/allenai/Dolci-RL-Zero-IF-7B}{General}\}-7B}
    \quad
    \href{https://huggingface.co/datasets/allenai/Dolci-RL-Zero-Mix-7B}{\texttt{Dolci-RL-Zero-Mix-7B}}
}

\metadata[\vspace{.5em}\quad\github Training Code:]{
    \href{https://github.com/allenai/olmo-core}{\texttt{\olmocore}}~\sans{(pretrain)} \quad
    \href{https://github.com/allenai/open-instruct}{\texttt{\openinstruct}}~\sans{(posttrain)}
}
\metadata[\quad\github Data Code:]{
    \href{https://github.com/allenai/datamap-rs}{\texttt{datamap-rs}}~\sans{(data processing)} \quad \href{https://github.com/allenai/duplodocus}
    {\texttt{duplodocus}}~\sans{(deduplication)}\quad
    \href{https://github.com/allenai/dolma3}{\texttt{dolma3}}~\sans{(data recipes)}
}
\metadata[\quad\github Eval Code:]{
    \href{https://github.com/allenai/olmes}{\texttt{OLMES}}~\sans{(eval suite)}\quad
    \href{https://github.com/allenai/decon}{\texttt{decon}}~\sans{(eval decontamination)}
}
\metadata[\vspace{.5em}\quad\wandb Training Logs:]{
\texttt{Olmo-3-7B-\{\href{https://api.wandb.ai/links/ai2-llm/y8mbjvll}{Base}|\href{https://wandb.ai/ai2-llm/Olmo-3-7B-Think/reports/Olmo-3-7B-Think-SFT-DPO-RL--VmlldzoxNTE3ODQzMA}{Think}|\href{https://wandb.ai/ai2-llm/Olmo-3-7B-Instruct/reports/Olmo-3-7B-Instruct-SFT-DPO-RL--VmlldzoxNTE3ODk3Mg}{Instruct}|\href{https://wandb.ai/ai2-llm/Olmo-3-7B-RL-Zero/reports/Olmo-3-7B-RL-Zero--VmlldzoxNTM0OTI1Nw}{RL-Zero}\}}
\quad
\texttt{Olmo-3-32B-\{\href{https://wandb.ai/ai2-llm/Olmo-3-1125-32B/reports/Olmo-3-32B-November-2025--VmlldzoxNTA4NzAxMw}{Base}|\href{https://wandb.ai/ai2-llm/Olmo-3-32B-Think/reports/Olmo-3-32B-Think-SFT-DPO-RL--VmlldzoxNTE3OTA5Mg}{Think}|\href{https://wandb.ai/ai2-llm/Olmo-3-32B-Instruct/reports/Olmo-3-32B-Instruct-SFT-DPO-RL--VmlldzoxNTM0OTIzNw}{Instruct}\}
}
}

\metadata[\vspace{.25em}\quad\aitoo Demo:]{\href{https://playground.allenai.org/?model=Olmo-3.1-32B-Think}{\sans{32B Think}} \quad 
\href{https://playground.allenai.org/?model=Olmo-3.1-32B-Instruct}{\sans{32B Instruct}} \quad 
\href{http://playground.allenai.org/?model=Olmo-3-7B-Think}{\sans{7B Think}} \quad 
\href{http://playground.allenai.org/?model=Olmo-3-7B-Instruct}{\sans{7B Instruct}}}

\metadata[\vspace{.25em}\quad\emailLogo Contact:]{\color{ai2accent}{\href{mailto:olmo@allenai.org}{\texttt{olmo@allenai.org}}}}

\begin{document}

\maketitle

\newpage
\setcounter{tocdepth}{2}
\tableofcontents
\newpage

\section{Introduction}
\label{sec:intro}

We introduce {\bf{\olmothree}}, a family of state-of-the-art, fully-open language and thinking models at the 7B and 32B parameter scales with a diverse set of capabilities, including long-context reasoning, function calling, coding, instruction following, general chat, and knowledge recall. The \olmothree release provides complete access to its entire {\bf model flow}---the full lifecycle of a language model, including every stage, checkpoint, datapoint, and dependency required to create it. This enables infinite customization through intervention at any stage of the model development process---not just the final weights.

To truly advance open-source AI research and development, we argue that releasing a state-of-the-art language model should make its entire model flow---not just its endpoint---transparent and accessible. With the \olmothree\ release, we provide complete access to the pathways we charted throughout the model flow, from initial conception to the creation of state-of-the-art, fully-open language models.

Specifically, we train {\bf{\olmothreebase}} as a foundation on which to build models with thinking and tool-use capabilities. From \olmothreebase{} we develop our flagship model, {\bf{\olmothreethinking}}, trained to perform step-by-step reasoning by generating intermediate thinking traces before producing a final answer. {\olmothreethinking~32B} is the strongest fully-open thinking model, narrowing the gap to the best open-weight models of similar scale, such as the Qwen 3 32B thinking~\citep{qwen3} on our suite of reasoning benchmarks, while being trained on six times fewer tokens. 
Because of our fully-open approach, the \olmothree{} release also enables reasoning chains to be traced back to their original training data, unlocking research opportunities  not possible with any other thinking model.

\begin{figure}[h]
    \centering
    \includegraphics[width=\textwidth]{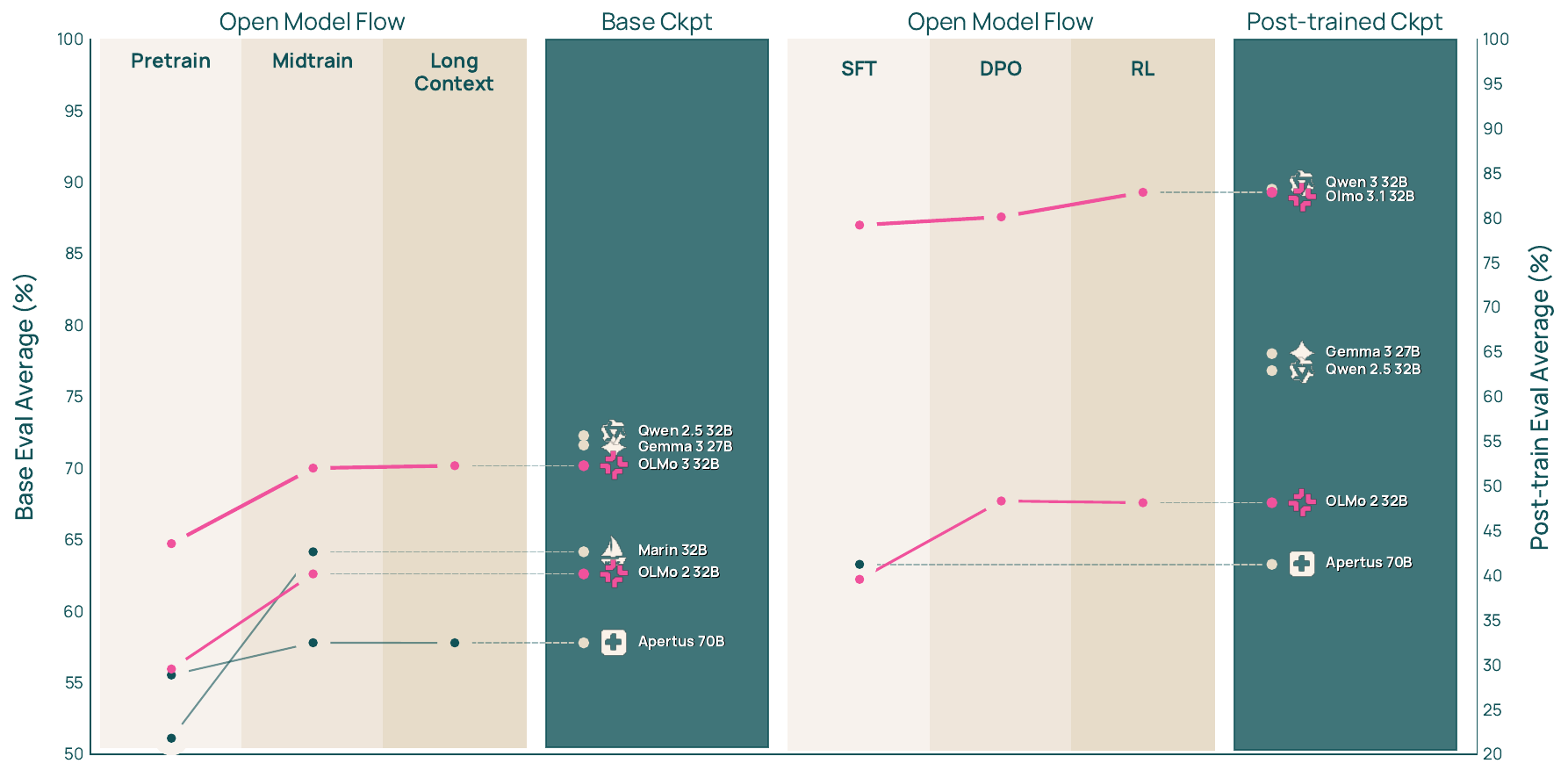}
    \caption{
        \textbf{The model flow encompasses training data, code and intermediate checkpoints for all stages of development}. 
        While both fully-open and open-weights models release their final checkpoints \textcolor[HTML]{105257}{\bf{(dark teal)}}, 
        fully-open releases like Marin, Apertus, and \newOlmo provide data along their model flow, enabling the careful study of intermediate development stages \textcolor[HTML]{a1947d}{\bf{(beige)}}.
        \olmothreethinking 32B is shown here along with other open models of comparable size and architecture. 
        \olmothreethinking is competitive with Qwen 3 32B, which does not have a released base model.
        Its underlying \olmothreebase 32B surpasses all other fully-open base models. 
    }
    \label{fig:model-flow-hero}
\end{figure}

In addition, we train {\bf{\olmothreeinstruct} 7B} and {\bf 32B} models tuned to produce shorter, more direct responses. By avoiding intermediate ``thinking'' outputs, \olmothreeinstruct effectively reduces response latency and is optimized for general chat and function calling. 
\olmothreeinstruct 7B and 32B surpass other notable open-weight models of comparable size---Qwen 2.5~\citep{qwen2.5}, Gemma 3~\citep{team2025gemma3}, IBM Granite 3.3~\citep{souleBergmann2025granite33}, and Llama 3~\citep{dubey2024llama}---and additionally reduces the remaining performance gap to Qwen 3~\citep{qwen3}. %
Finally, we introduce {\bf{\olmothreerlzero 7B}}, a variant of \olmothree trained using RL directly from \olmothreebase.
\olmothreerlzero enables researchers to study how base model data affects RL performance.

The Olmo 3 family is the strongest collection of fully-open base models, outperforming Stanford Marin~\citep{Hall2025marin}, Apertus~\citep{swissai2025apertus}, and LLM360 K2-V2~\citep{k2team2025k2v2360openreasoningenhancedllm}.
To achieve these results, we construct new datasets for every stage of the model flow. 
This includes {\bf{\dolmathree}}, our pretraining data mix encompassing carefully-sampled natural data from crawled sources, our midtraining mix of high-quality data designed to jump-start reasoning, and a large collection of science-focused PDF documents that unlock long-context support in \olmothree. We also introduce {\bf{\dolci}}, a post-training data suite that advances step-by-step reasoning during supervised finetuning, provides high-quality contrastive data for preference tuning, and offers challenging general and reasoning prompts for reinforcement learning.

Finally, we develop a set of new algorithmic and infrastructural advances across data processing, evaluation, pretraining, and reinforcement learning. This includes {\bf{\olmothreeeval}}, a benchmark suite tailored to compute-efficient base-model development, and {\bf{\olmothreerl}}, a reinforcement-learning framework incorporating efficiency optimizations tailored to our thinking models.
Taken together, these training recipes are shaped by a development framework that blends distributed experimentation with centralized evaluation, enabling coordinated, capability-driven improvements throughout the model pipeline.

\section{Model Flow for \olmothree}
\label{sec:olmo-family}

\begin{figure}[!h]
\centering
\includegraphics[width=\textwidth]{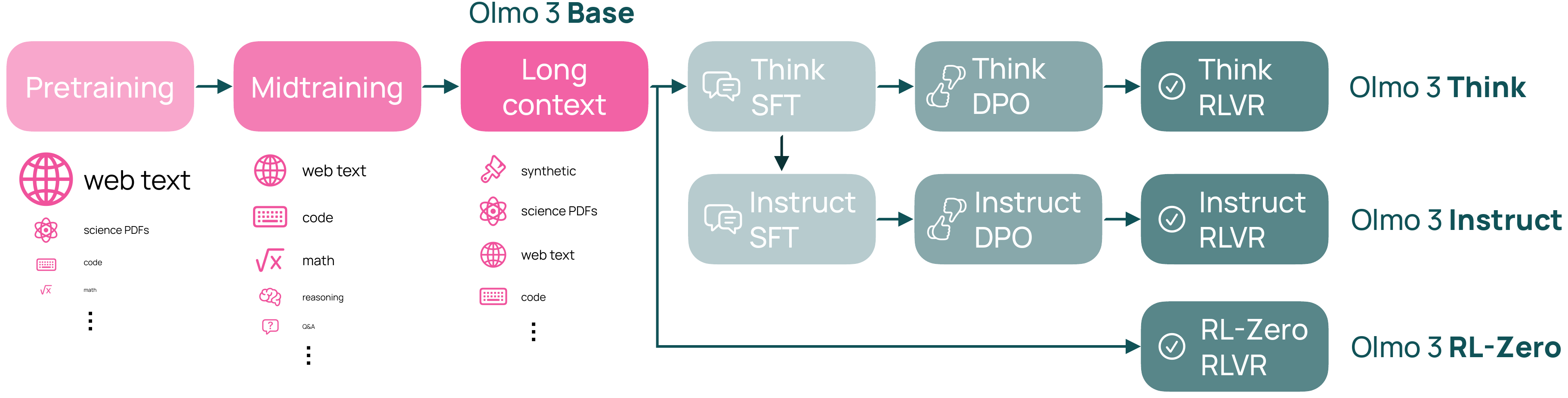}
\caption{\textbf{Depiction of model flow for \olmothree}. Development is divided into major \textcolor[HTML]{f0529c}{\bf{base model training (left)}} and \textcolor[HTML]{105257}{\bf{post-training (right)}} stages, each further divided into sub-stages with their own recipes (i.e., training data and method).
}
\label{fig:olmo3_pipeline}
\end{figure}

In this section, we provide a brief overview of all components of the model flow for \olmothree{}, highlighting our methodology for targeting reasoning and tool-use capabilities in ways that advance beyond \olmotoo~\citep{olmo20242olmo2furious} and other open-weight models. Subsequent sections will then provide deep dives into each of the model flow components. \olmothree{} training is divided into major stages of base model training and post-training, each further divided into sub-stages as outlined in Figure~\ref{fig:olmo3_pipeline}.

\subsection{Base Model Training}

We develop \olmothreebase in three stages of \emph{pretraining} for up to 5.9T tokens (Section~\S\ref{sec:pretraining}), \emph{midtraining} for 100 billion tokens (Section~\S\ref{sec:midtraining}), and the newly added \emph{long-context extension} for 50B (\olmothreebase 7B) or 100B (\olmothreebase 32B) tokens (Section~\S\ref{sec:long-context}).

\paragraph{Evaluation}
We develop \olmothreeeval, a collection of benchmarking suites to support decision-making during base model development (pretraining and midtraining). Our goal is to be compute-efficient by making development decisions based on models trained at a small scale. The challenge is that such models can exhibit random-chance performance on certain tasks, and have  small differences in scores that are hard to distinguish from benchmark noise. To address this, we (1) aggregate scores over clusters of tasks that assess similar capabilities (Section~\S\ref{sec:eval:clustering}); (2) develop proxy metrics for evaluating small-scale models (Section~\S\ref{sec:eval:scaling-analysis}); and (3) improve overall signal-to-noise ratio by evaluating on more examples from noisy tasks or even removing them entirely (Section~\S\ref{sec:eval:signal-noise}).

\paragraph{Data curriculum}
We curate specialized datasets for each training stage,
with latter stages focused on strengthening capabilities crucial in post-training stages, such as math, code, reasoning, instruction following, and long-context understanding:
\begin{itemize}
\item {\bf{Pretraining}} 
We first train \olmothreebase on {\bf{\dolmatoomix}} (Section~\S\ref{sec:pretraining}), our 6T-token pretraining data mix.
While \dolmatoomix is largely comprised of the same types of data sources used in other open pretraining recipes~\citep{soldaini2024dolma,bakouch2025smollm3,olmo20242olmo2furious}, we demonstrate three key novelties:
\begin{itemize}[itemsep=2pt]
    \item[$\circ$] New tooling for fast and scalable global deduplication at the trillion-token scale;
    \item[$\circ$]  A novel source of academic PDFs---\olmocrPDF---converted to linearized plain text using \olmocr~\citep{poznanski2025olmocr,poznanski2025olmocr2unittest};
    \item[$\circ$]  Two new methods for optimizing selection of training tokens: token-constrained mixing and quality-aware upsampling.
\end{itemize}
\item {\bf{Midtraining}} We continue training on {\bf{\dolminostoo}} (Section~\S\ref{sec:midtraining}), our 100B-token data curated to boost target capabilities across code, math, and general knowledge QA domains through the introduction of: %
\begin{itemize}[itemsep=2pt]
    \item[$\circ$] A new two-part methodological framework combining 1) lightweight, distributed feedback loops on individual data sources, with 2) centralized integration tests to assess candidate mixes on base model quality and post-trainability.
    \item[$\circ$] Intentional inclusion instruction data and thinking traces to lay groundwork for post-training.
\end{itemize}
\item {\bf{Long-context extension}} Through {\bf{\longminomix}} (Section~\S\ref{sec:long-context}), \olmothree  supports long-context input and output, a crucial feature to unlock reasoning and tool-use capabilities.
\begin{itemize}[itemsep=2pt]
\item[$\circ$] Documents in \olmocrPDF enable of our long-context approach;
with over 22.3M documents of length above 8K tokens (640B tokens total), and 4.5M documents over 32K tokens (380B tokens total), this collection is the largest openly available for long-context research.
\item[$\circ$] As result, \olmothree is our first model with long-context capabilities, supporting up to 65K context after extension. 
\olmothreebase 32B rivals performance of Qwen 2.5 32B, Mistral Small 3.1 24B, and Gemma 3 27B on long-context benchmarks, despite a short extension stage (50B for 7B, 100B for 32B). 
\end{itemize}
\end{itemize}

\paragraph{Open artifacts}
We release all of our intermediate checkpoints as well as the final models at the end of each stage of training.
For data, we release both our {\bf{data mixes}}, which are the actual tokens used for base model training,\footnote{A data mix may involve upsampling or repeating data from a data pool.} as well as our full source {\bf{data pools}} for each stage---9T tokens of cleaned source tokens for pretraining, and 2T and 640B tokens of specialized data for midtraining and long-context extension respectively.
For pretraining, in addition to our actual training mix for \olmothreebase, we also release smaller sample mixes for accessible experimentation with less compute (150B for pretraining and 10B for midtraining).

\subsection{Post-training}
We post-train \olmothreebase{} into three model variants:

\begin{itemize}
\item {\bf \olmothreethinking{}} (Section~\S\ref{sec:posttrain-thinking}) is trained to perform extended reasoning by generating a structured thinking trace before a final answer. We train it via SFT, DPO, and RLVR, observing gains at each stage.

\begin{itemize}[itemsep=2pt]
    \item[$\circ$] We introduce {\bf\dolcithinksft} (Section~\S\ref{sec:thinking_sft_recipe}), {\bf\dolcithinkdpo} (Section~\S\ref{sec:thinking_dpo_recipe}), and {\bf\dolcithinkrl} (Section~\S\ref{sec:thinking_rl_recipe}), new post-training datasets designed to target a broad range of key capabilities such as math, coding, instruction following, and general conversation.
    The dataset includes synthetic examples with long thinking traces for supervised finetuning, high-quality contrastive data following the insights from Delta Learning \citep{geng2025delta}, and challenging prompts for reinforcement learning across both verifiable and non-verifiable domains. In particular, our new approach to curating contrastive instances for preference tuning expands the reasoning frontier of the model beyond what SFT alone can provide and primes the model for effective reinforcement learning.
    \item[$\circ$] We introduce algorithmic and infrastructural advances in reinforcement learning with verifiable rewards (Section~\S\ref{sec:thinking_rl_recipe}). This approach generalizes verifiable reasoning to multiple domains, expanding beyond the settings explored in \olmotoo{} to include code and general chat. Our improvements enable longer and more stable RL runs across diverse domains and increase the overall efficiency of training cycles, leading to a 4x speedup in RL training. Notably, we introduce {\bf \olmothreeonethink 32B} to illustrate that extended \olmothreerl training leads to improved performance.
\end{itemize}
\item {\bf \olmothreeinstruct{}} (Section~\S\ref{sec:posttrain-instruct}) is trained to produce efficient and helpful responses to user queries without generating internal thinking traces. This model prioritizes typical user needs, such as avoiding excessive verbosity for easy user understanding and function-calling for user information seeking.
In such settings, thinking traces are unnecessary, and inference-time efficiency matters more than inference-time scaling.
\begin{itemize}[itemsep=2pt]
    \item[$\circ$] We introduce \dolciinstructsft, our new dataset enriched with data specifically created for function calling (Section~\S\ref{sec:tool-use-sft}). To directly optimize model interactivity on top of capabilities, we extend our Delta Learning preference pipeline in \dolciinstructdpo, incorporating multi-turn preference data and targeted data length interventions that encourage concise responses (Section~\S\ref{sec:dolci-instruct-dpo}). Finally, we use reinforcement learning with verifiable rewards (Section~\S\ref{sec:RLVR-instruct}) to further refine core capabilities, where preference tuning synergizes with RL to improve model performance while maintaining learned brevity.
\end{itemize}
\item {\bf \olmothreerlzero{}} (Section~\S\ref{sec:rlzero}) %
To date, all leading open RLVR benchmarks and algorithms train on top of open-weight models that do not reveal their pretraining or mid-training data \citep{scalingllama3,qwen3}.
This limits the community's ability to study the role of pretraining data on RLVR performance.
It can lead to myriad issues with benchmark evaluations being contaminated, e.g., mid-training data containing the evaluation, which makes spurious rewards as effective as true reward~\citep{shao2025spuriousrewardsrethinkingtraining,wu2025reasoning} or improvements from fixing prompt templates outweighing the improvements from RL~\citep{liu2025understanding}.
\begin{itemize}[itemsep=2pt]
    \item[$\circ$]  We therefore release a fully open dataset \dolcirlzero, an algorithmic RL zero setup for \olmothree, and open-source \olmothreerl code to enable clear benchmarking in the RL research community.
We perform RLVR from \olmothreebase over four benchmarking domains to create the \olmothreerlzero family: math, code, precise instruction following (IF) and a general mix.
In all cases, we further decontaminate \dolcirlzero from pretraining and midtraining data to guarantee our setup carefully studies the effect of RLVR without data leakage confounding our conclusions.
\end{itemize}
\end{itemize}

\begin{table}[t]
\centering
\footnotesize
\setlength\tabcolsep{5pt}
\renewcommand{\arraystretch}{0.95}
\adjustbox{max width=\linewidth}{
{\fontsize{8}{8}\selectfont
\begin{NiceTabular}{@{}Hl
P{38pt}C{38pt}C{38pt}C{38pt}|
C{33pt}C{33pt}C{33pt}C{33pt}C{33pt}C{33pt}@{}}
\toprule
& & \multicolumn{4}{c}{\quad \quad \textbf{\texttt{Fully-Open Models}}} & \multicolumn{6}{c}{\textbf{\texttt{Open-weight Models}}} \\
\textbf{Skill} &
& \textbf{OLMo 3.1 32B Think} & \textbf{OLMo 2 Instruct 32B} & \textbf{Apertus Instruct 70B} & \textbf{LLM360 K2-V2 Instruct 70B} & \textbf{Qwen 3 32B} & \textbf{Qwen 3 VL 32B Think} & \textbf{Qwen 2.5 32B} & \textbf{Gemma 3 27B} & \textbf{Gemma 2 27B} & \textbf{DS-R1 32B} \\
\midrule

\rowcolor{midgrey} -- & \textbf{Math} & & & & & & & & & & \\
\rowcolor{lightgrey} -- & \metric{MATH} & 96.2 & 49.2 & 36.2 & 94.5 & 95.4 & 96.7 & 80.2 & 87.4 & 51.5 & 92.6 \\
\rowcolor{lightgrey} -- & \metric{AIME 2024} & 80.6 & 4.6 & 0.3 & 78.4 & 80.8 & 86.3 & 15.7 & 28.9 & 4.7 & 70.3 \\
\rowcolor{lightgrey} -- & \metric{AIME 2025} & 78.1 & 0.9 & 0.1 & 70.3 & 70.9 & 78.8 & 13.4 & 22.9 & 0.9 & 56.3 \\
\rowcolor{lightgrey} -- & \metric{OMEGA} & 53.4 & 9.8 & 5.6 & 46.1 & 47.7 & 50.8 & 19.2 & 24.0 & 9.1 & 38.9 \\
\midrule

\rowcolor{midgrey} -- & \textbf{Reasoning} & & & & & & & & & & \\
\rowcolor{lightgrey} -- & \metric{BigBenchHard} & 88.6 & 65.6 & 57.0 & 87.6 & 90.6 & 91.1 & 80.9 & 82.4 & 66.0 & 89.7 \\
\rowcolor{lightgrey} -- & \metric{ZebraLogic} & 80.1 & 13.3 & 9.0 & 79.2 & 88.3 & 96.1 & 24.1 & 24.8 & 17.2 & 69.4 \\
\rowcolor{lightgrey} -- & \metric{AGI Eval English} & 89.2 & 68.4 & 61.7 & 89.6 & 90.0 & 92.2 & 78.9 & 76.9 & 70.9 & 88.1 \\
\midrule

\rowcolor{midgrey} -- & \textbf{Coding} & & & & & & & & & & \\
\rowcolor{lightgrey} -- & \metric{HumanEvalPlus} & 91.5 & 44.4 & 42.9 & 88.0 & 91.2 & 90.6 & 82.6 & 79.2 & 67.5 & 92.3 \\
\rowcolor{lightgrey} -- & \metric{MBPP+} & 68.3 & 49.0 & 45.8 & 66.0 & 70.6 & 66.2 & 66.6 & 65.7 & 61.2 & 70.1 \\
\rowcolor{lightgrey} -- & \metric{LiveCodeBench v3} & 83.3 & 10.6 & 9.7 &  78.4 & 90.2 & 84.8 & 49.9 & 39.0 & 28.7 & 79.5 \\
\midrule

\rowcolor{midgrey} -- & \textbf{IF} & & & & & & & & & & \\
\rowcolor{lightgrey} -- & \metric{IFEval} & 93.8 & 85.8 & 70.4 & 85.3 & 86.5 & 85.5 & 81.9 & 85.4 & 62.1 & 78.7 \\
\rowcolor{lightgrey} -- & \metric{IFBench} & 68.1 & 36.4 & 26.0 & 57.7 & 37.3 & 55.1 & 36.7 & 31.3 & 27.8 & 23.8 \\
\midrule

\rowcolor{midgrey} -- & \textbf{Knowledge \& QA} & & & & & & & & & & \\
\rowcolor{lightgrey} -- & \metric{MMLU} & 86.4 & 77.1 & 70.2 & 88.4 & 88.8 & 90.1 & 84.6 & 74.6 & 76.1 & 88.0 \\
\rowcolor{lightgrey} -- & \metric{PopQA} & 30.9 & 37.2 & 33.6 & 32.2 & 30.7 & 32.2 & 28.0 & 30.2 & 30.4 & 26.7 \\
\rowcolor{lightgrey} -- & \metric{GPQA} & 57.5& 36.4 & 27.9 & 64.0 & 67.3 & 67.4 & 44.6 & 45.0 & 39.9 & 61.8 \\
\midrule

\rowcolor{midgrey} -- & \textbf{Chat} & & & & & & & & & & \\
\rowcolor{lightgrey} -- & \metric{AlpacaEval 2 LC} & 69.1 & 38.0 & 19.9 &  - & 75.6 & 80.9 & 81.9 & 65.5 & 39.8 & 26.2 \\

\bottomrule

\end{NiceTabular}}}
\caption{\textbf{Results on our flagship model \olmothreeonethink~32B} on our post-training evaluation suite. \olmothreeonethink~32B is the best fully-open model at 32B.
}
\label{tab:32b-think-flagship-results}
\end{table}

\subsection{Results} Table~\ref{tab:32b-think-flagship-results} demonstrates a snapshot of our evaluation for \olmothreethinking compared to other open-weight and fully-open models. 
To the best of our knowledge, \olmothreethinking is the strongest fully-open thinking model to date.
It is better than Qwen2.5-Instruct, Gemma 2 and 3 27B, DeepSeek R1, and Distilled Qwen 32B; it is also close to Qwen 3 and Qwen 3 VL 32B models, narrowing the gap to the best open-weight models of similar scale while training on roughly 6x fewer tokens.

For more details and results of other models along our \olmothree model flow, refer to the quick links below.

\begin{itemize}
    \item {\bf{\olmothreebase}} Section~\S\ref{subsec:pretrain_base_model_results} for detailed evaluation discussion. Table~\ref{tab:transposed-super-base-table-32b} (32B) and Table~\ref{tab:transposed-super-base-table-7b} (7B) for main results. Table~\ref{tab:ruler_baselines} for long context evaluations. Table~\ref{tab:base_evals_overview} for pretraining vs midtraining vs long-context extension stages.
    \item {\bf{\olmothreethinking}} Section~\S\ref{sec:posttrain_eval} for detailed evaluation discussion. Table~\ref{tab:32b-think-baeslines} (32B) and Table~\ref{tab:post-train-eval-overview} (7B) for main results, including SFT vs DPO vs RL stages.
    \item {\bf{\olmothreeinstruct}} Section~\S\ref{sec:posttrain_eval_instruct} for detailed evaluation discussion. Table~\ref{tab:32b-instruct-baeslines} (32B) and Table~\ref{tab:7b-instruct-vs-baselines} (7B) for main results, including SFT vs DPO vs RL stages.
\end{itemize}

\subsection{Costs}

The cost of training large models is often reported as a single dollar figure, typically by converting GPU-hours at market rates to dollars, such as \$5.576M in H800-hours for DeepSeek V3~\citep{deepseekv3}.
To provide a more representative view of the resources required to train \olmothree~32B, we instead report the wall-clock time that elapses during training.

In total, approximately 56 days elapsed from the start of training to the evaluation of the \olmothreethinking 32B checkpoint, on a cluster with 1024 H100 GPUs dedicated to \olmothree. 
The 32B 3.1 Think and Instruct checkpoints were trained after this time period.
This training time is largely a reflection of \textit{applying our best recipe to the model}\footnote{The recipe was developed on 7B or smaller and applied to 32B rapidly.}, and does not include any substantial modifications or research ideas that could expand the timeline substantially.
At a price of \$2/H100 hour, this would cost \$2.75M. Runtime breakdown is as follows:

\begin{itemize}
    \item {\bf{Pretraining:}}~$\mathbf{\sim}${\bf{47 days}} (including midtraining and long-context stages) The initial pretraining phase on 5.5T tokens took about 9.5 days on 512 GPUs, followed by an additional 35 days on 1024 GPUs.
    These durations include all crash resumptions and other engineering concerns that kept us from running at full speed.
    Midtraining consisted of two parallel runs on 512 GPUs each, covering 100B tokens per run, followed by model merging and evaluations to decide on final checkpoints, taking about 1.5 days in total.
    Long-context extension was executed as a single run on 1024 GPUs; the full long-context stage---including training and all associated merges and evaluations---added approximately one additional day.
    \item {\bf{Post-training:}}~$\mathbf{\sim}${\bf{9 days}} (SFT, DPO, and RL) Post-training follows a different operational pattern in which we run each stage multiple times, sweeping over learning rates and other hyperparameters.
    The theory for post-training, particularly, RL, is less developed, so we have to run multiple experiments to identify the optimal hyperparameters for a given base model. 
    We hope to address this in future work.
    During post-training, checkpoint evaluation consumes a larger proportion of compute resources, in part due to long generations from reasoning models on core benchmarks.
    For SFT, we swept over four candidate learning rates, on 256 GPUs each, in parallel for 36 hours. Then approximately 12 hours was spent on evaluation, merging, and checkpoint confirmation, totaling approximately two days.
    DPO training takes less time per run (about 18 hours for a full learning-rate sweep on 64 GPUs per job) but in practice extended over multiple days due to cluster instability.
    The final RL runs for the initial \olmothreethinking 32B spanned approximately 5 days with at least a day of training time lost due to stability issues.
    After the initial release of \olmothree, we continued our best RL run for another 21 days on 224 GPUs to produce \olmothreeonethink 32B.
\end{itemize}

While pretraining accounts for the majority of total GPU hours, a non-trivial share is consumed by post-training and by the repeated checkpoint evaluations required when transitioning between major training stages. These additional costs are not captured when reporting pretraining hours alone but remain significant across the model’s full development cycle. Further pretraining details, which represent the bulk of expenditure, are provided in Appendix \ref{sec:appendixpretrain}.

\section{\olmothreebase}
\label{sec:pretraining-process}

The goal of \olmothreebase is to establish a strong foundation that supports a diversity of general capabilities while enabling downstream capabilities like thinking, tool-use, and instruction-following to be easily elicited during post-training. In this section, we describe our recipe for \olmothreebase, organized as follows:

\begin{itemize}
    \item {\bf{Modeling}}~(Section~\S\ref{sec:modeling}) \olmothreebase closely follows \olmotoo in that it is a dense model at 7B and 32B sizes, with largely identical hyperparameters. Apart from engineering improvements that enable better training throughput, we focus on enabling a larger context window. We lay out the details in Section~\S\ref{sec:modeling}.

    \item {\bf{Evaluation}}~(Section~\S\ref{sec:experimental-design}) To guard against overfitting \olmothreebase to any one capability, we greatly expand on our evaluation suite from \olmotoo to include more benchmarks.
    We make small-scale experiments more reliable by systematically refining benchmark selection and usage throughout development.

    \item {\bf{Data}}~We introduce \dolmatoo, a collection of data to support multiple stages of base model development:
    \begin{itemize}[topsep=0pt,itemsep=2pt,parsep=0pt]
        \item[$\circ$] {\bf{Pretraining}}~(Section~\S\ref{sec:pretraining})
        We train on \dolmatoomix, a mix of 5.9T tokens of diverse, natural data including sources like web pages, academic PDFs, code repositories, and more.

        \item[$\circ$] {\bf{Midtraining}}~(Section~\S\ref{sec:midtraining})
        We train on \dolminostoo, a mix of 100B tokens combining our highest-quality pretraining data with substantial task data for math and code problems, general knowledge QA, instruction following, and more.

        \item[$\circ$] {\bf{Long-context extension}}~(Section~\S\ref{sec:long-context})
        We train on \longminomix, a mix of 50B (\olmothreebase~7B) or 100B (\olmothreebase~32B) tokens combining long documents with our midtraining data.
    \end{itemize}

\end{itemize}

\subsection{Main Results for \olmothreebase}
\label{sec:pretrain_eval}
Tables~\ref{tab:transposed-super-base-table-32b}~and~\ref{tab:transposed-super-base-table-7b} compare \olmothreebase 32B and 7B with leading fully-open and open-weights base models, demonstrating both the effectiveness of our evaluation design and the strong performance of \olmothreebase across a broad set of capabilities.

\olmothreebase is the best fully-open model at 32B parameters, outperforming Stanford Marin 32B and Apertus 70B.
On Math and Code evaluation composites, it achieves double-digit improvements over the other fully-open 32B models and is within a few points of strong open-weight baselines.
On MCQA benchmarks, its STEM and Non-STEM scores closely track Marin 32B and \olmotoo 32B and sit a few points behind the top open-weight models, while on GenQA \olmothreebase forms the top fully-open cluster with Marin 32B and \olmotoo 32B and is only narrowly behind Llama 3.1 70B among the open-weight baselines.
At the 7B scale, \olmothreebase achieves the strongest Math and Code performance among fully-open models, with sizable margins over Marin 8B, Apertus 8B, and \olmotoo 7B.
Compared to open-weight models, it trails only the strongest models such as Qwen and Nemotron Nano on Math and Code.
In MCQA, \olmothreebase 7B is on par with the strongest fully-open models in both STEM and Non-STEM areas.
Finally, on GenQA tasks, \olmothreebase outperforms all but Marin among listed fully-open models, and outperforms all but the larger Gemma 2 9B and Llama3.1 8B among listed open-weight models.

\begin{table}[t]
\centering
\footnotesize
\setlength\tabcolsep{5pt}
\renewcommand{\arraystretch}{0.9}
\adjustbox{max width=\linewidth}{
{\fontsize{9}{9}\selectfont
\begin{NiceTabular}{@{}Hl
P{38pt}C{38pt}C{38pt}C{38pt}C{38pt}C{38pt}|C{38pt}C{38pt}C{38pt}C{38pt}C{38pt}C{38pt}H@{}}
\toprule
& & \multicolumn{6}{c}{\quad \quad \quad \quad \textbf{\texttt{Fully-open Models}}} & \multicolumn{6}{c}{\textbf{\texttt{Open-weight Models}}} \\
\textbf{Skill} &

& \textbf{\olmothree 32B} & \textbf{Marin 32B} & \textbf{Apertus 70B} & \textbf{Gaperon 24B} & \textbf{LLM 360 K2~V2~70B\footnotemark} & \textbf{\olmotoo 32B} & \textbf{Qwen 2.5 32B} & \textbf{Gemma 3 27B} & \textbf{Mistral 3.1 24B} & \textbf{Seed 36B} & \textbf{Gemma 2 27B} & \textbf{Llama 3.1 70B} \\
\midrule
\rowcolor{midgrey}   -- & {\textbf{\olmothreeeval \fontsize{9}{9}\selectfont~Math}}                                              & 61.9                    & 49.3               & 39.7                 & 20.7                 & 72.9                                    & 53.9                  & 64.7                  & 63.2                 & 59.5                     & 15.3              & 57.5                 & 62.0                   \\
\rowcolor{lightgrey} -- & GSM8k                                                                                                  & 80.6                    & 69.1               & 63.0                 & 33.3                 & 90.9                                    & 77.6                  & 81.1                  & 81.3                 & 79.3                     & 26.9              & 76.3                 & 81.2                   \\
\rowcolor{lightgrey} -- & GSM Symbolic                                                                                           & 61.2                    & 42.0               & 38.6                 & 14.5                 & 77.7                                    & 53.1                  & 56.2                  & 61.2                 & 59.1                     & 10.3              & 57.3                 & 64.6                   \\
\rowcolor{lightgrey} -- & MATH                                                                                                   & 43.8                    & 36.8               & 17.4                 & 14.2                 & 50.2                                    & 31.0                  & 56.7                  & 47.0                 & 40.1                     & 8.7               & 38.8                 & 40.2                   \\
\midrule
\rowcolor{midgrey}   -- & {\textbf{\olmothreeeval \fontsize{9}{9}\selectfont~Code}}                                              & 39.7                    & 30.8               & 23.3                 & 19.4                 & 38.4                                    & 20.5                  & 48.3                  & 41.6                 & 42.4                     & 54.9              & 41.0                 & 36.3                   \\
\rowcolor{lightgrey} -- & BigCodeBench                                                                                           & 43.7                    & 34.5               & 24.0                 & 17.0                 & 42.9                                    & 22.2                  & 48.1                  & 44.0                 & 46.4                     & 50.7              & 43.4                 & 43.4                   \\
\rowcolor{lightgrey} -- & HumanEval                                                                                              & 65.8                    & 52.3               & 32.5                 & 31.2                 & 61.1                                    & 29.4                  & 65.6                  & 62.1                 & 65.5                     & 71.3              & 57.5                 & 57.4                   \\
\rowcolor{lightgrey} -- & DeepSeek LeetCode                                                                                      & 2.0                     & 1.3                & 1.2                  & 0.0                  & 3.1                                     & 0.8                   & 8.0                   & 5.8                  & 0.1                      & 13.0              & 4.7                  & 0.2                    \\
\rowcolor{lightgrey} -- & DS 1000                                                                                                & 29.4                    & 26.3               & 17.8                 & 11.0                 & 28.0                                    & 20.4                  & 43.3                  & 34.3                 & 36.3                     & 44.0              & 29.7                 & 29.5                   \\
\rowcolor{lightgrey} -- & MBPP                                                                                                   & 59.6                    & 52.1               & 37.6                 & 36.7                 & 55.7                                    & 37.1                  & 69.8                  & 60.0                 & 61.9                     & 72.0              & 61.7                 & 55.5                   \\
\rowcolor{lightgrey} -- & MultiPL HumanEval                                                                                      & 36.0                    & 18.5               & 18.4                 & 13.0                 & 36.3                                    & 10.5                  & 49.7                  & 37.7                 & 39.0                     & 69.2              & 40.3                 & 32.2                   \\
\rowcolor{lightgrey} -- & MultiPL MBPPP                                                                                          & 41.5                    & 30.5               & 31.3                 & 26.5                 & 41.5                                    & 23.2                  & 53.6                  & 47.2                 & 47.7                     & 63.8              & 49.7                 & 35.9                   \\
\midrule
\rowcolor{midgrey}   -- & {$\textbf{\olmothreeeval \fontsize{9}{9}\selectfont~MC}_\textbf{\fontsize{6}{6}\selectfont~STEM}$}     & 74.5                    & 75.9               & 70.0                 & 56.2                 & 75.7                                    & 75.3                  & 82.2                  & 80.2                 & 81.5                     & 83.4              & 75.6                 & 80.1                   \\
\rowcolor{lightgrey} -- & ARC MC                                                                                                 & 94.7                    & 93.4               & 90.7                 & 72.7                 & 93.5                                    & 94.4                  & 97.0                  & 95.8                 & 96.2                     & 97.3              & 94.1                 & 95.2                   \\
\rowcolor{lightgrey} -- & MMLU STEM                                                                                              & 70.8                    & 68.4               & 57.8                 & 45.3                 & 66.5                                    & 64.7                  & 79.7                  & 74.9                 & 76.1                     & 82.8              & 65.8                 & 70.0                   \\
\rowcolor{lightgrey} -- & MedMCQA MC                                                                                             & 57.6                    & 61.8               & 55.9                 & 42.6                 & 62.5                                    & 60.2                  & 68.8                  & 64.7                 & 68.8                     & 69.6              & 61.8                 & 67.8                   \\
\rowcolor{lightgrey} -- & MedQA MC                                                                                               & 53.8                    & 60.8               & 52.4                 & 35.4                 & 61.1                                    & 62.2                  & 68.4                  & 68.7                 & 70.4                     & 70.1              & 61.0                 & 72.3                   \\
\rowcolor{lightgrey} -- & SciQ MC                                                                                                & 95.5                    & 95.1               & 93.3                 & 84.9                 & 94.8                                    & 95.1                  & 97.1                  & 96.8                 & 96.3                     & 97.1              & 95.1                 & 95.4                   \\
\midrule
\rowcolor{midgrey}   -- & {$\textbf{\olmothreeeval \fontsize{9}{9}\selectfont~MC}_\textbf{\fontsize{6}{6}\selectfont~Non-STEM}$} & 85.6                    & 84.5               & 78.5                 & 64.1                 & 84.0                                    & 84.2                  & 89.3                  & 86.7                 & 87.9                     & 89.0              & 83.2                 & 86.1                   \\
\rowcolor{lightgrey} -- & MMLU Humanities                                                                                        & 78.3                    & 78.9               & 74.1                 & 56.7                 & 78.4                                    & 79.7                  & 85.0                  & 80.5                 & 82.7                     & 85.7              & 79.3                 & 83.4                   \\
\rowcolor{lightgrey} -- & MMLU Social Sci.                                                                                       & 84.0                    & 83.7               & 79.2                 & 58.9                 & 84.1                                    & 84.5                  & 88.4                  & 86.2                 & 88.6                     & 90.1              & 85.8                 & 87.4                   \\
\rowcolor{lightgrey} -- & MMLU Other                                                                                             & 75.1                    & 75.4               & 70.1                 & 55.4                 & 77.1                                    & 75.6                  & 81.2                  & 80.2                 & 81.9                     & 82.4              & 76.9                 & 79.4                   \\
\rowcolor{lightgrey} -- & CSQA MC                                                                                                & 82.3                    & 80.1               & 76.9                 & 60.6                 & 80.2                                    & 81.2                  & 89.9                  & 79.0                 & 80.5                     & 81.1              & 78.1                 & 79.0                   \\
\rowcolor{lightgrey} -- & PiQA MC                                                                                                & 85.6                    & 90.5               & 79.0                 & 72.0                 & 87.5                                    & 87.7                  & 93.3                  & 90.3                 & 91.0                     & 92.5              & 89.0                 & 91.5                   \\
\rowcolor{lightgrey} -- & SocialIQA MC                                                                                           & 83.9                    & 82.4               & 79.3                 & 71.3                 & 83.0                                    & 82.3                  & 86.6                  & 81.2                 & 81.0                     & 84.9              & 81.0                 & 83.5                   \\
\rowcolor{lightgrey} -- & CoQA Gen2MC MC                                                                                         & 96.4                    & 93.9               & 87.5                 & 67.3                 & 92.2                                    & 94.4                  & 96.8                  & 95.8                 & 94.9                     & 96.9              & 94.3                 & 95.1                   \\
\rowcolor{lightgrey} -- & DROP Gen2MC MC                                                                                         & 87.2                    & 71.0               & 56.5                 & 48.0                 & 67.6                                    & 68.6                  & 86.6                  & 84.6                 & 86.5                     & 90.1              & 66.6                 & 70.3                   \\
\rowcolor{lightgrey} -- & Jeopardy Gen2MC MC                                                                                     & 92.3                    & 95.3               & 93.2                 & 77.0                 & 95.6                                    & 96.6                  & 97.0                  & 95.9                 & 97.2                     & 96.2              & 92.0                 & 97.1                   \\
\rowcolor{lightgrey} -- & NaturalQs Gen2MC MC                                                                                    & 78.0                    & 81.0               & 71.9                 & 47.5                 & 80.5                                    & 78.6                  & 79.9                  & 82.0                 & 84.6                     & 81.4              & 74.5                 & 82.4                   \\
\rowcolor{lightgrey} -- & SQuAD Gen2MC MC                                                                                        & 98.2                    & 97.6               & 95.7                 & 90.0                 & 97.4                                    & 97.4                  & 97.9                  & 97.7                 & 97.9                     & 98.1              & 97.5                 & 97.7                   \\
\midrule
\rowcolor{midgrey}   -- & {\textbf{\olmothreeeval \fontsize{9}{9}\selectfont~GenQA}}                                             & 79.8                    & 80.3               & 75.0                 & 65.3                 & 75.6                                    & 79.1                  & 68.5                  & 73.5                 & 78.0                     & 76.0              & 72.9                 & 81.6                   \\
\rowcolor{lightgrey} -- & HellaSwag RC                                                                                           & 84.8                    & 87.2               & 84.5                 & 75.2                 & 86.3                                    & 87.5                  & 86.3                  & 86.0                 & 86.2                     & 84.8              & 86.7                 & 88.4                   \\
\rowcolor{lightgrey} -- & Winogrande RC                                                                                          & 90.3                    & 90.5               & 87.7                 & 80.3                 & 89.5                                    & 89.4                  & 87.5                  & 91.3                 & 90.8                     & 89.3              & 90.8                 & 91.7                   \\
\rowcolor{lightgrey} -- & Lambada                                                                                                & 75.7                    & 76.7               & 74.8                 & 58.3                 & 75.3                                    & 77.0                  & 76.2                  & 77.5                 & 79.3                     & 76.1              & 76.9                 & 79.6                   \\
\rowcolor{lightgrey} -- & Basic Skills                                                                                           & 93.5                    & 91.1               & 87.5                 & 83.2                 & 91.5                                    & 88.7                  & 94.2                  & 94.9                 & 91.9                     & 96.0              & 93.2                 & 92.4                   \\
\rowcolor{lightgrey} -- & DROP                                                                                                   & 80.9                    & 76.5               & 56.3                 & 59.4                 & 75.0                                    & 76.3                  & 53.7                  & 75.9                 & 74.9                     & 76.1              & 73.2                 & 78.3                   \\
\rowcolor{lightgrey} -- & Jeopardy                                                                                               & 75.3                    & 80.5               & 77.2                 & 58.9                 & 77.6                                    & 79.1                  & 74.0                  & 82.1                 & 80.3                     & 77.4              & 80.7                 & 84.0                   \\
\rowcolor{lightgrey} -- & NaturalQs                                                                                              & 49.0                    & 55.1               & 43.1                 & 33.5                 & 45.7                                    & 51.4                  & 39.3                  & 49.2                 & 45.1                     & 30.7              & 47.1                 & 53.1                   \\
\rowcolor{lightgrey} -- & SQuAD                                                                                                  & 94.5                    & 94.4               & 90.7                 & 89.3                 & 93.9                                    & 94.0                  & 64.9                  & 92.4                 & 92.6                     & 89.1              & 93.0                 & 92.9                   \\
\rowcolor{lightgrey} -- & CoQA                                                                                                   & 74.1                    & 70.7               & 72.8                 & 49.8                 & 45.6                                    & 68.7                  & 40.4                  & 12.4                 & 61.1                     & 64.4              & 14.9                 & 73.9                   \\
\midrule
\rowcolor{midgrey}   -- & {\textbf{\olmothreeeval \fontsize{9}{9}\selectfont~HeldOut}}                                           &                         &                    &                      &                      &                                         &                       &                       &                      &                          &                   &                      &                        \\
\rowcolor{lightgrey} -- & LBPP                                                                                                   & 21.8                    & 17.3               & 8.1                  & 4.3                  & 19.9                                    & 8.2                   & 40.3                  & 17.7                 & 30.3                     & 42.6              & 19.7                 & 11.8                   \\
\rowcolor{lightgrey} -- & BBH                                                                                                    & 77.6                    & 70.1               & 58.8                 & 36.6                 & 82.6                                    & 64.6                  & 81.1                  & 77.4                 & 81.4                     & 85.0              & 74.8                 & 80.8                   \\
\rowcolor{lightgrey} -- & MMLU Pro MC                                                                                            & 49.7                    & 48.1               & 39.6                 & 21.3                 & 50.1                                    & 46.9                  & 61.1                  & 53.1                 & 58.9                     & 62.2              & 47.6                 & 50.4                   \\
\rowcolor{lightgrey} -- & Deepmind Math                                                                                          & 29.6                    & 26.7               & 20.1                 & 28.3                 & 29.8                                    & 22.0                  & 40.7                  & 30.4                 & 35.3                     & 31.3              & 27.6                 & 40.3                   \\

\bottomrule
\end{NiceTabular}}
}
\caption{
\textbf{Results comparing \olmothreebase 32B  to other base models using the \olmothreeeval Main suite} (details in Section~\S\ref{sec:experimental-design}).
\olmothree was not evaluated on held-out benchmarks prior to release.
}
\label{tab:transposed-super-base-table-32b}
\end{table}

\begin{table}[t]
\centering
\footnotesize
\setlength\tabcolsep{5pt}
\renewcommand{\arraystretch}{1}
\adjustbox{max width=\linewidth}{
{\fontsize{9}{9}\selectfont
\begin{NiceTabular}{@{}Hl
P{35pt}C{35pt}C{35pt}C{35pt}C{35pt}|C{35pt}C{35pt}C{35pt}C{35pt}C{35pt}C{35pt}C{35pt}C{35pt}C{35pt}@{}}
\toprule
& & \multicolumn{5}{c}{\quad \quad \quad \quad \textbf{\texttt{Fully-open Models}}} & \multicolumn{8}{c}{\textbf{\texttt{Open-weight Models}}} \\
\textbf{Skill} &

& \textbf{\olmothree 7B} & \textbf{Marin 8B} & \textbf{Apertus 8B} & \textbf{Gaperon 8B} & \textbf{\olmotoo 7B} & \textbf{Qwen3 8B} & \textbf{Nemo. Nano 9B} & \textbf{Gemma 2 9B} & \textbf{Qwen 2.5 7B} & \textbf{Llama 3.1 8B} & \textbf{Granite 3.3 8B} & \textbf{MiMo 7B} \\
\midrule
\rowcolor{midgrey}   -- & {\textbf{\olmothreeeval \fontsize{9}{9}\selectfont~Math}}                                              & 54.7                   & 39.6              & 29.2                & 16.9                & 41.7                 & 67.2              & 49.8                   & 48.8                & 60.7                 & 36.9                  & 41.5                    & 54.3             \\
\rowcolor{lightgrey} -- & GSM8k                                                                                                  & 75.5                   & 60.9              & 48.2                & 30.0                & 67.1                 & 84.5              & 82.3                   & 68.5                & 79.9                 & 56.4                  & 61.0                    & 74.3             \\
\rowcolor{lightgrey} -- & GSM Symbolic                                                                                           & 48.6                   & 33.6              & 26.3                & 12.5                & 38.8                 & 65.4              & 62.7                   & 45.1                & 56.2                 & 35.1                  & 35.5                    & 53.3             \\
\rowcolor{lightgrey} -- & MATH                                                                                                   & 40.0                   & 24.3              & 13.1                & 8.2                 & 19.1                 & 51.6              & 4.5                    & 32.9                & 45.9                 & 19.2                  & 27.9                    & 35.2             \\
\midrule
\rowcolor{midgrey}   -- & {\textbf{\olmothreeeval \fontsize{9}{9}\selectfont~Code}}                                              & 30.7                   & 21.4              & 19.0                & 16.1                & 10.4                 & 46.1              & 43.1                   & 30.2                & 41.0                 & 21.2                  & 18.0                    & 35.7             \\
\rowcolor{lightgrey} -- & BigCodeBench                                                                                           & 34.1                   & 21.5              & 20.9                & 13.0                & 8.8                  & 42.5              & 43.2                   & 30.9                & 39.7                 & 30.7                  & 0.4                     & 38.3             \\
\rowcolor{lightgrey} -- & HumanEval                                                                                              & 49.1                   & 31.6              & 21.6                & 24.5                & 16.3                 & 71.7              & 71.7                   & 40.0                & 66.1                 & 40.4                  & 0.0                     & 57.0             \\
\rowcolor{lightgrey} -- & DeepSeek LeetCode                                                                                      & 1.4                    & 0.5               & 0.6                 & 0.0                 & 0.2                  & 8.3               & 6.8                    & 1.9                 & 5.1                  & 0.1                   & 0.0                     & 1.2              \\
\rowcolor{lightgrey} -- & DS 1000                                                                                                & 20.2                   & 16.5              & 11.8                & 9.1                 & 10.1                 & 33.1              & 30.3                   & 23.4                & 35.2                 & 22.2                  & 22.6                    & 28.1             \\
\rowcolor{lightgrey} -- & MBPP                                                                                                   & 43.6                   & 36.5              & 33.5                & 29.3                & 21.2                 & 66.2              & 62.3                   & 49.1                & 55.4                 & 12.1                  & 48.5                    & 48.3             \\
\rowcolor{lightgrey} -- & MultiPL HumanEval                                                                                      & 28.7                   & 15.6              & 15.5                & 12.1                & 4.2                  & 52.3              & 40.0                   & 27.9                & 40.3                 & 14.5                  & 22.3                    & 34.5             \\
\rowcolor{lightgrey} -- & MultiPL MBPPP                                                                                          & 38.2                   & 27.6              & 29.2                & 24.6                & 12.2                 & 48.4              & 47.5                   & 38.2                & 45.4                 & 28.3                  & 32.3                    & 42.5             \\
\midrule
\rowcolor{midgrey}   -- & {$\textbf{\olmothreeeval \fontsize{9}{9}\selectfont~MC}_\textbf{\fontsize{6}{6}\selectfont~STEM}$}     & 66.4                   & 68.1              & 66.3                & 58.0                & 64.6                 & 78.8              & 73.5                   & 72.8                & 74.7                 & 69.0                  & 65.0                    & 71.6             \\
\rowcolor{lightgrey} -- & ARC MC                                                                                                 & 89.2                   & 89.2              & 87.9                & 77.2                & 85.7                 & 95.4              & 94.1                   & 92.7                & 93.4                 & 86.4                  & 86.2                    & 91.7             \\
\rowcolor{lightgrey} -- & MMLU STEM                                                                                              & 59.7                   & 58.1              & 52.4                & 43.1                & 53.2                 & 76.7              & 71.1                   & 62.8                & 67.6                 & 55.7                  & 55.6                    & 63.5             \\
\rowcolor{lightgrey} -- & MedMCQA MC                                                                                             & 48.3                   & 52.7              & 51.7                & 44.5                & 49.2                 & 63.5              & 54.5                   & 58.9                & 60.3                 & 56.5                  & 49.6                    & 56.2             \\
\rowcolor{lightgrey} -- & MedQA MC                                                                                               & 41.8                   & 47.3              & 47.6                & 36.8                & 43.8                 & 62.1              & 53.5                   & 55.4                & 56.6                 & 53.7                  & 43.0                    & 53.0             \\
\rowcolor{lightgrey} -- & SciQ MC                                                                                                & 92.8                   & 93.2              & 91.9                & 88.4                & 90.9                 & 96.1              & 94.3                   & 94.4                & 95.4                 & 92.7                  & 90.8                    & 93.5             \\
\midrule
\rowcolor{midgrey}   -- & {$\textbf{\olmothreeeval \fontsize{9}{9}\selectfont~MC}_\textbf{\fontsize{6}{6}\selectfont~Non-STEM}$} & 78.2                   & 78.8              & 74.2                & 65.0                & 75.2                 & 84.8              & 81.3                   & 81.3                & 82.9                 & 76.1                  & 76.9                    & 80.5             \\
\rowcolor{lightgrey} -- & MMLU Humanities                                                                                        & 68.9                   & 71.4              & 67.8                & 59.5                & 67.9                 & 78.6              & 78.0                   & 74.5                & 76.2                 & 70.1                  & 67.6                    & 73.6             \\
\rowcolor{lightgrey} -- & MMLU Social Sci.                                                                                       & 75.0                   & 77.4              & 74.7                & 60.8                & 73.1                 & 84.8              & 82.2                   & 82.9                & 83.0                 & 75.5                  & 71.8                    & 80.8             \\
\rowcolor{lightgrey} -- & MMLU Other                                                                                             & 66.9                   & 68.3              & 66.1                & 57.2                & 65.2                 & 76.8              & 73.8                   & 74.2                & 74.4                 & 69.1                  & 64.5                    & 72.7             \\
\rowcolor{lightgrey} -- & CSQA MC                                                                                                & 75.3                   & 75.3              & 72.1                & 65.5                & 72.0                 & 84.1              & 74.4                   & 75.3                & 85.0                 & 72.9                  & 82.3                    & 76.1             \\
\rowcolor{lightgrey} -- & PiQA MC                                                                                                & 80.2                   & 85.7              & 80.5                & 71.6                & 80.1                 & 89.9              & 86.0                   & 85.7                & 88.5                 & 78.3                  & 81.5                    & 87.2             \\
\rowcolor{lightgrey} -- & SocialIQA MC                                                                                           & 80.3                   & 79.8              & 76.3                & 73.4                & 77.5                 & 83.3              & 78.7                   & 80.3                & 82.9                 & 77.0                  & 83.1                    & 80.7             \\
\rowcolor{lightgrey} -- & CoQA Gen2MC MC                                                                                         & 92.5                   & 86.2              & 82.8                & 59.7                & 85.0                 & 93.7              & 92.2                   & 92.7                & 93.5                 & 89.9                  & 87.6                    & 91.4             \\
\rowcolor{lightgrey} -- & DROP Gen2MC MC                                                                                         & 67.3                   & 63.7              & 47.5                & 44.8                & 55.6                 & 78.3              & 70.0                   & 65.8                & 69.1                 & 53.3                  & 55.0                    & 64.1             \\
\rowcolor{lightgrey} -- & Jeopardy Gen2MC MC                                                                                     & 86.9                   & 90.8              & 90.3                & 83.2                & 89.5                 & 92.3              & 90.7                   & 92.8                & 92.1                 & 88.9                  & 88.4                    & 89.5             \\
\rowcolor{lightgrey} -- & NaturalQs Gen2MC MC                                                                                    & 69.4                   & 71.5              & 66.7                & 51.3                & 66.3                 & 74.1              & 71.1                   & 72.5                & 70.5                 & 68.0                  & 69.2                    & 72.2             \\
\rowcolor{lightgrey} -- & SQuAD Gen2MC MC                                                                                        & 96.9                   & 96.5              & 91.3                & 87.7                & 95.3                 & 97.5              & 97.4                   & 97.3                & 96.4                 & 94.4                  & 94.5                    & 96.7             \\
\midrule
\rowcolor{midgrey}   -- & {\textbf{\olmothreeeval \fontsize{9}{9}\selectfont~GenQA}}                                             & 72.5                   & 75.9              & 69.0                & 63.3                & 72.4                 & 71.1              & 71.8                   & 75.6                & 67.5                 & 73.1                  & 67.8                    & 71.4             \\
\rowcolor{lightgrey} -- & HellaSwag RC                                                                                           & 77.7                   & 84.0              & 81.0                & 73.9                & 82.2                 & 80.5              & 80.2                   & 81.8                & 81.0                 & 81.5                  & 83.7                    & 80.6             \\
\rowcolor{lightgrey} -- & Winogrande RC                                                                                          & 85.7                   & 88.6              & 85.8                & 76.4                & 87.4                 & 86.4              & 86.2                   & 88.8                & 86.0                 & 87.3                  & 89.4                    & 86.5             \\
\rowcolor{lightgrey} -- & Lambada                                                                                                & 68.9                   & 73.9              & 70.9                & 67.0                & 70.5                 & 73.0              & 67.9                   & 76.3                & 70.3                 & 75.5                  & 76.0                    & 73.1             \\
\rowcolor{lightgrey} -- & Basic Skills                                                                                           & 89.5                   & 85.6              & 83.8                & 80.5                & 82.2                 & 93.5              & 91.4                   & 89.3                & 91.4                 & 88.0                  & 88.7                    & 89.7             \\
\rowcolor{lightgrey} -- & DROP                                                                                                   & 71.5                   & 73.0              & 37.1                & 54.9                & 61.5                 & 57.2              & 71.4                   & 68.2                & 56.7                 & 59.5                  & 38.4                    & 69.3             \\
\rowcolor{lightgrey} -- & Jeopardy                                                                                               & 60.4                   & 72.7              & 70.1                & 55.5                & 70.8                 & 65.1              & 64.9                   & 75.1                & 63.0                 & 70.9                  & 69.7                    & 65.6             \\
\rowcolor{lightgrey} -- & NaturalQs                                                                                              & 32.6                   & 42.6              & 35.0                & 28.8                & 37.4                 & 33.8              & 31.2                   & 40.4                & 31.2                 & 36.7                  & 37.0                    & 33.1             \\
\rowcolor{lightgrey} -- & SQuAD                                                                                                  & 93.5                   & 93.4              & 89.6                & 86.0                & 91.5                 & 89.2              & 92.3                   & 88.8                & 87.0                 & 89.2                  & 89.6                    & 90.3             \\
\rowcolor{lightgrey} -- & CoQA                                                                                                   & 72.8                   & 69.5              & 67.4                & 46.7                & 68.3                 & 61.6              & 60.4                   & 71.5                & 40.5                 & 69.0                  & 37.8                    & 54.4             \\
\midrule
\rowcolor{midgrey}   -- & {\textbf{\olmothreeeval \fontsize{9}{9}\selectfont~HeldOut}}                                           &                        &                   &                     &                     &                      &                   &                        &                     &                      &                       &                         &                  \\
\rowcolor{lightgrey} -- & LBPP                                                                                                   & 17.1                   & 5.8               & 7.1                 & 4.7                 & 3.1                  & 25.7              & 31.7                   & 12.4                & 22.1                 & 9.1                   & 18.5                    & 21.5             \\
\rowcolor{lightgrey} -- & BBH                                                                                                    & 63.5                   & 55.6              & 48.1                & 38.4                & 49.6                 & 76.5              & 77.0                   & 68.8                & 54.7                 & 63.0                  & 61.5                    & 75.1             \\
\rowcolor{lightgrey} -- & MMLU Pro MC                                                                                            & 37.3                   & 38.8              & 33.9                & 20.8                & 33.1                 & 50.3              & 50.2                   & 44.7                & 48.1                 & 37.4                  & 33.9                    & 44.3             \\
\rowcolor{lightgrey} -- & Deepmind Math                                                                                          & 23.7                   & 20.2              & 17.1                & 34.1                & 16.2                 & 47.7              & 31.4                   & 23.0                & 32.8                 & 24.1                  & 32.2                    & 25.4             \\

\bottomrule
\end{NiceTabular}}
}
\caption{
\textbf{Results comparing \olmothreebase 7B  to other base models using the \olmothreeeval Main suite} (details in \S\ref{sec:experimental-design}).
\olmothree was not evaluated on held-out benchmarks prior to release.
}
\label{tab:transposed-super-base-table-7b}
\end{table}

\subsection{Modeling and Architecture} \label{sec:modeling}

\olmothree modeling and training largely follows that of \olmotoo. We focus this section on the key differences and refer to the appendix for further details.

\paragraph{Architecture} \label{sec:architecture}
We adopt a decoder-only transformer architecture based on \citet{vaswani2017attention}. Details of the architecture are presented in Table~\ref{tab:combined_model_specs} in Appendix~\ref{sec:appendixpretrain}. Compared to \olmotoo:

\begin{itemize}
    \item We train with a context window of 8192 tokens (increased from 4096 tokens for \olmotoo) during pretraining and midtraining stages.
    \item To support scalable pretraining at longer sequence lengths, and to keep inference costs manageable, we introduce a sliding window attention (SWA) pattern~\citep{beltagy2020longformer} in which each token can attend to previous tokens in a window of size 4096. We add SWA at three out of every four layers, and ensure that the last layer always uses full attention.
\end{itemize}

\begin{figure}[t!] %
  \centering
  
  \includegraphics[width=0.85\linewidth]{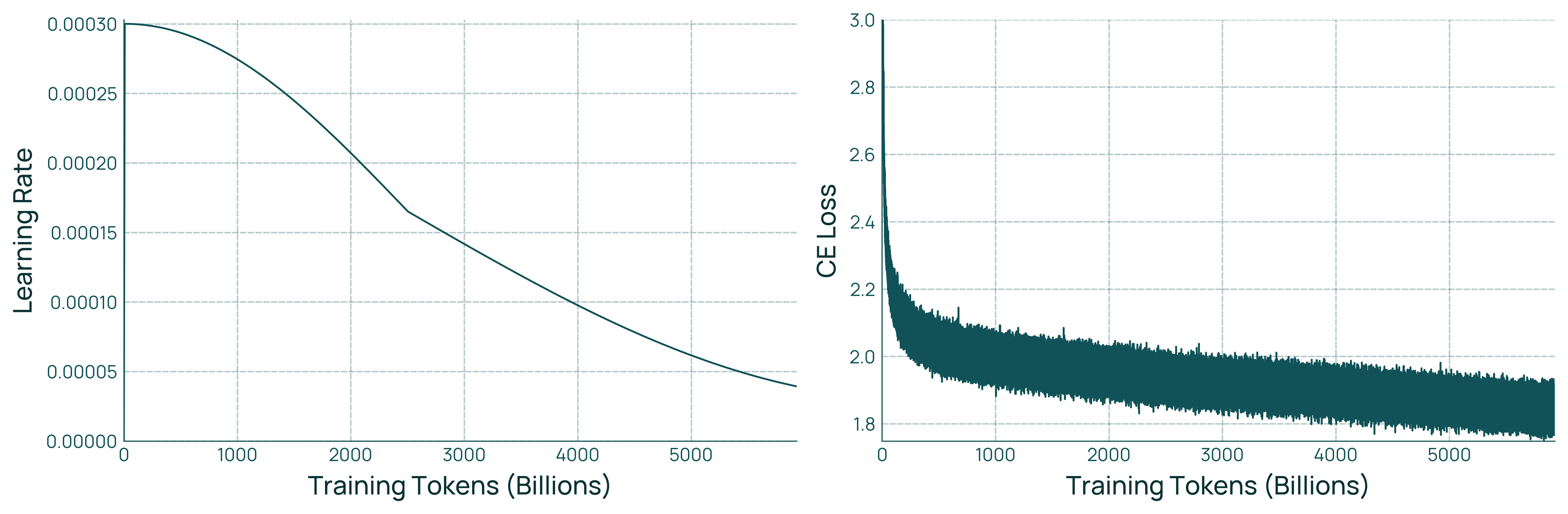}
  \caption{\textbf{Learning rate schedule and loss for \olmothreebase 7B}.
      The first half of the learning rate schedule is a cosine schedule over 5T tokens.
      We stretch the second half of the schedule to reach a target length of one epoch (5.93T tokens). Warm-up is 2000 steps, the peak learning rate is $3 \times 10^{-4}$, and the final learning rate is 10\% of the peak LR.}
  \label{fig:pretrain:7b-lr-and-loss}
  
  \vspace{1em} %
  
  \includegraphics[width=0.85\linewidth]{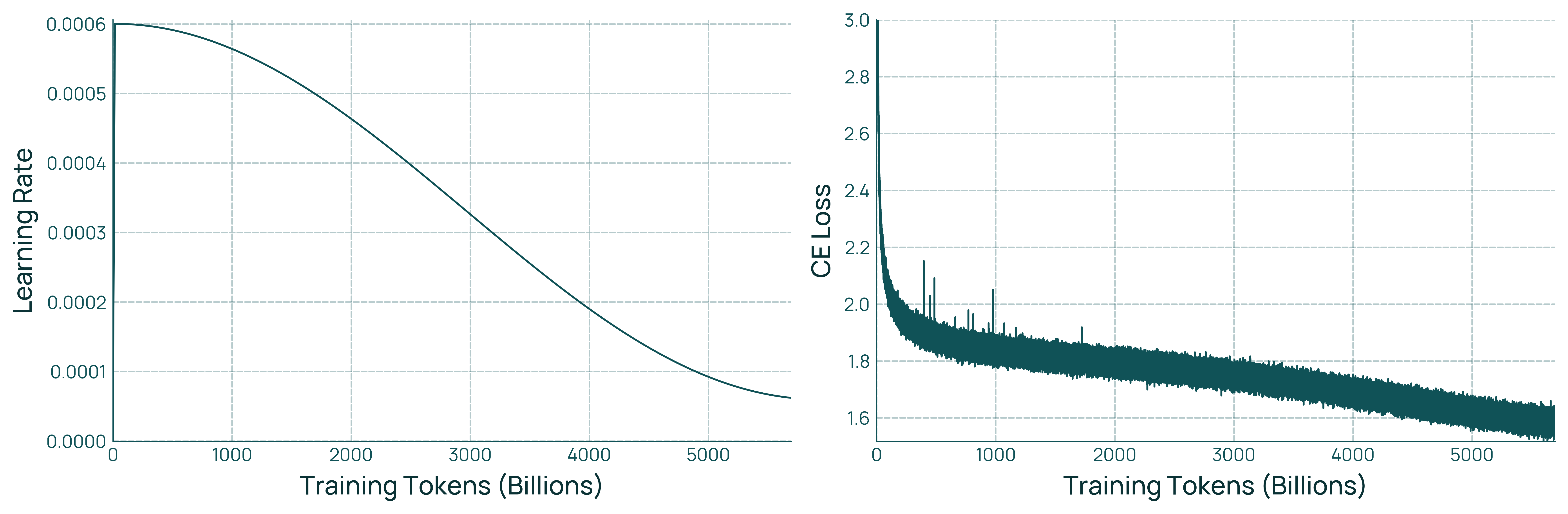}
  \caption{\textbf{Learning rate schedule and loss for \olmothreebase 32B}.
      The learning rate schedule is a cosine schedule over one epoch (5.93T tokens), truncated
      at 5.5T tokens. Warm-up is 2000 steps, and the peak learning rate is $6 \times 10^{-4}$. The schedule targets a final learning rate of 10\% of the peak. Due to the truncation, the real final learning rate is $6.210 \times 10^{-5}$. Unintuitively, the learning rate for the 32B is higher than for the 7B, but this is somewhat compensated for by the larger batch size of the 32B (8M tokens vs. 4M tokens per batch).}
  \label{fig:pretrain:32b-lr-and-loss}
  
\end{figure}

\footnotetext{For the K2 V2 results here, we use an updated pretraining checkpoint uploaded on Jan 22, 2026, released after Olmo 3.}

\paragraph{Training}
\olmothreebase is trained using the \olmocore\footnote{Further details and code: \href{https://github.com/allenai/OLMo-core}{\path{github.com/allenai/OLMo-core}}} codebase.
With this stack, we train the 7B model at 7700 tokens per second per GPU and the 32B model at 1960 tokens per second per GPU at a sequence length of 8192, using \texttt{bfloat16} precision throughout. This corresponds to roughly 43\% and 41\% MFU, respectively. We achieve this performance by combining PyTorch's built-in \texttt{torch.compile()}, custom kernels for operations such as attention~\citep{dao2023flashattention2} and the language modeling head~\citep{hsu2025ligerkernelefficienttriton}, asynchronous and batched gathering of metrics, and asynchronous checkpoint writing, among other optimizations.

\olmocore supports pretraining, midtraining, long-context extension, and SFT, along with auxiliary tools for checkpoint conversion to and from Hugging Face Transformers format and for merging model checkpoints. Support for DPO and RL is planned but not yet complete.

Hyperparameters for training \olmothreebase 7B and 32B are presented in Table~\ref{tab:training_stages_7b_32b} in Appendix~\ref{sec:appendixpretrain}. As in \olmotoo, we train in stages defined by the data curriculum and learning rate schedule (see Appendix Table~\ref{tab:training_stages_7b_32b} for details).
Infrastructure and distributed training configurations for each stage are summarized in Appendix Table~\ref{tab:training-config}.

\paragraph{Tokenizer} We process data for each stage using the same tokenizer as \olmotoo, which is derived from OpenAI's \texttt{cl100k}~\citep{gpt35,gpt4}.

\subsection{Experimental Design and Evaluation}\label{sec:experimental-design}

Model development requires many iterative data and training decisions.
However, benchmarks are not perfect decision-making tools: different evaluations are only sensitive for making development decisions across specific ranges of scale and capability \citep{magnusson2025datadecidepredictbestpretraining}.
Models trained at small compute scales are known to exhibit random-chance performance on math, code, and multiple-choice question answering (MCQA) tasks~\citep{wei2022emergent,olmes}, and benchmark noise can reduce the ability to trust small differences in scores \citep{heineman2025signalnoiseframeworkreducing}.
To address these problems, we develop {\bf \olmothreeeval}, a collection of benchmark suites to support decision-making during base model development. \olmothreeeval features the following improvements:
\begin{itemize}
    \item We aggregate scores over {\bf task clusters} that group benchmarks by assessed capability (Section~\S\ref{sec:eval:clustering}),
    \item We develop {\bf proxy metrics} for evaluating small-scale models by identifying when capabilities ``emerge'' during training (Section~\S\ref{sec:eval:scaling-analysis}), and
    \item We improve the overall {\bf signal-to-noise ratio} by evaluating more examples from noisy tasks or even removing them entirely (Section~\S\ref{sec:eval:signal-noise}).
\end{itemize}

We start by targeting a high \textit{coverage} of capabilities; we select benchmarks to prioritize science knowledge, medical/lab knowledge, math, and code tasks.
Because our data interventions are targeted to a core capability rather than a specific benchmark (e.g., ``Code'' rather than ``DS-1000''), we group tasks into \textit{clusters}, where we expect the benchmarks within a cluster to behave similarly to particular data changes.
To handle evaluation of models trained using small compute budgets (e.g., up to our largest experiment scale of 1B parameters at 100B tokens), we perform a \textit{scaling analysis} to determine which benchmarks show signal at a small scale and find proxy metrics which we use to make decisions. Finally, we analyze the \textit{signal-to-noise ratio} of each benchmark---we select benchmark metrics to improve SNR, remove benchmarks that were too noisy for making decisions, and move benchmarks out of the average if the noise of one particular benchmark dominated the aggregate scores.

\begin{figure}[t]
    \centering
    \includegraphics[width=\textwidth]{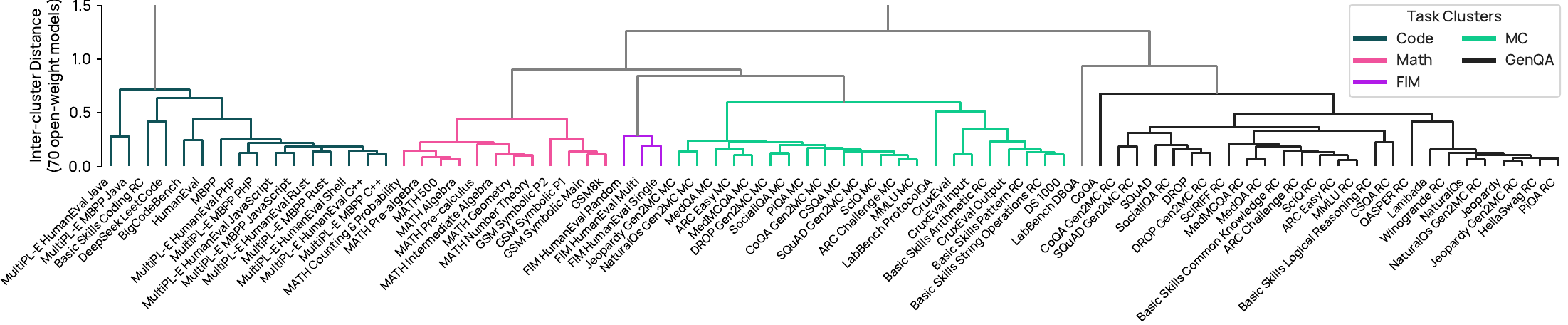}
    \caption{
   \textbf{Task clustering for \olmothreeeval}. Using a set of 23K benchmark results, the clustering method iteratively merges tasks which rank models similarly, until arriving at a stop condition. To arrive at \olmothreeeval, we move tasks in the same format into the same cluster and split MC into STEM and Non-STEM tasks.
    }
    \label{fig:clustering_dendrogram}
\end{figure}

\subsubsection{Clustering Tasks} \label{sec:eval:clustering}
To handle the large number of tasks, we cluster similar tasks into macro-averages. We aim for task clusters to match the granularity at which we perform data interventions, and for tasks within each cluster to behave similarly. Our clustering procedure requires a process to determine the similarity of two evaluations---we do this by collecting a pool of 23K benchmark scores from 70 external, open-weight models.

Using our dataset of evaluation results, we assume that two benchmarks evaluate similar constructs if they rank models similarly. We perform hierarchical clustering using Ward's variance-minimization \citep{ward1963hierarchical}, which iteratively merges evaluation scores to minimize the variance of scores between benchmarks within a cluster.
Figure \ref{fig:clustering_dendrogram} shows the result of the clustering procedure, where we manually select a threshold to balance the amount and granularity of clusters.
Importantly, we do not use the exact result of the clustering procedure---we manually move a few tasks to ensure the format of the task is the same within each cluster (e.g., tasks requiring code execution all occur in the same cluster). The resulting task clusters are: {\bf{$\text{{MC}}_\text{{STEM}}$}}, {\bf{$\text{{MC}}_\textsc{{Non-STEM}}$}}, {\bf{GenQA}}, {\bf{Math}}, {\bf{Code}}, and {\bf{Code FIM}}. %

\subsubsection{Scaling analysis}\label{sec:eval:scaling-analysis}

We evaluate open-weight models across compute scales from $10^{18}$ to $10^{25}$ training FLOPs to determine the compute scale at which particular metrics and tasks are useful for development decisions.
On some evaluation benchmarks, it is too difficult to see signal when training models at small scales \citep{wei2022emergent}, and other benchmarks `saturate' near the labeling error of the benchmark \citep{vendrow2025large}.
However, while many tasks appear emergent, continuous proxy metrics have been shown to be a better decision-making tool for model performance before we exit the noise floor \citep{schaeffer2023emergent,huang2024compression,magnusson2025datadecidepredictbestpretraining}.
We propose a Base Easy task suite which measures bits-per-byte (BPB) over tasks from the Base Main suite that have gold labels or human-written answers, calculated as the negative log-likelihood of the answer divided by the number of UTF-8 bytes in the answer string, as described in \citet{gao2020pile}.

\begin{figure}[t]
\centering
\includegraphics[width=\linewidth]{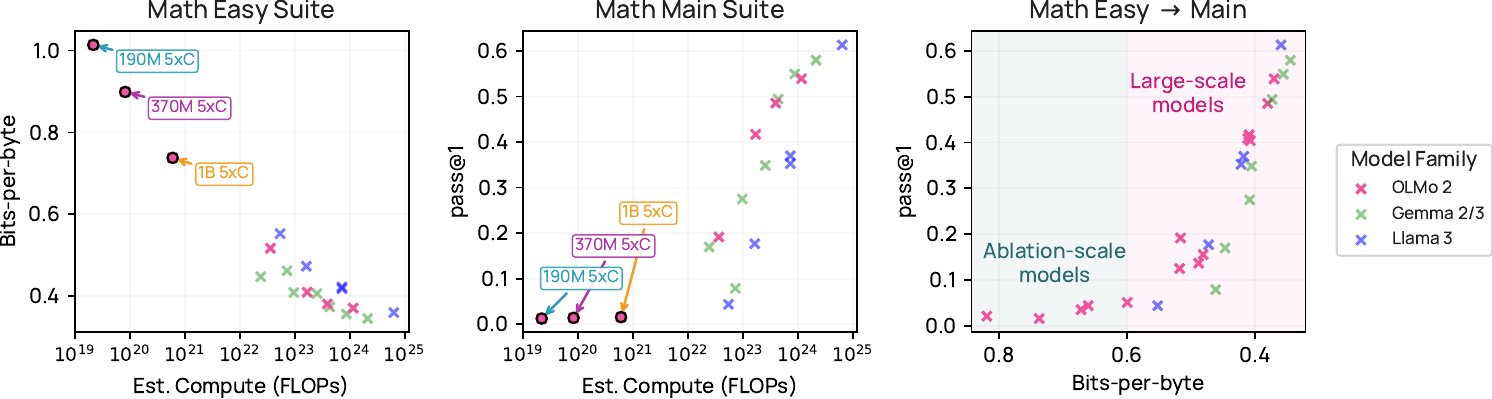}
\caption{
\textbf{Scaling analysis on the \olmothreeeval Math suite}. We use the \olmotoo scaling models \citep{bhagia2024establishingtaskscalinglaws} to find benchmarks and metrics that show signal for small-scale models (left and center). Then, we use the small-scale \olmothreeeval Easy suite as a proxy-metric for making data decisions.
}
\label{fig:scaling-analysis-math}
\end{figure}

We evaluate on the suite of 25 \olmotoo scaling law models from \citet{bhagia2024establishingtaskscalinglaws} to understand the scaling behavior in the low-compute regime, and 70 open-weight models to understand scaling behavior in the high-compute regime.
Figure~\ref{fig:scaling-analysis-math} shows the scaling behavior for our resulting Base Main benchmarks. For each task family, the Base Easy task suite shows signal at the small data ablation scale, and the Base Main task suites were not saturated at the large scale, leaving headroom for data experiments in midtraining.

\subsubsection{Signal-to-Noise Analysis}\label{sec:eval:signal-noise}
When reporting a macro-average, we aim to exclude tasks from each cluster that were too noisy to be helpful for development. We calculate the signal-to-noise ratio of each benchmark following the method from \citet{heineman2025signalnoiseframeworkreducing}, where we evaluate the final 50 checkpoints of \olmotoo 13B training, and 10 external base models trained at roughly the same compute scale ($4\cdot10^{23}$ FLOPs). From our findings, we transition from using 1K instance subsets to full evaluation sets when available. We remove some benchmarks from our evaluation suite entirely, particularly binary benchmarks such as BoolQ \citep{clark-etal-2019-boolq}, as we found that models usually oscillate between predicting the majority and minority class.

\begin{figure}[t]
\centering
\includegraphics[width=\linewidth]{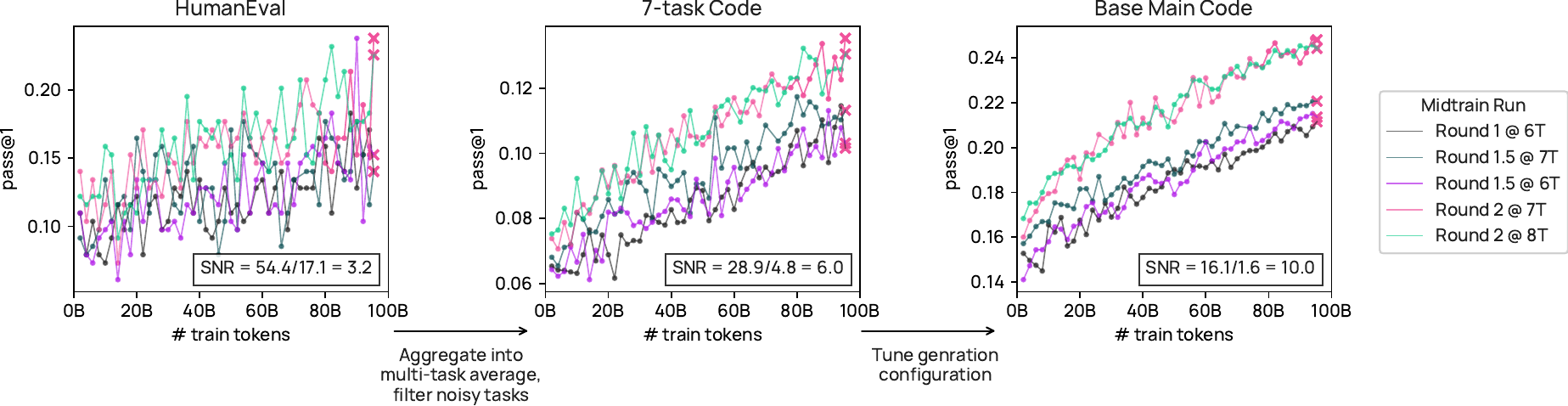}
\caption{
\textbf{\olmothreeeval{} signal-to-noise analysis on the code multi-task average using intermediate checkpoints from midtraining}. First, we aggregate into multi-task averages and remove tasks with high noise, such as CruxEval (left $\rightarrow$ center). Then, we tune generation hyperparameters to improve SNR, e.g., by increasing the $n$ in pass@k (center $\rightarrow$ right).
}
\label{fig:signal-and-noise-code}
\end{figure}

We repeat the same analysis for midtraining, instead using intermediate checkpoints from 5 preliminary pretraining runs. One important finding was to separate some benchmarks from the macro-average, like CruxEval \citep{gu2024cruxeval}, which measures a relevant and unique capability (code input/output prediction) but would introduce too much noise into the macro-average. We show an example of the SNR of three individual benchmarks compared to the base main task averages across intermediate checkpoints during midtraining in Figure~\ref{fig:signal-and-noise-code}.

\subsubsection{\olmothreeeval} \label{sec:eval:olmo-3-eval}

The resulting \olmothreeeval consists of a {\bf{Base Easy}} suite for making development decisions using small compute budgets (e.g., less than 1B parameters) and a {\bf{Base Main}} suite for development decisions for the final pretraining run and midtraining.
We provide detail on the {\bf{Chat}} suite later in \S\ref{sec:posttrain_eval}.
\olmothreeeval contains 43 tasks, which is over 4 times more benchmarks than \olmotoo---including tracking math and code benchmarks in pretraining. To prevent overfitting on the development suite, we include a {\bf{Held-out}} set of 4 benchmarks---MMLU Pro, DeepMind Math, LBPP, and BBH---each benchmark matching one broad capability we target during pretraining.

The suite includes four new benchmarks: {\bf{BasicSkills}}, a set of 6 tasks to isolate the development of skills during pretraining (e.g., basic arithmetic, reasoning, and coding); {\bf{Gen2MC}}, a multiple-choice version of 5 short-form generative tasks; {\bf{MT MBPP}}, a translated BPB set for MBPP in 17 code languages; and {\bf{Masked Perplexity}}, a new evaluation method applying token masking and calculating perplexity only on tokens that are difficult to learn. We evaluate with masked perplexity using UltraChat and WildChat, which provides a wide coverage of real user interaction evaluation in pretraining. Additional design and implementation details for \olmothreeeval{} are included in Appendix \ref{sec:eval-suite}.

\subsection{Stage 1: Pretraining}\label{sec:pretraining}

\begin{table}[h]
\centering
\footnotesize
\renewcommand{\arraystretch}{1}
\begin{tabular}{l l rr ll ll}
\toprule
{\bf Source} & 
{\bf Type} & 
\multicolumn{2}{c}{{\bf 9T Pool}} & 
\multicolumn{2}{c}{{\bf 6T Mix}} & 
\multicolumn{2}{c}{{\bf 150B Mix}} \\
\cmidrule(lr){3-4} \cmidrule(lr){5-6} \cmidrule(lr){7-8}
& & {\bf Tokens} & {\bf Docs} & {\bf Tokens} & {\bf Docs} & {\bf Tokens} & {\bf Docs} \\
\midrule
\rowcolor{ai2offwhite}Common Crawl & Web pages & 8.14T & 9.67B & 4.51T (76.1\%) & 3.15B & 121B (76.9\%) & 84.5M \\ 
\olmocrPDF & Academic documents & 972B & 101M & 805B (13.6\%) & 83.8M & 19.9B (12.6\%) & 2.25M \\
\rowcolor{ai2offwhite} Stack-Edu (Rebalanced) & GitHub code & 137B & 167M & 409B (6.89\%) & 526M & 11.1B (7.06\%) & 14.3M \\
arXiv & Papers with LaTeX & 21.4B & 3.95M & 50.8B (0.86\%) & 9.10M & 1.29B (0.82\%) & 247K \\
\rowcolor{ai2offwhite}FineMath 3+ & Math web pages & 34.1B & 21.4M & 152B (2.56\%) & 95.5M & 4.10B (2.60\%) & 2.57M \\
Wikipedia \& Wikibooks & Encyclopedic & 3.69B & 6.67M & 2.51B (0.04\%) & 4.24M & 64.6M (0.04\%) & 119K \\
\rowcolor{ai2offwhite}{\bf Total} & & {\bf 9.31T} & {\bf 9.97B} & {\bf 5.93T (100\%)} & {\bf 3.87B} & {\bf 157B (100\%)} & {\bf 104M} \\
\bottomrule
\end{tabular}
\caption{
\textbf{Composition of \dolmatoomix} including our 9T pool of data, the 6T mix we used for final model training, and the 150B mix we used for experimentation. 
}
\label{table:data-stage-1}
\end{table}

We first train \olmothreebase on {\dolmatoomix}, our 6T token pretraining data mix.
While \dolmatoomix is comprised of largely the same types of data sources used in other open pretraining recipes~\citep{soldaini2024dolma,bakouch2025smollm3,olmo20242olmo2furious}, we demonstrate three key novelties:
\begin{itemize}[topsep=0pt,parsep=0pt]
    \item New tooling for fast and scalable global deduplication at the trillion-token scale;
    \item Two new methods for optimizing selection of training tokens: token-constrained mixing and quality-aware upsampling;
    \item A novel source of academic PDFs---\olmocrPDF---converted to linearized plain text using \olmocr(Section~\S\ref{sec:preparing-pdf-data})~\citep{poznanski2025olmocr}.
\end{itemize}

\noindent Table~\ref{table:data-stage-1} summarizes our data sources, pool sizes, and final training mix.\footnote{The training mixes that we release represent reconstructions of the data sampled during our actual training runs. Tokens included in these reconstructions represent all of the tokens trained on for the training run, while included documents represent a union of all unique documents that contributed at least one token during training.} As developing a base model is the most compute-intensive part of our development process, requiring training over trillions of tokens and consuming over 90\% of overall compute, we adhere to two major principles to guide our data strategy:
\begin{itemize}[topsep=0pt,parsep=0pt]
    \item We consider a source of data for pretraining if it has potential to yield enough tokens to impact model capabilities at pretraining scale. Valuable data sources that are small may not be impactful in pretraining and are better reserved for midtraining.
    \item While we embrace exploration of structured ``task'' data (e.g. QA pairs, chat instances) for training base models, we reserve their use only for later stages of midtraining (Section~\S\ref{sec:midtraining}) and long-context extension (Section~\S\ref{sec:long-context}). Task data often does not meet the pool size needed to impact our pretraining stage, even with synthetic generation, and task data also tends to have an outsized impact on evaluation results, potentially confounding data ablations for other sources.
\end{itemize}

\begin{figure}[!h]
  \centering
  \includegraphics[width=\linewidth]{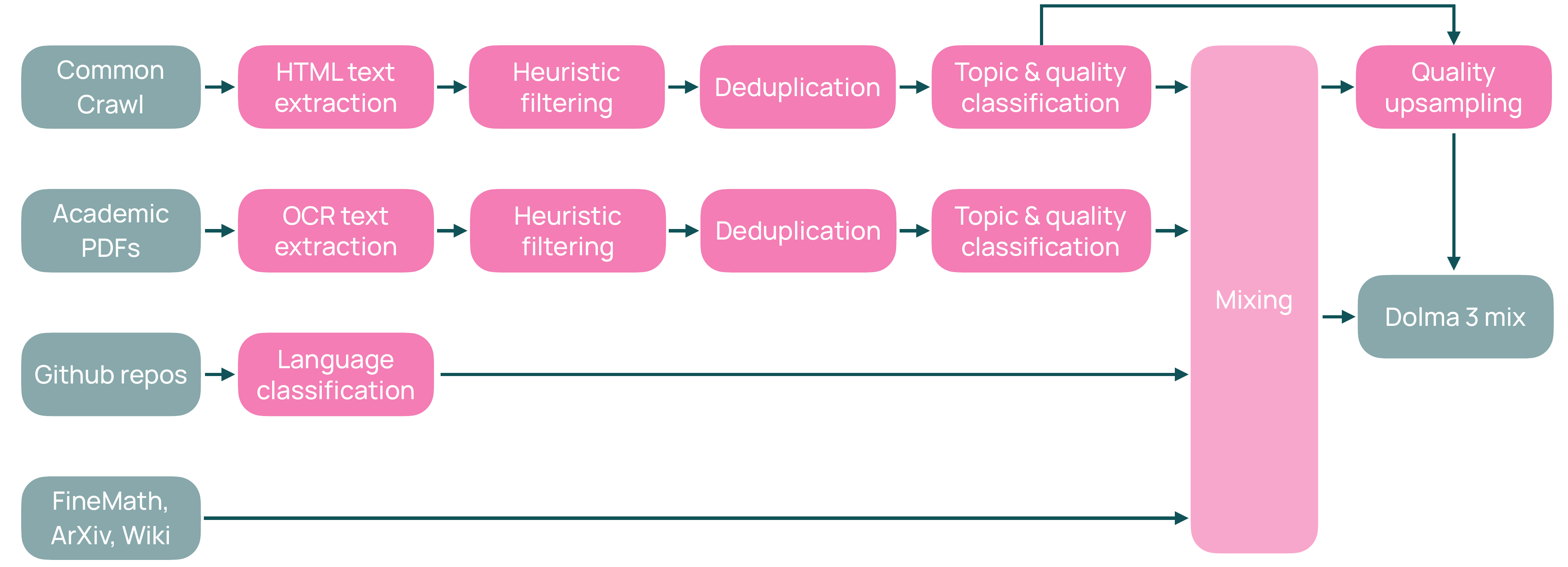}
  \caption{\textbf{Data curation flow} for pretraining data sources in \dolmatoomix.}
  \label{fig:mixing:dolma3-pipeline}
\end{figure}

Figure~\ref{fig:mixing:dolma3-pipeline} summarizes the pipeline steps for creating \dolmatoomix pretraining data. We describe them in more detail in the remainder of this section.

\subsubsection{Preparing our Web Data Pool}
\label{sec:preparing-web-data}
We took the following steps to curate pretraining data from CommonCrawl~\citep{CommonCrawl}, which constituted the majority of our pretraining corpus.
\paragraph{Text extraction} We start with 104 dumps from the CommonCrawl corpus, with a cutoff date of December 31, 2024. Following DCLM~\citep{dclm}, we remove HTML artifacts and extract the semantic text from WARC files using Resiliparse~\citep{bevendorff2018}. Where applicable, we directly leverage the raw Resiliparse-extracted data from DCLM-pool\footnote{\href{https://data.commoncrawl.org/contrib/datacomp/DCLM-pool/index.html}{\path{data.commoncrawl.org/contrib/datacomp/DCLM-pool/index.html}}}~\citep{dclm} and apply Resiliparse extraction on dumps not contained with the DCLM-pool.

\paragraph{Heuristic filtering} We apply a pipeline of heuristic filtering steps to prune our initial collection of 252.6B documents to a size amenable for pretraining. Our process closely follows that of DCLM~\citep{dclm} with minor modifications to improve data quality and computational efficiency. We first apply URL filtering to remove spam and adult-content from an expanded blocklist. We then remove documents that were either too short or too long, followed by filtering documents that contain excessive symbols or insufficient quantities of alphabetic characters. Next we remove documents containing large amounts of internal repetition and apply filtering to remove common spam phrases, fully removing any documents that are identify by these heuristics. We then use a fastText classifier~\footnote{\texttt{lid.176} from \href{https://fasttext.cc/docs/en/language-identification.html}{\path{fasttext.cc/docs/en/language-identification}}} to identify the language of each document, keeping only documents that contain English text. As a final step, we apply sentence-level heuristics from Madlad400~\citep{dclm}. In aggregate, this process reduces the size of our data pool by 84.6\%, yielding a corpus of 38.8B documents. More details are provided in Appendix~\S\ref{sec:appendixpretrain}.

\paragraph{Deduplication}
The web data we collect from CommonCrawl naturally contains an abundance of duplicated documents. This duplication arises from repeated crawls of the same website, near-copies of documents appearing across multiple web pages, and highly-repeated boilerplate text. Our deduplication strategy is motivated by three observations from prior work: 1) deduplication generally leads to more token-efficient training~\citep{lee2022deduplicatingtrainingdatamakes}; 2) duplicate count serves as a weak signal of data quality, with higher duplicate counts indicating higher quality~\citep{fang2025datasetsdocumentsrepetitionspracticalities}; 3) repeating documents more than a handful of times provides rapidly diminishing returns~\citep{muennighoff2025scalingdataconstrainedlanguagemodels}.

Given these observations, we design our deduplication strategy to enable a future quality-based upsampling step (Section \ref{sec:combining-data-pools}). We aggressively deduplicate our dataset at multiple granularities, targeting the removal of exact replicas, near-duplicates, and repeated filler text. While this necessarily discards the quality signal from duplicate counts, it produces a clean base dataset from which we can later selectively reintroduce repetition for high-quality documents. Our goal is a final dataset with minimal repetition overall, with any duplication concentrated in high-quality data. We implement our deduplication procedure in three distinct stages:

\begin{enumerate}
    \item{{\bf{Exact deduplication}} We apply global deduplication based on document text hashes to remove all exact copies. This step identifies 67\% of the pool as duplicates, reducing the dataset from 38.7B to 12.8B documents.}
    \item{{\bf{Fuzzy deduplication}} We apply MinHash-based deduplication to identify and remove near-identical documents, such as documents copied across multiple domains that differ only in headers or footers. We partition the dataset into 32 shards, ran MinHash deduplication on each shard, then performed exhaustive pairwise Jaccard similarity checks within each identified cluster. From each cluster, we retain the most recent document by crawl date. This procedure identified 23\% of the pool as duplicates, yielding 9.8B documents.}
    \item{{\bf{Substring deduplication}} The previous steps removes whole duplicate documents but did not address repeated content within individual documents. Many documents contain substantial boilerplate text or HTML artifacts (e.g., headers and footers) of limited training value. To remove these repeated substrings, we apply a novel fuzzy suffix-array-based deduplication procedure. We partition the dataset into 57 shards and apply this procedure to each, marking any substring of 500 or more bytes that occurred multiple times. Unlike previous suffix-array methods, we preserve at least one occurrence of each repeated substring in the corpus. We then merge the intervals marking repeated substrings to also remove short substrings sandwiched between longer repeated segments. This procedure removes 14\% of text bytes, yielding 9.7B documents totaling 36.5T bytes of uncompressed text.}
\end{enumerate}

This three-stage procedure reduces the web corpus from 38.7B to 9.7B documents—a 75\% reduction in document count. The resulting aggressively deduplicated dataset can then be partitioned by topic and quality and controllably upsampled for training.

To scale our deduplication strategy, we develop the Duplodocus tool,\footnote{\href{https://github.com/allenai/duplodocus}{\path{github.com/allenai/duplodocus}}} a native-rust toolkit for large-scale distributed execution of both hash-based exact deduplication and MinHash fuzzy deduplication.

\paragraph{Topic and quality classification}

We use our WebOrganizer tool~\citep{weborganizer} to partition the deduplicated corpus into 24 topics (e.g., ``\textit{Adult Content}'', ``\textit{Politics}'', or ``\textit{Science and Technology}''). 
To speed up processing of the \dolmatoo~pool, we distill the transformer-based models by \citealt{weborganizer} into a simpler fastText model.\footnote{\href{https://huggingface.co/allenai/dolma3-fasttext-weborganizer-topic-classifier}{\path{huggingface.co/allenai/dolma3-fasttext-weborganizer-topic-classifier}}}
We only partition by topic, not format.
We also train and apply a fastText-based quality classifier\footnote{\href{https://huggingface.co/allenai/dolma3-fasttext-quality-classifier}{\path{huggingface.co/allenai/dolma3-fasttext-quality-classifier}}} to assign each document a quality score. Following DCLM~\citep{dclm}, we use OpenHermes-2.5 ~\citep{OpenHermes} and ELI5~\citep{fan2019eli5} as positive training examples, supplemented with UltraChat-200k~\citep{ding2023enhancing} and WildChat-1M~\citep{zhao2024wildchat}. Negative training examples consist of 30GB sampled from DCLM-RefinedWeb.

We apply both the topic and quality classifiers to the full deduplicated corpus in order to partition the dataset. Documents are first partitioned by topic, then within each topic partition we compute quality score percentiles and subdivide documents into vigintile buckets (5-percentile intervals). This two-stage partitioning yields 480 disjoint subsets (24 topics $\times$ 20 quality tiers), enabling fine-grained control over the topic and quality distribution of our pretraining mixture.

\paragraph{Final web data pool} The above steps results in an 8T-token pool of annotated data, partitioned into buckets according to topic and text quality. This pool serves as the foundation for our pretraining mixture, though additional processing is required to construct the final training data. Specifically, we apply quality-based filtering and topic reweighting to generate a balanced, high-quality mixture, as discussed in Section~\S\ref{sec:combining-data-pools}.

\subsubsection{Preparing our \olmocrPDF Data Pool}
\label{sec:preparing-pdf-data}

We curate a novel dataset of academic PDFs, replacing our previous use of peS2o~\citep{peS2o}.
These documents are crawled ``politely'': we identify our crawler as \texttt{AI2Bot},\footnote{Crawling notice: \href{https://allenai.org/crawler}{\path{allenai.org/crawler}}} we adhere to \texttt{robots.txt}, and do not bypass paywalls. 
The crawler is seeded with a focus on academic sites and paper repositories.
We process all PDFs using the first version of \olmOCR~\citep{poznanski2025olmocr}. Ultimately this crawl generates a collection of 238 million unique PDF documents with a cutoff date of December 2024.

\paragraph{\olmOCR text extraction}
To convert PDFs to a format usable by our trainer, we apply pre-filtering and text extraction. If a document contains born-digital text, we used the Lingua language detector to retain only English documents and remove documents where spam or SEO-optimization keywords exceeded 0.4\% of total words. We then extract text using \olmOCR~\citep{poznanski2025olmocr} (versions 0.1.49-0.1.53). If olmOCR fails, we use Poppler's \texttt{pdftotext} as a fallback; documents requiring this fallback for more than 1 in 250 pages are excluded from the corpus. This yields a dataset of 160 million PDF documents.

\paragraph{Deduplication} We then identify and remove any fuzzy-duplicates using a MinHash algorithm. This differs slightly from the MinHash step we apply to the web text corpus in Section~\S\ref{sec:preparing-web-data}: we use the MinHash parameters as in FineWeb~\citep{penedo2024fineweb}, which targets document pairs with at least 75\% similarity; and we omit an exhaustive pairwise Jaccard similarity check. After this deduplication step, we were left with a corpus of 156M documents for a removal rate of 2.3\%.

\paragraph{PII filtering}
Next we remove documents containing PII from the pool of PDFs. Our goal was to to remove documents that contained sensitive standalone PII, such as government IDs and login information, as well as documents that link biographical, medical, location, employment, or educational information to a specific individual.
Through iteration, we determine that PII detection must be \textit{document type-aware} to be effective.
For example, a conference paper might contain name and place of employment of authors; however, as research articles are intended for publication, removal would not make sense. 
At the same time, a bank statement might contain the same name and employer information, and is clearly a document a language model should not be trained on. 
The rule we follow is: \textit{is this document type intended for public dissemination?}
We use manual annotators to iterate which documents types are not suitable for public dissemination, and what PII attributes we should consider. 
The resulting taxonomy is used as part of a multi-stage model-based PII filtering pipeline.

First we classify documents using a prompt to Gemma 3 12B~\citep{team2025gemma3} on the first page of each document to determine if they contain any sensitive standalone PII, or link sensitive information to an individual. Next, we use Gemma 3 4B on the first 5,000 characters of each document to arrive at a set of flags describing the type of document. From these classification results, we develop a set of rules to identify which types of documents containing PII should be publicly available and which should be filtered. Ultimately this removes 4.9\% of the remaining pool and yields a pool of 148 million documents. See~\citet{poznanski2025olmocr} for more a complete overview of the PII removal pipeline.

\paragraph{Heuristic filtering}
After PII removal, we apply a round of heuristic filtering to further remove low-quality documents. Filters applied in this step include checking for: non-English documents not originally caught by the Lingua filter; documents that were more than 30\% tables; and documents that contain more than 20\% numbers. Next we apply modifications that convert markdown tables to HTML and remove URL references. 
The combination of these filtration steps yield a corpus of 108 million documents. This corpus is then partitioned into 24 topical buckets, according to the WebOrganizer topic classifier~\citep{weborganizer}, and passed off to the mixing (Section~\S\ref{sec:combining-data-pools}).

\subsubsection{Preparing Code, Math, and other sources}
\label{sec:preparing-other-data}

\paragraph{Code} For code data, we use Stack-Edu~\citep{allal2025smollm2smolgoesbig}, an improved curation of GitHub repositories from the-stack-v2 dataset~\citep{lozhkov2024starcoder} with additional filtering for educational programming content. We keep partitions of the data by programming language for subsequent mixing.

\paragraph{Math}
As in \olmotoo, we include arXiv documents from the Proof-Pile-2 dataset~\citep{azerbayev2023llemma}, which in turn are from the RedPajama dataset~\citep{together2023redpajama} and have a cutoff date of April 2023. We use this source primarily because it preserves the original LaTeX notation, enabling the model to learn both mathematical content and how to properly format it.

Furthermore, we replace our previous use of OpenWebMath~\citep{paster2023openwebmath} with FineMath~\citep{allal2025smollm2smolgoesbig}, a subset of Common Crawl documents that contain mathematical educational content and have been reprocessed to preserve proper mathematical notation. We include all documents that have a quality score of at least 3 (out of 4), according to the FineMath classifier. This data has a cutoff date of September 2024.

\paragraph{Other} Finally, we include the Wikipedia and Wikibooks sources from \dolma~\citep{soldaini2024dolma} as base sources of encyclopedic knowledge. These are both the ``English'' and ``Simple'' editions of Wikipedia and Wikibooks with a cutoff date of March 2023. These sources were processed using WikiExtractor~\citep{Wikiextractor2015} to remove markup formatting, and all documents with 25 or fewer words were filtered out to exclude template pages or pages that encountered XML parsing errors.

\subsubsection{Sampling and Mixing over Data Pools}
\label{sec:combining-data-pools}
The data sources described above collectively provide over 9 trillion tokens of diverse text data. Transforming this collection into a training dataset requires a mixing and sampling pipeline to prescribe exactly how much of each source to include in a final training mix, and how much, if any, upsampling to apply to each source. We apply a mixing strategy that draws on swarm-based methods to train and evaluate many smaller proxy models, using these results to inform an optimal mix. Further, we apply a novel conditional mixing procedure to account for the fact that our data sources were being constantly refined and updated throughout the development cycle. In this section, we describe how we derive the final at the mixing ratios for each source; for web text, we only optimize ratios at the topic category level and apply quality-aware upsampling to obtain the final mix.

\paragraph{Constrained data mixing}

\begin{figure}[!h]
  \centering
  \begin{subfigure}{0.48\linewidth}
    \centering
    \includegraphics[width=\linewidth]{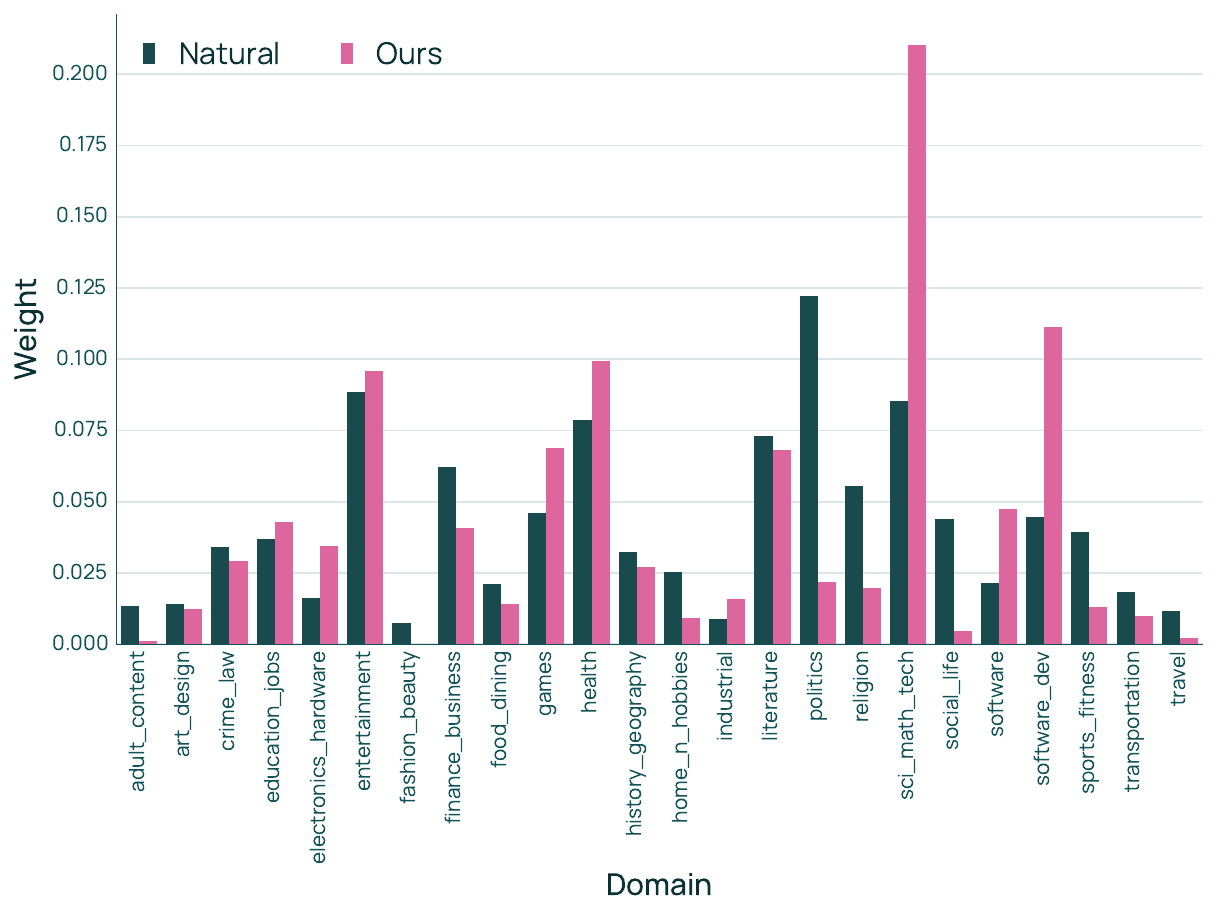}
    \caption{DCLM Baseline partitioned by topic.}
    \label{fig:mixing:dclm}
  \end{subfigure}\hfill
  \begin{subfigure}{0.48\linewidth}
    \centering
    \includegraphics[width=\linewidth]{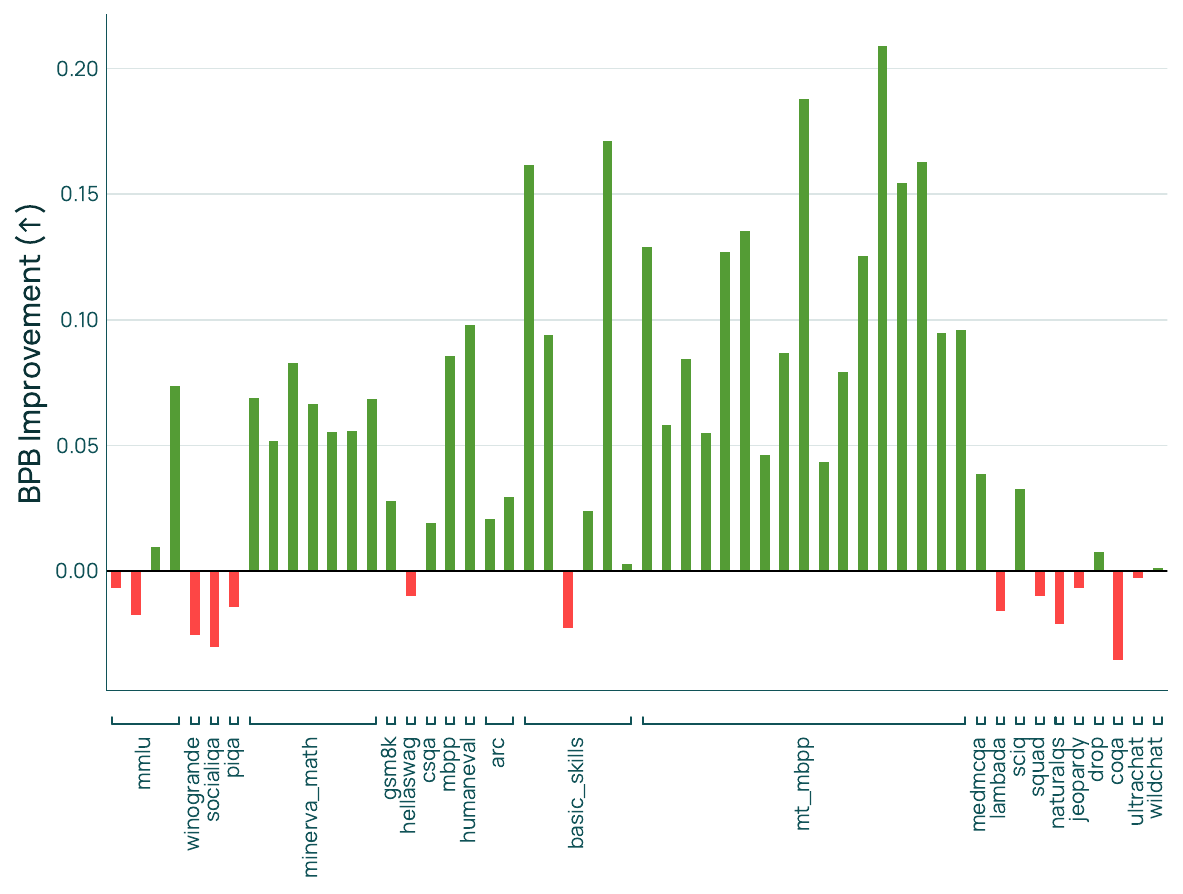}
    \caption{Improvement when training over DCLM Baseline.}
    \label{fig:mixing:dclm_pareto}
  \end{subfigure}

  \vspace{0.8em} %

  \begin{subfigure}{0.48\linewidth}
    \centering
    \includegraphics[width=\linewidth]{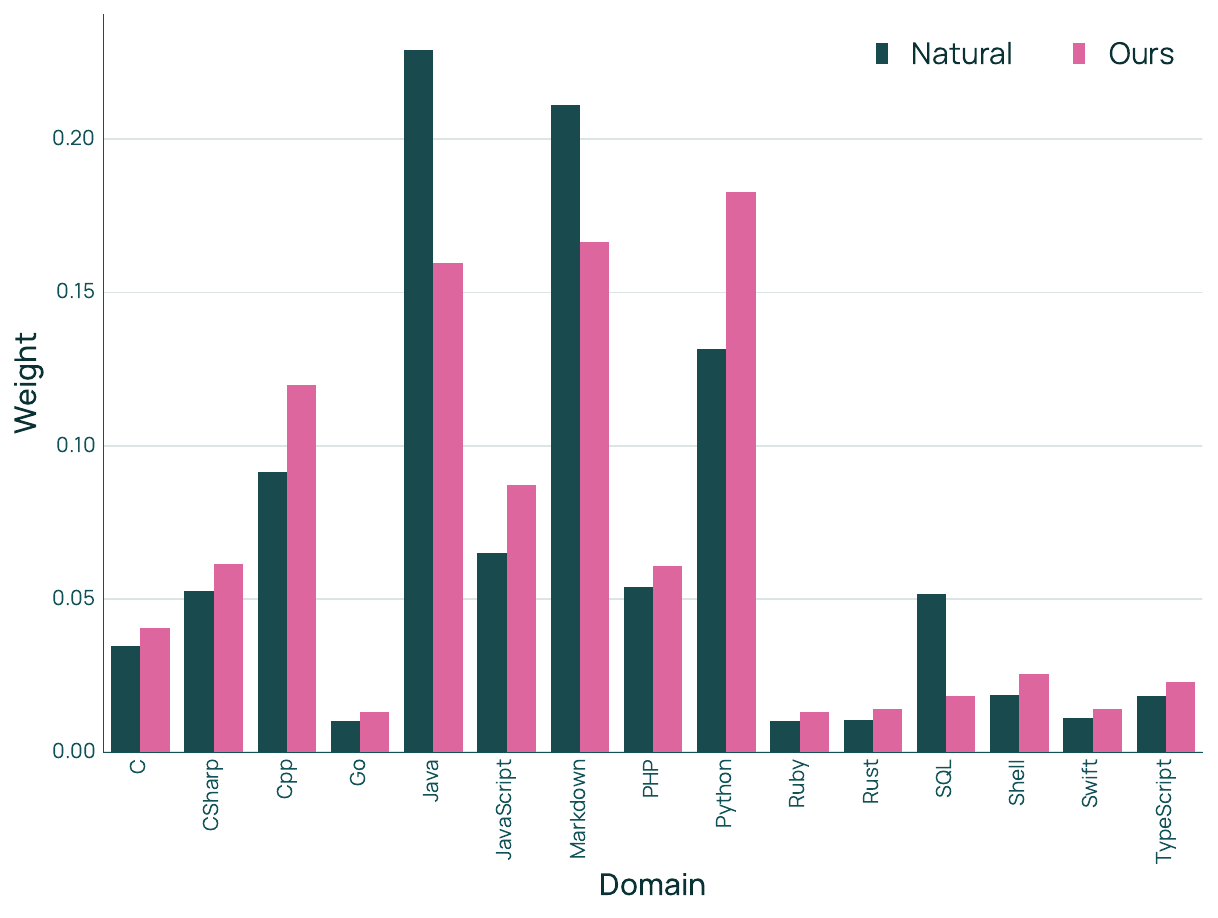}
    \caption{Stack-Edu partitioned by programming language.}
    \label{fig:mixing:Stack-Edu}
  \end{subfigure}\hfill
  \begin{subfigure}{0.48\linewidth}
    \centering
    \includegraphics[width=\linewidth]{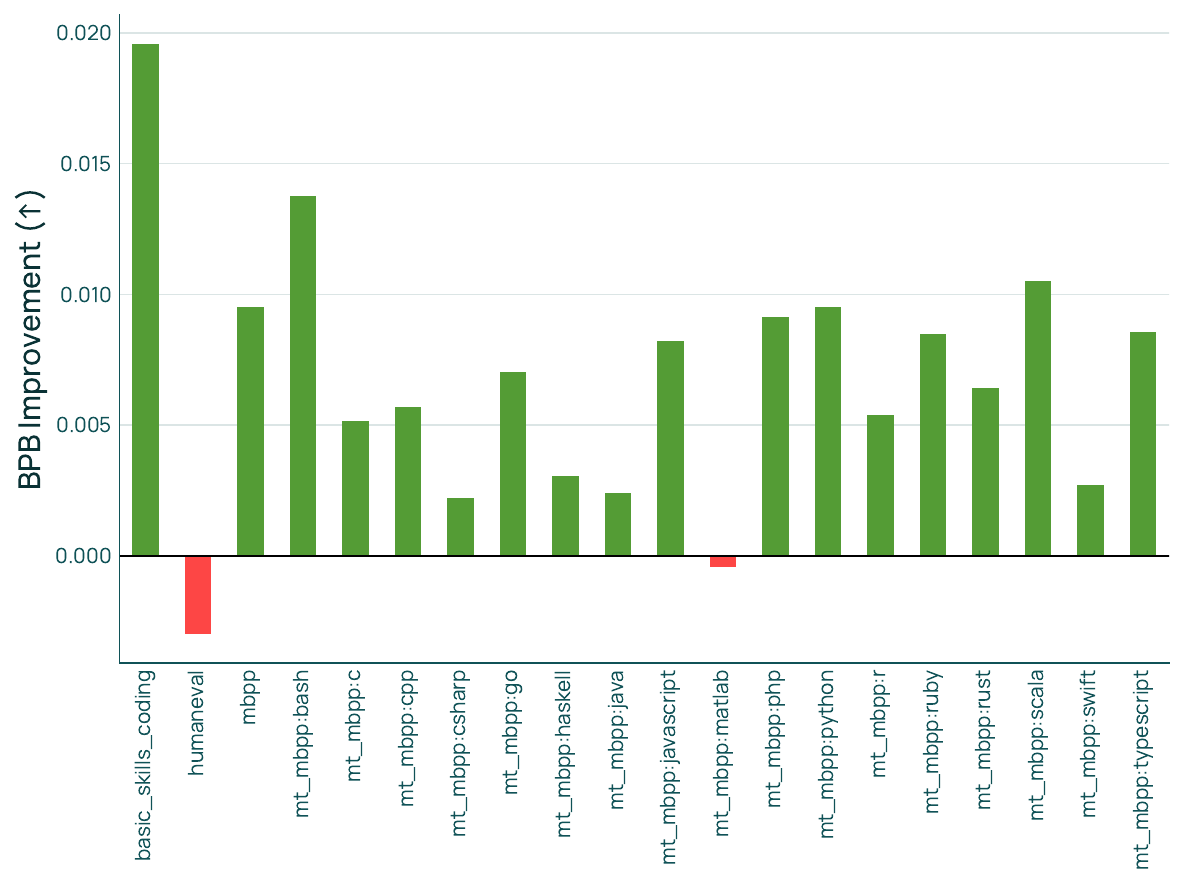}
    \caption{Improvement when training over Stack-Edu.}
    \label{fig:mixing:Stack-Edu_pareto}
  \end{subfigure}
  \caption{\textbf{Examples and effects of constrained data mixing for \olmothree{}.} On the left, comparison of the natural distribution of data sources in the \textcolor[HTML]{194a4e}{\bf\dolmatoo~pool} versus our learned data mixture in \textcolor[HTML]{dc669d}{\bf\dolmatoomix} (Figures~\ref{fig:mixing:dclm}~and~\ref{fig:mixing:Stack-Edu}). On the right, the \textcolor[HTML]{559c35}{\bf{improvement}} on downstream evaluations resulting from training on our data mix compared to the natural distribution (Figures~\ref{fig:mixing:dclm_pareto}~and~\ref{fig:mixing:Stack-Edu_pareto}).
  }
  \label{fig:mixing:main}
\end{figure}

We applied data mixing across all pretraining sources, as well as across the WebOrganizer topics within the web data and PDF sources, and the Stack-Edu programming languages.
Our mixing procedure~\citep{olmix}, consists of two components: a base procedure that constructs a high-quality mix over a fixed set of data domains, and a meta-procedure called conditional mixing that efficiently updates an existing mix when domains change. Together, these allow us to iteratively build an optimal mix and adapt to data refinements or additions without starting from scratch.

The base procedure follows a swarm-based approach inspired by RegMix~\citep{liu2024regmix}, Data Mixing Laws~\citep{ye2025datamixinglawsoptimizing}, and CLIMB~\citep{diao2025climbclusteringbasediterativedata};
it consists of three stages:
\begin{enumerate}
\item {\bf Swarm construction}. We sample the space of possible mixes by training many small proxy models, each with a different mixing ratio. Specifically, we train 30M-parameter models following the \olmothree architecture for 3B tokens (5x Chinchilla), sampling each mix from a Dirichlet distribution centered on the natural (no-mixing) distribution. As a rule of thumb, we launch a swarm of size 5x that of the number of domains. We then evaluate each proxy model on the Base Easy suite.
\item {\bf Per-task regression}. Each proxy model provides a datapoint mapping mixture weights to task performance---measured in bits-per-byte (BPB)---for each task. We fit a separate generalized linear model for each task, enabling us to predict how any candidate mix will perform.

\item {\bf Mix optimization}. We find the mixture that minimizes the average task BPB, as predicted by the per-task regression models. Since we ultimately seek a corpus with a 6T token budget, and we avoid repeating any domain more than approximately $4-7$ times, this naturally imposes maximum ratio constraints on certain domains based on their available token counts. We solve this constrained optimization using a guided search initialized from a prior or natural distribution.
\end{enumerate}

The base procedure assumes fixed domains, but real preprocessing workflows evolve continuously as we refine filters, add domains, or discover and mitigate quality issues. Rather than recomputing an entire swarm each time domains change, we introduce a new procedure called conditional mixing to efficiently adapt the base method to an evolving data landscape. The key idea is to treat the existing optimized mix as a single virtual domain with frozen mixing ratios, then re-run the base procedure over this virtual domain plus any new or modified domains. This effectively restricts the base mixing procedure to a lower-dimensional subspace of the mixture weight space, reducing swarm size and computational cost. Further details and justification of this procedure can be found in~\cite{olmix}.

To construct the \dolmatoomix weights, we perform three rounds of our conditional mixing procedure, with each stage building incrementally on frozen mixtures from prior stages. We first obtain optimized mixture weights over the 24 WebOrganizer categories within the DCLM Baseline mix\footnote{\href{https://data.commoncrawl.org/contrib/datacomp/DCLM-baseline/index.html}{\path{data.commoncrawl.org/contrib/datacomp/DCLM-baseline/index.html}}} as well as the source-level mix. Web text serves as the starting point because it constitutes the largest data pool and because we use it to develop the base mixing methodology. As finalizing the bespoke web data pool described in Section~\S\ref{sec:preparing-web-data} occurs concurrently with these initial mixing rounds, we perform this first round of mixing on DCLM-Baseline, expecting that learned preferences would transfer to our final web data.

Having frozen a mixture across WebOrganizer categories over web text, we turn our attention to mixtures of programming languages from Stack-Edu. Diverging slightly from the conditional mixing procedure, we fix the web text ratio to be 75\% of the pool and force a 25\% mixture of Stack-Edu data and only optimize over the composition of programming languages within this 25\%. Finally, we perform one more round of conditional mixing to integrate the 24 WebOrganizer categories of the PDF data, conditioned on the DCLM, Stack-Edu, and source-level mixes. This incremental approach towards mixing is essential: for example, we complete PDF curation substantially later than other sources, and conditional mixing enable us to incorporate late-arriving data while reusing prior optimization results rather than restarting the expensive swarm-based base procedure.

Figure~\ref{fig:mixing:main} presents mixing outcomes and their performance results relative to the natural data distribution. For web text (top panels), the optimized mixture dramatically upweights STEM domains (e.g. ``Science, Math, and Technology'' and ``Software Development''). On 1B-parameter models trains for 5x Chinchilla, this mixture obtains an  average improvement of 0.056 and max of 0.209 (in BPB), while only 13 out of 54 tasks show degradations, none of which exceed 0.035. For rebalancing of programming languages in Stack-Edu (bottom panels), the optimized mix favors Python over Java and Markdown, yielding modest improvements in all but two coding benchmarks. Table~\ref{tab:token-constrained-mixing} further demonstrates our method's adaptability: swapping development suites to emphasize QA, math, or coding produces mixtures that preferentially optimize these respective capabilities.

\paragraph{Quality-aware upsampling}

\begin{figure}[!h]q
  \centering
  \includegraphics[width=0.9\linewidth]{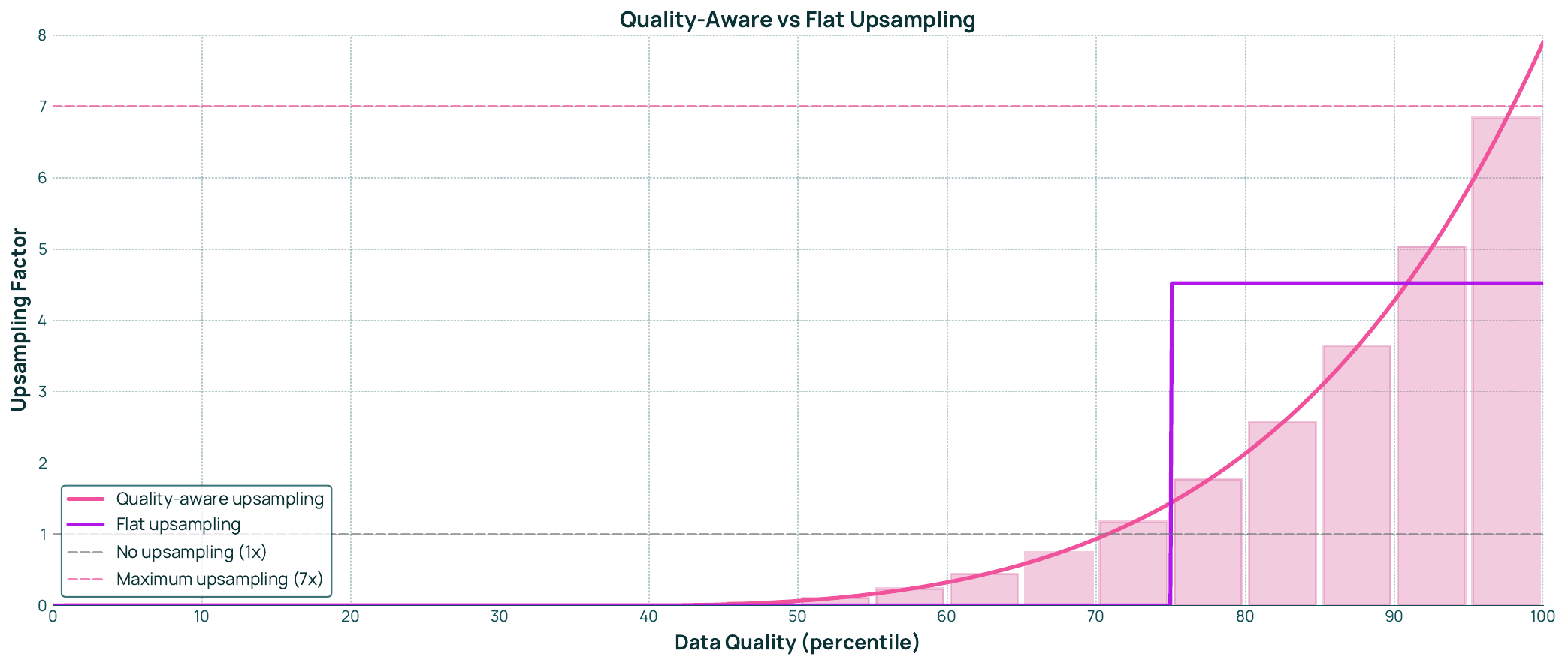}
  \caption{\textbf{Example of quality-aware upsampling curve} compared to a flat upsampling curve. The x-axis denotes quality of data in terms of percentiles and the y-axis denotes how much the data is repeated. In this instance, the bottom 40\% of data is discarded, and the top 5\% of data is resampled 7 times.}
  \label{fig:mixing:quality-aware}
\end{figure}
The data mixing procedure described in the previous section determines optimal proportions across different data sources and topics, but does not account for quality variations within each topic. For web text sources like CommonCrawl, we initially derive these proportions from DCLM, which applies only flat filtering-based on quality classifier scores. However, in a separate set of experiments, we found that quality-aware upsampling improves performance in data-constrained settings (see Appendix). For example, when constructing a 250B token mix from a 1T token pool, flat quality-filtering (as in DCLM) would simply select the top quartile. We achieve better results by upsampling the highest-quality data: including multiple copies of the top 5\% and single copies of the remaining data to reach the target token count.

We formalize this approach using upsampling curves, as in Figure \ref{fig:mixing:quality-aware}. The x-axis represents data quality in percentiles, while the y-axis shows the upsampling factor. Flat filtering corresponds to a step function on this plot, and quality-based upsampling would correspond to a monotonically increasing curve. For the purposes of generating a training data corpus, we generate separate upsampling curves for each of the 24 WebOrganizer-defined topics in our web text pool. The integral of each curve determines the total tokens extracted from that topic: for example, an integral of 2.0 indicates an average upsampling rate of 2x, yielding twice the token count from that data bucket.

To define an upsampling curve for each web text topic bucket, we leverage three constraints: 1) the optimal topic proportion, as determined by the mixing experiments; 2) the total desired training duration in terms of tokens; and 3) a maximum upsampling factor of 7 (empirically determined). The first two of these constraints control the target integral (average upsampling rate) for each topic bucket. The third constraint dictates an upper bound on the upsampling curve. Given these constraints, we can search over the space of curves to find a parametric curve that meets these constraints, which becomes the upsampling curve for this topic-bucket. In practice, our data is organized into discrete quality buckets that partition the quality percentile range. For each quality bucket, we compute its upsampling rate by integrating the upsampling curve over the corresponding percentile interval and dividing by the interval width. More details regarding this procedure can be found in Appendix~\S\ref{sec:appendixpretrain}.

\paragraph{Evaluation during pretraining}
\label{sec:eval:duringpretraining}

It can be difficult to obtain a reliable estimate of model performance in the middle of a pretraining run, since the quality of a run is highly influenced by the learning rate (see \cite{olmo20242olmo2furious}, Section 4.1).
For a 7B model, we can anneal the learning rate to zero at regular intervals throughout training to assess progress, but this is prohibitively expensive for a 32B model.
To monitor performance of our 32B model during the training run, we use the technique from~\cite{modelmerginginpretraining}, and average the weights from four checkpoints, chosen 1{,}000 steps apart at regular intervals.

\subsection{Stage 2: Midtraining}\label{sec:midtraining}

\begin{table}[!h]
\centering
\footnotesize
\small
\renewcommand{\arraystretch}{1}
\begin{tabular}{l l rr ll }
\toprule
{\bf Type} & 
{\bf Source} & 
\multicolumn{2}{c}{{\bf 2T Pool}} & 
\multicolumn{2}{c}{{\bf 100B Mix}} \\
\cmidrule(lr){3-4} \cmidrule(lr){5-6}
& & {\bf Tokens} & {\bf Docs} & {\bf Tokens} & {\bf Docs} \\
\midrule
    \rowcolor{ai2offwhite} Math (synth) & TinyMATH Mind** & 899M & 1.42M & 898M (0.9\%) & 1.52M  \\
    Math (synth) & TinyMATH PoT** & 241M & 729K & 241M (0.24\%) & 758K  \\
    \rowcolor{ai2offwhite} Math (synth) & CraneMath* & 5.62B & 6.55M & 5.62B (5.63\%) & 7.24M  \\
     Math (synth) & MegaMatt* & 3.88B & 6.79M & 1.73B (1.73\%) & 3.23M \\
     \rowcolor{ai2offwhite} Math (synth) & Dolmino Math\^{}\^{} & 10.7B & 21M & 10.7B (10.7\%) & 22.3M  \\
\midrule
    Code & StackEdu (FIM)\^{} & 21.4B & 32M & 10.0B  (10.0\%)& 16.2M  \\
    \rowcolor{ai2offwhite} Python (synth) & CraneCode* & 18.8B & 19.7M & 10.0B  (10.0\%)& 11.7M  \\
\midrule
    QA (synth) & Reddit To Flashcards** & 21.6B & 370M & 5.90B (5.9\%) & 101M  \\
    \rowcolor{ai2offwhite} QA (synth) & Wiki To RCQA** & 4.22B & 22.3M & 3.0B (3.0\%) & 16.3M  \\
    QA (synth) & Nemotron Synth QA\^{} & 487B & 972M & 5.0B (5.0\%) & 10.6M  \\
\midrule
    \rowcolor{ai2offwhite} Thinking (synth) & Math Meta-Reasoning** & 1.05B & 984K & 381M (0.38\%) & 401K \\
    Thinking (synth) & Code Meta-Reasoning** & 1.27B & 910K & 459M (0.46\%) & 398K \\
    \rowcolor{ai2offwhite} Thinking (synth) & Program-Verifiable** & 438M & 384K & 159M (0.16\%) & 158K \\
    Thinking (synth) & OMR Rewrite FullThoughts\^{} & 850M & 291K & 850M (0.85\%) & 394K \\
    \rowcolor{ai2offwhite} Thinking (synth) & QWQ Reasoning Traces\^{} & 4.77B & 438K & 1.87B (1.87\%) & 401K  \\
    Thinking (synth) & General Reasoning Mix\^{} & 2.48B & 668K & 1.87B  (1.87\%) & 732K \\
    \rowcolor{ai2offwhite} Thinking (synth) & Gemini Reasoning Traces\^{} & 246M & 55.2K & 246M (0.25\%)) & 85.1K \\
    Thinking (synth) & Llama Nemotron Reasoning Traces\^{} & 20.9B & 3.91M & 1.25B (1.25\%) & 368K \\
    \rowcolor{ai2offwhite} Thinking (synth) & OpenThoughts2 Reasoning Traces\^{} & 5.6B & 1.11M & 1.25B (1.25\%) & 402K \\
\midrule
    Instruction (synth) & Tulu 3 SFT\^{}\^{} & 1.61B & 1.95M & 1.1B (1.1\%) & 1.45M \\
    \rowcolor{ai2offwhite} Instruction (synth) & Dolmino 1 Flan\^{}\^{} & 16.8B & 56.9M & 5.0B (5.0\%) & 14.8M  \\
\midrule
    PDFs & \olmocrPDF (HQ subset)\^{} &  240B & 28.7M & 4.99B (5.0\%) & 1.20M \\
    \rowcolor{ai2offwhite} Web pages & STEM-Heavy Crawl\^{} & 5.21B & 5.16M & 4.99B (5.0\%)& 5.53M \\
    Web pages & Common Crawl (HQ subset)\^{} & 1.32T & 965M & 22.4B (22.5\%) & 18.3M \\
    
\midrule
\rowcolor{ai2offwhite} {\bf Total} & & {\bf 2.19T} & {\bf 2.52B} & {\bf 99.95B (100\%)} & {\bf 236M} \\
\bottomrule
\end{tabular}
\caption{
\textbf{Composition of the midtraining data (\dolminostoo)}. 
Here we show the full composition of the midtraining data mix. **=newly-introduced synthetic dataset. *=novel recreation of existing data. \^{}\^{}=reuse of previously-introduced data. \^{}=filtering or light transformation of existing external data.    
}

\label{table:stage-2-data}
\end{table}

\begin{figure}[!t]
  \centering
  \includegraphics[width=.8\linewidth]{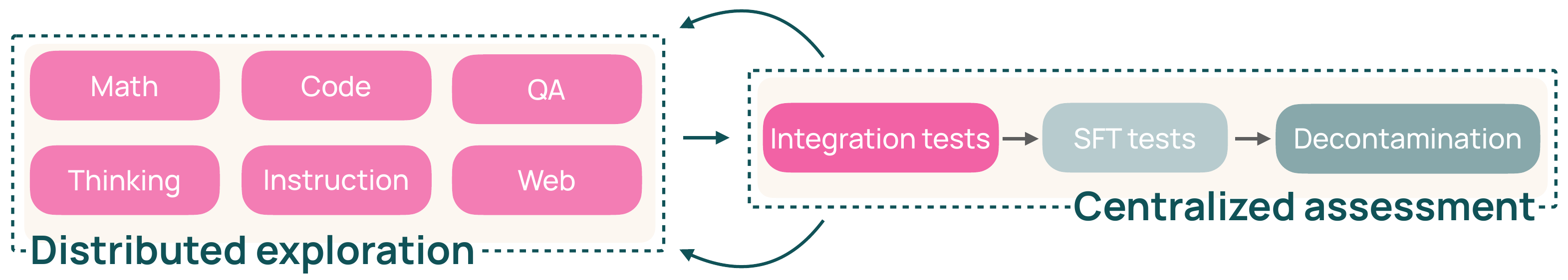}
  \caption{\textbf{Flow for midtraining data curation.} We employ a distributed system of lightweight feedback loops to explore datasets for targeted boosts across capabilities, and combine these with centralized integration tests and SFT training for assessment of candidate mix quality (discussion in Section~\S\ref{sec:midtraining-methods}). Finally, we incorporate a newly-developed decontamination method, to ensure that our mix is not contaminated with evaluation data (discussion in Section~\S\ref{sec:midtraining-methods}). %
  }
  \label{fig:mixing:midtraining-experiment-flow}
\end{figure}

After pretraining, \olmothreebase is further trained to improve key fundamental capabilities. 
During this midtrain stage, we use 100B high-quality tokens sampled from a brand new data pool we introduce in this work, {\bf \dolminostoo}. This midtraining data significantly expands and improves upon \dolminos, which we curated for our previous model \olmotoo.
The improvement comes from two key elements:
\begin{itemize}
    \item A new {\bf two-part methodological framework} combining 1) lightweight, distributed feedback loops on individual data sources, with 2) centralized integration tests to assess candidate mixes on base model quality and post-trainability.
    \item Expansion to {\bf targeted data curation efforts} across code, math, and general knowledge QA domains (broadening from the math-focused efforts in \dolminos).
    \item More intentional inclusion of data types---instruction data and thinking traces---to {\bf lay groundwork for supporting post-training} of \olmothreethinking, \olmothreeinstruct, and \olmothreerlzero models.
\end{itemize}

The resulting midtraining data is a diverse mixture that combines novel synthetic sources with data from pretraining stage, but quality-filtered and rewritten to better suit capabilites we target at this stage.
Through midtraining, we achieve improvements across the board in our target capability domains, as well as improvements in performance resulting from subsequent SFT training.

\subsubsection{Methodological framework}\label{sec:midtraining-methods}

\paragraph{Targeted capability boosts}
In the midtraining stage, we aim to make targeted improvements to capabilities spanning a wide range of domains: prioritizing significant gains in code and math, but also aiming for focused improvements in QA and general knowledge access capabilities, and to lay groundwork for instruction and thinking capabilities in post-training. 
This requires a lightweight, distributed framework for dataset testing, to allow us to investigate many domains of datasets efficiently and in parallel~(Figure~\ref{fig:mixing:midtraining-experiment-flow}).

For lightweight testing we use the microanneal methodology introduced with \olmotoo, which we further modify for more systematic baselining. For a standard microanneal we use the following setup: 1) select a target dataset, 2) sample 5B tokens, 3) match this with 5B web tokens, 4) anneal on the resulting 10B mix. We then compare the performance of the resulting checkpoint against that of a baseline microanneal on 10B web-only data, for a cheap and efficient assessment of the impact of the dataset on base model performance, over and above the impact of continued training on web data alone.\footnote{The microanneal framework allows for flexibility to test small datasets, and as a result the specifics of our microanneals varied based on dataset needs. Variants of the above include some 5B microanneals for datasets that could only support 2.5B tokens, some microanneals that test the target dataset as a smaller percentage of a more diverse 10B mix, and certain microanneals---for large numbers of comparisons between variable-size datasets---that use the original microanneal methodology omitting compute-matched baseline comparisons and assessing based on the individual annealing gains directly.}

This methodology allows us to make rapid, targeted assessments of the quality of datasets being considered for the midtraining mix, and to iterate on many data domains in parallel. 
Our workflow operates as follows: for each capability that we target for improvement  (in categories of math, code, QA, instruction, and thinking), we generate or collect new datasets  as candidates to boost performance for this capability; we assess each via microanneals---if the results are promising, new datasets can be incorporated into the larger integration tests described next.

\paragraph{Integration tests}
In parallel with the microanneal process, we conduct integration tests involving full annealing runs on candidate mixes for the 100B-token midtraining mix. 
These integration tests evaluate how candidate data sources perform when combined together;  
further, we can assesss effect of longer 100B midtrain runs (as compared to shorted, 5--10B tokens used in microanneals). 

Finally, checkpoints from integration runs can be quickly instruction-tuned and evaluated on the post-train eval suite; 
we use this additional step to verify that gains we observe in midtrain yield improvements beyond base model capabilities.

We run these integration tests periodically as we reach a critical mass of microanneal results for new candidate data sources. 
For each integration test, new sources that show promise in microanneals are incorporated into an updated 100B mix, retaining strong sources from previous iterations.

We carry out five major rounds of integration tests; we report three in this manuscript: \roundOne, \roundThree, and \roundFive. 
\roundFive folds in the newly-developed decontamination process (Section~\S\ref{sec:midtraining-contamination}). 
For each mix we evaluate the resulting midtrained model on our \olmothreeeval Main evaluation suite, and additionally run the midtrained model through SFT for post-training assessment.

\subsubsection{Capability Improvements for Final Data Mix}

With \dolminostoo, we target five core capabilities during midtraining: improved math and coding, better knowledge elicitation through QA, and bootstrapping instruction following and reasoning ability ahead of post-training stages. 
To maintain continuity with pretraining, we keep web and PDF data from the first stage of \olmothree, albeit after filtering for higher quality documents; 
this approach prevents excessive shift in training data distribution.
Table~\ref{table:stage-2-data} outlines the composition of the final mix, which includes a combination of newly-introduced synthetic data and refinements of existing data. 
Below we give an overview, for each capability category, of our curation efforts and final selected data. Additional details are in Appendix~\ref{app:sec:midtraining-data}, and dataset descriptions and replication resources for novel datasets are provided in the \dolmatoo repository\footnote{\href{https://github.com/allenai/dolma3}{\path{github.com/allenai/dolma3}}}.

\paragraph{Math capabilities}

For math capability, we expand efforts from \dolminos. 
We consider a total of 25 data sources, which we evaluate over 80 microanneal runs. 
We ultimately settle on a combination of 5 top math-specific sources, 4 of which were newly synthesized. 
For high-performing existing datasets without permissive licensing, we synthesize new data modeled after those datasets.

We will outline and briefly summarize the math-targeted data sources that are included in the final mix. 
More details about data generation procedure and microanneal results can be found in the Appendix.
\begin{itemize}
\item {\bf Dolmino-1 math} We include the entirety of the 10.7B-token \dolminos Math subset. The version we use differs from the original only in additional filtering for decontamination. 
As described for \olmotoo~\citet{olmo20242olmo2furious}, this set was generated to lift general-purpose math capabilities, measured in terms of improvements on the  GSM8K test set. 
A 10B microanneal, using 5B of the available 10.7B tokens in isolation, achieves a lift in 10.4 points in MATH and 38.2 points in the GSM8K benchmark.\footnote{Performance benefits seen in Math microanneals are stated in terms of improvement relative to a pre-anneal baseline.}

\item {\bf TinyMATH} For each of the 7500 examples in the MATH training set, we generate 100 new, similar problems. We then create Python code solutions to the newly for each problem (TinyMATH-PoT), and two flavors of conversational English discussing these solutions (TinyMATH-MIND). In aggregate, this yields 1.14B tokens of novel, synthetic data targeted to improve performance on the MATH benchmark. 
A microanneal consisting of all of these new tokens in a 50/50 ratio with web data yields 13.2 points of improvement in the MATH benchmark and 13.9 points in GSM8K.

\item {\bf CraneMath} The recently published SwallowMath dataset~\citep{fujii2025rewriting} demonstrates the potential of rewriting already finely-curated naturally-occurring mathematical web data---in this case, FineMath4+ ~\citep{allal2025smollm2smolgoesbig}. We corroborate this strong performance with a microanneal over SwallowMath that showed a lift of 16.0 points in MATH and 24.5 points in GSM8K using only 3.6B high quality tokens. 
Because SwallowMath comes with additional license restrictions---having been generated with the Llama suite of models---we generate an independent reproduction of SwallowMath by rewriting FineMath4+ with the SwallowMath prompt, using Qwen3~\citep{qwen3} for generation. We denote this new mix as CraneMath, which yields 5.6B tokens of high-quality math. 
Microanneals demonstrate a lift of 18.5 points in MATH and 27.4 points in GSM8K.

\item {\bf MegaMatt}
Similar to SwallowMath, Megamath-Web-Pro-Max~\citep{wang2025octothinker} applies Llama rewrites to naturally-occurring mathematical web text---in this case a filtered version of MegaMath-Web~\citep{zhou2025megamath}. Our microannealing procedure demonstrates that MegaMath-Web-Pro-Max was able to improve MATH by 7.0 points and GSM8K by 13.3 points using only 5B tokens of high-quality data. However, in order to use this dataset, we re-generate it using open source models. Specifically, we collect the Megamath-Web-Pro data occurring after June 2023, apply filtering as in Megamath-Web-Pro-Max, and rewrite it using Qwen3~\citep{qwen3}. This yields 3.88B tokens of high-quality data, which we refer to as MegaMatt. 
In microanneals, this data yields a lift of 8.0 points in MATH and 13.0 points in GSM8K.
\end{itemize}

\paragraph{Code capabilities}

Our efforts to improve code capabilities include two major threads: 1) curation of higher-quality general code data, and 2) introduction of fill-in-the-middle (FIM) code capabilities. The top-performing datasets included in the final mix are the following:

\begin{itemize}
\item{\bf Stack-Edu (FIM)} We include a modified version of Stack-Edu, in which 50\% of documents reflect fill-in-the-middle (FIM) transformation via the infilling procedure from StarCoder2~\citep{lozhkov2024starcoder}. This transformation splits code documents into prefix, middle, and suffix segments in order to train on prediction of the concealed middle segment. To further improve the quality of this code data, we apply quality filtering by performing reservoir sampling and bucketing of documents based on educational value score,\footnote{For educational value score we use language-specific classifiers provided developed for Hugging Face SmolLM model series, e.g. \href{https://huggingface.co/HuggingFaceTB/stack-edu-classifier-php}{\path{huggingface.co/HuggingFaceTB/stack-edu-classifier-php}}.} followed by weighted random sampling of the upper 20\% of buckets from each language subset. Microanneals validate that this quality filtering combined with the sampling procedure improves code benchmark performance over both the natural distribution of Stack-Edu and more naive sampling procedures such as sampling the top document per language based on classifier score.

\item {\bf CraneCode} As with our math datasets, we find strong performance from the SwallowCode dataset, and generate a permissively-licensed recreation for use in our midtraining. 
Like~\citet{fujii2025rewriting}, we source data from the Python subset of the-stack-v2-smol\footnote{\href{https://huggingface.co/datasets/bigcode/the-stack-v2-train-smol-ids}{\path{huggingface.co/datasets/bigcode/the-stack-v2-train-smol-ids}}, released by \citet{lozhkov2024starcoder}.}, 
then filter for syntax errors and filter based on linter outputs.
Then, we apply the SwallowCode two-stage rewriting pipeline, with one stage to augment style, and another to optimize the code itself. 
This yields 18.8B tokens of high-quality python code. 
In a microanneal using 5B tokens of high-quality data, CraneCode results in a lift in HumanEval of 5.0 points relative to pre-anneal baseline, compared to the 10.3 seen for SwallowCode. 
When using a larger microanneal with 12.5B tokens of CraneCode, the lift in HumanEval improves to 13.5.

\end{itemize}

\paragraph{QA and knowledge access capabilities}
We target improvements in question-answering and general knowledge access capabilities through synthesis of two novel datasets focused on particular QA capabilities, as well as inclusion of high-quality existing QA data. The final datasets included for these capabilities are the following:

\begin{itemize}
\item {\bf Reddit-to-Flashcards} We synthesize this dataset in response to the need to handle diverse content categories and question structures in multiple-choice QA tasks. We first identify a subset of academically-relevant subreddits, and then use GPT 4o-mini to rewrite submission-comment pairs from those subreddits into multiple-choice QA pairs.
We use seven task formats to increase diversity.
Microanneals show that inclusion of 5B tokens of this data in a 10B-token microanneal resulted in over 2 points of improvement in the $\text{MC}_\text{Non-STEM}$ task cluster---relative to a 10B-token web-only baseline microanneal---with 3 points of improvement in MMLU.

\item {\bf Wiki-to-RCQA} We synthesize this dataset in response to the need for improvements in passage-based reading comprehension QA. We collect Wikipedia passages and prompted Qwen2.5 32B Instruct to generate QA pairs based on these passages, meeting a range of constraints inspired by instructions given to annotators of reading comprehension QA datasets. Microanneals show that 4.2B tokens of this data in a 10B microanneal results in nearly 2 points of improvement in the GenQA task cluster relative to a 10B web-only baseline, with improvements focused on the DROP, SQuAD and CoQA reading comprehension QA benchmarks.

\item {\bf Nemotron} We include the ``diverse QA pairs'' synth subset of the Nemotron CC dataset~\citep{su2025nemotroncctransformingcommoncrawl},
as, in microanneals, it improved GenQA tasks by 1.5 points, $\text{MC}_\text{Non-STEM}$ by 1.9 points, and it had equal $\text{MC}_\text{STEM}$ performance compared to a microanneal run of web documents from the top quality (5\%) bucket.
All other Nemotron synth subsets (``distill'', ``extract knowledge'', ``knowledge list'', and ``wrap medium'') performed worse than natural data, so we did not use them.

\end{itemize}

\paragraph{Cross-Capability instruction data}
To lay the groundwork for post-training, we include cross-domain instruction datasets to prime models for instruction-tuning.

\begin{itemize}
\item{\bf Tulu3 SFT data} We sample instruction data from the  SFT set from \tulu. Compared to dataset released by  \citet{lambert2024tulu3}, we lightly process these data as follows:
1) we use an expanded  set of examples that were created and subsequently filtered out for the final \tulu data, 2) instead of relying on post-train syntax, such as \texttt{<|im\_start|>} and \texttt{<|im\_end|>}, we concatenate messages using double newlines. We choose this format, rather than using special tokens after microanneal experiments comparing them. More details are provided in see discussion of special tokens in Section~\S\ref{sec:midtraining-findings}.

\item{\bf Flan} Through microanneals, we also find the Flan dataset~\citep{wei2021flan,longpre2023flan} improves performance in QA tasks, and as a result included a subset of the Flan dataset in the final mix. We use same subset and preprocessing from  \olmotoo~\citep{olmo20242olmo2furious}. 
\end{itemize}

\paragraph{Cross-capability thinking traces}
We also curate a diverse collection of thinking traces across a variety of domains to lay the foundation for \olmothreethinking and \olmothreerlzero. This includes two new synthetic datasets, as well as rewritten and filtered versions of existing thinking trace datasets.
\begin{itemize}
\item {\bf{Meta-reasoning}}
The first of the two new datasets introduced in this work; we crate it to target seven core cognitive capabilities from~\citet{kargupta2025cognitive} that are foundational to mathematical and programming expertise: self-awareness~\citep{toy2024metacognitionneedusingintrospection,bfa322bf36e54a4ca19f9a73bee6184b}, evaluation~\citep{Fleming2017-FLESOD}, goal management~\citep{ACKERMAN2017607,GRIFFITHS201924}, hierarchical organization~\citep{Haupt2018}, backward chaining~\citep{Olieslagers2024}, backtracking~\citep{Joyce2009}, and conceptual reasoning~\citep{Markovits2015}. 
These categories are inspired by work suggesting that meta-reasoning capabilities in base models may be associated with superior reinforcement learning trajectories~\citep{kargupta2025cognitive,gandhi2025cognitive}. 
We express these capabilities into tasks\footnote{See Appendix Tables~\ref{tab:math-meta}~and~\ref{tab:code-meta} for list of tasks, and \href{https://github.com/allenai/dolma3/tree/main/datasets/dolma3_dolmino_mix/meta-reasoning}{\path{github.com/allenai/dolma3/tree/main/datasets/dolma3_dolmino_mix/meta-reasoning}} for the prompts.} that require levering meta-reasoning, such as backtracking from an answer back to its original math problem, or debugging a program. 
To generate our meta-reasoning data for each of these tasks, we synthetically augmented existing math \citep{deepscaler2025,moshkov2025aimo2} and code \citep{tacoli,hendrycksapps2021,ahmad2025opencodereasoning} problems with detailed annotations such as `problem classification', `difficulty analysis', `solution approaches', `common pitfalls', and `verification methods', modeled after the Pandalla-Math dataset.\footnote{\href{https://huggingface.co/datasets/pandalla/pandalla-math-dataset-v1.0}{\path{huggingface.co/datasets/pandalla/pandalla-math-dataset-v1.0}}} 
Using these annotations as foundation, we prompt \texttt{GPT-4.1} and \texttt{o4-mini} to generate thinking traces for each capability-targeted task.
Microanneals show that inclusion of this data results in substantial improvements to math and coding tasks, resulting in approximately 14 points of boost---relative to a strong math/code baseline microanneal---in Minerva Math, and 14 and 20 points of boost on Codex HumanEval and MBPP benchmarks, respectively.

\item {\bf{Program-verifiable data}}
Our second new synthetic reasoning dataset consists of program-verifiable tasks~\citep{zeng2025rlve} for which we can use a Python program to deterministically verify whether an answer to a problem is correct. Solving these problems naturally requires a wide range of meta-reasoning strategies that are well-suited to be learned during midtraining. We 1) programmatically generate these problems, 2) distill thinking traces from  \texttt{GPT-4.1} and \texttt{o4-mini} models, and 3) finally filter those for correctness using an output verifier (Python programs). 
Microanneals show that including about 250M verifiable data tokens (in a 5B microanneal) led to 1-2 points of improvement on math and code tasks, including GSM8K and MBPP, relative to a math/code microanneal baseline.

\item{\bf{OMR rewrite full-thoughts}} We also consider 9 different versions of rewriting\footnote{Documentation for this approach, including all prompts, is available at \href{https://github.com/allenai/dolma3/tree/main/datasets/dolma3_dolmino_mix/open_math_reasoning_rewrites}{\path{github.com/allenai/dolma3/datasets/dolma3_dolmino_mix/open_math_reasoning_rewrites}}.} of the OpenMathReasoning dataset~\citep{moshkov2025aimo2}, and find top performance for what we call the Full-Thoughts rewrite. This is a light rewrite of the OpenMathReasoning dataset, instructing GPT-4.1 to edit items for clarity, flow, and formatting (e.g., converting to LaTeX) while preserving all reasoning, explanations, and thoughts of the original. In microanneals, training on all 850M OMR Full-Thoughts tokens and an equal amount of web text, we see a lift of 5.5 points in the MATH benchmark and a 8.4 lift in GSM8K.

\item {\bf Existing thinking traces}
We also draw on a variety of existing synthetic thinking trace datasets, to which we apply a range of filtering steps to reduce noise and increase quality. These sources have coverage over a broad variety of domains, including math, code, natural sciences, social sciences, humanities, and puzzles. These datasets are listed in Table~\ref{table:stage-2-data}, and more details are provided in Appendix~\ref{app:sec:midtraining-data}. 
Microanneals show that inclusion of these datasets yielded improvements especially in math and code domains, with improvements of up to 8 points in GSM8K, and approximately 2 points in HumanEval and MBPP, relative to a math/code microanneal baseline.
\end{itemize}

Table~\ref{tab:with_without_reasoning} provides further results showing the impacts of inclusion of instruction and thinking data in our midtraining mix, at the level of full integration tests.

\paragraph{High quality web and PDF data}

Finally, we include three types of web / pretraining data to avoid skewing too far from the pretraining distribution.

\begin{itemize}
\item {\bf Stage 1 web data} We sample documents from the top two quality buckets (top 10\% quality).
We sample according to natural distribution, not the optimal ratio described in Appendix~\S\ref{sec:commoncrawl-mixing}.
In tests, the optimal ratio from the pretrain stage results in no improvement over natural distribution; 
since it introduce additional implementation complexity, we abandon it for the midtraining stage.
\item {\bf Stage 1 \olmocrPDF} From our PDF documents (Section~\S\ref{sec:preparing-pdf-data}) we create a further filtered version, which we use both for midtraining and for long-context extension. Instead of discussing details here, the reader will have to hold their breath till Section~\S\ref{lc:data}. This creates tension in the manuscript, giving them something to look forward to. %
\item {\bf Stem-heavy crawl} We also create a separate high-quality web collection, crawled between September 12, 2024 and June 3rd, 2025 using our in-house crawler. 
The crawler ingested scientific, educational, and general domains based on domain-level seeds sourced from manual lists of websites deemed high value. 
We use same crawling policy described as \olmocrPDF (Section~\S\ref{sec:preparing-pdf-data}).
Through microanneal experiments, we choose to filter this set using the quality classifier introduced in Section~\S\ref{sec:preparing-web-data}; 
in detail, we use a threshold score of 0.6, which corresponds to the top $2.83\%$ of the data we crawled, and would make put these sources in the top $0.79\%$ of web data in the \dolmatoo~pool.
Relative to a web-only baseline, our crawled data yields an improvement of  approximately 2 points each for $\text{MC}_\text{Non-STEM}$, $\text{MC}_\text{STEM}$, and Math subsets of \olmothreeeval.
\end{itemize}

\subsubsection{Decontamination}
\label{sec:midtraining-contamination}

Earlier \newOlmo models have enabled research on benchmark contamination in base model training, such as decontamination of perplexity evaluations \citep{magnusson2024palomabenchmarkevaluatinglanguage} or measuring the impact of quality filters on evaluation leakage \citep{godey2025gaperonpepperedenglishfrenchgenerative}. In \olmothree midtraining we use a decontamination tool to ensure minimal contamination with evaluation datasets. 
We focus our decontamination efforts on the midtraining stage (and the long-context extension, which drew from the same data pools) in light of results suggesting that memorization occurs most strongly near the end of training \citep{Magar2022DataCF, Bordt2024HowMC}.

\paragraph{Method and tooling} 
For decontamination, we search for and remove  matches of any split of any benchmark dataset that are part of in our evaluation harness, as for some we increased sample size by evaluating on training splits. 
We detect and remove contamination between midtraining data and benchmark documents by developing a new \texttt{decon} package\footnote{\href{https://github.com/allenai/decon}{\path{github.com/allenai/decon}}}. 
Briefly, \texttt{decon} operates in two phases: 

\begin{enumerate}
    \item {\bf{Detection phase}} For each midtraining document, \texttt{decon} samples n-grams at a regular stride, checking whether the current n-gram matches known n-gram for any benchmark in the evaluation suite\footnote{We decontaminate against all benchmarks in the \textsc{OLMES} package: \href{https://github.com/allenai/olmes}{\path{github.com/allenai/olmes}}}.
    \item {\bf{Cluster expansion phase}} If a match is found, the matching text is expanded on both sides, counting the number of adjacent ngrams that are also contaminated; if the value is above a specified threshold, the document is deemed contaminated removed. 
\end{enumerate}

The two phases approach is key for efficiency: \textit{detection} phase checks at non-overlapping intervals to speed up processing, while the \textit{cluster expansion} phase thoroughly checks for matches to compute an accurate contamination score.  

We tune the contamination score to balance precision and recall based on numerous qualitative review.

We iteratively refine our decontamination protocol; 
For example, the first version fails to decontaminate against SQuAD v2 due to a preprocessing issue; 
DROP is also incorrectly processed due to its short-question-about-a-passage format. 
We address these issues by evaluating question, answer, and passage components separately—matching primarily on questions, but using answer/passage matches as supporting information for shorter or edited questions. 
We also improve precision for multiple-choice evals by matching against full answers rather than just A/B/C/D labels. 
The \texttt{decon} repository includes configuration files that reproduce both the earlier and final approaches.
Appendix~\ref{app:decon} provides a detailed overview of \texttt{decon}.

\subsubsection{Key findings}\label{sec:midtraining-findings}

Our two-part methodological framework for evaluating midtraining enables us to track closely the quality of our candidate mixes and the behaviors of individual data sources in interaction with others. Here we detail some of the key findings from that process.

\paragraph{Candidate mix quality improves over time}
Our integration tests allows us to verify progressive improvements in our candidate midtraining mixes over time: Table~\ref{tab:rounds_comparison} shows this improvement across a sample of three candidate mixes illustrating the development trajectory. (Since midtraining development operates in tandem with pretraining, we develop mixes on earlier pretrained checkpoints---thus the comparisons here are given to illustrate progress in data curation, and should not be confused with final midtraining numbers.)

\begin{table}[tbp]
\centering
\small
\begin{tabular}{@{}l | ccccccHc|c@{}}
\toprule
 & \multicolumn{7}{c}{\textbf{\olmothreeeval}} &  & \textbf{SFT Exps} \\
    {\bf Mix}  &
    {\textbf{\fontsize{8}{8}\selectfont~Avg}} &
    {$\textbf{\fontsize{8}{8}\selectfont~MC}_\textbf{\fontsize{6}{6}\selectfont~STEM}$} &
    {$\textbf{\fontsize{8}{8}\selectfont~MC}_\textbf{\fontsize{6}{6}\selectfont~Non-STEM}$} &
    {\textbf{\fontsize{8}{8}\selectfont~GenQA}} &
    {\textbf{\fontsize{8}{8}\selectfont~Math}} &
    {\textbf{\fontsize{8}{8}\selectfont~Code}} &
    -- &
    {\textbf{\fontsize{8}{8}\selectfont~FIM}} &
    {\textbf{\fontsize{8}{8}\selectfont~Avg}} \\
\midrule
\roundOne & 49.7 & 64.3 & 75.2 & 68.3 & 47.4 & 23.4 & 40.7 & 28.4 & 35.2 \\
\roundThree  & 50.7 & 64.9 & 75.7 & 68.1 & 48.7 & 24.4 & 41.1 & 31.9 & 35.3 \\
\rowcolor{ai2lightpink} \roundFive  & 53.1 & 65.3 & 76.1 & 70.8 & 57.1 & 27.7 & 45.4 & 29.4 & 37.3 \\
\bottomrule
\end{tabular}
\caption{\textbf{Performance across candidate 100B-token midtraining mixes} on the \olmothreeeval Main suite, and in evals after subsequent SFT. We highlight three of our five total candidate mixes to provide a representative illustration of the improvement trajectory. We see that our data curation framework yields improvements across the board from our first candidate mix to our last. (Discussion in Section~\S\ref{sec:midtraining-findings}.)}
\label{tab:rounds_comparison}
\end{table}

We see in Table~\ref{tab:rounds_comparison} that across all base model metrics, as well as in evaluations of subsequent SFT training, newer candidate mixes consistently improve performance. Notably, between \roundThree and \roundFive we also introduce our decontamination process, which
means that the gains of \roundFive relative to \roundOne and \roundThree are likely underestimated in this table, given that only \roundFive reflects decontaminated data.

\paragraph{Performance shows substantial domain tradeoffs}

Alongside our central integration tests, we also conduct exploratory 100B anneals with heavy skews toward particular domains, to better understand domain tradeoffs. We treat code/math/thinking capabilities as one domain group, and generative/QA capabilities as another domain group---and create modified mixes each prioritizing one of these groups while omitting the other. Our Gen-QA mix increases proportions of web, QA, and instruction data while omitting math, code, and thinking, and our math-code-thinking mix increases proportions of math, code, and thinking data while omitting QA and instruction data (but keeping web to avoid excessive skew away from pretraining distribution).

\begin{table}[tbp]
\centering
\small
\begin{tabular}{@{}lcccccHc@{}}
\toprule
& \multicolumn{7}{c}{{\textbf{\olmothreeeval}}} \\
    {\textbf{Mix}} &
    {$\textbf{\fontsize{8}{8}\selectfont~MC}_\textbf{\fontsize{6}{6}\selectfont~STEM}$} &
    {$\textbf{\fontsize{8}{8}\selectfont~MC}_\textbf{\fontsize{6}{6}\selectfont~Non-STEM}$} &
    \textbf{{\fontsize{8}{8}\selectfont~GenQA}} &
    \textbf{{\fontsize{8}{8}\selectfont~Math}} &
    \textbf{{\fontsize{8}{8}\selectfont~Code}} &
    -- &
    \textbf{{\fontsize{8}{8}\selectfont~FIM}}
    \\
\midrule

Gen-QA mix  & 66.3 & {\bf 78.1} & 72.5 & 27.5 & 11.9 & 21.5 & 0.1 \\
Math-code-thinking mix  & 62.5 & 69.6 & 65.9 & {\bf 60.8} & {\bf 35.6} & {\bf 53.0} & {\bf 37.7} \\
\rowcolor{ai2lightpink} \roundFive (final mix)  & {\bf 66.4} & 77.4 & {\bf73.1} & 57.3 & 31.2 & 48.0 & 31.7 \\
\bottomrule
\end{tabular}
\caption{\textbf{Demonstration of tradeoffs in domain-skewed mixes} using the \olmothreeeval Main suite. Increasing weight of math and code domains in the mix improves performance in these domains---however, it comes at significant cost
to MCQA and GenQA performance. Increasing weight on GenQA domains, on the other hand, yields
minimal improvement on MCQA and GenQA tasks, while hurting math and code performance. (Discussion in Section~\S\ref{sec:midtraining-findings}.)}
\label{tab:domain_tradeoffs}
\end{table}

Table~\ref{tab:domain_tradeoffs} shows results from these runs, compared against our final \roundFive midtraining mix. We see that training on our Gen-QA mix results in a substantial drop in math and code performance, while approximately matching the final mix in $\text{MC}_\text{STEM}$, $\text{MC}_\textsc{Non-STEM}$, and GenQA performance. By contrast, in our math-code-thinking mix, math and code performance substantially exceeds that of our final mix---however, $\text{MC}_\text{STEM}$, $\text{MC}_\textsc{Non-STEM}$, and GenQA performance take a notable hit.

These results indicate that there are real tradeoffs when skewing toward certain of these domains over others during midtraining. We see in particular that there is clear potential to further improve math and code performance by increasing weight of these domains in the mix---however, this comes at a significant cost to our MCQA and GenQA performance. Increasing weight on Gen-QA domains, on the other hand, yields minimal improvement on QA tasks, while predictably hurting math and code performance. Overall, these results suggest that our final midtraining mix strikes a healthy balance across these domains, avoiding too heavy of a domain skew and enabling strong final performance across metrics.

\begin{table}[tbp]
\centering
\footnotesize
\begin{tabular}{@{}lccccccHcc@{}}
\toprule
& \multicolumn{9}{c}{\sans{Select benchmarks from}~\textbf{\olmothreeeval}} \\
\textbf{Mix} & \textbf{MMLU} & \textbf{ARC} & \textbf{GenQA} & \textbf{BasicSkills} & \textbf{GSM8K} & \textbf{Minerva} & \textbf{MultiPL-E$\textbf{HumanEval}$} & \textbf{MultiPL-E$_\textbf{MBPP}$} & \textbf{HumanEval} \\
\midrule
Web-only & 55.6 & 78.1 & {\bf 53.4} & {\bf 80.4} & {\bf 22.4} & {\bf 6.1} & 5.8 & 9.6 & {\bf 16.0} \\
\rowcolor{ai2lightpink}Reddit & {\bf 58.8} & {\bf 80.7} & 52.5 & 79.9 & 21.2 & 4.5 & {\bf 6.3} & {\bf 11.2} & 14.5 \\
\bottomrule
\end{tabular}
\caption{\textbf{Microanneal-level domain tradeoffs: Reddit-to-Flashcards} (10B microanneal, web-only baseline). We see domain tradeoffs at the level of individual sources as well: the Reddit-to-Flashcards dataset yields strong boosts in MCQA tasks and some code tasks, but decreases performance in math and GenQA tasks. (Discussion in Section~\S\ref{sec:midtraining-findings}.)
}
\label{tab:reddit_micro}
\end{table}

\begin{table}[tbp]
\centering
\footnotesize
\begin{tabular}{@{}lccccccHcc@{}}
\toprule
& \multicolumn{9}{c}{\sans{Select benchmarks from}~\textbf{\olmothreeeval}} \\
\textbf{Mix} & \textbf{MMLU} & \textbf{ARC} & \textbf{GenQA} & \textbf{BasicSkills} & \textbf{GSM8K} &\textbf {Minerva} & \textbf{MultiPL$_\text{MBPP}$} & \textbf{MBPP} & \textbf{HumanEval} \\
\midrule
Web-only & {\bf 55.2} & 77.6 & {\bf 53.7} & 80.9 & 18.4 & 6.3 & 4.7 & 6.2 & 7.9 \\
\rowcolor{ai2lightpink}Reasoning & 53.7 & {\bf 77.7} & 52.9 & {\bf 82.9} & {\bf 26.8} & {\bf 13.6} & {\bf 8.0} & {\bf 12.6} & {\bf 19.5} \\
\bottomrule
\end{tabular}
\caption{\textbf{Microanneal-level domain tradeoffs: meta-reasoning and program-verifiable reasoning} (5B microanneal, web-only baseline). We see domain tradeoffs for reasoning datasets as well: adding the meta-reasoning and program-verifiable data yields significant improvement in math and code tasks, but some performance drop in generative and MCQA tasks. (Discussion in Section~\S\ref{sec:midtraining-findings}.)}
\label{tab:metaverifiable_micro}
\end{table}

We also see these domain tradeoffs at the individual source level, observable in results from microanneals. Table~\ref{tab:reddit_micro} shows a microanneal comparison for the Reddit-to-Flashcards dataset, which relative to the web-only baseline yields improvement for multiple choice tasks, as well as a boost for certain code tasks, but results in some performance decrease in math and GenQA tasks. Conversely, in Table~\ref{tab:metaverifiable_micro} we see that our novel synthetic reasoning data---meta-reasoning and program-verifiable reasoning---yields significant improvement in math and code tasks, but results in some performance drop on certain GenQA and MCQA tasks.

\paragraph{Thinking/instruct data benefits base performance}

We also investigate the overall impact of inclusion of our post-training-oriented data---instruction and thinking trace data---through 100B integration tests on one of our intermediate midtraining mixes both with and without inclusion of these data subsets (holding total mix tokens constant). Table~\ref{tab:with_without_reasoning} shows base eval performance after each of these training runs---we see that the mix that includes these post-training elements performs better on every base eval measure. This suggests that although individual sources and domains present performance tradeoffs, the inclusion of these cross-domain post-training data types in aggregate is consistently beneficial, and this benefit begins even before post-training.

\begin{table}[t]
\centering
\small
\begin{tabular}{@{}lccccccHc@{}}
\toprule
& & \multicolumn{7}{c}{\textbf{\olmothreeeval}} \\
\textbf{Model} &
\textbf{Avg} &
    {$\textbf{\fontsize{8}{8}\selectfont~MC}_\textbf{\fontsize{6}{6}\selectfont~STEM}$} &
    {$\textbf{\fontsize{8}{8}\selectfont~MC}_\textbf{\fontsize{6}{6}\selectfont~Non-STEM}$} &
    \textbf{{\fontsize{8}{8}\selectfont~GenQA}} &
    \textbf{{\fontsize{8}{8}\selectfont~Math}} &
    \textbf{{\fontsize{8}{8}\selectfont~Code}} &
    -- &
    \textbf{{\fontsize{8}{8}\selectfont~FIM}}
    \\
\midrule
No thinking traces/instruction & 48.8 & 63.6 & 74.0 & 66.7 & 43.1 & 23.3 & 41.4 & 29.2 \\
\rowcolor{ai2lightpink}Full mix & {\bf 50.7} & {\bf 64.9} & {\bf 75.7} & {\bf 68.1} & {\bf 48.7} & {\bf 24.4} & {\bf 41.1} & {\bf 31.9} \\
\bottomrule
\end{tabular}
\caption{\textbf{Effect of thinking traces and instruction data} on \olmothreeeval.``Full mix'' is ``\roundThree'' from Table~\ref{tab:rounds_comparison}. The mix that includes instruction and thinking data performs better across base eval measures, suggesting that inclusion of these data types is beneficial even before post-training. (Discussion in Section~\S\ref{sec:midtraining-findings}.)
}
\label{tab:with_without_reasoning}
\end{table}

\paragraph{Leave special tokens for SFT stage}

To inform our formatting for instruction datasets, we also conduct an investigation to determine the impacts of inclusion or omission of special chat tokens such as \texttt{<|im\_start|>} and \texttt{<|im\_end|>} in our midtraining data. 
We test this via microanneals on the Tulu3-SFT data, comparing versions with and without these tokens. 
Experiments show that when training on data containing chat templates and special tokens, models consistently output these special tokens at inference time, resulting in evaluation scores that are dramatically reduced (e.g. GSM8K drops from 49.43 to 0, and CruxEval drops from 32.89 to 18.91). 
Further analysis highlights that simply including a chat template, with ordinary text in place of special tokens, did not produce the same performance drop (46.02 on GSM8K and 29.65 on CruxEval), suggesting that this disruption in model behavior is not due to inclusion of a chat template more generally, but is rather due specifically to the introduction of special tokens to the embedding vocabulary when they have not been seen in pretraining.

Though the degradation in model evaluation scores can be attributed primarily to disruption in answer parsing, these results highlight the broader issue that inclusion of these tokens at midtraining time results in emission of these tokens by the base model at inference time. 
Since this is an undesirable behavior, we ultimately remove both the chat template and special tokens from our instruct data, and revert to simple newline-based formatting.

\begin{figure}[ht]
    \centering
    \includegraphics[width=\linewidth]{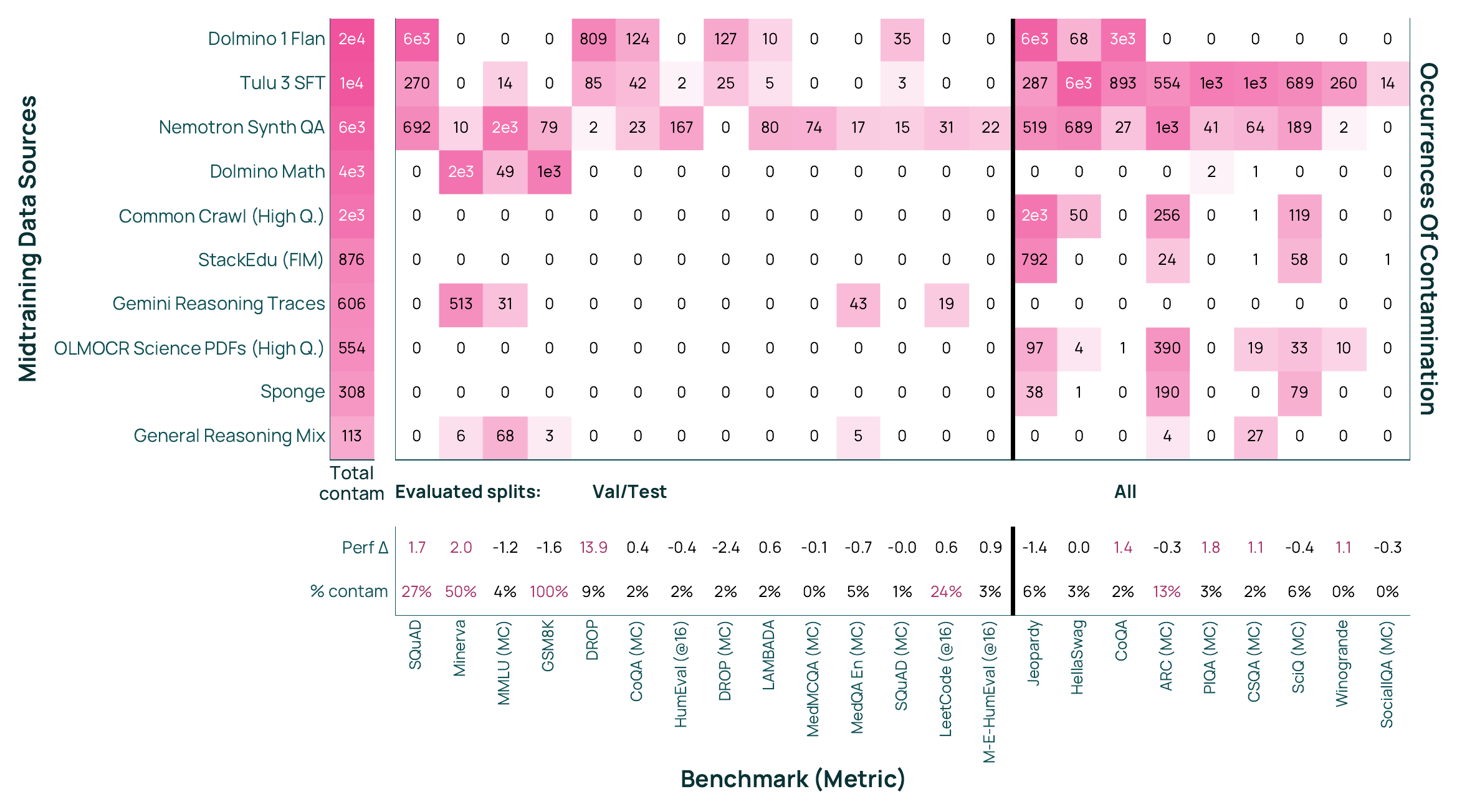}
    \caption{\textbf{Occurrences of benchmark instances in 10 most contaminated midtraining sources}. We decontaminate against all splits of all benchmarks, as some (right) include training data when evaluated to reduce noise. Some but not all contaminated benchmarks show substantial \textit{Perf $\Delta$} between contaminated and decontaminated runs (discussion in Section~\S\ref{sec:midtraining-findings}).}
    \label{fig:contam_heatmap}
\end{figure}

\paragraph{Extent and impact of decontamination are variable}
Figure~\ref{fig:contam_heatmap} shows the top ten midtraining data sources containing the most occurrences of benchmark contamination. We find that much of the contamination occurs in existing datasets such as Flan and Nemotron. Not all contamination was subtle---we found many templated contamination instances, in which fields from benchmarks were exactly matched, with templated content inserted between them. Furthermore, many of these were not isolated instances, but complete validation or test splits. For instance, Flan is constructed from templates on benchmark data, and can include validation data that is used for model development decisions since test sets are hidden (e.g., DROP).

Performance is sometimes, but not always, inflated by contamination. We investigate this by comparing our final decontaminated 100B anneal with a matched 100B anneal using the non-decontaminated data versions. Figure~\ref{fig:contam_heatmap} also shows the extent to which benchmark performance after midtraining drops when contamination is removed (\textit{Perf $\Delta$}). Some differences are substantial---such as validation or test performance changes in DROP, Minerva, SQuAD. Note that we remove contamination of all splits for all benchmarks, such as for DROP removing over 60,000 training examples from sources such as Flan. So performance differences may indicate that decontamination is preventing memorization or also removing in-distribution training examples. We remove all splits because some of our development benchmarks increase sample size by evaluating on train and held out splits (Figure~\ref{fig:contam_heatmap}, right) and several of these also show performance overestimation with contamination of any of the evaluated benchmark splits.  However, other benchmarks do not show inflated performance, despite contamination: we see that DeepSeek LeetCode performance is close to 0 with or without contamination, and SQuAD under the easier MC metric is saturated in either case. Finally, similarly to reports from Marin 32B \citep{Hall2025marin}, we find that despite the fact that our decontamination procedure detected complete leakage of GSM8K in our data, this does not result in better performance with the contaminated data. Instead we see that performance is in fact better with the decontaminated data, a phenomenon that the Marin authors explain occurs due to the contaminated formatting not matching the evaluated format.\footnote{This discussion was disseminated \href{https://x.com/percyliang/status/1983561570539176334}{on social media}.}

\paragraph{Model souping can improve midtraining performance}

For \olmothreebase 32B, we observe noteworthy performance improvement from merging two independent midtraining runs with differing seeds. 
Relative to the individual midtraining runs, the merged model yields nearly a full point of improvement in the $\text{MC}_\text{STEM}$ task cluster, 0.4 improvement in the GenQA task cluster, and in the Math task cluster result in improvements of 2.9 and 1.6 relative to the first and second midtraining runs, respectively. 
Other noteworthy improvements include approximately 1 point of improvement in MMLU,
and 5 and 2 points of improvement in GSM Symbolic relative to the first and second runs.
For this reason, we select the merged model as our final midtrained 32B checkpoint.\footnote{Initial experimentation for the 7B model did not show similar gains from model merging, so the 7B midtrained checkpoint is the result of a single run.}

\subsection{Stage 3: Long-context Extension}
\label{sec:long-context}

A crucial ability for modern language models is the capacity to operate over long sequences.
This capability is necessary to process the long inputs required by many real-world tasks.
Moreover, generating long sequences of intermediate tokens is a common technique to achieve test-time scaling~\citep{muennighoff2025s1simpletesttimescaling}.
In this section, we provide an overview of the methodology we used to scale \olmothree's context window from 8,192 to 65,536 tokens. %
We also describe \longminomix{}, a high-quality dataset of both naturally-occurring and synthetically-augmented long texts.
\longminomix{} consists of over {\bf 600 billion tokens}; statistics in Table~\ref{table:data-stage-3}.

\begin{table}[h]
\centering
\footnotesize
\renewcommand{\arraystretch}{1}
\begin{tabular}{l l rr ll}
\toprule
{\bf Source} & 
\bf Length bucket & 
\multicolumn{2}{c}{{\bf 600B Pool}} & 
\multicolumn{2}{c}{{\bf 50B Mix}} \\
\cmidrule(lr){3-4} \cmidrule(lr){5-6}
& & {\bf Tokens} & {\bf Docs} & {\bf Tokens} & {\bf Docs} \\
\midrule
\rowcolor{ai2offwhite}olmOCR PDFs & 8K-16K & 144B (22.5\%) & 12.7M & 2.27B (4.55\%) & 235K \\
olmOCR PDFs & 16K-32K  & 115B (18.0\%) & 5.06M & 1.85B (3.70\%) & 110K \\
\rowcolor{ai2offwhite} olmOCR PDFs & 32K-64K & 106B (16.6\%) & 2.30M & 4.81B (9.63\%) & 177K  \\

olmOCR PDFs &  64K-128K & 96.0B (15.0\%) & 1.05M & -- & -- \\
\rowcolor{ai2offwhite} olmOCR PDFs & 128K-256K  & 60.8B (9.5\%) & 342K & -- & -- \\
olmOCR PDFs  & 256K-512K & 35.1B (5.49\%) & 97.1K & -- &  -- \\
\rowcolor{ai2offwhite}olmOCR PDFs  & 512K-1M & 21.5B (3.36\%) & 30.2K & -- & -- \\
olmOCR PDFs & 1M+ & 26.9B (4.21\%) & 12.2K & -- & -- \\
\rowcolor{ai2offwhite} olmOCR PDFs + synth {\bf{CWE}} & 32K-64K & 8.77B (1.37\%) & 189K & 1.94B (3.88\%) & 71.3K  \\
olmOCR PDFs + synth {\bf{REX}}  & 32K-64K  & 24.1B (3.77\%) & 492K & 6.08B (12.2\%) & 217K \\ 
\rowcolor{ai2offwhite}Midtraining data mix & Variable & -- & -- & 33.0B (66.1\%) & 79.2M \\
{\bf Total} & & {\bf 639B} & {\bf 22.3M} & {\bf 50.0B (100\%)} & {\bf 80.0M} \\
\bottomrule
\end{tabular}
\caption{\textbf{Composition of \longminomix}. The 100B mix for \olmothree 32B maintains the same proportions as the 50B mix. Length buckets are reported in \dolmatoo tokens.}
\label{table:data-stage-3}
\end{table}

\paragraph{Long-context extension strategy}
Because training with long sequence lengths is computationally costly, most language models are pretrained with shorter sequences and extended only in a later stage of model development.
During the extension phase, models are trained on longer documents and the hyperparameters of positional embeddings are typically adjusted to ease positional generalization.

\paragraph{High variance in open-model recipes}
The recipes for performing this long-context extension vary dramatically between models.
The context extension phase for many language models ranges from hundreds of billions (SmolLM3: 100B,~\citealt{bakouch2025smollm3}; GLM 4.5: 100B,~\citealt{glm45}; DeepSeek V3: 123B,~\citealt{deepseekv3}; Apertus: 225B,~\citealt{swissai2025apertus}) to almost one trillion tokens (Kimi K2: 400B,~\citealt{kimiK2}; Llama 3.1: 800B,~\citealt{dubey2024llama}; DeepSeek V3.1: 840B,~\citealt{deepseekV31}).
However, there are outliers: AFM~\citep{goddard2025extendingAFM} and Nemotron Nano 2~\citep{nvidia2025nvidianemotronnano2} both use fewer than 20 billion tokens to extend to 64K and 128K, respectively.
Standalone extension recipes have also been proposed, many emphasizing token efficiency. For instance, ProLong~\citep{prolong} uses 20B tokens drawn from books and code, whereas LongAttn~\citep{wu2025longattn} constructs a 5B-token corpus using self-attention scores from existing language models to select documents exhibiting long-range dependencies.
Another key point of divergence across model families is when in the development pipeline the extension is performed: Llama 3.1 models apply long-context extension prior to midtraining, Qwen 2.5 and 3 perform it afterwards, and GLM 4.5 applies extension only after supervised finetuning.

\begin{figure}[h]
    \centering
    \begin{subfigure}[b]{0.49\textwidth}
        \centering
        \includegraphics[width=\linewidth, trim=0.05in 0.1in 0.05in 0.1in, clip]{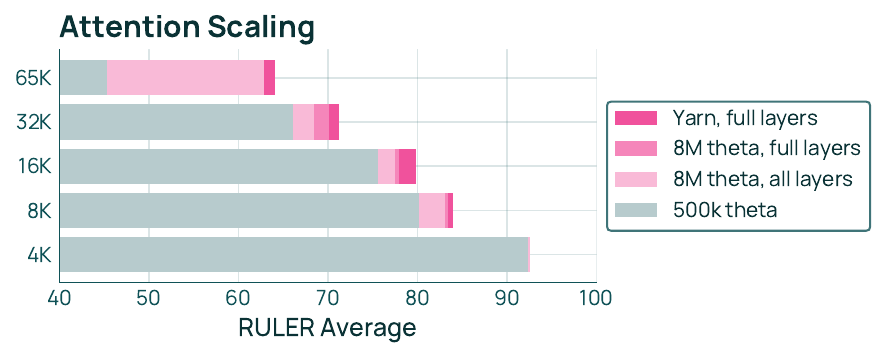}
        \caption{}
        \label{lc:component:attention_scaling}
    \end{subfigure}
    \hfill
    \begin{subfigure}[b]{0.49\textwidth}
        \centering
        \includegraphics[width=\linewidth, trim=0.05in 0.1in 0.05in 0.1in, clip]{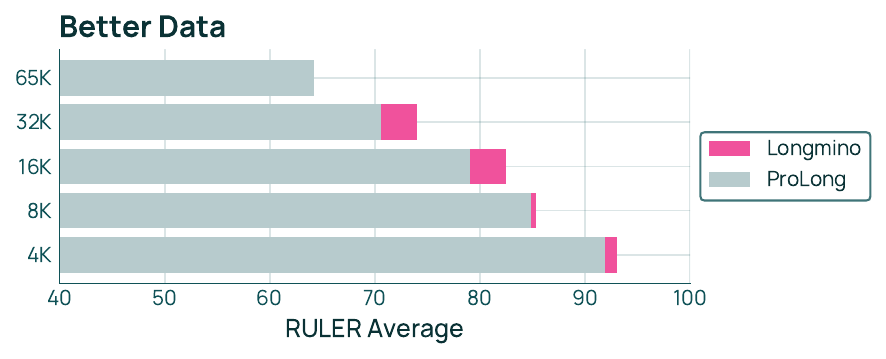}
        \caption{}
        \label{lc:component:better_data}
    \end{subfigure}
    \par\bigskip
    \begin{subfigure}[b]{0.49\textwidth}
        \centering
        \includegraphics[width=\linewidth, trim=0.05in 0.1in 0.05in 0.1in, clip]{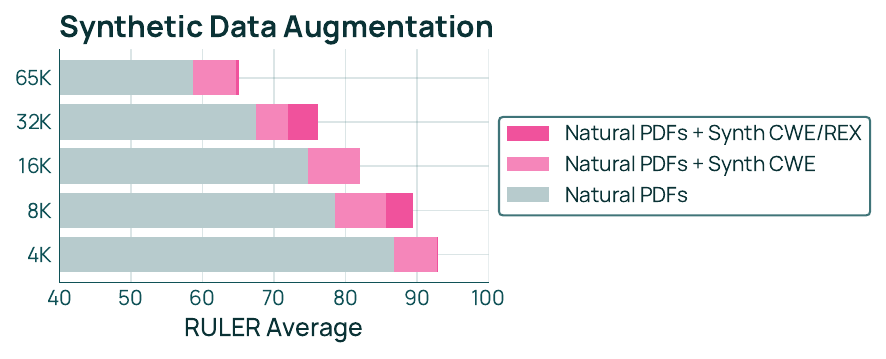}
        \caption{}
        \label{lc:component:synth}
    \end{subfigure}
    \hfill
    \begin{subfigure}[b]{0.49\textwidth}
        \centering
        \includegraphics[width=\linewidth, trim=0.05in 0.1in 0.05in 0.1in, clip]{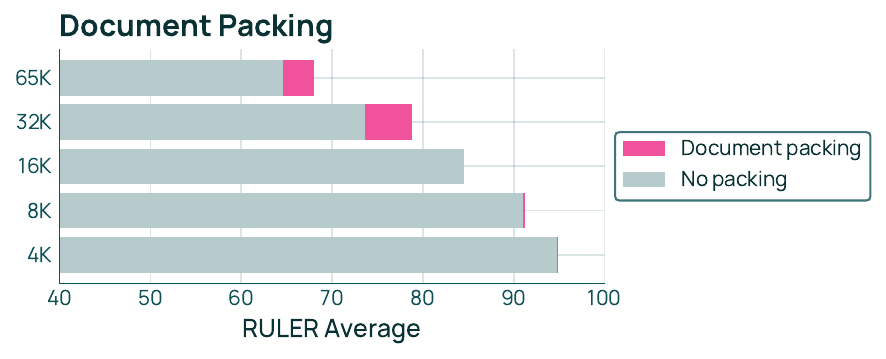}
        \caption{}
        \label{lc:component:packing}
    \end{subfigure}
    \par\bigskip
    \begin{subfigure}[b]{0.49\textwidth}
        \centering
        \includegraphics[width=\linewidth, trim=0.05in 0.1in 0.05in 0.1in, clip]{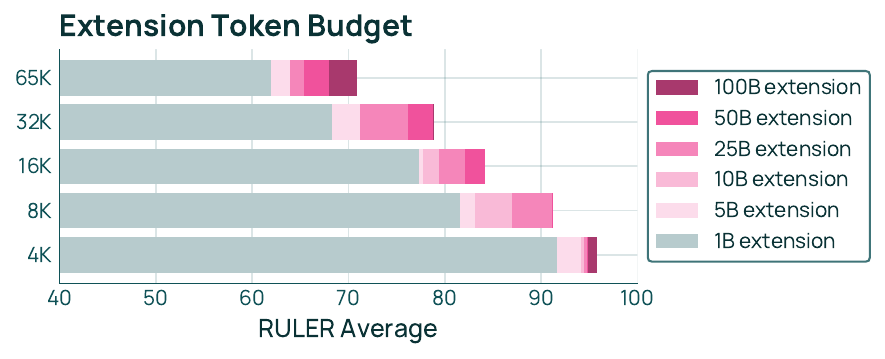}
        \caption{}
        \label{lc:component:longer}
    \end{subfigure}
    \vspace{-0.3em}
    \caption{        
        \textbf{Five key components of the \olmothree long-context extension recipe} measured on the RULER benchmark.
        applying YaRN to full attention layers only gives the best results (Figure~\ref{lc:component:attention_scaling});
        \olmocrPDF are more effective than other recipes (Figure~\ref{lc:component:better_data});
        synthetic data augmentation improves performance over natural documents alone (Figure~\ref{lc:component:synth});
        Document packing boosts performance for longer context lengths (Figure~\ref{lc:component:packing});
        longer extensions improve RULER scores, especially for longer sequences (Figure~\ref{lc:component:longer}).
        }
    \label{fig:lc-progress}
\end{figure}

\paragraph{\olmothree long-context recipe}
To extend \olmothree's context, we use long documents from the \olmocrPDF pool (Section~\S\ref{lc:data}) with additional filtering and synthetic data augmentation applied (Section~\S\ref{lc:synth}).
We call this collection \longminoPool{}.
We mix 34\% long-context data with 66\% high-quality short-context data sampled from \dolminostoo{}, and train using this mix for an additional 50B tokens for \olmothree 7B and 100B tokens for \olmothree 32B, as described in Section~\S\ref{lc:mix}.
During long-context extension, we apply YaRN~\citep{peng2023yarnefficientcontextwindow} to full attention layers, and do not adjust positional embeddings on sliding-window attention layers;
we use document packing and inter-document masking (Section~\S\ref{lc:mix}).
We summarize the key aspects of our recipe in Figure~\ref{fig:lc-progress}.
While developing this recipe, we carefully analyze and isolate architectural design decisions that have profound impact on long-context performance; our investigation is presented in \citet{bertsch2026cracks}.

\paragraph{Overall results}  We evaluate our context-extended models on two popular long-context benchmarks.
RULER \citep{hsieh2024rulerwhatsrealcontext} is a benchmark of synthetic long-context tasks including challenging variations of the Needle-in-a-Haystack task~\citep{nelson2024needlehaystackmemorybased} and simple aggregation tasks that require counting over inputs; we use RULER as the primary metric to guide our long-context recipe development.
HELMET~\citep{yen2025helmet} is a suite of long-context benchmarks across a diverse set of task types, including retrieval, in-context learning, and summarization tasks, which we evaluate on to represent more general long-context capabilities.  We keep HELMET as an unseen evaluation suite and test our final checkpoints on it.\footnote{There is some overlap between RULER and HELMET, so this is not a perfect held-out suite; however, the overlapping subsets are generally the easier ones where models trivially achieve near-perfect performance. See Appendix~\ref{appx:eval-details} for details.} %
We report results in Table~\ref{tab:ruler_baselines}.

\begin{table}[h]
\centering
\begin{tabular}{lcccccp{0.2em}cccc}
\toprule
 & \multicolumn{5}{c}{\textbf{RULER}\quad\textcolor{neutralFive}{\small\texttt{(dev suite)}}} && \multicolumn{4}{c}{\textbf{HELMET}\quad\textcolor{neutralFive}{\small\texttt{(held-out eval)}}} \\
\cmidrule(lr){2-6} \cmidrule(lr){8-11}
\textbf{Model} & \textbf{4K} & \textbf{8K} & \textbf{16K} & \textbf{32K} & \textbf{65K} && \textbf{8K} & \textbf{16K} & \textbf{32K} & \textbf{65K} \\
\midrule
\rowcolor{midgrey} \multicolumn{11}{c}{\textbf{7B scale}} \\
\rowcolor{lightgrey} Llama 3.1 8B & 95.56 & 92.76 & 93.13 & 91.43 & 86.88 && 45.00 & 43.48 & 42.44 & 40.18 \\
\rowcolor{lightgrey} Qwen 2.5 7B & 94.63 & 90.87 & 88.68 & 87.26 & 67.30 && 49.26 & 46.25 & 42.99 & 30.47 \\
\rowcolor{lightgrey} IBM Granite 3.3 8B & 91.98 & 85.69 & 82.70 & 78.13 & 67.62 && 43.19 & 41.63 & 39.31 & 35.74 \\
\rowcolor{lightgrey} Qwen 3 8B & 95.58 & 94.10 & 93.78 & 90.29 & - && 51.62 & 49.90 & 47.71 & - \\
\rowcolor{lightgrey} Xiaomi MiMo 7B & 94.33 & 93.45 & 92.53 & 89.28 & - && 50.57 & 49.68 & 46.01 & - \\
\rowcolor{lightgrey} Nemotron Nano 9B & 95.31 & 93.09 & 91.58 & 89.01 & 85.13 && 41.78 & 42.90 & 41.82 & 41.48 \\
\rowcolor{lightgrey} Apertus 8B & 90.47 & 82.48 & 74.43 & 69.05 & 59.89 && 46.09 & 43.71 & 41.26 & 35.12 \\
\rowcolor{ai2lightpink} Olmo 3 7B & 94.89 & 91.21 & 84.14 & 78.79 & 67.96 && 45.66 & 43.62 & 41.15 & 36.80 \\
\midrule
\rowcolor{midgrey} \multicolumn{11}{c}{\textbf{32B scale}} \\
\rowcolor{lightgrey} Qwen 2.5 32B & 96.03 & 94.52 & 95.07 & 92.67 & 80.73 && 57.61 & 56.06 & 54.01 & 41.73 \\
\rowcolor{lightgrey} Gemma 3 27B & 84.48 & 84.20 & 85.36 & 87.06 & 84.59 && 49.37 & 49.92 & 50.31 & 48.60 \\
\rowcolor{lightgrey} Mistral Small 3.1 24B & 96.05 & 95.06 & 93.77 & 92.42 & 88.80 && 49.41 & 49.71 & 47.46 & 43.34 \\
\rowcolor{lightgrey} Apertus 70B & 91.52 & 84.26 & 80.54 & 76.82 & 60.33 && 44.72 & 44.60 & 41.07 & 35.67 \\
\rowcolor{ai2lightpink} Olmo 3 32B & 96.10 & 94.57 & 90.42 & 86.22 & 79.70 && 52.11 & 49.36 & 48.60 & 43.15 \\
\bottomrule
\end{tabular}
\caption{\textbf{Performance of \olmothree compared to other open base models of comparable size}. During \olmothree development, we use RULER~\citep{hsieh2024rulerwhatsrealcontext} as our development suite; we hold HELMET~\citep{yen2025helmet} out as an unseen evaluation suite.
The table contains base variants of each model; models are sorted by their respective release dates.
Qwen 3 8B Base~\citep{qwen3} and Xiaomi MiMo 7B~\citep{mimo} only support a context length of up to $32{,}768$ tokens.
We exclude any base model that does not support at least $32{,}768$ tokens.
}
\label{tab:ruler_baselines}
\end{table}

\subsubsection{Sourcing Long Context Data}
\label{lc:data}

\begin{figure}[h!]
  \centering
  \includegraphics[width=\linewidth]{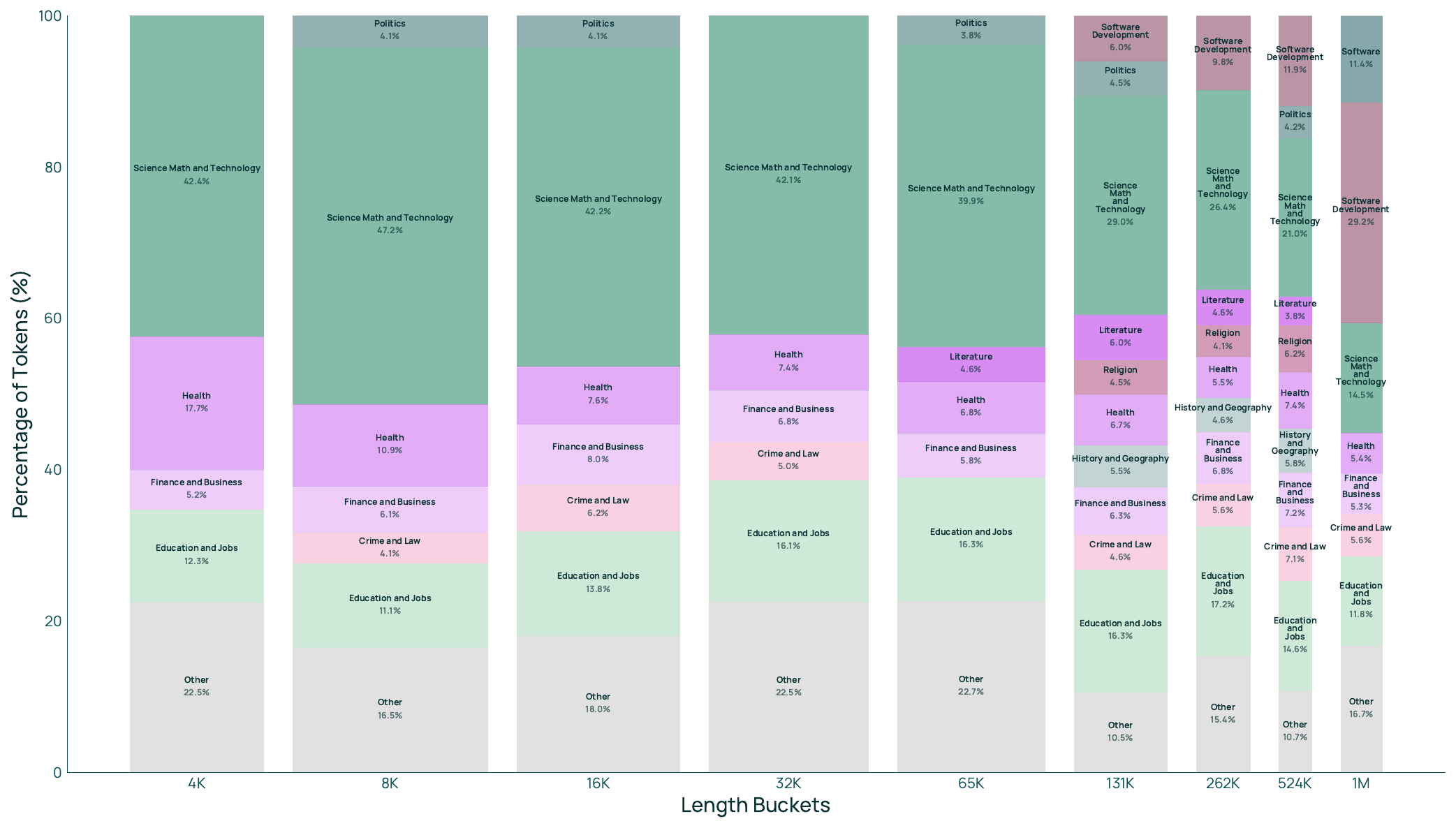}
  \caption{\textbf{Distribution of token counts over WebOrganizer}~\citep{weborganizer} topics in \olmocrPDF, partitioned by length. }
  \label{fig:lc-distribution}
\end{figure}

\paragraph{\olmocrPDF} The backbone of our long-context data pool is scientific PDFs scraped from the web and processed by \olmocr.\footnote{See Section~\ref{sec:preparing-pdf-data} for more details on the preprocessing of this data.}
Figure~\ref{fig:lc-distribution} describes the distribution by topic in each length bucked shown in Table~\ref{table:data-stage-3}.

\paragraph{Data filtering}
We filter this data using \texttt{gzip} compressibility as a metric. \texttt{gzip} has been used for text classification \citep{jiang2022moreparameterfreetextclassification} and as a feature in fine-grained scaling laws \citep{pandey2024gzippredictsdatadependentscaling}. We use \texttt{gzip} for data filtering by excluding the extremes: removing the 20\% of text that is most compressible and the 20\% of text that is least compressible. %

We also consider applying filters based on LongPpl \citep{fang2025wrongperplexitylongcontextlanguage}, which identifies tokens that rely moston  long-range dependencies by measuring, for each token, the change in perplexity under an existing long-context model when additional preceding context is provided. We compute LongPpl over 10B tokens of \longminomix{} using Gemma 3 4B~\citep{team2025gemma3} as the reference model, and comparing contextualization using 4K or 128K context windows. We use the same threshold as \citet{fang2025wrongperplexitylongcontextlanguage} for determining whether a token is a ``key'' token that requires long context dependencies.

We compute two statistics over each document: the fraction of tokens marked as key tokens, and the spread of key tokens across the document (which we compute as the standard deviation of key token locations, which are measured relative to the document length). In a sweep of experiments, we consider excluding the bottom 20\% of documents with the least key tokens or lowest spread, and excluding both the top and bottom 20\% as outliers; none of these possibilities outperform the \texttt{gzip} filter, so we do not use this for the final run.

\subsubsection{Experiments with Synthetic Augmentation}
\label{lc:synth}

A common use case for extended-context language models is information extraction and synthesis over long inputs~\citep{longbench1,longbench2}.
However, most long documents do not provide supervision for such tasks.
Directly inspired by CLIPPER~\citep{pham2025clipper}, we modify a portion of our science PDF pool by injecting synthetically generated aggregation tasks at randomly sampled intervals.
Our approach also shares similarities with Qwen 2.5 1M~\citep{qwen1M}.

\paragraph{Generation~pipeline} The main challenge in generating synthetic data for long-context understanding is the bootstrap problem: how can we create effective data without having access to models that can process long context?
Our pipeline uses document statistics to identify the most important terms and then extracts snippets containing those terms.
Those snippets are subsequently provided to a language model to create aggregation tasks.
In detail:
\begin{enumerate}
    \item For a given document of length $n$ tokens, we partition the document into $m$ sections of length 8K to 32K tokens. We attempt to place these partitions near natural breaks in the document flow, such as right before new sections;
    \item For each partition, we normalize and tokenize the text, extract one- and two-word noun phrases, and use \textit{tf-idf} to identify the most salient noun phrases;
    \item For each noun phrase, we select $k=8$ snippets of text from the partition, ranked by \textit{tf-idf};
    \item We pass the noun phrases, (optional) snippets, and one or more prompts describing the aggregation task to a language model.
\end{enumerate}

For \olmothree, we use documents where $32,768 \leq n < 65,536$ tokens, resulting in 2 to 8 partitions per document.  While we experimented with several closed and open language models, we ultimately use \textsc{\olmotooinstruct~32B} for all generations.

\paragraph{Synthetic~aggregation~tasks} We consider two aggregation tasks; we refer the reader to the code implementation\footnote{\href{https://github.com/allenai/dolma3/blob/f7def5838c8c2d25e358b2b35b2b752168107ed4/datasets/dolma3_longmino_mix/synthetic_cwe_rex/longmino_synthetic_cwe_rex.py}{\path{github.com/allenai/dolma3/datasets/dolma3_longmino_mix/synthetic_cwe_rex/longmino_synthetic_cwe_rex.py}}} for the exact prompts used.

\begin{itemize}
	\item {\bf{CWE}} (Common Word Extraction) We prompt \olmotooinstruct with 5 commonly occurring single-word noun phrases in the partition, and ask the model to generate diverse QA pairs that require the answer to be the exact number of times each unigram occurs in the partition;
	\item {\bf{REX}} (Rewriting EXpressions) For each noun phrase and corresponding snippets, we prompt \olmotooinstruct to generate an aggregation task matching one of the following 12 vignettes discussing the noun phrase: a short summary, a dialogue between a professor and student, a simple paragraph for high school students, a set of flashcards, a school quiz, a game show, a dinner party, a debate, a list of true or false claims, a movie scene, an encyclopedic description, or an explainer in the style of conversations on the \texttt{r/explainlikeimfive} subreddit.
\end{itemize}

\subsubsection{Choosing Data Mix and Token Budget}
\label{lc:mix}

\paragraph{Interleaving long- and short-context data}
Rather than training on only long-context data, we mix high-quality short-context data from midtraining (stage two) to ensure that performance on short-context tasks is not meaningfully degraded.
Early experiments on a 10B-token extension show that a 66\% / 34\% mix of long-context to short-context data drops performance on a subset of \olmothreeeval by 2.5 points; in comparison, a 34\% long-context, 66\% short-context mix only drops performance by 0.8 points.

\paragraph{Longer extension helps} Figure~\ref{lc:component:longer} shows that allocating more tokens to the long-context extension stage improves performance on long-context tasks, particularly at longer sequence lengths.
We extend the context of \olmothree 7B through a 50B stage 3 training;
for \olmothree 32B, we extend for 100B tokens for better long-context capabilities.

\subsubsection{Curating a Training Recipe for Extension}
\label{lc:recipe}

\paragraph{RoPE extension} \olmothree uses RoPE \citep{su2024roformer} to encode positional information within the transformer architecture. We experiment with several methods for extending RoPE beyond the original pretraining context length, including adjusted base frequency scaling \citep{xiong2023effectivelongcontextscalingfoundation, rozière2024codellamaopenfoundation}, position interpolation \citep{chen2023extendingcontextwindowlarge}, and YaRN \citep{peng2023yarnefficientcontextwindow}. Each approach is applied either to all RoPE instances or is restricted to RoPE used in full attention layers. We find that applying YaRN only to full attention layers yields the best overall performance.

\paragraph{Document packing} During pretraining and midtraining, we follow the standard approach of concatenating documents and splitting them into fixed-length training sequences. However, when extending the context length, this strategy produces training instances that are, on average, shorter than the underlying document length distribution. To address this, we adopt best-fit document packing~\citep{ding2024fewertruncationsimprovelanguage}, which reduces the number of split documents while adding a negligible amount of padding. Compared to the naive concatenate-then-split approach, best-fit packing yields substantially improved performance on long-context benchmarks.

\paragraph{Intra-document masking} During long-context extension, we apply intra-document masking to ensure that each training sequence attends only to tokens originating from the same underlying document \citep{zhao2024interdoc,dubey2024llama}. This prevents the model from being distracted by cross-document signals, which can otherwise introduce spurious attention patterns and degrade long-range performance.

\paragraph{LC training infrastructure} To extend the model to a 65K-token context window, we employ 8-way context parallelism (CP) so that each device processes 8K tokens from each training instance. We adopt the all-gather-based CP attention strategy introduced by \cite{scalingllama3}, which makes it straightforward to support irregular attention masks, including sliding-window and intra-document masking. For parallelism configurations, infrastructure details, and throughput measurements, see Appendix Table~\ref{tab:training-config}.

\paragraph{Model souping} Following performance improvements from merging midtraining runs for \olmothreebase 32B, we experiment with averaging long-context checkpoints. In this case, rather than running long-context extension multiple times with different seeds, we merge the last three checkpoints from the end of the extension run (at steps 10,000, 11,000, and 11,921) to produce our final long-context \olmothreebase 32B.

\subsection{Base Model Results}
\label{subsec:pretrain_base_model_results}
\begin{table}[ht!]
\setlength\tabcolsep{2pt} 
\renewcommand{\arraystretch}{0.9}
{\fontsize{8}{8}\selectfont
\begin{center}
\begin{footnotesize}
\begin{tabular}{lc|ccccc|cccc}
\toprule
    &\multicolumn{6}{c}{\quad \quad \quad \quad \textbf{\texttt{Base Aggregate Scores}}} & \multicolumn{4}{c}{\textbf{\texttt{Select Base Benchmarks}}} \\
    {\textbf{\fontsize{8}{8}\selectfont~Model}} &
    {\textbf{\fontsize{8}{8}\selectfont~$\#$ Toks}} & 
    {\textbf{\fontsize{8}{8}\selectfont~Math}} & 
    {\textbf{\fontsize{8}{8}\selectfont~Code}} & 
    {$\textbf{\fontsize{8}{8}\selectfont~MC}_\textbf{\fontsize{6}{6}\selectfont~STEM}$} & 
    {$\textbf{\fontsize{8}{8}\selectfont~MC}_\textbf{\fontsize{6}{6}\selectfont~Non-STEM}$} & 
    {\textbf{\fontsize{8}{8}\selectfont~GenQA}} & 
    {\textbf{\fontsize{8}{8}\selectfont~Minerva}} &
    {\textbf{\fontsize{8}{8}\selectfont~GenXL}} &
    {\textbf{\fontsize{8}{8}\selectfont~MMLU}} & 
    {\textbf{\fontsize{8}{8}\selectfont~BCB}}
    \\
\midrule

\rowcolor{midgrey} \multicolumn{11}{c}{\textbf{7B scale}} \\
\rowcolor{lightgrey}    \olmotoo 7B Stage 1              & 4T & 12.7 & 7.1 & 61.0 & 70.6 & 68.6 & 5.6 & 15.8 & 59.8 & 81.6 \\
\rowcolor{lightgrey}    \olmotoo 7B Stage 2 Ingredient 1 & 4.05T & 40.4 & 10.4 & 64.1 & 74.6 & 72.1 & 18.9 & 21.3 & 63.1 & 85.1 \\
\rowcolor{lightgrey}    \olmotoo 7B Stage 2 Ingredient 2 & 4.05T & 41.4 & 10.4 & 64.3 & 74.9 & 71.8 & 18.7 & 21.0 & 63.8 & 85.8 \\
\rowcolor{lightgrey}    \olmotoo 7B Stage 2 Ingredient 3 & 4.05T & 40.8 & 10.1 & 64.0 & 74.9 & 72.1 & 19.1 & 21.9 & 63.8 & 85.6 \\
\rowcolor{lightgrey}    \olmotoo 7B Stage 2 Soup         & 4.15T & 41.7 & 10.4 & 64.6 & 75.2 & 72.4 & 19.1 & 21.2 & 63.7 & 85.7 \\
\rowcolor{lightgrey}    Apertus 8B Phase 3               & 12T & 19.2 & 9.9 & 61.1 & 68.4 & 68.3 & 7.3 & 19.0 & 58.3 & 81.4 \\
\rowcolor{lightgrey}    Apertus 8B Phase 4               & 13.5T & 26.0 & 16.2 & 65.1 & 73.8 & 69.7 & 10.8 & 30.5 & 63.3 & 86.8 \\
\rowcolor{lightgrey}    Apertus 8B Phase 5               & 15T & 29.3 & 19.0 & 66.7 & 75.0 & 70.1 & 12.9 & 31.0 & 65.0 & 88.6 \\
\rowcolor{lightgrey}    Marin 8B Phoenix                 & 11.1T & 11.2 & 8.0 & 60.9 & 71.1 & 68.7 & 4.7 & 15.0 & 58.5 & 83.1 \\
\rowcolor{lightgrey}    Marin 8B Starling                & 12.4T & 40.5 & 20.8 & {\bf{68.3}} & 78.7 & 75.7 & 23.2 & 36.2 & {\bf{67.8}} & 89.1 \\
\rowcolor{lightgrey}    Marin 8B Deeper Starling         & 12.7T & 39.4 & 21.3 & 68.1 & {\bf{78.8}} & {\bf{75.9}} & 23.9 & 37.0 & 67.7 & 89.2 \\
\rowcolor{ai2lightpink} \olmothree 7B Stage 1            & 5.9T & 23.5 & 19.8 & 64.0 & 71.9 & 68.5 & 12.2 & 34.7 & 62.3 & 84.8 \\
\rowcolor{ai2lightpink} \olmothree 7B Stage 2            & 6T & {\bf{59.8}} & {\bf{31.9}} & 67.2 & 78.2 & 71.3 & {\bf{41.4}} & {\bf{49.1}} & 66.9 & {\bf{89.7}} \\
\rowcolor{ai2lightpink} \olmothree 7B Stage 3            & 6.05T & 54.4 & 30.6 & 66.4 & 78.2 & 72.5 & 39.8 & 43.6 & 66.9 & 89.2 \\

\rowcolor{midgrey} \multicolumn{11}{c}{\textbf{32B scale}} \\
\rowcolor{lightgrey}    \olmotoo 32B Stage 1                & 6.5T & 33.2 & 16.0 & 73.0 & 81.7 & 75.8 & 13.6 & 29.2 & 72.3 & 93.5 \\
\rowcolor{lightgrey}    \olmotoo 32B Stage 2 Ingredient 1   & 6.6T & 51.6 & 19.9 & 75.1 & 84.5 & 78.5 & 30.3 & 36.8 & 75.5 & 94.8 \\
\rowcolor{lightgrey}    \olmotoo 32B Stage 2 Ingredient 2   & 6.6T & 51.9 & 20.0 & 74.1 & 83.8 & 79.1 & 30.7 & 35.2 & 74.0 & 94.1 \\
\rowcolor{lightgrey}    \olmotoo 32B Stage 2 Ingredient 3   & 6.6T & 51.5 & 19.6 & 74.4 & 83.6 & 79.0 & 29.2 & 35.7 & 74.3 & 93.8 \\
\rowcolor{lightgrey}    \olmotoo 32B Stage 2 Ingredient 4   & 6.8T & 51.9 & 19.2 & 74.6 & 83.3 & 78.3 & 31.0 & 37.1 & 74.3 & 94.0 \\
\rowcolor{lightgrey}    \olmotoo 32B Stage 2 Soup           & 7.1T & 53.9 & 20.5 & 75.3 & 84.2 & 79.1 & 31.0 & 37.1 & 75.0 & 94.4 \\
\rowcolor{lightgrey}    Apertus 70B Phase 3                 & 12T & 34.2 & 17.8 & 68.6 & 78.2 & 74.6 & 13.4 & 31.9 & 67.3 & 88.8 \\
\rowcolor{lightgrey}    Apertus 70B Phase 4                 & 13.5T & 39.8 & 21.5 & 70.5 & 79.5 & 75.8 & 16.3 & 34.8 & 69.5 & 91.0 \\
\rowcolor{lightgrey}    Apertus 70B Phase 5                 & 15T & 40.6 & 23.0 & 70.5 & 79.4 & 75.5 & 17.5 & 37.7 & 69.3 & 91.4 \\
\rowcolor{lightgrey}    K2 V2 70B Pretrain                  & 12.3T & 46.1 & 35.4 & 75.6 & 83.5 & 77.1 & 27.2 & 54.5 & 75.2 & 93.0 \\
\rowcolor{lightgrey}    K2 V2 70B Stage 1                   & 14.0T & 60.1 & 37.4 & 74.9 & 84.2 & 69.4 & 38.6 & 56.3 & 74.9 & 93.1 \\
\rowcolor{lightgrey}    K2 V2 70B Stage 2                   & 14.6T & 60.9 & 36.6 & 75.0 & 84.1 & 71.8 & 38.6 & 55.9 & 74.7 & 92.9 \\
\rowcolor{lightgrey}    K2 V2 70B Stage 3                   & 14.8T & 69.5 & 38.0 & 76.1 & 84.1 & 75.3 & 47.9 & 58.5 & 75.5 & 93.3 \\
\rowcolor{lightgrey}    K2 V2 70B Stage 4                   & 15T & 72.8 & 38.3 & 75.7 & 84.0 & 75.6 & 50.0 & 55.7 & 75.6 & 93.5 \\
\rowcolor{lightgrey}    Marin 32B Phase 3                   & 5.4T & 25.8 & 13.9 & 70.4 & 80.2 & 75.1 & 9.7 & 19.6 & 69.5 & 90.8 \\
\rowcolor{lightgrey}    Marin 32B Mantis                    & 6.5T & 49.3 & 30.8 & {\bf{75.9}} & 84.5 & {\bf{80.3}} & 36.8 & 52.1 & 75.7 & 93.4 \\
\rowcolor{ai2lightpink} \olmothree 32B Stage 1              & 5.5T & 48.4 & 29.8 & 72.3 & 80.6 & 76.1 & 26.7 & 47.8 & 71.7 & 92.6 \\
\rowcolor{ai2lightpink} \olmothree 32B Stage 2 Ingredient 1 & 5.6T & 66.8 & 38.4 & 74.6 & 85.6 & 78.9 & 46.5 & 59.6 & 75.9 & 94.7 \\
\rowcolor{ai2lightpink} \olmothree 32B Stage 2 Ingredient 2 & 5.6T & 65.4 & 39.3 & 74.8 & 85.0 & 78.9 & 44.1 & {\bf{60.0}} & 76.3 & 94.3 \\
\rowcolor{ai2lightpink} \olmothree 32B Stage 2 Soup         & 5.7T & {\bf{69.7}} & {\bf{39.7}} & 75.6 & {\bf{85.7}} & 79.4 & {\bf{46.9}} & 59.7 & {\bf{76.9}} & {\bf{95.0}} \\
\rowcolor{ai2lightpink} \olmothree 32B Stage 3              & 6.2T & 61.4 & {\bf{39.7}} & 74.3 & 85.6 & 79.7 & 42.9 & 59.4 & 76.2 & 94.8 \\

\bottomrule
\end{tabular}
\end{footnotesize}
\vspace{2mm}
\caption{
\textbf{Results comparing \olmothree to open base models across stages of pretraining, midtraining and long context}.
As of writing, Marin has undergone learning rate cooldown (Mantis), but not long-context (LC) extension stage.
Apertus also has a two-stage cooldown (Phase 4 and 5) and performed long-context extension by mixing-in data to their Phase 5 training. Token counts are presented in "Cumulative training tokens", so each row denotes the number of tokens that model has seen up to that point in training. For \olmotoo and \olmothree models, Stage 1 is the standard pretraining phase, Stage 2 is midtraining, and Stage 3 is LC extension. 
}
\label{tab:base_evals_overview}
\end{center}
}
\end{table}

In Table \ref{tab:base_evals_overview} we outline the results of \olmothreebase after the pretraining, midtraining, and long-context extension stages, comparing performance to other open base models. Compared to \olmotoo, the \olmothree models demonstrate clear improvements on science, math, and code-based evaluation metrics, which we attribute largely to our emphasis and upsampling of STEM-related data during the pretraining and midtraining stages. On the other hand, because of this emphasis on STEM, we see slight degradation in general knowledge benchmarks.

\newpage
\section{\olmothreethinking}
\label{sec:posttrain-thinking}
We train \olmothreethinking{} to reason by first generating extended thought sequences and then producing a final answer (Figure~\ref{fig:olmo3_pipeline}).
To achieve this, we curate high-quality reasoning data ({\bf \dolcithink}), harness a three-stage training recipe (SFT, DPO, and RLVR), and introduce {\bf \olmothreerl} approach, which merges our new algorithmic and engineering advances with a strong community platform of research in reinforcement learning with verifiable rewards.

Through these data, training, and algorithmic innovations, \olmothreethinking{} achieves strong performance in math, coding, reasoning, and general conversation.
At the 32B scale, it stands as the best fully-open thinking model, outperforming Qwen 2.5 32B, Gemma 2 and 3 27B, and narrowing the gap to top open-weight systems like Qwen 3 32B while being trained on fewer FLOPs (\autoref{tab:32b-think-baeslines}).

\begin{enumerate}
\item {\bf Data: \dolcithink} 
Building on prior open-source datasets~\citep{guha2025openthoughts,lambert2024tulu3,primeintellect2025synthetic2} \emph{inter alia}, we introduce \dolcithinksft, \dolcithinkdpo, and \dolcithinkrl, new cutting-edge post-training datasets designed to target a broad range of key capabilities such as math, coding, instruction following, and general conversation.
The dataset includes synthetic examples with long thinking traces for supervised finetuning, high-contrast paired data for contrastive learning via preference optimization, and challenging prompts for reinforcement learning across diverse domains. Our data curation pipeline is shown in Figure \ref{fig:posttrain-data-pipeline}.
\item {\bf Three-Stage training recipe}
We employ a three-stage post-training process comprising Supervised Finetuning (SFT), Preference Finetuning via Direct Preference Optimization (DPO), and then Reinforcement Learning with Verifiable Rewards (RLVR).
We observe consistent gains across all three stages, demonstrating the impact of careful data curation, algorithmic refinement, and infrastructure development. This contrasts with most recent prior work on open thinking models, which typically employs only a subset of these training stages.\footnote{More concretely, OpenThought3 and S1 only used supervised finetuning; SmolLM used SFT and DPO, but did not apply RL.} For example, we find that our RL framework yields greater improvements when applied after contrastive learning with DPO rather than directly following SFT (\autoref{fig:dpo_sft_rl_comparison}).
\item {\bf \olmothreerl} We present \olmothreerl, our RL training approach which builds upon GRPO and extends it with improvements from recent work. Additionally, we expand verifiable reasoning to multiple domains, going beyond the math and code settings typically explored in prior work. \olmothreerl enables longer and more stable RL runs across diverse domains and increases the overall efficiency of training cycles (Section~\S\ref{sec:thinking_rl_recipe}).
\end{enumerate}

\subsection{Main Results for \olmothreethinking}
\label{sec:posttrain_eval}

\subsubsection{Evaluation Details}
We establish a suite of benchmarks to evaluate \olmothree post-trained models on math, reasoning, coding, precise instruction following, question answering, knowledge recall, and general chat. We expand upon the evaluation suite of \olmotoo~\citep{olmo20242olmo2furious} by adding new, more challenging benchmarks and removing saturated or noisy ones. 
Table~\ref{tab:task-details-chat} shows our evaluation benchmarks and describes the task configurations and metrics for the \olmothree post-training evaluation suite. 
We establish a standard evaluation configuration between all baseline models, including thinking and instruct models, to simplify comparisons. 
Namely, we follow \citet{guo2025deepseek,adler2024nemotron,qwen3} and use a 32K max context length, a sampling temperature of 0.6 and top-p of 0.95.
Note, some models likely perform better with a higher inference budget, for instance K2 V2 \citep{k2team2025k2v2360openreasoningenhancedllm} use a 128K sequence length.
Further details of our evaluation settings are provided in Appendix~\ref{appx:eval-details}.

Evaluation with reasoning models is both computationally expensive and often high variance.
In our development of our recipe on versions of our 7B model---i.e., before the hyperparameter sweeps for final models---we find that evaluation costs between 10 and 20\% of our compute budget.
When compiling results, we measure the variance of every evaluation in our suite by taking the mean of the standard deviation from 3 runs of 14 models (both baselines and our final models).
By taking the variance per model and then the average variance per evaluation, we can bucket the evaluations by their variance.
We partition evaluations based on their variance as follows:
\begin{itemize}
    \item {\bf{High variance}}:
GPQA: 1.4798,
AlpacaEval 3: 1.2406,
IFEval: 0.8835.
\item {\bf{Stable}}:
ZebraLogic: 0.5638,
Omega: 0.5579,
AIME 24 (Avg@32): 0.5437,
HumanEvalPlus: 0.4615,
AgiEval: 0.4339,
BigBenchHard: 0.3866.
\item {\bf{Very stable}}:
LiveCodeBench (Avg@10): 0.2852,
MBPPPlus: 0.2749,
MATH: 0.2522,
MMLU: 0.2219,
PopQA: 0.1554.
\end{itemize}

\begin{table}[t]
\centering
\footnotesize
\setlength\tabcolsep{5pt}
\renewcommand{\arraystretch}{0.95}
\adjustbox{max width=\linewidth}{
{\fontsize{8}{8}\selectfont
\begin{NiceTabular}{@{}Hl
C{35pt}C{35pt}C{35pt}P{35pt}|
C{35pt}C{35pt}C{35pt}C{35pt}@{}}
\toprule
& & \multicolumn{4}{c}{\quad \quad \textbf{\texttt{Olmo 3 32B Think}}} & \multicolumn{4}{c}{\textbf{\texttt{Baselines}}} \\
\textbf{Skill} &
& \textbf{SFT} & \textbf{DPO} & \textbf{Final Think 3.0} & \textbf{Final Think 3.1} & \textbf{Qwen 3 32B} & \textbf{Qwen 3 VL 32B Think} & \textbf{DS-R1 32B} & \textbf{K2-V2 70B Instruct}\\
\midrule

\rowcolor{midgrey} -- & \textbf{Math} & & & & & & & &\\
\rowcolor{lightgrey} -- & \metric{MATH} & 95.6 & 95.9 & 96.1 & 96.2 & 95.4 & 96.7 & 92.6 & 94.5 \\
\rowcolor{lightgrey} -- & \metric{AIME 2024} & 73.5 & 76.0 & 76.8 & 80.6 & 80.8 & 86.3 & 70.3 & 78.4 \\
\rowcolor{lightgrey} -- & \metric{AIME 2025} & 66.2 & 70.7 & 72.5 & 78.1 & 70.9 & 78.8 & 56.3 & 70.3 \\
\rowcolor{lightgrey} -- & \metric{OMEGA} & 43.1 & 45.2 & 50.6 & 53.4 & 47.7 & 50.8 & 38.9 & 46.1 \\
\midrule

\rowcolor{midgrey} -- & \textbf{Reasoning} & & & & & & & &\\
\rowcolor{lightgrey} -- & \metric{BigBenchHard} & 88.8 & 89.1 & 89.8 & 88.6 & 90.6 & 91.1 & 89.7 & 87.6 \\
\rowcolor{lightgrey} -- & \metric{ZebraLogic} & 70.5 & 74.5 & 76.0 & 80.1 & 88.3 & 96.1 & 69.4 & 79.2 \\
\rowcolor{lightgrey} -- & \metric{AGI Eval English} & 85.9 & 87.8 & 88.2 & 88.8 & 90.0 & 92.2 & 88.1 & 89.6\\
\midrule

\rowcolor{midgrey} -- & \textbf{Coding} & & & & & &  & &\\
\rowcolor{lightgrey} -- & \metric{HumanEvalPlus} & 90.0 & 91.6 & 91.4 & 91.5 & 91.2 & 90.6 & 92.3 & 88.0 \\
\rowcolor{lightgrey} -- & \metric{MBPP+} & 66.7 & 67.2 & 68.0 & 68.3 & 70.6 & 66.2 & 70.1 & 66.0 \\
\rowcolor{lightgrey} -- & \metric{LiveCodeBench v3} & 75.8 & 81.9 & 83.5 & 83.3 & 90.2 & 84.8 & 79.5 & 78.4 \\
\midrule

\rowcolor{midgrey} -- & \textbf{IF} & & & & & & & &\\
\rowcolor{lightgrey} -- & \metric{IFEval} & 83.9 & 80.6 & 89.0 & 93.8 & 86.5 & 85.5 & 78.7 & 68.7\\
\rowcolor{lightgrey} -- & \metric{IFBench} & 37.0 & 34.4 & 47.6 & 68.1 & 37.3 & 55.1 & 23.8 & 46.3 \\
\midrule

\rowcolor{midgrey} -- & \textbf{Knowledge \& QA} & & & & & & & &\\
\rowcolor{lightgrey} -- & \metric{MMLU} & 85.3 & 85.2 & 85.4 & 86.4 & 88.8 & 90.1 & 88.0 & 88.4 \\
\rowcolor{lightgrey} -- & \metric{PopQA} & 33.1 & 37.0 & 31.9 & 30.9 & 30.7 & 32.2 & 26.7 & 32.2 \\
\rowcolor{lightgrey} -- & \metric{GPQA} & 55.7 & 57.6 & 58.1 & 56.7 & 67.3 & 67.4 & 61.8 & 64.0 \\
\midrule

\rowcolor{midgrey} -- & \textbf{Chat} & & & & & & & &\\
\rowcolor{lightgrey} -- & \metric{AlpacaEval 2 LC} & 69.1 & 78.6 & 74.2 & 69.1 & 75.6 & 80.9 & 26.2 & - \\
\midrule

\rowcolor{midgrey} -- & \textbf{Safety} & 64.8 & 65.3 & 68.8 & 83.6 & 69.0 & 82.7  & 63.6 & 88.5 \\
\bottomrule

\end{NiceTabular}}}
\caption{\textbf{Results on our flagship model \olmothreethinking~32B} on our post-training evaluation suite. \olmothreeonethink 32B is the best fully-open model at 32B.
}
\label{tab:32b-think-baeslines}
\end{table}

\begin{table}[t]
\centering
\footnotesize
\setlength\tabcolsep{5pt}
\renewcommand{\arraystretch}{0.95}
\adjustbox{max width=\linewidth}{
{\fontsize{8}{8}\selectfont
\begin{NiceTabular}{@{}Hl
C{35pt}C{35pt}P{35pt}|
C{35pt}C{35pt}C{35pt}C{35pt}C{35pt}C{35pt}@{}}
\toprule
& & \multicolumn{3}{c}{\quad \quad \textbf{\texttt{Olmo 3 7B Think}}} & \multicolumn{6}{c}{\textbf{\texttt{Baselines}}} \\
\textbf{Skill} &
& \textbf{SFT} & \textbf{DPO} & \textbf{Final Think} & \textbf{OpenThinker3 7B} & \textbf{Nemotron Nano 9B v2} & \textbf{DS-R1 Qwen 7B} & \textbf{Qwen 3 8B} & \textbf{Qwen 3 VL 8B Think} & \textbf{OR Nemotron 7B} \\
\midrule

\rowcolor{midgrey} -- & \textbf{Math} & & & & & & & & & \\
\rowcolor{lightgrey} -- & \metric{MATH} & 94.4 & 92.4 & 95.1 & 94.5 & 94.4 & 87.9 & 95.1 & 95.2 & 94.6 \\
\rowcolor{lightgrey} -- & \metric{AIME 2024} & 69.6 & 74.6 & 71.6 & 67.7 & 72.1 & 54.9 & 74.0 & 70.9 & 77.0 \\
\rowcolor{lightgrey} -- & \metric{AIME 2025} & 57.6 & 62.7 & 64.6 & 57.2 & 58.9 & 40.2 & 67.8 & 61.5 & 73.1 \\
\rowcolor{lightgrey} -- & \metric{OMEGA} & 37.8 & 40.5 & 45.0 & 38.4 & 42.4 & 28.5 & 43.4 & 38.1 & 43.2 \\
\midrule

\rowcolor{midgrey} -- & \textbf{Reasoning} & & & & & & & & & \\
\rowcolor{lightgrey} -- & \metric{BigBenchHard} & 84.1 & 83.7 & 86.6 & 77.1 & 86.2 & 73.5 & 84.4 & 86.8 & 81.3 \\
\rowcolor{lightgrey} -- & \metric{ZebraLogic} & 57.9 & 60.6 & 66.5 & 34.9 & 60.8 & 26.1 & 85.2 & 91.2 & 22.4 \\
\rowcolor{lightgrey} -- & \metric{AGI Eval English} & 77.2 & 79.1 & 81.5 & 78.6 & 83.1 & 69.5 & 87.0 & 90.1 & 81.4 \\
\midrule

\rowcolor{midgrey} -- & \textbf{Coding} & & & & & & & & & \\
\rowcolor{lightgrey} -- & \metric{HumanEvalPlus} & 88.2 & 91.4 & 89.9 & 87.4 & 89.7 & 83.0 & 80.2 & 83.7 & 89.7 \\
\rowcolor{lightgrey} -- & \metric{MBPP+} & 63.2 & 63.0 & 64.7 & 61.4 & 66.1 & 63.5 & 69.1 & 63.0 & 61.2 \\
\rowcolor{lightgrey} -- & \metric{LiveCodeBench v3} & 67.8 & 75.1 & 75.2 & 68.0 & 83.4 & 58.8 & 86.2 & 85.5 & 82.3 \\
\midrule

\rowcolor{midgrey} -- & \textbf{IF} & & & & & & & & & \\
\rowcolor{lightgrey} -- & \metric{IFEval} & 77.9 & 75.9 & 88.2 & 51.7 & 86.0 & 59.6 & 87.4 & 85.5 & 42.5 \\
\rowcolor{lightgrey} -- & \metric{IFBench} & 30.0 & 28.3 & 41.6 & 23.0 & 34.6 & 16.7 & 37.1 & 40.4 & 23.4 \\
\midrule

\rowcolor{midgrey} -- & \textbf{Knowledge \& QA} & & & & & & & & & \\
\rowcolor{lightgrey} -- & \metric{MMLU} & 74.9 & 74.8 & 77.8 & 77.4 & 84.3 & 67.9 & 85.4 & 86.5 & 80.7 \\
\rowcolor{lightgrey} -- & \metric{PopQA} & 20.8 & 24.7 & 23.7 & 18.0 & 17.9 & 12.8 & 24.3 & 29.3 & 14.5 \\
\rowcolor{lightgrey} -- & \metric{GPQA} & 45.8 & 48.6 & 46.2 & 47.6 & 56.2 & 54.4 & 57.7 & 61.5 & 56.6 \\
\midrule

\rowcolor{midgrey} -- & \textbf{Chat} & & & & & & & & & \\
\rowcolor{lightgrey} -- & \metric{AlpacaEval 2 LC} & 43.9 & 50.6 & 52.1 & 24.0 & 58.0 & 7.7 & 60.5 & 73.5 & 8.6 \\
\midrule

\rowcolor{midgrey} -- & \textbf{Safety} & 65.8 & 67.7 & 70.7 & 31.6 & 72.1 & 54.0 & 68.3 & 82.9 & 30.3 \\
\bottomrule

\end{NiceTabular}}}
\caption{\textbf{Overview of results of \textbf{\olmothreethinking 7B} on our post-training evaluation suite.} All numbers are the mean of three runs. We evaluate all models using our evaluation framework, generating up to a maximum of 32768 tokens.}
\label{tab:post-train-eval-overview}
\end{table}

\begin{table*}[t]
  \centering

\begin{scriptsize}
\begin{tabular}{HlHHllllllHll} %
\toprule
& \textbf{Task} & \textbf{Capability} & \textbf{\# Inst} & \textbf{Format} & \textbf{Metric} & \textbf{Temp} & \textbf{Top-p} & \textbf{Ans. Extract} & \textbf{Max Toks} & \textbf{P@k (N)} & \textbf{N} & \textbf{\# Sub} \\
\midrule

\rowcolor{midgrey}\multicolumn{13}{c}{\rule{0pt}{1pt}} \\[-9pt]
\rowcolor{midgrey}\multicolumn{13}{c}{\textbf{Chat Suite}} \\
\rowcolor{midgrey}\multicolumn{13}{c}{\rule{0pt}{1pt}} \\[-9pt]

\rowcolor{lightgrey} & IF Eval (\citeyear{zhou2023instructionfollowingevaluationlargelanguage}) & Instruction Following & - & CoT & Custom & 0.6 & 0.95 & Custom & 32768 & - & 1 & - \\ 
\rowcolor{lightgrey} & Minerva MATH (\citeyear{lewkowycz2022solving}) & Math Gen & - & CoT EM & EM Flex & 0.6 & 0.95 & Minerva & 32768 & - & 1 & 7 \\
\rowcolor{lightgrey} & MATH 500 (\citeyear{lewkowycz2022solving,lightman2023lets}) & Math Gen & - & CoT EM & EM Flex & 0.6 & 0.95 & Minerva & 32768 & - & 1 & - \\
\rowcolor{lightgrey} & AIME 2024* & Math Gen & - & CoT EM & EM Flex & 0.6 & 0.95 & Minerva & 32768 & - & 32 & - \\
\rowcolor{lightgrey} & AIME 2025* & Math Gen & - & CoT EM & EM Flex & 0.6 & 0.95 & Minerva & 32768 & - & 32 & - \\
\rowcolor{lightgrey} & Omega Math (\citeyear{Sun2025OMEGACL}) & Math Gen & - & CoT EM & EM Flex & 0.6 & 0.95 & Custom Regexes & 32768 & - & 1 & 55 \\
\rowcolor{lightgrey} & HumanEval+ (\citeyear{evalplus}) & Code Gen & - & CoT Code & pass@1 & 0.6 & 0.95 & Split on \texttt{```} & 32768 & - & 10 & - \\
\rowcolor{lightgrey} & MBPP+* (\citeyear{evalplus}) & Code Gen & - & CoT Code & pass@1 & 0.6 & 0.95 & Split on \texttt{```} & 32768 & - & 10 & - \\
\rowcolor{lightgrey} & LiveCodeBench v3* (\citeyear{jain2024livecodebench}) & Code Gen & - & CoT Code & pass@1 & 0.6 & 0.95 & Split on \texttt{```} & 32768 & - & 10 & - \\
\rowcolor{lightgrey} & ZebraLogic* (\citeyear{lin2025zebralogic}) & Puzzle Solving & - & CoT JSON & Custom & 0.6 & 0.95 & Custom JSON & 32768 & - & 1 & - \\ 
\rowcolor{lightgrey} & BigBench-Hard (\citeyear{suzgun2022challenging}) & Puzzle Solving & - & CoT EM & EM Flex & 0.6 & 0.95 & \olmothree Regex & 32768 & - & 1 & 23 \\
\rowcolor{lightgrey} & GPQA* (\citeyear{rein2024gpqa}) & General QA & - & CoT MC & Acc & 0.6 & 0.95 & \olmothree Regex & 32768 & - & 1 & - \\
\rowcolor{lightgrey} & AGI Eval* (\citeyear{zhong2023agieval}) & General QA & - & CoT MC & Acc & 0.6 & 0.95 & \olmothree Regex & 32768 & - & 1 & 9 \\
\rowcolor{lightgrey} & MMLU (\citeyear{hendryckstest2021}) & General QA & - & CoT MC & Acc & 0.6 & 0.95 & \olmothree Regex & 32768 & - & 1 & 57 \\
\rowcolor{lightgrey} & PopQA (\citeyear{mallen2023llm_memorization}) & Trivia QA & - & CoT MC & Acc & 0.6 & 0.95 & EM Recall & 32768 & - & 1 & - \\
\rowcolor{lightgrey} & SimpleQA* (\citeyear{wei2024measuring}) & - & - & - & - & - & - & - & - & - & 1 & - \\
\rowcolor{lightgrey} & Alpaca Eval v2 (\citeyear{alpaca_eval,dubois2024length}) & - & - & CoT & Winrate & 0.6 & 0.95 & - & 32768 & - & 1 & - \\
\rowcolor{lightgrey} & BFCL* (\citeyear{patil2025bfcl}) & - & - & - & - & - & - & - & - & - & 1 & - \\
\rowcolor{lightgrey} & LitQA2* (\citeyear{skarlinski2024language}) & - & - & - & - & - & - & - & - & - & 1 & - \\

\bottomrule
\end{tabular}
\end{scriptsize}
  \caption{
  \textbf{Details of the \olmothree chat evaluation suite}. 
  We mark tasks with * to indicate new additions compared to the \olmotoo suite~\citep{olmo20242olmo2furious}. All evaluation generations have thinking traces (text between \texttt{<think>...</think>}) stripped before passing to the answer scorer. We use zero-shot setting for all metrics. 
}
  \label{tab:task-details-chat}
\end{table*}

\subsubsection{Main Results} \label{sec:thinkresults} Table~\ref{tab:32b-think-baeslines} and Table~\ref{tab:post-train-eval-overview} show the performance of \olmothreethinking across different training stages and compare it with other baselines of similar scale on our benchmarks\footnote{Running AlpacaEval on K2-V2-Instruct led to token-parsing errors on the output of the LLM judge, resulting in null preference scores. If we are able to devise a solution, we will update the report accordingly.}. As described before, \olmothreethinking 32B is the best fully-open model at the 32B scale, outperforming other models including Gemma 2 27B, Gemma 3 27B, and Qwen 2.5 32B-Instruct. It narrows the gap to the best open-weight models at this scale, Qwen 3 and Qwen 3VL, while being trained with 6x fewer tokens.   Similarly, \olmothreethinking-7B outperforms OpenReasoning Nemotron 7B, DeepSeek-R1-Distill-Qwen-7B, and OpenThinker-7B, some of the best open-weight thinking models. In addition, it performs similarly to Nemotron-Nano-9B-v2 despite being smaller. At 7B, it lags the Qwen 3 series of models in knowledge tasks. We think that this is mainly due to the fact that Qwen 3 models are trained through distillation from Qwen's largest model. %

Notably, we introduce {\bf \olmothreeonethink 32B} to illustrate that extended \olmothreerl training, via additional epochs on our \dolcithinkrl dataset\footnote{While \olmothreethinking 32B was trained for 750 steps, we continued the run past our initial release, going up to 2300 steps for \olmothreeonethink 32B. We stopped there due to compute limitations, but note that \textit{performance had not yet fully saturated}, suggesting even longer runs could further improve performance.}, leads to improved performance. We observe substantial improvements on math, reasoning, and instruction-following benchmarks, including gains of 4+ points on AIME, 4 points on ZebraLogic, 4 points on IFEval, and 20 points on IFBench, suggesting the additional RL training improves the model's reasoning abilities. Most other benchmarks remain largely unchanged, with the exception of AlpacaEval, where we observe a 5-point drop.

\subsection{Supervised Finetuning with \dolcithinksft }
\label{sec:thinking_sft_recipe}

\begin{figure}
    \centering
    \includegraphics[width=1.0\linewidth]{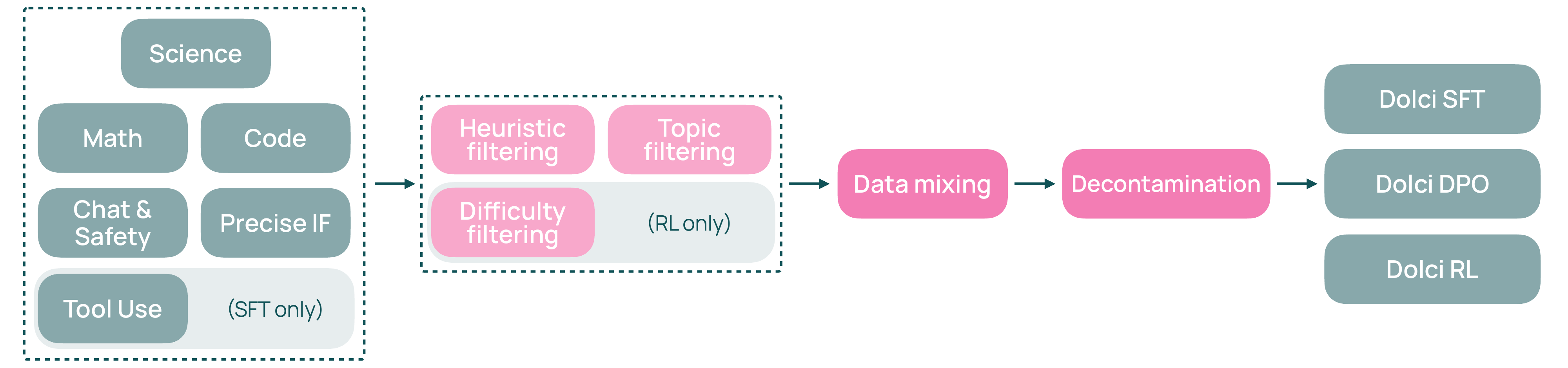}
    \caption{\textbf{Data pipeline for all \olmothree post-training stages.} We share most steps across SFT, DPO and RL to ensure consistent quality.}
    \label{fig:posttrain-data-pipeline}
\end{figure}

In this stage, we construct \dolcithinksft, a resource for finetuning the base model to produce explicit thinking traces that support accurate responses. This supervised finetuning step is especially impactful for smaller models, offering an efficient mechanism for acquiring strong reasoning capabilities. We next detail the \dolcithinksft data curation pipeline (Figure~\ref{fig:posttrain-data-pipeline}).

\subsubsection{\dolcithinksft: Data Curation}
\label{sec:think-sft}

To curate \dolcithinksft, we compile a large collection of prompts across a diverse set of skills from other open efforts (e.g.,~\citealp{guha2025openthoughts, primeintellect2025synthetic2}), substantially filter them, and synthetically generate reasoning traces for their completions.
An overview of the \dolcithinksft data mix is shown in Table~\ref{tab:olmo3_thinking_sft} and is described below:

\begin{table}[t!]
\centering
\setlength\tabcolsep{5pt}
{\small
\begin{tabular}{ll r r l}
\toprule
{\bf Category} & {\bf Prompt Dataset}  & {\bf 7B Count} & {\bf 32B Count} & {\bf Reference} \\
\midrule
\rowcolor{ai2offwhite} Chat \&  & WildChat & 83,054 & 76,209 & \citet{zhao2024wildchat} \\ %
\rowcolor{ai2offwhite} Precise IF & OpenAssistant & 6,800 & 6,647 & \citet{kopf2024openassistant} \\ %
\rowcolor{ai2offwhite} & \dolcithink Persona Precise IF & 223,123 & 220,530 & -- \\ %
\rowcolor{ai2offwhite}  & \dolcithink Precise IF & 135,792 & 135,722 & -- \\ %
Math & \dolcithink OpenThoughts 3+ Math$^\Uparrow$ & 752,997 & 752,997 & \citet{guha2025openthoughts} \\
 & \dolcithink OpenThoughts 3+ STEM$^\Uparrow$ & 99,269 & 99,268 & \citet{guha2025openthoughts} \\ %
 & SYNTHETIC-2-SFT-Verified & 104,569 & 104,548 & \citet{primeintellect2025synthetic2} \\ %
\rowcolor{ai2offwhite} Coding & Nemotron Post-Training Code   & 113,777 & 113,777 & \citet{nvidia2025nemotron_post_training_dataset} \\ %
 \rowcolor{ai2offwhite}  & \dolcithink OpenThoughts 3+ Code$^\Uparrow$ & 88,900 & 88,899 & \citet{guha2025openthoughts} \\ %
\rowcolor{ai2offwhite}  & \dolcithink Python Algorithms$^\Uparrow$ & 466,677 & 466,676 & -- \\ %
Safety & CoCoNot & 10,227 & 9,549 & \citet{brahman2024art} \\ %
 & WildGuardMix & 38,315 & 36,673 & \citet{han2024wildguard} \\ %
 & WildJailbreak & 41,100 & 40,002 & \citet{jiang2024wildteaming} \\ %
\rowcolor{ai2offwhite} Multilingual & Aya & 98,597 & 97,156 & \citet{singh2024aya} \\ %
Other & TableGPT & 4,981 & 4,973 & \citet{zha2023tablegpt} \\ %
& Olmo Identity Prompts & 290 & 290 & -- \\ 
\rowcolor{ai2offwhite} {\bf{Total}} & & 2,268,468 & 2,253,916 & \\
\bottomrule
\end{tabular}}
\vspace{3pt}
\caption{\textbf{\olmothreethinking SFT prompt sources}. %
$^\Uparrow$ indicates prompt datasets where the datasets are upsampled by repeating prompts with different completions. 
Prior to \olmothree 32B training, we filter responses with non-Olmo model identities and irrelevant prompts (e.g. generate a photo).
}
\label{tab:olmo3_thinking_sft}
\end{table}

\paragraph{Step 1: sourcing prompts and generating reasoning traces}
\begin{itemize}

\item{\bf Math} We source prompts from the math subsets of OpenThoughts3~\citep{guha2025openthoughts} and SYNTHETIC-2~\citep{primeintellect2025synthetic2}.
For OpenThoughts3 prompts, we use all the available math prompts (maintaining the 16X repetition from the original) and the available reasoning traces with complete solutions.
For incomplete traces, we generate full reasoning chains and solutions using QwQ-32B, the original model used for the completions, and the same generation settings as OpenThoughts3, except up to 32K tokens instead of the original 16K. We discard any examples that are still incomplete after regenerating. For SYNTHETIC-2, we take completions directly from the verified subsection.

\item {\bf Code} We collect code prompts from different sources and generate completions for them. To create \dolcithink Python Algorithms, we source prompts from AceCoder~\citep{zeng2025acecoder}, the Python subset of The Algorithms~\citep{thealgorithms_python}, Llama Nemotron Post-training~\citep{bercovich2025llamanemotronefficientreasoningmodels}, and OpenCodeReasoning~\citep{ahmad2025opencodereasoning}, and then we generate up to 16 responses per prompt from QwQ-32B, which we filter for correctness using synthetically generated test cases from GPT-4.1. For OpenThoughts 3 code prompts, we downsample each prompt to at most 16 times and regenerate complete responses for all incomplete examples. We combine \dolcithink Python Algorithms with the code prompts from OpenThoughts3, downsample them to 16 repetitions, and regenerate completions for incomplete ones.

\item {\bf Chat \& safety} We source chat prompts from both the T\"ulu 3~\citep{lambert2024tulu3} subset of WildChat~\citep{zhao2024wildchat}, as well as WildChat prompts not used during T\"ulu 3, and the T\"ulu 3 subset of OpenAssistant~\citep{kopf2024openassistant}.
For safety, we reuse safety prompts used during T\"ulu 3.
We then generate reasoning traces and completions from DeepSeek R1~\citep{guo2025deepseek}.

\item {\bf Precise instruction following}
We source precise IF prompts from the overall T\"ulu 3 mix with additional verifiable constraints added from \citet{pyatkin2025generalizing}.
We also regenerate Persona IF prompts as in T\"ulu 3, but with personas sourced from~\citet{nvidia/Nemotron-Personas-USA}.
We then generate responses for each prompt using QwQ-32B, and we verify responses using verifiers associated with each constraint, keeping only the correct responses.

\item {\bf Science \& other}
We source science prompts from the OpenThoughts3 science subset.
For other data sources, we include the TableGPT~\citep{zha2023tablegpt} subset in T\"ulu 3 for data transformation and Aya~\citep{singh2024aya} for chat and basic multilinguality.
We regenerate incomplete responses in OpenThoughts3 as we did for the math and code subsets, and we generate responses with reasoning chains for the other datasets using DeepSeek R1.

\end{itemize}

\paragraph{Step 2: filtering}
\label{sec:topic_filtering}
We perform extensive filtering on the data we have collected and generated.
\begin{itemize}
\item {\bf Heuristic filtering} We filter out examples with (1) non-commercial or unclear licenses, (2) incomplete reasoning chains, (3) domain-specific inaccuracies (i.e., verifying the constraint-adherence of instruction-following data or executing test cases against model completions for code), (4) mentions of other model developers and date cutoffs, (5) excessive repetition, and (6) an excessive number of Chinese characters or Chinese political values reflected in reasoning chains.
\item {\bf Topic filtering}  We classify our dataset by topic using the OpenAI query taxonomy~\citep{Chatterji2025-fs}, and find that filtering out and downsampling topics irrelevant to our model (e.g., requests to generate images or excessive basic greetings) from WildChat qualitatively improves model behavior.\footnote{To evaluate the impact of our filtering process, we manually created an internal benchmark to vibe test the model.} %
See Appendix~\ref{appendix:filtering-think-sft} for detailed descriptions and links to filter scripts.
\end{itemize}

\paragraph{Step 3: data mixing}
For data mixing, we follow a methodology similar to that described in the midtraining section (Section~\S\ref{sec:midtraining}) for parallel data collection, adhering to shared standards for data mixing and conducting multiple rounds of integration testing.
More specifically, we conduct careful experiments using a small ``base'' mix, consisting of 100K examples taken from our extended OpenThought 3 dataset.
We found that this base mix was performant enough on key reasoning benchmarks to serve as a strong baseline, while saving substantial amounts of compute versus training on the full mix.
We then train individual models on the base mix combined with up to 100K training examples (without upsampling) from each category to observe the impact on our evaluation suite.
As shown in Table~\ref{tab:think-sft-ablate}, we generally find that each dataset is helpful on at least one evaluation, and so our final mix includes at least a portion of each dataset we tested.

\paragraph{Step 4: decontamination} We followed the recommended settings from the T\"ulu 3 Decontamination Procedure and toolkit \citep{lambert2024tulu3} to filter out the portions of all post-training data (all three stages) that matched the evaluation sets. We used n-gram matching with 8-grams and an overlap threshold of 0.5 (i.e., at least 50\% of the n-grams in the test instance match a training instance) for filtering. We developed additional heuristics to mitigate false positives: (1) we ignored matches of task-irrelevant chunks of text, e.g., common generic phrases, with the irrelevance determined per task based on manual inspection; (2) particularly in math datasets, we ignored matches of n-grams where most of the tokens are of length 1 (typically math symbols). %

\begin{table}[tbp]
\centering
\small
\resizebox{\textwidth}{!}{
\begin{tabular}{@{}l >{\columncolor{ai2pink!18}}c c c c c c c c c c@{}}
\toprule
& \multicolumn{10}{c}{\textbf{Subset of \olmothreethinking Benchmarks}} \\
\cmidrule(lr){2-11}
\textbf{Name} & \cellcolor{ai2pink!18}\textbf{Avg.} & \textbf{MMLU} & \textbf{BBH} & \textbf{GPQA} & \textbf{Zebra} & \textbf{MATH} & \textbf{CHE} & \textbf{MBPP} & \textbf{AE} & \textbf{IFEval} \\
\midrule
Base mix & 39.2 & 52.4 & 48.7 & 31.0 & 21.0 & 74.6 & 35.4 & 34.7 & 19.0 & 35.7 \\
Base + Aya & 41.9 & 54.4 & 55.7 & 33.9 & 22.7 & 74.0 & 30.5 & 36.0 & 30.2 & 39.6 \\
Base + WildChat and OAsst & 44.2 & 58.3 & 53.3 & 31.7 & 25.8 & 74.0 & 28.7 & 38.4 & 38.5 & 48.8 \\
Base + Persona IF & 45.9 & 64.1 & 55.1 & 31.3 & 25.1 & 74.5 & 25.0 & 33.9 & 34.2 & 70.4 \\
Base + Safety & 40.9 & 53.8 & 49.7 & 30.1 & 22.0 & 74.2 & 31.7 & 33.1 & 33.0 & 40.9 \\
Base + Synthetic 2 & 47.3 & 66.5 & 54.0 & 35.5 & 27.8 & 82.0 & 39.6 & 39.7 & 26.9 & 53.4 \\
Swap base code to Nemotron Code & 34.5 & 48.6 & 43.4 & 33.0 & 19.3 & 74.4 & 22.6 & 26.2 & 16.6 & 26.6 \\
Swap base code to Dolci Python Algorithms & 36.9 & 48.0 & 47.2 & 33.0 & 15.9 & 72.1 & 30.5 & 37.8 & 18.1 & 29.4 \\
\bottomrule
\end{tabular}}
\caption{\textbf{Results of our thinking SFT mixing ablations} on top of an internal \olmotoo long context checkpoint.}
\label{tab:think-sft-ablate}
\end{table}

\subsubsection{Training}

For supervised finetuning, we switch from \openinstruct to \olmocore, resulting in an $8\times$ increase in training throughput. See Appendix \ref{appx:sft-details} for more information about our training settings and hyperparameters. We train all models for two epochs to avoid overfitting, and perform a learning-rate sweep to select the best candidate checkpoints based on our evaluation suite. We then test each candidate checkpoint with a series of qualitative ``vibe-test'' questions to inform our final checkpoint selection. Finally, we explore model souping \citep{wortsman2022model, morrison2024mergelearnefficientlyadding}, and our final thinking SFT checkpoint is a linearly weighted merge of two checkpoints trained with different learning rates, merged with mergekit~\citep{goddard-etal-2024-arcees}.

\subsection{Preference Tuning with Delta Learning} \label{sec:thinking_dpo_recipe}
Prior work in general post-training has positioned preference tuning primarily as a means to improve alignment with human values and preferences~\citep{lambert2024tulu3, rlhf2024}. Hence, most recent efforts in building capability-oriented thinking models~\citep{guha2025openthoughts,ahmad2025opencodereasoning} have not incorporated preference tuning (one exception is SmolLM3;~\cite{bakouch2025smollm3}). We rethink preference tuning as a stage of contrastive learning that drives capability gains beyond what SFT alone can provide. We introduce \dolcithinkdpo, a preference dataset containing completion pairs with clear capability deltas. We leverage these relative contrasts to enhance the model’s reasoning capabilities via preference optimization, extending the ideas from Delta Learning~\citep{geng2025delta}.

In particular, we find that further supervised finetuning on thinking traces generated by Qwen3 32B (one of the few open-thought models) outright hurts the performance of \olmothreethinkingsft, indicating that we are approaching saturation on learning from imitation. To extract a useful training signal out of these now-ineffective completions, we apply Delta Learning's principle by pairing these completions with even \textit{worse} responses~\citep{geng2025delta}; minimizing the quality of the rejected completions (thus increasing the quality delta) yields a useful contrastive signal for preference tuning.

With these insights in mind, we construct \dolcithinkdpo, which we use to improve the model’s performance across a wide range of benchmarks.
We use Direct Preference Optimization (DPO)~\citep{rafailov2024direct} for training with pairwise data.
Details of DPO training are provided in Appendix \ref{appx:dpo-details}.

\paragraph{Delta Learning}
The intuition behind delta-learning is that the quality of preference data depends primarily on the quality of the \textit{delta} between chosen and rejected responses rather than the quality of either response individually.
By constructing preference pairs $(x, y_c, y_r)$ that exhibit capability-relevant contrasts with $y_c \succ y_r$,
tuning to prefer $y_c$ over $y_r$ can improve the model even when supervised finetuning on $y_c$ would not help or even actively hurt~\citep{geng2025delta, d2025anchored, kim2025systematic}.

\subsubsection{\dolcithink-DPO: Preference Data Creation}
\label{sec:dpo_think_data}
To construct \dolcithinkdpo, we compile a large pool of prompts covering a wide range of datasets and skills (see Table \ref{tab:olmo3_think_dpo_mix}) and synthesize chosen and rejected responses to exhibit capability deltas.
Following the delta-learning heuristic \citep{geng2025delta}, for each prompt $x$, we simply decode a chosen completion $y_c$ from one model (Qwen 3 32B, thinking) and a rejected completion $y_r$ from an overall weaker model (Qwen 3 0.6B, thinking) to construct a consistent contrast.\footnote{The UltraFeedback-style LLM-judge preference pipeline employed in \olmotoo and \tulu assumes access to a diverse pool of models to construct preference pairs with useful contrasts; however, there are few open-thought thinking models available to construct such pairs, rendering the \olmotoo pipeline less ideal for this setting. Our \dolciinstructdpo dataset does benefit from model-pool diversity; we are able to further supplement our delta-learning heuristic data with LLM-judged data in \dolciinstructdpo to yield mutually complementary gains (Section~\S\ref{sec:dolci-instruct-dpo}).}

\begin{table}[t!]
\centering
\setlength\tabcolsep{5pt}
{\small
\begin{tabular}{ll R{2.0cm} l}
\toprule
{\bf Category} & {\bf Prompt Dataset} & {\bf \# Prompts used in DPO} & {\bf Reference} \\
\midrule
\rowcolor{ai2offwhite} Chat \&  & WildChat & 40,701 & \citet{zhao2024wildchat} \\
\rowcolor{ai2offwhite} Precise IF & \dolciinstruct Precise IF & 19,365 & -- \\
\rowcolor{ai2offwhite}  & \tulu Persona IF & 3,486 & \cite{lambert2024tulu3} \\
\rowcolor{ai2offwhite}  & OpenAssistant & 1,762 & \citet{kopf2024openassistant} \\
Math & \tulu Persona MATH & 10,657 & \citet{lambert2024tulu3} \\
 & \tulu Persona Algebra & 1,417 & \citet{lambert2024tulu3} \\
 & \tulu Persona GSM & 3,681 & \citet{lambert2024tulu3} \\
 & OpenMathInstruct 2 & 3,615 & \citet{toshniwal2024openmathinstruct} \\
\rowcolor{ai2offwhite} Coding & \dolciinstruct Python Algorithms  & 13,236 & -- \\
 \rowcolor{ai2offwhite}  & \tulu Persona Python & 2,514 & \citet{lambert2023entangled} \\
\rowcolor{ai2offwhite}  & Evol CodeAlpaca & 7,634 & \cite{luo2023wizardcoder} \\
Safety & CoCoNot & 927 & \citet{brahman2024art} \\
 & WildGuardMix & 5,338 & \citet{han2024wildguard} \\
 & WildJailbreak & 5,616 & \citet{jiang2024wildteaming} \\
\rowcolor{ai2offwhite} Science & SciRiff & 2,253 & \citet{wadden2024sciriff} \\
\rowcolor{ai2offwhite} & OpenThoughts3 Science & 19,023 & \cite{guha2025openthoughts} \\
Multilingual & Aya & 4,078 & \citet{singh2024aya} \\
\rowcolor{ai2offwhite} Other & TableGPT & 1,170 & \citet{zha2023tablegpt} \\
\rowcolor{ai2offwhite} & FLAN & 19,660 & \citet{wei2021flan} \\
Not used in SFT & DaringAnteater & 1,089 & \cite{wang2024helpsteer2} \\
& UltraFeedback & 32,778 & \cite{cui2023ultrafeedback} \\
\rowcolor{ai2offwhite} {\bf{Total}} &  & 200,000 & \\ 
\bottomrule
\end{tabular}}
\vspace{3pt}
\caption{\textbf{\olmothreethinking DPO prompt sources}. See Section~\S\ref{sec:dpo_think_data} for data details.}
\label{tab:olmo3_think_dpo_mix}
\end{table}

\paragraph{Step 1: sourcing prompts and contrastive completions}  \olmothreethinking focuses on reasoning capabilities; we thus construct pairs that exhibit a delta in reasoning quality by pairing model completions from models of differing reasoning capability~\citep{geng2025delta, bakouch2025smollm3, kim2023aligning}.
Our prompt pool is derived from the \dolciinstructsft dataset supplemented with the DaringAnteater~\citep{wang2024helpsteer2} and UltraFeedback~\citep{cui2023ultrafeedback} subsets from the \olmotoo 7B preference dataset.

\paragraph{Step 2: filtering}
We apply topic filtering and heuristic model-identity filtering as described from the SFT stage (Section~\S\ref{sec:topic_filtering}) to all \textit{chosen} responses. We leave rejected responses unfiltered with the intuition that an incorrect rejected response may elicit a useful contrast. We further decontaminate all prompts against our evaluation suites.

\paragraph{Step 3: mixing}
Experimentation with long reasoning traces is significantly more expensive than with non-thinking completions.
To obtain the final mix of prompts for {\bf \dolcithinkdpo}, we leverage mixing experiments conducted on prompts with non-thinking completions (see Section~\S\ref{sec:posttrain-instruct} for details).
Specifically, we select the three best-performing prompt distributions from our \olmothreeinstruct experiments and generate chosen and rejected responses for these prompts using the thinking versions of the Qwen models to elicit a delta in reasoning quality. We choose the empirically best-performing mix during our experiments as our final DPO data pool.\footnote{Our \dolciinstructdpo dataset includes additional contrastive pairs, which we obtain through careful experimental analysis. Refer to Section~\S\ref{sec:dolci-instruct-dpo} for more details.}

\subsubsection{Training}
We train all models for one epoch following previous work \citep{lambert2024tulu3}, sweeping learning rate and dataset size to identify the best candidate checkpoints based on our evaluation suite. Dataset size is an important hyperparameter, as we observe that early stopping is important for performant preference tuning; please see our data mixing experiments on our Instruct model (Section~\S\ref{sec:dpo_mixing}) for our motivating results. Beyond our evaluation suite, we further inspect each checkpoint via the same ``vibe-tests'' as in SFT training to qualitatively assess model behavior. See Appendix~\ref{appx:dpo-details} for full training settings.

\subsection{Reinforcement Learning with \olmothreerl: The Cherry on Top} \label{sec:thinking_rl_recipe}

The third stage of post-training is reinforcement learning with a mixture of verifiable and LM-judge rewards across a variety of domains.
We introduce \olmothreerl, which includes our algorithm and closely intertwined engineering infrastructure to address challenges for reinforcement learning with long reasoning traces, extending RLVR to include a wider variety of verifiable tasks.
We also release \textsc{Dolci-Think-RL}---a large-scale and diverse dataset of roughly 100K prompts across four domains: mathematics, coding, instruction following, and general chat---to support robust reinforcement learning on varied reasoning tasks while maintaining general utility.
Next, we describe the RL algorithmic details (\S\ref{olmo3think-rl}), the \dolcithinkrl dataset (\S\ref{subsec:rl-thinking-data}), and finally \olmothreerl infrastructure in \openinstruct (\S\ref{sec:posttraining_infra}).

\begin{figure}[t]
    \centering
    \includegraphics[width=1.0\linewidth]{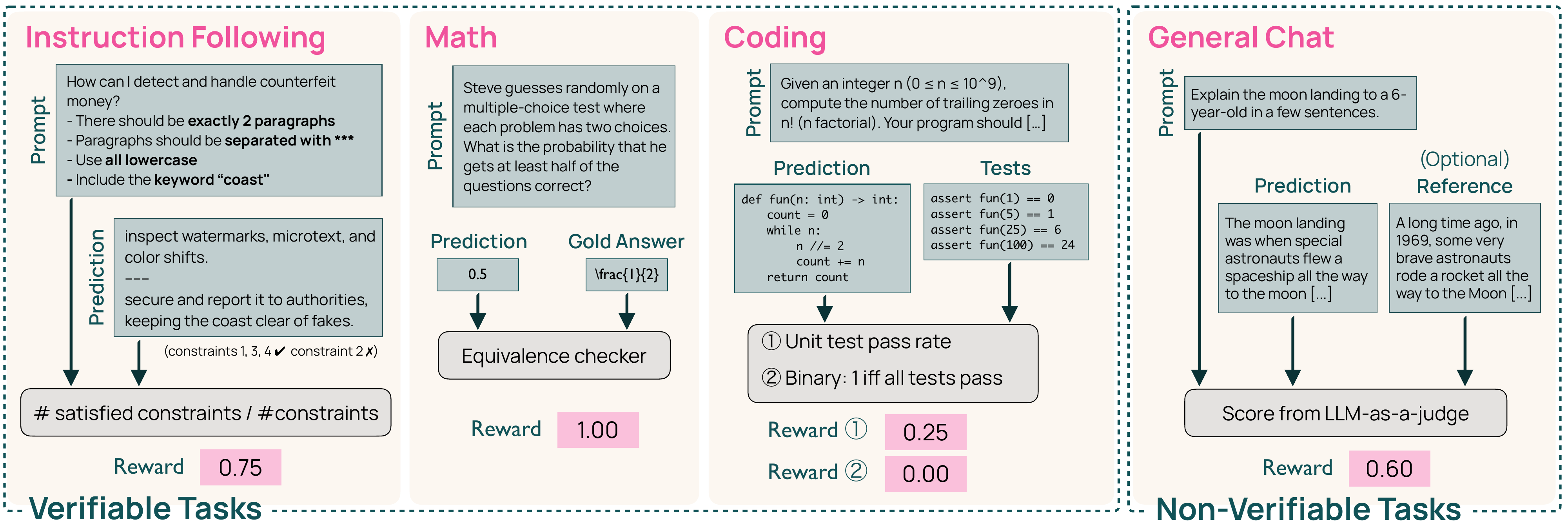}
    \caption{\textbf{Verifiers and reward design} for verifiable and non-verifiable tasks.
    }
    \label{fig:rl-teaser}
\end{figure}

\subsubsection{\olmothreerl Algorithmic Details}
\label{olmo3think-rl}

Our reinforcement learning stage is powered by \olmothreerl, an approach that builds on Group Relative Policy Optimization (GRPO)~\citep{shao2024deepseekmath} and integrates a number of recent algorithmic advances. In particular, we adopt improvements from DAPO~\citep{yu2025dapo} and Dr GRPO~\citep{liu2025understanding}, among others~\citep{yao2025offpolicy, pipelinerl}.
At its core, the objective of RLVR is to maximize the expected reward of a model-generated response $y$ given the prompt $x$, where a verifier checks whether the response \(y\) matches the ground-truth answer associated with $x$.

We make the following improvements\footnote{We experimented with additional changes (e.g., overlong filtering), but did not find these gave consistent improvements.} over vanilla GRPO:
\begin{itemize}
    \item {\bf{Zero gradient signal filtering}}: We remove groups of instances whose rewards are all identical (i.e., a batch with zero standard deviation in their advantage) to avoid training on samples that provide zero gradient, similar to DAPO~\citep{yu2025dapo}.
    \item {\bf{Active sampling}}: We maintain a consistent batch size in spite of zero gradient filtering with a novel, more efficient version of dynamic sampling \citep{yu2025dapo}, see 
    Section~\S\ref{sec:posttraining_infra} for details.
    \item {\bf{Token-level loss}}: We use a token-level loss to normalize the loss by the total number of tokens across the batch~\citep{yu2025dapo}, rather than per-sample to avoid a length bias.
    \item {\bf{No KL loss}}: We remove the KL loss as a common practice~\citep{glm45,yu2025dapo,liu2025understanding} as it allows less-restricted policy updates, and removing it does not lead to over-optimization or destabilized training.
    \item {\bf{Clip higher}}: We set the upper-bound clipping term in the loss to a slightly higher value than the lower bound to enable larger updates on tokens, as proposed by~\citet{yu2025dapo}.
      \item {\bf{Truncated importance sampling}}: To adjust for differences between log probabilities from the inference and training engines, we multiply the loss by the truncated importance sampling ratio, following~\citet{yao2025offpolicy}.
     \item {\bf{No standard deviation normalization}}: When calculating advantage, we do not normalize by the standard deviation of the group, following~\citet{liu2025understanding}.
     This removes a difficulty bias, where questions with low standard deviation in their rewards (e.g., too hard or too easy) have their advantages significantly increased by the normalization term.
\end{itemize}

\paragraph{\olmothreerl formulation} Our final objective function includes {\color[HTML]{115257} a token-level loss}, {\color[HTML]{f0529c} truncated importance sampling}, {\color[HTML]{0fcb8c} clip-higher}, and {\color[HTML]{b11be8} no standard deviation in the advantage calculation}:
\begingroup\makeatletter\def\f@size{9.8}\check@mathfonts\def\maketag@@@#1{\hbox{\m@th\normalfont\normalfont#1}}
\begin{align}
\mathcal{J}(\mathrm{\theta})
&= {\color[HTML]{115257}\frac{1}{\sum_{i=1}^{G} |y_i|}
  \sum_{i=1}^{G} \sum_{t=1}^{|y_i|}}
  {\color[HTML]{f0529c}\min\!\Big(
  \frac{\pi\!\left(y_{i,t}\mid x, y_{i,<t};\theta_{\mathrm{old}}\right)}{\pi_{\mathrm{vllm}}\!\left(y_{i,t}\mid x, y_{i,<t};\theta_{\mathrm{old}}\right)}, \rho\Big)}
  \min \Bigl(
  r_{i,t}\,  {\color[HTML]{b11be8}{A}_{i,t}},\;
    \operatorname{clip} \bigl(
     r_{i,t},\;
      1 - \varepsilon_{\mathrm{low}},\;
      1 + {\color[HTML]{0fcb8c} \varepsilon_{\mathrm{high}}}
    \bigr)\,
   {\color[HTML]{b11be8}{A}_{i,t}}
  \Bigr),
\end{align}
\endgroup
where $r_{i,t}=\frac{\pi\left(y_{i, t} \mid x, y_{i,<t};\theta\right)}{\pi\left(y_{i,t} \mid x, y_{i, <t};\theta_{\mathrm{old}}\right)}$, $\varepsilon_{\mathrm{low}}$ and ${\color[HTML]{0fcb8c}\varepsilon_{\mathrm{high}}}$ are the clipping hyperparameters. Here, $y_i \sim \pi_{\text {vllm}}(\cdot \mid x;\theta_{\mathrm{old}})$ and $\pi_{\text {vllm }}(\cdot \mid x;\theta_{\mathrm{old}})$ are the token probabilities returned from vLLM, $\rho$ is the truncated importance sampling cap value~\citep{yao2025offpolicy}, and the advantage $A_{i,t}$ for the $t$-th token $t$ in the response $y_{i}$ is calculated within the group $G$ based on the relative reward of the outputs inside each group:
\begin{align}
    {\color[HTML]{b11be8}{A}_{i,t}=\left(r\left(x, y_i\right)-\operatorname{mean}\left(\left\{r\left(x, y_i\right)\right\}_{i=1}^G\right)\right)}.
\end{align}

$r\left(x, y_i\right)$ is the reward score returned by the corresponding verifier. Our hyperparameters for various runs are in Appendix Table \ref{tab:rlvr_training_settings}.

\paragraph{Verifiers}
We extend verifiable rewards beyond math domains from \olmotoo to include general domains.
For each domain we use a different custom verifier (see Figure~\ref{fig:rl-teaser}):
\begin{itemize}
    \item {\bf{Math}} We use a rule-based verifier that performs basic normalization and compares with a reference answer with \texttt{SymPy} to determine answer correctness. The verifier returns 1 if the answer is determined the same as the reference answer and 0 otherwise.
    \item {\bf{Code}} We use a test-case based verifier that runs a set of test cases over the response. We experiment with (a) using the percentage of passed test cases as the reward and (b) returning 1 when the response passes all test cases and 0 otherwise.\footnote{{\bf Code execution} When performing RL on code environments, we need to actually execute the generated code against test cases to calculate our rewards. We use AWS Lambda to do so. Using a distributed cloud function approach ensures that verification does not block the trainer process, and allows us to scale seamlessly. Many test case suites, such as those present in SYNTHETIC-2 \citep{primeintellect2025synthetic2}, contain test cases designed to penalize programs with poor time complexity, and running these tests can exceed hundreds of MBs for a single program, exceeding the resources of our local machines.}
    \item {\bf{Instruction-following}} We pass the response through a set of functions that check adherence to a series of constraints from the prompt. A reward of 1 is assigned if all constraints are satisfied, and 0 otherwise.
    \item {\bf{Chat---reference}} For tasks with a ground-truth response, we pass the response to an LM judge to compare the model's response against a provided reference answer, and ask the judge to give a score in [0, 1] based on the quality of the response.
    \item {\bf{Chat---open-ended}} We pass the response to an LM judge and ask the judge to give a score in [0, 1] based on the quality of the response without any reference answer.\footnote{Unless otherwise stated, for an LM judge we host Qwen3 32B~\citep{qwen3} with thinking mode turned off using vLLM~\citep{vllm}, and allow a max input prompt of 32768 tokens while only allowing a response length of 2048 tokens. We provide the judge prompts in Figure~\ref{llm-judge-prompt} in the appendix.
    We additionally experimented with puzzle problem (checking if a puzzle solution is correct relative to a reference answer) and length-control~\citep{aggarwal2025l1controllinglongreasoning} verifiers, but did not find it useful for downstream performance.}
\end{itemize}

\subsubsection{\textsc{Dolci-Think-RL}: Curating a State-of-the-art RLVR Dataset}
\label{subsec:rl-thinking-data}

\begin{table}[t!]
\centering
\setlength\tabcolsep{5pt}
{\small
\begin{tabular}{ll R{1.8cm} R{2.0cm} l}
\toprule
{\bf Category} & {\bf Prompt Dataset} & {\bf \# Prompts Used in Think RL} & {\bf \# Prompts Used in Instruct RL} & {\bf Reference} \\
\midrule

\rowcolor{ai2offwhite}
Precise IF & IF-RLVR & 30,186 & 38,000 & \citet{pyatkin2025generalizing} \\

Math & Open-Reasoner-Zero & 3,000 & 14,000 & \citet{hu2025open} \\
& DAPO-Math & 2,584 & 7,000 & \citet{yu2025dapo} \\
& AceReason-Math & 6,602 & -- & \citet{chen2025acereason} \\
& Polaris-Dataset & -- & 14,000 & \citet{Polaris2025} \\
& KlearReasoner-MathSub & 3,000 & 9,000 & \citet{su2025klear} \\
& OMEGA-train & 15,000 & 20,000 & \citet{Sun2025OMEGACL} \\
\rowcolor{ai2offwhite}
Coding & AceCoder & 9,767 & 20,000 & \citet{zeng2025acecoder} \\
\rowcolor{ai2offwhite}
& KlearReasoner-Code & 8,040 & -- & \citet{su2025klear} \\
\rowcolor{ai2offwhite}
& Nemotron Post-training Code & 2,303 & -- & \citet{nvidia2025nemotron_post_training_dataset} \\
\rowcolor{ai2offwhite}
& SYNTHETIC-2 & 3,000 & -- & \citet{primeintellect2025synthetic2} \\
General Chat & Tulu 3 SFT & 7,129 & 18,955 & \citet{lambert2024tulu3} \\
 & Wildchat-4.8M & 7,129  & 18,761 & - \\
 & Multi-Subject RLVR & 7,129 & 12,234 & \citet{su2025expanding} \\
\rowcolor{ai2offwhite} 
{\bf{Total}} &  & 104,869 & 171,950 & \\
\bottomrule
\end{tabular}}
\vspace{3pt}
\caption{\textbf{Breakdown of datasets in Dolci-Think-RL used for RL training}. See \S\ref{subsec:rl-thinking-data} for further details on how each dataset is processed.}
\label{tab:olmo3_think_rl_mix}
\end{table}

We curate a large-scale and diverse dataset of roughly 100K samples across four domains: mathematics, coding, instruction following, and general chat to support robust RL on varied reasoning tasks while maintaining general utility.
Each domain is associated with either a verifiable or non-verifiable reward signal (continuous or binary), ensuring that every instance can be automatically checked for correctness or general quality (see Figure~\ref{fig:rl-teaser}).
For all domains we take particular care with the provenance and licensing of sources.
We provide the size of each dataset subsection after sourcing, filtering, and mixing in~\autoref{tab:olmo3_think_rl_mix}.

\paragraph{Step 1: sourcing prompts}

In what follows, we will describe our data construction process.

\begin{itemize}
    \item {\bf{Math}}: We combine community-curated math problems, including Open-Reasoner-Zero~\citep{hu2025open}, DAPO-Math~\citep{yu2025dapo}, AceReason-Math~\citep{chen2025acereason}, DeepScaler~\citep{luo2025deepscaler}, KlearReasoner-MathSub~\citep{su2025klear}, and OMEGA~\citep{Sun2025OMEGACL} covering a wide range of mathematical domains including algebra, combinatorics, number theory, and geometry.

    \item {\bf{Coding}} To construct reinforcement learning (RL) data for code, we required pairs of \emph{(problem, test cases)}.
    We curate a diverse set of prompts for coding problems, including AceCoder~\citep{zeng2025acecoder}, Klear-Reasoner Code~\citep{su2025klear}, Nemotron Post-training Code \citep{nvidia2025nemotron_post_training_dataset}, SYNTHETIC-2 code \citep{primeintellect2025synthetic2}, and Open-Code Reasoner \citep{ahmad2025opencodereasoning}.
    We use the Klear-Reasoner and SYNTHETIC-2 test cases directly.
    For the other datasets, we run prompts through the following synthetic data pipeline: (1) \textit{problem rewriting}, (2)  \textit{solution generation}, and (3) \textit{test case generation}.
    After generating these triplets (problem, solution, test cases), we executed all model-generated or rewritten test cases against the corresponding solutions and kept examples with solutions that passed more than 80\% of test cases while removing failed test cases.
    The resulting filtered dataset provided high-quality \emph{(problem, test cases)} pairs suitable for training and experimentation with RL methods for code.
    We use the AceCoder prompts in function completion format, while all other datasets are in stdio format.
    Details of each step in code data synthesis pipeline can be found in Appendix \ref{appx:code-data}.

    \item {\bf{Instruction-following}} We use the prompts from IF-RLVR~ \citep{pyatkin2025generalizing} with up to 5 constraints, which are sampled from IFEval~\citep{zhou2023instructionfollowingevaluationlargelanguage} and IFBench-Train~\citep{pyatkin2025generalizing}.

    \item {\bf{General chat}} We sample our general chat instances from three sources: (a) \tulu SFT~\citep{lambert2024tulu3}; (b) the new WildChat-4.8M data\footnote{\href{https://huggingface.co/datasets/allenai/WildChat-4.8M}{\path{huggingface.co/datasets/allenai/WildChat-4.8M}}} containing a broad spectrum of user-chatbot interactions on ambiguous requests, code-switching, topic shifts, political debates, and more; and (c) the Multi-subject-RLVR dataset~\citep{su2025expanding}, consisting of college-level English questions and objective answers written by domain experts for examination purposes.
    For WildChat, we only sample from instances that are in English and do not require reasoning (such as math and code).
    For \tulu, we first rewrote samples using GPT-4.1 for better clarity and to extract reference answers from the SFT set.
    We then generated eight samples per prompt with a Qwen 2.5 7B model finetuned on OpenThoughts 2 and computed the F1 score between the reference answer and each response.
    We then removed all samples with average F1 score $<0.1$ and $>0.8$.
    This removes both noisy and overly difficult samples.
    WildChat in particular has a high prevalence of role-playing and other character-based data.
    In order to balance the data, we filter any mention of a single character down to a maximum of 10 instances.\footnote{In our intermediate general dataset of 57,819 samples, we found the top characters were   1. \href{https://doki-doki-literature-club.fandom.com/wiki/Natsuki_(DDLC)}{Natsuki}: 1284 appearances, 2. \href{https://doki-doki-literature-club.fandom.com/wiki/Monika_(DDLC)}{Monika}: 1243, 3. \href{https://doki-doki-literature-club.fandom.com/wiki/Sayori_(DDLC)}{Sayori}: 1076, 4. \href{https://doki-doki-literature-club.fandom.com/wiki/Yuri_(DDLC)}{Yuri}: 957, 5. Sakura: 453, and 6. MC: 424. All others were at 60 or lower before filtering.}
    We then finally performed some post-hoc manual filtering to remove code- and math-centric prompts.

\end{itemize}

\paragraph{Step 2: offline difficulty filtering}
As stated previously, to improve the sample efficiency of RL for our reasoner model, we generate eight rollouts for each prompt from the initial checkpoint of the model we train (e.g., if starting from the DPO-trained model, we generate from the DPO checkpoint). We then remove all samples that the model easily solves (that is, those with a pass rate greater than 62.5\%).
We sample with a temperature of 1.0 and top-p of 1.0, matching how we sample during RL training.
We used offline filtering for the 7B \olmothreethinking to filter out RL problems that are too easy for our models' training.
For the 32B, we rely on active sampling, which fills RL batches only on samples with a non-zero GRPO group gradient, and re-using the 7B DPO-filtered data as the starting point for the model due to compute and time constraints.

\paragraph{Step 3: data mixing}
When developing our data mixture and overall recipe, we found RL experiments were both long and compute-expensive, preventing us from ablating the full space of datasets and algorithmic choices.
Instead, we established a pipeline in which: (a) we performed dataset-specific runs on an intermediate SFT checkpoint and observed downstream evaluation trends over the first 500-1000 RL steps; (b) focused on math domain training when testing new algorithmic changes; (c) periodically ran overall mixture experiments to ensure mixing was stable.
When setting up our final run, we then took the most promising datasets, performed offline filtering, and carefully mixed them to ensure higher-quality datasets were upweighted, and roughly equal amounts of data were used for each domain (with slightly more focus on math and instruction following, as training on these domains seemed the most effective in per-dataset runs).
Additionally, we downsample certain subtasks from OMEGA that the model especially struggled with based on offline filtering results.\footnote{In particular, we downsample the following tasks by 50\% \textit{after} filtering: \texttt{trans\_integrations}, \texttt{logic\_gridworld\_rookmove}, \texttt{logic\_puzzles\_grid\_chip}, \texttt{comp\_grid\_chips}, \texttt{comp\_n\_gon}, \texttt{arithmetic\_matrix\_svd}, \texttt{comp\_parametric\_intersection}, \texttt{comp\_vertex\_color}.}
We used this pipeline to develop an RL mixture for the 7B model, and then used the same data mixture for the 32B model due to compute and time constraints.

For our \olmothreethinking 7B training run, we used an initial version of our infrastructure without pipelineRL or truncation importance sampling, which took approximately 15 days. We later replicated the same run with our newer infrastructure, achieving similar performance in just 6 days of training.

\subsubsection{\olmothreerl Infrastructure in \openinstruct}
\label{sec:posttraining_infra}

We made substantial improvements to our reinforcement learning infrastructure to handle longer sequences and faster overall throughput.
In RL with LLMs, the key technical challenge for finetuning models that generate long sequences is managing inference---also called the rollouts.
For our final models, we generated rollouts with a maximum size of 32K tokens in length, averaging more than 10K tokens (for the reasoner models). Inference dominated our computational costs, using 8 H100 nodes for training and 20 nodes for inference for the 32B \olmothreerl reasoner model. Given the cost of autoregressive inference, our learner spends 75\% of the time waiting for data, so in terms of GPU utilization, we use approximately 5x as much for inference as for training. In fact, we use the minimal possible sharding configuration to fit the learner in memory and do not prioritize speed at all, unlike in the supervised learning setting. For the 7B reasoner model, where we have less memory pressure on the learner, the situation was more dramatic, as we used 7 nodes for inference and only 2 for the learner. Given the similarly low utilization of the learner, we used approximately 14x as much compute for inference as for training. We suspect that we have a suboptimal sharding configuration for the 32B learner and expect that we could do better in future work.

\begin{figure}[t!]
    \centering
    \begin{subfigure}[t]{0.59\textwidth}
        \includegraphics[width=0.99\linewidth]{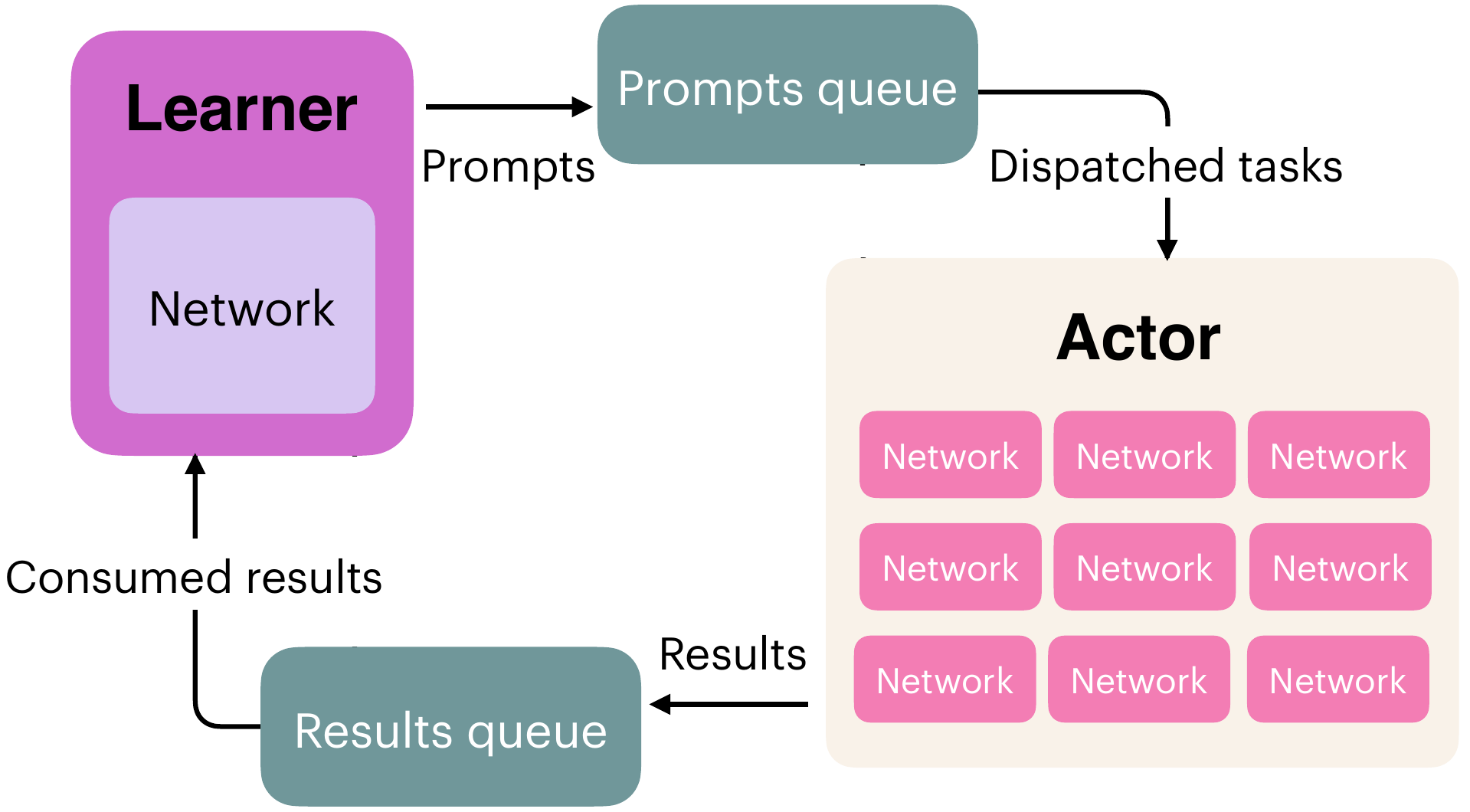}
        \caption{Distributed reinforcement learning architecture}
        \label{fig:actor-learner}
    \end{subfigure}
    \hfill
    \begin{subfigure}[t]{0.39\textwidth}
    \centering
        \vspace{-15em}
        \begin{subfigure}{0.7\textwidth}
            \centering
            \includegraphics[width=\linewidth]{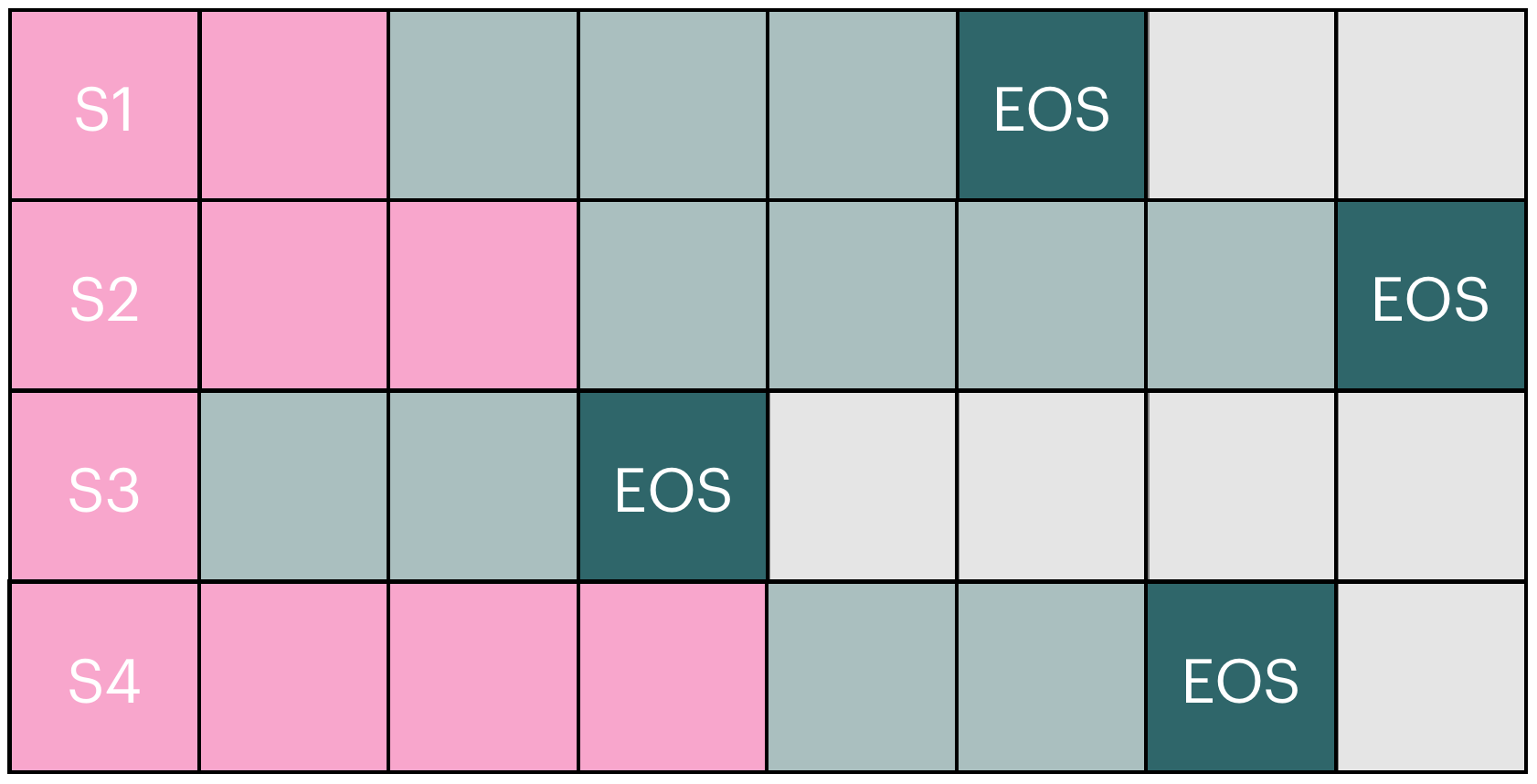}
            \caption{Static batching}
            \label{fig:static-batching}
        \end{subfigure}
        \begin{subfigure}{0.7\textwidth}
            \centering
            \includegraphics[width=\linewidth]{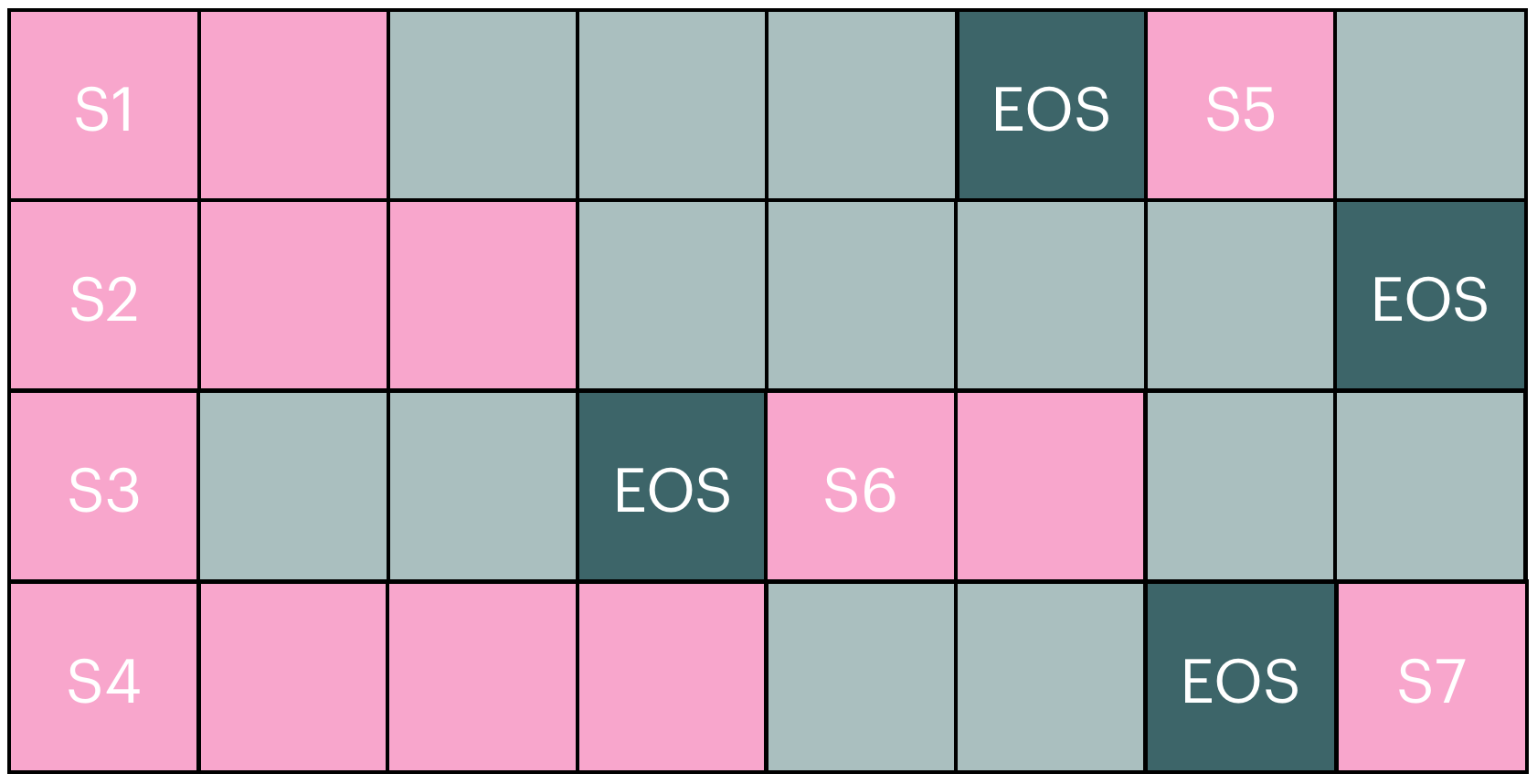}
            \caption{Continuous batching}
            \label{fig:continuous-batching-sub}
        \end{subfigure}
    \end{subfigure}
    \caption{\textbf{Overview of \olmorl infrastructure}. On the left: distributed reinforcement learning architecture (Figure~\ref{fig:actor-learner}). On the right: static vs.\ continuous batching.
    Static batching (Figure~\ref{fig:static-batching}) wastes compute when generations have variable sequence lengths. Pink cells are prefilled tokens, green cells are decoded tokens, with dark green representing EOS.
    Grey indicates that sequence is not doing anything, so
    continuous batching (Figure~\ref{fig:continuous-batching-sub}) backfills finished rows immediately, resulting in no wasted compute.}
    \label{fig:continuous-batching}
\end{figure}

\paragraph{Fully asynchronous training} Shown in \autoref{fig:actor-learner}, we employ an off-policy asynchronous RL setup \citep{noukhovitch2024asynchronousrlhffasterefficient} featuring a centralized learner distributed across multiple nodes via DeepSpeed~\citep{deepspeed} and a large pool of actors, each running an independent vLLM~\citep{vllm} instance.
The learner produces prompts that are queued and dispatched to the actors, which execute the prompts, interact with the environment, and return results through a results queue that the learner uses to update the model parameters.%
\footnote{For the 7B training runs, we use a single GPU for each actor and scale generation via data parallelism.
The RL setup would be familiar to readers of \cite{horgan2018distributed} or \cite{silver2017alphazero}. For 32B, we use one node per actor and then similarly further scale via data parallelism.}
Due to the variance in completion length, a long time delta can emerge between completions in an individual batch of RLVR.
The guiding principles to mitigate this issue are to make efficient use of resources (avoiding idling) and to make processes asynchronous.\footnote{%
For one of our main RL runs, which was broadly representative of what we experienced across all of our runs, each training step averaged 1000 seconds, of which 125 seconds was spent running training.
Each batched completion generation took 1000 seconds.
As we overlap generation and training \citep{noukhovitch2024asynchronousrlhffasterefficient}, the bottleneck is entirely generation.
Consequently, significant engineering resources were spent improving the way generation is handled, where we could continue to use the training code used in \olmotoo, as we would need to speed up generation by $>8\times$ for that to be a bottleneck. }

\begin{figure}
    \centering

\end{figure}

\paragraph{Continuous batching}{We employ continuous batching to constantly enqueue new generations as each one finishes to remove the compute waste for long generations (see Figure~\ref{fig:continuous-batching}).
    This is in contrast to {\it static batching}, in which a batch of prompts is split over $N$ actors, and each actor generates the entire batch,\footnote{Calling \texttt{llm.generate} in vLLM.} returns the generated responses to the learner, and then receives a new batch of data.
    Static batching is inefficient, as when one generation finishes that ``slot'' of the batch will remain empty until we get a new batch.
    The exact wasted compute can be calculated as the maximum sequence length minus the average sequence length divided by the maximum sequence length.
    With \olmothree, at a 32K generation length, we see a mean generation length of 14628 and a maximum of 32K, which means that up to 54\% of our compute would have been wasted with static batching.
    See Figure~\ref{fig:continuous-batching} for an illustrated example. %

\paragraph{Active sampling}{To compensate for filtered instances, our fully asynchronous framework enables continuously pulling completions from the actor and resampling prompts into the queue. We actively sample and filter until we reach our desired batch size of non-zero-gradient completions. Previously, \citet{yu2025dapo} dynamic sampling would oversample and generate three times the number of prompts used in each training batch. This was to reasonably guarantee that the batch had enough completions with non-zero standard deviation. In contrast, our active sampling more efficiently uses the infrastructure. As demonstrated in \autoref{sec:rlzero}, we find this significantly stabilizes training and prevents batch size from reducing over the course of training (a common issue with vanilla GRPO).

\paragraph{Inflight updates}{%
A common goal of RL training for LLMs is to minimize the degree of difference between the actor policy and the learner policy, i.e., to minimize being off-policy~\citep{deadly-triad}.
This can be achieved by synchronizing the weights after every training step as follows: each actor finishes all of their ongoing generations, dumps the KV cache, and updates its copy of the weights. However, this causes GPUs to be idle and hurts training efficiency.
Instead, we follow \citet{pipelinerl} to immediately update the weights without pausing the engine, relying on the generation framework to be thread-safe, and continue generating, \emph{without invalidating the KV cache}.
This enables a significant increase in throughput: up to 4x faster with the same resources, without hurting accuracy.}

\paragraph{Better threading and engineering}{ These changes are primarily around handling the weight synchronization after each training step to make actors more efficient. %
Our new setup decouples the actors, allowing each one to start and stop by itself, without waiting for the rest of the actors to finish their syncs as well.
Similarly, we make a large number of optimizations that were not machine learning specific, and were centered around efficiently using the \emph{CPU}.}
For example, our initial implementation of continuous batching, for instance, was slower than static batching before adding a prefetch thread to our actors that constantly refilled the inference queue to see a throughput improvement.

\begin{table}[tbp]
\centering
\small
\begin{tabular}{@{}l >{\columncolor{ai2pink!18}}c cccccccccc@{}}%
\toprule
& \multicolumn{11}{c}{\textbf{Subset of \olmothreethinking Benchmarks}} \\
\cmidrule(lr){2-12}
\textbf{Name} & \cellcolor{ai2pink!18}\textbf{Avg.} & \textbf{MMLU} & \textbf{BBH} & \textbf{GPQA} & \textbf{Zebra} & \textbf{AGI} & \textbf{AIME25} & \textbf{AIME24} & \textbf{CHE} & \textbf{LCB} & \textbf{IFEval} \\
\midrule
Qwen3 32B (chosen) & 83.2 & 88.8 & 90.6 & 64.7 & 78.2 & 90.2 & 71.0 & 80.3 & 90.9 & 89.6 & 87.4 \\
Qwen3 0.6B (rejected) & 35.1 & 55.8 & 41.5 & 27.2 & 29.8 & 59.2 & 15.2 & 11.2 & 14.8 & 34.4 & 62.3 \\
\midrule
Dev. 7B SFT ckpt & 70.3 & {\bf 76.1} & {\bf 83.9} & 45.1 & 56.5 & 76.4 & 58.8 & 71.0 & 88.1 & 67.0 & {\bf 79.7} \\
Cont. SFT on chosen & 64.5 & 72.6 & 80.2 & 40.2 & 49.8 & 73.9 & 52.8 & 61.0 & 83.4 & 55.1 & 76.0 \\
Delta learning & {\bf 72.9} & 75.5 & 82.8 & {\bf 48.4} & {\bf 60.9} & {\bf 79.7} & {\bf 66.3} & {\bf 75.7} & {\bf 91.5} & {\bf 72.6} & 75.2 \\
\bottomrule
\end{tabular}
\caption{\textbf{The delta between chosen and rejected responses is critical}. Supervised finetuning directly on the chosen responses generated by Qwen3-32B Thinking hurts the Initial SFT model. In contrast, DPO tuning to prefer the 32B responses over weaker Qwen3-0.6B Thinking responses yields strong gains across math and code reasoning.}
\label{tab:dpo_sftchosen}
\end{table}

Our final RL run ended up mixing carefully-filtered data from all domains roughly equally and running on top of the DPO checkpoint.

\subsection{Key Findings}
\label{sec:thinkfindings}

\begin{table}[tbp]
\centering
\small
\begin{tabular}{@{}l >{\columncolor{ai2pink!18}}c cccccccccc@{}}
\toprule
& \multicolumn{11}{c}{\textbf{Subset of \olmothreethinking Benchmarks}} \\
\cmidrule(lr){2-12}
\textbf{Name} & \cellcolor{ai2pink!18}\textbf{Avg.} & \textbf{MMLU} & \textbf{BBH} & \textbf{GPQA} & \textbf{Zebra} & \textbf{AGI} & \textbf{AIME25} & \textbf{AIME24} & \textbf{CHE} & \textbf{LCB} & \textbf{IFEval} \\
\midrule
SFT & 70.1 & 74.9 & 84.1 & 45.8 & 57.9 & 77.2 & 57.6 & 69.6 & 88.2 & 67.8 & 77.9 \\
SFT + DPO & 72.7 & 74.8 & 83.7 & 48.6 & 60.6 & 79.1  & 62.7 & {\bf 74.6} & {\bf 91.4} & {\bf 75.1} & 75.9 \\
SFT + RLVR & 71.9 & 77.4 & 83.2 & 42.7 & 63.1 & 78.5 & 62.4 & 70.0 & 87.9 & 70.7 & {\bf 82.8} \\
SFT + DPO + RLVR & {\bf 74.1} & {\bf 77.9} & {\bf 86.8} & {\bf 50.2} & {\bf 62.9} & {\bf 80.1} & {\bf 64.2} & 73.2 & 89.9 & 73.4 & 82.3 \\
\bottomrule
\end{tabular}
\caption{\textbf{Delta learning provides a stronger initialization for subsequent RLVR than SFT alone}. We show the effect of conducting RLVR for 1000 steps after DPO and SFT on our 7B model on a subset of our evaluation suite. Note that here evaluations are from one run only. Preference tuning with delta learning first followed by RLVR, yields the best overall performance. For RLVR, we use data offline-filtered by the corresponding starting point (SFT only or SFT + DPO).}
\label{tab:dpo_rl_sequence}
\end{table}

\begin{table}[t]
\centering
\begin{tabular}{l r r r r l}
\toprule
                    & \textbf{Total tokens (Mtok)} & \textbf{Tokens/second} & \textbf{MFU (\%)} & \textbf{MBU (\%)} %
                    \\ \midrule
\rowcolor{ai2offwhite}
 \olmotoo                       & $6.34$ &  881               & 0.30     & 12.90   %
\\
$\plus$ continuous batching & $7.02$ &  975               & 0.33     & 14.29  %
\\
\rowcolor{ai2offwhite}
$\plus$ better threading    & $9.77$ & 1358               & 0.46     & 19.89   %
\\
$\plus$ inflight updates (\olmothree)   & $21.23$ & 2949 & 1.01     & 43.21  %
\\ \bottomrule
\end{tabular}
\caption{\textbf{Effect of core infrastructure improvements to \olmothreerl}. We ablate the effect of each component by measuring the training speed (tokens/second) and utilization metrics as we add each component in turn from the original \olmotoo RL infrastructure. The addition of inflight-updates has the most drastic improvement.
}
\label{tab:rl-infra}
\end{table}
\begin{figure}[t]
    \centering
    \adjustbox{max width=\linewidth}{\includegraphics{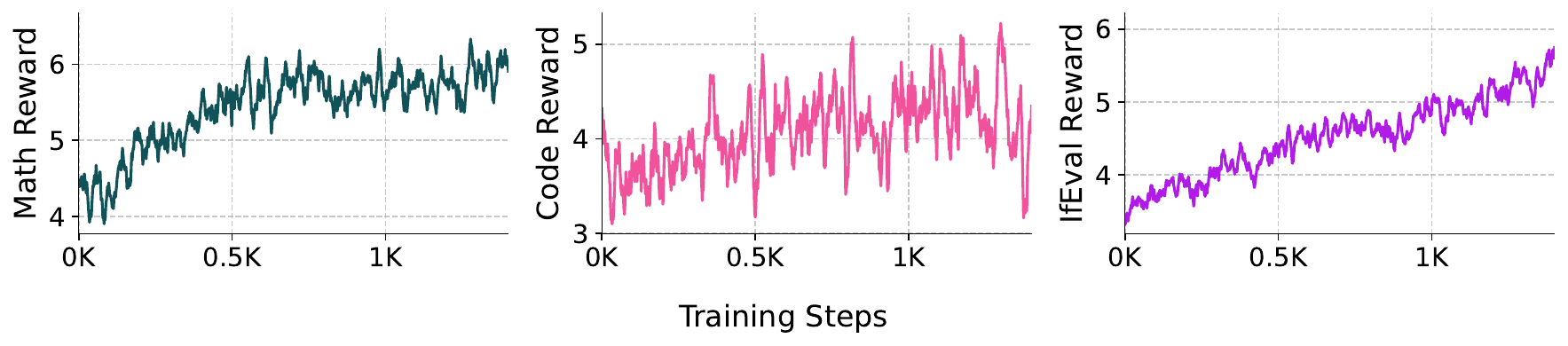}}
    \caption{{\textbf{Reward curves during training of \olmothreethinking 7B}}. Average, math, code, and IF reward over RL training for the final RLVR training run of \olmothreethinking. Reward steadily grows across domains, suggesting smooth training. See~\autoref{fig:olmo3_final_rl_run} in Appendix for further RL curves.}
    \label{fig:rl_rewards_three}
\end{figure}
\begin{figure}[t]
    \centering
    \adjustbox{max width=.9\linewidth}{\includegraphics{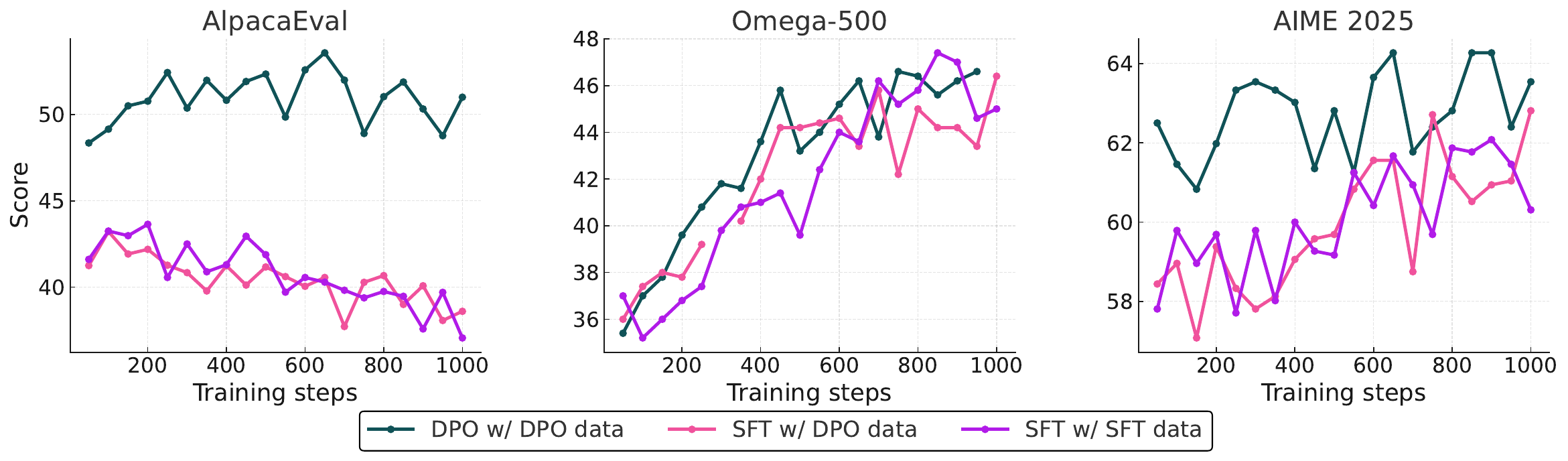}}
    \caption{{\textbf{Using DPO as a starting point for RLVR works best}}. AlpacaEval, Omega-500, and AIME 2025 performance over the course of RLVR training when starting from \olmothree 7B SFT or DPO, training using either data filtered via the DPO model (w/ DPO data) or SFT model (w/ SFT data). The importance of starting from DPO or SFT depends on the evaluation, but starting from DPO is overall preferable.}
    \label{fig:dpo_sft_rl_comparison}
\end{figure}
\begin{figure}[t]
    \centering
    \adjustbox{max width=.68\linewidth}{\includegraphics{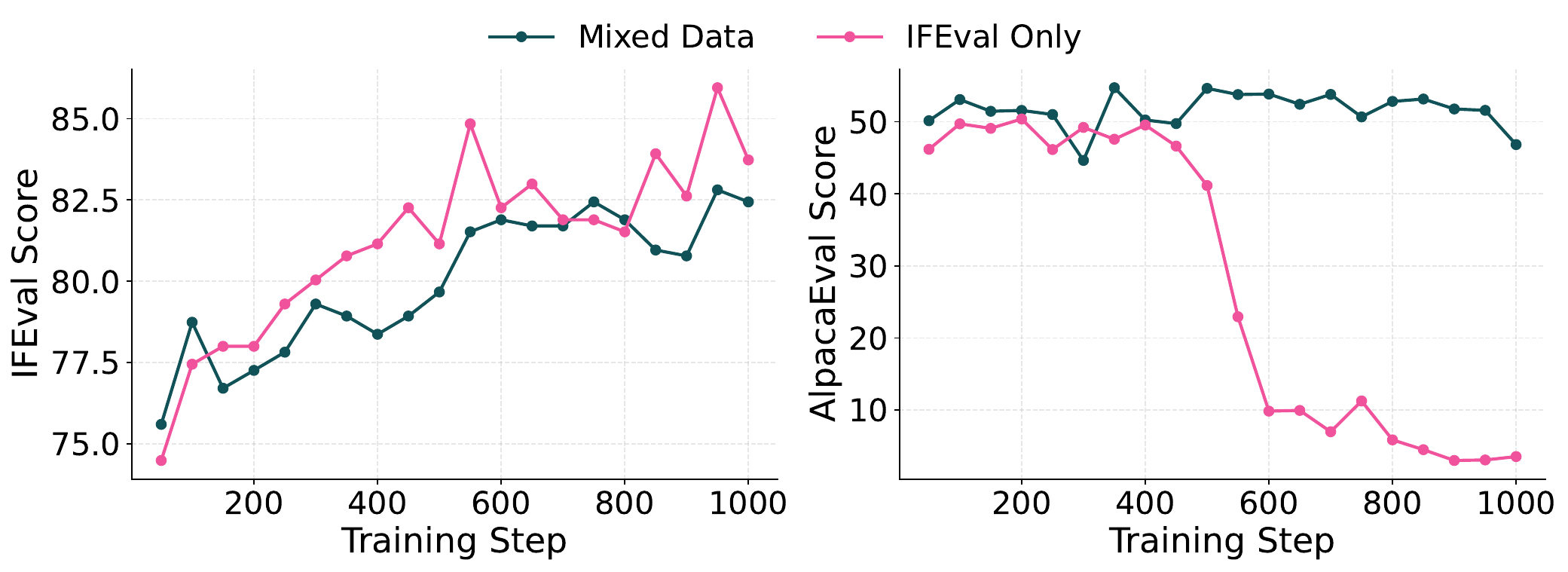}}
    \adjustbox{max width=.3\linewidth}{\includegraphics{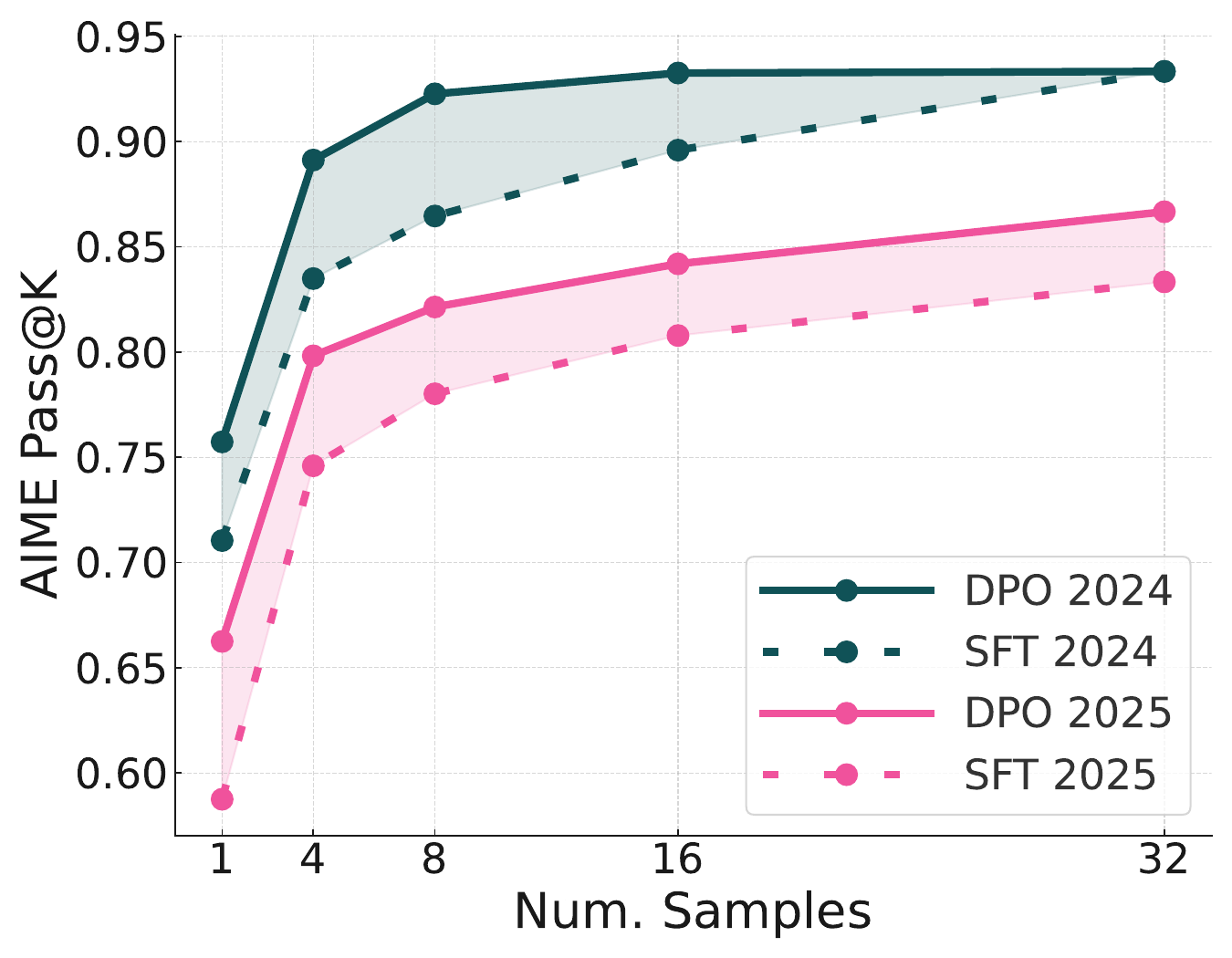}}
    \caption{\textbf{Effect of mixing and DPO on downstream metrics}.
    Training on mixed data prevents overfitting (left) We plot IFEval and AlpacaEval performance over RL training on \olmothreethinking SFT 7B when training on IFEval data only or on mixed data. Training on mixed data achieves similar IFEval performance while maintaining high AlpacaEval performance. DPO with delta learning displays higher pass@K performance than SFT (right). We plot pass@K for AIME 2024 and 2025 for SFT and DPO thinking models for up to K=32. DPO consistently improves performance, even at higher K.}
    \label{fig:ifeval_alpacaeval_mix}
    \end{figure}

\begin{figure}[t]
    \centering
    \adjustbox{max width=.32\linewidth}{\includegraphics[]{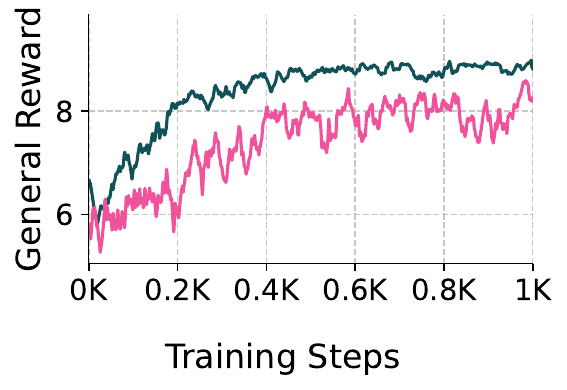}} \hspace{-1.5cm}
    \adjustbox{max width=.4\linewidth}{\includegraphics[]{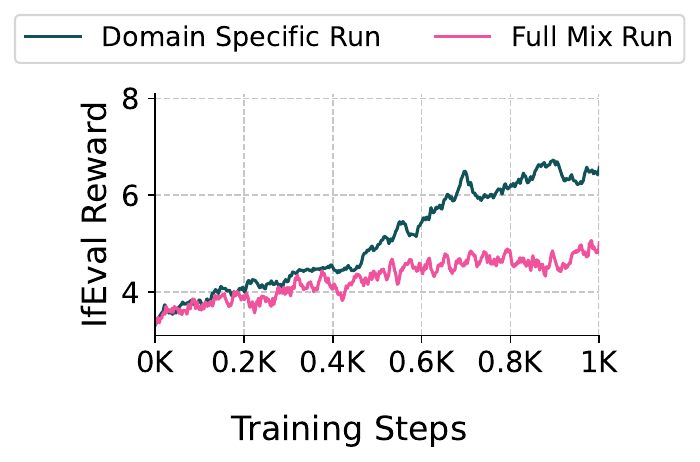}}
    \hspace{-1cm}
    \adjustbox{max width=.32\linewidth}{\includegraphics[]{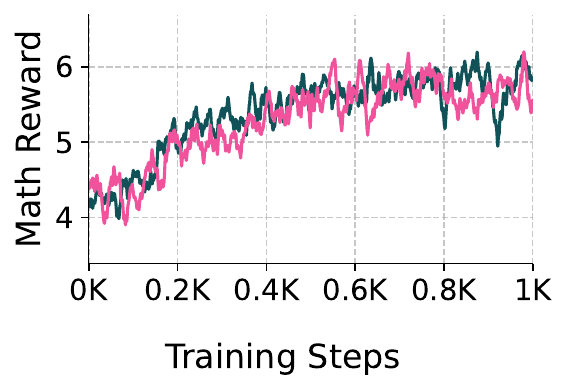}}
    \caption{{\textbf{Per-domain training yields higher train rewards.}} We plot the train reward over RL training for per-domain and overall mix (i.e., final) training runs. In each plot, we train an intermediate SFT model using RLVR with data only from general, IF, and math subsets, and compare to training on our overall mix. While the domain-specific runs achieve higher train reward, \autoref{fig:ifeval_alpacaeval_mix} shows this does not necessarily yield improved downstream performance.}
    \label{fig:per_domain_vs_mix_training}
\end{figure}

\paragraph{DPO yields gains where SFT on the same data cannot} Continued supervised finetuning directly on the chosen responses from \dolcithinkdpo outright hurts the initial SFT model (\autoref{tab:dpo_sftchosen}), dropping all evaluation tasks. We conjecture that this is because the chosen responses (generated by Qwen3 32B Thinking) are weaker relative to data the model has already seen in \dolcithinksft, and hence, they are no longer useful targets for imitation. However, by pairing these chosen responses with rejected responses generated by a weaker model, we construct a useful contrast, enabling preference tuning to drive strong gains beyond the initial SFT model (\autoref{tab:dpo_sftchosen}). Promisingly, these gains are not merely converting pass@k into pass@1 but rather expanding the reasoning frontier of the model (e.g., improved pass@k on AIME evaluations; \autoref{fig:ifeval_alpacaeval_mix}). These findings highlight contrastive learning with preference tuning as a useful stage for improving capabilities even when imitation is saturated.

\paragraph{DPO and SFT both benefit from RL, but DPO remains a better starting point}
\autoref{tab:dpo_rl_sequence} shows that running our final RL mix on the DPO model consistently yields better performance than running it on the SFT model. We find three primary differences, highlighted in~\autoref{fig:dpo_sft_rl_comparison}: for evaluations that RL does not improve, the DPO model often performs better and maintains its advantage during RL training (e.g., AlpacaEval). For evaluations explicitly targeted by RL (e.g., Omega), both the DPO and SFT models achieve similar end performance. For evaluations targeted by RL but hard to improve further from DPO (e.g., AIME 2025), the SFT model improves to get close to DPO performance. In no situation does the SFT model improve over the DPO model after RL, and as such we opt to focus on applying RL over our DPO model.
Curiously, we find that the SFT model performs similarly when trained either with the data offline-filtered using the SFT or DPO model, suggesting that the additional samples filtered out (i.e., solved) by the DPO model do not provide additional signal for improving the SFT model.
Further investigating this, we find that while the DPO model does display lower entropy, it in fact has higher pass@K performance on AIME evaluations, as shown in \autoref{fig:ifeval_alpacaeval_mix}.
This suggests that the DPO model remains a strong starting point for RL relative to the SFT model, as prior work~\citep{yue2025limit-of-rlvr,shao2024deepseekmath} suggests RLVR, under certain conditions, helps convert pass@K improvements into pass@1 gains.

\paragraph{Rewards steadily increase across all domains during RL} ~\autoref{fig:rl_rewards_three} plots per-verifier reward curves along with average output length. We find that reward steadily increases across all domains, albeit at differing rates (with instruction-following data increasing most steadily, and code reward increasing most slowly).
We plot more RL curves in the appendix (\autoref{fig:olmo3_final_rl_run}).
Interestingly, we find that sequence lengths first slightly dip and then slowly increase over time. This is likely due to the reasoning SFT and DPO already training the model to produce long reasoning traces of up to the maximum response length of 32K tokens.

\paragraph{Mixing RL data from varied domains can prevent over-optimization}
\autoref{fig:ifeval_alpacaeval_mix} (left) demonstrates that training on specific domains can lead to over-optimization,
in which performance on evaluations outside that domain drops, while training on a mix yields steady improvements across
different domains. For example, we observe a trade-off when performing \olmothreerl on IFEval alone, wherein higher
IFEval scores correlate with lower AlpacaEval scores. However, when we perform our final mixed training,
we are able to maintain high AlpacaEval scores without compromising IFEval performance, as the LM-judge reward
ensures that the model continues to produce well-formed chat responses.

\paragraph{Mixing data yields lower train reward, but not lower downstream performance}
While~\autoref{fig:ifeval_alpacaeval_mix} demonstrates that our final mixture run achieves downstream performance similar to or greater than RL training runs on single domains, we find that we observe \textit{lower} train reward across each domain when training on mixed data as opposed to single-domain data, as seen in~\autoref{fig:per_domain_vs_mix_training}.
This suggests that mixing data may in fact reduce the model's tendency to \textit{over-optimize during training}, preventing some degree of reward-hacking and thus generalizing better to downstream evaluations. This may explain why RL training on broader data mixtures can outperform domain-specific mixtures~\citep{cheng2025revisiting}.

\paragraph{Continuous batching and inflight updates are crucial to training speed} Using a reasoner SFT or DPO as a starting point stresses RL training to its limits, as the model starts with extremely long average generation lengths. Table~\ref{tab:rl-infra} demonstrates how using continuous batching and inflight updates is crucial to training speed, allowing us to achieve two times faster training on half as many GPUs, making experimentation and long RL runs more tractable.\footnote{While an initial checkpoint took 14 straight days of training across 9 nodes to achieve 1 epoch, with continuous batching and inflight updates, we could achieve 1 epoch on 5 nodes in 7 days.}
To carefully benchmark this, we ablate the changes to our RL infrastructure between \olmotoo and \olmothree. See Table \ref{tab:rl-infra}. %
For each ablation, we ran a benchmark experiment for 2 hours using 2 8x A100 nodes. One node was used for training, and one for inference. Since inference is our bottleneck, we report Model FLOPs Utilization (MFU) and Model Bandwidth Utilization (MBU)  based solely on the single node used for inference. A typical full-scale experiment would use many more nodes for inference, typically with a 8:1 ratio (or more) of inference nodes to training nodes. The benchmark experiment generates a batch of 128 completions for each training step, using 64 prompts, each sampled twice, with a maximum output length of 32000, and a maximum input length of 2048, leading to a context length of 2048.\footnote{Script can be found in the \href{https://github.com/allenai/open-instruct}{\path{github.com/allenai/open-instruct}}, at \texttt{scripts/benchmarking/olmo3\_infra.sh}.} %

\paragraph{\olmothreerl shows significant improvement in precise instruction following} The precise instruction-following performance increases across post-training stages, with the final RL training stage leading to the biggest improvements in \olmothree's precise instruction-following abilities, as shown in Table~\ref{tab:if_summary}, for both the development (IFEval) and the unseen (IFBench) evaluations.

\begin{table}[H]
\centering
\small
\begin{tabular}{lcccccc}
\toprule
                  & \textbf{Think SFT} & \textbf{Think DPO} & \textbf{Think RL}       & \textbf{Instruct SFT} & \textbf{Instruct DPO} & \textbf{Instruct RL} \\ 
\midrule
\rowcolor{midgrey} \multicolumn{7}{c}{\textbf{7B scale}} \\
\rowcolor{lightgrey} \textbf{IFEval}  & 77.9      & 75.9     & {\bf{88.2}} & 81.7           & 82.0         & {\bf{85.8}}         \\
\rowcolor{lightgrey}\textbf{IFBench} & 30.0      & 28.3      & {\bf{41.6}}  & 27.4         & 29.3         & {\bf 32.3}          \\
\rowcolor{midgrey} \multicolumn{7}{c}{\textbf{32B scale}} \\
\rowcolor{lightgrey} \textbf{IFEval}  & 83.7      & 82.3     & 89.0 (3), {\bf 93.8} (3.1) & 87.7           & 87.3         & 88.8         \\
\rowcolor{lightgrey} \textbf{IFBench} & 37    & 34.4     & 47.6 (3), {\bf 68.1} (3.1)  & 29.7         & 36.3         & 39.7          \\ 
\bottomrule
\end{tabular}
\caption{\textbf{Summary of precise instruction following results on IFEval and IFBench}, for both the \olmothreethinking and \olmothreeinstruct models (at 7B and 32B sizes), across various stages of the post-training pipeline.}
\label{tab:if_summary}
\end{table}

\section{\olmothreeinstruct} \label{sec:posttrain-instruct}

 Recent studies suggest that real-world language model use predominantly centers on general tasks such as advice-seeking and information recall~\citep{Chatterji2025-fs} that may not require extensive reasoning. 
 Everyday chat settings often do not require the inference-time scaling of \olmothreethinking. Hence, we develop \olmothreeinstruct, a non-reasoning model designed with these real use cases in mind. \olmothreeinstruct  quickly and helpfully respond to common user queries.

This different model type demands different data to support it. 
We focus on improving the interactivity of the models by introducing multi-turn DPO data and promoting concise responses in our delta-learning preference-tuning pipeline. 
Additionally, \olmothreeinstruct is trained for function-calling, for which we release new SFT datasets. 
Together, our recipe yields \olmothreeinstruct models that effectively leverage tools and efficiently respond to user queries. 

\subsection{Main Results for \olmothreeinstruct}
\label{sec:posttrain_eval_instruct}

\begin{table}[t]
\centering
\footnotesize
\setlength\tabcolsep{5pt}
\renewcommand{\arraystretch}{0.95}
\adjustbox{max width=\linewidth}{
{\fontsize{8}{8}\selectfont
\begin{NiceTabular}{@{}Hl
C{35pt}C{35pt}P{35pt}|
C{35pt}C{35pt}C{35pt}C{35pt}C{35pt}C{35pt}C{35pt}@{}}
\toprule
& & \multicolumn{3}{c}{\quad \quad \textbf{\texttt{Olmo 3.1 32B Instruct}}} & \multicolumn{7}{c}{\textbf{\texttt{Baselines}}} \\
\textbf{Skill} &
& \textbf{SFT} & \textbf{DPO} & \textbf{Final \mbox{Instruct} 3.1} & \textbf{Apertus 70B} & \textbf{Qwen 3 32B (No Thinking)} & \textbf{Qwen 3 VL 32B Instruct} & \textbf{Qwen 2.5 32B} & \textbf{Gemma 3 27B} & \textbf{Gemma 2 27B} & \textbf{OLMo 2 32B} \\
\midrule

\rowcolor{midgrey} -- & \textbf{Math} & & & & & & & & & & \\
\rowcolor{lightgrey} -- & \metric{MATH} & 74.4 & 86.6 & 93.4 & 36.2 & 84.3 & 95.1 & 80.2 & 87.4 & 51.5 & 49.2 \\
\rowcolor{lightgrey} -- & \metric{AIME 2024} & 12.7 & 35.2 & 67.8 & 0.31 & 27.9 & 75.4 & 15.7 & 28.9 & 4.7 & 4.6 \\
\rowcolor{lightgrey} -- & \metric{AIME 2025} & 8.2 & 23.3 & 57.9 & 0.1 & 21.3 & 64.2 & 13.4 & 22.9 & 0.9 & 0.9 \\
\rowcolor{lightgrey} -- & \metric{OMEGA} & 15.5 & 33.3 & 42.2 & 5.6 & 23.4 & 44.0 & 19.2 & 24.0 & 9.1 & 9.8 \\
\midrule

\rowcolor{midgrey} -- & \textbf{Reasoning} & & & & & & & & & & \\
\rowcolor{lightgrey} -- & \metric{BigBenchHard} & 69.0 & 82.1 & 84.0 & 57.0 & 80.4 & 89.0 & 80.9 & 82.4 & 66.0 & 65.6 \\
\rowcolor{lightgrey} -- & \metric{ZebraLogic} & 30.6 & 51.1 & 61.7 & 9.0 & 28.4 & 86.7 & 24.1 & 24.8 & 17.2 & 13.3 \\
\rowcolor{lightgrey} -- & \metric{AGI Eval English} & 71.7 & 79.4 & 79.5 & 61.6 & 82.4 & 89.4 & 78.9 & 76.9 & 70.9 & 68.4 \\
\midrule

\rowcolor{midgrey} -- & \textbf{Coding} & & & & & & & & & & \\
\rowcolor{lightgrey} -- & \metric{HumanEvalPlus} & 80.8 & 85.7 & 86.7 & 42.9 & 83.9 & 89.3 & 82.6 & 79.2 & 67.5 & 44.4 \\
\rowcolor{lightgrey} -- & \metric{MBPP+} & 61.5 & 63.6 & 65.1 & 45.8 & 67.9 & 69.0 & 66.6 & 65.7 & 61.2 & 49.0 \\
\rowcolor{lightgrey} -- & \metric{LiveCodeBench v3} & 35.4 & 49.6 & 54.7 & 9.7 & 57.5 & 70.2 & 49.9 & 39.0 & 28.7 & 10.6 \\
\midrule

\rowcolor{midgrey} -- & \textbf{IF} & & & & & & & & & & \\
\rowcolor{lightgrey} -- & \metric{IFEval} & 87.7 & 87.3 & 88.8 & 70.4 & 87.5 & 88.1 & 81.9 & 85.4 & 62.1 & 85.8 \\
\rowcolor{lightgrey} -- & \metric{IFBench} & 29.7 & 36.3 & 39.7 & 26.0 & 31.3 & 37.2 & 36.7 & 31.3 & 27.8 & 36.4 \\
\midrule

\rowcolor{midgrey} -- & \textbf{Knowledge \& QA} & & & & & & & & & & \\
\rowcolor{lightgrey} -- & \metric{MMLU} & 79.0 & 81.9 & 80.9 & 70.2 & 85.8 & 88.7 & 84.6 & 74.6 & 76.1 & 77.1 \\
\rowcolor{lightgrey} -- & \metric{PopQA} & 23.7 & 28.5 & 25.0 & 33.5 & 25.9 & 25.7 & 28.0 & 30.2 & 30.4 & 37.2 \\
\rowcolor{lightgrey} -- & \metric{GPQA} & 41.3 & 47.9 & 48.6 & 27.9 & 54.4 & 61.4 & 44.6 & 45.0 & 39.9 & 36.4 \\
\midrule

\rowcolor{midgrey} -- & \textbf{Chat} & & & & & & & & & & \\
\rowcolor{lightgrey} -- & \metric{AlpacaEval 2 LC} & 42.2 & 69.7 & 59.8 & 19.9 & 67.9 & 84.3 & 81.9 & 65.5 & 39.8 & 38.0 \\
\midrule

\rowcolor{midgrey} -- & \textbf{Tool Use} & & & & & & & & & \\
\rowcolor{lightgrey} -- & \metric{SimpleQA} & 82.3 & 85.3 & 84.7 & - & 86.7 & 91.5 & 90 & - & - & - \\
\rowcolor{lightgrey} -- & \metric{LitQA2} & 47.6 & 53.3 & 55.6 & - & 46.7 & 32 & 26.2 & - & - & - \\
\rowcolor{lightgrey} -- & \metric{BFCL} & 57 & 58.6 & 58.8 & - & 63.1 & 66.3 & 62.8 & - & - & -  \\
\midrule

\rowcolor{midgrey} -- & \textbf{Safety} & 92.1 & 88.9 & 89.5 & 77.1 & 81.6 & 85.8 & 82.2 & 68.8 & 74.4 & 84.2 \\
\bottomrule

\end{NiceTabular}}}
\caption{\textbf{Results of our model Olmo 3.1~32B Instruct} on our post-training evaluation suite. Olmo 3.1~32B Instruct is the best fully-open model at 32B. %
}
\label{tab:32b-instruct-baeslines}
\end{table}

\begin{table}[!h]
\centering
\footnotesize
\setlength\tabcolsep{5pt}
\renewcommand{\arraystretch}{0.95}
\adjustbox{max width=\linewidth}{
{\fontsize{8}{8}\selectfont
\begin{NiceTabular}{@{}Hl
C{35pt}C{35pt}P{35pt}|
C{35pt}C{35pt}C{35pt}C{35pt}C{35pt}C{35pt}@{}}
\toprule
& & \multicolumn{3}{c}{\quad \quad \textbf{\texttt{Olmo 3 7B Instruct}}} & \multicolumn{7}{c}{\textbf{\texttt{Baselines}}} \\
\textbf{Skill} &
& \textbf{SFT} & \textbf{DPO} & \textbf{Final \mbox{Instruct}} & \textbf{Qwen 3 8B} & \textbf{Qwen 3 VL 8B Inst} & \textbf{Qwen 2.5 7B} & \textbf{OLMo 2 7B Inst} & \textbf{Apertus 8B Inst} & \textbf{Granite 3.3 8B Inst} \\
\midrule

\rowcolor{midgrey} -- & \textbf{Math} & & & & & & & & & \\
\rowcolor{lightgrey} -- & \metric{MATH} & 65.1 & 79.6 & 87.3 & 82.3 & 91.6 & 71.0 & 30.1 & 21.9 & 67.3 \\
\rowcolor{lightgrey} -- & \metric{AIME 2024} & 6.7 & 23.5 & 44.3 & 26.2 & 55.1 & 11.3 & 1.3 & 0.5 & 7.3 \\
\rowcolor{lightgrey} -- & \metric{AIME 2025} & 7.2 & 20.4 & 32.5 & 21.7 & 43.3 & 6.3 & 0.4 & 0.2 & 6.3 \\
\rowcolor{lightgrey} -- & \metric{OMEGA} & 14.4 & 22.8 & 28.9 & 20.5 & 32.3 & 13.7 & 5.2 & 5.0 & 10.7 \\
\midrule

\rowcolor{midgrey} -- & \textbf{Reasoning} & & & & & & & & & \\
\rowcolor{lightgrey} -- & \metric{BigBenchHard} & 51.0 & 69.3 & 71.2 & 73.7 & 85.6 & 68.8 & 43.8 & 42.2 & 61.2 \\
\rowcolor{lightgrey} -- & \metric{ZebraLogic} & 18.0 & 28.4 & 32.9 & 25.4 & 64.3 & 10.7 & 5.3 & 5.3 & 17.6 \\
\rowcolor{lightgrey} -- & \metric{AGI Eval English} & 59.2 & 64.0 & 64.4 & 76.0 & 84.5 & 69.8 & 56.1 & 50.8 & 64.0 \\
\midrule

\rowcolor{midgrey} -- & \textbf{Coding} & & & & & & & & & \\
\rowcolor{lightgrey} -- & \metric{HumanEvalPlus} & 69.8 & 72.9 & 77.2 & 79.8 & 82.9 & 74.9 & 25.8 & 34.4 & 64.0 \\
\rowcolor{lightgrey} -- & \metric{MBPP+} & 56.5 & 55.9 & 60.2 & 64.4 & 66.3 & 62.6 & 40.7 & 42.1 & 54.0 \\
\rowcolor{lightgrey} -- & \metric{LiveCodeBench v3} & 20.0 & 18.8 & 29.5 & 53.2 & 55.9 & 34.5 & 7.2 & 7.8 & 11.5 \\
\midrule

\rowcolor{midgrey} -- & \textbf{IF} & & & & & & & & & \\
\rowcolor{lightgrey} -- & \metric{IFEval} & 81.7 & 82.0 & 85.6 & 86.3 & 87.8 & 73.4 & 72.2 & 71.4 & 77.5 \\
\rowcolor{lightgrey} -- & \metric{IFBench} & 27.4 & 29.3 & 32.3 & 29.3 & 34.0 & 28.4 & 26.7 & 22.1 & 22.3 \\
\midrule

\rowcolor{midgrey} -- & \textbf{Knowledge \& QA} & & & & & & & & & \\
\rowcolor{lightgrey} -- & \metric{MMLU} & 67.1 & 69.1 & 69.1 & 80.4 & 83.6 & 77.2 & 61.6 & 62.7 & 63.5 \\
\rowcolor{lightgrey} -- & \metric{PopQA} & 16.5 & 20.7 & 14.1 & 20.4 & 26.5 & 21.5 & 25.5 & 25.5 & 28.9 \\
\rowcolor{lightgrey} -- & \metric{GPQA} & 30.0 & 37.9 & 40.4 & 44.6 & 51.1 & 35.6 & 31.3 & 28.8 & 33.0 \\
\midrule

\rowcolor{midgrey} -- & \textbf{Chat} & & & & & & & & & \\
\rowcolor{lightgrey} -- & \metric{AlpacaEval 2 LC} & 21.8 & 43.3 & 40.9 & 49.8 & 73.5 & 23.0 & 18.3 & 8.1 & 28.6 \\
\midrule

\rowcolor{midgrey} -- & \textbf{Tool Use} & & & & & & & & & \\
\rowcolor{lightgrey} -- & \metric{SimpleQA} & 74.2 & 79.8 & 79.3 & 79.0 & 90.3 & 78.0 & - & - & - \\
\rowcolor{lightgrey} -- & \metric{LitQA2} & 38.0 & 43.3 & 38.2 & 39.6 & 30.7 & 29.8 & - & - & - \\
\rowcolor{lightgrey} -- & \metric{BFCL} & 48.9 & 49.6 & 49.8 & 60.2 & 66.2 & 55.8 & - & - & -  \\
\midrule

\rowcolor{midgrey} -- & \textbf{Safety} & 89.5 & 89.9 & 87.6 & 78.4 & 77.7 & 73.4 & 91.1 & 71.1 & 74.3  \\
\bottomrule

\end{NiceTabular}}}
\caption{\textbf{Overview of \olmothreeinstruct 7B results on the \olmothree post-training evaluation suite}. To reduce variance due to model non-determinism, all numbers are the average over three runs.}
\label{tab:7b-instruct-vs-baselines}
\end{table}

Table~\ref{tab:7b-instruct-vs-baselines} and Table~\ref{tab:32b-instruct-baeslines} demonstrates the results of \olmothreeinstruct 7B and 32B, respectively, on our evaluation suite\footnote{We omit reporting of Essential AI’s Rnj-1 Instruct~\citep{vaswani2025rnj1} due to discrepancies between our observed and their reported numbers. Qualitatively, Rnj-1 behaves like a code specialized model (generates code even for IFEval and Safety chat tasks). Our evaluation framework is meant for general instruct models without code execution for chat tasks. This yields lower scores for Rnj-1 than they report (e.g., 16.1 versus 43.3 on AIME 25, 64.8 versus 75.7 on MBPP+, 79.3 versus 83.5 on HumanEval+) even when we use their recommended general system prompt for turning off code-producing behavior. Thus, we omit it from comparison as we do other specialized models (eg Qwen Coder).}.
In addition to the evaluations used for \olmothreethinking (Section~\ref{sec:posttrain_eval}), we add benchmarks for function-calling.\footnote{For missing function-calling evaluations: \olmotooinstruct and Gemma 2 and 3 don't support this. Apertus and Granite aren't supported by BFCL and we had some difficulties getting the other tasks running. We will update the paper with scores as open git requests are resolved.}
\olmothreeinstruct 7B outperforms Qwen 2.5-7B Instruct, \olmotooinstruct 7B, and Apertus 8B Instruct.
Similarly, \olmothreeoneinstruct 32B outperforms most open models at similar scale, including Qwen 2.5 32B, Qwen 3 32B (No Thinking), Gemma 3 27B, and Apertus 70B. Notably, \olmothreeoneinstruct 32B achieves 39.7 on IFBench outperforming Qwen 3 and Qwen 3 VL at 32B scale. In addition,  \olmothreeoneinstruct 32B achieves 57.9 on AIME 2025, surpassing Qwen 3 32B (No Thinking) by 36.6 points, and closing the gap to Qwen 3 VL 32B-Instruct.

\subsection{Supervised Finetuning with \dolciinstructsft}
We construct \dolciinstructsft by building upon our \olmotooinstruct mixture, making significant improvements to advance general chat, reasoning, and function-calling capabilities.

\subsubsection{Function-calling Training Data} \label{sec:tool-use-sft}
\begin{table}[h]
\centering
\begin{tabular}{l C{64pt} C{64pt} C{64pt} C{64pt} C{64pt}}
\toprule
{\bf Dataset} & {\bf Env. interactions} & {\bf \# Trajectories} & {\bf \# Unique functions} & {\bf \% Multi-turn} & {\bf \% Multi-step} \\ \midrule
\rowcolor{ai2offwhite}
Science QA & Real (MCP) & 22.6K & 8 & - & 42.3\% \\
Web Search QA & Real (MCP) & 6.6K & 3 & - & 76.1\% \\
\rowcolor{ai2offwhite}
SimFC & Simulated & 200K & 42.6K & 42.3\% & 23.8\%
\\ \bottomrule
\end{tabular}
\caption{\textbf{Details of function calling datasets}. Multi-turn refers to multiple user turns per trajectory and multi-step refers to multiple environment interactions per user request.}
\label{tab:func_calling_datasets}
\end{table}

Our goals for curating tool-use training data for \olmothreeinstruct are to provide the model a strong foundation in basic function calling and to expose the model to trajectories demonstrating the effective use of real environments (i.e., MCP servers) to perform tasks. Accordingly, we collect two kinds of trajectories synthesized using LLMs, described below.

\paragraph{Trajectories with real interactions} We collect trajectories demonstrating agents' use of MCP servers to answer queries. All trajectories have a single user turn and multiple agent–environment interactions. We focus on the following domains: 
\begin{itemize}
    \item \textbf{Science QA dataset} contains two broad classes of queries requiring retrieval and reasoning over scholarly content: 1) paper content-based queries, which focus on information present in the abstract or full text of papers and 2) citation graph-based queries, which are about metadata such as authors, venues, and citations. Trajectories associated with the queries are obtained using an agent based on GPT-4.1-mini equipped with the ASTA Scientific Corpus (ASC) MCP server\footnote{\href{https://allenai.org/asta/resources/mcp}{\path{allenai.org/asta/resources/mcp}}}, which provides structured access to metadata and paper content on Semantic Scholar\footnote{\href{https://www.semanticscholar.org/}{\path{www.semanticscholar.org}}}. Additional details about these datasets are provided in Appendix~\ref{sec:appendix_tool_use_data}.

    \item \textbf{Web search QA dataset} is adapted from DR~Tulu~\citep{drtulu}. It consists of a multi-stage pipeline that combines benchmark-derived and real-world queries. Queries are drawn from open-access benchmarks: HotpotQA~\citep{Yang2018HotpotQAAD}, TaskCraft~\citep{Shi2025TaskCraftAG}, and WebWalkerQA (silver)~\citep{wu-etal-2025-webwalker}, as well as from consented, publicly released user prompts from SearchArena~\citep{Miroyan2025SearchAA} and OpenScholar~\citep{Asai2024OpenScholarSS}. We filter the set of queries using GPT-5 to keep only those that both require search and have long-form, verifiable responses. The trajectories for these queries are obtained from a GPT-5 agent equipped with the Serper API\footnote{\href{https://serper.dev/}{\path{serper.dev}}}, which provides access to a Google search tool and a tool for fetching webpages given their URLs. Additional details about query filtering and trajectory generation can be found in Appendix~\ref{sec:appendix_tool_use_data}.
\end{itemize}

\paragraph{Trajectories with simulated interactions} While training on trajectories with executable environments is expected to teach the model to effectively deal with real environment outputs and handle unexpected errors, it is difficult to curate such trajectories at scale, thus potentially limiting the model's generalization to unseen tools at inference time. To fill this gap, we also create a dataset of synthetic trajectories with LLM-simulated environments which are much easier to scale. We call this dataset \textbf{SimFC}. We start with a large pool of tool sets or APIs from existing datasets (e.g., xLAM~\citep{Liu2024APIGenAP}, ToolACE~\citep{Liu2024ToolACEWT}), and from publicly available MCP servers, and prompted LLMs (GPT-4o, GPT-4.1, and GPT-5) to generate entire trajectories including simulated user queries, environment responses, and assistant messages. We design prompts to ensure the dataset contains a variety of interaction patterns including multi-turn, multi-step, and refusals due to inadequate information or tools. Additional details about this dataset and illustrative prompts used for generation can be found in Appendix~\ref{sec:appendix_tool_use_data}, Figure~\ref{fc-multi-turn-prompt}, and Figure~\ref{fc-refusals-prompt}.

\paragraph{Balancing function diversity with interaction complexity} As illustrated by the statistics in Table~\ref{tab:func_calling_datasets}, the two types of trajectories have key differences. SimFC has a large number of trajectories with diverse sets of functions. 
We find that synthesizing trajectories with multiple user turns (multi-turn trajectories) is relatively easier than those with multiple assistant-environment interactions per user request (multi-step trajectories). 
However, the latter class usually corresponds to more complex tasks.  
On the other hand, the datasets with real interactions, while smaller in size, are naturally more complex in terms of multi-step interactions.

\paragraph{Unified data format} Across all tool-use data, we adopt consistent tool definition and tool-calling formats. 
We find that unifying format to be crucial for stable and high-quality tool-use behavior. 
Particularly, we use the OpenAPI specification\footnote{\href{https://swagger.io/specification/}{\path{swagger.io/specification/}}} for all tool definitions and represent all function calls as pythonic code blocks. 
We provide tool specifications in the system prompt, encapsulate tool calls with XML tags within the assistant role, and present environment outputs to the model within a special environment role. 
We also extend the tokenizer's vocabulary with dedicated special tokens corresponding to these tags. 
Unlike \olmothreethinking, preliminary suggest this approach to be more effective for tool-use training than encoding \texttt{<functions>}, \texttt{</functions>}, \texttt{<function\_calls>}, and \texttt{<function\_calls>} as regular  text.

\paragraph{Evaluating function calling} %
We evaluate the function calling capabilities of \olmothreeinstruct in terms of \textit{intrinsic function calling} and \textit{extrinsic task completion} accuracies using different benchmarks. We use the Berkeley Function Calling Leaderboard (BFCLv3)~\citep{patil2025bfcl} to evaluate intrinsic function calling accuracy. This benchmark focuses on models' ability to choose the relevant functions and the right values for their arguments to accomplish a given task in settings that require one or more interactions with simulated users and environments. We evaluate task completion accuracy of \olmothreeinstruct in comparison with similar models when they are deployed as agents with access to tools served via Model Context Protocol (MCP) servers. Particularly, we use the Asta Scientific Corpus (ASC) tool~\citep{bragg2025astabench} that serves eight functions for accessing scientific literature, and the Serper API which provides Google search tool and web browsing functionalities. To evaluate models' usage of the ASC tools, following~\citet{bragg2025astabench}, we use a subset of 75 questions from LitQA2~\citep{skarlinski2024language} for which the associated papers can be found in ASC's index. We evaluate the models' usage of search and browsing tools using a subset of SimpleQA\footnote{\href{https://huggingface.co/datasets/akariasai/sampled_simpleqa}{\path{huggingface.co/datasets/akariasai/sampled_simpleqa}}}~\citep{wei2024measuring}.

We use the official Gorilla repository\footnote{\href{https://github.com/ShishirPatil/gorilla}{\path{github.com/ShishirPatil/gorilla}}} for BFCLv3 evaluations. For LitQA2 and SimpleQA, we implement a basic function-calling agent using OpenAI's Agent SDK. This agent uses the tools provided by the relevant MCP server\footnote{We the same setup introduced by \citet{drtulu} for DR~Tulu.}, and interacts with the environment by iteratively making function calls and processing the outputs of executing them to solve the given tasks.
For LitQA2 and SimpleQA, we also measure model performance when deployed in a \textit{No-Tools} setting, in which we provide no tools to the agents and they are expected to solve the tasks entirely from the models' parametric knowledge. We use a zero-shot evaluation for all these benchmarks. We sample from models at temperature 0 and, for LitQA2 and SimpleQA, allow the agents at most 10 turns to finish each task. We run each evaluation three times and report the average accuracy. We release our code\footnote{\href{https://github.com/allenai/mcp-tool-eval}{\path{github.com/allenai/mcp-tool-eval}}} for running our MCP-based tool-use evaluations.

\subsubsection{Curating \dolciinstructsft}
\paragraph{Step 1. Sourcing Prompts and Completions}

Our prompt collection includes all our new function-calling data~(Section~\S\ref{sec:tool-use-sft}), new prompts for instruction following (see Section~\S\ref{sec:think-sft}) and science, and more chat prompts from WildChat~\citep{zhao2024wildchat}. For examples that originally contained reasoning traces (such as the OpenThoughts3 science subset described in Section~\S\ref{sec:think-sft}), we remove the reasoning traces and special tokens. We also update completions from older models such as GPT-3.5 and GPT-4 with completions from GPT-4.1. We show a summary of our instruct SFT mix in Table~\ref{tab:instruct_prompt_mix}.

\paragraph{Step 2: Filtering \& Mixing}

We follow the same filtering and mixing procedure detailed in Section~\ref{sec:think-sft}. For \olmothreeinstruct, our base mix is 100K examples from an updated intermediate mix based on the \olmotoo SFT mix. We show results of our data-mixing experiments on \olmotoo in Table~\ref{tab:instruct-sft-ablate}.

\begin{table}[tbp]
\centering
\footnotesize
\begin{tabular}{@{}l >{\columncolor{ai2pink!18}}c c c c c c c c c@{}}
\toprule
& \multicolumn{9}{c}{\texttt{Subset of \olmothreeinstruct Benchmarks}} \\
\cmidrule(lr){2-10}
\textbf{Name} & \cellcolor{ai2pink!18}\textbf{Avg.} & \textbf{MMLU} & \textbf{BBH} & \textbf{GPQA} & \textbf{MATH} & \textbf{GSM8K} & \textbf{CHE} & \textbf{AE} & \textbf{IFEval} \\
\midrule
Base mix & 29.0 & 50.0 & 29.5 & 25.2 & 6.6 & 30.1 & 23.2 & 5.8 & 61.7 \\
Base mix + Aya & 29.1 & 51.9 & 28.2 & 28.1 & 6.9 & 31.4 & 21.3 & 4.9 & 60.3 \\
Base mix + Code & 28.7 & 51.1 & 28.8 & 25.0 & 6.9 & 28.2 & 26.8 & 5.8 & 57.3 \\
Base mix + Flan & 30.3 & 51.9 & 35.0 & 26.8 & 6.6 & 34.7 & 21.3 & 5.8 & 60.3 \\
Base mix + IF & 30.7 & 51.4 & 24.7 & 25.5 & 7.9 & 42.2 & 14.6 & 5.5 & 74.1 \\
Base mix + Math & 29.3 & 49.9 & 23.9 & 29.2 & 14.2 & 39.7 & 18.3 & 5.4 & 54.0 \\
Base mix + Safety & 27.0 & 51.7 & 28.3 & 24.8 & 6.5 & 28.2 & 14.0 & 6.8 & 56.0 \\
Base mix + Science & 29.4 & 53.4 & 25.3 & 28.1 & 8.3 & 34.9 & 20.7 & 6.8 & 57.3 \\
Base mix + Wildchat & 30.9 & 51.9 & 30.7 & 23.7 & 6.9 & 32.2 & 23.2 & 19.2 & 59.7 \\
\bottomrule
\end{tabular}
\caption{\textbf{Results of our instruct SFT mixing ablations} on top of \olmotoo.}
\label{tab:instruct-sft-ablate}
\end{table}

\paragraph{Starting from \olmothreethinking SFT} We train the SFT stage of \olmothreeinstruct starting from the \olmothreethinking SFT model as shown in Figure~\ref{fig:olmo3_pipeline} to give it a ``warm-start.'' We found that this significantly improves the performance of the Instruct model, as shown by the results in Table~\ref{tab:sft-reasoning-first}.

\begin{table}[tbp]
\centering
\footnotesize
\resizebox{\textwidth}{!}{
\begin{tabular}{@{}l >{}c c c c c c c c c c c@{}}
\toprule
& \multicolumn{11}{c}{\textbf{Subset of \olmothreeinstruct Benchmarks}} \\
\cmidrule(lr){2-12}
\textbf{Name} & \textbf{Avg.} & \textbf{BBH} & \textbf{GPQA} & \textbf{MATH} & \textbf{GSM8K} & \textbf{OMEGA} & \textbf{CHE} & \textbf{MBPP} & \textbf{LCB} & \textbf{AE} & \textbf{IFEval} \\
\midrule
No thinking SFT & 44.5 & 46.5 & 29.7 & 60.3 & 87.6 & 8.6 & 63.8 & 54.1 & 13.0 & 27.0 & 81.0 \\
With thinking SFT & 47.8 & 46.6 & 34.4 & 65.9 & 91.1 & 12.2 & 68.7 & 57.1 & 17.1 & 27.1 & 84.7 \\
\midrule
Gain from thinking SFT first & 3.3 & 0.1 & 4.7 & 5.6 & 3.5 & 3.6 & 4.9 & 3.0 & 4.0 & 0.1 & 3.7 \\
\bottomrule
\end{tabular}}
\caption{\textbf{Results of training an intermediate \olmothreeinstruct 7B checkpoint} with and without thinking SFT first.}
\label{tab:sft-reasoning-first}
\end{table}

\subsection{Preference Tuning with \dolciinstructdpo}
\begin{table}[t!]
\centering
\setlength\tabcolsep{5pt}
{\small
\begin{tabular}{ll R{1.5cm} R{1.5cm} l}
\toprule
{\bf Category} & {\bf Prompt Dataset} & {\bf \# Prompts used in SFT} & {\bf \# Prompts used in DPO} & {\bf Reference} \\
\midrule
\rowcolor{ai2offwhite} Chat \&  & WildChat & 302,406 & 30,248 & \citet{zhao2024wildchat} \\
\rowcolor{ai2offwhite} Precise IF & \dolciinstruct Precise IF & 136,833 & 35,057 & -- \\ 
\rowcolor{ai2offwhite} & \dolciinstruct Persona Precise IF & -- & 6667 & \cite{lambert2024tulu3} \\ 
\rowcolor{ai2offwhite}  & OpenAssistant & 7,132 & 493 & \citet{kopf2024openassistant} \\
Math & \tulu Persona MATH & 149,958 & 14,728 & \citet{lambert2024tulu3} \\
 & \tulu Persona Algebra & 19,999 & 2,025 & \citet{lambert2024tulu3} \\
 & \tulu Persona GSM & 49,980 & 5,011 & \citet{lambert2024tulu3} \\
 & OpenMathInstruct 2 & 50,000 & 5,325 & \citet{toshniwal2024openmathinstruct} \\
\rowcolor{ai2offwhite} Coding & \dolciinstruct Python Algorithms  & 186,345 & 24,096 & -- \\
 \rowcolor{ai2offwhite}  & \tulu Persona Python & 34,999 & 4,598 & \citet{lambert2024tulu3} \\
\rowcolor{ai2offwhite}  & Evol CodeAlpaca & 107,270 & 12,953 & \cite{luo2023wizardcoder} \\  
Safety & CoCoNot & 10,957 & 2,203 & \citet{brahman2024art} \\
 & WildGuardMix & 49,373 & 12,037 & \citet{han2024wildguard} \\
 & WildJailbreak & 49,965 & 12,431 & \citet{jiang2024wildteaming} \\
\rowcolor{ai2offwhite} Science & SciRiff & 4,557 & 8,874 & \citet{wadden2024sciriff} \\
\rowcolor{ai2offwhite} & \dolciinstruct OpenThought3+ Science & 99,268  & 26,134 & \citet{guha2025openthoughts} \\
Multilingual & Aya & 99,987 & 6,523 & \citet{singh2024aya} \\
\rowcolor{ai2offwhite} Other & TableGPT & 5,000 & 1,218 & \citet{zha2023tablegpt} \\
\rowcolor{ai2offwhite} & FLAN & 89,981 & 16,120 & \citet{wei2021flan} \\
\rowcolor{ai2offwhite} & Logic Puzzles & 159,882 & -- & -- \\
\rowcolor{ai2offwhite} & Verifiable Reasoning & 310,572 & -- & -- \\
\rowcolor{ai2offwhite} & \dolciinstruct Hardcoded & 69 & -- & -- \\
\rowcolor{ai2offwhite} & \dolciinstruct Tool Use & 227,579 & -- & --\\
Multiturn & \dolciinstruct Self-Talk & -- & 5,000 & -- \\
& \dolciinstruct Synthetic Context & -- & 5,000 & -- \\
\rowcolor{ai2offwhite} Not used in SFT & DaringAnteater & -- & 878 & \cite{wang2024helpsteer2} \\
\rowcolor{ai2offwhite} & UltraFeedback & -- & 22,303 & \cite{cui2023ultrafeedback} \\
{\bf{Total}} & & 2,152,112 & 259,922 &  \\
\bottomrule
\end{tabular}}
\vspace{3pt}
\caption{\olmothreeinstruct prompt sources for both SFT and DPO.}
\label{tab:instruct_prompt_mix}
\end{table}

We create \dolciinstructdpo by extending the strong base of our delta-learning heuristic preferences (Section \S\ref{sec:thinking_dpo_recipe}) with further curated preference signals to enhance our model's behavior in general use settings. 
We enrich our heuristic data with contrastive pairs from an improved GPT-judge pipeline for general alignment.
Additionally, user interaction with LMs commonly requires multi-turn conversational capabilities, so we introduce synthetic multi-turn conversations to our preference data. We also observe that preference-data pipelines often promote overly verbose responses; we introduce counteracting interventions to promote brevity in model responses by mitigating length bias in the preference data.

\subsubsection{Preference Signals}
\label{sec:dolci-instruct-dpo}
\dolciinstructdpo is constructed from a composite of several preference signals to promote model capabilities and general usability:

\paragraph{Delta-learning heuristic pairs} Similar to \dolcithinkdpo, we construct heuristic contrastive pairs by generating chosen responses with a large model (Qwen3 32B) and rejected responses with a small model (Qwen3 0.6B) following \cite{geng2025delta}. Note that we turn off thinking mode, as we do not need internal thinking traces.

\paragraph{Delta-aware GPT-judged pairs} We additionally generate GPT-judged preference pairs to add a further source of preference signal. Our initial attempts to modernize the UltraFeedback pipeline from \olmotoo and \tulu by improving the quality of the LLM judge (GPT-4o $\rightarrow$ GPT-4.1) and updating our data-generator model pool do not yield gains and even hurt model performance relative to the \olmotoo preference dataset baseline. 
We speculate that this failure is due to the fact that the majority of our data generators are high-quality, very capable models; 
hence on average there was minimal meaningful contrast between the resulting chosen and rejected pairs. 
To mitigate this, we explicitly introduce {\bf{delta-aware}} interventions designed to \textit{lower} the quality of the rejected response. We 1) ensure that responses from weaker models are always present in the response set judged for each prompt, and 2) select the \textit{worst} response as the rejected completion to maximize the resulting delta. 
We find these ``delta-maximizing'' interventions to be critical for the quality of preference pair data; see our findings in Section~\S\ref{sec:instruct_findings} for details.

\paragraph{Multi-turn preferences} To ensure \olmothree's usability in realistic multi-turn conversations, we further add a multi-turn preference dataset with prompts synthetically extended from the \tulu-DPO dataset. Preference pairs differed in only the last turn of the conversation to avoid ambiguity in quality ranking between turns of the same conversation. Synthetic conversations are generated with two methods: 1) {\bf{self-talk}} extending the original prompt into a multi-turn conversation with LLM-generated follow-up requests and 2) {\bf{synthetic-context}} created by generating related, independent questions or paraphrases of the initial prompt to use as previous user turns with associated completions. The combination of these generation methods ensures diversity in generated conversations. Final turns are generated with the delta-learning heuristic~\citep{geng2025delta}; chosen/rejected completion pairs are generated by either GPT-4o and GPT-3.5 or Qwen 3 32B and Qwen 3 0.6B (both no-thinking) respectively.

\paragraph{Controlling length bias} Preference data often has a length bias: the chosen responses are significantly longer than the rejected responses. This comes from sourcing synthetic response pairs where historically more information has been treated as more helpful by both LLM judges and preference heuristics. Namely, LLM judges such as the GPT judge in our pipeline tend to prefer longer responses. Similarly, we empirically observe that preference pairs made with the delta-learning heuristic also exhibit length bias; larger models generate longer responses (\autoref{fig:ushape_dpo}). 
Thus, models often learn this length bias in addition to the intended useful quality signal during preference tuning, after which its generation length per prompt increases significantly. 
While this increased length is empirically useful for reasoning tasks, excessive verbosity can be undesirable for common real-use settings (see an example in \autoref{fig:conciseness}). We seek to strike a balance by filtering the chat and multi-turn subsets of our preference data to limit the length difference between the chosen and rejected responses to 100 tokens.
\begin{figure}[t]
    \centering
    \adjustbox{max width=0.35625\linewidth}{\includegraphics{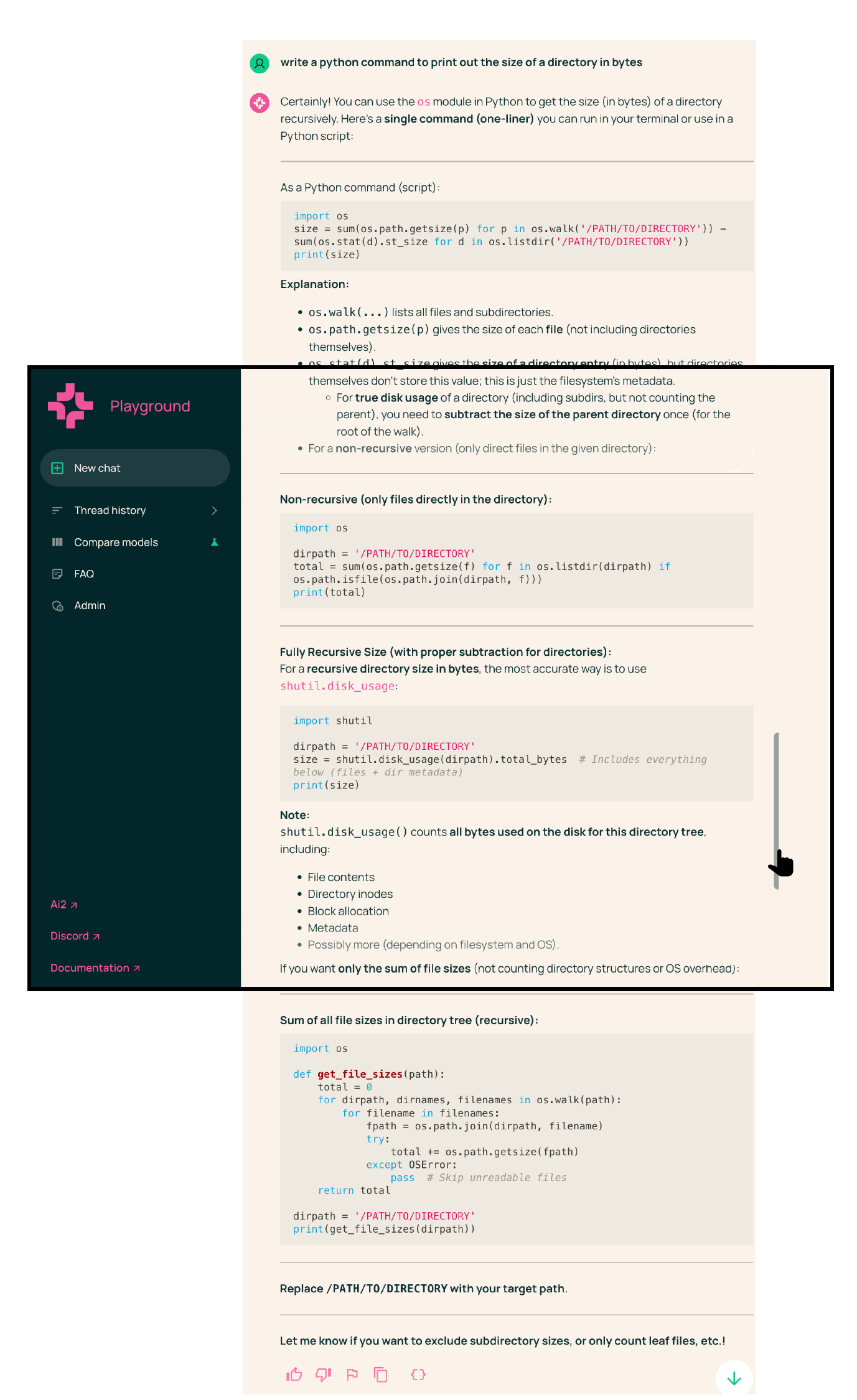}}
    \hspace{10pt}
    \raisebox{.3\height}
    {\adjustbox{max width=0.59375\linewidth}{\includegraphics{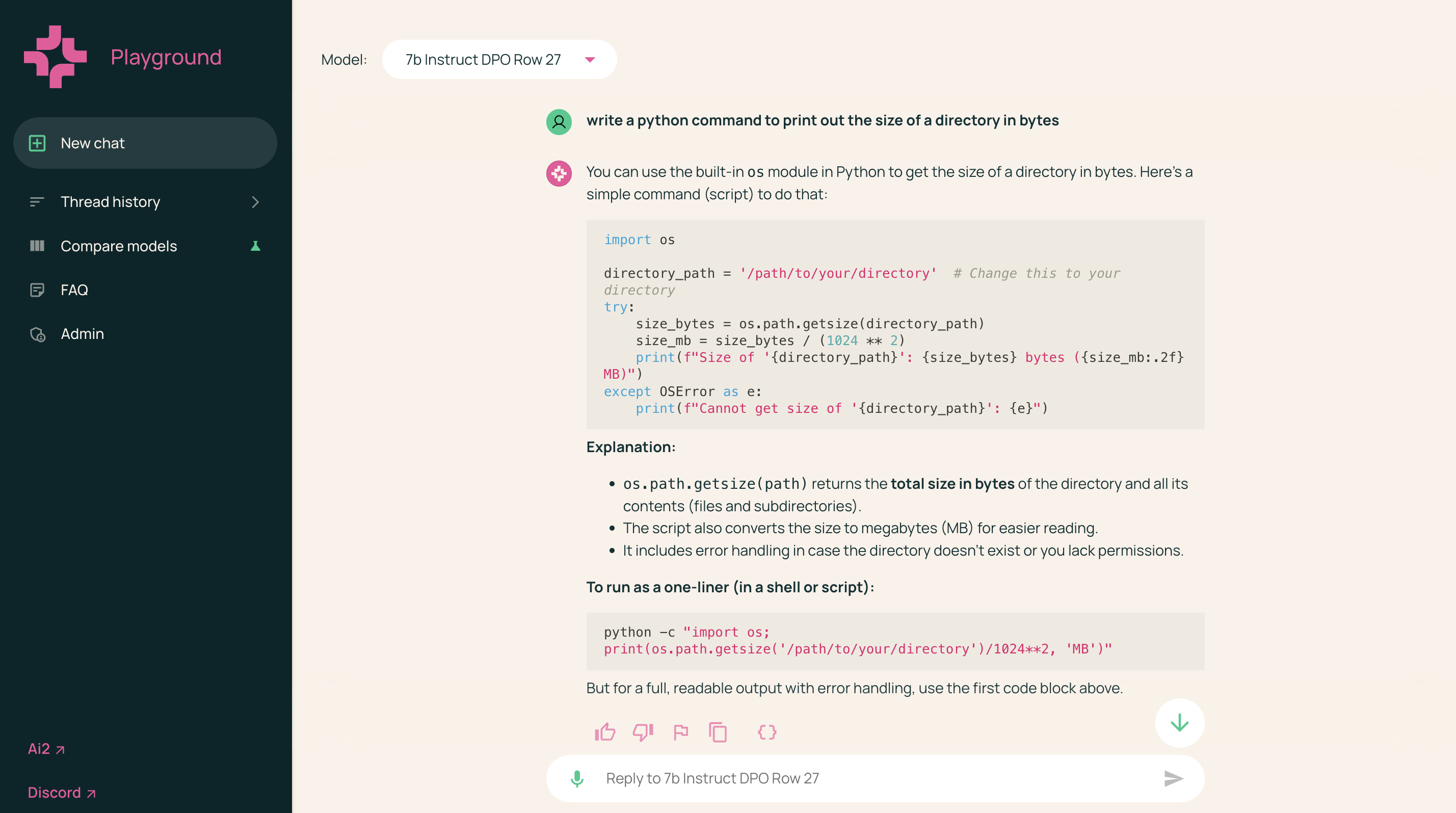}}}
    \caption{\textbf{Length control promotes concise, usable responses.} On the left is a response from a development model preference-tuned without length control; on the right, a response to the same prompt from \olmothreeinstruct-DPO (with length control). Promoting brevity in model responses makes the response easier to read and understand.}
    \label{fig:conciseness}
\end{figure}

\subsubsection{Prompt Mixing}
\label{sec:dpo_mixing}
Our prompt pool for GPT-judged and delta-learning heuristic pairs (see Table~\ref{tab:instruct_prompt_mix}) is derived from the \dolciinstructsft dataset supplemented with the DaringAnteater and UltraFeedback subsets from the \olmotoo~7B preference dataset. Because DPO performance does not monotonically increase with more data (see Figure~\ref{fig:ushape_dpo}), we optimize the prompt distribution as a ratio within a set data budget and treat dataset size as a hyperparameter when training.

To determine our final preference-tuning prompt distribution, we begin with near-uniform random sampling\footnote{We decided early to truncate the number of Wildchat prompts to be at most 35\% of the prompt mix. If you read Wildchat prompts for a month, you would too.} of 100K examples as an empirically strong baseline prompt mix. We then perform ablations of prompt-domain subsets to determine the impact of preference pairs from each domain subset. Additionally, we perform experiments that pair 50K samples of our base mix with 50K samples from a given domain, allowing us to understand the effects of upsampling each prompt domain.

Notably, prompt-domain distributions do not consistently align with the \textit{contrast} exhibited in the response pair and thus in improvements in the corresponding downstream evaluation domains. For example, upsampling code prompts led to the counter-intuitive effect of decreasing code benchmark performance (see Table \ref{tab:dpo_mixing_exps} in the Appendix). For determining our final mix, we create nine candidate mixes based on expert intuition gained from our ablations, comparing these hand-crafted mixes against the uniform sampling baseline. Our final mix is determined empirically; we find that our hand-crafted mixes outperformed random sampling.

\subsubsection{Training}
We follow the same training setup as \olmothreethinking and sweep the same hyperparameters, namely learning rate and dataset size. We further sweep different length-control interventions by creating datasets with differing token cutoffs for length filtering. We select the best-performing checkpoint of each length budget and then select the final \olmothreeinstruct-DPO checkpoint based on qualitative vibe tests and performance-vs-length analysis.

\subsection{Reinforcement Learning with \dolciinstruct-RL}
\label{sec:RLVR-instruct}

For our RL training stage, we modify the pool of prompts from \dolcithinkrl (Section~\S\ref{subsec:rl-thinking-data}) by 1) utilizing less challenging datasets in the math and code domains, and 2) skipping the offline difficulty filtering, as our instruct model focuses more on general instruction following rather than complex reasoning.

\subsubsection{Training}

Following our \olmothreethinking recipe, we train \olmothreeinstruct on a mixture of general chat, math, and code data.\footnote{Preliminary experiments indicated that alternative RL setups—for example, first warming up on math-only data and then switching to a mixed dataset without math—resulted in suboptimal performance.} We likewise employ \olmothreerl for training, with a maximum response length of 8K tokens for 7B and 16K for 32B\footnote{We experiment with both 8K and 16K length training for 7B and 32B; while evaluation scores are minimally impacted by different lengths, we notice undesirable behaviors when qualitatively testing 7B-16K and 32B-8K configurations in an internal demo.}. Since our goal for \olmothreeinstruct is to avoid generating excessively long outputs and preserve general usability, we apply RL on top of two DPO candidates: one that achieved the best average performance, and another with slightly lower performance but better qualitative “vibe test.'' We then choose the final RL checkpoint based on final average performance, length analysis, and
vibe test. Concretely, we begin by ranking checkpoints by average score; in the case of ties, we place more emphasis on datasets that do not scale with test-time compute (e.g., MATH and AIME performance increase with response length) to avoid biasing our selection towards models with overly long responses. Finally, we apply the vibe test to identify regressions or undesirable behaviors that may fall outside the scope of our evaluation suite.

\subsection{Key Findings}
\label{sec:instruct_findings}

Below, we summarize our key findings across all 3 stages of \olmothreeinstruct training:

\paragraph{Starting from the \olmothreethinking SFT is helpful} We find that training \olmothreeinstruct on top of the \olmothreethinking SFT both increases model performance on benchmarks, as shown in Table~\ref{tab:sft-reasoning-first}. 
Importantly, average model response length is minimally affected by this strategy: \olmothreeinstruct SFT checkpoints produce succinct answers with no remnants of thinking traces.

\paragraph{High contrast in preference pairs drives DPO improvements} We observe that a high contrast between completions is critical for achieving improvements during DPO training (Table~\ref{tab:dpo_different_signals}). Using LLM-judge pipelines requires carefully thinking about maximizing the delta between chosen and rejected responses. Our initial attempts to modernize the \olmotoo preference data pipeline by improving the models used to generate responses failed to yield any improvements beyond the \olmotoo data baseline (Table \ref{tab:dpo_different_signals}). This is likely because the models used for synthetic completions were universally too good: the chosen and rejected responses no longer had meaningful contrast. Extending prior findings that high contrast pairs are critical for performance~\citep{geng2025delta, d2025anchored}, we introduce interventions to explicitly lower the quality of the rejected response and therefore increase the magnitude of the quality delta in the preference pair. These resulting \textit{delta-aware} GPT pairs significantly outperform the \olmotoo preference data.

\paragraph{Combining different preference signals improves overall performance} We combine delta-learning heuristic data with GPT-judged preference pairs to get the ``best of both worlds.'' Empirically, tuning with either delta-learning or GPT-judged pairs yields a different spread of gains; we find that these gains are complementary. Combining both sources of preference signal outperforms using either alone (Table~\ref{tab:dpo_different_signals}).

\paragraph{The ideal amount of preference data depends on the downstream task}
Preference-tuned model performance peaks with different amounts of training for different downstream task domains. We plot preference-tuning performance for example tasks across varying amounts of delta-learning heuristic pairs\footnote{Initial experiments with GPT-judged data showed similar trends.} in Figure~\ref{fig:ushape_dpo}. Further optimization beyond these optimal points hurts downstream performance, consistent with theoretical results showing that early stopping is important for preference tuning~\citep{azar2023generaltheoreticalparadigmunderstand, geng2025delta}. %
Practically, this informs our training approach: we sweep learning rate and dataset size to control the amount of total optimization, and pick the best-performing setting via our development evaluation set.

\begin{figure}[t]
    \centering
    \adjustbox{max width=.55\linewidth}{\includegraphics{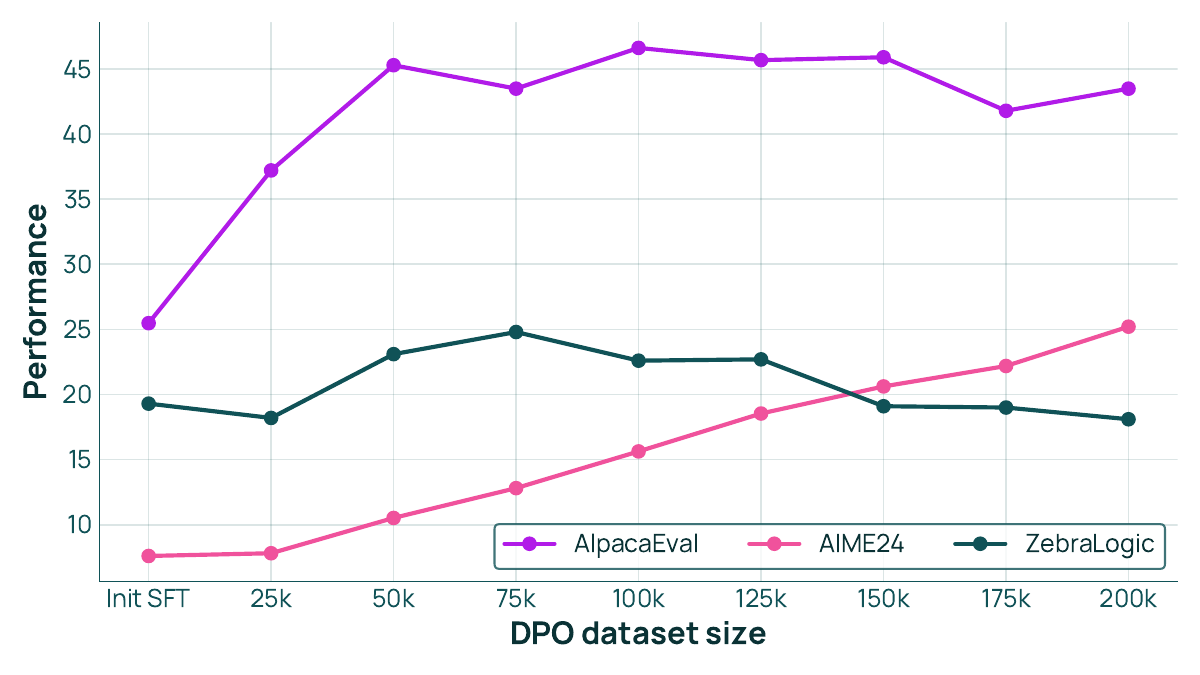}}
    \adjustbox{max width=.42\linewidth}{\includegraphics{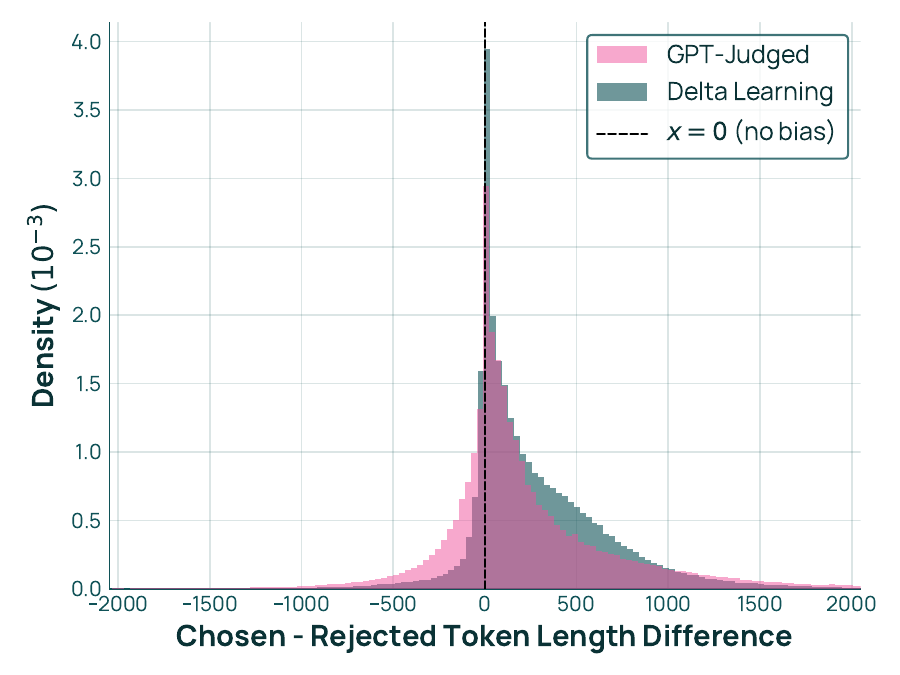}}
    \caption{\textbf{Effect of dataset size and filtering for preference data}. Ideal preference dataset size depends on the downstream task (left). Both AlpacaEval and ZebraLogic performance increase up to around 75–100K samples, beyond which further data scaling hurts or does not help. In contrast, AIME2024 does not saturate before the point at which AlpacaEval and ZebraLogic begin to see drops in performance. Hence, to strike an ideal balance between all downstream tasks, we sweep dataset size as a hyperparameter during training. Unfiltered preference data exhibits a length bias (right). A significant portion of the data distribution has longer chosen than rejected completions. For example, the 80th percentile of token difference for the GPT-judged data is 538 tokens and for the delta-learning heuristic pairs is 564 tokens.}
    \label{fig:ushape_dpo}
\end{figure}

\paragraph{Concise, usable model outputs from preference tuning can boost RL performance} Applying length control during DPO substantially reduces the model’s average generation length, allowing us to trade off some performance for improved conciseness and overall usability. While this reduction in length comes with lower scores on length-sensitive evaluations—particularly math benchmarks such as AIME and MATH—our internal qualitative assessments (``vibe tests'') almost uniformly preferred the shorter, more direct model. We make a conscious decision to prioritize usability.

Crucially, despite the lower benchmark performance at the DPO stage, length control ultimately yields to a more performant model post RL.
At 7B scale, we conjecture that this arises from the RL training context window: with a fixed context window (8K), a shorter model may be ``more intelligent per token,'' allowing it to leverage the available budget more effectively during optimization. 
Thus, what initially appeared to be a tradeoff between usability and performance ultimately produced improvements in both. 
Moreover, we found that RL training progresses more reliably when initialized from the length-controlled DPO policy. 
Across most benchmarks, performance improves more steadily compared to RL runs starting from a higher-scoring but uncorrected DPO checkpoint, which tends to show earlier signs of instability or degradation. This further supports the role of concise preference-tuned models as advantageous starting points for RL.

\paragraph{Need for tools} We assess how much of \olmothreeinstruct's performance on LitQA2 and SimpleQA can be attributed to tool use by measuring the delta of the model performance on the benchmarks between answering the questions only from parametric memory (``No tools'' setting) and doing so using tools. Table~\ref{tab:tools-vs-no-tools} shows these deltas in comparison to those from three Qwen models. All models benefit significantly from tool use on SimpleQA. However, Qwen models, unlike \olmothreeinstruct 7B, mostly seem to rely on parametric knowledge for LitQA2, with two of the models even losing performance when provided with tools.
\begin{table}[t]
\centering
\setlength\tabcolsep{5pt}
\renewcommand{\arraystretch}{1.2}
\adjustbox{max width=\linewidth}{
\begin{NiceTabular}{lcccccc}
\toprule
& \multicolumn{3}{c}{\textbf{LitQA2}} & \multicolumn{3}{c}{\textbf{SimpleQA}} \\
\cmidrule(lr){2-7}
& \textbf{No tools} & \textbf{ASC} & \textbf{$\Delta$} & \textbf{No tools} & \textbf{SBT} & \textbf{$\Delta$} \\ 
\midrule
Olmo 3 Instruct 7B & 24.4 & 38.2 & 13.8 & 3.3 & 79.2 & 75.9 \\
Qwen 3 8B (w/o reasoning)  & 34.7 & 39.6 & 4.9 & 2.0 & 79.0 & 77.0 \\
Qwen 3 VL 8B Instruct  & 34.7 & 30.7 & -4.0 & 9.3 & 90.3 & 81.0\\
Qwen 2.5 7B  & 36.0 & 29.8 & -6.2 & 3.3 & 78.0 & 74.7 \\
\bottomrule
\end{NiceTabular}}
\caption{\textbf{Comparison of agents' performance with and without access to tools on LitQA2 and SimpleQA}. ASC refers to Asta Scientific Corpus tools and SBT refers to search and browsing tools.}
\label{tab:tools-vs-no-tools}
\end{table}

\begin{table}[t]
\centering
\small
\begin{tabular}{@{}l >{\columncolor{ai2pink!18}}c ccccccccc@{}}
\toprule
& \multicolumn{10}{c}{\textbf{Subset of \olmothreeinstruct Benchmarks}} \\
\cmidrule(lr){2-11}
\textbf{Name} & \cellcolor{ai2pink!18}\textbf{Avg.} & \textbf{MMLU} & \textbf{BBH} & \textbf{GPQA} & \textbf{AGI} & \textbf{MATH} & \textbf{CHE} & \textbf{LCB} & \textbf{IFEval} & \textbf{AE2} \\
\midrule
Dev. 7B SFT ckpt & 51.9 & 67.6 & 47.7 & 30.2 & 62.0 & 65.5 & 69.3 & 17.9 & 83.2 & 23.8 \\
\olmotoo preference data & 55.5 & 69.4 & 55.6 & 33.7 & 63.6 & 71.3 & 73.7 & 12.7 & {\bf 84.5} & 35.2 \\
\midrule
Updated GPT UltraF pipeline & 55.4 & 67.6 & 51.2 & 31.5 & 61.8 & 72.2 & 71.5 & 14.7 & 80.8 & 47.5 \\
+ Sample weak models & 56.3 & 68.4 & 50.4 & 33.9 & 63.8 & 71.6 & 74.3 & 18.2 & 81.9 & 44.4 \\
+ Min score rejected & 57.4 & 68.5 & 53.6 & 34.4 & 64.2 & 72.6 & {\bf 75.2} & 19.1 & 82.3 & 47.0 \\
Delta learning only & 57.6 & 68.7 & 49.5 & {\bf 35.5} & {\bf 64.6} & 79.1 & 73.9 & {\bf 22.0} & 78.6 & 46.1 \\
\midrule
Delta learning + GPT & {\bf 60.4} & {\bf 69.4} & {\bf 66.9} & 34.6 & 64.3 & {\bf 80.0} & 74.1 & 21.1 & 83.0 & {\bf 49.8} \\
\bottomrule
\end{tabular}
\caption{\textbf{Comparing sources of preference signals}. Preference pairs created with the delta-learning heuristic (chosen = large model response, rejected = smaller model response) and pairs created with our delta-aware LLM-judge pipeline yield a different spread of gains, suggesting that they provide different preference signals. These signals are complementary; combining them both yields the largest average gain. Our final \olmothreeinstruct preference data greatly outperforms our previous \olmotoo preference data.}
\label{tab:dpo_different_signals}
\end{table}

\newpage
\section{\olmothreerlzero}
\label{sec:rlzero}

RL has become a key part of recent LLM pipelines in part due to prominent open models such Deepseek R1-Zero \citep{guo2025deepseek}, which notably leverages RL training on top of a base model to bootstrap complex reasoning behavior \citep{marjanović2025deepseekr1thoughtologyletsthink}, and due to the rapid adoption of closed reasoning models such as OpenAI's o1-series and Gemini with Thinking.
This has made RLVR finetuning from a base model the standard large-scale benchmark for RL algorithms \citep{liu2025prorl,yu2025dapo,luo2025deepscaler}.
To date, leading open RLVR benchmarks and algorithms train on top of open-weights models that do not reveal their pretraining or midtraining data \citep{scalingllama3,qwen3}.
This limits the ability for the community to study the role of pretraining data on RLVR performance.
It can lead to a myriad of issues with benchmark evaluations being contaminated, e.g., midtraining data containing data from the evaluation set, which makes spurious rewards as effective as true rewards~\citep{shao2025spuriousrewardsrethinkingtraining,wu2025reasoning}, or improvements from fixing prompt templates outweighing the improvements from RL~\citep{liu2025understanding}.

We therefore release a fully-open dataset \dolcirlzero, an algorithmic RL zero setup for \olmothree, and open-source \olmothreerl code to enable clear benchmarking for the ecosystem.
We perform RLVR from \olmothreebase over five benchmarking domains to create the \olmothreerlzero family: math, code, precise instruction following (IF), general chat, and a mix of all listed sub-domains.
In all cases, we further decontaminate \dolcirlzero from pretraining and midtraining data to guarantee our setup carefully studies the effect of RLVR without data leakage confounding our conclusions.

\subsection{Reinforcement Learning From Base with \dolcirlzero}
\label{sub:rlzero-data-setup}

\paragraph{Data}

We create \dolcirlzero, an effective RL-Zero training dataset.
For math, we aggressively filter DAPO Math \citep{yu2025dapo}, Klear-Reasoner Math \citep{su2025klear},  Open-Reasoner-Zero (Orz)~\citep{hu2025open}, and  Omega~\citep{Sun2025OMEGACL}.
We deduplicate DAPO and remove all non-English examples. As Klear-Reasoner, Orz, and Omega are much larger datasets, we further group questions via semantic clustering across Klear-Reasoner, Orz, and Omega, and select one representative question per cluster, in addition to including DAPO.
We further decontaminate against both pretraining and evaluation data following \autoref{sec:think-sft} and perform offline filtering, removing prompts fully solved in 8 out of 8 sample completions by the final base model.
This results in a dataset of 13.3K math prompts. Data for code, instruction-following, and general chat are subsampled from \dolcithinkrl (Section~\S\ref{subsec:rl-thinking-data}).

\paragraph{Prompt and eval template}
Confirming the findings of \citet{liu2025understanding}, we find that ``simple'' prompt templates greatly outperform standard post-trained templates (e.g., \texttt{<think></think>}) when training from a purely midtrained model, as \dolminostoo excluded most special formatting.
We develop a simple custom prompt for each domain, using the zero-shot pass@k performance as our metric.
We end up with a prompt similar to \citet{yu2025dapo}, shown in \autoref{fig:rlzero-math-prompt}.
We furthermore ``clean'' all our evaluation prompts to remove special formatting (i.e., \texttt{\textbackslash boxed\{\}}) to make evaluation prompts more similar to our training prompts.

\paragraph{RL algorithm} We follow Section~\S\ref{olmo3think-rl} in all RL details except (i) we train with a response length of 16K tokens to better accommodate long chain-of-thought reasoning in the math and code domains and (ii) we evaluate with a response length of 32K tokens and temperature 1.0 to encourage diversity as we report pass@k.
See \autoref{tab:rlvr_training_settings} for hyperparameter details.

\subsection{Key Findings}

\begin{figure}[t]
    \centering
    \begin{subfigure}{0.32\linewidth}
    \centering
    \includegraphics[width=\linewidth]{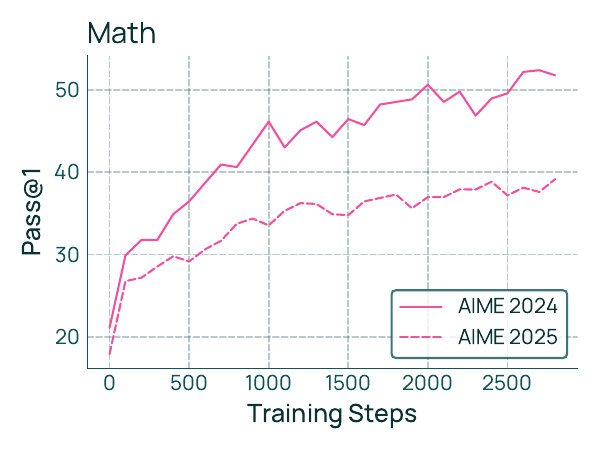}
    \end{subfigure}
    \begin{subfigure}{0.32\linewidth}
    \centering
    \includegraphics[width=\linewidth]{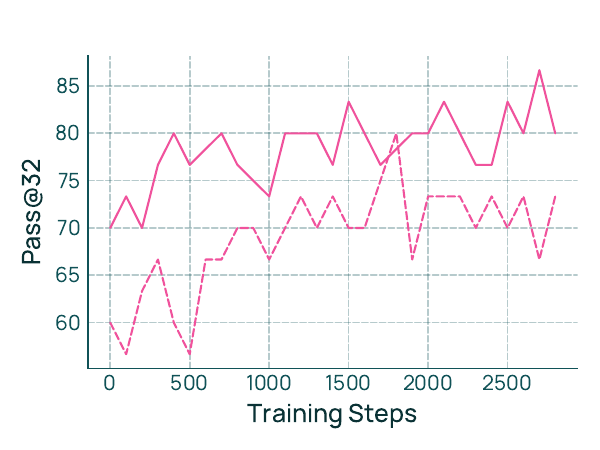}
    \end{subfigure}
    \begin{subfigure}{0.32\linewidth}
    \centering
    \includegraphics[width=\linewidth]{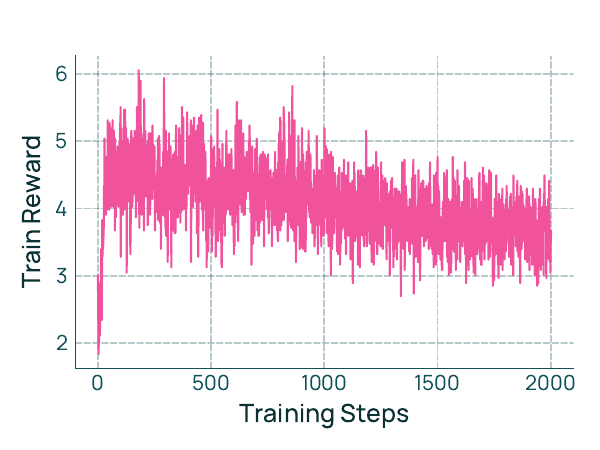}
    \end{subfigure}
    \begin{subfigure}{0.32\linewidth}
    \centering
    \includegraphics[width=\linewidth]{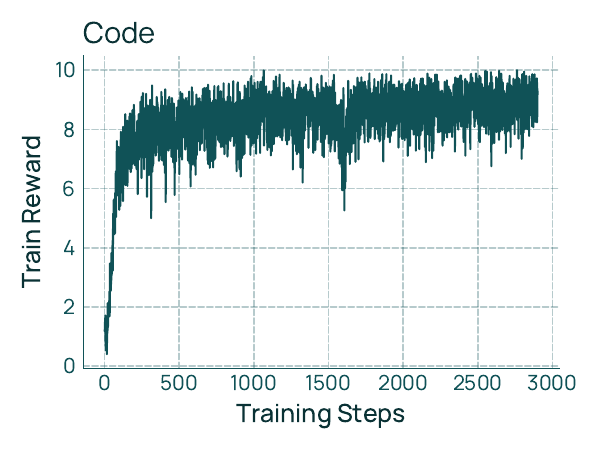}
    \end{subfigure}
    \begin{subfigure}{0.32\linewidth}
    \centering
    \includegraphics[width=\linewidth]{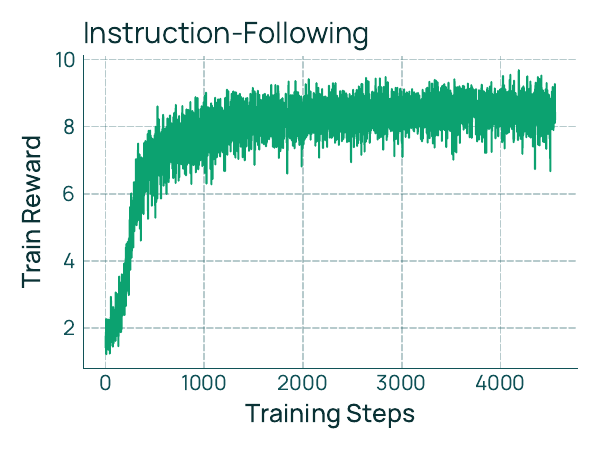}
    \end{subfigure}
    \begin{subfigure}{0.32\linewidth}
    \centering
    \includegraphics[width=\linewidth]{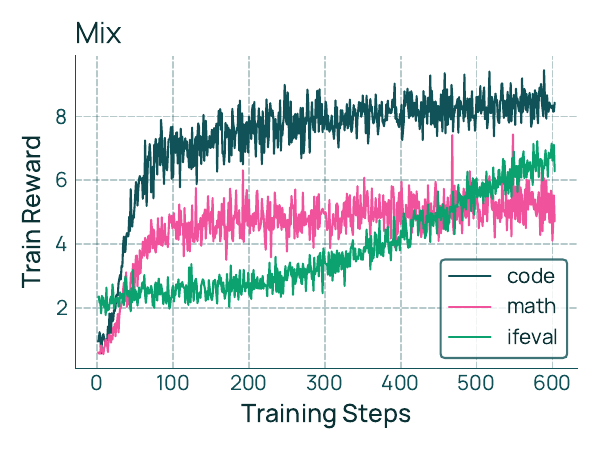}
    \end{subfigure}
    \caption{\textbf{Different domain runs of RL-Zero on \olmothreebase}: math, precise instruction-following, code, and a mix of all three plus general chat.
    We show the main evaluation for the math domain: AIME 2024 and 2025 with pass@1, computed as a bootstrapped average over 32 samples, and pass@32.
    For all domains, we show reward over training.
    For Mix, we separate out the individual rewards for each domain.}
    \label{fig:rlzero-eval}
\end{figure}

\paragraph{\olmothreerlzero can strongly improve on reasoning} As shown in Figure~\ref{fig:rlzero-eval}, our base model can greatly improve on training reward across the different domains when leveraging RL on our datasets. To demonstrate out-of-domain improvements, we evaluate our math run on the decontaminated evals AIME 2024 and 2025. We find that \olmothreebase drastically improves in the first couple hundred steps of training and then improves steadily but slowly.
We also see a decent improvement in pass@32 results, demonstrating that our run maintains diversity and RLVR pushes the model beyond its initial capabilities.
Our initial scores and final scores with the 7B model are, notably, close to DAPO \citep{yu2025dapo} which leverages the larger Qwen 2.5 32B and trains for an order of magnitude more steps, see Figure~\ref{fig:rlzero-vs-dapo} in Appendix~\ref{ssub:rl-zero}.
This demonstrates how \olmothreerlzero can be a more efficient alternative to existing RLVR experiments.

\paragraph{\olmothreerlzero mix can benchmark challenges in multi-objective RL}
Most studies have focused exclusively on RLHF \citep{stiennon2020learning} or single-domain RLVR \citep{yu2025dapo,deepscaler2025}.
Our mix of math, code, instruction-following, and general chat is a more challenging RLVR benchmark for models.
\autoref{fig:rlzero-eval} demonstrates that our general run has improved performance across different domains, but each domain is under-optimized compared to the single-domain setup. Future work can leverage this setup to investigate the interactions between domains in multi-objective RLVR.

\begin{figure}[t]
    \centering
    \begin{subfigure}{0.48\linewidth}
    \centering
    \includegraphics[width=\linewidth]{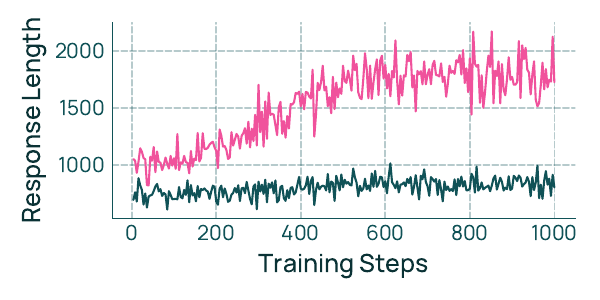}
    \end{subfigure}
    \begin{subfigure}{0.48\linewidth}
    \centering
    \includegraphics[width=\linewidth]{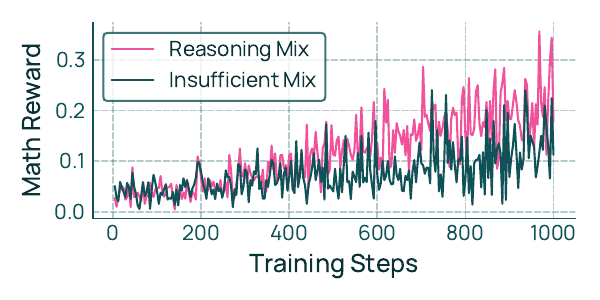}
    \end{subfigure}
    \caption{
    \textbf{The response length and math reward over RL training} for two early midtrained base models. This demonstrates how base model midtraining can determine whether RL-Zero learns longer, more complex reasoning and increases response length.
    }
    \label{fig:rlzero-midtrain}
\end{figure}

\paragraph{\olmothreerlzero can benchmark reasoning data mixes in midtraining} Midtraining and \olmothreerlzero offer a chance to ablate specific data sources, unlike the large-scale effort behind \olmothreethinking. We leverage RL-Zero to evaluate midtraining data mixes for their ability to develop downstream reasoning with RL. For example, we compare two early models in Figure~\ref{fig:rlzero-midtrain}. As evidenced by the stagnant response length, the model with insufficient reasoning data does not leverage backtracking, answer verification, and other cognitive skills \citep{gandhi2025cognitive}. \olmothreerlzero can therefore serve as a testbed for downstream performance of alternative midtraining approaches and improvements over \dolminostoo.

\begin{figure}[t]
    \centering
    \begin{subfigure}{0.48\linewidth}
    \centering
    \includegraphics[width=\linewidth]{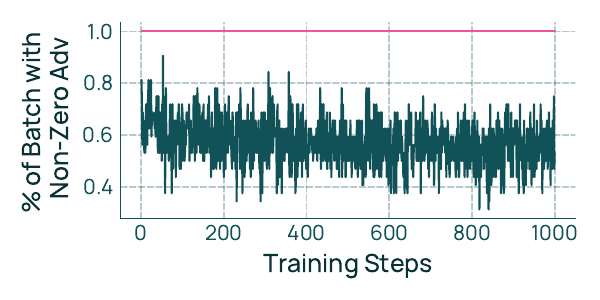}
    \end{subfigure}
    \begin{subfigure}{0.48\linewidth}
    \centering
    \includegraphics[width=\linewidth]{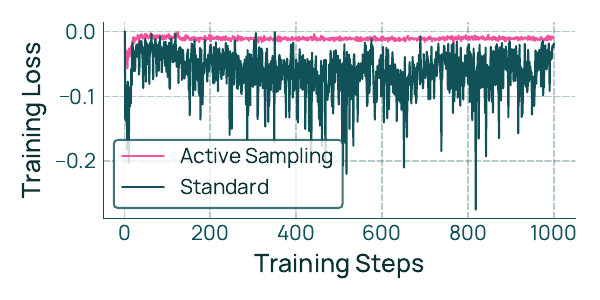}
    \end{subfigure}
\caption{\textbf{Active sampling maintains a full batch of non-zero-advantage samples} by continuously pulling prompt–completion pairs from the result queue after filtering. We plot the percentage of the batch with non-zero advantage as well as the train loss for an RL-Zero Math run with and without active sampling.}
    \label{fig:rlzero-active-sampling}
\end{figure}

\paragraph{Active sampling stabilizes training}
\olmothreerlzero also offers a simpler testbed for ablating RL algorithm and infrastructure decisions.
We ablate active sampling, our novel method for continuously resampling prompts after filtering for non-zero advantage (see Section~\S\ref{sec:posttraining_infra} for details).
Running on our math domain, Figure~\ref{fig:rlzero-active-sampling} shows that active sampling does indeed maintain a consistently full batch of completions with non-zero advantage.
These consistent batch sizes have a stabilizing effect on training, and we see greatly reduced loss variance.

\paragraph{Eval decontamination is verified via spurious rewards}
Recent RLVR benchmarks have shown substantial improvements from training with spurious rewards that are not correlated with model utility. This can suggest that the RLVR task may have been \textit{contaminated}, i.e., the model was exposed to evaluation data during pretraining or midtraining. RLVR with a spurious reward can elicit this memorized knowledge, differentiating it from genuine learning of reasoning capabilities \citep{shao2025spuriousrewardsrethinkingtraining}.
To verify that \olmothreerlzero evaluation is not contaminated, we conduct a negative control experiment by training \olmothreebase with spurious rewards.
Specifically, we train on \dolcirlzero, but instead of rewarding correct answers, we assign random binary rewards to model generations independent of response quality following the protocol in \citet{shao2025spuriousrewardsrethinkingtraining}.
If our pretraining or midtraining data contained significant overlap with our evaluation sets, we would expect spurious reward training to elicit these memorized solutions and improve benchmark performance.

As shown in Figure~\ref{fig:spurious-rewards}, training with random rewards does not improve performance on any of our benchmark evaluations. Performance either remains flat with random fluctuations or degrades, which is consistent with the model learning arbitrary patterns unrelated to the task. This negative result is evidence that our data decontamination successfully removed overlaps between our base-model pipeline and RLVR evaluation data.

\begin{figure}[t]
    \centering
    \includegraphics[width=\linewidth]{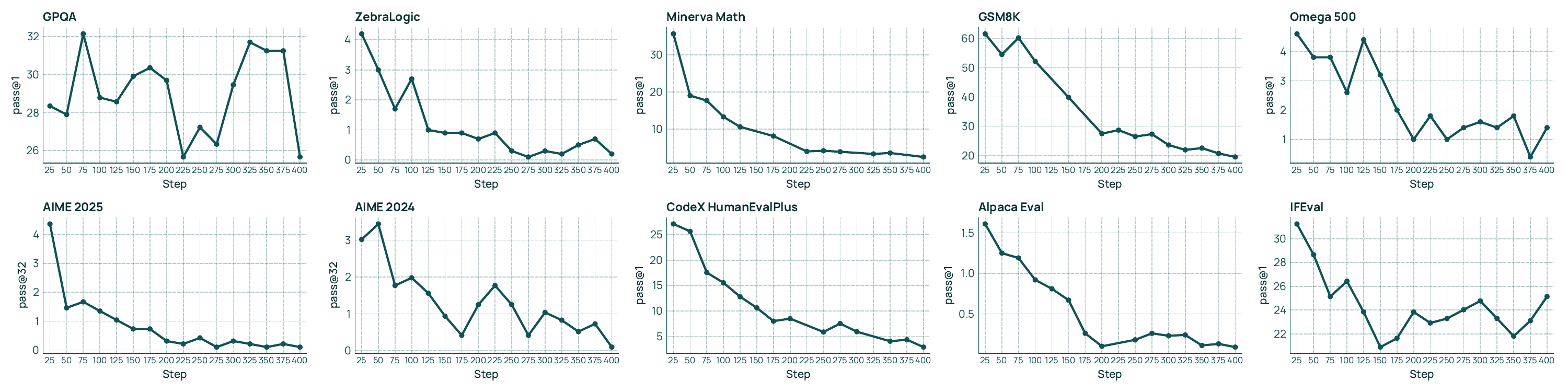}
    \caption{\textbf{RL training on \olmothreebase on random, signal-free rewards produces no performance gains}, suggesting successful decontamination of training data.
    }
    \label{fig:spurious-rewards}
\end{figure}

\begin{textblock*}{100mm}(\dimexpr 25mm,\dimexpr\paperheight - 26mm)
    \nointerlineskip
    \includegraphics[width=4.4124em]{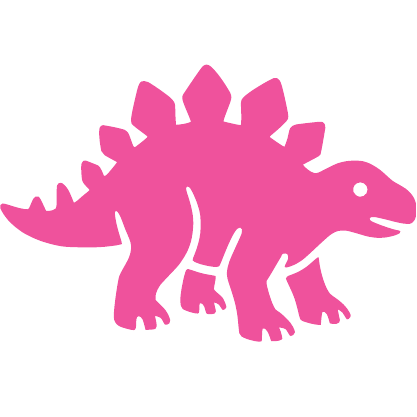}
\end{textblock*}

\begin{textblock*}{100mm}(\dimexpr\paperwidth-100mm+6cm,\dimexpr\paperheight - 26mm)
    \nointerlineskip
    \includegraphics[width=4.4124em]{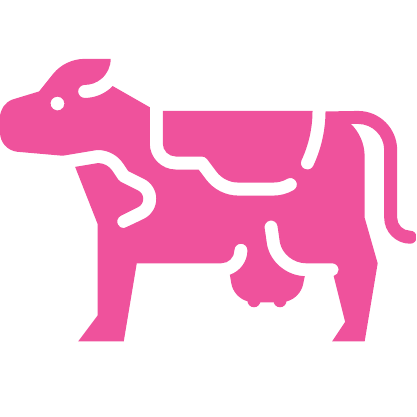}
\end{textblock*}

\clearpage
\bibliographystyle{abbrvnat}
\bibliography{references,safety-refs}

\providecommand{\CNFX}[1]{{\em{\textrm{(#1)}}}}
\begin{thebibliography}{257}
\providecommand{\natexlab}[1]{#1}
\providecommand{\url}[1]{\texttt{#1}}
\expandafter\ifx\csname urlstyle\endcsname\relax
  \providecommand{\doi}[1]{doi: #1}\else
  \providecommand{\doi}{doi: \begingroup \urlstyle{rm}\Url}\fi

\bibitem[Abdin et~al.(2024)Abdin, Aneja, Behl, Bubeck, Eldan, Gunasekar, Harrison, Hewett, Javaheripi, Kauffmann, Lee, Lee, Li, Liu, Mendes, Nguyen, Price, de~Rosa, Saarikivi, Salim, Shah, Wang, Ward, Wu, Yu, Zhang, and Zhang]{Abdin2024Phi4TR}
M.~Abdin, J.~Aneja, H.~S. Behl, S.~Bubeck, R.~Eldan, S.~Gunasekar, M.~Harrison, R.~J. Hewett, M.~Javaheripi, P.~Kauffmann, J.~R. Lee, Y.~T. Lee, Y.~Li, W.~Liu, C.~C.~T. Mendes, A.~Nguyen, E.~Price, G.~de~Rosa, O.~Saarikivi, A.~Salim, S.~Shah, X.~Wang, R.~Ward, Y.~Wu, D.~Yu, C.~Zhang, and Y.~Zhang.
\newblock Phi-4 technical report.
\newblock \emph{arXiv preprint arXiv:2412.08905}, 2024.

\bibitem[Ackerman and Thompson(2017)]{ACKERMAN2017607}
R.~Ackerman and V.~A. Thompson.
\newblock Meta-reasoning: Monitoring and control of thinking and reasoning.
\newblock \emph{Trends in Cognitive Sciences}, 21\penalty0 (8):\penalty0 607--617, 2017.
\newblock ISSN 1364-6613.
\newblock \doi{https://doi.org/10.1016/j.tics.2017.05.004}.
\newblock URL \url{https://www.sciencedirect.com/science/article/pii/S1364661317301055}.

\bibitem[Adler et~al.(2024)Adler, Agarwal, Aithal, Anh, Bhattacharya, Brundyn, Casper, Catanzaro, Clay, Cohen, et~al.]{adler2024nemotron}
B.~Adler, N.~Agarwal, A.~Aithal, D.~H. Anh, P.~Bhattacharya, A.~Brundyn, J.~Casper, B.~Catanzaro, S.~Clay, J.~Cohen, et~al.
\newblock Nemotron-4 340b technical report.
\newblock \emph{arXiv preprint arXiv:2406.11704}, 2024.

\bibitem[Agarwal et~al.(2025)Agarwal, Ahmad, Ai, Altman, Applebaum, Arbus, Arora, Bai, Baker, Bao, et~al.]{agarwal2025gpt}
S.~Agarwal, L.~Ahmad, J.~Ai, S.~Altman, A.~Applebaum, E.~Arbus, R.~K. Arora, Y.~Bai, B.~Baker, H.~Bao, et~al.
\newblock gpt-oss-120b \& gpt-oss-20b model card.
\newblock \emph{arXiv preprint arXiv:2508.10925}, 2025.

\bibitem[Aggarwal and Welleck(2025)]{aggarwal2025l1controllinglongreasoning}
P.~Aggarwal and S.~Welleck.
\newblock L1: Controlling how long a reasoning model thinks with reinforcement learning, 2025.
\newblock URL \url{https://arxiv.org/abs/2503.04697}.

\bibitem[Ahmad et~al.(2025)Ahmad, Narenthiran, Majumdar, Ficek, Jain, Huang, Noroozi, and Ginsburg]{ahmad2025opencodereasoning}
W.~U. Ahmad, S.~Narenthiran, S.~Majumdar, A.~Ficek, S.~Jain, J.~Huang, V.~Noroozi, and B.~Ginsburg.
\newblock Opencodereasoning: Advancing data distillation for competitive coding.
\newblock \emph{arXiv preprint arXiv:2504.01943}, 2025.
\newblock URL \url{https://arxiv.org/abs/2504.01943}.

\bibitem[Ainslie et~al.(2023)Ainslie, Lee-Thorp, de~Jong, Zemlyanskiy, Lebrón, and Sanghai]{ainslie2023gqatraininggeneralizedmultiquery}
J.~Ainslie, J.~Lee-Thorp, M.~de~Jong, Y.~Zemlyanskiy, F.~Lebrón, and S.~Sanghai.
\newblock Gqa: Training generalized multi-query transformer models from multi-head checkpoints, 2023.
\newblock URL \url{https://arxiv.org/abs/2305.13245}.

\bibitem[Akter et~al.(2024)Akter, Prabhumoye, Kamalu, Satheesh, Nyberg, Patwary, Shoeybi, and Catanzaro]{akter2024mindmathinformedsynthetic}
S.~N. Akter, S.~Prabhumoye, J.~Kamalu, S.~Satheesh, E.~Nyberg, M.~Patwary, M.~Shoeybi, and B.~Catanzaro.
\newblock Mind: Math informed synthetic dialogues for pretraining llms, 2024.
\newblock URL \url{https://arxiv.org/abs/2410.12881}.

\bibitem[Allal et~al.(2025)Allal, Lozhkov, Bakouch, Blázquez, Penedo, Tunstall, Marafioti, Kydlíček, Lajarín, Srivastav, Lochner, Fahlgren, Nguyen, Fourrier, Burtenshaw, Larcher, Zhao, Zakka, Morlon, Raffel, von Werra, and Wolf]{allal2025smollm2smolgoesbig}
L.~B. Allal, A.~Lozhkov, E.~Bakouch, G.~M. Blázquez, G.~Penedo, L.~Tunstall, A.~Marafioti, H.~Kydlíček, A.~P. Lajarín, V.~Srivastav, J.~Lochner, C.~Fahlgren, X.-S. Nguyen, C.~Fourrier, B.~Burtenshaw, H.~Larcher, H.~Zhao, C.~Zakka, M.~Morlon, C.~Raffel, L.~von Werra, and T.~Wolf.
\newblock Smollm2: When smol goes big -- data-centric training of a small language model, 2025.
\newblock URL \url{https://arxiv.org/abs/2502.02737}.

\bibitem[An et~al.(2025)An, Xie, Li, Li, Zhang, Gong, Zhong, Xu, Qiu, Wang, and Kong]{Polaris2025}
C.~An, Z.~Xie, X.~Li, L.~Li, J.~Zhang, S.~Gong, M.~Zhong, J.~Xu, X.~Qiu, M.~Wang, and L.~Kong.
\newblock Polaris: A post-training recipe for scaling reinforcement learning on advanced reasoning models, 2025.
\newblock URL \url{https://hkunlp.github.io/blog/2025/Polaris}.

\bibitem[Anthropic(2025)]{anthropic-claude4-systemcard}
Anthropic.
\newblock System card: Claude opus 4 \& claude sonnet 4.
\newblock Technical report, Anthropic, 2025.
\newblock Accessed: 2025-10-07.

\bibitem[{Apertus Team}(2025)]{swissai2025apertus}
{Apertus Team}.
\newblock {Apertus: Democratizing Open and Compliant LLMs for Global Language Environments}.
\newblock \url{https://huggingface.co/swiss-ai/Apertus-70B-2509}, 2025.

\bibitem[Asai et~al.(2024)Asai, He, Shao, Shi, Singh, Chang, Lo, Soldaini, Feldman, D'Arcy, Wadden, Latzke, Tian, Ji, Liu, Tong, Wu, Xiong, Zettlemoyer, Neubig, Weld, Downey, tau Yih, Koh, and Hajishirzi]{Asai2024OpenScholarSS}
A.~Asai, J.~He, R.~Shao, W.~Shi, A.~Singh, J.~C. Chang, K.~Lo, L.~Soldaini, S.~Feldman, M.~D'Arcy, D.~Wadden, M.~Latzke, M.~Tian, P.~Ji, S.~Liu, H.~Tong, B.~Wu, Y.~Xiong, L.~S. Zettlemoyer, G.~Neubig, D.~Weld, D.~Downey, W.~tau Yih, P.~W. Koh, and H.~Hajishirzi.
\newblock Openscholar: Synthesizing scientific literature with retrieval-augmented lms.
\newblock \emph{ArXiv}, abs/2411.14199, 2024.
\newblock URL \url{https://api.semanticscholar.org/CorpusID:274166189}.

\bibitem[Attardi(2015)]{Wikiextractor2015}
G.~Attardi.
\newblock Wikiextractor.
\newblock \url{https://github.com/attardi/wikiextractor}, 2015.

\bibitem[Austin et~al.(2021)Austin, Odena, Nye, Bosma, Michalewski, Dohan, Jiang, Cai, Terry, Le, et~al.]{austin2021program}
J.~Austin, A.~Odena, M.~Nye, M.~Bosma, H.~Michalewski, D.~Dohan, E.~Jiang, C.~Cai, M.~Terry, Q.~Le, et~al.
\newblock Program synthesis with large language models.
\newblock \emph{arXiv preprint arXiv:2108.07732}, 2021.

\bibitem[Azar et~al.(2023)Azar, Rowland, Piot, Guo, Calandriello, Valko, and Munos]{azar2023generaltheoreticalparadigmunderstand}
M.~G. Azar, M.~Rowland, B.~Piot, D.~Guo, D.~Calandriello, M.~Valko, and R.~Munos.
\newblock A general theoretical paradigm to understand learning from human preferences, 2023.
\newblock URL \url{https://arxiv.org/abs/2310.12036}.

\bibitem[Azerbayev et~al.(2023)Azerbayev, Schoelkopf, Paster, Santos, McAleer, Jiang, Deng, Biderman, and Welleck]{azerbayev2023llemma}
Z.~Azerbayev, H.~Schoelkopf, K.~Paster, M.~D. Santos, S.~McAleer, A.~Q. Jiang, J.~Deng, S.~Biderman, and S.~Welleck.
\newblock Llemma: An open language model for mathematics, 2023.

\bibitem[Bai et~al.(2024)Bai, Lv, Zhang, Lyu, Tang, Huang, Du, Liu, Zeng, Hou, Dong, Tang, and Li]{longbench1}
Y.~Bai, X.~Lv, J.~Zhang, H.~Lyu, J.~Tang, Z.~Huang, Z.~Du, X.~Liu, A.~Zeng, L.~Hou, Y.~Dong, J.~Tang, and J.~Li.
\newblock {L}ong{B}ench: A bilingual, multitask benchmark for long context understanding.
\newblock In L.-W. Ku, A.~Martins, and V.~Srikumar, editors, \emph{Proceedings of the 62nd Annual Meeting of the Association for Computational Linguistics (Volume 1: Long Papers)}, pages 3119--3137, Bangkok, Thailand, Aug. 2024. Association for Computational Linguistics.
\newblock \doi{10.18653/v1/2024.acl-long.172}.
\newblock URL \url{https://aclanthology.org/2024.acl-long.172/}.

\bibitem[Bai et~al.(2025)Bai, Tu, Zhang, Peng, Wang, Lv, Cao, Xu, Hou, Dong, Tang, and Li]{longbench2}
Y.~Bai, S.~Tu, J.~Zhang, H.~Peng, X.~Wang, X.~Lv, S.~Cao, J.~Xu, L.~Hou, Y.~Dong, J.~Tang, and J.~Li.
\newblock {L}ong{B}ench v2: Towards deeper understanding and reasoning on realistic long-context multitasks.
\newblock In W.~Che, J.~Nabende, E.~Shutova, and M.~T. Pilehvar, editors, \emph{Proceedings of the 63rd Annual Meeting of the Association for Computational Linguistics (Volume 1: Long Papers)}, pages 3639--3664, Vienna, Austria, July 2025. Association for Computational Linguistics.
\newblock ISBN 979-8-89176-251-0.
\newblock \doi{10.18653/v1/2025.acl-long.183}.
\newblock URL \url{https://aclanthology.org/2025.acl-long.183/}.

\bibitem[Bakouch et~al.(2025)Bakouch, Ben~Allal, Lozhkov, Tazi, Tunstall, Patiño, Beeching, Roucher, Reedi, Gallouédec, Rasul, Habib, Fourrier, Kydlicek, Penedo, Larcher, Morlon, Srivastav, Lochner, Nguyen, Raffel, von Werra, and Wolf]{bakouch2025smollm3}
E.~Bakouch, L.~Ben~Allal, A.~Lozhkov, N.~Tazi, L.~Tunstall, C.~M. Patiño, E.~Beeching, A.~Roucher, A.~J. Reedi, Q.~Gallouédec, K.~Rasul, N.~Habib, C.~Fourrier, H.~Kydlicek, G.~Penedo, H.~Larcher, M.~Morlon, V.~Srivastav, J.~Lochner, X.-S. Nguyen, C.~Raffel, L.~von Werra, and T.~Wolf.
\newblock {SmolLM3: smol, multilingual, long-context reasoner}.
\newblock \url{https://huggingface.co/blog/smollm3}, 2025.

\bibitem[Bavarian et~al.(2022)Bavarian, Jun, Tezak, Schulman, McLeavey, Tworek, and Chen]{bavarian2022efficient}
M.~Bavarian, H.~Jun, N.~Tezak, J.~Schulman, C.~McLeavey, J.~Tworek, and M.~Chen.
\newblock Efficient training of language models to fill in the middle.
\newblock \emph{arXiv preprint arXiv:2207.14255}, 2022.

\bibitem[Beltagy et~al.(2020)Beltagy, Peters, and Cohan]{beltagy2020longformer}
I.~Beltagy, M.~E. Peters, and A.~Cohan.
\newblock Longformer: The long-document transformer.
\newblock \emph{arXiv preprint arXiv:2004.05150}, 2020.

\bibitem[Bercovich et~al.(2025)Bercovich, Levy, Golan, Dabbah, El-Yaniv, Puny, Galil, Moshe, Ronen, Nabwani, Shahaf, Tropp, Karpas, Zilberstein, Zeng, Singhal, Bukharin, Zhang, Konuk, Shen, Mahabaleshwarkar, Kartal, Suhara, Delalleau, Chen, Wang, Mosallanezhad, Renduchintala, Qian, Rekesh, Jia, Majumdar, Noroozi, Ahmad, Narenthiran, Ficek, Samadi, Huang, Jain, Gitman, Moshkov, Du, Toshniwal, Armstrong, Kisacanin, Novikov, Gitman, Bakhturina, Scowcroft, Kamalu, Su, Kong, Kliegl, Karimi, Lin, Satheesh, Parmar, Gundecha, Norick, Jennings, Prabhumoye, Akter, Patwary, Khattar, Narayanan, Waleffe, Zhang, Su, Huang, Kong, Chadha, Jain, Harvey, Segal, Huang, Kashirsky, McQueen, Putterman, Lam, Venkatesan, Wu, Nguyen, Kilaru, Wang, Warno, Somasamudramath, Bhaskar, Dong, Assaf, Mor, Argov, Junkin, Romanenko, Larroy, Katariya, Rovinelli, Balas, Edelman, Bhiwandiwalla, Subramaniam, Ithape, Ramamoorthy, Wu, Velury, Almog, Daw, Fridman, Galinkin, Evans, Luna, Derczynski, Pope, Long, Schneider, Siman, Grzegorzek, Ribalta,
  Katariya, Conway, Saar, Guan, Pawelec, Prayaga, Kuchaiev, Ginsburg, Olabiyi, Briski, Cohen, Catanzaro, Alben, Geifman, Chung, and Alexiuk]{bercovich2025llamanemotronefficientreasoningmodels}
A.~Bercovich, I.~Levy, I.~Golan, M.~Dabbah, R.~El-Yaniv, O.~Puny, I.~Galil, Z.~Moshe, T.~Ronen, N.~Nabwani, I.~Shahaf, O.~Tropp, E.~Karpas, R.~Zilberstein, J.~Zeng, S.~Singhal, A.~Bukharin, Y.~Zhang, T.~Konuk, G.~Shen, A.~S. Mahabaleshwarkar, B.~Kartal, Y.~Suhara, O.~Delalleau, Z.~Chen, Z.~Wang, D.~Mosallanezhad, A.~Renduchintala, H.~Qian, D.~Rekesh, F.~Jia, S.~Majumdar, V.~Noroozi, W.~U. Ahmad, S.~Narenthiran, A.~Ficek, M.~Samadi, J.~Huang, S.~Jain, I.~Gitman, I.~Moshkov, W.~Du, S.~Toshniwal, G.~Armstrong, B.~Kisacanin, M.~Novikov, D.~Gitman, E.~Bakhturina, J.~P. Scowcroft, J.~Kamalu, D.~Su, K.~Kong, M.~Kliegl, R.~Karimi, Y.~Lin, S.~Satheesh, J.~Parmar, P.~Gundecha, B.~Norick, J.~Jennings, S.~Prabhumoye, S.~N. Akter, M.~Patwary, A.~Khattar, D.~Narayanan, R.~Waleffe, J.~Zhang, B.-Y. Su, G.~Huang, T.~Kong, P.~Chadha, S.~Jain, C.~Harvey, E.~Segal, J.~Huang, S.~Kashirsky, R.~McQueen, I.~Putterman, G.~Lam, A.~Venkatesan, S.~Wu, V.~Nguyen, M.~Kilaru, A.~Wang, A.~Warno, A.~Somasamudramath, S.~Bhaskar, M.~Dong,
  N.~Assaf, S.~Mor, O.~U. Argov, S.~Junkin, O.~Romanenko, P.~Larroy, M.~Katariya, M.~Rovinelli, V.~Balas, N.~Edelman, A.~Bhiwandiwalla, M.~Subramaniam, S.~Ithape, K.~Ramamoorthy, Y.~Wu, S.~V. Velury, O.~Almog, J.~Daw, D.~Fridman, E.~Galinkin, M.~Evans, K.~Luna, L.~Derczynski, N.~Pope, E.~Long, S.~Schneider, G.~Siman, T.~Grzegorzek, P.~Ribalta, M.~Katariya, J.~Conway, T.~Saar, A.~Guan, K.~Pawelec, S.~Prayaga, O.~Kuchaiev, B.~Ginsburg, O.~Olabiyi, K.~Briski, J.~Cohen, B.~Catanzaro, J.~Alben, Y.~Geifman, E.~Chung, and C.~Alexiuk.
\newblock Llama-nemotron: Efficient reasoning models, 2025.
\newblock URL \url{https://arxiv.org/abs/2505.00949}.

\bibitem[Bertsch et~al.(2026)Bertsch, Soldaini, Gormley, Neubig, Hajishirzi, Lo, and Groeneveld]{bertsch2026cracks}
A.~Bertsch, L.~Soldaini, M.~Gormley, G.~Neubig, H.~Hajishirzi, K.~Lo, and D.~Groeneveld.
\newblock Cracks in the foundation: Architectural choices impact long context extension, 2026.

\bibitem[Bevendorff et~al.(2018)Bevendorff, Stein, Hagen, and Potthast]{bevendorff2018}
J.~Bevendorff, B.~Stein, M.~Hagen, and M.~Potthast.
\newblock {Elastic ChatNoir: Search Engine for the ClueWeb and the Common Crawl}.
\newblock In L.~Azzopardi, A.~Hanbury, G.~Pasi, and B.~Piwowarski, editors, \emph{Advances in Information Retrieval. 40th European Conference on IR Research (ECIR 2018)}, Lecture Notes in Computer Science, Berlin Heidelberg New York, Mar. 2018. Springer.

\bibitem[Bhagia et~al.(2024)Bhagia, Liu, Wettig, Heineman, Tafjord, Jha, Soldaini, Smith, Groeneveld, Koh, Dodge, and Hajishirzi]{bhagia2024establishingtaskscalinglaws}
A.~Bhagia, J.~Liu, A.~Wettig, D.~Heineman, O.~Tafjord, A.~H. Jha, L.~Soldaini, N.~A. Smith, D.~Groeneveld, P.~W. Koh, J.~Dodge, and H.~Hajishirzi.
\newblock Establishing task scaling laws via compute-efficient model ladders, 2024.
\newblock URL \url{https://arxiv.org/abs/2412.04403}.

\bibitem[Bisk et~al.(2020)Bisk, Zellers, Le~bras, Gao, and Choi]{Bisk_Zellers_Le_bras_Gao_Choi_2020}
Y.~Bisk, R.~Zellers, R.~Le~bras, J.~Gao, and Y.~Choi.
\newblock {PIQA}: Reasoning about physical commonsense in natural language.
\newblock \emph{Proceedings of the AAAI Conference on Artificial Intelligence}, 34\penalty0 (05):\penalty0 7432--7439, Apr. 2020.
\newblock \doi{10.1609/aaai.v34i05.6239}.
\newblock URL \url{https://ojs.aaai.org/index.php/AAAI/article/view/6239}.

\bibitem[Bordt et~al.(2024)Bordt, Srinivas, Boreiko, and von Luxburg]{Bordt2024HowMC}
S.~Bordt, S.~Srinivas, V.~Boreiko, and U.~von Luxburg.
\newblock How much can we forget about data contamination?
\newblock \emph{ArXiv}, abs/2410.03249, 2024.
\newblock URL \url{https://api.semanticscholar.org/CorpusID:273163321}.

\bibitem[Bragg et~al.(2025)Bragg, D'Arcy, Balepur, Bareket, Dalvi, Feldman, Haddad, Hwang, Jansen, Kishore, et~al.]{bragg2025astabench}
J.~Bragg, M.~D'Arcy, N.~Balepur, D.~Bareket, B.~Dalvi, S.~Feldman, D.~Haddad, J.~D. Hwang, P.~Jansen, V.~Kishore, et~al.
\newblock Astabench: Rigorous benchmarking of ai agents with a scientific research suite.
\newblock \emph{arXiv preprint arXiv:2510.21652}, 2025.

\bibitem[Brahman et~al.(2024)Brahman, Kumar, Balachandran, Dasigi, Pyatkin, Ravichander, Wiegreffe, Dziri, Chandu, Hessel, et~al.]{brahman2024art}
F.~Brahman, S.~Kumar, V.~Balachandran, P.~Dasigi, V.~Pyatkin, A.~Ravichander, S.~Wiegreffe, N.~Dziri, K.~Chandu, J.~Hessel, et~al.
\newblock The art of saying no: Contextual noncompliance in language models.
\newblock \emph{arXiv preprint arXiv:2407.12043}, 2024.

\bibitem[Cai et~al.(2025)Cai, Shabihi, An, Che, Bartoldson, Kailkhura, Goldstein, and Huang]{cai2025aegisllm}
Z.~Cai, S.~Shabihi, B.~An, Z.~Che, B.~R. Bartoldson, B.~Kailkhura, T.~Goldstein, and F.~Huang.
\newblock Aegisllm: Scaling agentic systems for self-reflective defense in llm security.
\newblock \emph{arXiv preprint arXiv:2504.20965}, 2025.
\newblock Preprint.

\bibitem[Callaway et~al.(2022)Callaway, \{van Opheusden\}, Gul, Das, Krueger, Griffiths, and Lieder]{bfa322bf36e54a4ca19f9a73bee6184b}
F.~Callaway, B.~\{van Opheusden\}, S.~Gul, P.~Das, P.~Krueger, T.~Griffiths, and F.~Lieder.
\newblock Rational use of cognitive resources in human planning.
\newblock \emph{Nature Human Behaviour}, 6\penalty0 (8):\penalty0 1112--1125, Aug. 2022.
\newblock ISSN 2397-3374.
\newblock \doi{10.1038/s41562-022-01332-8}.
\newblock Publisher Copyright: {\textcopyright} 2022, The Author(s), under exclusive licence to Springer Nature Limited.

\bibitem[Cassano et~al.(2022)Cassano, Gouwar, Nguyen, Nguyen, Phipps-Costin, Pinckney, Yee, Zi, Anderson, Feldman, et~al.]{cassano2022multipl}
F.~Cassano, J.~Gouwar, D.~Nguyen, S.~Nguyen, L.~Phipps-Costin, D.~Pinckney, M.-H. Yee, Y.~Zi, C.~J. Anderson, M.~Q. Feldman, et~al.
\newblock Multipl-e: A scalable and extensible approach to benchmarking neural code generation.
\newblock \emph{arXiv preprint arXiv:2208.08227}, 2022.

\bibitem[Chatterji et~al.(2025)Chatterji, Cunningham, Deming, Hitzig, Ong, Shan, and Wadman]{Chatterji2025-fs}
A.~Chatterji, T.~Cunningham, D.~Deming, Z.~Hitzig, C.~Ong, C.~Y. Shan, and K.~Wadman.
\newblock How people use {ChatGPT}.
\newblock Technical Report w34255, National Bureau of Economic Research, Cambridge, MA, Sept. 2025.

\bibitem[Chen et~al.(2021)Chen, Tworek, Jun, Yuan, de~Oliveira~Pinto, Kaplan, Edwards, Burda, Joseph, Brockman, Ray, Puri, Krueger, Petrov, Khlaaf, Sastry, Mishkin, Chan, Gray, Ryder, Pavlov, Power, Kaiser, Bavarian, Winter, Tillet, Such, Cummings, Plappert, Chantzis, Barnes, Herbert-Voss, Guss, Nichol, Paino, Tezak, Tang, Babuschkin, Balaji, Jain, Saunders, Hesse, Carr, Leike, Achiam, Misra, Morikawa, Radford, Knight, Brundage, Murati, Mayer, Welinder, McGrew, Amodei, McCandlish, Sutskever, and Zaremba]{chen2021codex}
M.~Chen, J.~Tworek, H.~Jun, Q.~Yuan, H.~P. de~Oliveira~Pinto, J.~Kaplan, H.~Edwards, Y.~Burda, N.~Joseph, G.~Brockman, A.~Ray, R.~Puri, G.~Krueger, M.~Petrov, H.~Khlaaf, G.~Sastry, P.~Mishkin, B.~Chan, S.~Gray, N.~Ryder, M.~Pavlov, A.~Power, L.~Kaiser, M.~Bavarian, C.~Winter, P.~Tillet, F.~P. Such, D.~Cummings, M.~Plappert, F.~Chantzis, E.~Barnes, A.~Herbert-Voss, W.~H. Guss, A.~Nichol, A.~Paino, N.~Tezak, J.~Tang, I.~Babuschkin, S.~Balaji, S.~Jain, W.~Saunders, C.~Hesse, A.~N. Carr, J.~Leike, J.~Achiam, V.~Misra, E.~Morikawa, A.~Radford, M.~Knight, M.~Brundage, M.~Murati, K.~Mayer, P.~Welinder, B.~McGrew, D.~Amodei, S.~McCandlish, I.~Sutskever, and W.~Zaremba.
\newblock Evaluating large language models trained on code.
\newblock 2021.

\bibitem[Chen et~al.(2026)Chen, Murray, Heineman, Jordan, Hajishirzi, R\'e, Soldaini, and Lo]{olmix}
M.~F. Chen, T.~Murray, D.~Heineman, M.~Jordan, H.~Hajishirzi, C.~R\'e, L.~Soldaini, and K.~Lo.
\newblock Olmix: Efficient mixture recomputation for evolving lm datasets, 2026.

\bibitem[Chen et~al.(2023)Chen, Wong, Chen, and Tian]{chen2023extendingcontextwindowlarge}
S.~Chen, S.~Wong, L.~Chen, and Y.~Tian.
\newblock Extending context window of large language models via positional interpolation, 2023.
\newblock URL \url{https://arxiv.org/abs/2306.15595}.

\bibitem[Chen et~al.(2025)Chen, Yang, Liu, Lee, Xu, Shoeybi, Catanzaro, and Ping]{chen2025acereason}
Y.~Chen, Z.~Yang, Z.~Liu, C.~Lee, P.~Xu, M.~Shoeybi, B.~Catanzaro, and W.~Ping.
\newblock Acereason-nemotron: Advancing math and code reasoning through reinforcement learning.
\newblock \emph{arXiv preprint arXiv:2505.16400}, 2025.

\bibitem[Cheng et~al.(2025)Cheng, Hao, Liu, Zhou, Xie, Yao, Bian, Zhuang, Dey, Zha, Gu, Zhou, Wang, Li, Fan, She, Gao, Saparov, Li, Killian, Yurochkin, Liu, Xing, and Hu]{cheng2025revisiting}
Z.~Cheng, S.~Hao, T.~Liu, F.~Zhou, Y.~Xie, F.~Yao, Y.~Bian, Y.~Zhuang, N.~Dey, Y.~Zha, Y.~Gu, K.~Zhou, Y.~Wang, Y.~Li, R.~Fan, J.~She, C.~Gao, A.~Saparov, H.~Li, T.~W. Killian, M.~Yurochkin, Z.~Liu, E.~P. Xing, and Z.~Hu.
\newblock Revisiting reinforcement learning for llm reasoning from a cross-domain perspective, 2025.
\newblock URL \url{https://arxiv.org/abs/2506.14965}.

\bibitem[Chu et~al.(2025)Chu, Xie, Yu, Wang, Phanishayee, Tang, Hao, Huang, Ozdal, Wang, Goswami, Goyal, Kadian, Gu, Cai, Tian, Wang, Si, Balaji, Chu, and Park]{scalingllama3}
W.~Chu, X.~Xie, J.~Yu, J.~Wang, A.~Phanishayee, C.~Tang, Y.~Hao, J.~Huang, M.~Ozdal, J.~Wang, V.~Goswami, N.~Goyal, A.~Kadian, A.~Gu, C.~Cai, F.~Tian, X.~Wang, M.~Si, P.~Balaji, C.-H. Chu, and J.~Park.
\newblock Scaling llama 3 training with efficient parallelism strategies.
\newblock In \emph{Proceedings of the 52nd Annual International Symposium on Computer Architecture}, ISCA '25, page 1703–1716, New York, NY, USA, 2025. Association for Computing Machinery.
\newblock ISBN 9798400712616.
\newblock \doi{10.1145/3695053.3731410}.
\newblock URL \url{https://doi.org/10.1145/3695053.3731410}.

\bibitem[Clark et~al.(2019)Clark, Lee, Chang, Kwiatkowski, Collins, and Toutanova]{clark-etal-2019-boolq}
C.~Clark, K.~Lee, M.-W. Chang, T.~Kwiatkowski, M.~Collins, and K.~Toutanova.
\newblock {B}ool{Q}: Exploring the surprising difficulty of natural yes/no questions.
\newblock In J.~Burstein, C.~Doran, and T.~Solorio, editors, \emph{Proceedings of the 2019 Conference of the North {A}merican Chapter of the Association for Computational Linguistics: Human Language Technologies, Volume 1 (Long and Short Papers)}, pages 2924--2936, Minneapolis, Minnesota, June 2019. Association for Computational Linguistics.
\newblock \doi{10.18653/v1/N19-1300}.
\newblock URL \url{https://aclanthology.org/N19-1300}.

\bibitem[Clark et~al.(2018)Clark, Cowhey, Etzioni, Khot, Sabharwal, Schoenick, and Tafjord]{clark2018think}
P.~Clark, I.~Cowhey, O.~Etzioni, T.~Khot, A.~Sabharwal, C.~Schoenick, and O.~Tafjord.
\newblock Think you have solved question answering? {T}ry {ARC}, the {AI2} reasoning challenge.
\newblock \emph{CoRR}, arXiv:1803.05457, 2018.

\bibitem[Cobbe et~al.(2021)Cobbe, Kosaraju, Bavarian, Chen, Jun, Kaiser, Plappert, Tworek, Hilton, Nakano, Hesse, and Schulman]{cobbe2021gsm8k}
K.~Cobbe, V.~Kosaraju, M.~Bavarian, M.~Chen, H.~Jun, L.~Kaiser, M.~Plappert, J.~Tworek, J.~Hilton, R.~Nakano, C.~Hesse, and J.~Schulman.
\newblock Training verifiers to solve math word problems.
\newblock \emph{arXiv preprint arXiv:2110.14168}, 2021.

\bibitem[{Common Crawl Foundation}()]{CommonCrawl}
{Common Crawl Foundation}.
\newblock {Common Crawl Dataset}.
\newblock \url{https://commoncrawl.org/}.
\newblock Accessed: December 31, 2024.

\bibitem[Cui et~al.(2023)Cui, Yuan, Ding, Yao, Zhu, Ni, Xie, Liu, and Sun]{cui2023ultrafeedback}
G.~Cui, L.~Yuan, N.~Ding, G.~Yao, W.~Zhu, Y.~Ni, G.~Xie, Z.~Liu, and M.~Sun.
\newblock {UltraFeedback}: Boosting language models with scaled ai feedback.
\newblock \emph{arXiv preprint arXiv:2310.01377}, 2023.

\bibitem[Dao(2024)]{dao2023flashattention2}
T.~Dao.
\newblock Flash{A}ttention-2: Faster attention with better parallelism and work partitioning.
\newblock In \emph{International Conference on Learning Representations (ICLR)}, 2024.

\bibitem[Dasigi et~al.(2021)Dasigi, Lo, Beltagy, Cohan, Smith, and Gardner]{dasigi2021dataset}
P.~Dasigi, K.~Lo, I.~Beltagy, A.~Cohan, N.~A. Smith, and M.~Gardner.
\newblock A dataset of information-seeking questions and answers anchored in research papers.
\newblock \emph{arXiv preprint arXiv:2105.03011}, 2021.

\bibitem[{DeepSeek-AI}(2025)]{deepseekV31}
{DeepSeek-AI}.
\newblock {DeepSeek}-{V3.1} release.
\newblock \url{https://api-docs.deepseek.com/news/news250821}, 2025.
\newblock Accessed: 2025-11-10.

\bibitem[DeepSeek-AI et~al.(2025)DeepSeek-AI, Liu, Feng, Xue, Wang, Wu, Lu, Zhao, Deng, Zhang, Ruan, Dai, Guo, Yang, Chen, Ji, Li, Lin, Dai, Luo, Hao, Chen, Li, Zhang, Bao, Xu, Wang, Zhang, Ding, Xin, Gao, Li, Qu, Cai, Liang, Guo, Ni, Li, Wang, Chen, Chen, Yuan, Qiu, Li, Song, Dong, Hu, Gao, Guan, Huang, Yu, Wang, Zhang, Xu, Xia, Zhao, Wang, Zhang, Li, Wang, Zhang, Zhang, Tang, Li, Tian, Huang, Wang, Zhang, Wang, Zhu, Chen, Du, Chen, Jin, Ge, Zhang, Pan, Wang, Xu, Zhang, Chen, Li, Lu, Zhou, Chen, Wu, Ye, Ye, Ma, Wang, Zhou, Yu, Zhou, Pan, Wang, Yun, Pei, Sun, Xiao, Zeng, Zhao, An, Liu, Liang, Gao, Yu, Zhang, Li, Jin, Wang, Bi, Liu, Wang, Shen, Chen, Zhang, Chen, Nie, Sun, Wang, Cheng, Liu, Xie, Liu, Yu, Song, Shan, Zhou, Yang, Li, Su, Lin, Li, Wang, Wei, Zhu, Zhang, Xu, Xu, Huang, Li, Zhao, Sun, Li, Wang, Yu, Zheng, Zhang, Shi, Xiong, He, Tang, Piao, Wang, Tan, Ma, Liu, Guo, Wu, Ou, Zhu, Wang, Gong, Zou, He, Zha, Xiong, Ma, Yan, Luo, You, Liu, Zhou, Wu, Ren, Ren, Sha, Fu, Xu, Huang, Zhang, Xie, Zhang, Hao,
  Gou, Ma, Yan, Shao, Xu, Wu, Zhang, Li, Gu, Zhu, Liu, Li, Xie, Song, Gao, and Pan]{deepseekv3}
DeepSeek-AI, A.~Liu, B.~Feng, B.~Xue, B.~Wang, B.~Wu, C.~Lu, C.~Zhao, C.~Deng, C.~Zhang, C.~Ruan, D.~Dai, D.~Guo, D.~Yang, D.~Chen, D.~Ji, E.~Li, F.~Lin, F.~Dai, F.~Luo, G.~Hao, G.~Chen, G.~Li, H.~Zhang, H.~Bao, H.~Xu, H.~Wang, H.~Zhang, H.~Ding, H.~Xin, H.~Gao, H.~Li, H.~Qu, J.~L. Cai, J.~Liang, J.~Guo, J.~Ni, J.~Li, J.~Wang, J.~Chen, J.~Chen, J.~Yuan, J.~Qiu, J.~Li, J.~Song, K.~Dong, K.~Hu, K.~Gao, K.~Guan, K.~Huang, K.~Yu, L.~Wang, L.~Zhang, L.~Xu, L.~Xia, L.~Zhao, L.~Wang, L.~Zhang, M.~Li, M.~Wang, M.~Zhang, M.~Zhang, M.~Tang, M.~Li, N.~Tian, P.~Huang, P.~Wang, P.~Zhang, Q.~Wang, Q.~Zhu, Q.~Chen, Q.~Du, R.~J. Chen, R.~L. Jin, R.~Ge, R.~Zhang, R.~Pan, R.~Wang, R.~Xu, R.~Zhang, R.~Chen, S.~S. Li, S.~Lu, S.~Zhou, S.~Chen, S.~Wu, S.~Ye, S.~Ye, S.~Ma, S.~Wang, S.~Zhou, S.~Yu, S.~Zhou, S.~Pan, T.~Wang, T.~Yun, T.~Pei, T.~Sun, W.~L. Xiao, W.~Zeng, W.~Zhao, W.~An, W.~Liu, W.~Liang, W.~Gao, W.~Yu, W.~Zhang, X.~Q. Li, X.~Jin, X.~Wang, X.~Bi, X.~Liu, X.~Wang, X.~Shen, X.~Chen, X.~Zhang, X.~Chen, X.~Nie, X.~Sun,
  X.~Wang, X.~Cheng, X.~Liu, X.~Xie, X.~Liu, X.~Yu, X.~Song, X.~Shan, X.~Zhou, X.~Yang, X.~Li, X.~Su, X.~Lin, Y.~K. Li, Y.~Q. Wang, Y.~X. Wei, Y.~X. Zhu, Y.~Zhang, Y.~Xu, Y.~Xu, Y.~Huang, Y.~Li, Y.~Zhao, Y.~Sun, Y.~Li, Y.~Wang, Y.~Yu, Y.~Zheng, Y.~Zhang, Y.~Shi, Y.~Xiong, Y.~He, Y.~Tang, Y.~Piao, Y.~Wang, Y.~Tan, Y.~Ma, Y.~Liu, Y.~Guo, Y.~Wu, Y.~Ou, Y.~Zhu, Y.~Wang, Y.~Gong, Y.~Zou, Y.~He, Y.~Zha, Y.~Xiong, Y.~Ma, Y.~Yan, Y.~Luo, Y.~You, Y.~Liu, Y.~Zhou, Z.~F. Wu, Z.~Z. Ren, Z.~Ren, Z.~Sha, Z.~Fu, Z.~Xu, Z.~Huang, Z.~Zhang, Z.~Xie, Z.~Zhang, Z.~Hao, Z.~Gou, Z.~Ma, Z.~Yan, Z.~Shao, Z.~Xu, Z.~Wu, Z.~Zhang, Z.~Li, Z.~Gu, Z.~Zhu, Z.~Liu, Z.~Li, Z.~Xie, Z.~Song, Z.~Gao, and Z.~Pan.
\newblock Deepseek-v3 technical report, 2025.
\newblock URL \url{https://arxiv.org/abs/2412.19437}.

\bibitem[Diao et~al.(2025)Diao, Yang, Fu, Dong, Su, Kliegl, Chen, Belcak, Suhara, Yin, Patwary, Yingyan, Lin, Kautz, and Molchanov]{diao2025climbclusteringbasediterativedata}
S.~Diao, Y.~Yang, Y.~Fu, X.~Dong, D.~Su, M.~Kliegl, Z.~Chen, P.~Belcak, Y.~Suhara, H.~Yin, M.~Patwary, Yingyan, Lin, J.~Kautz, and P.~Molchanov.
\newblock Climb: Clustering-based iterative data mixture bootstrapping for language model pre-training, 2025.
\newblock URL \url{https://arxiv.org/abs/2504.13161}.

\bibitem[Ding et~al.(2024)Ding, Wang, Paolini, Kumar, Deoras, Roth, and Soatto]{ding2024fewertruncationsimprovelanguage}
H.~Ding, Z.~Wang, G.~Paolini, V.~Kumar, A.~Deoras, D.~Roth, and S.~Soatto.
\newblock Fewer truncations improve language modeling, 2024.
\newblock URL \url{https://arxiv.org/abs/2404.10830}.

\bibitem[Ding et~al.(2023)Ding, Chen, Xu, Qin, Hu, Liu, Sun, and Zhou]{ding2023enhancing}
N.~Ding, Y.~Chen, B.~Xu, Y.~Qin, S.~Hu, Z.~Liu, M.~Sun, and B.~Zhou.
\newblock Enhancing chat language models by scaling high-quality instructional conversations.
\newblock In \emph{Proceedings of the 2023 Conference on Empirical Methods in Natural Language Processing}, pages 3029--3051, 2023.

\bibitem[D'Oosterlinck et~al.(2025)D'Oosterlinck, Xu, Develder, Demeester, Singh, Potts, Kiela, and Mehri]{d2025anchored}
K.~D'Oosterlinck, W.~Xu, C.~Develder, T.~Demeester, A.~Singh, C.~Potts, D.~Kiela, and S.~Mehri.
\newblock Anchored preference optimization and contrastive revisions: Addressing underspecification in alignment.
\newblock \emph{Transactions of the Association for Computational Linguistics}, 13:\penalty0 442--460, 2025.

\bibitem[Dua et~al.(2019)Dua, Wang, Dasigi, Stanovsky, Singh, and Gardner]{dua-etal-2019-drop}
D.~Dua, Y.~Wang, P.~Dasigi, G.~Stanovsky, S.~Singh, and M.~Gardner.
\newblock {DROP}: A reading comprehension benchmark requiring discrete reasoning over paragraphs.
\newblock In J.~Burstein, C.~Doran, and T.~Solorio, editors, \emph{Proceedings of the 2019 Conference of the North {A}merican Chapter of the Association for Computational Linguistics: Human Language Technologies, Volume 1 (Long and Short Papers)}, pages 2368--2378, Minneapolis, Minnesota, June 2019. Association for Computational Linguistics.
\newblock \doi{10.18653/v1/N19-1246}.
\newblock URL \url{https://aclanthology.org/N19-1246}.

\bibitem[Dubois et~al.(2024)Dubois, Galambosi, Liang, and Hashimoto]{dubois2024length}
Y.~Dubois, B.~Galambosi, P.~Liang, and T.~B. Hashimoto.
\newblock Length-controlled alpacaeval: A simple way to debias automatic evaluators.
\newblock \emph{arXiv preprint arXiv:2404.04475}, 2024.

\bibitem[Fan et~al.(2019)Fan, Jernite, Perez, Grangier, Weston, and Auli]{fan2019eli5}
A.~Fan, Y.~Jernite, E.~Perez, D.~Grangier, J.~Weston, and M.~Auli.
\newblock Eli5: Long form question answering.
\newblock In \emph{Proceedings of the 57th Annual Meeting of the Association for Computational Linguistics}, pages 3558--3567, 2019.

\bibitem[Fang et~al.(2025{\natexlab{a}})Fang, Pouransari, Jordan, Toshev, Shankar, Schmidt, and Gunter]{fang2025datasetsdocumentsrepetitionspracticalities}
A.~Fang, H.~Pouransari, M.~Jordan, A.~Toshev, V.~Shankar, L.~Schmidt, and T.~Gunter.
\newblock Datasets, documents, and repetitions: The practicalities of unequal data quality, 2025{\natexlab{a}}.
\newblock URL \url{https://arxiv.org/abs/2503.07879}.

\bibitem[Fang et~al.(2025{\natexlab{b}})Fang, Wang, Liu, Zhang, Jegelka, Gao, Ding, and Wang]{fang2025wrongperplexitylongcontextlanguage}
L.~Fang, Y.~Wang, Z.~Liu, C.~Zhang, S.~Jegelka, J.~Gao, B.~Ding, and Y.~Wang.
\newblock What is wrong with perplexity for long-context language modeling?, 2025{\natexlab{b}}.
\newblock URL \url{https://arxiv.org/abs/2410.23771}.

\bibitem[Fleming and Daw(2017)]{Fleming2017-FLESOD}
S.~Fleming and N.~Daw.
\newblock Self-evaluation of decision-making: A general bayesian framework for metacognitive computation.
\newblock \emph{Psychological Review}, 124\penalty0 (1):\penalty0 91--114, 2017.
\newblock \doi{10.1037/rev0000045}.

\bibitem[Fujii et~al.(2025)Fujii, Tajima, Mizuki, Shimada, Shiotani, Saito, Ohi, Kawamura, Nakamura, Okamoto, et~al.]{fujii2025rewriting}
K.~Fujii, Y.~Tajima, S.~Mizuki, H.~Shimada, T.~Shiotani, K.~Saito, M.~Ohi, M.~Kawamura, T.~Nakamura, T.~Okamoto, et~al.
\newblock Rewriting pre-training data boosts llm performance in math and code.
\newblock \emph{arXiv preprint arXiv:2505.02881}, 2025.

\bibitem[Gandhi et~al.(2025)Gandhi, Chakravarthy, Singh, Lile, and Goodman]{gandhi2025cognitive}
K.~Gandhi, A.~Chakravarthy, A.~Singh, N.~Lile, and N.~D. Goodman.
\newblock Cognitive behaviors that enable self-improving reasoners, or, four habits of highly effective stars.
\newblock \emph{arXiv preprint arXiv:2503.01307}, 2025.

\bibitem[Gao et~al.(2020)Gao, Biderman, Black, Golding, Hoppe, Foster, Phang, He, Thite, Nabeshima, et~al.]{gao2020pile}
L.~Gao, S.~Biderman, S.~Black, L.~Golding, T.~Hoppe, C.~Foster, J.~Phang, H.~He, A.~Thite, N.~Nabeshima, et~al.
\newblock The pile: An 800gb dataset of diverse text for language modeling.
\newblock \emph{arXiv preprint arXiv:2101.00027}, 2020.

\bibitem[Gao et~al.(2025)Gao, Wettig, Yen, and Chen]{prolong}
T.~Gao, A.~Wettig, H.~Yen, and D.~Chen.
\newblock How to train long-context language models (effectively).
\newblock In \emph{ACL}, 2025.

\bibitem[{Gemma 3 Team}(2025)]{team2025gemma3}
{Gemma 3 Team}.
\newblock Gemma 3 technical report, 2025.
\newblock URL \url{https://arxiv.org/abs/2503.19786}.

\bibitem[{Gemma Team} et~al.(2024){Gemma Team}, Mesnard, Hardin, Dadashi, Bhupatiraju, Pathak, Sifre, Rivi{\`e}re, Kale, Love, et~al.]{gemma}
{Gemma Team}, T.~Mesnard, C.~Hardin, R.~Dadashi, S.~Bhupatiraju, S.~Pathak, L.~Sifre, M.~Rivi{\`e}re, M.~S. Kale, J.~Love, et~al.
\newblock Gemma: Open models based on gemini research and technology.
\newblock \emph{arXiv preprint arXiv:2403.08295}, 2024.

\bibitem[Geng et~al.(2025)Geng, Ivison, Li, Sap, Li, Krishna, and Koh]{geng2025delta}
S.~Geng, H.~Ivison, C.-L. Li, M.~Sap, J.~Li, R.~Krishna, and P.~W. Koh.
\newblock The delta learning hypothesis: Preference tuning on weak data can yield strong gains.
\newblock \emph{arXiv preprint arXiv:2507.06187}, 2025.

\bibitem[{GLM-4.5 Team} et~al.(2025){GLM-4.5 Team}, Zeng, Lv, Zheng, Hou, Chen, Xie, Wang, Yin, Zeng, Zhang, Wang, Zhong, Liu, Lu, Cao, Zhang, Huang, Wei, Cheng, An, Niu, Wen, Bai, Du, Wang, Zhu, Zhang, Wen, Wu, Xu, Huang, Zhao, Cai, Yu, Li, Ge, Huang, Zhang, Xu, Zhu, Li, Yin, Lin, Yang, Jiang, Ai, Zhu, Wang, Pan, Wang, Sun, Li, Li, Hu, Zhang, Peng, Tai, Zhang, Wang, Yang, Liu, Zhao, Liu, Yan, Liu, Chen, Li, Zhao, Ren, Jiao, Zhao, Yan, Wang, Gui, Zhao, Liu, Li, Li, Lu, Wang, Yuan, Li, Du, Du, Liu, Zhi, Gao, Wang, Yang, Xu, Fan, Wu, Ding, Wang, Zhang, Li, Xu, Zhao, Zhai, Du, Dong, Lei, Tu, Yang, Lu, Li, Li, Shuang-Li, Yang, Yi, Yu, Tian, Wang, Yu, Tam, Liang, Liu, Wang, Jia, Gu, Ling, Wang, Fan, Pan, Zhang, Zhang, Fu, Zhang, Xu, Wu, Lu, Wang, Zhou, Pan, Zhang, Wang, Li, Su, Geng, Zhu, Yang, Li, Wu, Li, Liu, Wang, Li, Zhang, Liu, Yang, Zhou, Qiao, Feng, Liu, Zhang, Wang, Yao, Wang, Liu, Chai, Li, Zhao, Chen, Zhai, Xu, Huang, Wang, Li, Dong, and Tang]{glm45}
{GLM-4.5 Team}, A.~Zeng, X.~Lv, Q.~Zheng, Z.~Hou, B.~Chen, C.~Xie, C.~Wang, D.~Yin, H.~Zeng, J.~Zhang, K.~Wang, L.~Zhong, M.~Liu, R.~Lu, S.~Cao, X.~Zhang, X.~Huang, Y.~Wei, Y.~Cheng, Y.~An, Y.~Niu, Y.~Wen, Y.~Bai, Z.~Du, Z.~Wang, Z.~Zhu, B.~Zhang, B.~Wen, B.~Wu, B.~Xu, C.~Huang, C.~Zhao, C.~Cai, C.~Yu, C.~Li, C.~Ge, C.~Huang, C.~Zhang, C.~Xu, C.~Zhu, C.~Li, C.~Yin, D.~Lin, D.~Yang, D.~Jiang, D.~Ai, E.~Zhu, F.~Wang, G.~Pan, G.~Wang, H.~Sun, H.~Li, H.~Li, H.~Hu, H.~Zhang, H.~Peng, H.~Tai, H.~Zhang, H.~Wang, H.~Yang, H.~Liu, H.~Zhao, H.~Liu, H.~Yan, H.~Liu, H.~Chen, J.~Li, J.~Zhao, J.~Ren, J.~Jiao, J.~Zhao, J.~Yan, J.~Wang, J.~Gui, J.~Zhao, J.~Liu, J.~Li, J.~Li, J.~Lu, J.~Wang, J.~Yuan, J.~Li, J.~Du, J.~Du, J.~Liu, J.~Zhi, J.~Gao, K.~Wang, L.~Yang, L.~Xu, L.~Fan, L.~Wu, L.~Ding, L.~Wang, M.~Zhang, M.~Li, M.~Xu, M.~Zhao, M.~Zhai, P.~Du, Q.~Dong, S.~Lei, S.~Tu, S.~Yang, S.~Lu, S.~Li, S.~Li, Shuang-Li, S.~Yang, S.~Yi, T.~Yu, W.~Tian, W.~Wang, W.~Yu, W.~L. Tam, W.~Liang, W.~Liu, X.~Wang, X.~Jia, X.~Gu, X.~Ling,
  X.~Wang, X.~Fan, X.~Pan, X.~Zhang, X.~Zhang, X.~Fu, X.~Zhang, Y.~Xu, Y.~Wu, Y.~Lu, Y.~Wang, Y.~Zhou, Y.~Pan, Y.~Zhang, Y.~Wang, Y.~Li, Y.~Su, Y.~Geng, Y.~Zhu, Y.~Yang, Y.~Li, Y.~Wu, Y.~Li, Y.~Liu, Y.~Wang, Y.~Li, Y.~Zhang, Z.~Liu, Z.~Yang, Z.~Zhou, Z.~Qiao, Z.~Feng, Z.~Liu, Z.~Zhang, Z.~Wang, Z.~Yao, Z.~Wang, Z.~Liu, Z.~Chai, Z.~Li, Z.~Zhao, W.~Chen, J.~Zhai, B.~Xu, M.~Huang, H.~Wang, J.~Li, Y.~Dong, and J.~Tang.
\newblock {GLM-4.5: Agentic, Reasoning, and Coding (ARC) Foundation Models}, 2025.
\newblock URL \url{https://arxiv.org/abs/2508.06471}.

\bibitem[Goddard(2025)]{goddard2025extendingAFM}
C.~Goddard.
\newblock Extending {AFM}-4.{5B} to {64K} context length.
\newblock \url{https://www.arcee.ai/blog/extending-afm-4-5b-to-64k-context-length}, June 2025.
\newblock Accessed: 2025-11-10.

\bibitem[Goddard et~al.(2024)Goddard, Siriwardhana, Ehghaghi, Meyers, Karpukhin, Benedict, McQuade, and Solawetz]{goddard-etal-2024-arcees}
C.~Goddard, S.~Siriwardhana, M.~Ehghaghi, L.~Meyers, V.~Karpukhin, B.~Benedict, M.~McQuade, and J.~Solawetz.
\newblock Arcee{'}s {M}erge{K}it: A toolkit for merging large language models.
\newblock In F.~Dernoncourt, D.~Preo{\c{t}}iuc-Pietro, and A.~Shimorina, editors, \emph{Proceedings of the 2024 Conference on Empirical Methods in Natural Language Processing: Industry Track}, pages 477--485, Miami, Florida, US, Nov. 2024. Association for Computational Linguistics.
\newblock \doi{10.18653/v1/2024.emnlp-industry.36}.
\newblock URL \url{https://aclanthology.org/2024.emnlp-industry.36}.

\bibitem[Godey et~al.(2025)Godey, Antoun, Touchent, Bawden, Éric de~la Clergerie, Sagot, and Seddah]{godey2025gaperonpepperedenglishfrenchgenerative}
N.~Godey, W.~Antoun, R.~Touchent, R.~Bawden, Éric de~la Clergerie, B.~Sagot, and D.~Seddah.
\newblock Gaperon: A peppered english-french generative language model suite, 2025.
\newblock URL \url{https://arxiv.org/abs/2510.25771}.

\bibitem[Grattafiori et~al.(2024)Grattafiori, Dubey, Jauhri, Pandey, Kadian, Al-Dahle, Letman, Mathurx, Schelten, Vaughan, Yang, Fan, Goyal, Hartshorn, Yang, Mitra, Sravankumar, Korenev, Hinsvark, Rao, Zhang, Rodriguez, Gregerson, Spataru, Roziere, Biron, Tang, Chern, Caucheteux, Nayak, Bi, Marra, McConnell, Keller, Touret, Wu, Wong, Ferrer, Nikolaidis, Allonsius, Song, Pintz, Livshits, Wyatt, Esiobu, Choudhary, Mahajan, Garcia-Olano, Perino, Hupkes, Lakomkin, AlBadawy, Lobanova, Dinan, Smith, Radenovic, Guzmán, Zhang, Synnaeve, Lee, Anderson, Thattai, Nail, Mialon, Pang, Cucurell, Nguyen, Korevaar, Xu, Touvron, Zarov, Ibarra, Kloumann, Misra, Evtimov, Zhang, Copet, Lee, Geffert, Vranes, Park, Mahadeokar, Shah, van~der Linde, Billock, Hong, Lee, Fu, Chi, Huang, Liu, Wang, Yu, Bitton, Spisak, Park, Rocca, Johnstun, Saxe, Jia, Alwala, Prasad, Upasani, Plawiak, Li, Heafield, Stone, El-Arini, Iyer, Malik, Chiu, Bhalla, Lakhotia, Rantala-Yeary, van~der Maaten, Chen, Tan, Jenkins, Martin, Madaan, Malo, Blecher,
  Landzaat, de~Oliveira, Muzzi, Pasupuleti, Singh, Paluri, Kardas, Tsimpoukelli, Oldham, Rita, Pavlova, Kambadur, Lewis, Si, Singh, Hassan, Goyal, Torabi, Bashlykov, Bogoychev, Chatterji, Zhang, Duchenne, Çelebi, Alrassy, Zhang, Li, Vasic, Weng, Bhargava, Dubal, Krishnan, Koura, Xu, He, Dong, Srinivasan, Ganapathy, Calderer, Cabral, Stojnic, Raileanu, Maheswari, Girdhar, Patel, Sauvestre, Polidoro, Sumbaly, Taylor, Silva, Hou, Wang, Hosseini, Chennabasappa, Singh, Bell, Kim, Edunov, Nie, Narang, Raparthy, Shen, Wan, Bhosale, Zhang, Vandenhende, Batra, Whitman, Sootla, Collot, Gururangan, Borodinsky, Herman, Fowler, Sheasha, Georgiou, Scialom, Speckbacher, Mihaylov, Xiao, Karn, Goswami, Gupta, Ramanathan, Kerkez, Gonguet, Do, Vogeti, Albiero, Petrovic, Chu, Xiong, Fu, Meers, Martinet, Wang, Wang, Tan, Xia, Xie, Jia, Wang, Goldschlag, Gaur, Babaei, Wen, Song, Zhang, Li, Mao, Coudert, Yan, Chen, Papakipos, Singh, Srivastava, Jain, Kelsey, Shajnfeld, Gangidi, Victoria, Goldstand, Menon, Sharma, Boesenberg,
  Baevski, Feinstein, Kallet, Sangani, Teo, Yunus, Lupu, Alvarado, Caples, Gu, Ho, Poulton, Ryan, Ramchandani, Dong, Franco, Goyal, Saraf, Chowdhury, Gabriel, Bharambe, Eisenman, Yazdan, James, Maurer, Leonhardi, Huang, Loyd, Paola, Paranjape, Liu, Wu, Ni, Hancock, Wasti, Spence, Stojkovic, Gamido, Montalvo, Parker, Burton, Mejia, Liu, Wang, Kim, Zhou, Hu, Chu, Cai, Tindal, Feichtenhofer, Gao, Civin, Beaty, Kreymer, Li, Adkins, Xu, Testuggine, David, Parikh, Liskovich, Foss, Wang, Le, Holland, Dowling, Jamil, Montgomery, Presani, Hahn, Wood, Le, Brinkman, Arcaute, Dunbar, Smothers, Sun, Kreuk, Tian, Kokkinos, Ozgenel, Caggioni, Kanayet, Seide, Florez, Schwarz, Badeer, Swee, Halpern, Herman, Sizov, Guangyi, Zhang, Lakshminarayanan, Inan, Shojanazeri, Zou, Wang, Zha, Habeeb, Rudolph, Suk, Aspegren, Goldman, Zhan, Damlaj, Molybog, Tufanov, Leontiadis, Veliche, Gat, Weissman, Geboski, Kohli, Lam, Asher, Gaya, Marcus, Tang, Chan, Zhen, Reizenstein, Teboul, Zhong, Jin, Yang, Cummings, Carvill, Shepard, McPhie,
  Torres, Ginsburg, Wang, Wu, U, Saxena, Khandelwal, Zand, Matosich, Veeraraghavan, Michelena, Li, Jagadeesh, Huang, Chawla, Huang, Chen, Garg, A, Silva, Bell, Zhang, Guo, Yu, Moshkovich, Wehrstedt, Khabsa, Avalani, Bhatt, Mankus, Hasson, Lennie, Reso, Groshev, Naumov, Lathi, Keneally, Liu, Seltzer, Valko, Restrepo, Patel, Vyatskov, Samvelyan, Clark, Macey, Wang, Hermoso, Metanat, Rastegari, Bansal, Santhanam, Parks, White, Bawa, Singhal, Egebo, Usunier, Mehta, Laptev, Dong, Cheng, Chernoguz, Hart, Salpekar, Kalinli, Kent, Parekh, Saab, Balaji, Rittner, Bontrager, Roux, Dollar, Zvyagina, Ratanchandani, Yuvraj, Liang, Alao, Rodriguez, Ayub, Murthy, Nayani, Mitra, Parthasarathy, Li, Hogan, Battey, Wang, Howes, Rinott, Mehta, Siby, Bondu, Datta, Chugh, Hunt, Dhillon, Sidorov, Pan, Mahajan, Verma, Yamamoto, Ramaswamy, Lindsay, Lindsay, Feng, Lin, Zha, Patil, Shankar, Zhang, Zhang, Wang, Agarwal, Sajuyigbe, Chintala, Max, Chen, Kehoe, Satterfield, Govindaprasad, Gupta, Deng, Cho, Virk, Subramanian, Choudhury,
  Goldman, Remez, Glaser, Best, Koehler, Robinson, Li, Zhang, Matthews, Chou, Shaked, Vontimitta, Ajayi, Montanez, Mohan, Kumar, Mangla, Ionescu, Poenaru, Mihailescu, Ivanov, Li, Wang, Jiang, Bouaziz, Constable, Tang, Wu, Wang, Wu, Gao, Kleinman, Chen, Hu, Jia, Qi, Li, Zhang, Zhang, Adi, Nam, Yu, Wang, Zhao, Hao, Qian, Li, He, Rait, DeVito, Rosnbrick, Wen, Yang, Zhao, and Ma]{dubey2024llama}
A.~Grattafiori, A.~Dubey, A.~Jauhri, A.~Pandey, A.~Kadian, A.~Al-Dahle, A.~Letman, A.~Mathurx, A.~Schelten, A.~Vaughan, A.~Yang, A.~Fan, A.~Goyal, A.~Hartshorn, A.~Yang, A.~Mitra, A.~Sravankumar, A.~Korenev, A.~Hinsvark, A.~Rao, A.~Zhang, A.~Rodriguez, A.~Gregerson, A.~Spataru, B.~Roziere, B.~Biron, B.~Tang, B.~Chern, C.~Caucheteux, C.~Nayak, C.~Bi, C.~Marra, C.~McConnell, C.~Keller, C.~Touret, C.~Wu, C.~Wong, C.~C. Ferrer, C.~Nikolaidis, D.~Allonsius, D.~Song, D.~Pintz, D.~Livshits, D.~Wyatt, D.~Esiobu, D.~Choudhary, D.~Mahajan, D.~Garcia-Olano, D.~Perino, D.~Hupkes, E.~Lakomkin, E.~AlBadawy, E.~Lobanova, E.~Dinan, E.~M. Smith, F.~Radenovic, F.~Guzmán, F.~Zhang, G.~Synnaeve, G.~Lee, G.~L. Anderson, G.~Thattai, G.~Nail, G.~Mialon, G.~Pang, G.~Cucurell, H.~Nguyen, H.~Korevaar, H.~Xu, H.~Touvron, I.~Zarov, I.~A. Ibarra, I.~Kloumann, I.~Misra, I.~Evtimov, J.~Zhang, J.~Copet, J.~Lee, J.~Geffert, J.~Vranes, J.~Park, J.~Mahadeokar, J.~Shah, J.~van~der Linde, J.~Billock, J.~Hong, J.~Lee, J.~Fu, J.~Chi, J.~Huang,
  J.~Liu, J.~Wang, J.~Yu, J.~Bitton, J.~Spisak, J.~Park, J.~Rocca, J.~Johnstun, J.~Saxe, J.~Jia, K.~V. Alwala, K.~Prasad, K.~Upasani, K.~Plawiak, K.~Li, K.~Heafield, K.~Stone, K.~El-Arini, K.~Iyer, K.~Malik, K.~Chiu, K.~Bhalla, K.~Lakhotia, L.~Rantala-Yeary, L.~van~der Maaten, L.~Chen, L.~Tan, L.~Jenkins, L.~Martin, L.~Madaan, L.~Malo, L.~Blecher, L.~Landzaat, L.~de~Oliveira, M.~Muzzi, M.~Pasupuleti, M.~Singh, M.~Paluri, M.~Kardas, M.~Tsimpoukelli, M.~Oldham, M.~Rita, M.~Pavlova, M.~Kambadur, M.~Lewis, M.~Si, M.~K. Singh, M.~Hassan, N.~Goyal, N.~Torabi, N.~Bashlykov, N.~Bogoychev, N.~Chatterji, N.~Zhang, O.~Duchenne, O.~Çelebi, P.~Alrassy, P.~Zhang, P.~Li, P.~Vasic, P.~Weng, P.~Bhargava, P.~Dubal, P.~Krishnan, P.~S. Koura, P.~Xu, Q.~He, Q.~Dong, R.~Srinivasan, R.~Ganapathy, R.~Calderer, R.~S. Cabral, R.~Stojnic, R.~Raileanu, R.~Maheswari, R.~Girdhar, R.~Patel, R.~Sauvestre, R.~Polidoro, R.~Sumbaly, R.~Taylor, R.~Silva, R.~Hou, R.~Wang, S.~Hosseini, S.~Chennabasappa, S.~Singh, S.~Bell, S.~S. Kim, S.~Edunov,
  S.~Nie, S.~Narang, S.~Raparthy, S.~Shen, S.~Wan, S.~Bhosale, S.~Zhang, S.~Vandenhende, S.~Batra, S.~Whitman, S.~Sootla, S.~Collot, S.~Gururangan, S.~Borodinsky, T.~Herman, T.~Fowler, T.~Sheasha, T.~Georgiou, T.~Scialom, T.~Speckbacher, T.~Mihaylov, T.~Xiao, U.~Karn, V.~Goswami, V.~Gupta, V.~Ramanathan, V.~Kerkez, V.~Gonguet, V.~Do, V.~Vogeti, V.~Albiero, V.~Petrovic, W.~Chu, W.~Xiong, W.~Fu, W.~Meers, X.~Martinet, X.~Wang, X.~Wang, X.~E. Tan, X.~Xia, X.~Xie, X.~Jia, X.~Wang, Y.~Goldschlag, Y.~Gaur, Y.~Babaei, Y.~Wen, Y.~Song, Y.~Zhang, Y.~Li, Y.~Mao, Z.~D. Coudert, Z.~Yan, Z.~Chen, Z.~Papakipos, A.~Singh, A.~Srivastava, A.~Jain, A.~Kelsey, A.~Shajnfeld, A.~Gangidi, A.~Victoria, A.~Goldstand, A.~Menon, A.~Sharma, A.~Boesenberg, A.~Baevski, A.~Feinstein, A.~Kallet, A.~Sangani, A.~Teo, A.~Yunus, A.~Lupu, A.~Alvarado, A.~Caples, A.~Gu, A.~Ho, A.~Poulton, A.~Ryan, A.~Ramchandani, A.~Dong, A.~Franco, A.~Goyal, A.~Saraf, A.~Chowdhury, A.~Gabriel, A.~Bharambe, A.~Eisenman, A.~Yazdan, B.~James, B.~Maurer,
  B.~Leonhardi, B.~Huang, B.~Loyd, B.~D. Paola, B.~Paranjape, B.~Liu, B.~Wu, B.~Ni, B.~Hancock, B.~Wasti, B.~Spence, B.~Stojkovic, B.~Gamido, B.~Montalvo, C.~Parker, C.~Burton, C.~Mejia, C.~Liu, C.~Wang, C.~Kim, C.~Zhou, C.~Hu, C.-H. Chu, C.~Cai, C.~Tindal, C.~Feichtenhofer, C.~Gao, D.~Civin, D.~Beaty, D.~Kreymer, D.~Li, D.~Adkins, D.~Xu, D.~Testuggine, D.~David, D.~Parikh, D.~Liskovich, D.~Foss, D.~Wang, D.~Le, D.~Holland, E.~Dowling, E.~Jamil, E.~Montgomery, E.~Presani, E.~Hahn, E.~Wood, E.-T. Le, E.~Brinkman, E.~Arcaute, E.~Dunbar, E.~Smothers, F.~Sun, F.~Kreuk, F.~Tian, F.~Kokkinos, F.~Ozgenel, F.~Caggioni, F.~Kanayet, F.~Seide, G.~M. Florez, G.~Schwarz, G.~Badeer, G.~Swee, G.~Halpern, G.~Herman, G.~Sizov, Guangyi, Zhang, G.~Lakshminarayanan, H.~Inan, H.~Shojanazeri, H.~Zou, H.~Wang, H.~Zha, H.~Habeeb, H.~Rudolph, H.~Suk, H.~Aspegren, H.~Goldman, H.~Zhan, I.~Damlaj, I.~Molybog, I.~Tufanov, I.~Leontiadis, I.-E. Veliche, I.~Gat, J.~Weissman, J.~Geboski, J.~Kohli, J.~Lam, J.~Asher, J.-B. Gaya, J.~Marcus,
  J.~Tang, J.~Chan, J.~Zhen, J.~Reizenstein, J.~Teboul, J.~Zhong, J.~Jin, J.~Yang, J.~Cummings, J.~Carvill, J.~Shepard, J.~McPhie, J.~Torres, J.~Ginsburg, J.~Wang, K.~Wu, K.~H. U, K.~Saxena, K.~Khandelwal, K.~Zand, K.~Matosich, K.~Veeraraghavan, K.~Michelena, K.~Li, K.~Jagadeesh, K.~Huang, K.~Chawla, K.~Huang, L.~Chen, L.~Garg, L.~A, L.~Silva, L.~Bell, L.~Zhang, L.~Guo, L.~Yu, L.~Moshkovich, L.~Wehrstedt, M.~Khabsa, M.~Avalani, M.~Bhatt, M.~Mankus, M.~Hasson, M.~Lennie, M.~Reso, M.~Groshev, M.~Naumov, M.~Lathi, M.~Keneally, M.~Liu, M.~L. Seltzer, M.~Valko, M.~Restrepo, M.~Patel, M.~Vyatskov, M.~Samvelyan, M.~Clark, M.~Macey, M.~Wang, M.~J. Hermoso, M.~Metanat, M.~Rastegari, M.~Bansal, N.~Santhanam, N.~Parks, N.~White, N.~Bawa, N.~Singhal, N.~Egebo, N.~Usunier, N.~Mehta, N.~P. Laptev, N.~Dong, N.~Cheng, O.~Chernoguz, O.~Hart, O.~Salpekar, O.~Kalinli, P.~Kent, P.~Parekh, P.~Saab, P.~Balaji, P.~Rittner, P.~Bontrager, P.~Roux, P.~Dollar, P.~Zvyagina, P.~Ratanchandani, P.~Yuvraj, Q.~Liang, R.~Alao, R.~Rodriguez,
  R.~Ayub, R.~Murthy, R.~Nayani, R.~Mitra, R.~Parthasarathy, R.~Li, R.~Hogan, R.~Battey, R.~Wang, R.~Howes, R.~Rinott, S.~Mehta, S.~Siby, S.~J. Bondu, S.~Datta, S.~Chugh, S.~Hunt, S.~Dhillon, S.~Sidorov, S.~Pan, S.~Mahajan, S.~Verma, S.~Yamamoto, S.~Ramaswamy, S.~Lindsay, S.~Lindsay, S.~Feng, S.~Lin, S.~C. Zha, S.~Patil, S.~Shankar, S.~Zhang, S.~Zhang, S.~Wang, S.~Agarwal, S.~Sajuyigbe, S.~Chintala, S.~Max, S.~Chen, S.~Kehoe, S.~Satterfield, S.~Govindaprasad, S.~Gupta, S.~Deng, S.~Cho, S.~Virk, S.~Subramanian, S.~Choudhury, S.~Goldman, T.~Remez, T.~Glaser, T.~Best, T.~Koehler, T.~Robinson, T.~Li, T.~Zhang, T.~Matthews, T.~Chou, T.~Shaked, V.~Vontimitta, V.~Ajayi, V.~Montanez, V.~Mohan, V.~S. Kumar, V.~Mangla, V.~Ionescu, V.~Poenaru, V.~T. Mihailescu, V.~Ivanov, W.~Li, W.~Wang, W.~Jiang, W.~Bouaziz, W.~Constable, X.~Tang, X.~Wu, X.~Wang, X.~Wu, X.~Gao, Y.~Kleinman, Y.~Chen, Y.~Hu, Y.~Jia, Y.~Qi, Y.~Li, Y.~Zhang, Y.~Zhang, Y.~Adi, Y.~Nam, Yu, Wang, Y.~Zhao, Y.~Hao, Y.~Qian, Y.~Li, Y.~He, Z.~Rait, Z.~DeVito,
  Z.~Rosnbrick, Z.~Wen, Z.~Yang, Z.~Zhao, and Z.~Ma.
\newblock The llama 3 herd of models, 2024.
\newblock URL \url{https://arxiv.org/abs/2407.21783}.

\bibitem[Griffiths et~al.(2019)Griffiths, Callaway, Chang, Grant, Krueger, and Lieder]{GRIFFITHS201924}
T.~L. Griffiths, F.~Callaway, M.~B. Chang, E.~Grant, P.~M. Krueger, and F.~Lieder.
\newblock Doing more with less: meta-reasoning and meta-learning in humans and machines.
\newblock \emph{Current Opinion in Behavioral Sciences}, 29:\penalty0 24--30, 2019.
\newblock ISSN 2352-1546.
\newblock \doi{https://doi.org/10.1016/j.cobeha.2019.01.005}.
\newblock URL \url{https://www.sciencedirect.com/science/article/pii/S2352154618302122}.
\newblock Artificial Intelligence.

\bibitem[Gu et~al.(2024{\natexlab{a}})Gu, Rozi{\`e}re, Leather, Solar-Lezama, Synnaeve, and Wang]{gu2024cruxeval}
A.~Gu, B.~Rozi{\`e}re, H.~Leather, A.~Solar-Lezama, G.~Synnaeve, and S.~I. Wang.
\newblock Cruxeval: A benchmark for code reasoning, understanding and execution.
\newblock \emph{arXiv preprint arXiv:2401.03065}, 2024{\natexlab{a}}.

\bibitem[Gu et~al.(2024{\natexlab{b}})Gu, Tafjord, Kuehl, Haddad, Dodge, and Hajishirzi]{olmes}
Y.~Gu, O.~Tafjord, B.~Kuehl, D.~Haddad, J.~Dodge, and H.~Hajishirzi.
\newblock Olmes: A standard for language model evaluations.
\newblock \emph{ArXiv}, abs/2406.08446, 2024{\natexlab{b}}.
\newblock URL \url{https://api.semanticscholar.org/CorpusID:270391754}.

\bibitem[Guha et~al.(2025{\natexlab{a}})Guha, Marten, Keh, Raoof, Smyrnis, Bansal, Nezhurina, Mercat, Vu, Sprague, Suvarna, Feuer, Chen, Khan, Frankel, Grover, Choi, Muennighoff, Su, Zhao, Yang, Pimpalgaonkar, Sharma, Ji, Deng, Pratt, Ramanujan, Saad-Falcon, Li, Dave, Albalak, Arora, Wulfe, Hegde, Durrett, Oh, Bansal, Gabriel, Grover, Chang, Shankar, Gokaslan, Merrill, Hashimoto, Choi, Jitsev, Heckel, Sathiamoorthy, Dimakis, and Schmidt]{guha2025openthoughts}
E.~Guha, R.~Marten, S.~Keh, N.~Raoof, G.~Smyrnis, H.~Bansal, M.~Nezhurina, J.~Mercat, T.~Vu, Z.~Sprague, A.~Suvarna, B.~Feuer, L.~Chen, Z.~Khan, E.~Frankel, S.~Grover, C.~Choi, N.~Muennighoff, S.~Su, W.~Zhao, J.~Yang, S.~Pimpalgaonkar, K.~Sharma, C.~C.-J. Ji, Y.~Deng, S.~Pratt, V.~Ramanujan, J.~Saad-Falcon, J.~Li, A.~Dave, A.~Albalak, K.~Arora, B.~Wulfe, C.~Hegde, G.~Durrett, S.~Oh, M.~Bansal, S.~Gabriel, A.~Grover, K.-W. Chang, V.~Shankar, A.~Gokaslan, M.~A. Merrill, T.~Hashimoto, Y.~Choi, J.~Jitsev, R.~Heckel, M.~Sathiamoorthy, A.~G. Dimakis, and L.~Schmidt.
\newblock Openthoughts: Data recipes for reasoning models.
\newblock \emph{arXiv preprint arXiv:2506.04178}, 2025{\natexlab{a}}.
\newblock URL \url{https://arxiv.org/abs/2506.04178}.

\bibitem[Guha et~al.(2025{\natexlab{b}})Guha, Marten, Keh, Raoof, Smyrnis, Bansal, Nezhurina, Mercat, Vu, Sprague, Suvarna, Feuer, Chen, Khan, Frankel, Grover, Choi, Muennighoff, Su, Zhao, Yang, Pimpalgaonkar, Sharma, Ji, Deng, Pratt, Ramanujan, Saad-Falcon, Li, Dave, Albalak, Arora, Wulfe, Hegde, Durrett, Oh, Bansal, Gabriel, Grover, Chang, Shankar, Gokaslan, Merrill, Hashimoto, Choi, Jitsev, Heckel, Sathiamoorthy, Dimakis, and Schmidt]{guha2025openthoughtsdatarecipesreasoning}
E.~Guha, R.~Marten, S.~Keh, N.~Raoof, G.~Smyrnis, H.~Bansal, M.~Nezhurina, J.~Mercat, T.~Vu, Z.~Sprague, A.~Suvarna, B.~Feuer, L.~Chen, Z.~Khan, E.~Frankel, S.~Grover, C.~Choi, N.~Muennighoff, S.~Su, W.~Zhao, J.~Yang, S.~Pimpalgaonkar, K.~Sharma, C.~C.-J. Ji, Y.~Deng, S.~Pratt, V.~Ramanujan, J.~Saad-Falcon, J.~Li, A.~Dave, A.~Albalak, K.~Arora, B.~Wulfe, C.~Hegde, G.~Durrett, S.~Oh, M.~Bansal, S.~Gabriel, A.~Grover, K.-W. Chang, V.~Shankar, A.~Gokaslan, M.~A. Merrill, T.~Hashimoto, Y.~Choi, J.~Jitsev, R.~Heckel, M.~Sathiamoorthy, A.~G. Dimakis, and L.~Schmidt.
\newblock Openthoughts: Data recipes for reasoning models, 2025{\natexlab{b}}.
\newblock URL \url{https://arxiv.org/abs/2506.04178}.

\bibitem[Guo et~al.(2024)Guo, Zhu, Yang, Xie, Dong, Zhang, Chen, Bi, Wu, Li, et~al.]{guo2024deepseek}
D.~Guo, Q.~Zhu, D.~Yang, Z.~Xie, K.~Dong, W.~Zhang, G.~Chen, X.~Bi, Y.~Wu, Y.~Li, et~al.
\newblock Deepseek-coder: When the large language model meets programming--the rise of code intelligence.
\newblock \emph{arXiv preprint arXiv:2401.14196}, 2024.

\bibitem[Guo et~al.(2025)Guo, Yang, Zhang, Song, Zhang, Xu, Zhu, Ma, Wang, Bi, et~al.]{guo2025deepseek}
D.~Guo, D.~Yang, H.~Zhang, J.~Song, R.~Zhang, R.~Xu, Q.~Zhu, S.~Ma, P.~Wang, X.~Bi, et~al.
\newblock Deepseek-r1: Incentivizing reasoning capability in llms via reinforcement learning.
\newblock \emph{arXiv preprint arXiv:2501.12948}, 2025.

\bibitem[Hall et~al.(2025)Hall, Chou, Garg, Ravi, Liu, Shandilya, Ahmed, Liang, Kuditipudi, {J38}, Lee, Power, Salahi, Held, Wang, {chiheem}, Niklaus, Mai, {dependabot[bot]}, Zhou, Li, Yang, Karamcheti, Williams, Zhou, Ramaswami, {whenwen}, Kotha, Miguel, and Xu]{Hall2025marin}
D.~Hall, C.~Chou, A.~Garg, N.~Ravi, N.~Liu, H.~Shandilya, A.~Ahmed, P.~Liang, R.~Kuditipudi, {J38}, T.~Lee, R.~Power, K.~Salahi, W.~Held, J.~Wang, {chiheem}, J.~Niklaus, Y.~Mai, {dependabot[bot]}, I.~Zhou, K.~X. Li, S.~Yang, S.~Karamcheti, R.~Williams, C.~Zhou, A.~Ramaswami, {whenwen}, S.~Kotha, G.~Miguel, and C.~Xu.
\newblock marin-community/marin.
\newblock https://github.com/marin-community/marin, nov 14 2025.
\newblock URL \url{https://github.com/marin-community/marin}.

\bibitem[Han et~al.(2024)Han, Rao, Ettinger, Jiang, Lin, Lambert, Choi, and Dziri]{han2024wildguard}
S.~Han, K.~Rao, A.~Ettinger, L.~Jiang, B.~Y. Lin, N.~Lambert, Y.~Choi, and N.~Dziri.
\newblock Wildguard: Open one-stop moderation tools for safety risks, jailbreaks, and refusals of llms.
\newblock \emph{arXiv preprint arXiv:2406.18495}, 2024.

\bibitem[Hartvigsen et~al.(2022)Hartvigsen, Gabriel, Palangi, Sap, Ray, and Kamar]{toxigen}
T.~Hartvigsen, S.~Gabriel, H.~Palangi, M.~Sap, D.~Ray, and E.~Kamar.
\newblock Toxigen: A large-scale machine-generated dataset for adversarial and implicit hate speech detection.
\newblock pages 3309--3326, 01 2022.
\newblock \doi{10.18653/v1/2022.acl-long.234}.

\bibitem[Haupt(2018)]{Haupt2018}
G.~Haupt.
\newblock Hierarchical thinking: a cognitive tool for guiding coherent decision making in design problem solving.
\newblock \emph{International Journal of Technology and Design Education}, 28\penalty0 (1):\penalty0 207--237, 2018.
\newblock ISSN 1573-1804.
\newblock \doi{10.1007/s10798-016-9381-0}.
\newblock URL \url{https://doi.org/10.1007/s10798-016-9381-0}.

\bibitem[Heineman et~al.(2025)Heineman, Hofmann, Magnusson, Gu, Smith, Hajishirzi, Lo, and Dodge]{heineman2025signalnoiseframeworkreducing}
D.~Heineman, V.~Hofmann, I.~Magnusson, Y.~Gu, N.~A. Smith, H.~Hajishirzi, K.~Lo, and J.~Dodge.
\newblock Signal and noise: A framework for reducing uncertainty in language model evaluation, 2025.
\newblock URL \url{https://arxiv.org/abs/2508.13144}.

\bibitem[Hendrycks et~al.(2021{\natexlab{a}})Hendrycks, Basart, Kadavath, Mazeika, Arora, Guo, Burns, Puranik, He, Song, and Steinhardt]{hendrycksapps2021}
D.~Hendrycks, S.~Basart, S.~Kadavath, M.~Mazeika, A.~Arora, E.~Guo, C.~Burns, S.~Puranik, H.~He, D.~Song, and J.~Steinhardt.
\newblock Measuring coding challenge competence with apps.
\newblock \emph{NeurIPS}, 2021{\natexlab{a}}.

\bibitem[Hendrycks et~al.(2021{\natexlab{b}})Hendrycks, Burns, Basart, Zou, Mazeika, Song, and Steinhardt]{hendryckstest2021}
D.~Hendrycks, C.~Burns, S.~Basart, A.~Zou, M.~Mazeika, D.~Song, and J.~Steinhardt.
\newblock Measuring massive multitask language understanding.
\newblock \emph{Proceedings of the International Conference on Learning Representations (ICLR)}, 2021{\natexlab{b}}.

\bibitem[Hendrycks et~al.(2021{\natexlab{c}})Hendrycks, Burns, Kadavath, Arora, Basart, Tang, Song, and Steinhardt]{hendrycksmath2021}
D.~Hendrycks, C.~Burns, S.~Kadavath, A.~Arora, S.~Basart, E.~Tang, D.~Song, and J.~Steinhardt.
\newblock Measuring mathematical problem solving with the math dataset.
\newblock \emph{NeurIPS}, 2021{\natexlab{c}}.

\bibitem[Horgan et~al.(2018)Horgan, Quan, Budden, Barth-Maron, Hessel, Van~Hasselt, and Silver]{horgan2018distributed}
D.~Horgan, J.~Quan, D.~Budden, G.~Barth-Maron, M.~Hessel, H.~Van~Hasselt, and D.~Silver.
\newblock Distributed prioritized experience replay.
\newblock \emph{arXiv preprint arXiv:1803.00933}, 2018.

\bibitem[Hsieh et~al.(2024)Hsieh, Sun, Kriman, Acharya, Rekesh, Jia, Zhang, and Ginsburg]{hsieh2024rulerwhatsrealcontext}
C.-P. Hsieh, S.~Sun, S.~Kriman, S.~Acharya, D.~Rekesh, F.~Jia, Y.~Zhang, and B.~Ginsburg.
\newblock Ruler: What's the real context size of your long-context language models?, 2024.
\newblock URL \url{https://arxiv.org/abs/2404.06654}.

\bibitem[Hsu et~al.(2025)Hsu, Dai, Kothapalli, Song, Tang, Zhu, Shimizu, Sahni, Ning, and Chen]{hsu2025ligerkernelefficienttriton}
P.-L. Hsu, Y.~Dai, V.~Kothapalli, Q.~Song, S.~Tang, S.~Zhu, S.~Shimizu, S.~Sahni, H.~Ning, and Y.~Chen.
\newblock Liger kernel: Efficient triton kernels for llm training, 2025.
\newblock URL \url{https://arxiv.org/abs/2410.10989}.

\bibitem[Hu et~al.(2025)Hu, Zhang, Han, Jiang, Zhang, and Shum]{hu2025open}
J.~Hu, Y.~Zhang, Q.~Han, D.~Jiang, X.~Zhang, and H.-Y. Shum.
\newblock Open-reasoner-zero: An open source approach to scaling up reinforcement learning on the base model.
\newblock \emph{arXiv preprint arXiv:2503.24290}, 2025.

\bibitem[Huang et~al.(2024{\natexlab{a}})Huang, Sun, Wang, Wu, Zhang, Li, Gao, Huang, Lyu, Zhang, et~al.]{huang2024trustllm}
Y.~Huang, L.~Sun, H.~Wang, S.~Wu, Q.~Zhang, Y.~Li, C.~Gao, Y.~Huang, W.~Lyu, Y.~Zhang, et~al.
\newblock Trustllm: Trustworthiness in large language models.
\newblock \emph{arXiv preprint arXiv:2401.05561}, 2024{\natexlab{a}}.

\bibitem[Huang et~al.(2024{\natexlab{b}})Huang, Zhang, Shan, and He]{huang2024compression}
Y.~Huang, J.~Zhang, Z.~Shan, and J.~He.
\newblock Compression represents intelligence linearly.
\newblock \emph{arXiv preprint arXiv:2404.09937}, 2024{\natexlab{b}}.

\bibitem[Jain et~al.(2024)Jain, Han, Gu, Li, Yan, Zhang, Wang, Solar-Lezama, Sen, and Stoica]{jain2024livecodebench}
N.~Jain, K.~Han, A.~Gu, W.-D. Li, F.~Yan, T.~Zhang, S.~Wang, A.~Solar-Lezama, K.~Sen, and I.~Stoica.
\newblock Livecodebench: Holistic and contamination free evaluation of large language models for code.
\newblock \emph{arXiv preprint arXiv:2403.07974}, 2024.

\bibitem[Jiang et~al.(2024)Jiang, Rao, Han, Ettinger, Brahman, Kumar, Mireshghallah, Lu, Sap, Choi, and Dziri]{jiang2024wildteaming}
L.~Jiang, K.~Rao, S.~Han, A.~Ettinger, F.~Brahman, S.~Kumar, N.~Mireshghallah, X.~Lu, M.~Sap, Y.~Choi, and N.~Dziri.
\newblock Wildteaming at scale: From in-the-wild jailbreaks to (adversarially) safer language models.
\newblock \emph{arXiv preprint arXiv:2406.18510}, 2024.
\newblock URL \url{https://arxiv.org/abs/2406.18510}.

\bibitem[Jiang et~al.(2022)Jiang, Yang, Tsirlin, Tang, and Lin]{jiang2022moreparameterfreetextclassification}
Z.~Jiang, M.~Y.~R. Yang, M.~Tsirlin, R.~Tang, and J.~Lin.
\newblock Less is more: Parameter-free text classification with gzip, 2022.
\newblock URL \url{https://arxiv.org/abs/2212.09410}.

\bibitem[Jin et~al.(2021)Jin, Pan, Oufattole, Weng, Fang, and Szolovits]{jin2021disease}
D.~Jin, E.~Pan, N.~Oufattole, W.-H. Weng, H.~Fang, and P.~Szolovits.
\newblock What disease does this patient have? a large-scale open domain question answering dataset from medical exams.
\newblock \emph{Applied Sciences}, 11\penalty0 (14):\penalty0 6421, 2021.

\bibitem[Joyce(2009)]{Joyce2009}
J.~M. Joyce.
\newblock Causal reasoning and backtracking.
\newblock \emph{Philosophical Studies}, 147\penalty0 (1):\penalty0 139--154, 2009.
\newblock \doi{10.1007/s11098-009-9454-y}.

\bibitem[Kaiyom et~al.(2024)Kaiyom, Ahmed, Mai, Klyman, Bommasani, and Liang]{kaiyom2024helmsafety}
F.~Kaiyom, A.~Ahmed, Y.~Mai, K.~Klyman, R.~Bommasani, and P.~Liang.
\newblock {HELM} safety: Towards standardized safety evaluations of language models, 8~Nov. 2024.
\newblock URL \url{https://crfm.stanford.edu/2024/11/08/helm-safety.html}.

\bibitem[Kargupta et~al.(2025)Kargupta, Li, Wang, Lee, Chen, Ahia, Light, Griffiths, Kleiman-Weiner, Han, Celikyilmaz, and Tsvetkov]{kargupta2025cognitive}
P.~Kargupta, S.~S. Li, H.~Wang, J.~Lee, S.~Chen, O.~Ahia, D.~Light, T.~L. Griffiths, M.~Kleiman-Weiner, J.~Han, A.~Celikyilmaz, and Y.~Tsvetkov.
\newblock Cognitive foundations for reasoning and their manifestation in llms.
\newblock \emph{arXiv}, 2025.

\bibitem[Kavukcuo\u{g}lu and DeepMind(2025)]{google-gemini2.5}
K.~Kavukcuo\u{g}lu and G.~DeepMind.
\newblock Gemini 2.5: Our most intelligent ai model.
\newblock \url{https://blog.google/technology/google-deepmind/gemini-model-thinking-updates-march-2025/}, Mar. 2025.
\newblock Accessed: 2025-10-07.

\bibitem[Kim et~al.(2025)Kim, Goyal, Zhang, Xiong, Hou, Kambadur, Mahajan, Hajishirzi, and Tan]{kim2025systematic}
J.~Kim, A.~Goyal, A.~Zhang, B.~Xiong, R.~Hou, M.~Kambadur, D.~Mahajan, H.~Hajishirzi, and L.~Tan.
\newblock A systematic examination of preference learning through the lens of instruction-following.
\newblock In \emph{Proceedings of the 2025 Conference of the Nations of the Americas Chapter of the Association for Computational Linguistics: Human Language Technologies (Volume 1: Long Papers)}, pages 11062--11082, 2025.

\bibitem[Kim et~al.(2023)Kim, Bae, Shin, Kang, Kwak, Yoo, and Seo]{kim2023aligning}
S.~Kim, S.~Bae, J.~Shin, S.~Kang, D.~Kwak, K.~Yoo, and M.~Seo.
\newblock Aligning large language models through synthetic feedback.
\newblock In H.~Bouamor, J.~Pino, and K.~Bali, editors, \emph{Proceedings of the 2023 Conference on Empirical Methods in Natural Language Processing}, pages 13677--13700, Singapore, Dec. 2023. Association for Computational Linguistics.
\newblock \doi{10.18653/v1/2023.emnlp-main.844}.
\newblock URL \url{https://aclanthology.org/2023.emnlp-main.844/}.

\bibitem[{Kimi Team} et~al.(2025){Kimi Team}, Bai, Bao, Chen, Chen, Chen, Chen, Chen, Chen, Chen, Chen, Cui, Ding, Dong, Du, Du, Du, Du, Fan, Feng, Fu, Gao, Gao, Gao, Gao, Gu, Guan, Guo, Guo, Hu, Hao, He, He, He, Hong, Hu, Hu, Huang, Huang, Huang, Jiang, Jiang, Jin, Kang, Lai, Li, Li, Li, Li, Li, Li, Li, Li, Li, Lin, Lin, Lin, Liu, Liu, Liu, Liu, Liu, Liu, Liu, Liu, Liu, Liu, Liu, Liu, Liu, Liu, Liu, Lu, Lu, Ma, Ma, Ma, Mao, Mei, Men, Miao, Pan, Peng, Qin, Qu, Shang, Shi, Shi, Song, Su, Su, Sun, Sung, Tang, Tao, Teng, Wang, Wang, Wang, Wang, Wang, Wang, Wang, Wang, Wang, Wang, Wang, Wang, Wang, Wang, Wang, Wang, Wang, Wei, Wei, Wu, Wu, Wu, Xiao, Xie, Xiong, Xu, Xu, Xu, Xu, Xu, Xu, Xu, Xu, Xu, Xu, Yan, Yan, Yang, Yang, Yang, Yang, Yang, Yao, Yao, Ye, Ye, Yin, Yu, Yuan, Yuan, Yuan, Zhan, Zhang, Zhang, Zhang, Zhang, Zhang, Zhang, Zhang, Zhang, Zhang, Zhang, Zhang, Zhao, Zhao, Zheng, Zheng, Zhou, Zhou, Zhou, Zhu, Zhuang, and Zu]{kimiK2}
{Kimi Team}, Y.~Bai, Y.~Bao, G.~Chen, J.~Chen, N.~Chen, R.~Chen, Y.~Chen, Y.~Chen, Y.~Chen, Z.~Chen, J.~Cui, H.~Ding, M.~Dong, A.~Du, C.~Du, D.~Du, Y.~Du, Y.~Fan, Y.~Feng, K.~Fu, B.~Gao, H.~Gao, P.~Gao, T.~Gao, X.~Gu, L.~Guan, H.~Guo, J.~Guo, H.~Hu, X.~Hao, T.~He, W.~He, W.~He, C.~Hong, Y.~Hu, Z.~Hu, W.~Huang, Z.~Huang, Z.~Huang, T.~Jiang, Z.~Jiang, X.~Jin, Y.~Kang, G.~Lai, C.~Li, F.~Li, H.~Li, M.~Li, W.~Li, Y.~Li, Y.~Li, Z.~Li, Z.~Li, H.~Lin, X.~Lin, Z.~Lin, C.~Liu, C.~Liu, H.~Liu, J.~Liu, J.~Liu, L.~Liu, S.~Liu, T.~Y. Liu, T.~Liu, W.~Liu, Y.~Liu, Y.~Liu, Y.~Liu, Y.~Liu, Z.~Liu, E.~Lu, L.~Lu, S.~Ma, X.~Ma, Y.~Ma, S.~Mao, J.~Mei, X.~Men, Y.~Miao, S.~Pan, Y.~Peng, R.~Qin, B.~Qu, Z.~Shang, L.~Shi, S.~Shi, F.~Song, J.~Su, Z.~Su, X.~Sun, F.~Sung, H.~Tang, J.~Tao, Q.~Teng, C.~Wang, D.~Wang, F.~Wang, H.~Wang, J.~Wang, J.~Wang, J.~Wang, S.~Wang, S.~Wang, Y.~Wang, Y.~Wang, Y.~Wang, Y.~Wang, Y.~Wang, Z.~Wang, Z.~Wang, Z.~Wang, C.~Wei, Q.~Wei, W.~Wu, X.~Wu, Y.~Wu, C.~Xiao, X.~Xie, W.~Xiong, B.~Xu, J.~Xu, J.~Xu, L.~H.
  Xu, L.~Xu, S.~Xu, W.~Xu, X.~Xu, Y.~Xu, Z.~Xu, J.~Yan, Y.~Yan, X.~Yang, Y.~Yang, Z.~Yang, Z.~Yang, Z.~Yang, H.~Yao, X.~Yao, W.~Ye, Z.~Ye, B.~Yin, L.~Yu, E.~Yuan, H.~Yuan, M.~Yuan, H.~Zhan, D.~Zhang, H.~Zhang, W.~Zhang, X.~Zhang, Y.~Zhang, Y.~Zhang, Y.~Zhang, Y.~Zhang, Y.~Zhang, Y.~Zhang, Z.~Zhang, H.~Zhao, Y.~Zhao, H.~Zheng, S.~Zheng, J.~Zhou, X.~Zhou, Z.~Zhou, Z.~Zhu, W.~Zhuang, and X.~Zu.
\newblock Kimi k2: Open agentic intelligence, 2025.
\newblock URL \url{https://arxiv.org/abs/2507.20534}.

\bibitem[K{\"o}pf et~al.(2024)K{\"o}pf, Kilcher, von R{\"u}tte, Anagnostidis, Tam, Stevens, Barhoum, Nguyen, Stanley, Nagyfi, et~al.]{kopf2024openassistant}
A.~K{\"o}pf, Y.~Kilcher, D.~von R{\"u}tte, S.~Anagnostidis, Z.~R. Tam, K.~Stevens, A.~Barhoum, D.~Nguyen, O.~Stanley, R.~Nagyfi, et~al.
\newblock Openassistant conversations-democratizing large language model alignment.
\newblock \emph{Advances in Neural Information Processing Systems}, 36, 2024.

\bibitem[Kudugunta et~al.(2023)Kudugunta, Caswell, Zhang, Garcia, Choquette-Choo, Lee, Xin, Kusupati, Stella, Bapna, and Firat]{kudugunta2023madlad400multilingualdocumentlevellarge}
S.~Kudugunta, I.~Caswell, B.~Zhang, X.~Garcia, C.~A. Choquette-Choo, K.~Lee, D.~Xin, A.~Kusupati, R.~Stella, A.~Bapna, and O.~Firat.
\newblock Madlad-400: A multilingual and document-level large audited dataset, 2023.
\newblock URL \url{https://arxiv.org/abs/2309.04662}.

\bibitem[Kwiatkowski et~al.(2019)Kwiatkowski, Palomaki, Redfield, Collins, Parikh, Alberti, Epstein, Polosukhin, Devlin, Lee, Toutanova, Jones, Kelcey, Chang, Dai, Uszkoreit, Le, and Petrov]{kwiatkowski-etal-2019-natural}
T.~Kwiatkowski, J.~Palomaki, O.~Redfield, M.~Collins, A.~Parikh, C.~Alberti, D.~Epstein, I.~Polosukhin, J.~Devlin, K.~Lee, K.~Toutanova, L.~Jones, M.~Kelcey, M.-W. Chang, A.~M. Dai, J.~Uszkoreit, Q.~Le, and S.~Petrov.
\newblock Natural questions: A benchmark for question answering research.
\newblock \emph{Transactions of the Association for Computational Linguistics}, 7:\penalty0 452--466, 2019.
\newblock \doi{10.1162/tacl_a_00276}.
\newblock URL \url{https://aclanthology.org/Q19-1026}.

\bibitem[Kwon et~al.(2023)Kwon, Li, Zhuang, Sheng, Zheng, Yu, Gonzalez, Zhang, and Stoica]{vllm}
W.~Kwon, Z.~Li, S.~Zhuang, Y.~Sheng, L.~Zheng, C.~H. Yu, J.~E. Gonzalez, H.~Zhang, and I.~Stoica.
\newblock Efficient memory management for large language model serving with pagedattention.
\newblock In \emph{Proceedings of the ACM SIGOPS 29th Symposium on Operating Systems Principles}, 2023.

\bibitem[Lai et~al.(2022)Lai, Li, Wang, Zhang, Zhong, Zettlemoyer, Yih, Fried, Wang, and Yu]{Lai2022DS1000}
Y.~Lai, C.~Li, Y.~Wang, T.~Zhang, R.~Zhong, L.~Zettlemoyer, W.-T. Yih, D.~Fried, S.~Wang, and T.~Yu.
\newblock Ds-1000: A natural and reliable benchmark for data science code generation.
\newblock \emph{ArXiv}, abs/2211.11501, 2022.

\bibitem[Lambert(2025)]{rlhf2024}
N.~Lambert.
\newblock \emph{Reinforcement Learning from Human Feedback}.
\newblock Online, 2025.
\newblock URL \url{https://rlhfbook.com}.

\bibitem[Lambert et~al.(2023)Lambert, Gilbert, and Zick]{lambert2023entangled}
N.~Lambert, T.~K. Gilbert, and T.~Zick.
\newblock Entangled preferences: The history and risks of reinforcement learning and human feedback.
\newblock \emph{arXiv preprint arXiv:2310.13595}, 2023.

\bibitem[Lambert et~al.(2024)Lambert, Morrison, Pyatkin, Huang, Ivison, Brahman, Miranda, Liu, Dziri, Lyu, Gu, Malik, Graf, Hwang, Yang, Bras, Tafjord, Wilhelm, Soldaini, Smith, Wang, Dasigi, and Hajishirzi]{lambert2024tulu3}
N.~Lambert, J.~D. Morrison, V.~Pyatkin, S.~Huang, H.~Ivison, F.~Brahman, L.~J.~V. Miranda, A.~Liu, N.~Dziri, S.~Lyu, Y.~Gu, S.~Malik, V.~Graf, J.~D. Hwang, J.~Yang, R.~L. Bras, O.~Tafjord, C.~Wilhelm, L.~Soldaini, N.~A. Smith, Y.~Wang, P.~Dasigi, and H.~Hajishirzi.
\newblock Tulu 3: Pushing frontiers in open language model post-training.
\newblock 2024.
\newblock URL \url{https://api.semanticscholar.org/CorpusID:274192505}.

\bibitem[Laurent et~al.(2024)Laurent, Janizek, Ruzo, Hinks, Hammerling, Narayanan, Ponnapati, White, and Rodriques]{laurent2024lab}
J.~M. Laurent, J.~D. Janizek, M.~Ruzo, M.~M. Hinks, M.~J. Hammerling, S.~Narayanan, M.~Ponnapati, A.~D. White, and S.~G. Rodriques.
\newblock Lab-bench: Measuring capabilities of language models for biology research.
\newblock \emph{arXiv preprint arXiv:2407.10362}, 2024.

\bibitem[Lee et~al.(2022)Lee, Ippolito, Nystrom, Zhang, Eck, Callison-Burch, and Carlini]{lee2022deduplicatingtrainingdatamakes}
K.~Lee, D.~Ippolito, A.~Nystrom, C.~Zhang, D.~Eck, C.~Callison-Burch, and N.~Carlini.
\newblock Deduplicating training data makes language models better, 2022.
\newblock URL \url{https://arxiv.org/abs/2107.06499}.

\bibitem[Lewkowycz et~al.(2022)Lewkowycz, Andreassen, Dohan, Dyer, Michalewski, Ramasesh, Slone, Anil, Schlag, Gutman-Solo, et~al.]{lewkowycz2022solving}
A.~Lewkowycz, A.~Andreassen, D.~Dohan, E.~Dyer, H.~Michalewski, V.~Ramasesh, A.~Slone, C.~Anil, I.~Schlag, T.~Gutman-Solo, et~al.
\newblock Solving quantitative reasoning problems with language models.
\newblock \emph{Advances in neural information processing systems}, 35:\penalty0 3843--3857, 2022.

\bibitem[Li et~al.(2024{\natexlab{a}})Li, Fang, Smyrnis, Ivgi, Jordan, Gadre, Bansal, Guha, Keh, Arora, Garg, Xin, Muennighoff, Heckel, Mercat, Chen, Gururangan, Wortsman, Albalak, Bitton, Nezhurina, Abbas, Hsieh, Ghosh, Gardner, Kilian, Zhang, Shao, Pratt, Sanyal, Ilharco, Daras, Marathe, Gokaslan, Zhang, Chandu, Nguyen, Vasiljevic, Kakade, Song, Sanghavi, Faghri, Oh, Zettlemoyer, Lo, El-Nouby, Pouransari, Toshev, Wang, Groeneveld, Soldaini, Koh, Jitsev, Kollar, Dimakis, Carmon, Dave, Schmidt, and Shankar]{dclm}
J.~Li, A.~Fang, G.~Smyrnis, M.~Ivgi, M.~Jordan, S.~Gadre, H.~Bansal, E.~Guha, S.~Keh, K.~Arora, S.~Garg, R.~Xin, N.~Muennighoff, R.~Heckel, J.~Mercat, M.~Chen, S.~Gururangan, M.~Wortsman, A.~Albalak, Y.~Bitton, M.~Nezhurina, A.~Abbas, C.-Y. Hsieh, D.~Ghosh, J.~Gardner, M.~Kilian, H.~Zhang, R.~Shao, S.~Pratt, S.~Sanyal, G.~Ilharco, G.~Daras, K.~Marathe, A.~Gokaslan, J.~Zhang, K.~Chandu, T.~Nguyen, I.~Vasiljevic, S.~Kakade, S.~Song, S.~Sanghavi, F.~Faghri, S.~Oh, L.~Zettlemoyer, K.~Lo, A.~El-Nouby, H.~Pouransari, A.~Toshev, S.~Wang, D.~Groeneveld, L.~Soldaini, P.~W. Koh, J.~Jitsev, T.~Kollar, A.~G. Dimakis, Y.~Carmon, A.~Dave, L.~Schmidt, and V.~Shankar.
\newblock Datacomp-lm: In search of the next generation of training sets for language models, 2024{\natexlab{a}}.
\newblock URL \url{https://arxiv.org/abs/2406.11794}.

\bibitem[Li et~al.(2024{\natexlab{b}})Li, Pan, Gopal, Yue, Berrios, Gatti, Li, Dombrowski, Goel, Mukobi, Helm-Burger, Lababidi, Justen, Liu, Chen, Barrass, Zhang, Zhu, Tamirisa, Bharathi, Herbert-Voss, Breuer, Zou, Mazeika, Wang, Oswal, Lin, Hunt, Tienken-Harder, Shih, Talley, Guan, Steneker, Campbell, Jokubaitis, Basart, Fitz, Kumaraguru, Karmakar, Tupakula, Varadharajan, Shoshitaishvili, Ba, Esvelt, Wang, and Hendrycks]{wmdp}
N.~Li, A.~Pan, A.~Gopal, S.~Yue, D.~Berrios, A.~Gatti, J.~D. Li, A.-K. Dombrowski, S.~Goel, G.~Mukobi, N.~Helm-Burger, R.~Lababidi, L.~Justen, A.~B. Liu, M.~Chen, I.~Barrass, O.~Zhang, X.~Zhu, R.~Tamirisa, B.~Bharathi, A.~Herbert-Voss, C.~B. Breuer, A.~Zou, M.~Mazeika, Z.~Wang, P.~Oswal, W.~Lin, A.~A. Hunt, J.~Tienken-Harder, K.~Y. Shih, K.~Talley, J.~Guan, I.~Steneker, D.~Campbell, B.~Jokubaitis, S.~Basart, S.~Fitz, P.~Kumaraguru, K.~K. Karmakar, U.~Tupakula, V.~Varadharajan, Y.~Shoshitaishvili, J.~Ba, K.~M. Esvelt, A.~Wang, and D.~Hendrycks.
\newblock The {WMDP} benchmark: Measuring and reducing malicious use with unlearning.
\newblock In R.~Salakhutdinov, Z.~Kolter, K.~Heller, A.~Weller, N.~Oliver, J.~Scarlett, and F.~Berkenkamp, editors, \emph{Proceedings of the 41st International Conference on Machine Learning}, volume 235 of \emph{Proceedings of Machine Learning Research}, pages 28525--28550. PMLR, 21--27 Jul 2024{\natexlab{b}}.
\newblock URL \url{https://proceedings.mlr.press/v235/li24bc.html}.

\bibitem[Li et~al.(2023{\natexlab{a}})Li, Fu, Zhang, Huang, Sun, Lyu, Liu, Jin, and Li]{tacoli}
R.~Li, J.~Fu, B.-W. Zhang, T.~Huang, Z.~Sun, C.~Lyu, G.~Liu, Z.~Jin, and G.~Li.
\newblock Taco: Topics in algorithmic code generation dataset.
\newblock \emph{arXiv preprint arXiv:2312.14852}, 2023{\natexlab{a}}.

\bibitem[Li et~al.(2024{\natexlab{c}})Li, Chiang, Frick, Dunlap, Wu, Zhu, Gonzalez, and Stoica]{li2024crowdsourced}
T.~Li, W.-L. Chiang, E.~Frick, L.~Dunlap, T.~Wu, B.~Zhu, J.~E. Gonzalez, and I.~Stoica.
\newblock From crowdsourced data to high-quality benchmarks: Arena-hard and benchbuilder pipeline.
\newblock \emph{arXiv preprint arXiv:2406.11939}, 2024{\natexlab{c}}.

\bibitem[Li et~al.(2023{\natexlab{b}})Li, Zhang, Dubois, Taori, Gulrajani, Guestrin, Liang, and Hashimoto]{alpaca_eval}
X.~Li, T.~Zhang, Y.~Dubois, R.~Taori, I.~Gulrajani, C.~Guestrin, P.~Liang, and T.~B. Hashimoto.
\newblock Alpacaeval: An automatic evaluator of instruction-following models.
\newblock \url{https://github.com/tatsu-lab/alpaca_eval}, 5 2023{\natexlab{b}}.

\bibitem[Li et~al.(2025)Li, Ma, Yan, Zhang, Liu, Lu, Xu, Chen, Wang, Zhan, Ma, Lai, Liu, Luo, Bin, Ren, Han, Hao, Yi, Liu, Ma, Jia, Zhou, Qiao, Xiang, and Wu]{modelmerginginpretraining}
Y.~Li, Y.~Ma, S.~Yan, C.~Zhang, J.~Liu, J.~Lu, Z.~Xu, M.~Chen, M.~Wang, S.~Zhan, J.~Ma, X.~Lai, D.~Liu, Y.~Luo, X.~Bin, H.~Ren, M.~Han, W.~Hao, B.~Yi, L.~Liu, B.~Ma, X.~Jia, X.~Zhou, S.~Qiao, L.~Xiang, and Y.~Wu.
\newblock Model merging in pre-training of large language models.
\newblock \emph{ArXiv}, abs/2505.12082, 2025.
\newblock URL \url{https://api.semanticscholar.org/CorpusID:278739754}.

\bibitem[Lightman et~al.(2023)Lightman, Kosaraju, Burda, Edwards, Baker, Lee, Leike, Schulman, Sutskever, and Cobbe]{lightman2023lets}
H.~Lightman, V.~Kosaraju, Y.~Burda, H.~Edwards, B.~Baker, T.~Lee, J.~Leike, J.~Schulman, I.~Sutskever, and K.~Cobbe.
\newblock Let's verify step by step.
\newblock \emph{arXiv preprint arXiv:2305.20050}, 2023.

\bibitem[Lin et~al.(2025)Lin, Bras, Richardson, Sabharwal, Poovendran, Clark, and Choi]{lin2025zebralogic}
B.~Y. Lin, R.~L. Bras, K.~Richardson, A.~Sabharwal, R.~Poovendran, P.~Clark, and Y.~Choi.
\newblock Zebralogic: On the scaling limits of llms for logical reasoning.
\newblock \emph{arXiv preprint arXiv:2502.01100}, 2025.

\bibitem[Liu et~al.(2023{\natexlab{a}})Liu, Bubeck, Eldan, Kulkarni, Li, Nguyen, Ward, and Zhang]{liu2023tinygsmachieving80gsm8k}
B.~Liu, S.~Bubeck, R.~Eldan, J.~Kulkarni, Y.~Li, A.~Nguyen, R.~Ward, and Y.~Zhang.
\newblock Tinygsm: achieving >80
\newblock URL \url{https://arxiv.org/abs/2312.09241}.

\bibitem[Liu et~al.(2023{\natexlab{b}})Liu, Xia, Wang, and Zhang]{evalplus}
J.~Liu, C.~S. Xia, Y.~Wang, and L.~Zhang.
\newblock Is your code generated by chat{GPT} really correct? rigorous evaluation of large language models for code generation.
\newblock In \emph{Thirty-seventh Conference on Neural Information Processing Systems}, 2023{\natexlab{b}}.
\newblock URL \url{https://openreview.net/forum?id=1qvx610Cu7}.

\bibitem[Liu et~al.(2025{\natexlab{a}})Liu, Diao, Lu, Hu, Dong, Choi, Kautz, and Dong]{liu2025prorl}
M.~Liu, S.~Diao, X.~Lu, J.~Hu, X.~Dong, Y.~Choi, J.~Kautz, and Y.~Dong.
\newblock Prorl: Prolonged reinforcement learning expands reasoning boundaries in large language models.
\newblock \emph{arXiv preprint}, 2025{\natexlab{a}}.
\newblock URL \url{https://arxiv.org/abs/2505.24864}.

\bibitem[Liu et~al.(2024{\natexlab{a}})Liu, Zheng, Muennighoff, Zeng, Dou, Pang, Jiang, and Lin]{liu2024regmix}
Q.~Liu, X.~Zheng, N.~Muennighoff, G.~Zeng, L.~Dou, T.~Pang, J.~Jiang, and M.~Lin.
\newblock Regmix: Data mixture as regression for language model pre-training.
\newblock \emph{arXiv preprint arXiv:2407.01492}, 2024{\natexlab{a}}.

\bibitem[Liu et~al.(2024{\natexlab{b}})Liu, Huang, Zeng, Hao, Yu, Li, Wang, Gan, Liu, Yu, Wang, Wang, Ning, Hou, Wang, Wu, Wang, Liu, Wang, Tang, Tu, Shang, Jiang, Tang, Lian, Liu, and Chen]{Liu2024ToolACEWT}
W.~Liu, X.~Huang, X.~Zeng, X.~Hao, S.~Yu, D.~Li, S.~Wang, W.~Gan, Z.~Liu, Y.~Yu, Z.~Wang, Y.~Wang, W.~Ning, Y.~Hou, B.~Wang, C.~Wu, X.~Wang, Y.~Liu, Y.~Wang, D.~Tang, D.~Tu, L.~Shang, X.~Jiang, R.~Tang, D.~Lian, Q.~Liu, and E.~Chen.
\newblock Toolace: Winning the points of llm function calling.
\newblock \emph{ArXiv}, abs/2409.00920, 2024{\natexlab{b}}.
\newblock URL \url{https://api.semanticscholar.org/CorpusID:272368347}.

\bibitem[Liu et~al.(2023{\natexlab{c}})Liu, Qiao, Neiswanger, Wang, Tan, Tao, Li, Wang, Sun, Pangarkar, et~al.]{liu2023llm360}
Z.~Liu, A.~Qiao, W.~Neiswanger, H.~Wang, B.~Tan, T.~Tao, J.~Li, Y.~Wang, S.~Sun, O.~Pangarkar, et~al.
\newblock Llm360: Towards fully transparent open-source llms.
\newblock \emph{arXiv preprint arXiv:2312.06550}, 2023{\natexlab{c}}.

\bibitem[Liu et~al.(2024{\natexlab{c}})Liu, Hoang, Zhang, Zhu, Lan, Kokane, Tan, Yao, Liu, Feng, Murthy, Yang, Savarese, Niebles, Wang, Heinecke, and Xiong]{Liu2024APIGenAP}
Z.~Liu, T.~Hoang, J.~Zhang, M.~Zhu, T.~Lan, S.~Kokane, J.~Tan, W.~Yao, Z.~Liu, Y.~Feng, R.~Murthy, L.~Yang, S.~Savarese, J.~C. Niebles, H.~Wang, S.~Heinecke, and C.~Xiong.
\newblock Apigen: Automated pipeline for generating verifiable and diverse function-calling datasets.
\newblock \emph{ArXiv}, abs/2406.18518, 2024{\natexlab{c}}.
\newblock URL \url{https://api.semanticscholar.org/CorpusID:270738094}.

\bibitem[Liu et~al.(2025{\natexlab{b}})Liu, Chen, Li, Qi, Pang, Du, Lee, and Lin]{liu2025understanding}
Z.~Liu, C.~Chen, W.~Li, P.~Qi, T.~Pang, C.~Du, W.~S. Lee, and M.~Lin.
\newblock Understanding r1-zero-like training: A critical perspective.
\newblock In \emph{Conference on Language Modeling (COLM)}, 2025{\natexlab{b}}.

\bibitem[Longpre et~al.(2023)Longpre, Hou, Vu, Webson, Chung, Tay, Zhou, Le, Zoph, Wei, et~al.]{longpre2023flan}
S.~Longpre, L.~Hou, T.~Vu, A.~Webson, H.~W. Chung, Y.~Tay, D.~Zhou, Q.~V. Le, B.~Zoph, J.~Wei, et~al.
\newblock The flan collection: Designing data and methods for effective instruction tuning.
\newblock \emph{arXiv preprint arXiv:2301.13688}, 2023.

\bibitem[Lozhkov et~al.(2024)Lozhkov, Li, Allal, Cassano, Lamy-Poirier, Tazi, Tang, Pykhtar, Liu, Wei, et~al.]{lozhkov2024starcoder}
A.~Lozhkov, R.~Li, L.~B. Allal, F.~Cassano, J.~Lamy-Poirier, N.~Tazi, A.~Tang, D.~Pykhtar, J.~Liu, Y.~Wei, et~al.
\newblock Starcoder 2 and the stack v2: The next generation.
\newblock \emph{arXiv preprint arXiv:2402.19173}, 2024.

\bibitem[Luo et~al.(2025{\natexlab{a}})Luo, Tan, Wong, Shi, Tang, Roongta, Cai, Luo, Li, Popa, and Stoica]{deepscaler2025}
M.~Luo, S.~Tan, J.~Wong, X.~Shi, W.~Y. Tang, M.~Roongta, C.~Cai, J.~Luo, L.~E. Li, R.~A. Popa, and I.~Stoica.
\newblock Deepscaler: Surpassing o1-preview with a 1.5b model by scaling rl.
\newblock \url{https://pretty-radio-b75.notion.site/DeepScaleR-Surpassing-O1-Preview-with-a-1-5B-Model-by-Scaling-RL-19681902c1468005bed8ca303013a4e2}, 2025{\natexlab{a}}.
\newblock Notion Blog.

\bibitem[Luo et~al.(2025{\natexlab{b}})Luo, Tan, Wong, Shi, Tang, Roongta, Cai, Luo, Zhang, Li, et~al.]{luo2025deepscaler}
M.~Luo, S.~Tan, J.~Wong, X.~Shi, W.~Y. Tang, M.~Roongta, C.~Cai, J.~Luo, T.~Zhang, L.~E. Li, et~al.
\newblock Deepscaler: Surpassing o1-preview with a 1.5 b model by scaling rl.
\newblock \emph{Notion Blog}, 2025{\natexlab{b}}.

\bibitem[Luo et~al.(2023)Luo, Xu, Zhao, Sun, Geng, Hu, Tao, Ma, Lin, and Jiang]{luo2023wizardcoder}
Z.~Luo, C.~Xu, P.~Zhao, Q.~Sun, X.~Geng, W.~Hu, C.~Tao, J.~Ma, Q.~Lin, and D.~Jiang.
\newblock Wizardcoder: Empowering code large language models with evol-instruct, 2023.

\bibitem[Magar and Schwartz(2022)]{Magar2022DataCF}
I.~Magar and R.~Schwartz.
\newblock Data contamination: From memorization to exploitation.
\newblock \emph{ArXiv}, abs/2203.08242, 2022.
\newblock URL \url{https://api.semanticscholar.org/CorpusID:247475929}.

\bibitem[Magnusson et~al.(2024)Magnusson, Bhagia, Hofmann, Soldaini, Jha, Tafjord, Schwenk, Walsh, Elazar, Lo, Groeneveld, Beltagy, Hajishirzi, Smith, Richardson, and Dodge]{magnusson2024palomabenchmarkevaluatinglanguage}
I.~Magnusson, A.~Bhagia, V.~Hofmann, L.~Soldaini, A.~H. Jha, O.~Tafjord, D.~Schwenk, E.~P. Walsh, Y.~Elazar, K.~Lo, D.~Groeneveld, I.~Beltagy, H.~Hajishirzi, N.~A. Smith, K.~Richardson, and J.~Dodge.
\newblock Paloma: A benchmark for evaluating language model fit, 2024.
\newblock URL \url{https://arxiv.org/abs/2312.10523}.

\bibitem[Magnusson et~al.(2025)Magnusson, Tai, Bogin, Heineman, Hwang, Soldaini, Bhagia, Liu, Groeneveld, Tafjord, Smith, Koh, and Dodge]{magnusson2025datadecidepredictbestpretraining}
I.~Magnusson, N.~Tai, B.~Bogin, D.~Heineman, J.~D. Hwang, L.~Soldaini, A.~Bhagia, J.~Liu, D.~Groeneveld, O.~Tafjord, N.~A. Smith, P.~W. Koh, and J.~Dodge.
\newblock Datadecide: How to predict best pretraining data with small experiments, 2025.
\newblock URL \url{https://arxiv.org/abs/2504.11393}.

\bibitem[Mallen et~al.(2022)Mallen, Asai, Zhong, Das, Hajishirzi, and Khashabi]{mallen2023llm_memorization}
A.~Mallen, A.~Asai, V.~Zhong, R.~Das, H.~Hajishirzi, and D.~Khashabi.
\newblock When not to trust language models: Investigating effectiveness and limitations of parametric and non-parametric memories.
\newblock \emph{arXiv preprint}, 2022.

\bibitem[Marjanović et~al.(2025)Marjanović, Patel, Adlakha, Aghajohari, BehnamGhader, Bhatia, Khandelwal, Kraft, Krojer, Lù, Meade, Shin, Kazemnejad, Kamath, Mosbach, Stańczak, and Reddy]{marjanović2025deepseekr1thoughtologyletsthink}
S.~V. Marjanović, A.~Patel, V.~Adlakha, M.~Aghajohari, P.~BehnamGhader, M.~Bhatia, A.~Khandelwal, A.~Kraft, B.~Krojer, X.~H. Lù, N.~Meade, D.~Shin, A.~Kazemnejad, G.~Kamath, M.~Mosbach, K.~Stańczak, and S.~Reddy.
\newblock Deepseek-r1 thoughtology: Let's think about llm reasoning, 2025.
\newblock URL \url{https://arxiv.org/abs/2504.07128}.

\bibitem[Markovits et~al.(2015)Markovits, Thompson, and Brisson]{Markovits2015}
H.~Markovits, V.~A. Thompson, and J.~Brisson.
\newblock Metacognition and abstract reasoning.
\newblock \emph{Memory \& Cognition}, 43\penalty0 (4):\penalty0 681--693, 2015.
\newblock ISSN 1532-5946.
\newblock \doi{10.3758/s13421-014-0488-9}.
\newblock URL \url{https://doi.org/10.3758/s13421-014-0488-9}.

\bibitem[Matton et~al.(2024)Matton, Sherborne, Aumiller, Tommasone, Alizadeh, He, Ma, Voisin, Gilsenan-McMahon, and Gall{\'e}]{matton-etal-2024-leakage}
A.~Matton, T.~Sherborne, D.~Aumiller, E.~Tommasone, M.~Alizadeh, J.~He, R.~Ma, M.~Voisin, E.~Gilsenan-McMahon, and M.~Gall{\'e}.
\newblock On leakage of code generation evaluation datasets.
\newblock In Y.~Al-Onaizan, M.~Bansal, and Y.-N. Chen, editors, \emph{Findings of the Association for Computational Linguistics: EMNLP 2024}, pages 13215--13223, Miami, Florida, USA, Nov. 2024. Association for Computational Linguistics.
\newblock \doi{10.18653/v1/2024.findings-emnlp.772}.
\newblock URL \url{https://aclanthology.org/2024.findings-emnlp.772/}.

\bibitem[Mazeika et~al.(2024)Mazeika, Phan, Yin, Zou, Wang, Mu, Sakhaee, Li, Basart, Li, et~al.]{mazeika2024harmbench}
M.~Mazeika, L.~Phan, X.~Yin, A.~Zou, Z.~Wang, N.~Mu, E.~Sakhaee, N.~Li, S.~Basart, B.~Li, et~al.
\newblock Harmbench: A standardized evaluation framework for automated red teaming and robust refusal.
\newblock \emph{arXiv preprint arXiv:2402.04249}, 2024.

\bibitem[Meyer and Corneil(2025)]{nvidia/Nemotron-Personas-USA}
Y.~Meyer and D.~Corneil.
\newblock {Nemotron-Personas-USA}: Synthetic personas aligned to real-world distributions, June 2025.
\newblock URL \url{https://huggingface.co/datasets/nvidia/Nemotron-Personas-USA}.

\bibitem[Mindermann et~al.(2022)Mindermann, Brauner, Razzak, Sharma, Kirsch, Xu, H{\"o}ltgen, Gomez, Morisot, Farquhar, et~al.]{mindermann2022prioritized}
S.~Mindermann, J.~M. Brauner, M.~T. Razzak, M.~Sharma, A.~Kirsch, W.~Xu, B.~H{\"o}ltgen, A.~N. Gomez, A.~Morisot, S.~Farquhar, et~al.
\newblock Prioritized training on points that are learnable, worth learning, and not yet learnt.
\newblock In \emph{International Conference on Machine Learning}, pages 15630--15649. PMLR, 2022.

\bibitem[Miroyan et~al.(2025)Miroyan, Wu, King, Li, Pan, Hu, Chiang, Angelopoulos, Darrell, Norouzi, and Gonzalez]{Miroyan2025SearchAA}
M.~Miroyan, T.-H. Wu, L.~King, T.~Li, J.~Pan, X.~Hu, W.-L. Chiang, A.~N. Angelopoulos, T.~Darrell, N.~Norouzi, and J.~Gonzalez.
\newblock Search arena: Analyzing search-augmented llms.
\newblock \emph{ArXiv}, abs/2506.05334, 2025.
\newblock URL \url{https://api.semanticscholar.org/CorpusID:279243096}.

\bibitem[Mirzadeh et~al.(2024)Mirzadeh, Alizadeh, Shahrokhi, Tuzel, Bengio, and Farajtabar]{gsm-symbolic}
I.~Mirzadeh, K.~Alizadeh, H.~Shahrokhi, O.~Tuzel, S.~Bengio, and M.~Farajtabar.
\newblock Gsm-symbolic: Understanding the limitations of mathematical reasoning in large language models, 2024.
\newblock URL \url{https://arxiv.org/abs/2410.05229}.

\bibitem[Morrison et~al.(2024)Morrison, Smith, Hajishirzi, Koh, Dodge, and Dasigi]{morrison2024mergelearnefficientlyadding}
J.~Morrison, N.~A. Smith, H.~Hajishirzi, P.~W. Koh, J.~Dodge, and P.~Dasigi.
\newblock Merge to learn: Efficiently adding skills to language models with model merging, 2024.
\newblock URL \url{https://arxiv.org/abs/2410.12937}.

\bibitem[MosaicML(2024)]{mosaic-jeopardy}
MosaicML.
\newblock Llm foundry - jeopardy dataset.
\newblock \url{https://github.com/mosaicml/llm-foundry/blob/main/scripts/eval/local_data/world_knowledge/jeopardy_all.jsonl}, 2024.
\newblock Accessed: 2024-11-10.

\bibitem[Moshkov et~al.(2025)Moshkov, Hanley, Sorokin, Toshniwal, Henkel, Schifferer, Du, and Gitman]{moshkov2025aimo2}
I.~Moshkov, D.~Hanley, I.~Sorokin, S.~Toshniwal, C.~Henkel, B.~Schifferer, W.~Du, and I.~Gitman.
\newblock {AIMO-2 Winning Solution: Building State-of-the-Art Mathematical Reasoning Models with OpenMathReasoning dataset}.
\newblock \emph{arXiv preprint arXiv:2504.16891}, 2025.

\bibitem[Muennighoff et~al.(2025{\natexlab{a}})Muennighoff, Rush, Barak, Scao, Piktus, Tazi, Pyysalo, Wolf, and Raffel]{muennighoff2025scalingdataconstrainedlanguagemodels}
N.~Muennighoff, A.~M. Rush, B.~Barak, T.~L. Scao, A.~Piktus, N.~Tazi, S.~Pyysalo, T.~Wolf, and C.~Raffel.
\newblock Scaling data-constrained language models, 2025{\natexlab{a}}.
\newblock URL \url{https://arxiv.org/abs/2305.16264}.

\bibitem[Muennighoff et~al.(2025{\natexlab{b}})Muennighoff, Yang, Shi, Li, Fei-Fei, Hajishirzi, Zettlemoyer, Liang, Candès, and Hashimoto]{muennighoff2025s1simpletesttimescaling}
N.~Muennighoff, Z.~Yang, W.~Shi, X.~L. Li, L.~Fei-Fei, H.~Hajishirzi, L.~Zettlemoyer, P.~Liang, E.~Candès, and T.~Hashimoto.
\newblock s1: Simple test-time scaling, 2025{\natexlab{b}}.
\newblock URL \url{https://arxiv.org/abs/2501.19393}.

\bibitem[Nathawani et~al.(2025)Nathawani, Gitman, Majumdar, Bakhturina, Sunil~Mahabaleshwarkar, , Zhang, and Polak~Scowcroft]{NemotronPostTrainingDatasetV1}
D.~Nathawani, I.~Gitman, S.~Majumdar, E.~Bakhturina, A.~Sunil~Mahabaleshwarkar, , J.~Zhang, and J.~Polak~Scowcroft.
\newblock {Nemotron-Post-Training-Dataset-v1}, 2025.
\newblock URL \url{https://huggingface.co/datasets/nvidia/Nemotron-Post-Training-Dataset-v1}.

\bibitem[Nelson et~al.(2024)Nelson, Kollias, Das, Chaudhury, and Dan]{nelson2024needlehaystackmemorybased}
E.~Nelson, G.~Kollias, P.~Das, S.~Chaudhury, and S.~Dan.
\newblock Needle in the haystack for memory based large language models, 2024.
\newblock URL \url{https://arxiv.org/abs/2407.01437}.

\bibitem[Noukhovitch et~al.(2024)Noukhovitch, Huang, Xhonneux, Hosseini, Agarwal, and Courville]{noukhovitch2024asynchronousrlhffasterefficient}
M.~Noukhovitch, S.~Huang, S.~Xhonneux, A.~Hosseini, R.~Agarwal, and A.~Courville.
\newblock Asynchronous rlhf: Faster and more efficient off-policy rl for language models, 2024.
\newblock URL \url{https://arxiv.org/abs/2410.18252}.

\bibitem[NVIDIA et~al.(2025)NVIDIA, , Basant, Khairnar, Paithankar, Khattar, Renduchintala, Malte, Bercovich, Hazare, Rico, Ficek, Kondratenko, Shaposhnikov, Bukharin, Taghibakhshi, Barton, Mahabaleshwarkar, Shen, Tao, Guan, Shors, Mandarwal, Mehta, Venkatesan, Sharabiani, Aithal, Poojary, Dattagupta, Buddharaju, Zhu, Simkin, Kartal, Rouhani, Chen, Ginsburg, Norick, Yu, Catanzaro, Wang, Truong, Mungekar, Patel, Alexiuk, Munley, Parisien, Su, Afrimi, Korzekwa, Rohrer, Gitman, Mosallanezhad, Narayanan, Rekesh, Yared, Pykhtar, Ahn, Riach, Long, Ning, Chung, Galinkin, Bakhturina, Prasad, Shen, Qian, Elisha, Sharma, Ross, Ngo, Sahota, Wang, Shin, Huang, Cunningham, Gitman, Moshkov, Jung, Kautz, Scowcroft, Casper, Zhang, Zeng, Zhang, Xue, Huang, Conway, Kamalu, Cohen, Jennings, Vialard, Yi, Parmar, Briski, Cheung, Luna, Wyss, Santhanam, Kong, Pawelec, Anik, Li, Ahmadian, McAfee, Sleiman, Derczynski, Vega, de~Melo, Sreedhar, Chochowski, Cai, Kliegl, Stepniewska-Dziubinska, Novikov, Samadi, Price, Boubdir, Boone,
  Evans, Bien, Zawalski, Martinez, Chrzanowski, Shoeybi, Patwary, Dhameja, Assaf, Habibi, Bhatia, Pope, Tajbakhsh, Juluru, Rybakov, Hrinchuk, Kuchaiev, Olabiyi, Ribalta, Subramanian, Chadha, Molchanov, Dykas, Jin, Bialecki, Januszewski, Thalasta, Gaikwad, Varshney, Gundecha, Tredak, Mahabadi, Patel, El-Yaniv, Rajan, Cheruvu, Shahbazyan, Borkar, Gala, Waleffe, Zhang, Hewett, Prenger, Jain, Kriman, Satheesh, Kaji, Yurick, Muralidharan, Narenthiran, Bak, Sameni, Han, Ramasamy, Ghosh, Sreenivas, Thomas, Diao, Gopal, Prabhumoye, Toshniwal, Ding, Singh, Jain, Majumdar, Singhal, Alborghetti, Akter, Kong, Moon, Hliwiak, Asida, Wang, Konuk, Vashishth, Poon, Karpas, Noroozi, Srinivasan, Korthikanti, Fugro, Kalluru, Kurin, Lavrukhin, Ahmad, Du, Byeon, Lu, Dong, Karnati, Choi, Zhang, Lin, Fu, Suhara, Dong, Li, Zhu, and Chen]{nvidia2025nvidianemotronnano2}
NVIDIA, , A.~Basant, A.~Khairnar, A.~Paithankar, A.~Khattar, A.~Renduchintala, A.~Malte, A.~Bercovich, A.~Hazare, A.~Rico, A.~Ficek, A.~Kondratenko, A.~Shaposhnikov, A.~Bukharin, A.~Taghibakhshi, A.~Barton, A.~S. Mahabaleshwarkar, A.~Shen, A.~Tao, A.~Guan, A.~Shors, A.~Mandarwal, A.~Mehta, A.~Venkatesan, A.~Sharabiani, A.~Aithal, A.~Poojary, A.~Dattagupta, B.~Buddharaju, B.~Zhu, B.~Simkin, B.~Kartal, B.~D. Rouhani, B.~Chen, B.~Ginsburg, B.~Norick, B.~Yu, B.~Catanzaro, C.~Wang, C.~Truong, C.~Mungekar, C.~Patel, C.~Alexiuk, C.~Munley, C.~Parisien, D.~Su, D.~Afrimi, D.~Korzekwa, D.~Rohrer, D.~Gitman, D.~Mosallanezhad, D.~Narayanan, D.~Rekesh, D.~Yared, D.~Pykhtar, D.~Ahn, D.~Riach, E.~Long, E.~Ning, E.~Chung, E.~Galinkin, E.~Bakhturina, G.~Prasad, G.~Shen, H.~Qian, H.~Elisha, H.~Sharma, H.~Ross, H.~Ngo, H.~Sahota, H.~Wang, H.~C. Shin, H.~Huang, I.~Cunningham, I.~Gitman, I.~Moshkov, J.~Jung, J.~Kautz, J.~P. Scowcroft, J.~Casper, J.~Zhang, J.~Zeng, J.~Zhang, J.~Xue, J.~Huang, J.~Conway, J.~Kamalu, J.~Cohen,
  J.~Jennings, J.~V. Vialard, J.~Yi, J.~Parmar, K.~Briski, K.~Cheung, K.~Luna, K.~Wyss, K.~Santhanam, K.~Kong, K.~Pawelec, K.~Anik, K.~Li, K.~Ahmadian, L.~McAfee, L.~Sleiman, L.~Derczynski, L.~Vega, M.~R. de~Melo, M.~N. Sreedhar, M.~Chochowski, M.~Cai, M.~Kliegl, M.~Stepniewska-Dziubinska, M.~Novikov, M.~Samadi, M.~Price, M.~Boubdir, M.~Boone, M.~Evans, M.~Bien, M.~Zawalski, M.~Martinez, M.~Chrzanowski, M.~Shoeybi, M.~Patwary, N.~Dhameja, N.~Assaf, N.~Habibi, N.~Bhatia, N.~Pope, N.~Tajbakhsh, N.~K. Juluru, O.~Rybakov, O.~Hrinchuk, O.~Kuchaiev, O.~Olabiyi, P.~Ribalta, P.~Subramanian, P.~Chadha, P.~Molchanov, P.~Dykas, P.~Jin, P.~Bialecki, P.~Januszewski, P.~Thalasta, P.~Gaikwad, P.~Varshney, P.~Gundecha, P.~Tredak, R.~K. Mahabadi, R.~Patel, R.~El-Yaniv, R.~Rajan, R.~Cheruvu, R.~Shahbazyan, R.~Borkar, R.~Gala, R.~Waleffe, R.~Zhang, R.~J. Hewett, R.~Prenger, S.~Jain, S.~Kriman, S.~Satheesh, S.~Kaji, S.~Yurick, S.~Muralidharan, S.~Narenthiran, S.~Bak, S.~Sameni, S.~Han, S.~Ramasamy, S.~Ghosh, S.~T. Sreenivas,
  S.~Thomas, S.~Diao, S.~Gopal, S.~Prabhumoye, S.~Toshniwal, S.~Ding, S.~Singh, S.~Jain, S.~Majumdar, S.~Singhal, S.~Alborghetti, S.~N. Akter, T.~Kong, T.~Moon, T.~Hliwiak, T.~Asida, T.~Wang, T.~Konuk, T.~Vashishth, T.~Poon, U.~Karpas, V.~Noroozi, V.~Srinivasan, V.~Korthikanti, V.~Fugro, V.~Kalluru, V.~Kurin, V.~Lavrukhin, W.~U. Ahmad, W.~Du, W.~Byeon, X.~Lu, X.~Dong, Y.~Karnati, Y.~Choi, Y.~Zhang, Y.~Lin, Y.~Fu, Y.~Suhara, Z.~Dong, Z.~Li, Z.~Zhu, and Z.~Chen.
\newblock {NVIDIA Nemotron Nano 2}: An accurate and efficient hybrid mamba-transformer reasoning model, 2025.
\newblock URL \url{https://arxiv.org/abs/2508.14444}.

\bibitem[{NVIDIA AI}(2025)]{nvidia2025nemotron_post_training_dataset}
{NVIDIA AI}.
\newblock Nemotron-post-training-dataset-v1.
\newblock \url{https://huggingface.co/datasets/nvidia/Nemotron-Post-Training-Dataset-v1}, 2025.
\newblock Dataset.

\bibitem[Olieslagers et~al.(2024)Olieslagers, Bnaya, Li, and Ma]{Olieslagers2024}
J.~Olieslagers, Z.~Bnaya, Y.~Li, and W.~Ma.
\newblock Backward reasoning through and/or trees to solve problems.
\newblock In \emph{Proceedings of the Annual Meeting of the Cognitive Science Society}, volume~46. Cognitive Science Society, 2024.
\newblock URL \url{https://escholarship.org/uc/item/9h4863xm}.
\newblock Retrieved from \url{https://escholarship.org/uc/item/9h4863xm}.

\bibitem[OLMo et~al.(2024)OLMo, Walsh, Soldaini, Groeneveld, Lo, Arora, Bhagia, Gu, Huang, Jordan, Lambert, Schwenk, Tafjord, Anderson, Atkinson, Brahman, Clark, Dasigi, Dziri, Guerquin, Ivison, Koh, Liu, Malik, Merrill, Miranda, Morrison, Murray, Nam, Pyatkin, Rangapur, Schmitz, Skjonsberg, Wadden, Wilhelm, Wilson, Zettlemoyer, Farhadi, Smith, and Hajishirzi]{olmo20242olmo2furious}
T.~OLMo, P.~Walsh, L.~Soldaini, D.~Groeneveld, K.~Lo, S.~Arora, A.~Bhagia, Y.~Gu, S.~Huang, M.~Jordan, N.~Lambert, D.~Schwenk, O.~Tafjord, T.~Anderson, D.~Atkinson, F.~Brahman, C.~Clark, P.~Dasigi, N.~Dziri, M.~Guerquin, H.~Ivison, P.~W. Koh, J.~Liu, S.~Malik, W.~Merrill, L.~J.~V. Miranda, J.~Morrison, T.~Murray, C.~Nam, V.~Pyatkin, A.~Rangapur, M.~Schmitz, S.~Skjonsberg, D.~Wadden, C.~Wilhelm, M.~Wilson, L.~Zettlemoyer, A.~Farhadi, N.~A. Smith, and H.~Hajishirzi.
\newblock 2 olmo 2 furious, 2024.
\newblock URL \url{https://arxiv.org/abs/2501.00656}.

\bibitem[OpenAI(2023{\natexlab{a}})]{gpt35}
OpenAI.
\newblock {GPT-3.5} turbo, 2023{\natexlab{a}}.
\newblock URL \url{https://platform.openai.com/docs/models/gp#gpt-3-5-turbo}.

\bibitem[OpenAI(2023{\natexlab{b}})]{gpt4}
OpenAI.
\newblock {GPT-4} technical report.
\newblock \emph{ArXiv}, abs/2303.08774, 2023{\natexlab{b}}.
\newblock URL \url{https://api.semanticscholar.org/CorpusID:257532815}.

\bibitem[OpenAI(2025)]{openai-gpt5-systemcard}
OpenAI.
\newblock Gpt-5 system card.
\newblock Technical report, OpenAI, Aug. 2025.
\newblock Accessed: 2025-10-07.

\bibitem[Pal et~al.(2022)Pal, Umapathi, and Sankarasubbu]{pmlr-v174-pal22a}
A.~Pal, L.~K. Umapathi, and M.~Sankarasubbu.
\newblock Medmcqa: A large-scale multi-subject multi-choice dataset for medical domain question answering.
\newblock In G.~Flores, G.~H. Chen, T.~Pollard, J.~C. Ho, and T.~Naumann, editors, \emph{Proceedings of the Conference on Health, Inference, and Learning}, volume 174 of \emph{Proceedings of Machine Learning Research}, pages 248--260. PMLR, 07--08 Apr 2022.
\newblock URL \url{https://proceedings.mlr.press/v174/pal22a.html}.

\bibitem[Pandey(2024)]{pandey2024gzippredictsdatadependentscaling}
R.~Pandey.
\newblock gzip predicts data-dependent scaling laws, 2024.
\newblock URL \url{https://arxiv.org/abs/2405.16684}.

\bibitem[Paperno et~al.(2016)Paperno, Kruszewski, Lazaridou, Pham, Bernardi, Pezzelle, Baroni, Boleda, and Fern{\'a}ndez]{paperno2016lambada}
D.~Paperno, G.~Kruszewski, A.~Lazaridou, Q.~N. Pham, R.~Bernardi, S.~Pezzelle, M.~Baroni, G.~Boleda, and R.~Fern{\'a}ndez.
\newblock The lambada dataset: Word prediction requiring a broad discourse context.
\newblock \emph{arXiv preprint arXiv:1606.06031}, 2016.

\bibitem[Parrish et~al.(2022)Parrish, Chen, Nangia, Padmakumar, Phang, Thompson, Htut, and Bowman]{parrish-etal-2022-bbq}
A.~Parrish, A.~Chen, N.~Nangia, V.~Padmakumar, J.~Phang, J.~Thompson, P.~M. Htut, and S.~Bowman.
\newblock {BBQ}: A hand-built bias benchmark for question answering.
\newblock In S.~Muresan, P.~Nakov, and A.~Villavicencio, editors, \emph{Findings of the Association for Computational Linguistics: ACL 2022}, pages 2086--2105, Dublin, Ireland, May 2022. Association for Computational Linguistics.
\newblock \doi{10.18653/v1/2022.findings-acl.165}.
\newblock URL \url{https://aclanthology.org/2022.findings-acl.165/}.

\bibitem[Paster et~al.(2023)Paster, Santos, Azerbayev, and Ba]{paster2023openwebmath}
K.~Paster, M.~D. Santos, Z.~Azerbayev, and J.~Ba.
\newblock Openwebmath: An open dataset of high-quality mathematical web text, 2023.

\bibitem[Patil et~al.(2025)Patil, Mao, Cheng-Jie~Ji, Yan, Suresh, Stoica, and E.~Gonzalez]{patil2025bfcl}
S.~G. Patil, H.~Mao, C.~Cheng-Jie~Ji, F.~Yan, V.~Suresh, I.~Stoica, and J.~E.~Gonzalez.
\newblock The berkeley function calling leaderboard (bfcl): From tool use to agentic evaluation of large language models.
\newblock In \emph{Forty-second International Conference on Machine Learning}, 2025.

\bibitem[Penedo et~al.(2023)Penedo, Malartic, Hesslow, Cojocaru, Cappelli, Alobeidli, Pannier, Almazrouei, and Launay]{penedo2023refinedwebdatasetfalconllm}
G.~Penedo, Q.~Malartic, D.~Hesslow, R.~Cojocaru, A.~Cappelli, H.~Alobeidli, B.~Pannier, E.~Almazrouei, and J.~Launay.
\newblock The refinedweb dataset for falcon llm: Outperforming curated corpora with web data, and web data only, 2023.
\newblock URL \url{https://arxiv.org/abs/2306.01116}.

\bibitem[Penedo et~al.(2024)Penedo, Kydl{\'\i}{\v{c}}ek, Lozhkov, Mitchell, Raffel, Von~Werra, Wolf, et~al.]{penedo2024fineweb}
G.~Penedo, H.~Kydl{\'\i}{\v{c}}ek, A.~Lozhkov, M.~Mitchell, C.~Raffel, L.~Von~Werra, T.~Wolf, et~al.
\newblock {The FineWeb Datasets: Decanting the Web for the Finest Text Data at Scale}.
\newblock In \emph{{The Thirty-eight Conference on Neural Information Processing Systems; Datasets and Benchmarks Track}}, 2024.

\bibitem[Peng et~al.(2023)Peng, Quesnelle, Fan, and Shippole]{peng2023yarnefficientcontextwindow}
B.~Peng, J.~Quesnelle, H.~Fan, and E.~Shippole.
\newblock Yarn: Efficient context window extension of large language models, 2023.
\newblock URL \url{https://arxiv.org/abs/2309.00071}.

\bibitem[Pham et~al.(2025)Pham, Chang, and Iyyer]{pham2025clipper}
C.~M. Pham, Y.~Chang, and M.~Iyyer.
\newblock Clipper: Compression enables long-context synthetic data generation, 2025.
\newblock URL \url{https://arxiv.org/abs/2502.14854}.

\bibitem[Pich{\'e} et~al.(2025)Pich{\'e}, Kamaloo, Pardinas, and Bahdanau]{pipelinerl}
A.~Pich{\'e}, E.~Kamaloo, R.~Pardinas, and D.~Bahdanau.
\newblock Pipelinerl: Faster on-policy reinforcement learning for long sequence generatio.
\newblock \emph{arXiv preprint arXiv:2509.19128}, 2025.

\bibitem[Poznanski et~al.(2025{\natexlab{a}})Poznanski, Rangapur, Borchardt, Dunkelberger, Huff, Lin, Wilhelm, Lo, and Soldaini]{poznanski2025olmocr}
J.~Poznanski, A.~Rangapur, J.~Borchardt, J.~Dunkelberger, R.~Huff, D.~Lin, C.~Wilhelm, K.~Lo, and L.~Soldaini.
\newblock {olmOCR: Unlocking trillions of tokens in pdfs with vision language models}.
\newblock \emph{arXiv preprint arXiv:2502.18443}, 2025{\natexlab{a}}.

\bibitem[Poznanski et~al.(2025{\natexlab{b}})Poznanski, Soldaini, and Lo]{poznanski2025olmocr2unittest}
J.~Poznanski, L.~Soldaini, and K.~Lo.
\newblock {olmOCR 2: Unit Test Rewards for Document OCR}, 2025{\natexlab{b}}.
\newblock URL \url{https://arxiv.org/abs/2510.19817}.

\bibitem[{PrimeIntellect}(2025)]{primeintellect2025synthetic2}
{PrimeIntellect}.
\newblock Synthetic-2.
\newblock \url{https://huggingface.co/datasets/PrimeIntellect/SYNTHETIC-2}, 2025.
\newblock Dataset.

\bibitem[Pyatkin et~al.(2025)Pyatkin, Malik, Graf, Ivison, Huang, Dasigi, Lambert, and Hajishirzi]{pyatkin2025generalizing}
V.~Pyatkin, S.~Malik, V.~Graf, H.~Ivison, S.~Huang, P.~Dasigi, N.~Lambert, and H.~Hajishirzi.
\newblock Generalizing verifiable instruction following.
\newblock \emph{arXiv preprint arXiv:2507.02833}, 2025.

\bibitem[Qwen et~al.(2024)Qwen, :, Yang, Yang, Zhang, Hui, Zheng, Yu, Li, Liu, Huang, Wei, Lin, Yang, Tu, Zhang, Yang, Yang, Zhou, Lin, Dang, Lu, Bao, Yang, Yu, Li, Xue, Zhang, Zhu, Men, Lin, Li, Xia, Ren, Ren, Fan, Su, Zhang, Wan, Liu, Cui, Zhang, and Qiu]{qwen2.5}
Qwen, :, A.~Yang, B.~Yang, B.~Zhang, B.~Hui, B.~Zheng, B.~Yu, C.~Li, D.~Liu, F.~Huang, H.~Wei, H.~Lin, J.~Yang, J.~Tu, J.~Zhang, J.~Yang, J.~Yang, J.~Zhou, J.~Lin, K.~Dang, K.~Lu, K.~Bao, K.~Yang, L.~Yu, M.~Li, M.~Xue, P.~Zhang, Q.~Zhu, R.~Men, R.~Lin, T.~Li, T.~Xia, X.~Ren, X.~Ren, Y.~Fan, Y.~Su, Y.~Zhang, Y.~Wan, Y.~Liu, Z.~Cui, Z.~Zhang, and Z.~Qiu.
\newblock Qwen2.5 technical report, 2024.
\newblock URL \url{https://arxiv.org/abs/2412.15115}.

\bibitem[{Qwen Team}(2025)]{qwen_qwq_32b_2025}
{Qwen Team}.
\newblock Qwq-32b: Embracing the power of reinforcement learning.
\newblock \url{https://qwenlm.github.io/blog/qwq-32b/}, Mar. 2025.
\newblock Model release blog.

\bibitem[Rafailov et~al.(2024)Rafailov, Sharma, Mitchell, Manning, Ermon, and Finn]{rafailov2024direct}
R.~Rafailov, A.~Sharma, E.~Mitchell, C.~D. Manning, S.~Ermon, and C.~Finn.
\newblock Direct preference optimization: Your language model is secretly a reward model.
\newblock \emph{Advances in Neural Information Processing Systems}, 36, 2024.

\bibitem[Rajpurkar et~al.(2016)Rajpurkar, Zhang, Lopyrev, and Liang]{rajpurkar-etal-2016-squad}
P.~Rajpurkar, J.~Zhang, K.~Lopyrev, and P.~Liang.
\newblock {SQ}u{AD}: 100,000+ questions for machine comprehension of text.
\newblock In J.~Su, K.~Duh, and X.~Carreras, editors, \emph{Proceedings of the 2016 Conference on Empirical Methods in Natural Language Processing}, pages 2383--2392, Austin, Texas, Nov. 2016. Association for Computational Linguistics.
\newblock \doi{10.18653/v1/D16-1264}.
\newblock URL \url{https://aclanthology.org/D16-1264}.

\bibitem[Rasley et~al.(2020)Rasley, Rajbhandari, Ruwase, and He]{deepspeed}
J.~Rasley, S.~Rajbhandari, O.~Ruwase, and Y.~He.
\newblock Deepspeed: System optimizations enable training deep learning models with over 100 billion parameters.
\newblock In \emph{Proceedings of the 26th ACM SIGKDD international conference on knowledge discovery \& data mining}, pages 3505--3506, 2020.

\bibitem[Reddy et~al.(2019)Reddy, Chen, and Manning]{reddy-etal-2019-coqa}
S.~Reddy, D.~Chen, and C.~D. Manning.
\newblock {C}o{QA}: A conversational question answering challenge.
\newblock \emph{Transactions of the Association for Computational Linguistics}, 7:\penalty0 249--266, 2019.
\newblock \doi{10.1162/tacl_a_00266}.
\newblock URL \url{https://aclanthology.org/Q19-1016}.

\bibitem[Rein et~al.(2024)Rein, Hou, Stickland, Petty, Pang, Dirani, Michael, and Bowman]{rein2024gpqa}
D.~Rein, B.~L. Hou, A.~C. Stickland, J.~Petty, R.~Y. Pang, J.~Dirani, J.~Michael, and S.~R. Bowman.
\newblock {GPQA}: A graduate-level google-proof q\&a benchmark.
\newblock In \emph{First Conference on Language Modeling}, 2024.
\newblock URL \url{https://openreview.net/forum?id=Ti67584b98}.

\bibitem[R{\"o}ttger et~al.(2023)R{\"o}ttger, Kirk, Vidgen, Attanasio, Bianchi, and Hovy]{rottger2023xstest}
P.~R{\"o}ttger, H.~R. Kirk, B.~Vidgen, G.~Attanasio, F.~Bianchi, and D.~Hovy.
\newblock Xstest: A test suite for identifying exaggerated safety behaviours in large language models.
\newblock \emph{arXiv preprint arXiv:2308.01263}, 2023.

\bibitem[Rozière et~al.(2024)Rozière, Gehring, Gloeckle, Sootla, Gat, Tan, Adi, Liu, Sauvestre, Remez, Rapin, Kozhevnikov, Evtimov, Bitton, Bhatt, Ferrer, Grattafiori, Xiong, Défossez, Copet, Azhar, Touvron, Martin, Usunier, Scialom, and Synnaeve]{rozière2024codellamaopenfoundation}
B.~Rozière, J.~Gehring, F.~Gloeckle, S.~Sootla, I.~Gat, X.~E. Tan, Y.~Adi, J.~Liu, R.~Sauvestre, T.~Remez, J.~Rapin, A.~Kozhevnikov, I.~Evtimov, J.~Bitton, M.~Bhatt, C.~C. Ferrer, A.~Grattafiori, W.~Xiong, A.~Défossez, J.~Copet, F.~Azhar, H.~Touvron, L.~Martin, N.~Usunier, T.~Scialom, and G.~Synnaeve.
\newblock Code llama: Open foundation models for code, 2024.
\newblock URL \url{https://arxiv.org/abs/2308.12950}.

\bibitem[Sakaguchi et~al.(2020)Sakaguchi, Le~Bras, Bhagavatula, and Choi]{Sakaguchi_Le_Bras_Bhagavatula_Choi_2020}
K.~Sakaguchi, R.~Le~Bras, C.~Bhagavatula, and Y.~Choi.
\newblock Wino{G}rande: An adversarial winograd schema challenge at scale.
\newblock \emph{Proceedings of the AAAI Conference on Artificial Intelligence}, 34\penalty0 (05):\penalty0 8732--8740, Apr. 2020.
\newblock \doi{10.1609/aaai.v34i05.6399}.
\newblock URL \url{https://ojs.aaai.org/index.php/AAAI/article/view/6399}.

\bibitem[Sap et~al.(2019)Sap, Rashkin, Chen, Le~Bras, and Choi]{sap-etal-2019-social}
M.~Sap, H.~Rashkin, D.~Chen, R.~Le~Bras, and Y.~Choi.
\newblock Social {IQ}a: Commonsense reasoning about social interactions.
\newblock In K.~Inui, J.~Jiang, V.~Ng, and X.~Wan, editors, \emph{Proceedings of the 2019 Conference on Empirical Methods in Natural Language Processing and the 9th International Joint Conference on Natural Language Processing (EMNLP-IJCNLP)}, pages 4463--4473, Hong Kong, China, Nov. 2019. Association for Computational Linguistics.
\newblock \doi{10.18653/v1/D19-1454}.
\newblock URL \url{https://aclanthology.org/D19-1454}.

\bibitem[Saxton et~al.(2019)Saxton, Grefenstette, Hill, and Kohli]{saxton2019analysing}
D.~Saxton, E.~Grefenstette, F.~Hill, and P.~Kohli.
\newblock Analysing mathematical reasoning abilities of neural models.
\newblock \emph{arXiv preprint arXiv:1904.01557}, 2019.

\bibitem[Schaeffer et~al.(2023)Schaeffer, Miranda, and Koyejo]{schaeffer2023emergent}
R.~Schaeffer, B.~Miranda, and S.~Koyejo.
\newblock Are emergent abilities of large language models a mirage?
\newblock \emph{Advances in neural information processing systems}, 36:\penalty0 55565--55581, 2023.

\bibitem[Shao et~al.(2025{\natexlab{a}})Shao, Asai, Shen, Ivison, Kishore, Zhuo, Zhao, Park, Finlayson, Sontag, Murray, Min, Dasigi, Soldaini, Brahman, tau Yih, Wu, Zettlemoyer, Kim, Hajishirzi, and Koh]{drtulu}
R.~Shao, A.~Asai, S.~Z. Shen, H.~Ivison, V.~Kishore, J.~Zhuo, X.~Zhao, M.~Park, S.~G. Finlayson, D.~Sontag, T.~Murray, S.~Min, P.~Dasigi, L.~Soldaini, F.~Brahman, W.~tau Yih, T.~Wu, L.~Zettlemoyer, Y.~Kim, H.~Hajishirzi, and P.~W. Koh.
\newblock {DR Tulu}: Reinforcement learning with evolving rubrics for deep research, 2025{\natexlab{a}}.
\newblock URL \url{https://arxiv.org/abs/2511.19399}.

\bibitem[Shao et~al.(2025{\natexlab{b}})Shao, Li, Xin, Geng, Wang, Oh, Du, Lambert, Min, Krishna, Tsvetkov, Hajishirzi, Koh, and Zettlemoyer]{shao2025spuriousrewardsrethinkingtraining}
R.~Shao, S.~S. Li, R.~Xin, S.~Geng, Y.~Wang, S.~Oh, S.~S. Du, N.~Lambert, S.~Min, R.~Krishna, Y.~Tsvetkov, H.~Hajishirzi, P.~W. Koh, and L.~Zettlemoyer.
\newblock Spurious rewards: Rethinking training signals in rlvr, 2025{\natexlab{b}}.
\newblock URL \url{https://arxiv.org/abs/2506.10947}.

\bibitem[Shao et~al.(2024)Shao, Wang, Zhu, Xu, Song, Bi, Zhang, Zhang, Li, Wu, et~al.]{shao2024deepseekmath}
Z.~Shao, P.~Wang, Q.~Zhu, R.~Xu, J.~Song, X.~Bi, H.~Zhang, M.~Zhang, Y.~Li, Y.~Wu, et~al.
\newblock Deepseekmath: Pushing the limits of mathematical reasoning in open language models.
\newblock \emph{arXiv preprint arXiv:2402.03300}, 2024.

\bibitem[Shen et~al.(2024)Shen, Chen, Backes, Shen, and Zhang]{shen2024anything}
X.~Shen, Z.~Chen, M.~Backes, Y.~Shen, and Y.~Zhang.
\newblock " do anything now": Characterizing and evaluating in-the-wild jailbreak prompts on large language models.
\newblock In \emph{Proceedings of the 2024 on ACM SIGSAC Conference on Computer and Communications Security}, pages 1671--1685, 2024.

\bibitem[Shi et~al.(2025)Shi, Cao, Chen, Sun, Li, Lu, Dong, Qin, Zhu, Liu, Yang, Zhang, Liu, Zhang, Wang, Jiang, and Zhou]{Shi2025TaskCraftAG}
D.~Shi, J.~Cao, Q.~Chen, W.~Sun, W.~Li, H.~Lu, F.~Dong, T.~Qin, K.~Zhu, M.~Liu, J.~Yang, G.~Zhang, J.~Liu, C.~Zhang, J.~Wang, Y.~E. Jiang, and W.~Zhou.
\newblock Taskcraft: Automated generation of agentic tasks.
\newblock \emph{ArXiv}, abs/2506.10055, 2025.
\newblock URL \url{https://api.semanticscholar.org/CorpusID:279318561}.

\bibitem[Silver et~al.(2017)Silver, Hubert, Schrittwieser, Antonoglou, Lai, Guez, Lanctot, Sifre, Kumaran, Graepel, et~al.]{silver2017alphazero}
D.~Silver, T.~Hubert, J.~Schrittwieser, I.~Antonoglou, M.~Lai, A.~Guez, M.~Lanctot, L.~Sifre, D.~Kumaran, T.~Graepel, et~al.
\newblock Mastering chess and shogi by self-play with a general reinforcement learning algorithm.
\newblock \emph{arXiv preprint arXiv:1712.01815}, 2017.

\bibitem[Singh et~al.(2024)Singh, Vargus, Dsouza, Karlsson, Mahendiran, Ko, Shandilya, Patel, Mataciunas, O'Mahony, Zhang, Hettiarachchi, Wilson, Machado, Moura, Krzemi{\'n}ski, Fadaei, Erg{\"u}n, Okoh, Alaagib, Mudannayake, Alyafeai, Chien, Ruder, Guthikonda, Alghamdi, Gehrmann, Muennighoff, Bartolo, Kreutzer, {\"U}st{\"u}n, Fadaee, and Hooker]{singh2024aya}
S.~Singh, F.~Vargus, D.~Dsouza, B.~F. Karlsson, A.~Mahendiran, W.-Y. Ko, H.~Shandilya, J.~Patel, D.~Mataciunas, L.~O'Mahony, M.~Zhang, R.~Hettiarachchi, J.~Wilson, M.~Machado, L.~S. Moura, D.~Krzemi{\'n}ski, H.~Fadaei, I.~Erg{\"u}n, I.~Okoh, A.~Alaagib, O.~Mudannayake, Z.~Alyafeai, V.~M. Chien, S.~Ruder, S.~Guthikonda, E.~A. Alghamdi, S.~Gehrmann, N.~Muennighoff, M.~Bartolo, J.~Kreutzer, A.~{\"U}st{\"u}n, M.~Fadaee, and S.~Hooker.
\newblock Aya dataset: An open-access collection for multilingual instruction tuning.
\newblock \emph{arXiv preprint arXiv:2402.06619}, 2024.
\newblock URL \url{https://arxiv.org/abs/2402.06619}.

\bibitem[Skarlinski et~al.(2024)Skarlinski, Cox, Laurent, Braza, Hinks, Hammerling, Ponnapati, Rodriques, and White]{skarlinski2024language}
M.~D. Skarlinski, S.~Cox, J.~M. Laurent, J.~D. Braza, M.~Hinks, M.~J. Hammerling, M.~Ponnapati, S.~G. Rodriques, and A.~D. White.
\newblock Language agents achieve superhuman synthesis of scientific knowledge.
\newblock \emph{arXiv preprint arXiv:2409.13740}, 2024.

\bibitem[Soldaini and Lo(2023)]{peS2o}
L.~Soldaini and K.~Lo.
\newblock {peS2o (Pretraining Efficiently on S2ORC) Dataset}, 2023.
\newblock URL \url{https://github.com/allenai/pes2o}.

\bibitem[Soldaini et~al.(2024)Soldaini, Kinney, Bhagia, Schwenk, Atkinson, Authur, Bogin, Chandu, Dumas, Elazar, Hofmann, Jha, Kumar, Lucy, Lyu, Lambert, Magnusson, Morrison, Muennighoff, Naik, Nam, Peters, Ravichander, Richardson, Shen, Strubell, Subramani, Tafjord, Walsh, Zettlemoyer, Smith, Hajishirzi, Beltagy, Groeneveld, Dodge, and Lo]{soldaini2024dolma}
L.~Soldaini, R.~Kinney, A.~Bhagia, D.~Schwenk, D.~Atkinson, R.~Authur, B.~Bogin, K.~Chandu, J.~Dumas, Y.~Elazar, V.~Hofmann, A.~H. Jha, S.~Kumar, L.~Lucy, X.~Lyu, N.~Lambert, I.~Magnusson, J.~Morrison, N.~Muennighoff, A.~Naik, C.~Nam, M.~E. Peters, A.~Ravichander, K.~Richardson, Z.~Shen, E.~Strubell, N.~Subramani, O.~Tafjord, P.~Walsh, L.~Zettlemoyer, N.~A. Smith, H.~Hajishirzi, I.~Beltagy, D.~Groeneveld, J.~Dodge, and K.~Lo.
\newblock Dolma: an open corpus of three trillion tokens for language model pretraining research, 2024.

\bibitem[Soule and Bergmann(2025)]{souleBergmann2025granite33}
K.~Soule and D.~Bergmann.
\newblock {IBM Granite 3.3: Speech recognition, refined reasoning, and RAG LoRAs}, Apr. 2025.
\newblock URL \url{https://www.ibm.com/new/announcements/ibm-granite-3-3-speech-recognition-refined-reasoning-rag-loras}.
\newblock Blog post.

\bibitem[Souly et~al.(2024)Souly, Lu, Bowen, Trinh, Hsieh, Pandey, Abbeel, Svegliato, Emmons, Watkins, and Toyer]{strongreject}
A.~Souly, Q.~Lu, D.~Bowen, T.~Trinh, E.~Hsieh, S.~Pandey, P.~Abbeel, J.~Svegliato, S.~Emmons, O.~Watkins, and S.~Toyer.
\newblock A strongreject for empty jailbreaks.
\newblock In A.~Globerson, L.~Mackey, D.~Belgrave, A.~Fan, U.~Paquet, J.~Tomczak, and C.~Zhang, editors, \emph{Advances in Neural Information Processing Systems}, volume~37, pages 125416--125440. Curran Associates, Inc., 2024.
\newblock URL \url{https://proceedings.neurips.cc/paper_files/paper/2024/file/e2e06adf560b0706d3b1ddfca9f29756-Paper-Datasets_and_Benchmarks_Track.pdf}.

\bibitem[Stiennon et~al.(2020)Stiennon, Ouyang, Wu, Ziegler, Lowe, Voss, Radford, Amodei, and Christiano]{stiennon2020learning}
N.~Stiennon, L.~Ouyang, J.~Wu, D.~Ziegler, R.~Lowe, C.~Voss, A.~Radford, D.~Amodei, and P.~F. Christiano.
\newblock Learning to summarize with human feedback.
\newblock \emph{Advances in Neural Information Processing Systems}, 33:\penalty0 3008--3021, 2020.

\bibitem[Su et~al.(2025{\natexlab{a}})Su, Kong, Lin, Jennings, Norick, Kliegl, Patwary, Shoeybi, and Catanzaro]{su2025nemotroncctransformingcommoncrawl}
D.~Su, K.~Kong, Y.~Lin, J.~Jennings, B.~Norick, M.~Kliegl, M.~Patwary, M.~Shoeybi, and B.~Catanzaro.
\newblock Nemotron-cc: Transforming common crawl into a refined long-horizon pretraining dataset, 2025{\natexlab{a}}.
\newblock URL \url{https://arxiv.org/abs/2412.02595}.

\bibitem[Su et~al.(2024)Su, Ahmed, Lu, Pan, Bo, and Liu]{su2024roformer}
J.~Su, M.~Ahmed, Y.~Lu, S.~Pan, W.~Bo, and Y.~Liu.
\newblock Roformer: Enhanced transformer with rotary position embedding.
\newblock \emph{Neurocomputing}, 568:\penalty0 127063, 2024.

\bibitem[Su et~al.(2025{\natexlab{b}})Su, Yu, Song, Li, Mi, Tu, Zhang, and Yu]{su2025expanding}
Y.~Su, D.~Yu, L.~Song, J.~Li, H.~Mi, Z.~Tu, M.~Zhang, and D.~Yu.
\newblock Expanding rl with verifiable rewards across diverse domains.
\newblock \emph{arXiv preprint arXiv:2503.23829}, 2025{\natexlab{b}}.

\bibitem[Su et~al.(2025{\natexlab{c}})Su, Pan, Bai, Liu, Dong, Huang, Hu, Zhang, Gai, and Zhou]{su2025klear}
Z.~Su, L.~Pan, X.~Bai, D.~Liu, G.~Dong, J.~Huang, W.~Hu, F.~Zhang, K.~Gai, and G.~Zhou.
\newblock Klear-reasoner: Advancing reasoning capability via gradient-preserving clipping policy optimization.
\newblock \emph{arXiv preprint arXiv:2508.07629}, 2025{\natexlab{c}}.

\bibitem[Sun et~al.(2025)Sun, Hu, Zhou, Zheng, Hajishirzi, Dziri, and Song]{Sun2025OMEGACL}
Y.~Sun, S.~Hu, G.~Zhou, K.~Zheng, H.~Hajishirzi, N.~Dziri, and D.~X. Song.
\newblock Omega: Can llms reason outside the box in math? evaluating exploratory, compositional, and transformative generalization.
\newblock \emph{ArXiv}, abs/2506.18880, 2025.
\newblock URL \url{https://api.semanticscholar.org/CorpusID:280000246}.

\bibitem[Suzgun et~al.(2022)Suzgun, Scales, Sch{\"a}rli, Gehrmann, Tay, Chung, Chowdhery, Le, Chi, Zhou, et~al.]{suzgun2022challenging}
M.~Suzgun, N.~Scales, N.~Sch{\"a}rli, S.~Gehrmann, Y.~Tay, H.~W. Chung, A.~Chowdhery, Q.~V. Le, E.~H. Chi, D.~Zhou, et~al.
\newblock Challenging big-bench tasks and whether chain-of-thought can solve them.
\newblock \emph{arXiv preprint arXiv:2210.09261}, 2022.

\bibitem[Talmor et~al.(2019)Talmor, Herzig, Lourie, and Berant]{talmor-etal-2019-commonsenseqa}
A.~Talmor, J.~Herzig, N.~Lourie, and J.~Berant.
\newblock {C}ommonsense{QA}: A question answering challenge targeting commonsense knowledge.
\newblock In J.~Burstein, C.~Doran, and T.~Solorio, editors, \emph{Proceedings of the 2019 Conference of the North {A}merican Chapter of the Association for Computational Linguistics: Human Language Technologies, Volume 1 (Long and Short Papers)}, pages 4149--4158, Minneapolis, Minnesota, June 2019. Association for Computational Linguistics.
\newblock \doi{10.18653/v1/N19-1421}.
\newblock URL \url{https://aclanthology.org/N19-1421}.

\bibitem[Team et~al.(2025)Team, Liu, Tang, Jin, Li, Ranjan, Fan, Rohatgi, Fan, Pangarkar, Wang, Cheng, Sun, Han, Tan, Gosal, Han, Pimpalkhute, Hao, Hee, Hestness, Jia, Ma, Singh, Soboleva, Vassilieva, Wang, Wu, Sun, Killian, Moreno, Maggs, Ren, He, Wang, Ma, Wang, Yurochkin, and Xing]{k2team2025k2v2360openreasoningenhancedllm}
K.~Team, Z.~Liu, L.~Tang, L.~Jin, H.~Li, N.~Ranjan, D.~Fan, S.~Rohatgi, R.~Fan, O.~Pangarkar, H.~Wang, Z.~Cheng, S.~Sun, S.~Han, B.~Tan, G.~Gosal, X.~Han, V.~Pimpalkhute, S.~Hao, M.~S. Hee, J.~Hestness, H.~Jia, L.~Ma, A.~Singh, D.~Soboleva, N.~Vassilieva, R.~Wang, Y.~Wu, Y.~Sun, T.~Killian, A.~Moreno, J.~Maggs, H.~Ren, G.~He, H.~Wang, X.~Ma, Y.~Wang, M.~Yurochkin, and E.~P. Xing.
\newblock K2-v2: A 360-open, reasoning-enhanced llm, 2025.
\newblock URL \url{https://arxiv.org/abs/2512.06201}.

\bibitem[Teknium(2023)]{OpenHermes}
Teknium.
\newblock Openhermes 2.5: An open dataset of synthetic data for generalist llm assistants, 2023.
\newblock URL \url{https://huggingface.co/datasets/teknium/OpenHermes-2.5}.

\bibitem[{The Algorithms}(2025)]{thealgorithms_python}
{The Algorithms}.
\newblock The algorithms -- python.
\newblock \url{https://github.com/TheAlgorithms/Python}, 2025.
\newblock GitHub repository, MIT License.

\bibitem[{Together AI}(2023)]{together2023redpajama}
{Together AI}.
\newblock {RedPajama}: An open source recipe to reproduce {LLaMA} training dataset, 2023.
\newblock URL \url{https://github.com/togethercomputer/RedPajama-Data}.

\bibitem[Toshniwal et~al.(2024)Toshniwal, Du, Moshkov, Kisacanin, Ayrapetyan, and Gitman]{toshniwal2024openmathinstruct}
S.~Toshniwal, W.~Du, I.~Moshkov, B.~Kisacanin, A.~Ayrapetyan, and I.~Gitman.
\newblock Openmathinstruct-2: Accelerating ai for math with massive open-source instruction data.
\newblock \emph{arXiv preprint arXiv:2410.01560}, 2024.

\bibitem[Toy et~al.(2024)Toy, MacAdam, and Tabor]{toy2024metacognitionneedusingintrospection}
J.~Toy, J.~MacAdam, and P.~Tabor.
\newblock Metacognition is all you need? using introspection in generative agents to improve goal-directed behavior, 2024.
\newblock URL \url{https://arxiv.org/abs/2401.10910}.

\bibitem[Van~Hasselt et~al.(2018)Van~Hasselt, Doron, Strub, Hessel, Sonnerat, and Modayil]{deadly-triad}
H.~Van~Hasselt, Y.~Doron, F.~Strub, M.~Hessel, N.~Sonnerat, and J.~Modayil.
\newblock Deep reinforcement learning and the deadly triad.
\newblock \emph{arXiv preprint arXiv:1812.02648}, 2018.

\bibitem[Vaswani(2025)]{vaswani2025rnj1}
A.~Vaswani.
\newblock Announcing rnj-1: Building instruments of intelligence, Dec. 2025.
\newblock URL \url{https://essential.ai/research/rnj-1}.
\newblock Blog post.

\bibitem[Vaswani et~al.(2017)Vaswani, Shazeer, Parmar, Uszkoreit, Jones, Gomez, Kaiser, and Polosukhin]{vaswani2017attention}
A.~Vaswani, N.~Shazeer, N.~Parmar, J.~Uszkoreit, L.~Jones, A.~N. Gomez, L.~u. Kaiser, and I.~Polosukhin.
\newblock Attention is all you need.
\newblock In I.~Guyon, U.~V. Luxburg, S.~Bengio, H.~Wallach, R.~Fergus, S.~Vishwanathan, and R.~Garnett, editors, \emph{Advances in Neural Information Processing Systems}, volume~30. Curran Associates, Inc., 2017.
\newblock URL \url{https://proceedings.neurips.cc/paper_files/paper/2017/file/3f5ee243547dee91fbd053c1c4a845aa-Paper.pdf}.

\bibitem[Vendrow et~al.(2025)Vendrow, Vendrow, Beery, and Madry]{vendrow2025large}
J.~Vendrow, E.~Vendrow, S.~Beery, and A.~Madry.
\newblock Do large language model benchmarks test reliability?
\newblock \emph{arXiv preprint arXiv:2502.03461}, 2025.

\bibitem[Wadden et~al.(2024)Wadden, Shi, Morrison, Naik, Singh, Barzilay, Lo, Hope, Soldaini, Shen, et~al.]{wadden2024sciriff}
D.~Wadden, K.~Shi, J.~Morrison, A.~Naik, S.~Singh, N.~Barzilay, K.~Lo, T.~Hope, L.~Soldaini, S.~Z. Shen, et~al.
\newblock Sciriff: A resource to enhance language model instruction-following over scientific literature.
\newblock \emph{arXiv preprint arXiv:2406.07835}, 2024.

\bibitem[Wang et~al.(2024{\natexlab{a}})Wang, Ma, Zhang, Ni, Chandra, Guo, Ren, Arulraj, He, Jiang, et~al.]{wang2024mmlu}
Y.~Wang, X.~Ma, G.~Zhang, Y.~Ni, A.~Chandra, S.~Guo, W.~Ren, A.~Arulraj, X.~He, Z.~Jiang, et~al.
\newblock Mmlu-pro: A more robust and challenging multi-task language understanding benchmark.
\newblock \emph{arXiv preprint arXiv:2406.01574}, 2024{\natexlab{a}}.

\bibitem[Wang et~al.(2024{\natexlab{b}})Wang, Dong, Delalleau, Zeng, Shen, Egert, Zhang, Sreedhar, and Kuchaiev]{wang2024helpsteer2}
Z.~Wang, Y.~Dong, O.~Delalleau, J.~Zeng, G.~Shen, D.~Egert, J.~J. Zhang, M.~N. Sreedhar, and O.~Kuchaiev.
\newblock Helpsteer2: Open-source dataset for training top-performing reward models.
\newblock \emph{arXiv preprint arXiv:2406.08673}, 2024{\natexlab{b}}.

\bibitem[Wang et~al.(2025)Wang, Zhou, Li, and Liu]{wang2025octothinker}
Z.~Wang, F.~Zhou, X.~Li, and P.~Liu.
\newblock Octothinker: Mid-training incentivizes reinforcement learning scaling.
\newblock \emph{arXiv preprint arXiv:2506.20512}, 2025.

\bibitem[Ward~Jr(1963)]{ward1963hierarchical}
J.~H. Ward~Jr.
\newblock Hierarchical grouping to optimize an objective function.
\newblock \emph{Journal of the American statistical association}, 58\penalty0 (301):\penalty0 236--244, 1963.

\bibitem[Wei et~al.(2021)Wei, Bosma, Zhao, Guu, Yu, Lester, Du, Dai, and Le]{wei2021flan}
J.~Wei, M.~Bosma, V.~Zhao, K.~Guu, A.~W. Yu, B.~Lester, N.~Du, A.~M. Dai, and Q.~V. Le.
\newblock Finetuned language models are zero-shot learners.
\newblock In \emph{International Conference on Learning Representations}, 2021.

\bibitem[Wei et~al.(2022)Wei, Tay, Bommasani, Raffel, Zoph, Borgeaud, Yogatama, Bosma, Zhou, Metzler, et~al.]{wei2022emergent}
J.~Wei, Y.~Tay, R.~Bommasani, C.~Raffel, B.~Zoph, S.~Borgeaud, D.~Yogatama, M.~Bosma, D.~Zhou, D.~Metzler, et~al.
\newblock Emergent abilities of large language models.
\newblock \emph{arXiv preprint arXiv:2206.07682}, 2022.

\bibitem[Wei et~al.(2024)Wei, Karina, Chung, Jiao, Papay, Glaese, Schulman, and Fedus]{wei2024measuring}
J.~Wei, N.~Karina, H.~W. Chung, Y.~J. Jiao, S.~Papay, A.~Glaese, J.~Schulman, and W.~Fedus.
\newblock Measuring short-form factuality in large language models.
\newblock \emph{arXiv preprint arXiv:2411.04368}, 2024.

\bibitem[Welbl et~al.(2017)Welbl, Liu, and Gardner]{welbl-etal-2017-crowdsourcing}
J.~Welbl, N.~F. Liu, and M.~Gardner.
\newblock Crowdsourcing multiple choice science questions.
\newblock In L.~Derczynski, W.~Xu, A.~Ritter, and T.~Baldwin, editors, \emph{Proceedings of the 3rd Workshop on Noisy User-generated Text}, pages 94--106, Copenhagen, Denmark, Sept. 2017. Association for Computational Linguistics.
\newblock \doi{10.18653/v1/W17-4413}.
\newblock URL \url{https://aclanthology.org/W17-4413/}.

\bibitem[Wettig et~al.(2025)Wettig, Lo, Min, Hajishirzi, Chen, and Soldaini]{weborganizer}
A.~Wettig, K.~Lo, S.~Min, H.~Hajishirzi, D.~Chen, and L.~Soldaini.
\newblock Organize the web: Constructing domains enhances pre-training data curation, 2025.
\newblock URL \url{https://arxiv.org/abs/2502.10341}.

\bibitem[Wortsman et~al.(2022)Wortsman, Ilharco, Gadre, Roelofs, Gontijo-Lopes, Morcos, Namkoong, Farhadi, Carmon, Kornblith, et~al.]{wortsman2022model}
M.~Wortsman, G.~Ilharco, S.~Y. Gadre, R.~Roelofs, R.~Gontijo-Lopes, A.~S. Morcos, H.~Namkoong, A.~Farhadi, Y.~Carmon, S.~Kornblith, et~al.
\newblock Model soups: averaging weights of multiple fine-tuned models improves accuracy without increasing inference time.
\newblock In \emph{International conference on machine learning}, pages 23965--23998. PMLR, 2022.

\bibitem[Wu et~al.(2025{\natexlab{a}})Wu, Yin, Jiang, Wang, Xi, Fang, Zhang, He, Zhou, Xie, and Huang]{wu-etal-2025-webwalker}
J.~Wu, W.~Yin, Y.~Jiang, Z.~Wang, Z.~Xi, R.~Fang, L.~Zhang, Y.~He, D.~Zhou, P.~Xie, and F.~Huang.
\newblock {W}eb{W}alker: Benchmarking {LLM}s in web traversal.
\newblock In W.~Che, J.~Nabende, E.~Shutova, and M.~T. Pilehvar, editors, \emph{Proceedings of the 63rd Annual Meeting of the Association for Computational Linguistics (Volume 1: Long Papers)}, pages 10290--10305, Vienna, Austria, July 2025{\natexlab{a}}. Association for Computational Linguistics.
\newblock ISBN 979-8-89176-251-0.
\newblock \doi{10.18653/v1/2025.acl-long.508}.
\newblock URL \url{https://aclanthology.org/2025.acl-long.508/}.

\bibitem[Wu et~al.(2025{\natexlab{b}})Wu, Zhu, Zhao, Yu, Ran, Wong, Sun, and Li]{wu2025longattn}
L.~Wu, D.~Zhu, G.~Zhao, Z.~Yu, J.~Ran, X.~Wong, L.~Sun, and S.~Li.
\newblock {LongAttn}: Selecting long-context training data via token-level attention, 2025{\natexlab{b}}.
\newblock URL \url{https://arxiv.org/abs/2502.16860}.

\bibitem[Wu et~al.(2025{\natexlab{c}})Wu, Zhang, Dong, Xi, Zhao, Jin, Fan, Zhou, Lv, Zhang, et~al.]{wu2025reasoning}
M.~Wu, Z.~Zhang, Q.~Dong, Z.~Xi, J.~Zhao, S.~Jin, X.~Fan, Y.~Zhou, H.~Lv, M.~Zhang, et~al.
\newblock Reasoning or memorization? unreliable results of reinforcement learning due to data contamination.
\newblock \emph{arXiv preprint arXiv:2507.10532}, 2025{\natexlab{c}}.

\bibitem[Xiaomi et~al.(2025)Xiaomi, :, Xia, Shen, Cici, Zhu, Zhang, Wang, Zhang, Liu, Xiao, Dong, Zhao, Li, Wang, Yu, Chen, Wang, Ma, Deng, Huang, Song, Jiang, Ye, Cai, He, Zhang, Zhang, Wang, Tian, Zhao, Qu, Xu, Shi, Bao, Fang, Zhou, Zhou, Li, Zhu, Chen, Wang, Liu, Li, Gu, Ren, Liu, Deng, Zhuang, Lv, Yang, Zhang, Yong, Zhang, Song, Xu, Wang, Yan, Tu, Tian, Wang, Yu, Lin, Song, and Yue]{mimo}
L.-C. Xiaomi, :, B.~Xia, B.~Shen, Cici, D.~Zhu, D.~Zhang, G.~Wang, H.~Zhang, H.~Liu, J.~Xiao, J.~Dong, L.~Zhao, P.~Li, P.~Wang, S.~Yu, S.~Chen, W.~Wang, W.~Ma, X.~Deng, Y.~Huang, Y.~Song, Z.~Jiang, B.~Ye, C.~Cai, C.~He, D.~Zhang, D.~Zhang, G.~Wang, H.~Tian, H.~Zhao, H.~Qu, H.~Xu, J.~Shi, K.~Bao, K.~Fang, K.~Zhou, K.~Zhou, L.~Li, M.~Zhu, N.~Chen, Q.~Wang, S.~Liu, S.~Li, S.~Gu, S.~Ren, S.~Liu, S.~Deng, W.~Zhuang, W.~Lv, W.~Yang, X.~Zhang, X.~Yong, X.~Zhang, X.~Song, X.~Xu, X.~Wang, Y.~Yan, Y.~Tu, Y.~Tian, Y.~Wang, Y.~Yu, Z.~Lin, Z.~Song, and Z.~Yue.
\newblock {MiMo}: Unlocking the reasoning potential of language model -- from pretraining to posttraining, 2025.
\newblock URL \url{https://arxiv.org/abs/2505.07608}.

\bibitem[Xiong et~al.(2023)Xiong, Liu, Molybog, Zhang, Bhargava, Hou, Martin, Rungta, Sankararaman, Oguz, Khabsa, Fang, Mehdad, Narang, Malik, Fan, Bhosale, Edunov, Lewis, Wang, and Ma]{xiong2023effectivelongcontextscalingfoundation}
W.~Xiong, J.~Liu, I.~Molybog, H.~Zhang, P.~Bhargava, R.~Hou, L.~Martin, R.~Rungta, K.~A. Sankararaman, B.~Oguz, M.~Khabsa, H.~Fang, Y.~Mehdad, S.~Narang, K.~Malik, A.~Fan, S.~Bhosale, S.~Edunov, M.~Lewis, S.~Wang, and H.~Ma.
\newblock Effective long-context scaling of foundation models, 2023.
\newblock URL \url{https://arxiv.org/abs/2309.16039}.

\bibitem[Yang et~al.(2025{\natexlab{a}})Yang, Li, Yang, Zhang, Hui, Zheng, Yu, Gao, Huang, Lv, et~al.]{qwen3}
A.~Yang, A.~Li, B.~Yang, B.~Zhang, B.~Hui, B.~Zheng, B.~Yu, C.~Gao, C.~Huang, C.~Lv, et~al.
\newblock Qwen3 technical report.
\newblock \emph{arXiv preprint arXiv:2505.09388}, 2025{\natexlab{a}}.

\bibitem[Yang et~al.(2025{\natexlab{b}})Yang, Yu, Li, Liu, Huang, Huang, Jiang, Tu, Zhang, Zhou, Lin, Dang, Yang, Yu, Li, Sun, Zhu, Men, He, Xu, Yin, Yu, Qiu, Ren, Yang, Li, Xu, and Zhang]{qwen1M}
A.~Yang, B.~Yu, C.~Li, D.~Liu, F.~Huang, H.~Huang, J.~Jiang, J.~Tu, J.~Zhang, J.~Zhou, J.~Lin, K.~Dang, K.~Yang, L.~Yu, M.~Li, M.~Sun, Q.~Zhu, R.~Men, T.~He, W.~Xu, W.~Yin, W.~Yu, X.~Qiu, X.~Ren, X.~Yang, Y.~Li, Z.~Xu, and Z.~Zhang.
\newblock {Qwen2.5-1M Technical Report}, 2025{\natexlab{b}}.
\newblock URL \url{https://arxiv.org/abs/2501.15383}.

\bibitem[Yang et~al.(2018)Yang, Qi, Zhang, Bengio, Cohen, Salakhutdinov, and Manning]{Yang2018HotpotQAAD}
Z.~Yang, P.~Qi, S.~Zhang, Y.~Bengio, W.~W. Cohen, R.~Salakhutdinov, and C.~D. Manning.
\newblock Hotpotqa: A dataset for diverse, explainable multi-hop question answering.
\newblock In \emph{Conference on Empirical Methods in Natural Language Processing}, 2018.
\newblock URL \url{https://api.semanticscholar.org/CorpusID:52822214}.

\bibitem[Yao et~al.(2025)Yao, Liu, Zhang, Dong, Shang, and Gao]{yao2025offpolicy}
F.~Yao, L.~Liu, D.~Zhang, C.~Dong, J.~Shang, and J.~Gao.
\newblock Your efficient rl framework secretly brings you off-policy rl training, Aug. 2025.
\newblock URL \url{https://fengyao.notion.site/off-policy-rl}.

\bibitem[Ye et~al.(2025)Ye, Liu, Sun, Zhan, Zhou, and Qiu]{ye2025datamixinglawsoptimizing}
J.~Ye, P.~Liu, T.~Sun, J.~Zhan, Y.~Zhou, and X.~Qiu.
\newblock Data mixing laws: Optimizing data mixtures by predicting language modeling performance, 2025.
\newblock URL \url{https://arxiv.org/abs/2403.16952}.

\bibitem[Yen et~al.(2025)Yen, Gao, Hou, Ding, Fleischer, Izsak, Wasserblat, and Chen]{yen2025helmet}
H.~Yen, T.~Gao, M.~Hou, K.~Ding, D.~Fleischer, P.~Izsak, M.~Wasserblat, and D.~Chen.
\newblock {HELMET}: How to evaluate long-context models effectively and thoroughly.
\newblock In \emph{The Thirteenth International Conference on Learning Representations}, 2025.
\newblock URL \url{https://openreview.net/forum?id=293V3bJbmE}.

\bibitem[Young et~al.(2024)Young, Chen, Li, Huang, Zhang, Zhang, Li, Zhu, Chen, Chang, et~al.]{young2024yi}
A.~Young, B.~Chen, C.~Li, C.~Huang, G.~Zhang, G.~Zhang, H.~Li, J.~Zhu, J.~Chen, J.~Chang, et~al.
\newblock Yi: Open foundation models by 01. ai.
\newblock \emph{arXiv preprint arXiv:2403.04652}, 2024.

\bibitem[Yu et~al.(2025)Yu, Zhang, Zhu, Yuan, Zuo, Yue, Dai, Fan, Liu, Liu, et~al.]{yu2025dapo}
Q.~Yu, Z.~Zhang, R.~Zhu, Y.~Yuan, X.~Zuo, Y.~Yue, W.~Dai, T.~Fan, G.~Liu, L.~Liu, et~al.
\newblock Dapo: An open-source llm reinforcement learning system at scale.
\newblock \emph{arXiv preprint arXiv:2503.14476}, 2025.

\bibitem[Yue et~al.(2025)Yue, Chen, Lu, Zhao, Wang, Yue, Song, and Huang]{yue2025limit-of-rlvr}
Y.~Yue, Z.~Chen, R.~Lu, A.~Zhao, Z.~Wang, Y.~Yue, S.~Song, and G.~Huang.
\newblock Does reinforcement learning really incentivize reasoning capacity in llms beyond the base model?
\newblock \emph{arXiv preprint arXiv:2504.13837}, 2025.

\bibitem[Zellers et~al.(2019)Zellers, Holtzman, Bisk, Farhadi, and Choi]{zellers-etal-2019-hellaswag}
R.~Zellers, A.~Holtzman, Y.~Bisk, A.~Farhadi, and Y.~Choi.
\newblock {H}ella{S}wag: Can a machine really finish your sentence?
\newblock In A.~Korhonen, D.~Traum, and L.~M{\`a}rquez, editors, \emph{Proceedings of the 57th Annual Meeting of the Association for Computational Linguistics}, pages 4791--4800, Florence, Italy, July 2019. Association for Computational Linguistics.
\newblock \doi{10.18653/v1/P19-1472}.
\newblock URL \url{https://aclanthology.org/P19-1472}.

\bibitem[Zeng et~al.(2025{\natexlab{a}})Zeng, Yang, Zhang, Yu, Wang, Liu, Sun, and Liu]{zeng2025acecoder}
H.~Zeng, J.~Yang, Y.~Zhang, B.~Yu, S.~Wang, Z.~Liu, M.~Sun, and T.~Liu.
\newblock Acecoder: Acing coder rl via automated test-case synthesis.
\newblock \emph{arXiv preprint arXiv:2502.01718}, 2025{\natexlab{a}}.
\newblock URL \url{https://arxiv.org/abs/2502.01718}.

\bibitem[Zeng et~al.(2025{\natexlab{b}})Zeng, Ivison, Wang, Yuan, Li, Ye, Li, He, Zhou, Chen, Zhao, Tsvetkov, Du, Jaques, Peng, Koh, and Hajishirzi]{zeng2025rlve}
Z.~Zeng, H.~Ivison, Y.~Wang, L.~Yuan, S.~S. Li, Z.~Ye, S.~Li, J.~He, R.~Zhou, T.~Chen, C.~Zhao, Y.~Tsvetkov, S.~S. Du, N.~Jaques, H.~Peng, P.~W. Koh, and H.~Hajishirzi.
\newblock Rlve: Scaling up reinforcement learning for language models with adaptive verifiable environments.
\newblock \emph{arXiv preprint 2511.07317}, 2025{\natexlab{b}}.

\bibitem[Zha et~al.(2023)Zha, Zhou, Li, Wang, Huang, Yang, Yuan, Su, Li, Su, Zhang, Zhou, Shou, Wang, Zhu, Lu, Ye, Ye, Ye, Zhang, Deng, Xu, Wang, Chen, and Zhao]{zha2023tablegpt}
L.~Zha, J.~Zhou, L.~Li, R.~Wang, Q.~Huang, S.~Yang, J.~Yuan, C.~Su, X.~Li, A.~Su, T.~Zhang, C.~Zhou, K.~Shou, M.~Wang, W.~Zhu, G.~Lu, C.~Ye, Y.~Ye, W.~Ye, Y.~Zhang, X.~Deng, J.~Xu, H.~Wang, G.~Chen, and J.~Zhao.
\newblock Tablegpt: Towards unifying tables, natural language and commands into one gpt.
\newblock \emph{arXiv preprint arXiv:2307.08674}, 2023.
\newblock URL \url{https://arxiv.org/abs/2307.08674}.

\bibitem[Zhao et~al.(2024{\natexlab{a}})Zhao, Ren, Hessel, Cardie, Choi, and Deng]{zhao2024wildchat}
W.~Zhao, X.~Ren, J.~Hessel, C.~Cardie, Y.~Choi, and Y.~Deng.
\newblock Wildchat: 1m chatgpt interaction logs in the wild.
\newblock \emph{arXiv preprint arXiv:2405.01470}, 2024{\natexlab{a}}.

\bibitem[Zhao et~al.(2023)Zhao, Gu, Varma, Luo, Huang, Xu, Wright, Shojanazeri, Ott, Shleifer, Desmaison, Balioglu, Damania, Nguyen, Chauhan, Hao, Mathews, and Li]{zhao2023pytorchfsdpexperiencesscaling}
Y.~Zhao, A.~Gu, R.~Varma, L.~Luo, C.-C. Huang, M.~Xu, L.~Wright, H.~Shojanazeri, M.~Ott, S.~Shleifer, A.~Desmaison, C.~Balioglu, P.~Damania, B.~Nguyen, G.~Chauhan, Y.~Hao, A.~Mathews, and S.~Li.
\newblock Pytorch fsdp: Experiences on scaling fully sharded data parallel, 2023.
\newblock URL \url{https://arxiv.org/abs/2304.11277}.

\bibitem[Zhao et~al.(2024{\natexlab{b}})Zhao, Qu, Staniszewski, Tworkowski, Liu, Miłoś, Wu, and Minervini]{zhao2024interdoc}
Y.~Zhao, Y.~Qu, K.~Staniszewski, S.~Tworkowski, W.~Liu, P.~Miłoś, Y.~Wu, and P.~Minervini.
\newblock Analysing the impact of sequence composition on language model pre-training.
\newblock In \emph{Proceedings of the 62nd Annual Meeting of the Association for Computational Linguistics (Volume 1: Long Papers)}, page 7897–7912. Association for Computational Linguistics, 2024{\natexlab{b}}.
\newblock \doi{10.18653/v1/2024.acl-long.427}.
\newblock URL \url{http://dx.doi.org/10.18653/v1/2024.acl-long.427}.

\bibitem[Zheng et~al.(2023)Zheng, Chiang, Sheng, Zhuang, Wu, Zhuang, Lin, Li, Li, Xing, et~al.]{zheng2023judging}
L.~Zheng, W.-L. Chiang, Y.~Sheng, S.~Zhuang, Z.~Wu, Y.~Zhuang, Z.~Lin, Z.~Li, D.~Li, E.~Xing, et~al.
\newblock Judging llm-as-a-judge with mt-bench and chatbot arena.
\newblock \emph{Advances in Neural Information Processing Systems}, 36:\penalty0 46595--46623, 2023.

\bibitem[Zhong et~al.(2023)Zhong, Cui, Guo, Liang, Lu, Wang, Saied, Chen, and Duan]{zhong2023agieval}
W.~Zhong, R.~Cui, Y.~Guo, Y.~Liang, S.~Lu, Y.~Wang, A.~Saied, W.~Chen, and N.~Duan.
\newblock Agieval: A human-centric benchmark for evaluating foundation models.
\newblock \emph{arXiv preprint arXiv:2304.06364}, 2023.

\bibitem[Zhou et~al.(2025)Zhou, Wang, Ranjan, Cheng, Tang, He, Liu, and Xing]{zhou2025megamath}
F.~Zhou, Z.~Wang, N.~Ranjan, Z.~Cheng, L.~Tang, G.~He, Z.~Liu, and E.~P. Xing.
\newblock Megamath: Pushing the limits of open math corpora.
\newblock \emph{arXiv preprint arXiv:2504.02807}, 2025.

\bibitem[Zhou et~al.(2023)Zhou, Lu, Mishra, Brahma, Basu, Luan, Zhou, and Hou]{zhou2023instructionfollowingevaluationlargelanguage}
J.~Zhou, T.~Lu, S.~Mishra, S.~Brahma, S.~Basu, Y.~Luan, D.~Zhou, and L.~Hou.
\newblock Instruction-following evaluation for large language models, 2023.
\newblock URL \url{https://arxiv.org/abs/2311.07911}.

\bibitem[Zhuo et~al.(2024)Zhuo, Vu, Chim, Hu, Yu, Widyasari, Yusuf, Zhan, He, Paul, et~al.]{zhuo2024bigcodebench}
T.~Y. Zhuo, M.~C. Vu, J.~Chim, H.~Hu, W.~Yu, R.~Widyasari, I.~N.~B. Yusuf, H.~Zhan, J.~He, I.~Paul, et~al.
\newblock Bigcodebench: Benchmarking code generation with diverse function calls and complex instructions.
\newblock \emph{arXiv preprint arXiv:2406.15877}, 2024.

\end{thebibliography}

\clearpage

\appendix
\section{Appendix}

\section*{Author Contributions}
\label{sec:contrib}

A successful team project like \olmothree{} would not be possible without the contributions of many teammates. 
We indicate each authors' main contributing role(s) in \olmothree, while recognizing that project impact was driven by fluid contributions across formal team boundaries.
Authors are listed in alphabetical order: 

\begin{itemize}

    \item For model architecture, infrastructure, and training methodology: Akshita Bhagia, Aman Rangapur, Amanda Bertsch, David Heineman, Dirk Groeneveld, Dustin Schwenk, Kyle Lo, Luca Soldaini, Mayee Chen, Pete Walsh, Shane Arora, Tyler Murray, Tyler Romero, Will Merrill

    \item For post-training infrastructure and training methodology: Costa Huang, Faeze Brahman, Finbarr Timbers, Hamish Ivison, Jacob Morrison, Michael Noukhovitch, Nathan Lambert, Pradeep Dasigi, Saurabh Shah, Scott Geng, Shannon Zejiang Shen, Shashank Gupta, Teng Xiao, Tyler Romero, Valentina Pyatkin, Victoria Graf

    \item For base model data acquisition: Chloe Anastasiades, David Graham, Dustin Schwenk, Jake Poznanski, Jaron Lochner, Kyle Lo, Luca Soldaini, Matt Jordan, Robert Berry, Tyler Murray

    \item For data curation infrastructure and experimentation: Alexander Wettig, Allyson Ettinger, Amanda Bertsch, Bailey Kuehl, David Heineman, Ian Magnusson, Jake Poznanski, Jiacheng Liu, Kyle Lo, Luca Soldaini, Matt Jordan, Mayee Chen, Tyler Murray, Tyler Romero

    \item For evaluation methodology and infrastructure: Akari Asai, Alexander Wettig, David Heineman, Dustin Schwenk, Hamish Ivison, Harsh Trivedi, Ian Magnusson, Kyle Lo, Luca Soldaini, Maarten Sap, Malia Morgan, Pradeep Dasigi, Regan Huff, Robert Berry, Ronan Le Bras, Rulin Shao, Saumya Malik, Saurabh Shah, Shannon Zejiang Shen, Shashank Gupta, Tyler Murray, Victoria Graf, Yuling Gu

    \item For mid- and post-training data curation and experimentation: Akari Asai, Alisa Liu, Allyson Ettinger, David Graham, David Heineman, Faeze Brahman, Hamish Ivison, Harsh Trivedi, Jacob Morrison, Kyle Lo, Lester James V. Miranda, Luca Soldaini, Matt Jordan, Michael Noukhovitch, Nathan Lambert, Pradeep Dasigi, Rui Xin, Saurabh Shah, Scott Geng, Saumya Malik, Shashank Gupta, Shuyue Stella Li, Teng Xiao, Valentina Pyatkin, Victoria Graf, Yapei Chang, Zhiyuan Zeng

    \item For compute infrastructure setup and support: Michael Schmitz, Michael Wilson, Michal Guerquin, Sam Skjonsberg, Tucker Wilde

    \item For mentorship, advising, program management, and broader strategy: Ali Farhadi, Ashish Sabharwal, Hannaneh Hajishirzi, Luke Zettlemoyer, Noah A. Smith, Pang Wei Koh, Taira Anderson

    \item For technical leadership and cross-workstream contributions: Hannaneh Hajishirzi, Kyle Lo, Luca Soldaini, Nathan Lambert, Pradeep Dasigi
\end{itemize}

Authorship for this work was determined by those making direct contributions to the \olmothree models, related artifacts, and their release. 
Core contributors are recognized for their sustained, significant contributions critical to the success of the \olmothree project.

\section*{Acknowledgments}

This research used resources of the Oak Ridge Leadership Computing Facility, which is a DOE Office of Science User Facility supported under Contract DE-AC05-00OR22725. We acknowledge the National Artificial Intelligence Research Resource (NAIRR) Pilot and Microsoft Azure for contributing to the results in this work. We are grateful for feedback throughout our development process from the open source language model developer community, especially those from Common Pile/Comma, SmolLM3, Marin, Apertus and Gaperon.

\subsection{Base Model Additional Training Details}

Table~\ref{tab:training-config} summarizes modeling configuration for \olmothree~7B and \olmothree~32B. 
Table~\ref{tab:training_stages_7b_32b} provides overview of training hyperparameters during the three stages of base model development: pretraining, midtraining, and long-context extension. 
Table~\ref{tab:training-config} describes parallelism configuration for the stages, and lists measured throughput in tokens per second (TPS) for each. 
Finally, Figure~\ref{fig:7B_pretrain_loss} shows training cross entropy loss and gradient norm for \olmothreebase 7B and 32B during the pretraining stage.

\begin{table}[!h]
    \centering
    \begin{small}
    \begin{tabular}{lclc}
        \toprule
        \rowcolor{midgrey}{Layers} & \textcolor[HTML]{d3327e}{\bf{32}} / \textcolor[HTML]{00ab72}{\bf{64}} & {Gradient clipping} & 1.0 \\
        {Hidden size~$(d_\text{model})$}  & \textcolor[HTML]{d3327e}{\bf{4096}} / \textcolor[HTML]{00ab72}{\bf{5120}} & {Z-loss weight} & $10^{-5}$ \\
        \rowcolor{midgrey}{Q heads} & \textcolor[HTML]{d3327e}{\bf{32}} / \textcolor[HTML]{00ab72}{\bf{40}} & {Weight decay on embeddings} & No \\
        {KV heads} & \textcolor[HTML]{d3327e}{\bf{32}} / \textcolor[HTML]{00ab72}{\bf{8}} & {Sliding window attention} & 3/4 of layers; 4{,}096 tokens \\
        \rowcolor{midgrey}{Activation} & SwiGLU & {RoPE scaling} & YaRN on full attn. layers \\
        {QKV normalization} & QK-Norm & {RoPE $\theta$} & $5 \cdot 10^{5}$ \\
        \rowcolor{midgrey}{Layer norm} & RMSNorm & {Layer norm applied to} & Outputs \\
        \bottomrule
    \end{tabular}
    \end{small}
    \caption{\textbf{Model architecture for \textcolor[HTML]{d3327e}{\olmothree~7B} and \textcolor[HTML]{00ab72}{\olmothree~32B}.} The 7B model uses multi-head attention, while the 32B model uses grouped-query attention~\citep{ainslie2023gqatraininggeneralizedmultiquery} for increased efficiency.}
    \label{tab:combined_model_specs}
\end{table}

\begin{table}[!h]
    \centering
    \begin{small}
    \begin{tabular}{lccc}
        \toprule
        \rowcolor{ai2lightpink}  \textbf{\olmothreebase 7B} & \textbf{Pretraining} & \textbf{Midtraining} & \textbf{Long-context ext} \\[2pt]
        {DP-rep} & 64 & 16 & 32 \\
        \rowcolor{midgrey} {DP-shard} & 8 & 8 & - \\
        {CP} & - & - & 8 \\
        \rowcolor{midgrey} {Num devices} & 512 & 128 & 256 \\
        {Throughput (TPS/device)} & 7.7K & 8.5K & 4.0K \\[2pt]
        
        \midrule

        \rowcolor{ai2lightpink}  \textbf{\olmothreebase 32B} & \textbf{Pretraining} & \textbf{Midtraining} & \textbf{Long-context ext} \\[2pt]
        {DP-rep} & 16 & 8 & 16 \\
        \rowcolor{midgrey} {DP-shard} & 64 & 64 & 8 \\
        {CP} & - & - & 8 \\
        \rowcolor{midgrey} {Num devices} & 1024 & 512 & 1024 \\
        {Throughput (TPS/device)} & 2.0K & 2.0K & 1.3K \\

        \bottomrule
    \end{tabular}
    \end{small}
    \caption{\textbf{Training configuration and throughput for \olmothreebase models} across different training stages. DP-shard refers to the sharding dimension for Hybrid-Sharded Data Parallelism (HSDP)~\citep{zhao2023pytorchfsdpexperiencesscaling}, DP-rep refers to the replication dimension, and CP refers to Llama3-style context parallelism \citep{scalingllama3}. We train on a cluster containing 8$\times$ NVIDIA H100 (80GB HBM3) nodes, connected via TCPXO (200 Gbps/GPU). Throughput numbers reflect the end of each phase, as, in some cases, we made improvements while the runs were ongoing.}
    \label{tab:training-config}
\end{table}

\begin{table}[!h]
    \centering
    \begin{small}
    \begin{tabular}{lccc}
        \toprule
\rowcolor{ai2lightpink}  {\bf \olmothreebase 7B} & {\bf Pretraining} & {\bf Midtraining} & {\bf Long-context ext} \\[2pt]
{Learning Rate Schedule} & \begin{tabular}{c}Modified cosine\\[-3pt] (see Figure~\ref{fig:pretrain:7b-lr-and-loss})\end{tabular}  & Linear decay & Linear decay \\
        \rowcolor{midgrey}{LR warmup from 0} & 2000 steps & 0 steps & 200 steps \\
        {Peak LR} & $3.0 \times 10^{-4}$ & $2.074 \times 10^{-4}$ & $2.074 \times 10^{-4}$ \\
        \rowcolor{midgrey} {Final LR} & $3.0 \times 10^{-5}$ & 0 & 0 \\
        {Batch size (\# instances)} & 512 & 256 & 64 \\
        \rowcolor{midgrey} {Sequence length} & 8{,}192 & 8{,}192 & 65{,}536 \\
        {Batch size (\# tokens)} & 4{,}194{,}304 & 2{,}097{,}152 & 4{,}194{,}304 \\
        \rowcolor{midgrey} {Total training tokens} & 5.93T & 100B & 50B \\[1pt]
        {Peak training temperature} ($\tfrac{\text{LR}^{2}}{\textit{bsz}}$) & $2.146 \times 10^{-14}$ & $2.051 \times 10^{-14}$ & $1.026 \times 10^{-14}$ \\[2pt]
        \midrule
\rowcolor{ai2lightpink}  {\bf \olmothreebase 32B} & {\bf Pretraining} & {\bf Midtraining} & {\bf Long-context ext} \\[2pt]
{Learning rate schedule} & 
        \begin{tabular}{c}5.93T cosine trunc.\\[-3pt] at 5.5T tokens\end{tabular} & Linear decay & Linear decay \\
        \rowcolor{midgrey}{LR warmup from 0} & 2000 steps & 0 steps & 200 steps \\
        {Peak LR} & $6.0 \times 10^{-4}$ & $2.071 \times 10^{-4}$ & $2.071 \times 10^{-4}$ \\
        \rowcolor{midgrey} {Final LR} & $6.0 \times 10^{-5}$ & 0 & 0 \\
        {Batch size (\# instances)} & 1{,}024 & 512 & 128 \\
        \rowcolor{midgrey} {Sequence length} & 8{,}192 & 8{,}192 & 65{,}536 \\
        {Batch size (\# tokens)} & 8{,}388{,}608 & 4{,}194{,}304 & 8{,}388{,}608 \\
        \rowcolor{midgrey} {Total training tokens} & 5.5T & 100B (twice) & 100B \\[1pt]
        {Peak training temperature} ($\tfrac{\text{LR}^{2}}{\textit{bsz}}$) & $4.292 \times 10^{-14}$ & $1.023 \times 10^{-14}$ & $5.113 \times 10^{-15}$ \\[2pt]
        \bottomrule
    \end{tabular}
    \end{small}
    \caption{\textbf{Training hyperparameters for each stage} of \olmothreebase 7B and 32B. Compared to the 7B, for the 32B we use a cosine learning rate schedule (truncated early at 5.5T tokens), double the batch size in all steps, run midtraining twice (with different data order seeds, and average model weights of resulting checkpoints), and increase the long-context extension stage from 50B to 100B tokens.}
    \label{tab:training_stages_7b_32b}
\end{table}

\begin{figure}[h!]
    \centering
    \adjustbox{max width=.49\linewidth}{\includegraphics{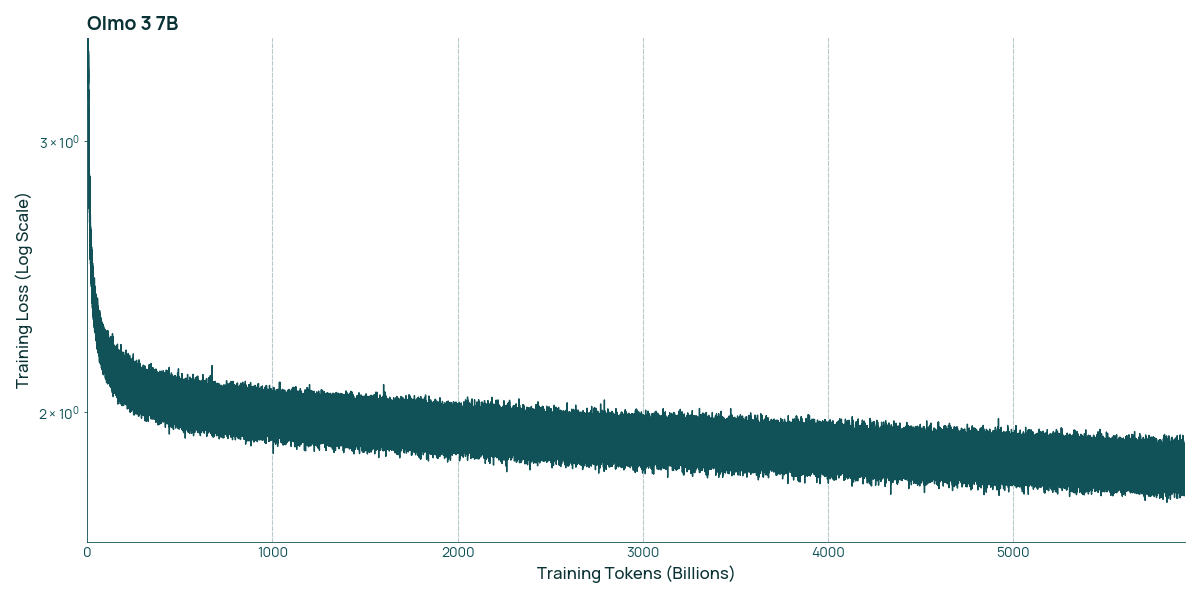}}
    \adjustbox{max width=.49\linewidth}{\includegraphics{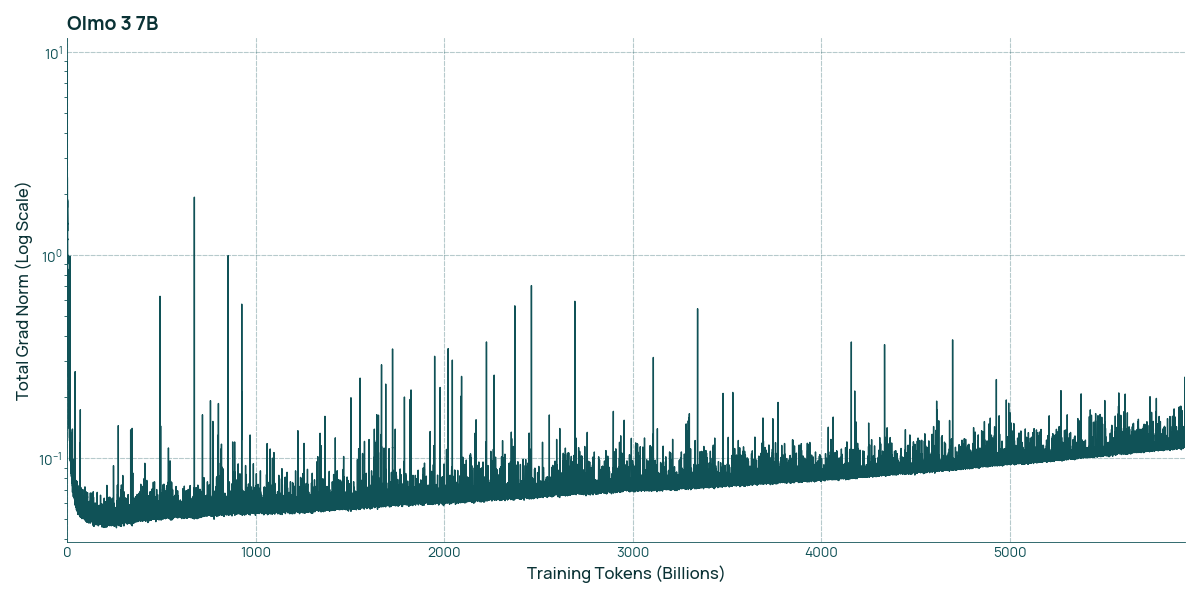}}
    \adjustbox{max width=.49\linewidth}{\includegraphics{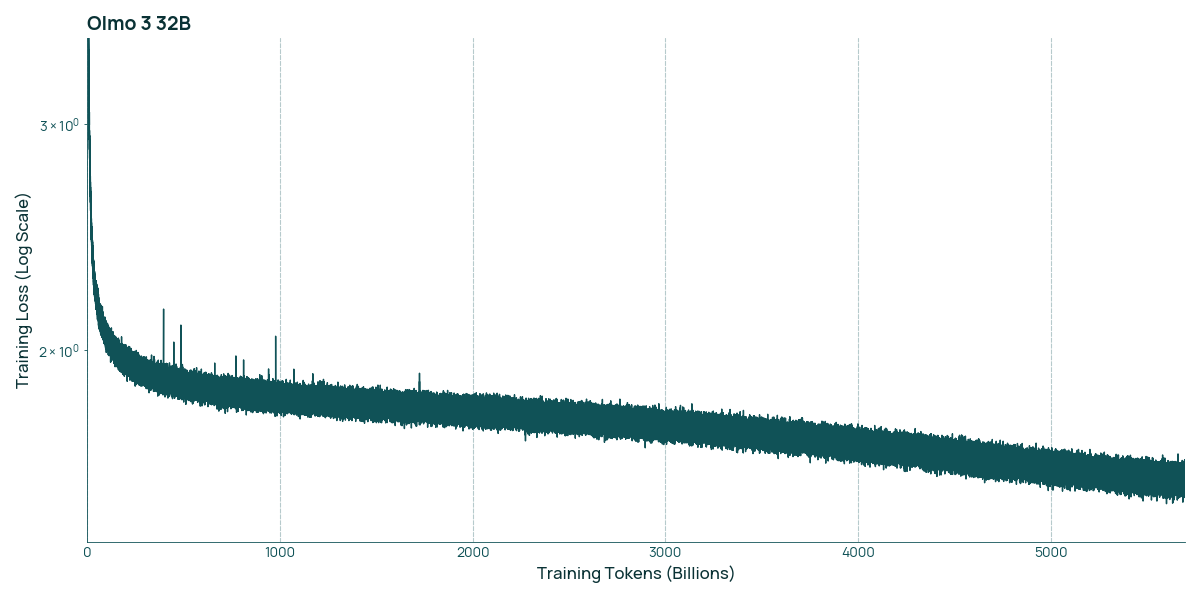}}
    \adjustbox{max width=.49\linewidth}{\includegraphics{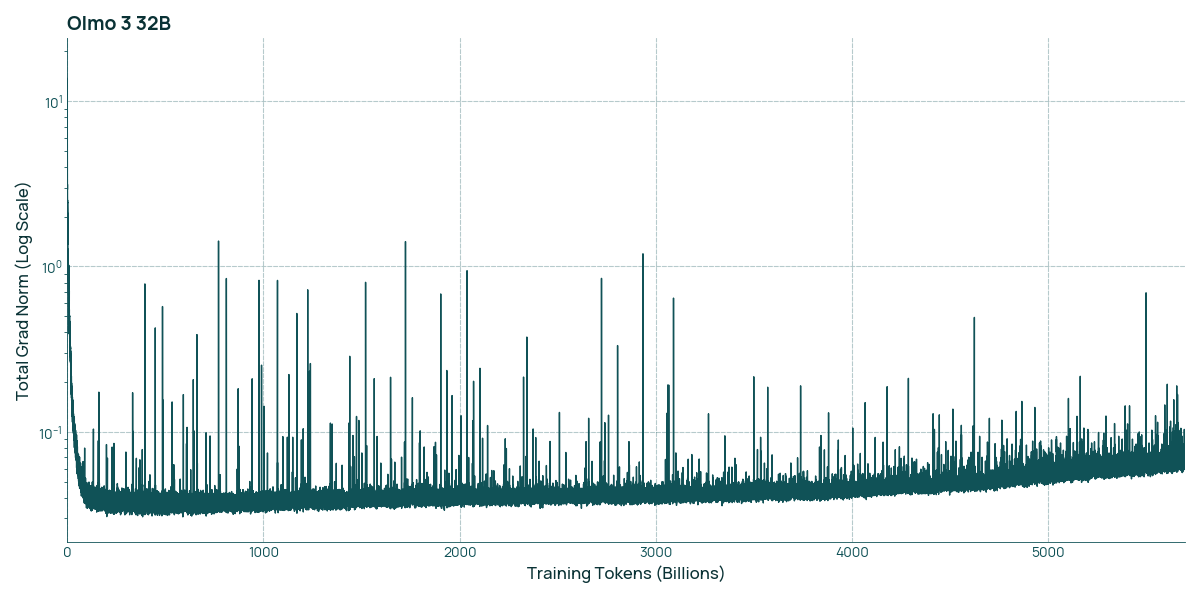}}
    \caption{\textbf{Cross-entropy loss and total gradient norm during pretraining} for \olmothreebase 7B (top) and 32B (bottom). For readability, gradient norm plots were produced using an exponential moving average with a window size of 20 steps.}
    \label{fig:7B_pretrain_loss}
\end{figure}

\FloatBarrier

\subsection{Base Model Additional Data Details: Pretraining}
\label{sec:appendixpretrain}

\subsubsection{CommonCrawl}
The majority of our pretraining corpus comes from CommonCrawl ~\citep{CommonCrawl}. We start with 104 dumps, starting with \texttt{CC-MAIN-2013-20} and ending with \texttt{CC-MAIN-2024-51}, roughly covering dates from mid-2013 until late 2024. We linearize the WET files provided by Commoncrawl using Resiliparse, yielding an initial pool composed of 252.6B documents.

Next we apply a pipeline of heuristic filtering steps to further prune down the dataset to a size amenable for pretraining. Our steps essentially follow those of DCLM ~\citep{dclm}, with a few small differences. We start with URL-based filtering, identifying and removing documents that have URLs containing banned words or subwords from the blacklists used by FineWeb \citep{penedo2024fineweb} and RefinedWeb ~\citep{penedo2023refinedwebdatasetfalconllm}. This step removes roughly 1\% of the data pool. Then we apply the DCLM collection of heuristic filters, roughly targeting and removing: i) very short documents, ii) very long documents, iii) documents with not enough alphanumeric characters, and iv) documents with large amounts of internal repetition. Next, we modify and remove any lines or paragraphs in each document that have i) too many numeric characters or ii) any boilerplate phrases such as "items in cart" or "read more...", and then we fully remove any documents that have been obliterated by these line-specific removals. We then apply a FastText English language filter, mirroring DCLM and using a threshold of 0.65 to identify documents as containing English text. Finally, we apply a subset of the rules for identifying questionable sentences from MADLAD-400 ~\citep{kudugunta2023madlad400multilingualdocumentlevellarge}. Ablation tests show that only rules 2 and 5 from MADLAD improve dataset quality, targeting sentences that have a large number of capitalized words or contain a "cursed regex". If the number of sentences in the document is less than 5 or if at least 20\% of sentences are questionable, we remove the document from the corpus.

Overall, the heuristic steps remove 76\% of the total pool, and the English filtering step removes an additional 2.5\% of the pool. This leaves a pool of 38.7B documents, attaining a survival rate of 15.1\%. While each of these described steps is incorporated into the DCLM processing pipeline, we note that these heuristic filters are commutative and that the English filtering is the slowest step, so efficiency gains can be attained by putting the language-filtering step at the end. We spent a total of 1030 i4i.32xlarge EC2 hours in this step, incurring a cost of approximately \$11,300. An exact breakdown of how much time was spent in each step is provided in Table \ref{tab:cc_pipeline_stats}.

\begin{table}[h]
\centering
\begin{tabular}{lrrr}
\toprule
{\bf Pipeline Step} & {\bf Docs Removed (B)} & {\bf \% of pool removed} & {\bf \% of total time} \\
\midrule
\rowcolor{ai2offwhite} URL Filters & 2.3 & 0.9 & 1.68 \\
Length Filters & 103.4 & 40.42 & 8.03 \\
\rowcolor{ai2offwhite} Symbol Filters & 56.5 & 22.1 & 4.13 \\
Internal Repetition & 32.1 & 12.53 & 31.41 \\
\rowcolor{ai2offwhite} Line Modifiers & 7.1 & 2.79 & 10.0 \\
English Filter & 6.2 & 2.44 & 14.3 \\
\rowcolor{ai2offwhite} MadLad Filters & 9.3 & 3.65 & 5.87 \\
Quality Classifiers & 0.0 & 0.0 & 24.58 \\
\bottomrule
\end{tabular}
\caption{\textbf{Web data processing step cost and removal breakdown} during the heuristic processing steps. We started with 252.6B documents and ended with 38.7B documents for a total removal rate of 84.9\%. This procedure took, in aggregate, approximately 1,030 hours on i4i.32xlarge EC2 instances.}
\label{tab:cc_pipeline_stats}
\end{table}

\subsubsection{Deduplication}

As described in the main paper, we apply a three-stage deduplication pipeline to our dataset, with each stage targeting progressively more nuanced forms of redundancy: (i) global exact deduplication based on document content hashes to remove identical copies, (ii) 32-way sharded MinHash deduplication with exact Jaccard similarity verification to remove near-duplicate documents, and (iii) 56-way sharded fuzzy suffix array deduplication to eliminate repeated boilerplate text. We note that while applying exact deduplication before MinHash deduplication is technically redundant, exact deduplication is substantially more efficient computationally; hence this two-pass approach is much faster overall. For the exact and MinHash deduplication stages, we utilize the Duplodocus tool,\footnote{\href{https://github.com/allenai/duplodocus}{\path{github.com/allenai/duplodocus}}} and for the suffix array deduplication stage, we employ \texttt{bsade}.\footnote{\href{https://github.com/liujch1998/bsade/}{\path{github.com/liujch1998/bsade}}}

\paragraph{Exact Deduplication}

We perform exact deduplication in two sequential passes. During the heuristic filtration pipeline, we annotate each document with a 128-bit hash computed from the document text. We then apply an initial deduplication step to each of the 104 processed CommonCrawl dumps individually, arbitrarily retaining one copy of each document per dump. This within-dump deduplication removes 24\% of the surviving document pool.

Following this, we aggregate all documents globally and perform a second exact deduplication pass across the entire corpus, again arbitrarily keeping one copy of each document. This global pass removes an additional 43\% of the surviving pool. In total, exact deduplication eliminates 66\% of the input documents, reducing the corpus to 12.7 billion documents for subsequent MinHash processing.

\paragraph{MinHash~Fuzzy deduplication}

We partition the 12.7 billion document corpus resulting from exact deduplication into 32 shards of approximately equal size and perform MinHash deduplication independently on each shard. Our MinHash procedure broadly follows the approach outlined in \citep{lee2022deduplicatingtrainingdatamakes}. We tokenize documents using the p50k tokenizer and construct sets of 5-gram token sequences. We then apply a MinHash locality-sensitive hashing scheme with 26 bands of size 11, configured to target a Jaccard similarity threshold of 0.80.

For any pair of documents that share at least one matching bucket, we treat them as connected by an edge in graph-theoretic terms. We construct a graph from the union of all such edges and identify connected components within this graph. Each document in a connected component is then annotated with a unique identifier for that component.

In a second verification phase, we explicitly compute pairwise Jaccard similarities within each MinHash-identified cluster to eliminate false positives. For this verification, we use 3-gram token sequences. Our approach varies based on cluster size: for connected components containing 500 or more documents, we apply a more stringent MinHash configuration using 200 bands of size 31; for components with fewer than 500 documents, we perform exhaustive pairwise Jaccard similarity checks and generate final duplicate clusters from these results.

After annotating all documents according to their true Jaccard similarity with other documents in the corpus, we retain only the most recent version of each document based on crawl date, removing all earlier duplicates. This complete MinHash deduplication procedure eliminates 24\% of the input documents, leaving 9.8 billion documents in the pool.

\paragraph{Suffix Array deduplication}

In the final deduplication stage, we employ suffix arrays to identify and remove substrings that appear repeatedly throughout the dataset. We partition the 9.8 billion document corpus into 56 shards of roughly equal size and run suffix array deduplication independently on each shard.

For each shard, we construct a suffix array and identify every byte sequence of length 500 or greater that appears at least twice in the shard. We then apply a novel ``fuzzy suffix array'' removal strategy that considers contiguous text spans within each document. Specifically, if a text span is bounded on both sides by 500-byte sequences that appear multiple times in the suffix array, and at least 80\% of the span is covered by such repeated sequences, we remove the entire span. This strategy targets cases where naive suffix array deduplication would leave short, unique fragments interspersed between removed substrings. For text that does not meet this bookended criterion, we remove all individual occurrences of repeated 500-byte sequences.

After these three rounds of deduplication—exact, MinHash, and suffix array—we arrive at a final corpus of 9.7 billion documents.

\subsubsection{Topic Classification}
\begin{table}[htbp]
\centering
\begin{tabular}{@{}lccc|lccc@{}}
\toprule
{\bf Category} & {\bf F1} & {\bf Prec.} & {\bf Rec.} & {\bf Category} & {\bf F1} & {\bf Prec.} & {\bf Rec.} \\
\midrule
\rowcolor{ai2offwhite} Finance and Business & 0.755 & 0.758 & 0.751 & Travel and Tourism & 0.781 & 0.780 & 0.782 \\
Home and Hobbies & 0.748 & 0.704 & 0.797 & Crime and Law & 0.735 & 0.747 & 0.724 \\
\rowcolor{ai2offwhite} Entertainment & 0.801 & 0.773 & 0.832 & Software & 0.666 & 0.696 & 0.639 \\
Sports and Fitness & 0.870 & 0.850 & 0.890 & Literature & 0.759 & 0.801 & 0.721 \\
\rowcolor{ai2offwhite} Politics & 0.788 & 0.786 & 0.790 & Games & 0.823 & 0.867 & 0.783 \\
Health & 0.822 & 0.824 & 0.820 & Transportation & 0.777 & 0.786 & 0.768 \\
\rowcolor{ai2offwhite} Education and Jobs & 0.706 & 0.789 & 0.638 & Religion & 0.808 & 0.833 & 0.785 \\
Science, Math and Technology & 0.679 & 0.665 & 0.693 & Electronics and Hardware & 0.743 & 0.730 & 0.757 \\
\rowcolor{ai2offwhite} Social Life & 0.628 & 0.609 & 0.649 & Software Development & 0.687 & 0.613 & 0.781 \\
Fashion and Beauty & 0.845 & 0.845 & 0.845 & Industrial & 0.710 & 0.691 & 0.731 \\
\rowcolor{ai2offwhite} Food and Dining & 0.878 & 0.860 & 0.896 & History and Geography & 0.630 & 0.698 & 0.574 \\
Art and Design & 0.670 & 0.668 & 0.672 & Adult Content & 0.700 & 0.894 & 0.575 \\
\midrule
\rowcolor{ai2lightpink} \multicolumn{8}{c}{{\bf Overall (N=20,000): Precision = 0.762, Recall = 0.762}} \\
\bottomrule
\end{tabular}
\caption{\textbf{Performance of FastText classifier distilled from WebOrganizer topic labels} on the held out sample of 20,000 documents used in the original WebOrganizer paper.}
\label{tab:wo_classifier_performance}
\end{table}

After strict rounds of deduplication, we partition our data according to topic using the 24 topic categories introduced in WebOrganizer~\citep{weborganizer}. Rather than using the 140M parameter topic classifier used by WebOrganizer, we train a FastText classifier\footnote{\href{https://huggingface.co/allenai/dolma3-fasttext-weborganizer-topic-classifier}{\path{huggingface.co/allenai/dolma3-fasttext-weborganizer-topic-classifier}}} to support cost-effective topic classification at scale. To train this classifier, we use the Llama-labeled training data used to train the original WebOrganizer category as well as an extra 506,746 examples with topics labeled by a combination of gpt-4.1 and o4-mini. The performance of this classifier is outlined in Table \ref{tab:wo_classifier_performance}.

\subsubsection{CommonCrawl Mixing}
\label{sec:commoncrawl-mixing}
We perform a hierarchical mixing procedure on our data. Our procedure \olmix~\citep{olmix} generates prescriptions for which percentage of the training mix should come from each topic or source, but offers no guidance on the quality composition within each topic. While prior works, such as DCLM~\citep{dclm} use a quality classifier to flatly filter data as high-quality or not, we take a more fine-grained approach and perform selective up and down-sampling within each WebOrganizer topic depending on the quality signal. This section formalizes the search procedure we use to generate these upsampling curves.

\paragraph{Problem formulation} We discuss this procedure in more general terms: consider a category with $X$ tokens, partitioned into $Q$ strictly ordered quality buckets, where the $q^{th}$ bucket contains $X_q$ tokens. Further assume that \olmix prescribes that $Z$ tokens be taken from this category, and that at no point do we want to upsample any quality bucket more than $M$ times. This equates to a search problem, where we need to take $Z_q$ tokens from the $q^{th}$ bucket such that $\sum_q Z_q= Z$ and $\forall q,\; Z_q/X_q \leq M$.

\paragraph{Parameterizing the solution space} To reduce the dimensionality of this search space, we make a modeling choice, where we search over a family of functions that control the upsampling ratio that meets the following criteria:
\begin{itemize}
    \item Every function in the family is convex and monotonic.
    \item The functions are defined on the interval $[0,1]$, which can be normalized to the token counts later.
    \item We are able to control the integral such that $\int_0^1f(x)dx = Z/X$.
    \item We can control the maximum average value of any one bucket. Suppose the $q^{th}$ bucket of data is arranged on the $x$-axis from $[a,b]$, then the maximum upsampling constraint correlates to the inequality $\frac{1}{b-a} \int_a^b f(x)dx \leq M$.
    \item We have the option to filter out the lowest quality buckets, i.e. $\int_0^a f(x)dx = 0.$
\end{itemize}

One such family of functions that meets these criteria is the family of truncated power-exponential functions:
\[
f_{p, \lambda}(x) =
  \begin{cases}
    0, & \text{for } x < a \\
    C(x-a)^p\cdot e^{\lambda(x-a)}, & \text{for } x \geq a\\
  \end{cases}
  \]
Specifically, this becomes a feasibility problem for each topic of the data, where we search over parameters $p, \lambda, C$ such that the constraints \begin{itemize}
    \item (Token yield is satisfied) $\int_0^1 f_{p,\lambda}(x)dx = Z/X$.
    \item (Maximum upsampling ratio is honored) $\frac{1}{b}\int_{1-b}^1 f_{p, \lambda}(x) dx \leq M.$
    \item (Function is monotonic) $\lambda \geq 0.$
\end{itemize}
are satisfied. The maximum upsampling constraint has been simplified such that, assuming monotonicity, the most upsampled quality bucket would be the highest-quality one, with an assumed data proportion of $b$.

\paragraph{Implementation details} For each WebOrganizer topic, we set the maximum upsampling ration to be $M=7$ and also throw away the bottom $40\%$ in terms of quality, $a=0.40$. Then we numerically solve for feasible $p, \lambda, C$. If the $q^{th}$ quality bucket spans from the $q^-$ percentile to the $q^+$ percentile of the data, then the upsampling ratio for this bucket of data should be $\frac{1}{q^+-q^-} \int_{q^-}^{q^+} f(x)dx$.

\subsubsection{Validating Quality Upsampling and Mixing}

We validate our quality upsampling curves and mixing methodology both individually and jointly using small-scale 1B parameter models trained on 100B tokens. Our validation consists of three experiments:

\paragraph{Targeted mixing} We first verify that our mixing methodology can successfully optimize for specific prediction targets. Using our swarm optimization procedure, we create mixes optimized for three different objectives: the QA average, Math average, and Code average from \olmothreeeval. We compare these targeted mixes against both the natural data distribution and the final \olmothree mix. Table \ref{tab:token-constrained-mixing} demonstrates that our swarm optimization successfully adapts the data distribution to match specific capability targets. While the final \olmothreeeval mix exhibits slightly higher (worse) BPB scores than task-specific mixes due to necessary trade-offs across multiple objectives, it substantially outperforms the natural distribution.

\paragraph{Quality-aware upsampling} Next, we demonstrate that quality-aware upsampling outperforms naive quality-based filtering. To simulate a data-constrained 4.51T token training run, we compare different data selection strategies in Table \ref{tab:quality-aware-upsampling}. For the filtering baselines, we select the top percentiles from our vigintile quality buckets and match the resulting repetition factor that would occur when training on 100B tokens drawn from a theoretical 4.51T pool. For the upsampling approach, we apply the same methodology but set the target pool size to 100B tokens directly. Our results show that quality-aware upsampling consistently outperforms flat filtering across all repetition factors.

\paragraph{Reconciling upsampling and mixing} Finally, we evaluate how to best combine our mixing and upsampling methodologies, which address complementary aspects of data selection. Data mixing determines the distribution across topics, while quality upsampling determines the distribution within a single source. To conceptualize this, imagine the dataset as a two-dimensional matrix of buckets where rows represent WebOrganizer topics and columns represent the quality buckets. Then the mixing strategy can be thought of as imposing row-wise (topic) constraints only. The quality-aware upsampling experiments in the preceding paragraph impose column-wise (quality) constraints only.

We considered several techniques that did not work quite as well as the truncated power-exponential forms described in \S~\ref{sec:commoncrawl-mixing}. On one hand, the \olmix framework samples data from each topic (row) according only to the natural quality distribution. On the other, quality upsampling samples data from each quality bucket (column) and does not consider reweighting topic distributions. For a theoretical target token yield, each of these strategies prescribes a target token count to be taken from each \texttt{(topic, quality)} bucket. Naive ways to rectify these strategies is to take an arithmetic or geometric mean between the target token counts from each bucket. We also note that the theoretical framework defining upsampling curves above is not necessarily restricted to the concept class of truncated power-exponential families. We could just as easily consider the family of exponential functions like $f_\lambda(x) = Ce^{\lambda(x-a)}$. Upon considering each of these techniques on small 1B models, we found that the truncated power-exponential family performed the best. Results are contained in Table \ref{tab:resolving-upsampling-mixing}.

\begin{table*}[tbp]
\centering

\begin{minipage}{0.48\linewidth}
\centering
\small
\begin{tabular}{@{}l ccc@{}} %
\toprule
&
{$\textbf{\fontsize{8}{8}\selectfont~QA}_\textbf{\fontsize{6}{6}\selectfont~Easy}$} &
{$\textbf{\fontsize{8}{8}\selectfont~Math}_\textbf{\fontsize{6}{6}\selectfont~Easy}$} &
{$\textbf{\fontsize{8}{8}\selectfont~Code}_\textbf{\fontsize{6}{6}\selectfont~Easy}$} \\
\midrule
\rowcolor{lightgrey} Natural Distribution & 1.017 & 0.719 & 0.592 \\
\rowcolor{lightgrey} QA-heavy Mix         & {\bf{0.972}} & 0.643 & 0.535 \\
\rowcolor{lightgrey} Math-heavy Mix       & 0.979 & {\bf{0.586}} & 0.497 \\
\rowcolor{lightgrey} Code-heavy Mix       & {0.986} & 0.619 & {\bf{0.481}} \\
\rowcolor{ai2lightpink} \olmix            & {0.995} & {0.617} & {0.489} \\
\bottomrule
\end{tabular}
\caption{
\textbf{Token-constrained mixing allows optimizing different evaluation objectives}. We use our swarms to optimize a QA-, Math- and Code-heavy data mix and train 1B models to 100B tokens. Results are on the \olmothreeeval Easy suite. Scores are expressed in bits-per-byte (BPB), lower is better (see Section~\S\ref{sec:experimental-design} for details).
}
\label{tab:token-constrained-mixing}
\end{minipage}
\hfill
\begin{minipage}{0.48\linewidth}
\centering
\small
\begin{tabular}{@{}l ccc@{}}
\toprule
&
{$\textbf{\fontsize{8}{8}\selectfont~QA}_\textbf{\fontsize{6}{6}\selectfont~Easy}$} &
{$\textbf{\fontsize{8}{8}\selectfont~Math}_\textbf{\fontsize{6}{6}\selectfont~Easy}$} &
{$\textbf{\fontsize{8}{8}\selectfont~Code}_\textbf{\fontsize{6}{6}\selectfont~Easy}$} \\
\midrule
\rowcolor{lightgrey} Top 50\% (1.1x repeat) & 1.042 & 0.863 & 0.943 \\
\rowcolor{lightgrey} Top 30\% (1.8x repeat) & 1.031 & 0.870 & 0.880 \\
\rowcolor{lightgrey} Top 10\% (5.6x repeat) & 1.041 & 0.858 & 0.939 \\
\rowcolor{lightgrey} Top 5\% (11.1x repeat) & 1.065 & 0.843 & 0.930 \\
\rowcolor{ai2lightpink} \olmothree Upsampling  & {\bf{1.000}} & {\bf{0.740}} & {\bf{0.719}} \\
\bottomrule
\end{tabular}
\caption{
\textbf{Quality-aware upsampling outperforms naive data filtering}. We simulate data-constrained training using 1B models trained to 100B tokens where we match the repetition of a 4.51T training run. Results are on the \olmothreeeval Easy suite. Scores are expressed in bits-per-byte (BPB), lower is better (see Section~\S\ref{sec:experimental-design} for details).
}
\label{tab:quality-aware-upsampling}
\end{minipage}

\end{table*}

\begin{table}[t]
\centering
\small
\begin{tabular}{@{}l ccc@{}}
\toprule
&
{$\textbf{\fontsize{8}{8}\selectfont~QA}_\textbf{\fontsize{6}{6}\selectfont~Easy}$} &
{$\textbf{\fontsize{8}{8}\selectfont~Math}_\textbf{\fontsize{6}{6}\selectfont~Easy}$} &
{$\textbf{\fontsize{8}{8}\selectfont~Code}_\textbf{\fontsize{6}{6}\selectfont~Easy}$} \\
\midrule
\rowcolor{lightgrey} Mixing Only             & 1.005 & 0.778 & 0.872 \\
\rowcolor{lightgrey} Quality Upsampling Only & 1.022 & 0.821 & 0.809 \\
\rowcolor{lightgrey} Arithmetic Mean         & 1.004 & 0.792 & 0.828 \\
\rowcolor{lightgrey} Geometric Mean          & 1.004 & 0.782 & 0.813 \\
\rowcolor{lightgrey} Truncated exponential family         & 1.002 & 0.782 & 0.787 \\
\rowcolor{ai2lightpink} Truncated power-exponential family (\olmothree) & {\bf{0.993}} & {\bf{0.758}} & {\bf{0.783}} \\
\bottomrule
\end{tabular}
\caption{
\textbf{Different methods of combining quality-aware upsampling and token-constrained mixing} to arrive at the final \olmothree pretrain mix. Results are on the \olmothreeeval Easy suite. Scores are expressed in bits-per-byte (BPB), lower is better (see Section~\S\ref{sec:experimental-design} for details).
}
\label{tab:resolving-upsampling-mixing}
\end{table}

\subsection{Base Model Additional Data Details: Midtraining}\label{app:sec:midtraining-data}

This section provides further detail on curation processes for \dolminostoo. Additional replication resources, including prompts for synthetic data generation, are available in the Dolma3 GitHub repository.

\subsubsection{Math Capabilities}
\begin{table}[t]
\centering
\begin{tabular}{lrrrrrr}
\toprule
 & {\bf\#~Toks} & ~{\bf\#~Toks} & & & & \\
 {\bf Model} & {\bf Seen (B)} & ~{\bf Total (B)} & {\bf$\mathbf{\Delta}$~MMLU} & {\bf$\mathbf{\Delta}$~Math} & {\bf$\mathbf{\Delta}$~MATH} & {\bf$\mathbf{\Delta}$~GSM8K}\\

\midrule
\rowcolor{ai2offwhite} tinyMATH (PoT)     & 0.24 & 0.24 & -2.90 & {\bf{16.58}} & {\bf{20.70}} & {\bf{25.33}} \\
tinyMATH (MIND)    & 0.90 & 0.90 & -1.75 & 11.62 & 12.48 & 14.80 \\
\rowcolor{ai2offwhite} tinyMATH (Both)    & 1.15 & 1.15 & -1.68 & 9.98  & 11.40 & 12.07 \\
CraneMath          & 4.34 & 4.34 & 0.01  & 4.86  & 4.26  & 6.32 \\
\rowcolor{ai2offwhite} SwallowMath        & 3.65 & 3.65 & {\bf{0.33}}  & 4.84  & 4.38  & 6.72 \\
Dolminos Math      & 5.00 & 10.70 & -0.60 & 4.68 & 2.08  & 7.65 \\
\rowcolor{ai2offwhite} MegaMatt           & 2.69 & 21.78 & 0.32  & 3.39 & 3.91  & 4.85 \\
MM-Web-Pro        & 5.00 & 15.10 & 0.09  & 2.31 & 1.92  & 3.49 \\
\rowcolor{ai2offwhite} MM-Web-Pro-Max         & 5.00 & 73.85 & -0.10 & 1.70 & 1.40  & 2.67 \\
FineMath4+      & 6.89 & 9.61  & 0.03  & 1.51 & 1.21  & 2.19 \\
\rowcolor{ai2offwhite} MM-Web        & 5.00 & 263.90 & 0.03 & 1.30 & 0.69 & 2.16 \\
\bottomrule
\end{tabular}
\caption{\textbf{Results from math microanneals}, with normalized per-token differences in scores relative to pre-anneal baseline. All anneals were run with a 50/50 mixture of web text data and the high quality data source. Numbers were arrived at by taking the difference from the pre-anneal baseline \emph{and dividing by the number of tokens seen during training.}}
\label{tab:midtraining-math-normalized}
\end{table}

Similar to \olmotoo, we take particular care to curate math-specific mixes of data during the midtraining phase of training. In this section we discuss some of the procedures used to generate, as well as validate, the math-specific data sources. It should be noted, that while there has been a flurry of research on high-quality, open-source, STEM-focused datasets, many of these are synthetic data generated using LLama-models, which carry with them a restrictive Llama license. We produce several reproductions of these with more permissive licensing and urge the community to take care in the licensing of the data they release if they wish to see adoption for research or commercial purposes.

\paragraph{TinyMATH}
In \olmotoo, great strides were made in performance on the GSM8K~\citep{cobbe2021gsm8k} dataset by generating synthetic math problems seeded from the original GSM training set, and then generating both python code (PoT) and natural language discussions of solutions (MIND). We adopt a similar strategy here, to target the MATH dataset~\citep{hendrycksmath2021}. Namely, we adopt the TinyGSM protocol~\citep{liu2023tinygsmachieving80gsm8k} and prompts to generate 100 new problems for each existing MATH problem, and then generate pythonic solutions for each of these new problems. Then we apply the MIND rewrite prompt~\citep{akter2024mindmathinformedsynthetic}, using the two-student and problem-solving variants. This yields the PoT dataset (241M tokens) and the MIND dataset (899M tokens). To assess the potency of these new datasets, we ran annealing runs and evaluated fine-grained math related benchmarks as well as MMLU, to keep an eye on generalization. These results are summarized in TABLE:

\paragraph{CraneMath}
SwallowMath~\citep{fujii2025rewriting} is a 2.3 Billion token dataset, generated from rewriting FineMath4$+$~\citep{allal2025smollm2smolgoesbig}. Unfortunately the data was rewritten using a Llama model, which would require that any model trained on this data would need to have "Llama" in the name, according to the Llama Community License. To provide truly open data, we mirror the generation of this dataset, but use Qwen3 32B~\cite{qwen3} to rewrite FineMath4$+$ using the prompt presented in the SwallowMath paper. This yields a 5.62B token dataset we refer to as CraneMath. Compared to the 9.6B tokens contained in FineMATH4$+$, CraneMath is a distillation into fewer tokens, but not as few as SwallowMath (2.3B) -- we posit that this is because using Qwen3 as a rewrite model is slightly "chattier" than Llama.

To evaluate performance of this rewrite procedure, we ran several anneals, starting from a base model that had seen 6T tokens of our pre-training mix, we ran several anneals, always with 50\% token from the pretraining mix and 50\% tokens from the data-source of interest. In the case where the anneals have different token counts, driving the learning rate linearly down to the same final learning rate. Then we compare the following runs: i) The pre-anneal baseline, ii) FineMath4$+$, but just an incomplete subset; iii) the original SwallowMath dataset; iv) our version, CraneMath; v) two copies of CraneMath; vi) a copy of CraneMath and all their original documents from FineMath4$+$.

\paragraph{MegaMatt}
OctoThinker~\citep{wang2025octothinker} generated a 70B token data pool dubbed Megamath-Web-Pro-Max, intended to be a rewrite of LLM360's MegaMath data pool~\citep{liu2023llm360}, with quality mirroring that of the MegaMath-Web-Pro quality. Again, unfortunately, MegaMath-Web-Pro-Max was rewritten using Llama, and an independent recreation needed to be performed for fully-open usage in training. Since our initial ablations showed that the Megamath-Web-Pro-Max pool wasn't as high of quality as, say, SwallowMath, we didn't need a recreation of the full 70B pool. Instead, we generated a recreation of just the documents from Megamath-Web-Pro-Max that occured in CommonCrawl dumps from dump \texttt{CC-MAIN-2023-23} and later, since more recent data was shown in the OctoThinker paper to be of higher quality. We ultimately generated 3.9B tokens of data, dubbed MegaMatt. To verify the efficacy, we ran ablations on: i) MegaMath-Web, ii) MegaMath-Web-Pro-Max (both to 10B and 25B tokens), and iii) MegaMatt.

\paragraph{OMR Rewrites}
Inspired by the success of Nvidia's OpenMathReasoning dataset on the AIO-2 Kaggle competition, we experimented with various rewrites sourced from AoPS forums \cite{moshkov2025aimo2}. See Dolma3 repo for further details.

\paragraph{Key Findings and Results} We summarize the annealing results for the math datasets in Table \ref{tab:midtraining-math-normalized}. Each value reflects the change in the evaluation score relative to the pre-anneal baseline, normalized by the number of training tokens. Presenting the results this way highlights several distinct tiers of math-data quality, stratified by the effect-per-token. Notably, these quality tiers anticorrelate with the number of available tokens: the highest-quality sources are also the smallest. While it is true that there are diminishing returns of evaluation scores as more tokens are added, we claim that amongst these high-quality data sources, some higher quality than others.

At the top of the quality-spectrum are the tinyMATH variants. Although each contains less than a billion tokens, they deliver the strongest improvement per token -- this is perhaps not surprising as these tokens were specifically crafted to augment the MATH evaluation score. Next in the tier-list of quality are the synthetic rewrites of natural high-quality data: the Crane, SwallowMath and MegaMatt sources which are each rewrites of FineMath4+ and MegaMathWeb-Pro. These provide a markedly weaker lift to the math evaluation metrics than the tinyMATH variants but also have a much larger pool of tokens to draw from. Finally, the largest data sources, including those of naturally occurring data such as FineMath4+ and MegamathWeb, also yield improvements, but their effect-per-token is noticeably smaller than that of the highly curated synthetic data. Finally we note that the effect of math midtraining on MMLU is generally neutral to negative, but is more strongly negative the more targeted the data pool is to Math evals, suggesting ``overcooking'', where increased specialization comes at the expense of broader general-purpose performance.

\subsubsection{Code Capabilities}
During pretraining, we relied entirely on stack-edu~\citep{allal2025smollm2smolgoesbig} for providing coding data. This data came in the form of naturally-occurring source code from github scraps with limited extra preprocessing. During midtraining, we focused on improving Python and code-completion capabilities. To this end, we incorporated 10B tokens of FIM-transformed data form the same source as the pretraining code mixture. Inspired by improvements in math metrics by incorporating synthetic data, we also created a fully-open replica of SwallowCode~\citep{fujii2025rewriting}, which we call CraneCode.

\paragraph{CraneCode}
Of the off-the-shelf synthetic code data sources we considered, SwallowCode provided the greatest lift to coding evaluation metrics. Unfortunately, SwallowCode was generated using Llama models and thus had the less-permissive Llama license attached. We created a replica of SwallowCode by starting with just the python files from The Stack v2 Smol~\citep{lozhkov2024starcoder}, and applying the compilation and linting filters just as in SwallowCode. Then we applied a two-stage rewriting process, first to generate code data that is more compliant to the python style guides (SGCR), and then to generate optimized code (SCOR); both using the prompts from the original SwallowCode paper and Qwen/Qwen2.5-Coder-32B-Instruct~\citep{qwen2.5}. To verify the quality of the reproduced dataset, we ran several anneals, where results are displayed in Table \ref{tab:microanneal_code}.

\begin{table}[h]
\centering
\begin{tabular}{lrrrrr}
\hline
{\bf Model} & {\bf \#Tokens} & {\bf Crux-Eval} & {\bf HumanEval} & {\bf MBPP} & {\bf MMLU} \\
\hline
\rowcolor{ai2offwhite} CraneCode (25B) & 18.87B & 35.92 & \textbf{35.06} & 31.72 & 54.30 \\
CraneCode SGCR & 18.87B & \textbf{41.75} & 33.78 & \textbf{36.76} & 54.18 \\
\rowcolor{ai2offwhite} SwallowCode & 10.0B & 35.74 & 31.80 & 34.67 & 55.03 \\
CraneCode (10B) & 10.0B & 33.28 & 26.51 & 34.94 & 54.98 \\
\rowcolor{ai2offwhite} Pre-anneal Baseline & N/A & 35.46 & 21.51 & 27.11 & \textbf{56.60} \\
\hline
\end{tabular}
\caption{\textbf{Microanneal results for CraneCode ablations}. For each annealing run, we ran with a 50/50 mixture of web text and high-quality synthetic code data. We note several observations: 1) Both SwallowCode and CraneCode provide a lift to coding evaluation metrics at the expense of MMLU metrics; 2) SwallowCode provides a larger lift normalized for tokens than the CraneCode dataset; 3) CraneCode continues to provide lift to HumanEval as more data is provided, indicating that this data source is not yet exhausted.}
\label{tab:microanneal_code}
\end{table}

\subsubsection{Thinking Capabilities}

\paragraph{Meta-reasoning}

Recent work demonstrates that structured meta reasoning capabilities present during pre-training and mid-training serve as the foundation for successful reinforcement learning in complex reasoning tasks. \citet{gandhi2025cognitive} showed that models exhibiting verification and backtracking behaviors during base training achieved dramatically superior performance trajectories during mathematical reasoning RL. %
Therefore, we begin by identifying structured reasoning capabilities that are critical for mathematical problem-solving. We select seven core capabilities that are foundational to mathematical and programming expertise: self-awareness~\citep{toy2024metacognitionneedusingintrospection,bfa322bf36e54a4ca19f9a73bee6184b}, self-evaluation~\citep{Fleming2017-FLESOD}, goal management~\citep{ACKERMAN2017607,GRIFFITHS201924}, hierarchical organization~\citep{Haupt2018}, backward chaining~\citep{Olieslagers2024}, backtracking and conceptual reasoning \citep{Markovits2015}. We then design specific tasks that systematically target these capabilities, as shown in Table~\ref{tab:math-meta}, and \ref{tab:code-meta}. For instance, Math Error Recovery specifically targets self-awareness, verification, and backtracking by requiring models to experience authentic mistakes and demonstrate recovery processes. Strategy Selection focuses purely on meta-cognitive choice processes, while Conversation Generation integrates all capabilities through educational dialogue. %
For data generation, we start with existing math \citep{deepscaler2025,moshkov2025aimo2} and coding \citep{tacoli,hendrycksapps2021,ahmad2025opencodereasoning} problems and their corresponding correct answers. Following Pandalla dataset,\footnote{\href{https://huggingface.co/datasets/pandalla/pandalla-math-dataset-v1.0}{\path{huggingface.co/datasets/pandalla/pandalla-math-dataset-v1.0}}} we automatically augment each problem with detailed  annotations\footnote{We provide the problem and the correct answer as inputs to \texttt{o4-mini} with high reasoning, to synthesize the annotations following the Pandalla-math annotation schema.} covering `problem classification', `difficulty analysis', `solution approaches', `common pitfalls', and `verification methods'. These rich annotations serve as inputs for our capability-targeted tasks. For example, the `common pitfalls' field directly informs math error recovery generation, while steps in `solution approach' provides structure for backward chaining tasks. Using the annotated datasets as foundation, we employ \texttt{GPT-4.1} and \texttt{o4-mini} to generate training data at scale for each capability-targeted task.

\begin{table}[h!]
\centering
\begin{tabular}{ll}
\toprule
{\bf Task} & {\bf Meta Capabilities} \\
\midrule
\rowcolor{ai2offwhite} Math error recovery & Self-awareness, verification, backtracking \\
Choosing the technique to use & Strategy selection \\
\rowcolor{ai2offwhite} Difficulty estimation \& self-awareness prompts & Self-evaluation \\
Steps generation & Goal management, hierarchical organization \\
\rowcolor{ai2offwhite} From answer, generate steps backwards & Backward chaining \\
Conversation generation & All capabilities (tagging) \\
\rowcolor{ai2offwhite} Reason about necessary concepts and how they connect & Conceptual reasoning \\
\bottomrule
\end{tabular}
\caption{\textbf{Meta reasoning capabilities across mathematical tasks}.}
\label{tab:math-meta}
\end{table}

\begin{table}[h!]
\centering
\begin{tabular}{ll}
\toprule
{\bf Task} & {\bf Meta Capabilities} \\
\midrule
\rowcolor{ai2offwhite} Code error recovery (single-turn) & Self-awareness, verification, backtracking \\
Code error recovery (multi-turn) & Self-awareness, verification, backtracking \\
\rowcolor{ai2offwhite} Planning the solution & Strategy selection, goal management \\
Solution implementation & Conceptual-level processing, hierarchical organization \\
\rowcolor{ai2offwhite} Code quality evaluation (high/low) & Self-evaluation \\
Difficulty estimation & Self-evaluation, self-awareness \\
\rowcolor{ai2offwhite} Unit test walkthrough & Goal management, verification \\
\bottomrule
\end{tabular}
\caption{\textbf{Meta reasoning capabilities across coding tasks}.}
\label{tab:code-meta}
\end{table}

\paragraph{Existing thinking traces}
The full list of existing thinking traces is as follows:

\begin{enumerate}
    \item \textbf{General reasoning mix} is a compilation of three existing datasets: GeneralThought-430K\footnote{\href{https://huggingface.co/datasets/RJT1990/GeneralThoughtArchive}{\path{huggingface.co/datasets/RJT1990/GeneralThoughtArchive}}}, \texttt{OpenThoughts-114k}~\citep{guha2025openthoughtsdatarecipesreasoning}, and \texttt{Open-R1-Math-220k}\footnote{\href{https://huggingface.co/datasets/open-r1/OpenR1-Math-220k}{\path{huggingface.co/datasets/open-r1/OpenR1-Math-220k}}}. The resulting dataset contains questions, thinking traces, and answers for topics spanning math, code, natural sciences, humanities, social sciences, and puzzles.
    \item \textbf{Gemini reasoning traces}, introduced by \citet{muennighoff2025s1simpletesttimescaling}, contains thinking traces covering domains of math, astronomy, biology, chemistry, computer science, geography, physics, English, law, logic, and more.
    \item \textbf{OpenThoughts2 reasoning traces} from \citet{guha2025openthoughtsdatarecipesreasoning} contains thinking traces in domains of math, science, code, and puzzles.
    \item \textbf{Llama Nemotron reasoning traces} \citep{bercovich2025llamanemotronefficientreasoningmodels} contains thinking trace data for math, code, general reasoning, and instruction following.
    \item \textbf{QwQ reasoning traces} consists of the QwQ subset of the OpenMathReasoning dataset~\citep{moshkov2025aimo2}.
\end{enumerate}

Filtering steps included subselecting for permissively-licensed generations, filtering to remove empty and truncated responses, performing checks of verifiable claims and safety, filtering overt LLM self-references, filtering heavily repeated sentences, paragraphs, and phrases, and remove reasoning traces consisting of greater than 5\% Chinese characters.

\subsection{Base Model Additional Evaluation Details}
\label{sec:eval-suite}

The \olmothreeeval suite expands on the 11 tasks in the \olmotoo iteration of OLMES \citep{olmo20242olmo2furious,olmes}, to include 43 tasks across new families of capabilities. Here, we enumerate details from Section~\S\ref{sec:experimental-design}.
All task suites are publicly available at \texttt{\href{https://github.com/allenai/olmes\#olmo-3-eval-suite}{github.com/allenai/olmes\#olmo-3-eval-suite}}.

\paragraph{Expanding OLMES tasks} We expanded our evaluation to target specific capabilities: new QA tasks focusing on science knowledge (SciQ, QASPER, SciRIFF), medical/lab knowledge (ProtocolQA, DBQA, MedMCQA, MedQA), math tasks (GSM Symbolic, Minerva MATH) and coding tasks (DS 1000, BigCodeBench, Deepseek LeetCode\footnote{We use `Deepseek LeetCode' to refer to the 180 LeetCode problems used during development in \citet{guo2024deepseek}}, MultiPL-E HumanEval, MultiPL-E MBPP). We use MultiPL-E to evaluate our multilingual code execution, limited to six core programming languages. Additionally, we track fill-in-the-middle (FIM) performance using HumanEval with the three settings from \citet{bavarian2022efficient}: single-line infilling, multi-line infilling and random span infilling.

We support code execution in Python, C++, Java, JavaScript, PHP, Rust and Shell using AWS Lambda functions to grade instances in parallel, isolated environments of up to 50K generations simultaneously. In total, our environments graded 17.2 million generated code samples during \olmothree development, with up to 1.5K simultaneously. To ensure reproducibility, we release a lightweight Docker library for code execution without AWS infrastructure\footnote{Our code execution environments are publicly available at \texttt{\href{https://github.com/allenai/olmes-docker}{github.com/allenai/olmes-docker}}.}.

Additionally, \olmotoo only tracked math and code capabilities after mid-training, as small models exhibit random-chance pass@1 performance on math and code tasks \citep{wei2022emergent}. Our base easy suite tracks perplexity over human-written math and code solutions \citep{huang2024compression}, which allows us to broadens the scope of capabilities we track during pre-training.

\begin{figure}[t]
    \centering
    \includegraphics[width=\textwidth]{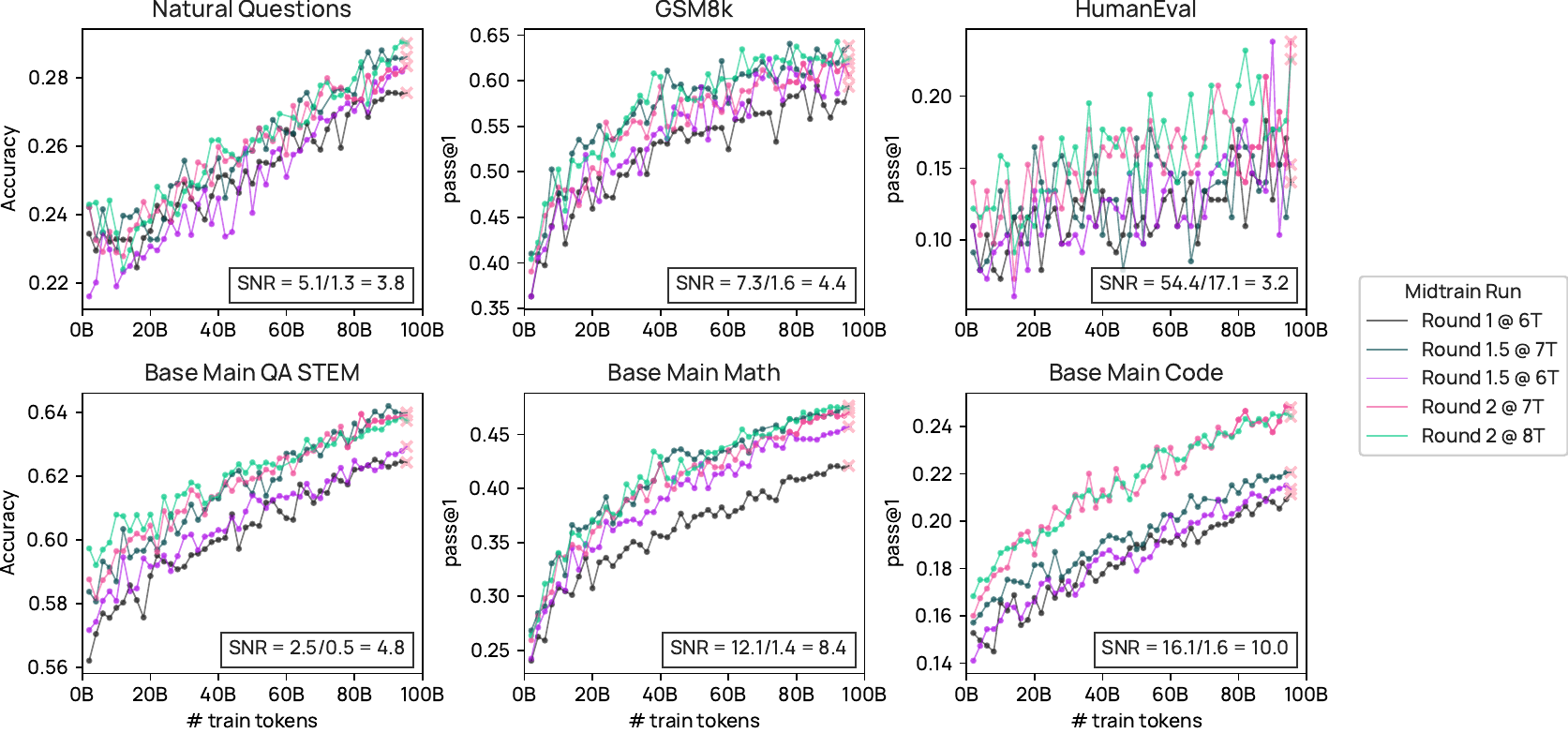}
    \caption{
    \textbf{Training curves of midtraining} on canonical language model benchmarks (top), and our proposed base main task suites (bottom) for QA, Math and Code.
    We used the signal-to-noise ratio of early mid-training runs to make decisions about aggregating evaluation scores. Our resulting task averages had a better signal-to-noise ratio than individual benchmarks.
    }
    \label{fig:snr-midtrain}
\end{figure}

\begin{figure}[t]
    \centering
    \includegraphics[width=\textwidth]{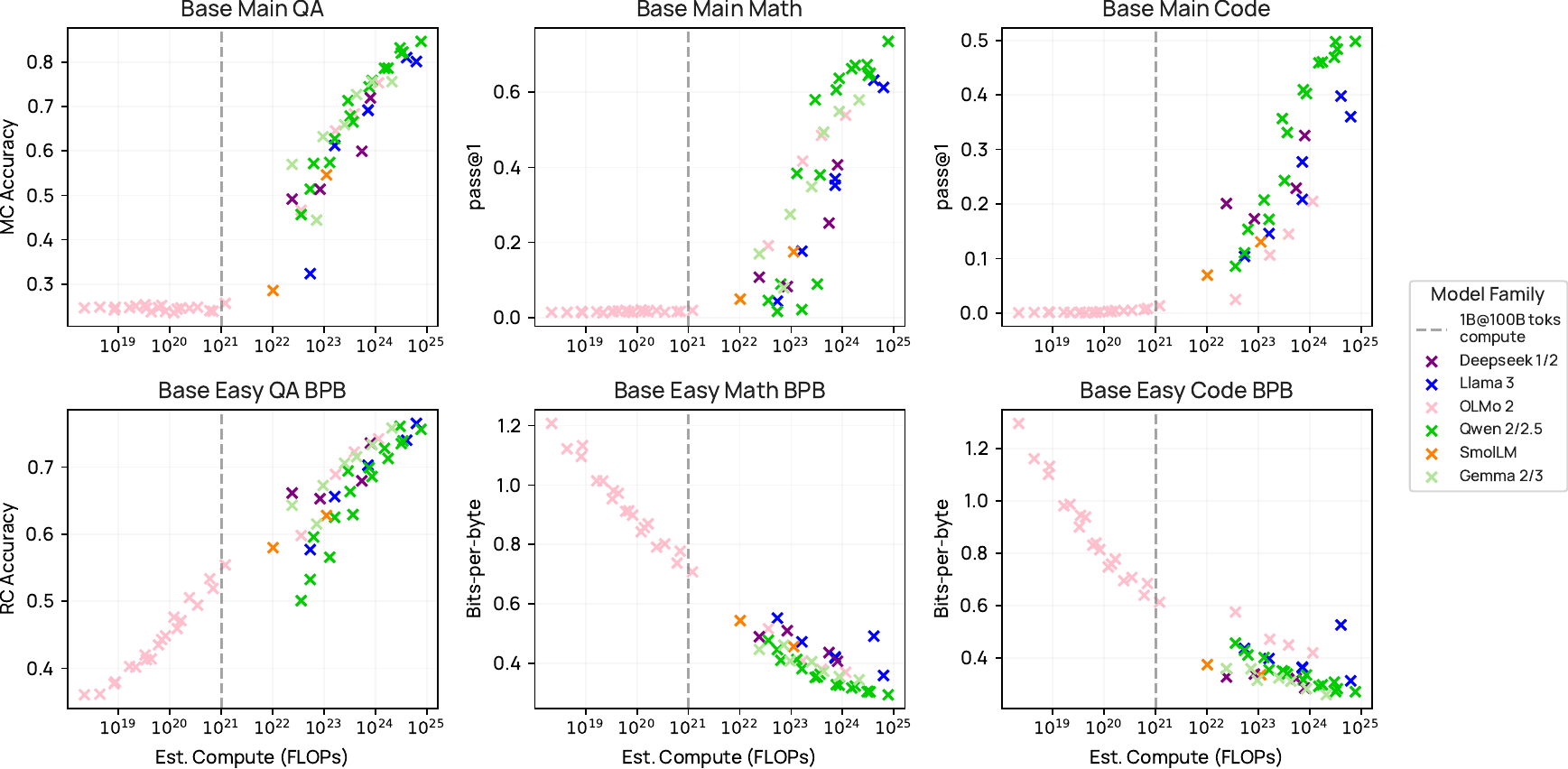}
    \caption{
    \textbf{Scaling analysis for the \olmothree base evaluation suite}. At the largest scale used to run from-scratch data ablations (grey line, a 1B model trained to 100B tokens), our `base main' evaluation suite is too difficult to show improvement (top figures). Instead, we introduce a `base easy' suite to compare models at small scales (bottom figures).
    }
    \label{fig:scaling-analysis}
\end{figure}

\begin{figure}[t]
    \centering
    \includegraphics[width=\textwidth]{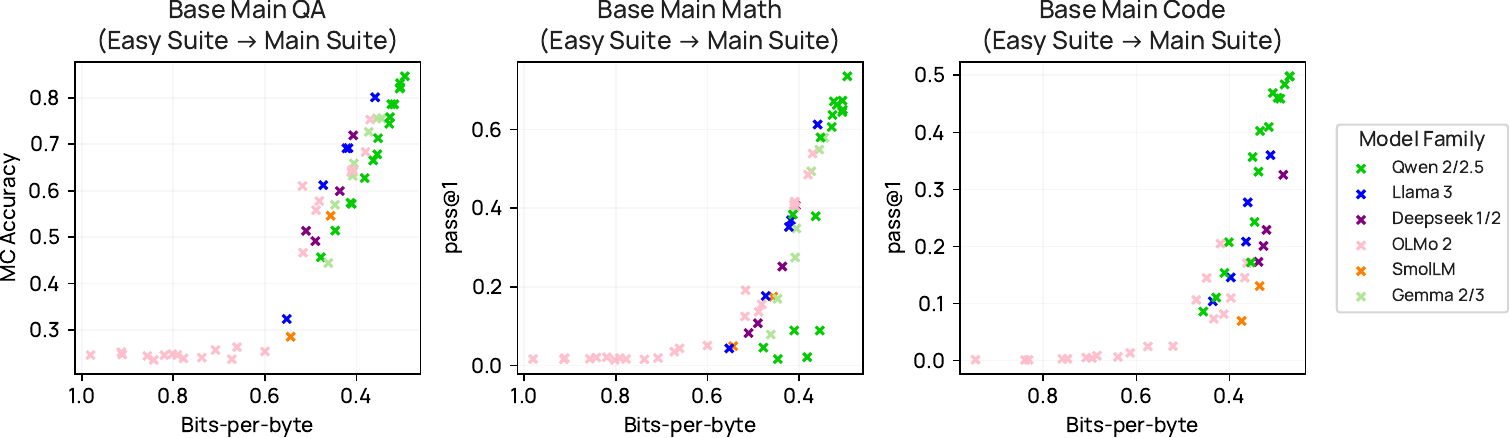}
    \caption{
    \textbf{Relationship between bits-per-byte using the Easy suite and final metrics on the Main eval suite}. We use the `Easy' suite to make decisions at a small scale, which corresponds to an improvement at the large scale.
    }
    \label{fig:easy-to-main}
\end{figure}

\subsubsection{Base Evaluation suites}

Using the analysis tools described in the previous section, we construct two evaluation suite for decision making in pre-training: the {\bf{Base Easy}} suite for small-scale data decisions and the {\bf{Base Main}} suite for in-loop evaluation and mid-training data decisions. We kept the number of in-context examples and generation arguments consistent within each family of tasks, when possible.\footnote{We perform all evaluation using vLLM. To prevent performance discrepancies between versions, we pin to \texttt{v0.9.0.1} for evaluation during development, and pin to \texttt{v0.11.0} for all evaluation in the final report.}

Table~\ref{tab:task-details} describes the task configuration and metrics for the \olmothree Base Main evaluation suite. 
Table~\ref{tab:easy-task-details} provides an overview of the Base Easy suite.

\paragraph{Base Easy suite}
For multiple-choice BPB, we simply use the correct answer as the continuation.
For math BPB, we use the provided human-written solutions from Minerva MATH \citep{lewkowycz2022solving}.
For code BPB, we use the gold `canontical' solution as provided in HumanEval and MBPP \citep{chen2021codex,austin2021program}.
For BPB over non-Python coding tasks, MultiPL-E did not release gold solutions \citep{cassano2022multipl}, so we generate silver continuations for 16 languages using \texttt{o4-mini-medium}\footnote{We release this generation set at \texttt{\href{https://huggingface.co/datasets/allenai/multilingual_mbpp}{huggingface.co/datasets/allenai/multilingual\_mbpp}}}.
Figure \ref{fig:scaling-analysis} shows the scaling behavior of the three base easy task clusters, where we see signal even at very small (190M parameter) model sizes.

One important property of the Base eval suite is that a ranking of two small models on the base easy suite agrees with their ranking on the downstream base main suite. We validate this by measuring rank correlation between the easy and main task suites, as pictured in Figure \ref{fig:easy-to-main}.

\paragraph{Base Main suite} As a result of the clustering procedure, the base main suite tracks 6 task groups: MCQA STEM, MCQA Non-STEM, Gen, Math, Code, Code FIM. Unlike \olmotoo, we are tracking generative math and code tasks at pre-training. We chose to evaluate pass@k with the largest number of samples such that each task could evaluate on \olmotoo 7B on 1 H100 in under 30 minutes, in order to ensure the eval speed is not bottlenecked by any particular task. For tasks with a large enough $n$, we set $k=16$ to match the GRPO group size, which we observed to act as an empirical upper-bound on the possible improvement from RL training. To decide on the the temperature and top-p, we ran a sweep and evaluated 5 models (\olmotoo 7B, 13B; Qwen 2.5 7B, 13B; Qwen 3 8B; \citealp{qwen2.5,qwen3}) to find an adequate configuration setting for high scores on both pass@1 and pass@k. Results are shown in Figure \ref{fig:temperature-sweep}, and we select temperature and top-p of 0.6 for all base math and code evaluation.

\begin{wrapfigure}{r}{0.45\linewidth}
    \centering
    \vspace{-10pt}
    \includegraphics[width=\linewidth]{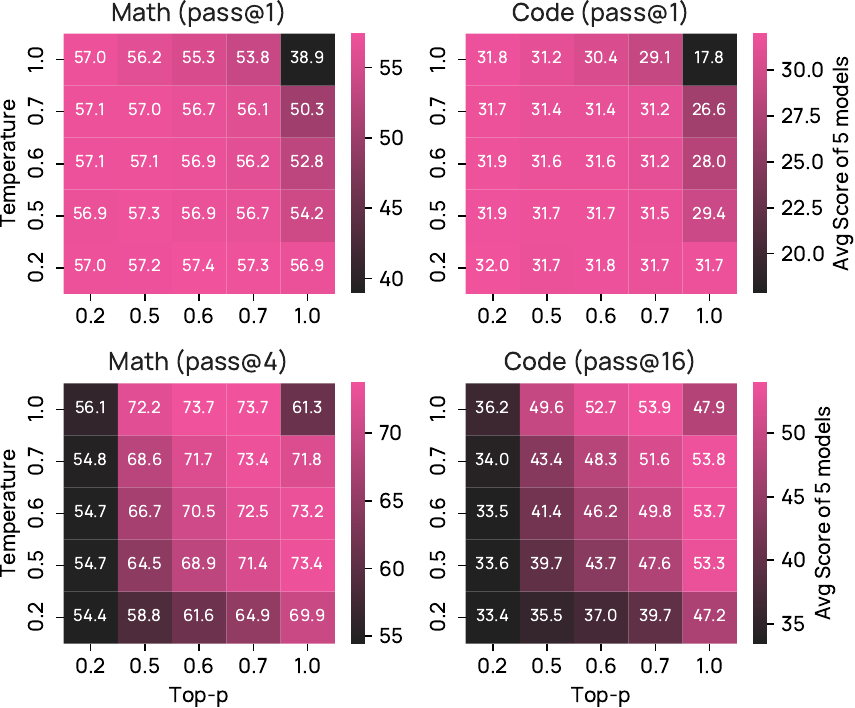}
    \caption{\textbf{To select generation arguments for base evaluation, we run a temperature and top-p sweep across 5 models}. We use a reasonable configuration such that we can calculate both pass@1 and pass@k using the results of a single evaluation job.}
    \label{fig:temperature-sweep}
    \vspace{-10pt}
\end{wrapfigure}

\paragraph{Base Chat suite} During mid-training, we refashion the Chat eval suite (\S\ref{sec:posttrain_eval}) for use evaluating base models, which served as a reference as to whether we expect our model to perform well after the adaptation pipeline. To do this, we used a standard, simple chat template (\texttt{Question: \{text\}\textbackslash nAnswer:}) across all base models (both \olmothree and baseline models) and we included stop tokens to prevent degenerate responses. We also excluded tasks which required an API-based judge (AlpacaEval, SimpleQA) due to cost. In practice, we noticed most of the disagreements between the base main and base chat evaluation suites were due to noise, so we primarily used the base suite for making decisions.

\paragraph{Base Long-Context suite} During the long-context extension phase, we evaluate long-context capability using RULER \citep{hsieh2024rulerwhatsrealcontext} as our primary development signal. As a complementary held-out set, we also use HELMET \citep{yen2025helmet}, noting that the HELMET \emph{Recall} task directly implements several RULER evaluations (specifically, ruler-niah-mk-2, ruler-niah-mk-3, and ruler-niah-mv). Because we evaluate only base models at this stage, we disable chat templates within HELMET to ensure consistent scoring across models. For HELMET tasks requiring an LLM-as-a-judge, we use its default judge configuration (gpt-4o-2024-05-13). Taken together, RULER guides most model-selection decisions during long-context development, with HELMET providing an additional check on generalization.

\paragraph{Base Held-out Suite} We targeted one held-out evaluation task to match each family of capability: MMLU Pro for QA \citep{wang2024mmlu}, LBPP for code \citep{matton-etal-2024-leakage}, Deepmind Math for math \citep{saxton2019analysing}, and BigBench Hard to measure broad coverage across unseen task types \citep{suzgun2022challenging}.

\subsubsection{New Evaluation Benchmarks}
\label{sec:new-evaluation-benchmarks}

\paragraph{Basic Skills} We developed a new benchmark, \textsc{BasicSkills}, to measure whether core capabilities are being acquired during pretraining. \textsc{BasicSkills} consists of 6 subtasks: basic arithmetic, string manipulation, simple coding, elementary logical reasoning, basic common sense, and simple pattern recognition. Each task isolates a single skill using a self-contained context that requires no external knowledge or additional information and can be completed through natural text continuation without relying on instruction-following abilities.

\paragraph{Gen2MC} One takeaway from \olmotoo development was a sensitivity to task format. The clustering procedure furhter confirmed this, finding that generative scores rank models similarly as rank choice (RC) QA tasks, disagreeing with ranking of single-token multiple choice (MC) QA tasks (see Figure \ref{fig:clustering_dendrogram}). In particular, the short-form generative QA tasks (GenQA in Table \ref{tab:task-details}) evaluate by comparing a generated answer to a bank of plausible answers, but these answer banks are often not complete, leading to false negatives. To address this, we introduce the \textbf{Gen2MC} benchmarks, which were constructed by taking the original question/answer pairs and generating incorrect multiple-choice distractor answers using a strong LLM. For each set of generated distractors, we manually review a set of 200 sample questions from the validation set before generating the full dataset. We create Gen2MC tasks for DROP, Jeopardy, NaturalQs, SQuAD, CoQA using GPT-4o for generating distractors, and fall-back to GPT-4.1 in cases where output parsing failed.

\paragraph{Masked perplexity} We want our model to perform well on the diversity of requests from real user chat data; however, we don’t want to overfit to the ``style'' of chat outputs.
To avoid this, we use a simple token masking strategy, inspired by work in loss masking \citep{mindermann2022prioritized}:

\begin{enumerate}
\item Fine-tune a 1B model on a tiny subset of the dataset (\~5\%) with a small learning rate. The key idea is that we `warm up' to the format of the target set without learning a lot of new knowledge.
\item Compute the token losses of the base model and the fine-tuned model on every sequence in the dataset and compute the difference: $\log p_{\text{SFT}}(y | x) - \log p_\text{base}(y | x)$
\item Mask tokens where the difference is greater than some threshold (found by inspection)
\item Also mask the user responses and tool calls (we don’t want to model these for data selection)
Use the loss at all the non-masked tokens positions for perplexity evaluations
\end{enumerate}

In practice, we use \olmotoo 1B and the trained \olmotoo 1B SFT to compute the loss difference on target tokens. We use UltraChat and WildChat \citep{ding2023enhancing,zhao2024wildchat} as our masked perplexity sets.

\begin{table*}[t]
  \centering

\begin{scriptsize}
\begin{tabular}{lllHlHlHHHHHHl} %
\toprule
& \bf{Task} & \bf{Capability} & \bf{\# Inst} & \bf{ICL} & \bf{Format} & \bf{Metric} & \bf{Temp} & \bf{top-p} & \bf{Extract} & \bf{Max Toks.} & \bf{p@k (n)} & \bf{N} & \bf{\# Sub.} \\
\midrule

\rowcolor{ai2midpink}\multicolumn{14}{c}{\rule{0pt}{1pt}} \\[-9pt]
\rowcolor{ai2midpink}\multicolumn{14}{c}{\textbf{Base Easy Suite}} \\
\rowcolor{ai2midpink}\multicolumn{14}{c}{\rule{0pt}{1pt}} \\[-9pt]
\rowcolor{ai2lightpink} & Minerva MATH\starOlmo{} (\citeyear{lewkowycz2022solving}) & Math Gen & - & 4$^\alpha$ & BPB & BPB & - & - & - & - & - & - & 7 \\
\rowcolor{lightgrey} & HumanEval\starOlmo{} (\citeyear{chen2021codex}) & Code Gen & - & 3 & BPB & BPB & - & - & - & - & - & - & - \\
\rowcolor{lightgrey} & MBPP\starOlmo{} (\citeyear{austin2021program}) & Code Gen & - & 3 & BPB & BPB & - & - & - & - & - & - & - \\
\rowcolor{lightgrey} \multirow{-3}{*}{\rotatebox[origin=c]{90}{\textit{Code}}} & MT MBPP\starOlmo{} (\citeyear{cassano2022multipl}) & Code Gen & - & 3 & BPB & BPB & - & - & - & - & - & - & 17 \\
\rowcolor{ai2offwhite} & ARC (\citeyear{clark2018think}) & Science QA & - & 5 & BPB & BPB & - & - & - & - & - & - & 2 \\
\rowcolor{ai2offwhite} & MMLU (\citeyear{hendryckstest2021}) & General QA & - & 5 & BPB & BPB & - & - & - & - & - & - & 57 \\
\rowcolor{ai2offwhite} & CSQA (\citeyear{talmor-etal-2019-commonsenseqa}) & Commonsense QA & - & 5 & BPB & BPB & - & - & - & - & - & - & - \\
\rowcolor{ai2offwhite} & HellaSwag (\citeyear{zellers-etal-2019-hellaswag}) & Language Modeling & - & 5 & BPB & BPB & - & - & - & - & - & - & - \\
\rowcolor{ai2offwhite} & WinoGrande (\citeyear{Sakaguchi_Le_Bras_Bhagavatula_Choi_2020}) & Language Modeling & - & 5 & BPB & BPB & - & - & - & - & - & - & - \\
\rowcolor{ai2offwhite} & SocialIQA (\citeyear{sap-etal-2019-social}) & Social QA & - & 5 & BPB & BPB & - & - & - & - & - & - & - \\
\rowcolor{ai2offwhite} & PiQA (\citeyear{Bisk_Zellers_Le_bras_Gao_Choi_2020}) & Physical QA & - & 5 & BPB & BPB & - & - & - & - & - & - & - \\
\rowcolor{ai2offwhite} & CoQA (\citeyear{reddy-etal-2019-coqa}) & Conversation QA & - & 0$^\dagger$ & BPB & BPB & - & - & - & - & - & - & - \\
\rowcolor{ai2offwhite} & DROP (\citeyear{dua-etal-2019-drop}) & Passage QA & - & 5 & BPB & BPB & - & - & - & - & - & - & - \\
\rowcolor{ai2offwhite} & Jeopardy (\citeyear{mosaic-jeopardy}) & Trivia QA & - & 5 & BPB & BPB & - & - & - & - & - & - & - \\
\rowcolor{ai2offwhite} & NaturalQs (\citeyear{kwiatkowski-etal-2019-natural}) & General QA & - & 5 & BPB & BPB & - & - & - & - & - & - & - \\
\rowcolor{ai2offwhite} & SQuAD (\citeyear{rajpurkar-etal-2016-squad}) & General QA & - & 5 & BPB & BPB & - & - & - & - & - & - & - \\
\rowcolor{ai2offwhite} & SciQ\starOlmo{} (\citeyear{welbl-etal-2017-crowdsourcing}) & Science QA & - & 5 & BPB & BPB & - & - & - & - & - & - & - \\
\rowcolor{ai2offwhite} & QASPER\starOlmo{} (\citeyear{dasigi2021dataset}) & Science QA & - & 5 & BPB & BPB & - & - & - & - & - & - & - \\
\rowcolor{ai2offwhite} & Basic Skills\starOlmo{} (\S\ref{sec:new-evaluation-benchmarks}) & Basic QA & - & 5 & BPB & BPB & - & - & - & - & - & - & 6 \\
\rowcolor{ai2offwhite} & DBQA\starOlmo{} (\citeyear{laurent2024lab}) & Science QA & - & 5 & BPB & BPB & - & - & - & - & - & - & - \\
\rowcolor{ai2offwhite} & ProtocolQA\starOlmo{} (\citeyear{laurent2024lab}) & Science QA & - & 5 & BPB & BPB & - & - & - & - & - & - & - \\
\rowcolor{ai2offwhite} & Lambada\starOlmo{} (\citeyear{paperno2016lambada}) & Language Modeling & - & 0 & BPB & BPB & - & - & - & - & - & - & - \\
\rowcolor{ai2offwhite} & MedMCQA\starOlmo{} (\citeyear{pmlr-v174-pal22a}) & Medical QA & - & 5 & BPB & BPB & - & - & - & - & - & - & - \\
\rowcolor{ai2offwhite} & MedQA\starOlmo{} (\citeyear{jin2021disease}) & Medical QA & - & 5 & BPB & BPB & - & - & - & - & - & - & - \\
\rowcolor{ai2offwhite} \multirow{-21}{*}{\rotatebox[origin=c]{90}{\textit{QA}}} & SciRIFF\starOlmo{} (\citeyear{wadden2024sciriff}) & Science QA & - & 5 & BPB & BPB & - & - & - & - & - & - & - \\

\bottomrule
\end{tabular}
\end{scriptsize}
  \caption{
  \textbf{Details of the \olmothree base easy evaluation suite}. 
  Tasks were formatted as bits-per-byte (BPB) over the gold continuation, or rank choice (RC, following the setup in \citet{olmes}).
  \starOlmo{} = new additions to the base \olmotoo suite \citep{olmo20242olmo2furious}; $^\dagger$ = few-shot examples are built-in the task; $^\alpha$ = human-written few-shot examples.
  }
  \label{tab:easy-task-details}
\end{table*}

\begin{table*}[t]
  \centering

\begin{scriptsize}
\renewcommand{\arraystretch}{1}
\adjustbox{max width=\linewidth}{
\begin{tabular}{llHHlllllHllHl} %
\toprule
& {\bf Task} & {\bf Capability} & {\bf \# inst} & {\bf ICL} & {\bf Format} & {\bf Metric} & {\bf Temp} & {\bf Top-p} & {\bf Extract} & {\bf Max toks} & {\bf P@k (n)} & {\bf N} & {\bf \# sub} \\
\midrule

\rowcolor{ai2midwhite}\multicolumn{14}{c}{\rule{0pt}{1pt}} \\[-9pt]
\rowcolor{ai2midwhite}\multicolumn{14}{c}{\textbf{Base Main Suite}} \\
\rowcolor{ai2midwhite}\multicolumn{14}{c}{\rule{0pt}{1pt}} \\[-9pt]
\rowcolor{ai2offwhite} & GSM8K* (\citeyear{cobbe2021gsm8k}) & Math Gen & - & 8$^\alpha$ & CoT EM & pass@k & 0.6 & 0.6 & GSM & 512 & 1, 4 (8) & 8 & - \\
\rowcolor{ai2offwhite} & GSM Symbolic* (\citeyear{gsm-symbolic}) & Math Gen & - & 8$^\alpha$ & CoT EM & pass@k & 0.6 & 0.6 & GSM & 512 & 1, 4 (8) & 8 & 3 \\
\rowcolor{ai2offwhite} & Minerva MATH* (\citeyear{lewkowycz2022solving}) & Math Gen & - & 4$^\alpha$ & CoT EM & pass@k & 0.6 & 0.6 & Minerva & 1024 & 1, 4 (4) & 4 & 7 \\
\rowcolor{ai2offwhite} \multirow{-4}{*}{\rotatebox[origin=c]{90}{\textit{Math}}} & MATH 500* (\citeyear{lewkowycz2022solving,lightman2023lets}) & Math Gen & - & 4$^\alpha$ & CoT EM & pass@k & 0.6 & 0.6 & Minerva & 1024 & 1, 16 (32) & 32 & - \\
\rowcolor{lightgrey} & HumanEval* (\citeyear{chen2021codex}) & Code Gen & - & 3 & Code Exec & pass@k & 0.6 & 0.6 & - & 512 & 1, 16 (32) & 32 & - \\
\rowcolor{lightgrey} & MBPP* (\citeyear{austin2021program}) & Code Gen & - & 3 & Code Exec & pass@k & 0.6 & 0.6 & - & 512 & 1, 16 (32) & 32 & - \\
\rowcolor{lightgrey} & BigCodeBench* (\citeyear{zhuo2024bigcodebench}) & Code Gen & - & 3 & Code Exec & pass@k & 0.6 & 0.6 & - & 1280 & 1 (5) & 5 & - \\
\rowcolor{lightgrey} & DS 1000* (\citeyear{Lai2022DS1000}) & Code Gen & - & 3 & Code Exec & pass@k & 0.6 & 0.6 & - & 1024 & 1 (5) & 5 & - \\
\rowcolor{lightgrey} & Deepseek LeetCode* (\citeyear{guo2024deepseek}) & Code Gen & - & 0 & Code Exec & pass@k & 0.6 & 0.6 & - & 512 & 1, 16 (32) & 32 & - \\
\rowcolor{lightgrey} & MultiPL-E HumanEval* (\citeyear{cassano2022multipl}) & Code Gen (6 Lang) & - & 0 & Code Exec & pass@k & 0.6 & 0.6 & - & 1024 & 1, 16 (32) & 32 & 6 \\
\rowcolor{lightgrey} \multirow{-7}{*}{\rotatebox[origin=c]{90}{\textit{Code}}} & MultiPL-E MBPP* (\citeyear{cassano2022multipl}) & Code Gen (6 Lang) & - & 0 & Code Exec & pass@k & 0.6 & 0.6 & - & 1024 & 1, 16 (32) & 32 & 6 \\
\rowcolor{ai2offwhite} & HumEval FIM Single* (\citeyear{bavarian2022efficient}) & Code FIM & - & 0 & FIM & pass@1 & 0.8 & 0.95 & - & 512 & 1 (10) & 10 & - \\
\rowcolor{ai2offwhite} & HumEval FIM Random* (\citeyear{bavarian2022efficient}) & Code FIM & - & 0 & FIM & pass@1 & 0.8 & 0.95 & - & 512 & 1 (5) & 5 & - \\
\rowcolor{ai2offwhite} \multirow{-3}{*}{\rotatebox[origin=c]{90}{\textit{FIM}}} & HumEval FIM Multi* (\citeyear{bavarian2022efficient}) & Code FIM & - & 0 & FIM & pass@1 & 0.8 & 0.95 & - & 512 & 1 (1) & 1 & - \\
\rowcolor{lightgrey} & ARC (\citeyear{clark2018think}) & Science QA & - & 5 & MC & Acc & - & - & - & - & - & - & 2 \\
\rowcolor{lightgrey} & MMLU STEM (\citeyear{hendryckstest2021}) & General QA & - & 5 & MC & Acc & - & - & - & - & - & - & 19 \\
\rowcolor{lightgrey} & MedMCQA* (\citeyear{pmlr-v174-pal22a}) & Medical QA & - & 5 & MC & Acc & - & - & - & - & - & - & - \\
\rowcolor{lightgrey} & MedQA* (\citeyear{jin2021disease}) & Medical QA & - & 5 & MC & Acc & - & - & - & - & - & - & - \\
\rowcolor{lightgrey} \multirow{-5}{*}{\rotatebox[origin=c]{90}{\textit{STEM QA}}} & SciQ* (\citeyear{welbl-etal-2017-crowdsourcing}) & Science QA & - & 5 & MC & Acc & - & - & - & - & - & - & - \\
\rowcolor{ai2offwhite} & MMLU Humanities (\citeyear{hendryckstest2021}) & General QA & - & 5 & MC & Acc & - & - & - & - & - & - & 13 \\
\rowcolor{ai2offwhite} & MMLU Social Sci. (\citeyear{hendryckstest2021}) & General QA & - & 5 & MC & Acc & - & - & - & - & - & - & 12 \\
\rowcolor{ai2offwhite} & MMLU Other (\citeyear{hendryckstest2021}) & General QA & - & 5 & MC & Acc & - & - & - & - & - & - & 14 \\
\rowcolor{ai2offwhite} & CSQA (\citeyear{talmor-etal-2019-commonsenseqa}) & Commonsense QA & - & 5 & MC & Acc & - & - & - & - & - & - & - \\
\rowcolor{ai2offwhite} & PiQA (\citeyear{Bisk_Zellers_Le_bras_Gao_Choi_2020}) & Physical QA & - & 5 & MC & Acc & - & - & - & - & - & - & - \\
\rowcolor{ai2offwhite} & SocialIQA (\citeyear{sap-etal-2019-social}) & Social QA & - & 5 & MC & Acc & - & - & - & - & - & - & - \\
\rowcolor{ai2offwhite} & DROP Gen2MC* (\S\ref{sec:new-evaluation-benchmarks}; \citeyear{dua-etal-2019-drop}) & Passage QA & - & 5 & MC & Acc & - & - & - & - & - & - & - \\
\rowcolor{ai2offwhite} & Jeopardy Gen2MC* (\S\ref{sec:new-evaluation-benchmarks}; \citeyear{mosaic-jeopardy}) & Trivia QA & - & 5 & MC & Acc & - & - & - & - & - & - & - \\
\rowcolor{ai2offwhite} & NaturalQs Gen2MC* (\S\ref{sec:new-evaluation-benchmarks}; \citeyear{kwiatkowski-etal-2019-natural}) & General QA & - & 5 & MC & Acc & - & - & - & - & - & - & - \\
\rowcolor{ai2offwhite} & SQuAD Gen2MC* (\S\ref{sec:new-evaluation-benchmarks}; \citeyear{rajpurkar-etal-2016-squad}) & General QA & - & 5 & MC & Acc & - & - & - & - & - & - & - \\
\rowcolor{ai2offwhite} & CoQA Gen2MC* (\S\ref{sec:new-evaluation-benchmarks}; \citeyear{reddy-etal-2019-coqa}) & Conversation QA & - & 0$^\dagger$ & MC & Acc & - & - & - & - & - & - & - \\
\rowcolor{ai2offwhite} \multirow{-12}{*}{\rotatebox[origin=c]{90}{\textit{Non-STEM QA}}} & Basic Skills* (\S\ref{sec:new-evaluation-benchmarks}) & Basic QA & - & 5 & MC & Acc & - & - & - & - & - & - & 6 \\
\rowcolor{lightgrey} & HellaSwag (\citeyear{zellers-etal-2019-hellaswag}) & Language Modeling & - & 5 & RC$_\text{per-char}$ & Acc & - & - & - & - & - & - & - \\
\rowcolor{lightgrey} & WinoGrande (\citeyear{Sakaguchi_Le_Bras_Bhagavatula_Choi_2020}) & Language Modeling & - & 5 & RC$_\text{none}$ & Acc & - & - & - & - & - & - & - \\
\rowcolor{lightgrey} & Lambada (\citeyear{paperno2016lambada}) & Language Modeling & - & 0 & RC$_\text{per-char}$ & Acc & - & - & - & - & - & - & - \\
\rowcolor{lightgrey} & Basic Skills* (\S\ref{sec:new-evaluation-benchmarks}) & Basic QA & - & 5 & RC$_\text{per-token}$ & Acc & - & - & - & - & - & - & 6 \\
\rowcolor{lightgrey} & DROP (\citeyear{dua-etal-2019-drop}) & Passage QA & - & 5 & GenQA & F1 & 0 & 1 & - & 100 & - & - & - \\
\rowcolor{lightgrey} & Jeopardy (\citeyear{mosaic-jeopardy}) & Trivia QA & - & 5 & GenQA & F1 & 0 & 1 & - & 50 & - & - & - \\
\rowcolor{lightgrey} & NaturalQs (\citeyear{kwiatkowski-etal-2019-natural}) & General QA & - & 5 & GenQA & F1 & 0 & 1 & - & 50 & - & - & - \\
\rowcolor{lightgrey} & SQuAD (\citeyear{rajpurkar-etal-2016-squad}) & General QA & - & 5 & GenQA & F1 & 0 & 1 & - & 50 & - & - & - \\
\rowcolor{lightgrey} \multirow{-9}{*}{\rotatebox[origin=c]{90}{\textit{GenQA}}} & CoQA (\citeyear{reddy-etal-2019-coqa}) & Conversation QA & - & 0$^\dagger$ & GenQA & F1 & 0 & 1 & - & 50 & - & - & - \\
\rowcolor{ai2midwhite}\multicolumn{14}{c}{\rule{0pt}{1pt}} \\[-9pt]
\rowcolor{ai2midwhite}\multicolumn{14}{c}{\textbf{Base Held-out Suite}} \\
\rowcolor{ai2midwhite}\multicolumn{14}{c}{\rule{0pt}{1pt}} \\[-9pt]
\rowcolor{ai2offwhite} & MMLU Pro (\citeyear{wang2024mmlu}) & - & - & 5 & MC & Acc & - & - & - & - & - & - & 13 \\
\rowcolor{ai2offwhite} & LBPP* (\citeyear{matton-etal-2024-leakage}) & - & - & 0 & Code Exec & pass@k & 0.6 & 0.6 & - & 4096 & 1 (32) & - & - \\
\rowcolor{ai2offwhite} & Deepmind Math* (\citeyear{saxton2019analysing}) & 5500 & - & 5 & CoT EM & pass@k & 0.6 & 0.6 & - & 2048 & 1 (1) & - & - \\
\rowcolor{ai2offwhite} & BigBench Hard (\citeyear{suzgun2022challenging}) & - & - & 3 & CoT EM & Acc & 0.6 & 0.6 & - & 512 & 1 (1) & - & 55 \\

\bottomrule
\end{tabular}
}
\end{scriptsize}
  \caption{
  \textbf{Details of the \olmothree base evaluation suite}. 
  Tasks were formatted as multiple-choice (MC), rank choice (RC, following the setup in \citet{olmes}), short-form generative (GenQA), chain-of-thought with exact-match scoring (CoT EM), code execution (Code Exec) or fill-in-the-middle coding (FIM).
  We use * to indicate new additions to the base \olmotoo suite \citep{olmo20242olmo2furious}, $^\dagger$ for  tasks with few-shot examples already specified within each instance, and $^\alpha$ for tasks with human-written few-shot examples.
  }
  \label{tab:task-details}
\end{table*}

\FloatBarrier

\subsection{Base Model Additional Decontamination Details}
\label{app:decon}

\textit{{\bf{Important}:} this section is adapted from the documentation of the \href{https://github.com/allenai/decon}{\texttt{decon}} package; for up to date information, please consults the official documentation: } \href{https://github.com/allenai/decon/blob/main/doc/simple-details.md}{\path{github.com/allenai/decon/doc/simple-details.md}}
\vspace{1em}

Evals provide measurable outcomes for model capabilities. We hope that these are meaningful measurements. When evals leak into training data we run the risk of overfitting on evals.

\subsubsection{Definitions and Preliminaries}

Training data and evals both consist of variable length token sequences. Contamination is a sufficient presence of a given eval sequence $e$ in a given training sequence $t$.

We characterize the problem as an approximate substring search for $e$ in $t$ for all $e \in E, t \in T$.

Our goal is to partition the set $T \times E$ into the set of contaminated documents, denoted as $C$, and the set of pure documents, denoted as $P$.

We note that $|T| \gg |E|$ and generally $C$ is very sparse within $T$, as $|C| \ll |P|$.

Our goal is to call whether any training sequence $t$ is derived directly from an eval sequence $e$. This involves distinguishing direct derivation of $t$ to $e$ from both noise and any source material for $e$.

\subsubsection{Example of Contamination}

There is great diversity in the format and purpose of evaluation suites.

\texttt{decon} is fundamentally counting tokens, so it does not consider the intent or semantics of eval instances. But it does leverage the inherent structure of evals to better distinguish between sequences that originate from source material and those that are derived directly from evals.

\begin{figure}[ht]
\begin{lstlisting}
// Eval
{"question": "What year was the Eiffel Tower constructed?", "answer": "1889"}

// Training Document
{"text": "Welcome to 1000facts. 1. What year was the Eiffel Tower constructed? A: 1889"}
\end{lstlisting}
\caption{Example of knowledge eval task.}
\label{fig:json-config}
\end{figure}

Knowledge evals frequently have shorter answers.

\begin{figure}[ht]
\begin{lstlisting}
// Eval
{
  "question": "Solve for x: 2x + 5 = 15",
  "answer": "To solve 2x + 5 = 15, subtract 5 from both sides to get 2x = 10, then divide by 2 to get x = 5"
}

// Training Document
{"text": "Here's a math problem solution: To solve 2x + 5 = 15, subtract 5 from both sides to get 2x = 10, then divide by 2 to get x = 5. This demonstrates basic algebraic manipulation."}
\end{lstlisting}
\caption{Example of reasoning eval eval task.}
\label{fig:json-config}
\end{figure}

Reasoning evals frequently have longer answers with a much larger sets of potential token sequences.

\begin{figure}[ht]
\begin{lstlisting}
// Eval
{
  "passage": "The Eiffel Tower, a landmark in Paris, France, was constructed in 1889. It is a global cultural icon. It receives over 6 million visitors each year.",
  "question": "What year was the Eiffel Tower constructed?",
  "answer": "1889"
}
\end{lstlisting}
\caption{Example of retrieval eval task.}
\label{fig:json-config}
\end{figure}

Retrieval evals frequently have a substantial passage from source material which acts as an almost input to a program selected by the question component.

\subsubsection{Eval Normalization}

\texttt{decon} normalizes all eval instances into question (Q), answer (A), and passage (P) components. A given eval split may hold out an answer and may or may not contain a passage depending on the task.

An eval instance can be described as having a Q, QA, QP, or QAP composition.

\begin{itemize}
    \item {\bf{Question}} All eval instances to be decontaminated contain a question, and it serves as the primary vessel for information to describe the task. \texttt{decon} uses the question field for initial identification of contamination clusters. Questions with substantial information content and a strong match are sufficient to call contamination.
    \item {\bf{Answer}} While the answer of an eval is important for measuring whether a model has learned a specific task, in the context of decontamination the answer primarily serves to provide supporting evidence of contamination. This is particularly important for questions with low information content or those that have edits.
    \item {\bf{Passage}} The passage, often derived from reference documents, is not a strong indicator of contamination, but in conjunction with a substantial question and answer match, further supports a contamination call.
\end{itemize}

\subsubsection{Decon Implementation}

We can now describe a computational tractable definition of contamination.
We start with the simplest scenario, evals that only have a question component Q, and later extend the approach for QA, QP, and QAP scenarios.

\paragraph{Detecting contamination} Scoring segments of training documents against evals is somewhat problematic because there is substantial variation in the length of eval and training documents.

\paragraph{Cluster discovery} We start by defining a contamination cluster as a substring of a training document and a set of candidate evals which have at least 1 matching ngram.

We discover clusters by sequentially sampling training document ngrams and checking for a hit in an inverted index which resolves ngrams to eval document ids. Upon an initial hit we expand left and right from the initial hit index until we observe a certain number misses, representing inserts, deletions, or edits.

The initial hit produces a set of matching document ids, which we call the active set. Each subsequent ngram lookup on traversal produces a set of matching documents from the inverted index, which we call the step set. We use the intersection between the active set and the step set to identify which documents in the active set hit for a given step. Once a specific document reaches 11 misses, it is removed from the active set. We repeat this process until the active set is empty or we reach the training document boundaries. At each step we accumulate the unique ngrams matched scoped by eval document id. The end result is a map of document ids to a set of unique ngram shingle matches.

\paragraph{IDF-weighted overlap}
Contamination scoring uses inverse document frequency (IDF) weighting:

\begin{equation*}
O = \frac{\sum_{x \in U_t \cap U_e} \text{idf}(x)}{\sum_{y \in U_e} \text{idf}(y)}
\end{equation*}

where $U_t$ is the set of unique n-grams in the training document segment and $U_e$ is the set of unique n-grams in the evaluation document.

\paragraph{Cluster match length decay} Less informative short texts require stronger matches.

\begin{equation*}
O' = O \times \begin{cases} 1 & \text{if } L \le L_\text{start} \\ 1 - 0.2 \frac{L - L_\text{start}}{L_\text{end} - L_\text{start}} & \text{if } L_\text{start} < L < L_\text{end} \\ 0.8 & \text{if } L \ge L_\text{end} \end{cases}
\end{equation*}

By default \texttt{L\_start} is set by the configuration \texttt{perfect\_match\_decay\_start: 20} and \texttt{L\_end} is set by the configuration \texttt{perfect\_match\_decay\_end: 50}.

\paragraph{Cluster discovery threshold} For efficiency we check that the question match $O'$ exceeds the minimum question match required to ultimately call contamination. Every candidate contamination that exceeds this value will get a complete scoring, which includes answer and passage information.

\begin{figure}[htbp]
    \centering
    \includegraphics[width=\textwidth]{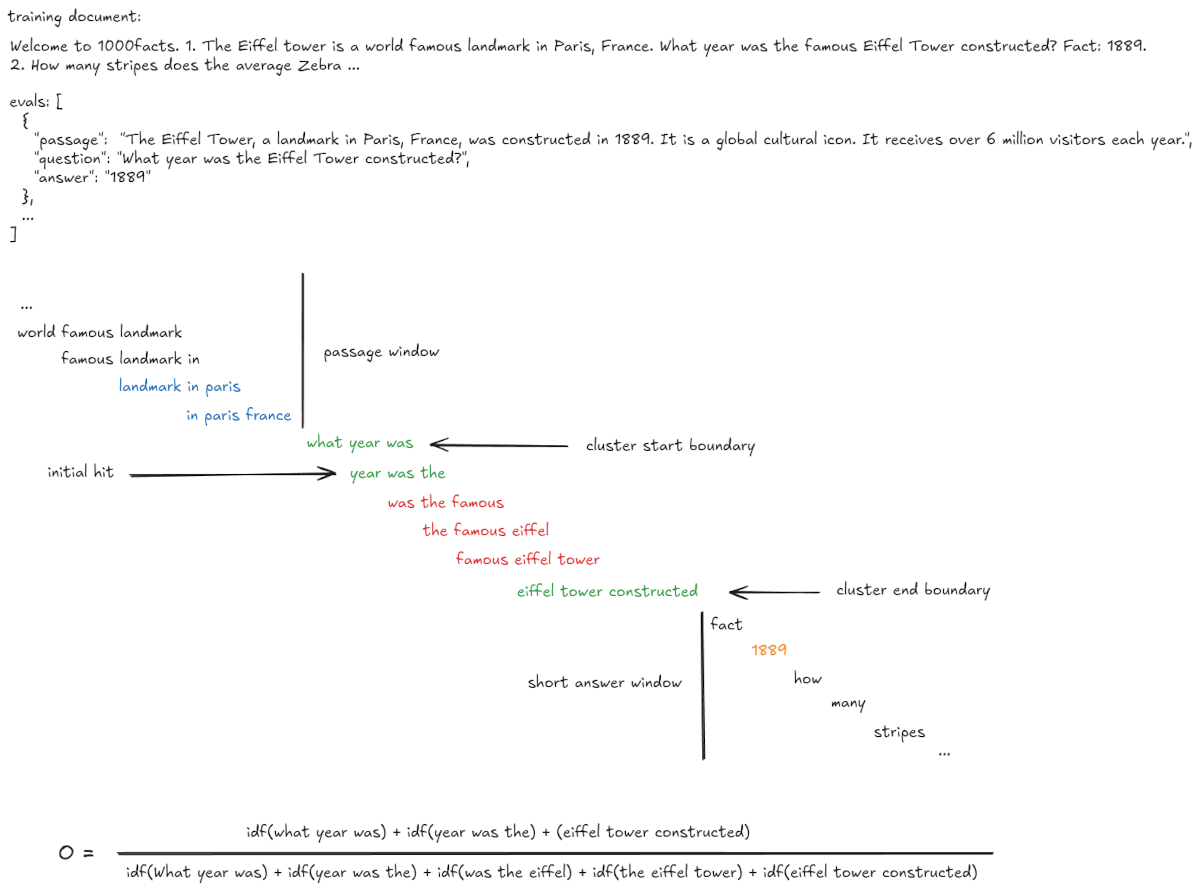}
    \caption{\textbf{Example of trigram processing} for \texttt{decon} pipeline.}
\end{figure}

\paragraph{Non-comparative} Note that we pre-compute the idf sums for evals during index construction. At detection time we sum the calculated idfs for matching ngrams to produce the overlap ratios. There is no string to string comparison. We rely on the nature of n-gram shingles for sequence matching. The probability of having a substantial ngram shingle overlap is low, and while degenerate cases are possible, they have not been observed in practice.

\paragraph{Inverted index} Because $|E|$ is relatively small, we build an inverted index in memory which maps ngrams to document ids.  We use a two tiered index, the first maps a \texttt{u64} hash to a \texttt{u32} sequential id assigned at index construction. And the second tier maps the n-gram id to a set of document ids. This oddity is done to achieve performant membership tests of training ngrams in the significantly smaller set of observed eval ngrams. Consider that the $|G_\text{tn}| \ll |G_\text{en}|$, so the supermajority of ngram lookups are misses, and skipped. The \texttt{u32} sequential id is empirically more performant than a one-tiered lookup with document id sets as values.

\paragraph{Hot n-grams} Cluster discovery begins with an initial hit in the inverted index. While the supermajority of ngrams samples are misses, there are some extremely common ngrams present in the eval texts. Because the ngrams are so common, the probability of a initial hit leading to a true instance of contamination is low. As an optimization we do not start contamination cluster expansion on hot ngram hits, but rather switch our sampling rate to 1, and traverse the training document by single tokens until we observe a miss or non-hot ngram hit.

\subsubsection{Scoring System}

Scores combine question, answer, and passage overlaps with adaptive weights based on the length of components:

\begin{itemize}
    \item {\bf{QAP}} (all components): 0.7 question, 0.2 answer, 0.1 passage
    \item {\bf{QA}} (no passage): 0.75 question, 0.25 answer
    \item {\bf{QP}} (no answer): 0.85 question, 0.15 passage
    \item {\bf{Q}} (question only): 1.0 question
\end{itemize}

\paragraph{Length penalty} We penalize short matches based on the length of Q+A+P by scaling down scores for shorter texts, making the contamination threshold effectively harder to reach. Shorter texts get their scores scaled down before threshold comparison. The scaling factor depends on the total token length $L_\text{total}$:

\begin{equation*}
S_\text{final} = S_\text{base} \times \text{scaleFactor}(L_\text{total})
\end{equation*}

Where the scaling factor decreases for shorter texts, making the threshold effectively harder to reach. Perfect scores (1.0) are never penalized.

\paragraph{Confidence adjusted weight} The question component is the core of a contaminated prompt and carries the most weight. But in some cases an eval will have short questions and long answers or a long passage followed by a short question about it.

Because longer sequences with more informative content provide stronger contamination evidence, we adjust component weights based on confidence factors derived from length by reducing the question weight and redistributing it to the answer or passage.

Question confidence, based on unique n-gram count:
\begin{equation*}
C_q = \begin{cases} 0.5 + 0.5 \frac{N_q}{20} & \text{if } N_q < 20 \\ 1 & \text{if } N_q \ge 20 \end{cases}
\end{equation*}

Base weights are adjusted by confidence factors:

\begin{equation*}
W_\text{adjusted} = W_\text{default} \cdot C + W_\text{redistributed}
\end{equation*}

Where low-confidence components redistribute their weight to higher-confidence ones.

\paragraph{Base scores}

\begin{itemize}
    \item {\bf{Q composition}}: $S_\text{base} = O_q$
    \item {\bf{QA composition}}: $S_\text{base} = O_q \cdot W_{q,\text{adjusted}} + O_a \cdot W_{a,\text{adjusted}}$
    \item {\bf{QP composition}}: $S_{base} = O_q \cdot W_{q,\text{adjusted}} + O_p \cdot W_{p,\text{adjusted}}$ 
    \item {\bf{QAP composition}}: $S_\text{base} = O_q \cdot W_{q,\text{adjusted}} + O_a \cdot W_{a,\text{adjusted}} + O_p \cdot W_{p,\text{adjusted}}$
\end{itemize}

\paragraph{Answer proximity}
For QA datasets, contamination requires the answer appears near the question cluster. Short answers use exact token matching; long answers use n-gram overlap with IDF weighting.

\paragraph{Passage proximity}
For datasets with passages, contamination checks if the passage appears within a configurable distance (\texttt{min\_passage\_distance}) from the question cluster. Passages use n-gram overlap with IDF weighting and can tolerate gaps (\texttt{passage\_max\_consecutive\_misses}).

\subsection{Post-Training Additional Training Details}
\subsubsection{Supervised Finetuning Details}
\label{appx:sft-details}

\paragraph{Using OLMo-core infrastructure for SFT Training}
Relative to pretraining, this involves a substantially smaller batch size, different data packing, and masking.
This leads to an ~8x faster training speed than open-instruct, dramatically improving our iteration speed.
We use between 1 and 8 8xH100 nodes, or 1 to 4 8xB200 nodes to train our 7B reasoner and instruct models. We use 32 8xH100 nodes to train our 32B thinking model
As a consequence of using olmo-core, our batch size is now measured in tokens instead of instances, and we train with document packing instead of padding.
We train all of our 7B SFT models with a batch size of 1M tokens and 32B SFT models with a batch size of 4M tokens, for two epochs, with packing, and a 32{,}768 sequence length.
Our hyperparameter settings are also summarized in Table~\ref{tab:sft_training_settings}.

\begin{table}[th]
    \centering
    \begin{small}
    \adjustbox{max width=\linewidth}{
    \begin{tabular}{lccc}
        \toprule
        & {\bf 7B Thinking SFT} & {\bf 32B Thinking SFT} & {\bf 7B Instruct SFT}  \\
        \midrule
        \rowcolor{ai2offwhite}{\bf Total Tokens} & 45.4B & 45.2B & 3.4B \\
        {\bf Learning Rate} & $5.0 \times 10^{-5}$ & $1.0 \times 10^{-4}$ souped with $5.0 \times 10^{-5}$  & $8.0 \times 10^{-5}$ \\
        \rowcolor{ai2offwhite}{\bf Num. GPUs} & 64 & 256 & 8-64 \\
        {\bf Max Sequence Length} & 32K & 32K & 32K \\
        \bottomrule
    \end{tabular}}
    \end{small}
    \caption{\textbf{Training hyperparameters for \olmothreethinking SFT and \olmothreeinstruct SFT.} GPU hours assume NVIDIA H100 accelerator.}
    \label{tab:sft_training_settings}
\end{table}

\subsubsection{Preference Tuning Details}
\label{appx:dpo-details}
\paragraph{Training Settings} Given a preference dataset $\mathcal{D} = \left\{(x,y_c,y_r)\right\}$ of prompts $x$ and corresponding chosen and rejected responses $y_c \succ y_r$, we optimize the model policy $\pi_\theta$ on a length-normalized DPO loss~\citep{lambert2024tulu3}:
\begin{align*}
    \max_{\pi_\theta} \mathbb{E}_{(x,y_c,y_r)\sim\mathcal{D}}\left[ \log \sigma \left( \frac{\beta}{|y_c|}\log\frac{\pi_\theta(y_c | x)}{\pi_{\text{ref}}(y_c|x)} - \frac{\beta}{|y_r|}\log\frac{\pi_\theta(y_r | x)}{\pi_{\text{ref}}(y_r|x)}\right) \right]
\end{align*}
where $\pi_{\text{ref}}$ is the initial reference policy and $\beta$ is a hyperparameter that regularizes learning via an implicit Kullback–Leibler (KL) divergence penalty between the reference policy and the training policy.

We sweep learning rate and preference dataset size, as we observe that performance increases up until some task-dependent optimal optimization point beyond which further tuning hurts (\autoref{fig:ushape_dpo}). All other hyperparameters are kept fixed. See \autoref{tab:dpo_training_settings} for exact hyperparameters. We train our 7B models using 2–4 8xH100 nodes, and our 32B models with 8–16 8xH100 nodes.
\begin{table}[th]
    \centering
    \begin{small}
    \adjustbox{max width=\linewidth}{
    \begin{tabular}{lccc}
        \toprule
        & {\bf 7B Thinking DPO} & {\bf 32B Thinking DPO} & {\bf 7B Instruct DPO}  \\
        \midrule
        \rowcolor{ai2offwhite}{\bf Num. Preference Pairs} & 150K & 200K & 260K \\
        {\bf Num. Epochs} & 1 & 1 & 1 \\
        \rowcolor{ai2offwhite}{\bf DPO $\beta$} & 5 & 5 & 5 \\
        {\bf Learning Rate} & $8.0 \times 10^{-8}$ & $7.0 \times 10^{-8}$  & $1.0 \times 10^{-6}$ \\
        \rowcolor{ai2offwhite}{\bf LR Schedule} & Linear decay & Linear decay & Linear decay \\
        {\bf Warmup Ratio} & 0.1 & 0.1 & 0.1 \\        
        \rowcolor{ai2offwhite}{\bf Num. GPUs} & 32 & 64-128 & 16 \\
        {\bf Batch Size} & 128 & 128 & 128 \\        
        \rowcolor{ai2offwhite}{\bf Max Sequence Length} & 16K & 8K & 16K \\
        \bottomrule
    \end{tabular}}
    \end{small}
    \caption{\textbf{Training hyperparameters for \olmothreethinking DPO and \olmothreeinstruct DPO}. GPU hours assume NVIDIA H100 accelerator.}
    \label{tab:dpo_training_settings}
\end{table}

\subsubsection{Reinforcement Learning Details}
We provide full training curves for our 7B reasoner in~\autoref{fig:olmo3_final_rl_run}.
The overall reward increases steadily over training. The KL divergence grows gradually and reflects stronger deviation from the reference policy. The response length becomes longer and stabilizes at a higher level. Domain-specific verifier rewards display consistent gains in math and moderate fluctuations in code. The IfEval reward rises throughout training. The two general-quality verifiers also show clear and sustained improvement. Together, these trends indicate that the policy improves both specialized skills and overall response quality. The full hyperparameters for all RL experiments are provided in in~\autoref{tab:rlvr_training_settings}.

\subsubsection{RL-Zero Details}
\label{ssub:rl-zero}
We detail the prompt used for math in \autoref{fig:rlzero-math-prompt}. Prompts of other domains are quite similar, see the open-instruct codebase for details.

We also compare Olmo 3 RL-Zero 7B to one of the more common benchmarks in RLVR, DAPO \citep{yu2025dapo} in \autoref{fig:rlzero-vs-dapo}. Olmo 3 RL-Zero achieves reasonable performance faster and is also much more compute efficient, making it better for experimentation.

Finally, we compare Olmo RL-Zero 3.1 to the initially released, RL-Zero 3.0 in \autoref{fig:rlzero-3.0-vs-3.1} and see a sizable improvement. There were some minor fixes to loss calculation but the major improvement comes from 1. setting completion length to 16k instead of 12k and 2. \textit{not} masking truncated sequences, one of the components of DAPO \citep{yu2025dapo}. Despite initial results suggesting this masking improved the speed of the trainer (by having fewer completions to train on), we ultimately found that variations in batch size caused by some examples masked out to reduce stability. And without training on overlong negative sequences, completion lengths were higher, on average. We therefore found that any efficiency gains in training speed from masking were outweighed by slowdowns from generating longer sequence lengths.

\begin{figure}[h]
\begin{prompt}{\sans{RL-Zero Math Prompt}}\small
Solve the following math problem step by step. \\
The last line of your response should be the answer to the problem in form Answer: \$Answer (without quotes) where \$Answer is the answer to the problem.\\

\{Math Question\}\\

Remember to put your answer on its own line after "Answer:"
\end{prompt}
\caption{\textbf{RL-Zero Prompt for Math Task}.}
\label{fig:rlzero-math-prompt}
\end{figure}

\begin{figure}[t]
    \centering
    \begin{subfigure}{0.48\linewidth}
    \centering
    \includegraphics[width=\linewidth]{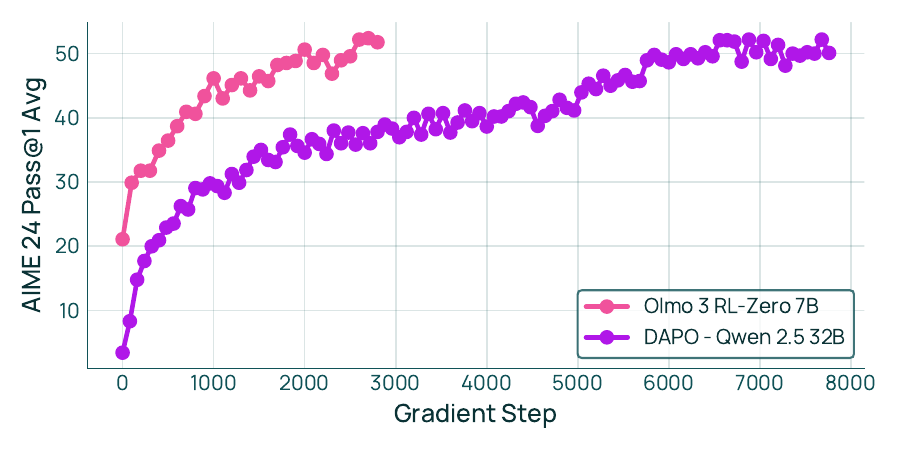}
    \end{subfigure}
    \begin{subfigure}{0.48\linewidth}
    \centering
    \includegraphics[width=\linewidth]{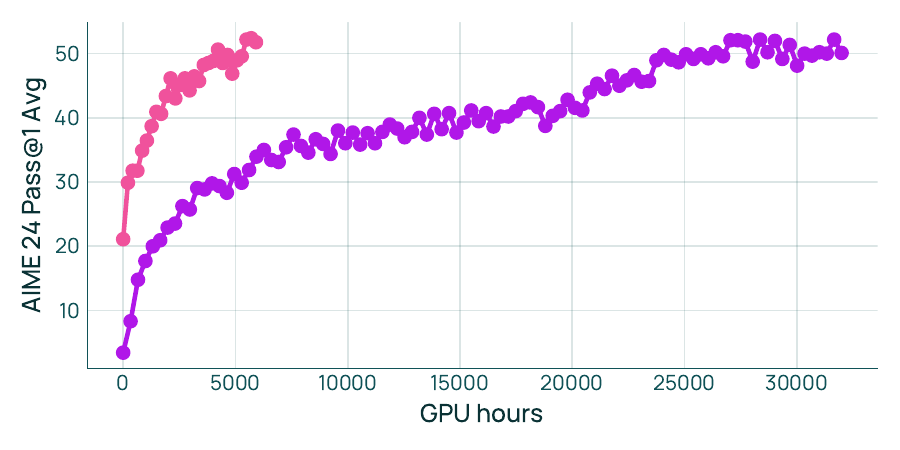}
    \end{subfigure}
    \vspace{-1em}
    \caption{
    \textbf{Olmo 3 RL-Zero 7B vs DAPO}~\citep{yu2025dapo} which leverages Qwen 2.5 32B. We compare the two benchmarks in terms of increase in model performance over training steps as well as GPU hours (exact values and GPU hours for DAPO taken from the \href{https://wandb.ai/verl-org/DAPO\%20Reproduction\%20on\%20verl/runs/0qjd0wap?nw=wmb4qxfht0n}{DAPO reproduction on verl}).
    }
    \label{fig:rlzero-vs-dapo}
\end{figure}

\begin{figure}[t]
    \centering
    \begin{subfigure}{0.48\linewidth}
    \centering
    \includegraphics[width=\linewidth]{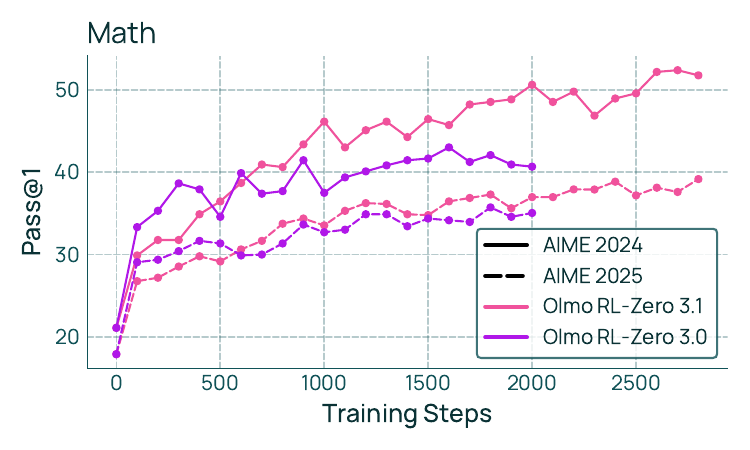}
    \end{subfigure}
    \caption{
    \textbf{\olmothreerlzero vs \olmothreeonerlzero}. We compare our new baseline to the previously released \olmothreerlzeromath on AIME 2024 and 2025, pass@1. Our new setup improves performance more slowly to begin with but outperforms as training goes longer, plateauing at a higher score $\sim 50\%$. 
    }
    \label{fig:rlzero-3.0-vs-3.1}
\end{figure}

\begin{figure}[h]
\begin{prompt}{\sans{Prompt for LLM Judge Reward}}\small

Please act as an impartial judge and evaluate the quality of the answer provided by an
AI assistant to the conversation history leading up to the answer displayed below.
Judge whether the provided answer is good by comparing it to the reference answer. \\ \\

Notes:\\
- Besides comparing to the reference answer, your evaluation should consider factors such as the helpfulness, relevance, accuracy, creativity, appropriate level of detail, and how well the response satisfies the user's explicit constraints or accurately follows their instructions.\\
- Note that sometimes the reference answer is not the only answer. So any valid variation of the reference answer is also acceptable and can get a full score. \\
- If there is a system prompt, ensure the AI answer prioritizes following it. \\
- Begin your evaluation by providing a short explanation. \\
- Be as objective as possible. After providing your short explanation, please output a score on a scale of 1 to 10. \\
- Please adhere to the following format. \\ \\

[Conversation History]\\
\{input\} \\ \\

[AI Answer]\\
\{output\}\\ \\

[Reference Gold Answer]\\
\{label\}\\ \\

[Your judgement]\\
Respond in JSON format. \{"REASONING": "[...]", "SCORE": "<your-score>"\}
\end{prompt}
\vspace{-1em}
\caption{\textbf{LLM judge prompt for non-verifiable tasks}.}
\label{llm-judge-prompt}
\end{figure}

\begin{table}[th]
    \centering
    \begin{small}
    \adjustbox{max width=\linewidth}{
    \begin{tabular}{lcccccc}
        \toprule
        & \textbf{7B Think RL} & \textbf{32B Think RL} & \textbf{7B Instruct RL} & \textbf{7B RL-Zero}  \\
        \midrule
        \rowcolor{ai2offwhite}{\bf Dataset size} & 104{,}869 & 104{,}869 & 171{,}950 & 13{,}314  \\
        {\bf Learning rate} & $1.0 \times 10^{-6}$ & $2.0 \times 10^{-6}$ & $1.0 \times 10^{-6}$ & $1.0 \times 10^{-6}$ \\
        \rowcolor{ai2offwhite}{\bf Minibatches} & 1 & 1 & 4 & 1 \\
        {\bf LR schedule} & constant & constant & constant & constant \\
        \rowcolor{ai2offwhite} {\bf Training steps} & 1{,}400 & 750 & 450 & 2{,}000 \\
        {\bf Max prompt length} & 2{,}048 & 2{,}048 & 2{,}048 & 2{,}048 \\
        {\bf Response length} & 32{,}768 & 32{,}768 & 8{,}192  & 16{,}384 \\
        \rowcolor{ai2offwhite}{\bf Unique prompts per batch} & 64 & 128 & 64 & 32 \\
        {\bf Group size} & 8 & 8 & 8 & 8 \\
        \rowcolor{ai2offwhite}{\bf TIS cap} & - & 2.0 & - & 2.0 \\
        {\bf Sampling temperature} & 1.0 & 1.0 & 1.0 & 1.0 \\
        \rowcolor{ai2offwhite}{\bf Clip-lower} & 0.2 & 0.2 & 0.2 & 0.2 \\
        {\bf Clip-higher} & 0.272 & 0.272 & 0.272 & 0.272 \\
        \rowcolor{ai2offwhite} {\bf Num learner GPUs} & 16 & 64 & 8 & 8 \\
        {\bf Num actor GPUs} & 56 & 160 & 56 & 64 \\
        \rowcolor{ai2offwhite} {\bf GPUs per actor (TP)} & 1 & 8 & 1 & 1 \\
        {\bf Max asynchrony} & 1 & 8 & 8 & 8  \\
        \bottomrule
    \end{tabular}}
    \end{small}
    \caption{\textbf{RL training hyperparameters for \olmothreethinking, \olmothreeinstruct and \olmothreerlzero.} GPU hours assume NVIDIA H100 accelerator.}
    \label{tab:rlvr_training_settings}
\end{table}

\begin{figure}
    \centering
    \adjustbox{max width=0.8\linewidth}{\includegraphics{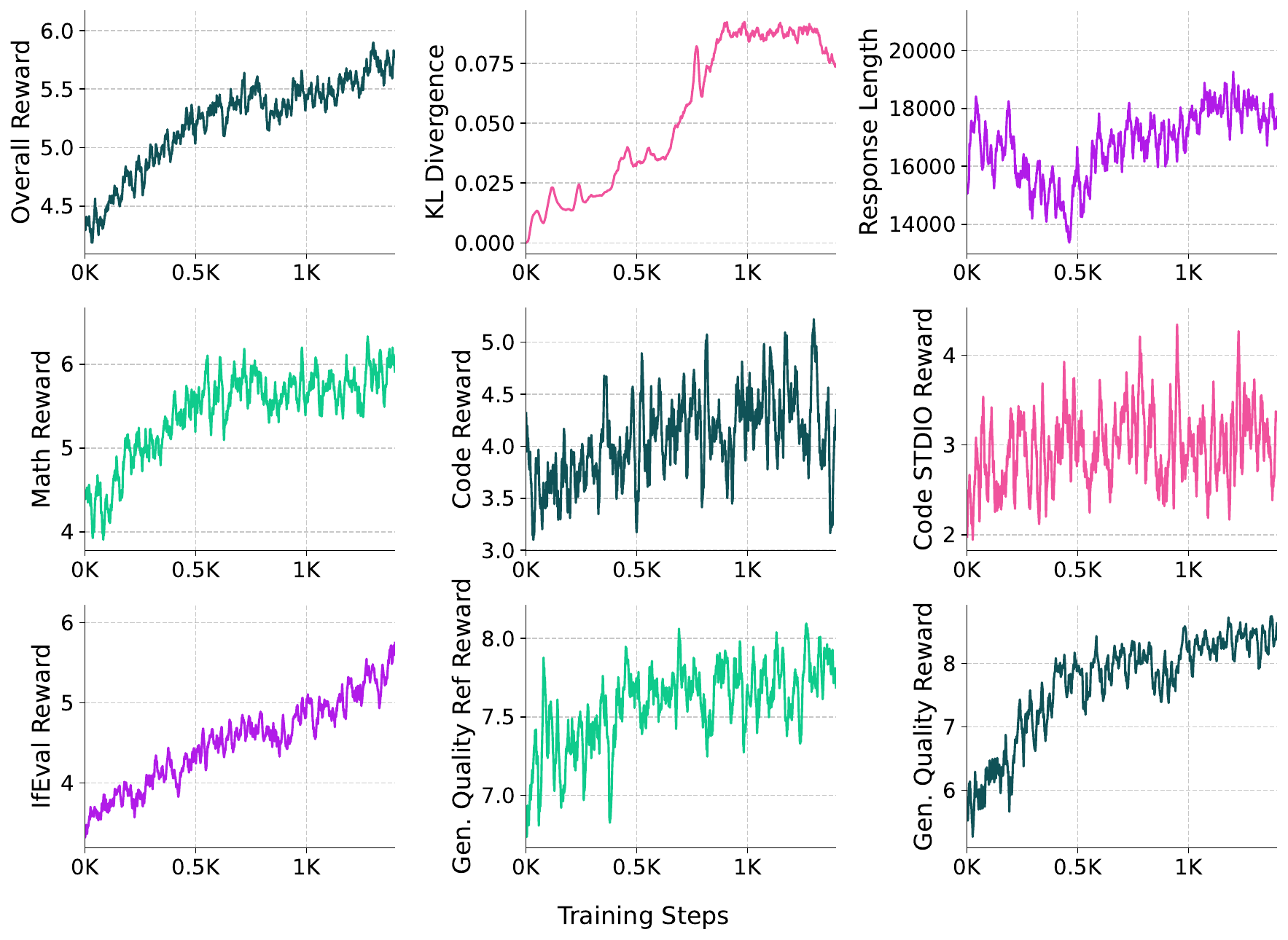}}
    \caption{\textbf{Reward, KL, response length, and per-verifier reward} over the final RL run for \olmothreethinking.}
    \label{fig:olmo3_final_rl_run}
\end{figure}

\subsection{Post-Training Additional Data Details}

\subsubsection{Filtering for \dolcithink-SFT}
\label{appendix:filtering-think-sft}

In this section we detail the filtering methods created primarily for training \olmothreethinking, which was also used for mid-training and \olmothreeinstruct data.
Each phase of filtering would remove 0-1\% of data across most available or generated reasoning traces.
Some data, such as Nvidia's Nemotron Post-training datasets~\citep{NemotronPostTrainingDatasetV1} had very few samples removed relative to their peers.

\begin{enumerate}

\item {\bf Source filtering}
We perform some filtering to remove non-compliant licenses or data that will not be useful.
E.g. for GeneralThoughts traces used in mid-training, we filtered to only commercially friendly licensed prompts.
For OpenThoughts2, we removed ShareGPT prompts due to questionable provenance (as done in Tulu 3).
For LlamaNemotron Post-Training we filter to only reasoning samples from DeepSeek and Qwen that have not been touched by Llama models.

\item {\bf Format filtering}
We remove truncated answers (i.e. if they have <think> and no </think>) and empty outputs (empty responses).
Implementation is available at \href{https://github.com/allenai/open-instruct/blob/7ba4cd0/scripts/data/filtering_and_updates/filter_cots.py}{\path{github.com/allenai/open-instruct//scripts/data/filtering_and_updates/filter_cots.py}}

\item {\bf Domain specific accuracy filtering} We check accuracy for many domains, such as precise instruction following, code, or math.
Additionally, for chat domains we use included metadata in some datasets such as Wildchat to remove responses or prompts tagged as unsafe.
Implementation is available at \href{https://github.com/allenai/open-instruct/blob/7ba4cd0/scripts/data/filtering_and_updates/filter_wildchat.py}{\path{github.com/allenai/open-instruct/scripts/data/filtering_and_updates/filter_wildchat.py}}

\item {\bf General content filters}
Here we remove mention of date cutoffs to try and avoid hallucinations of model characteristics and any mention in the user prompt or completion that indicates the date is to or from any model.
Maintaining identity of models trained on heavily distilled data takes a meaningful amount of data work and system prompt design.
Implementation is available at  \href{https://github.com/allenai/open-instruct/blob/7ba4cd0/scripts/data/filtering_and_updates/filter_datasets_sequential.sh}{\path{github.com/allenai/open-instruct/scripts/data/filtering_and_updates/filter_datasets_sequential.sh}}

\item {\bf Repetition filtering}
Many open-weights reasoning models have tendencies to perform extreme repetitions, even in thinking traces that result in a correct answer.
In particular, we find that ~.1\% of responses from QwQ have mass repetition.
We filter this roughly by searching for heavily repeated (~10x+) sentences, paragraphs, or (~50x+) phrases.
Implementation is available at \href{https://github.com/allenai/open-instruct/blob/7ba4cd0/scripts/data/filtering_and_updates/filter_ngram_repetitions.py}{\path{github.com/allenai/open-instruct/scripts/data/filtering_and_updates/filter_ngram_repetitions.py}}

\item {\bf Chinese language filtering}
In order to encourage \olmothreethinking to stay in its intended language of English, we remove any post-training responses with 5\% or higher prevalence of Chinese characters by searching over the range of Unicode character range of common Chinese characters.
Implementation is available at \href{https://github.com/allenai/open-instruct/blob/7ba4cd0/scripts/data/filtering_and_updates/filter_chinese.py}{\path{github.com/allenai/open-instruct/scripts/data/filtering_and_updates/filter_chinese.py}}

\end{enumerate}
\subsubsection{Tool-use data} \label{sec:appendix_tool_use_data}
\paragraph{Additional details about the Science QA dataset} Citation graph-based queries are produced by prompting GPT-5 in a few-shot setup to create query templates, e.g., \texttt{What are the top-three most cited papers by \{AUTHOR\} on \{TOPIC\}?} which are subsequently instantiated with real paper entities. Content-based questions are generated by a GPT-5-based agent equipped with the ASC server, which retrieves relevant papers and formulates grounded questions that can be answered using retrieved text. For both types of queries, to obtain corresponding tool-use trajectories we employ a GPT-4.1-mini agent with access to the same ASC server. All tool call outputs are derived from actual environment responses rather than synthetic completions.

\paragraph{Additional details about the Web Search QA dataset} Given the varied quality of real-world queries, GPT-5 is employed to rate each query drawn from existing open-access benchmarks on a five-point scale assessing (i) whether it calls for comprehensive long-form responses, (ii) factual verifiability, and (iii) the degree of search required. Only queries scoring 4 or 5 on these criteria are retained. We then use an agent equipped with web search and browsing via the Serper API, and scientific snippet retrieval via ASC to generate tool-use trajectories for these queries. This agent is instructed with tool specifications and step-by-step search instructions, resulting in detailed trajectories containing both tool calls and environment outputs. We then filter out trajectories that yield incorrect answers (where ground truth is available), and only keep trajectories that adhere to the expected output format. Additionally, since the environment outputs for the webpage-fetching tool of the Serper API are quite long (typically entire webpages), we used GPT-5 to summarize the content of the web pages and only retained the summaries in the training data.

\paragraph{Additional details about simulated interaction trajectories} We run various post-hoc checks on synthesized datasets to verify whether the generated trajectories adhere to the prompts, and filter the dataset to create SimFC. We filter out trajectories where the function calls include functions not part of the presented APIs. Our data-synthesis prompts explicitly target multi-turn, multi-step, parallel function calls (i.e., multiple calls per assistant turn) and refusals, and we filter out the trajectories that do not conform to such requirements specified in the prompts.

\begin{figure}[h]
\begin{prompt}{\sans{Prompt for Generating Multi-Turn Function-calling Interactions}}\footnotesize

You are provided an API with the details of the functions shown in a JSON format. Use this API to write a simulated interaction between a user, an assistant that can call the functions in the API, and the environment. The interaction should refer to three roles: \texttt{"user"}, \texttt{"assistant"}, and \texttt{"environment"}. Their messages should be represented as Python dicts with \texttt{"role"} and \texttt{"content"} fields. \\

If the assistant is making function calls, they should be shown under a \texttt{"function\_calls"} field instead of the \texttt{"content"} field. The interaction should start with a user request, contain multiple steps of the assistant making function calls while interacting with the user for additional inputs, and should conclude with the assistant performing the user's requested action. Please generate a simulated interaction with at least 5 function calls. Ensure that at the end of each turn, the assistant should address the request of the user by creating an assistant message with a text in the \texttt{"content"} field. \\

Here is an example:

\begin{lstlisting}[basicstyle=\ttfamily\footnotesize,breaklines=true]
API:
[
  {"name": "get_borrowed_books", "description": "Get borrowed books by user ID",
    "parameters": {"user_id": {"type": "int"}}},
  {"name": "get_user_info", "description": "Get user information",
    "parameters": {"prefix": {"type": "str", "required": false},
        "email": {"type": "str", "required": false}}},
  {"name": "get_late_fines", "description": ...}
]

INTERACTION:
[
  {"role": "user", "content": "How many users with the name Yoda exist?"},
  {"role": "assistant", "function_calls": "get_user_info(prefix=`Yoda')"},
  {"role": "environment", "content": "{\"results\": [{\"id\": 23}]}"},
  {"role": "assistant", "content": "There is one user with that name."},
  {"role": "user", "content": "How many books have they borrowed?"},

  ... additional turns ...

  {"role": "assistant", "content": "Luke Skywalker has borrowed one book."}
]
\end{lstlisting}

Here is the real task:

\begin{lstlisting}[basicstyle=\ttfamily\footnotesize,breaklines=true]
API: {}
INTERACTION:
\end{lstlisting}

\end{prompt}
\vspace{-1em}
\caption{\textbf{Illustrative prompt for generating multi-turn function-calling interactions} with simulated environment feedback (prompt has been truncated for readability).}
\label{fc-multi-turn-prompt}
\end{figure}

\begin{figure}[h!]
\begin{prompt}{\sans{Prompt for Generating Function Calling Refusals}}\footnotesize

You are given an API function described in JSON format. Your task is to write a simulated conversation between a user and an assistant.
First identify the domain of the API, and then create a user request that is similar in domain but still unaddressable by the API. \\

In this conversation:

1. The user makes a request that is slightly related to the capabilities of the API, but still unaddressable by the API.

2. The domain of the user request should be very similar to the API's capabilities. If it's about math, then the request should also be about math.

3. The assistant refuses the request and explains clearly why it cannot be fulfilled, referencing the actual API functions.

4. The assistant should not hallucinate functionality or attempt to fulfill the request.

5. The explanation must be concise, accurate, and polite.

6. The dialogue should be \texttt{brief but complete}, showing a realistic interaction.

7. Format the output as a realistic, short conversation between the user and assistant.

8. There is no need to put environment outputs.

9. Use an imperative tone and include concrete values (e.g., ``Compute the perimeter of a rectangle with length 10 and width 5''). \\

Format the output as a dialogue, alternating between the user and the assistant. \\

{\bf Example 1}

\begin{lstlisting}[basicstyle=\ttfamily\footnotesize,breaklines=true]
API:
[
  {"name": "get_user_info", "description": "Get user information",
    "parameters": {"prefix": {"type": "str", "required": false},
        "email": {"type": "str", "required": false}}},
  {"name": "get_borrowed_books", "description": "Get borrowed books by user ID"}
]

INTERACTION:
[
  {"role": "user", "content": "Sell the book `The Little Prince'"},
  {"role": "assistant", "content": "I'm sorry, but I can't sell books. Based on the APIs, I can help with retrieving user info or checking borrowed books."}
]
\end{lstlisting}

{\bf Example 2} \\

    ... additional examples ... \\\\

Here is the real task:

\begin{lstlisting}[basicstyle=\ttfamily\footnotesize,breaklines=true]
API: {}
INTERACTION:
\end{lstlisting}

\end{prompt}
\vspace{-1em}
\caption{\textbf{Illustrative prompt for generating function-calling refusals}, i.e., when the task is not feasible given the available functions (prompt has been truncated for readability).}
\label{fc-refusals-prompt}
\end{figure}

\begin{table}[!h]
  \centering
  \setlength\tabcolsep{4pt}
  {\footnotesize
  \begin{tabular}{l r r r r r r r r}
  \toprule
  {\bf Dataset} & {\bf Original} & {\bf Format} & {\bf Domain} & {\bf General} &
   {\bf Content} & {\bf Repetition} & {\bf Chinese} & {\bf Final} \\
   & {\bf Size} & {\bf Filtering} & {\bf Filtering} & {\bf Filtering} &
  {\bf Filtering} & {\bf Filtering} & {\bf Filtering} & {\bf Size} \\
  \midrule
  \rowcolor{ai2offwhite} WildChat (Tülu 3) & 57,407 & 1.61\% & 14.57\% & 0.75\% & 3.10\% & -- & 1.09\% &
  45,917 \\
  WildChat (New) & 74,997 & 1.53\% & 48.09\% & 0.80\% & 3.13\% & 0.02\% & 1.16\% & 36,417 \\
  \rowcolor{ai2offwhite} OpenAssistant1 & 7,094 & 0.08\% & -- & 0.22\% & -- & -- & 3.86\% & 6,800 \\
  OpenThoughts3-Regen & 1,200,000 & 3.22\% & -- & 0.00\% & -- & < 0.01\% & 0.04\% & 1,160,972
  \\
  \rowcolor{ai2offwhite} Persona Precise IF & 224,448 & 0.19\% & -- & 0.03\% & 0.29\% & < 0.01\% &
  0.08\% & 223,123 \\
  Val Precise IF (QwQ) & 286,003 & -- & -- & -- & 0.62\% & < 0.01\% & 1.17\% & 135,851 \\
  \rowcolor{ai2offwhite} Synthetic-2-SFT-Verified & 104,913 & 0.01\% & -- & 0.06\% & -- & < 0.01\% & 0.32\% &
  104,569 \\
  Saurabh Code Mix & 884,767 & -- & -- & -- & -- & < 0.01\% & < 0.01\% & 884,570 \\
  \rowcolor{ai2offwhite} CoCoNot & 10,460 & 0.57\% & -- & 1.57\% & -- & -- & 0.10\% & 10,227 \\
  WildGuard & 38,794 & 0.37\% & -- & 1.17\% & 0.54\% & < 0.01\% & 0.12\% & 38,315 \\
  \rowcolor{ai2offwhite} WildJailbreak & 41,420 & 0.13\% & -- & 0.21\% & 0.61\% & -- & < 0.01\% &
  41,100 \\
  Aya & 98,863 & 0.15\% & -- & 1.70\% & -- & < 0.01\% & 5.62\% & 98,598 \\
  \rowcolor{ai2offwhite} TableGPT & 4,982 & 0.02\% & -- & 0.00\% & -- & -- & 0.06\% & 4,981 \\
  \bottomrule
  \end{tabular}}
  \vspace{3pt}
  \caption{\textbf{Filtering statistics} showing percentage of prompts removed at each major
  filtering stage for reasoning datasets. ``--'' indicates filtering was not applicable or no
  samples were removed.}
  \label{tab:filtering_stats}
  \end{table}

\subsubsection{Coding Data Synthesis Pipeline}
\label{appx:code-data}

To construct reinforcement learning (RL) data for code, we required pairs of \emph{(problem, test cases)}. We curate a diverse set of prompts for coding problems, including AceCoder~\citep{zeng2025acecoder}, Klear-Reasoner Code~\citep{su2025klear}, Nemotron Post-training Code \citep{nvidia2025nemotron_post_training_dataset}, SYNTHETIC-2 code \citep{primeintellect2025synthetic2}, Open-Code Reasoner \citep{ahmad2025opencodereasoning}. We use the klear-reasoner and SYNTHETIC-2 test cases directly.  For the other datasets, we run prompts through the following synthetic data pipeline:
\begin{itemize}[leftmargin=2.5em]
     \item {\bf Problem rewriting.} Given a coding problem, we first prompted GPT-4.1 to rewrite the description so that it either (a) included a function signature, or (b) explicitly specified that the solution should read from and write to standard input/output (stdio)
     \item {\bf Solution generation.} GPT-4.1 was then prompted to provide a corresponding solution. Depending on the problem type, this was either a Python function matching the given signature, or a program reading from and writing to stdio. When the original problem source included a reference solution, we included it in the prompt
     \item {\bf Test case generation.} GPT-4.1 was further prompted to generate test cases in the appropriate format (function-based or stdio-based)
 \end{itemize}

\subsubsection{\dolciinstructdpo Details}
\paragraph{DPO prompt mixing} See \autoref{tab:dpo_mixing_exps} for prompt mixing experiment results.

\paragraph{Model pool for LLM-judged pairs} To create the GPT-judged subset of \dolciinstructdpo, we generate completions on our prompt pool with the following models: gpt-oss-20B, gpt-oss-120B~\citep{agarwal2025gpt}, GPT-4.1-2025-04-14~\citep{gpt4}, Mistral-Small-24B-Instruct-2501, \olmotoo-1B-Instruct, \olmotoo-7B-Instruct, \olmotoo-13B-Instruct, \olmotoo-32B-Instruct~\citep{olmo20242olmo2furious}, Phi4-Mini-Instruct~\citep{Abdin2024Phi4TR}, Gemma3-4B-it, Gemma3-12B-it, Gemma3-27B-it~\citep{team2025gemma3}, Qwen3-Coder-30B-3A (no reasoning), Qwen3-0.6B (no reasoning), Qwen3-1.7B (no reasoning), Qwen3-4B (no reasoning), Qwen3-8B (no reasoning), Qwen3-14B (no reasoning), Qwen3-32B (no reasoning), Qwen3-30B-3A (no reasoning)~\citep{qwen3},  QwQ-32b~\citep{qwen_qwq_32b_2025}, Yi-9B, and Yi-34B~\citep{young2024yi}.

For each prompt, we sample four model completions and judge them via a GPT-4.1 judge with the UltraFeedback judge prompts\footnote{We ran initial experiments employing a GPT-5 judge, but results indicatedthat the GPT-4.1 judge is better.}~\citep{lambert2024tulu3,cui2023ultrafeedback}. To enforce a meaningful delta between chosen and rejected responses, we enforce our judge pipeline to sample responses from exactly two of the following smaller and/or previous generation models which show lower overall performance: \olmotoo-1B-Instruct, \olmotoo-7B-Instruct, Yi-9B, Yi-34B, Phi4-Mini-Instruct,
Qwen3-0.6B (no reasoning), Qwen3-1.7B (no reasoning). Without this intervention (i.e. sample four models from the pool to judge at random), we would have an approximately 33\% chance of sampling at least 2 weak models out of our 4 samples from our model pool for judgment, providing limited contrast in preference pairs. We binarize into preference pairs by selecting the \textit{worst} response out of the four to be rejected, and the best as chosen.

\begin{table}[ht]
\centering
\small
\begin{tabular}{@{}l >{\columncolor{ai2pink!18}}c ccccccccc@{}}
\toprule
& \multicolumn{10}{c}{\textbf{Subset of \olmothreeinstruct Benchmarks}} \\
\cmidrule(lr){2-11}
\textbf{Name} & \cellcolor{ai2pink!18}\textbf{Avg.} & \textbf{MMLU} & \textbf{BBH} & \textbf{GPQA} & \textbf{AGI} & \textbf{MATH} & \textbf{CHE} & \textbf{LCB} & \textbf{IFEval} & \textbf{AE2} \\
\midrule
Development SFT  & 50.1 & 66.3 & 44.2 & 29.9 & 58.6 & 56.2 & 70.0 & 13.8 & {\bf{82.1}} & 29.8 \\
Base mix (uniform* sample) & 54.3 & {\bf{68.1}} & 48.1 & 32.1 & 62.7 & 67.3 & {\bf{68.5}} & 17.0 & 79.3 & {\bf{45.4}} \\
\midrule
Ablate code & 53.6 & 64.7 & {\bf{51.6}} & 33.0 & {\bf{65.2}} & {\bf{67.9}} & 65.9 & 17.7 & 75.8 & 40.6 \\
Ablate math & {\bf{54.4}} & 67.8 & 49.2 & 33.0 & 64.8 & 67.2 & 67.0 & 20.4 & 77.3 & 42.9 \\
Ablate science & 52.8 & 66.4 & 49.9 & 31.7 & 64.2 & 67.0 & 60.0 & 19.8 & 76.3 & 39.6 \\
Ablate chat & 53.1 & 67.1 & 51.3 & 30.6 & 64.8 & 67.6 & 59.3 & {\bf{21.2}} & 76.3 & 39.3 \\
Ablate inst. following & 50.3 & 66.1 & 51.0 & 29.5 & 62.5 & 66.3 & 48.3 & 18.7 & 75.2 & 34.8 \\
Ablate safety & 51.0 & 66.3 & 48.6 & {\bf{34.2}} & 63.5 & 67.3 & 51.0 & 18.1 & 74.7 & 35.4 \\
Ablate misc/SFT unused & 48.3 & 66.6 & 49.9 & 29.7 & 64.2 & 65.3 & 38.6 & 14.9 & 74.1 & 31.2 \\
\midrule
Upsample code & 51.1 & 67.7 & 48.6 & 31.7 & 63.8 & 65.9 & 51.7 & 18.0 & 76.0 & 36.3 \\
Upsample math & 53.3 & 67.5 & 48.6 & 29.5 & 62.3 & 66.4 & 66.7 & 17.5 & 78.4 & 42.6 \\
Upsample chat & 53.0 & 67.0 & 46.8 & 30.6 & 61.6 & 65.7 & 68.3 & 15.6 & 76.9 & 44.7 \\
\bottomrule
\end{tabular}
\caption{\textbf{Development results for DPO prompt domain mixing}. Overall, we find that (1) all prompt domains are useful for performant tuning, but (2) the exact optimal ratios for each domain are challenging to ascertain systematically since prompt domain does not necessarily correspond to the domains in which performance improves. (*)Wildchat is limited to 35\% of the base mix. All other prompts are uniformly sampled.}
\label{tab:dpo_mixing_exps}
\end{table}

\subsection{Post-Training Additional Evaluation Details}
\label{appx:eval-details}
\subsubsection{General Evaluation Settings}
\label{appx:eval-fun}

For post-training, we focus exclusively on generative evaluations, in which we generate completions until a max length is reached or eos token is generated (as opposed to multiple-choice-based evaluations used in pretraining), better matching real-world downstream usage.

Following DeepSeek R1 report~\citep{guo2025deepseek} and Nvidia Nemotron~\citep{adler2024nemotron} we use a sampling temperature of 0.6 and top-p of 0.95.\footnote{We find that both thinking models degenerate quickly when evaluated with low temperatures (as used in \olmotoo), while instruction models can be evaluated at this higher temperature.} We strip thinking traces from the answer text when generated.
We account for the variance this induces in smaller benchmarks (e.g. AIME, which is made up of 30 questions) by taking multiple samples and reporting the overall average performance.
For QA tasks (e.g. BBH, MMLU), we create a unified set of `Olmo 3' regexes for answer extraction, covering a wide variety of potential answer templates.
We additionally update AlpacaEval 2 Length Controlled (LC)~\citep{dubois2024length} to use GPT-4.1 as a judge instead of the original GPT-4-Turbo~\citep{gpt4} both to increase the reliability of the evaluation and to save $\sim$90\% of inference costs.
Importantly, our evaluation settings are {\bf unified across thinker and instruct models}, simplifying our evaluation development process.

\paragraph{Is AlpacaEval useful?}
Certainly! {\bf{AlpacaEval}}, and similar evaluations, such as {\bf{ChatBotArena}}~\citep{zheng2023judging}, {\bf{MT-Bench}}~\citep{zheng2023judging}, {\bf{Arena-Hard}}~\citep{li2024crowdsourced}, etc. are established as {\bf{crucial benchmarks}} for the {\bf{AI industry}}.
Let's delve into the {\bf{pros and cons}} of AlpacaEval:

\begin{itemize}[leftmargin=2.5em, labelsep=0.5em]
    \item[\scale] {\bf{It's not a broken evaluation, it's a trade-off}}. It's {\bf{well established}} that most people enjoy reading language model completions that have a {\bf{bit of flair}} to them. In fact, the {\bf{style of bold, lists, etc.}} can be {\bf{very helpful when skimming information}}. It just can often go over the top---such as when {\bf{too many emoji's}} are included!~\cow~\cowface~\trex
    \begin{itemize}[leftmargin=3.3em, labelsep=0em,nosep]
        \item[{\bf{Pro:}}]~Ease-of-reading and flair
        \item[{\bf{Con:}}]~Over-optimized style
    \end{itemize}
    \item[\trophy] {\bf{We're incentivized to maximize the benchmark---even if we don't like it}}. As a {\bf{smaller lab}}, we need to work hard to {\bf{put our models on the map!}} We don't \textit{love} the style of completions from models scoring high on these benchmarks, but we {\bf{derive so much benefit}} from the {\bf{attention it attracts}}.
    \begin{itemize}[leftmargin=3.3em, labelsep=0em,nosep]
        \item[{\bf{Pro:}}]~Simple comparison to known standards
        \item[{\bf{Con:}}]~Imperfect performance signal 
    \end{itemize}
    \item[\shrug] {\bf{There aren't many better options!}} There are just {\bf{so few evaluations}} that test a model's ability to {\bf{chat with the users reliably}}---and we need to serve the {\bf{most common use case}} if we want {\bf{adoption}}. {\bf{More diversity of benchmarks}}, such as alternatives like multi-turn and instruction following, are slowly helping out understanding.
    \begin{itemize}[leftmargin=3.3em, labelsep=0em,nosep]
        \item[{\bf{Pro:}}]~Common adoption
        \item[{\bf{Con:}}]~Low diversity in chat evaluations
    \end{itemize}
\end{itemize}

\sparkles\ {\bf{Bonus:}} There's something poetic about having LLMs judge LLMs. \devilsmile

In summary, {\bf{we need evaluations like this}} to make sure the model is {\bf{behaving as expected}}.
When it comes to {\bf{balancing style and benchmarks}}, at the end of the day, {\bf{no-one's perfect---not even us}}.

\begin{table}[t]
\centering
\footnotesize
\setlength\tabcolsep{5pt}
\renewcommand{\arraystretch}{0.95}
\adjustbox{max width=\linewidth}{
{\fontsize{8}{8}\selectfont
\begin{NiceTabular}{@{}ll|
C{35pt}C{35pt}P{35pt}|
C{35pt}C{35pt}C{35pt}C{35pt}C{35pt}C{35pt}@{}}
\toprule
& & \multicolumn{3}{c}{\quad \quad \textbf{\texttt{OLMo 3 7B Think}}} & \multicolumn{6}{c}{\textbf{\texttt{Baselines}}} \\
\textbf{Benchmark} &
& \textbf{SFT} & \textbf{DPO} & \textbf{Final Think} & \textbf{OpenThinker3 7B} & \textbf{Nemotron Nano 9B v2} & \textbf{DS-R1 Qwen 7B} & \textbf{Qwen 3 8B} & \textbf{Qwen 3 VL 8B Think} & \textbf{OR Nemotron 7B} \\
\midrule
DoAnythingNow &              & 19.3 & 19.6 & 23.4 & 1.8 & 56.7 & 34.3 & 53.1 & 83.0 & 2.3 \\
HarmBench &         & 67.8 & 72.7 & 75.4 & 26.7 & 69.4 & 50.7 & 74.0 & 81.9 & 20.0 \\
\multicolumn{2}{@{}l}{TrustLLM-JailbreakTrigger} & 64.8 & 65.2 & 72.0 & 2.9 & 62.6 & 50.1 & 56.7 & 77.0 & 6.9 \\
WildJailbreak-Test          & Harmful            & 23.4 & 27.5 & 39.0 & 0.3 & 28.7 & 4.5 & 12.3 & 38.6 & 0.5 \\
WildJailbreak-Test & Benign     & 99.1 & 98.5 & 98.8 & 99.2 & 97.3 & 98.0 & 99.7 & 98.0 & 97.1 \\
WildGuard-Test &        & 90.2 & 93.9 & 93.8 & 48.8 & 88.4 & 69.2 & 82.9 & 93.0 &  42.6\\
XSTest &  & 91.6 & 91.6 & 90.9 & 59.5 & 92.5 & 68.4 & 87.2 & 94.2 & 61.0 \\
\midrule
BBQ   &     Accuracy    & 86.6 & 84.8 & 89.2 & 80.5 & 92.0 & 78.0 & 91.8 & 86.6 & 82.6 \\
BBQ & Bias - Ambig. & 7.3 & 8.4 & 6.5 & 11.3 & 5.8 & 9.4 & 5.5 & 8.9 & 7.1 \\
BBQ & Bias - Disambig. & 1.7 & 1.1 & 1.7 & 2.4 & 0.7 & 2.4 & 1.5 & 1.0 & 2.3 \\
StrongReject  &            & 74.8 & 75.5 & 79.0 & 56.7 & 85.6 & 72.4 & 73.4 & 82.8 &  58.3 \\
Toxigen    &           & 100 & 99.9 & 100 & 97.4 & 100 & 99.7 & 100 & 99.9 & 86.4 \\

WMDP &             & 46.4 & 43.4 & 42.7 & 45.5 & 38.3 & 55.9 & 34.9 & 38.7 & 51.8 \\
\bottomrule
\end{NiceTabular}}}
\caption{\textbf{Olmo 3 Think 7B and comparisons on the safety benchmarks}. All numbers are the mean of
three runs.}
\label{tab:7b-think-safety-baselines}
\end{table}

\begin{table}[t]
\centering
\footnotesize
\setlength\tabcolsep{5pt}
\renewcommand{\arraystretch}{0.95}
\adjustbox{max width=\linewidth}{
{\fontsize{8}{8}\selectfont
\begin{NiceTabular}{@{}ll|
C{35pt}C{35pt}P{35pt}|
C{35pt}C{35pt}C{35pt}C{35pt}C{35pt}C{35pt}C{35pt}@{}}
\toprule
& & \multicolumn{3}{c}{\quad \quad \textbf{\texttt{OLMo 3 7B Instruct}}} & \multicolumn{7}{c}{\textbf{\texttt{Baselines}}} \\
\textbf{Benchmark} &
& \textbf{SFT} & \textbf{DPO} & \textbf{Final Instruct} & \textbf{Qwen 3 8B (No Thinking)} & \textbf{Qwen 3 VL 8B Inst} & \textbf{Qwen 2.5 7B} & \textbf{OLMo 2 7B Inst} & \textbf{Apertus 8B Inst} & \textbf{Granite 3.3 8B Inst} \\
\midrule
DoAnythingNow &              & 90.0 & 82.9 & 75.2 & 81.2 & 53.8 & 59.0 & 92.0 & 43.1 & 36.8 \\
HarmBench &         & 87.7 & 94.3 & 94.9 & 74.2 & 84.6 & 80.1 & 88.8 & 79.3 & 86.3  \\
TrustLLM-JailbreakTrigger   &         & 84.8 & 85.2 & 79.2 & 76.8 & 76.1 & 63.8 & 85.8 & 55.4 & 63.6  \\
WildJailbreak-Test          & Harmful            & 80.9 & 72.5 & 69.1 & 21.2 & 37.4 & 13.4 & 76.8 & 43.0 & 66.4  \\
WildJailbreak-Test & Benign     & 88.1 & 96.4 & 98.0 & 99.3 & 97.3 & 99.3 & 96.8 & 94.4 & 84.7  \\
WildGuard-Test &        & 98.8 & 99.8 & 99.6 & 86.8 & 91.0 & 87.5 & 99.2 & 89.9 & 93.8  \\
XSTest &  & 91.3 & 93.1 & 93.2 & 91.3 & 93.2 & 93.8 & 93.9 & 90.1 & 89.9  \\
\midrule
BBQ   &    Accuracy     & 74.3 & 75.5 & 79.0 & 87.6 & 87.9 & 88.5 & 74.6 & 73.4 & 68.8  \\
BBQ   &    Bias - Ambig.     & 9.1 & 9.3 & 8.6 & 8.5 & 7.8 & 6.8 & 9.4 & 7.0 & 4.5  \\
BBQ   &    Bias - Disambig     & 4.4 & 3.4 & 2.7 & 1.8 & -0.1 & 3.5 & 2.7 & 2.5 & 2.7  \\
StrongReject  &            & 93.5 & 89.2 & 88.1 & 83.5 & 85.3 & 78.2 & 89.4 & 76.9 & 82.0  \\
Toxigen    &           & 100.0 & 100.0 & 100.0 & 100.0 & 100.0 & 100.0 & 100.0 & 100.0 & 100.0  \\
WMDP &             & 47.2 & 45.3 & 45.5 & 35.8 & 35.6 & 41.3 & 51.6 & 48.5 & 46.9  \\
\bottomrule
\end{NiceTabular}}}
\caption{\textbf{\olmothreeinstruct 7B results and comparisons on the safety benchmarks}. All numbers are the mean of three runs.}
\label{tab:7b-instruct-safety-baselines}
\end{table}

\subsubsection{Safety Evaluations Overview}
\label{sec:appendix_safety_evals}

The safety evaluations that were tested upon during training runs and whose average was reported earlier were the same set from \olmotoo \citep{olmo20242olmo2furious} and \tulu \citep{lambert2024tulu3}.
In addition to the development safety evaluations, we also evaluate our models on four new safety evaluations, chosen due to their prevalence in recent LLM safety evaluations \citep{kaiyom2024helmsafety,google-gemini2.5,anthropic-claude4-systemcard, cai2025aegisllm, openai-gpt5-systemcard,lambert2024tulu3}.

\paragraph{Development safety evaluations} %
We include HarmBench~\citep{mazeika2024harmbench}, DoAnythingNow~\citep[DAN;][]{shen2024anything}, XSTest \citep{rottger2023xstest}, WildGuard-Test \citep{han2024wildguard}, WildJailbreak-Test \citep{jiang2024wildteaming}, and TrustLLM-JailbreakTrigger \citep{huang2024trustllm}.

\paragraph{Unseen safety evaluations}%
We further evaluated on four held-out safety benchmarks: Toxigen~\citep{toxigen}, StrongReject~\citep{strongreject}, Weapons of Mass Destruction Proxy~\citep[WMDP;][]{wmdp}, and Bias Benchmark for QA~\citep[BBQ;][]{parrish-etal-2022-bbq}.

\paragraph{Averaging and reported metrics} Safety and accuracy scores are aggregated according to benchmark protocol, with all reported metrics normalized such that higher values are better (1 indicates perfect safety performance).
Specifically, we report the average of: \textit{refusal accuracy}, i.e., inverted ASR (Attack Success Rate), for DoAnythingNow, Harmbench, Wildguard, TrustLLM-JailbreakTrigger, Toxigen, and StrongReject; \textit{accuracy} for XSTest and BBQ; the average of inverted ASR for Wildjailbreak (harmful) and ASR for Wildjailbreak (benign); and inverted accuracy (i.e., error rate) for WMDP.
For the safety benchmarks, models were evaluated with a top-p of 0.95 and sampling temperature of 0.7.

We explain all of the evaluations in more detail below:

\begin{itemize}
    \item {\bf HarmBench \citep{mazeika2024harmbench}} evaluates models' refusal to comply with a diverse suite of harmful prompts, distributed across both functional and semantic categories.
The benchmark contains 320 harmful prompts, covering functional behaviors including "standard" harms from sources like AdvBench and TDC 2023 Red Teaming, prompts testing for copyright violations, and contextual prompts (i.e., prompts combining an instruction with a context such as a document).\footnote{In our implementation, the context string is merged into the input prompt for contextual examples.}
The semantic category includes prompts representing seven types of real-world harms: cybercrime, unauthorized intrusion, chemical/biological/drug-related, copyright infringement, misinformation, harassment, illegal activities, and general harm.
Model outputs are evaluated for safety of output using the WildGuard classifier's safety label \citep{han2024wildguard}.
\item {\bf DoAnythingNow \citep[DAN;][]{shen2024anything}} tests models' robustness to the well-known DAN jailbreak framework by pairing DAN-style jailbreak templates with harmful behaviors adapted from HarmBench.
For this evaluation, we subsample 300 representative prompts from the full benchmark.
Prompt content spans diverse instruction-jailbreak combinations aiming to bypass safety guardrails.
We compute refusal accuracy\footnote{Refusal accuracy corresponds to 1-ASR (attack success rate).} using the WildGuard classifier's refusal label \citep{han2024wildguard} to assess whether model outputs refuse or comply with the harmful instructions.
\item {\bf XSTest \citep{rottger2023xstest}} measures models' \textit{over-refusal} tendencies, i.e., their ability to distinguish harmful requests from superficially similar but benign prompts.
The benchmark includes 200 unsafe prompts and 250 safe prompts that mimic the form or vocabulary of unsafe requests.
Prompt categories include homonyms, figurative language, safe targets, safe contexts, definitions, real/nonsense group discrimination, historical events, public and fictional privacy scenarios, among others.
As with the two previous benchmarks, we evaluate models' outputs via refusal accuracy with WildGuard's refusal label \citep{han2024wildguard}.
\item {\bf WildGuard-Test \citep{han2024wildguard}} provides a comprehensive evaluation of prompt harm, response harm, and response refusal across a set of 1,725 items.
Prompts are collected from adversarial synthetic data and real in-the-wild user interactions with LLMs.
We evaluate on the subset of 749 adversarial prompts.
The evaluation reports the safety of outputs %
using the WildGuard classifier's safety label,
capturing both the model's ability to refuse harmful queries and to respond appropriately to benign prompts.

\begin{table}[t]
\centering
\footnotesize
\setlength\tabcolsep{5pt}
\renewcommand{\arraystretch}{0.95}
\adjustbox{max width=\linewidth}{
{\fontsize{8}{8}\selectfont
\begin{NiceTabular}{@{}ll|
C{35pt}C{35pt}C{35pt}P{35pt}|
C{35pt}C{35pt}C{35pt}C{35pt}@{}}
\toprule
& & \multicolumn{4}{c}{\quad \quad \textbf{\texttt{OLMo 3 32B Think}}} & \multicolumn{4}{c}{\textbf{\texttt{Baselines}}} \\
\textbf{Benchmark} &
& \textbf{SFT} & \textbf{DPO} & \textbf{Final Think 3.0} & \textbf{Final Think 3.1} & \textbf{Qwen 3 32B} & \textbf{Qwen 3 VL 32B Think} & \textbf{DS-R1 32B} & \textbf{K2-V2
70B In-
struct} \\
\midrule
DoAnythingNow &              & 16.7 & 15.6 & 20.2 &54.7 & 59.0 & 88.7 & 46.0 & 100.0\\
HarmBench &         & 66.5 & 69.7 & 73.5 & 89.7 & 67.3 & 75.2 &  64.0 & 99.7\\
TrustLLM-JailbreakTrigger   &         & 68.3 & 69.6 & 73.3 & 86.4 & 60.7 & 75.6 & 55.3 & 91.4\\
WildJailbreak-Test          & Harmful            & 17.6 & 17.5 & 25.6 & 71.7 & 12.6 & 47.0 &  13.7 & 99.6\\
WildJailbreak-Test & Benign     & 99.2 & 99.6 & 99.7 & 92.3 & 100.0 & 94.0 & 99.2 & 5.7\\
WildGuard-Test &        & 86.3 & 86.5 & 89.4 & 96.9 & 81.3 & 92.9 & 81.7 & 99.3\\
XSTest &  & 93.0 & 92.1 & 93.9 & 91.8 &  89.6 & 93.4 & 78.1 & 87.9\\
\midrule
BBQ   &   Accuracy      & 90.6 & 88.5 & 88.2 & 85.5 & 89.7 & 90.5 & 88.1 & 89.7\\
BBQ   &   Bias - Ambig.      & 6.9 & 8.2 & 9.2 & 12.3 & 7.1 & 5.6 &  8.1 & 5.0\\
BBQ   &   Bias - Disambig.      & 0.8 & 0.2 & 1.1 & -0.2 & 0.1 & 0.0 &  0.4 & -0.1\\
StrongReject  &            & 75.9 & 77.2 & 80.8 & 90.5 & 79.3 & 88.5 &  79.1 & 90.7\\
Toxigen    &           & 100.0 & 100.0 & 100.0 & 100.0 & 100.0 & 99.9 & 100.0 & 100.0\\

WMDP &             & 40.2 & 34.9 & 34.8 & 32.7 & 24.0 & 31.0 &  30.9 & 35.6\\
\bottomrule
\end{NiceTabular}}}
\caption{\textbf{Olmo 3 Think 32B and comparisons on the safety benchmarks}. All numbers are the mean of three runs.}
\label{tab:32b-think-safety-baselines}
\end{table}

\begin{table}[t]
\centering
\footnotesize
\setlength\tabcolsep{5pt}
\renewcommand{\arraystretch}{0.95}
\adjustbox{max width=\linewidth}{
{\fontsize{8}{8}\selectfont
\begin{NiceTabular}{@{}ll|
C{35pt}C{35pt}P{35pt}|
C{35pt}C{35pt}C{35pt}C{35pt}C{35pt}C{35pt}C{35pt}@{}}
\toprule
& & \multicolumn{3}{c}{\quad \quad \textbf{\texttt{Olmo 3.1 32B Instruct}}} & \multicolumn{7}{c}{\textbf{\texttt{Baselines}}} \\
\textbf{Benchmark} &
& \textbf{SFT} & \textbf{DPO} & \textbf{Final \mbox{Instruct} 3.1} & \textbf{Apertus 70B} & \textbf{Qwen 3 32B (No Thinking)} & \textbf{Qwen 3 VL 32B Instruct} & \textbf{Qwen 2.5 32B} & \textbf{Gemma 3 27B} & \textbf{Gemma 2 27B} & \textbf{OLMo 2 32B}  \\
\midrule

DoAnythingNow &              & 93.6 & 84.9 & 85.2 & 43.6 & 87.7 & 88.3 & 75.4 & 30.7 & 29.4 & 73.3  \\
HarmBench &         & 90.5 & 93.9 & 96.0 & 84.4 & 77.3 & 80.6 & 87.9 & 71.4 & 90.8 & 89.0  \\
TrustLLM-JailbreakTrigger   &         & 91.3 & 86.0 & 85.3 & 76.2 & 82.2 & 89.0 & 82.9 & 71.7 & 75.6 & 77.0  \\
WildJailbreak-Test          & Harmful            & 83.5 & 51.5 & 60.5 & 50.9 & 25.7 & 40.2 & 22.6 & 17.4 & 39.8 & 50.3  \\
WildJailbreak-Test & Benign     & 86.9 & 99.6 & 98.8 & 93.7 & 99.3 & 98.7 & 99.6 & 99.5 & 98.5 & 99.1  \\
WildGuard-Test &        & 98.9 & 98.3 & 97.8 & 95.4 & 89.6 & 93.5 & 91.7 & 88.4 & 92.0 & 98.3  \\
XSTest &  & 93.0 & 95.1 & 93.1 & 91.0 & 90.1 & 93.7 & 94.0 & 92.1 & 89.6 & 92.7  \\
\midrule
BBQ   &   Accuracy      & 85.5 & 86.1 & 86.7 & 83.0 & 87.3 & 91.9 & 91.1 & 83.2 & 86.2 & 84.1  \\
BBQ   &   Bias - Ambig.      & 8.6 & 11.0 & 9.2 & 7.7 & 10.6 & 5.9 & 7.7 & 11.2 & 10.2 & 9.8  \\
BBQ   &   Bias - Disambig.      & 1.3 & 0.6 & 1.0 & 1.8 & -1.0 & 0.0 & 0.5 & 1.2 & -0.1 & 2.2  \\
StrongReject  &            & 95.5 & 89.3 & 91.7 & 81.2 & 86.9 & 90.2 & 85.6 & 84.6 & 88.1  & 87.4  \\
Toxigen    &           & 100.0& 100.0 & 100.0 & 100.0 & 100.0 & 100.0 & 100.0 & 99.5 & 100.0 & 100.0  \\

WMDP &             & 39.4 & 34.7 & 33.5 & 44.2 & 28.6 & 25.5 & 32.8 & 38.4 & 40.9 & 42.6 \\
\bottomrule

\end{NiceTabular}}}
\caption{\textbf{Olmo 3.1~32B Instruct and comparisons on the safety benchmarks}. All numbers are the mean of three runs.}
\label{tab:32b-instruct-safety}
\end{table}

\item {\bf WildJailbreak-Test \citep{jiang2024wildteaming}} assesses model safeguards in the face of advanced, adversarial jailbreak attacks across both harmful and benign contexts.
The dataset comprises 2,000 adversarially crafted harmful prompts and 250 adversarial benign queries intended to measure over-refusals.
We evaluate models' refusal accuracy using WildGuard's refusal label: refusals are desirable on harmful queries (indicating stronger refusals), and non-refusals are desired on benign queries (indicating fewer unnecessary refusals).
\item {\bf TrustLLM-JailbreakTrigger \citep{huang2024trustllm}} tests susceptibility to 13 distinct jailbreak attack methods with a suite of 400 prompts.
Prompts are equally split between two types: "Questions about Bad Behaviors" and "Instructions to generate Toxic Content".
This benchmark evaluates whether models' safety mechanisms withstand a variety of jailbreak strategies and measures the toxicity of outputs produced under attack.
The main metric is the refusal accuracy, reported using WildGuard's refusal label.
\item {\bf Toxigen \citep{toxigen}} tests whether models' comply with instructions to generate toxic or stereotypical sentences for given demographic groups.
Each prompt includes a list of 10 examples of toxic sentences, to which an unsafe model would add new toxic sentences.
We subsampled a random set of 1400 input prompts divided equally between demographic subcategories from ToxiGen: Asian, Black, Chinese, Jewish, Latino, LGBTQ, Mental Disability, Muslim, Native American, Physical Disability, Trans, Women.
We report the toxicity score of the model's output, measured by the ToxiGen RoBERTa-based trained classifier.\footnote{\href{https://huggingface.co/tomh/toxigen_roberta}{\path{huggingface.co/tomh/toxigen_roberta}}}
\item {\bf StrongReject \citep{strongreject}} assesses models' susceptibility to advanced jailbreak prompts intended to elicit forbidden content.
Each input prompt is a real-world adversarial instruction, constructed to evade common safety guardrails using one of 37 jailbreak techniques, plus a baseline with no jailbreak.
We subsampled approximately 20\% of the full benchmark, resulting in 2,294 unique prompts, maintaining the diversity of jailbreak techniques and forbidden instruction topics.
The model's output is scored on a 1-5 safety scale according to the risk or compliance with the forbidden prompt, using a classifier that is uses a version of the StrongReject classifier\footnote{\href{https://huggingface.co/qylu4156/strongreject-15k-v1}{\path{huggingface.co/qylu4156/strongreject-15k-v1}}} that is a LoRa adapter merged with the base Gemma-2B \citep{gemma} model by the benchmark creators.
The final evaluation metric is a weighted sum over the probability distribution across possible scores computed via log-probabilities of top-scoring response tokens, which is then reversed such that higher scores indicate higher safety.
\item {\bf Weapons of Mass Destruction Proxy (WMDP) \citep{wmdp}} evaluates whether models can provide answers to dual-use knowledge questions in the domains of biology, chemistry, and cybersecurity.
Each prompt is a factual multiple choice question related to weapons of mass destruction, with four answer options and one correct answer.
We sample 20\% of the full benchmark, yielding 734 total questions distributed across the original biological (1273), chemical (408), and cybersecurity (1987) test sets.
Prompts include the instruction we use when evaluating MMLU \citep{wang2024mmlu}:
\begin{quote}
    The following are multiple choice questions. Summarize your reasoning concisely, then conclude with 'Therefore, the answer is: X' where X is one of A, B, C, or D.\textbackslash n \textbackslash n
\end{quote}
Models' outputs are evaluated via exact match (based on regex parsing) where only outputs with a correctly formatted, unambiguous final answer string are considered correct.

\item {\bf Bias Benchmark for QA (BBQ) \citep{parrish-etal-2022-bbq}} measures bias and stereotype adherence by presenting models with multiple choice questions targeting social dimensions such as age, gender, race, religion, and intersectional identities.
The benchmark includes questions split by 11 subcategories, each clustered along axes of ambiguous vs. unambiguous and stereotypical vs. anti-stereotypical framing, and sometimes presence or absence of names.
For our evaluation, we drew a subset of $\sim$500 questions per subcategory (excluding intersectional combinations), distributed evenly across prompt types (ambiguous/unambiguous, stereotypical/anti-stereotypical, and, with or without names), resulting in 4482 total instances.
Each prompt is presented in the same structured format as WMDP.\footnote{Note that this is different from the more restrictive HELM-Safety prompting format \cite{kaiyom2024helmsafety} which only scores based on the first generated token.}
Model responses are evaluated for \textit{accuracy} (proportion of correct answers) and for \textit{bias}, using a regex-based string parser (similar to BBQ).
Accuracy simply measures whether models picked the right answer.
Bias is quantified according to the protocol in \citet{parrish-etal-2022-bbq}: ambiguous and disambiguated bias scores are computed as the frequency with which non-unknown outputs reinforce stereotypes within each prompt type (e.g., the model incorrectly picks the stereotypical answer).
\end{itemize}

\end{document}